%% file: main.tex
\documentclass[11pt]{article}
\usepackage[utf8]{inputenc}
\usepackage[T1]{fontenc}
\usepackage{helvet}

\usepackage[a4paper,margin=2.3cm]{geometry}
\usepackage{graphicx}
\graphicspath{{./}{figures/}}
\usepackage{booktabs}
\usepackage{tabularx}
\usepackage{array}
\usepackage{amsmath,amssymb}
\usepackage{xcolor}
\usepackage{caption}
\usepackage[numbers,sort&compress,super]{natbib}
\usepackage{jabbrv}
\usepackage[hidelinks]{hyperref}
\hypersetup{pdftitle={ICare-DT: a digital twin for continuous forecasting of intensive care patients},pdfauthor={Yanke Li, Yahang Qi, Yumi Kim, Diego Paez-Granados}}
\usepackage{lineno}
\usepackage{setspace}
\newcommand{\spo}{SpO\textsubscript{2}}
\newcommand{\sao}{SaO\textsubscript{2}}
\title{\textbf{ICare-DT: a digital twin for continuous forecasting of intensive care patients}}
\author{Yanke Li\textsuperscript{1,2,*},
Yahang Qi\textsuperscript{1,3},
Yumi Kim\textsuperscript{1} and
Diego Paez-Granados\textsuperscript{1,2,*}\\[4pt]
\small\textsuperscript{1}SCAI Lab, Department of Health Science and Technology, ETH Zurich, Switzerland\\
\small\textsuperscript{2}Swiss Paraplegic Research, Nottwil, Switzerland\\
\small\textsuperscript{3}Department of Statistical Sciences, University of Toronto, Canada\\
\small\textsuperscript{*}Corresponding authors: yanke.li@hest.ethz.ch, diego.paez@hest.ethz.ch}
\date{}

\begin{document}
\maketitle

\begin{abstract}
Digital twins of intensive care patients should update forecasts from new data and predict treatment responses, but few studies test whether routine records support either.
We propose ICare-DT, a Transformer-based twin that updates a latent patient state hourly from new measurements without retraining.
We trained it on 140,473 eICU-CRD stays to forecast vital signs crossing their training-set means.
On test data, later forecasts beat the twin's own hour-6 forecasts of the same windows in all 12 settings ($+0.073$ to $+0.153$ AUROC).
On validation data, controls that withheld or swapped the new hours traced the gain to the patient's own measurements.
Without refitting, the model kept its gain on a second database (8 of 9 settings, $+0.025$ to $+0.114$ AUROC) in a prespecified evaluation of the previously analyzed MIMIC-IV test split.
Refitted within MIMIC-IV, a twin state fine-tuned on clinical endpoints gave gradient boosting a significant gain for acute kidney injury ($+0.0035$ AUROC; hypotension $+0.0010$, not significant).
To test response prediction, we built a benchmark that replaces the treatments in real records with simulated doses and injects responses of known size, visible only at measurement times.
Ridge models supervised with injected effects recovered them from real records only through frequently measured variables (heart rate) read at their measurement times.
We asked standard learners to predict these effects for a dosing regime no patient had received.
No standard learner outperformed predicting zero effect at an out-of-support query.
When told how the simulated responses were generated, a separate learner recovered the effects from the same records without seeing their true values ($R^2$ 0.30 to 0.40 at one effect size).
Routine records support forecasts that improve with each patient's new data, and our benchmark is a testbed for models that learn how responses are generated.

\end{abstract}

\section*{Introduction}

Patients in intensive care can deteriorate within hours, and clinicians adjust vasopressors, fluids and ventilation as they watch each patient respond. Two questions recur at the bedside: where is this patient heading, and how would this patient respond if the treatment were changed? A digital twin of the patient has been proposed to help answer both\cite{katsoulakis2024digital,laubenbacher2022building,kamel2021digital,sel2025vvuq}. In the National Academies' definition, a digital twin is a model of one individual that is updated dynamically from that individual's data, is personalized and informs decisions\cite{nasem2024digital}.

\begin{figure}[tbp]
\centering
\includegraphics[width=0.94\textwidth]{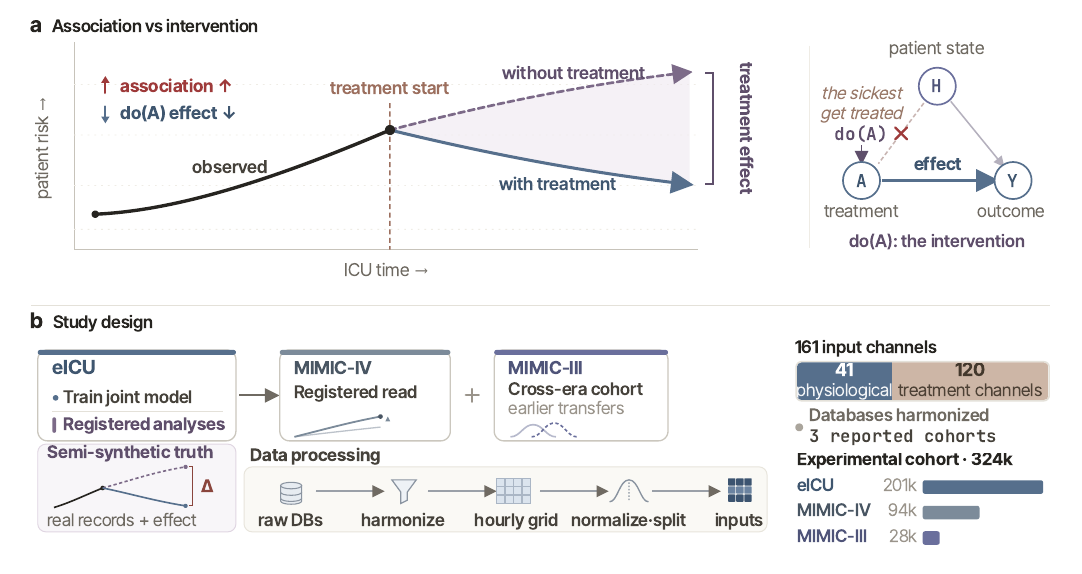}\\[2pt]
\includegraphics[width=0.94\textwidth]{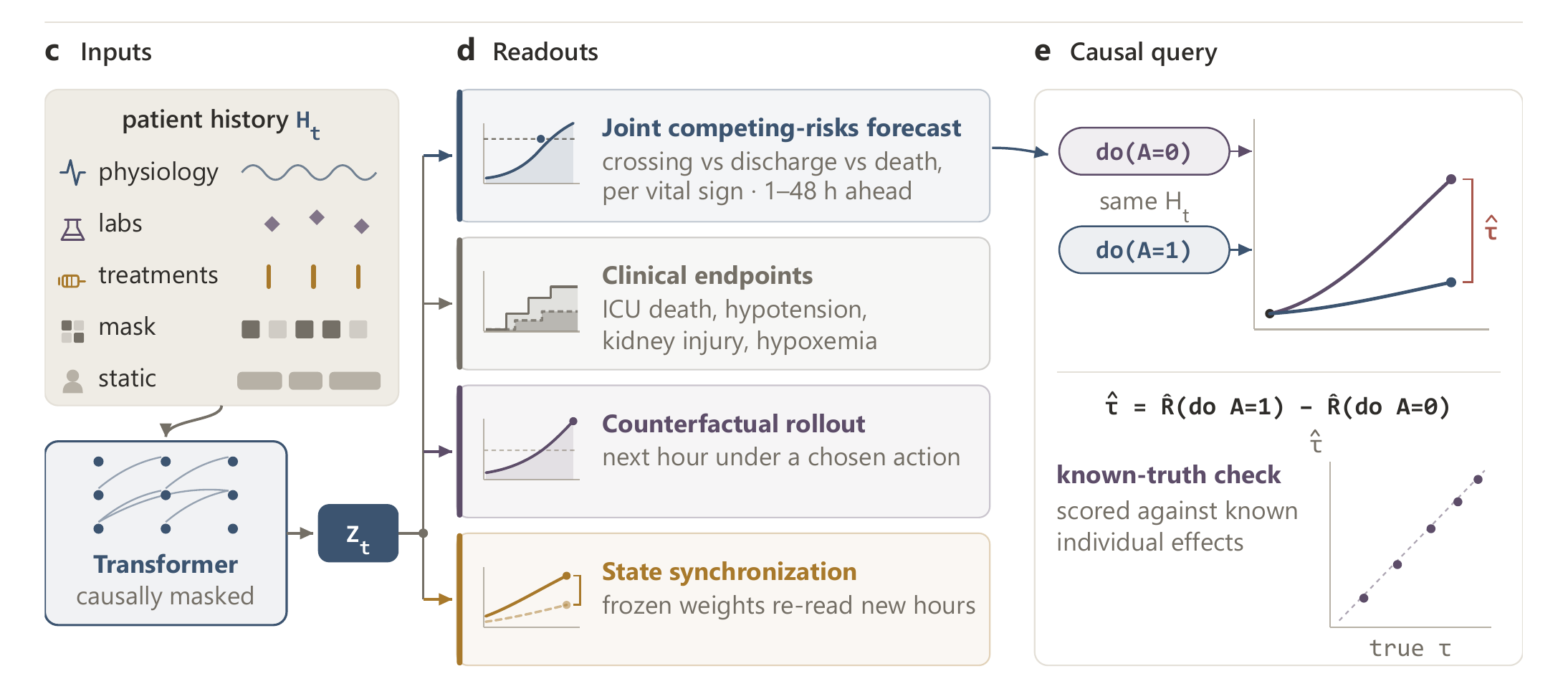}
\caption{\textbf{ICare-DT and its evaluation.} \textbf{a}, Association and intervention. Because the sickest patients are treated, a treatment can be associated with higher risk even when it lowers risk. The operator do($A$) sets the treatment $A$ by intervention. \textbf{b}, Data and design. Three harmonized databases are placed on an hourly grid of 161 input channels (41 physiological, 120 treatment; Methods), and bars give each database's stays in the experimental cohort (Table~\ref{tab:cohorts}). ICare-DT is trained on eICU-CRD (eICU in the panel), and every analysis uses eICU-CRD except one registered read of the previously analyzed MIMIC-IV test split. That read scored the two prediction results (synchronization and the clinical-endpoint probes), with disclosed secondary outcomes (Supplementary Note~7). MIMIC-III appears only in earlier transfer analyses that were not registered (Supplementary Note~8). A semi-synthetic generator writes injected responses onto real eICU-CRD records. Its generated dataset passed a locked realism audit (Fig.~\ref{fig:benchmark}b). \textbf{c}, Inputs. A causally masked Transformer encodes the patient history $H_t$ (physiology, laboratory values, treatments, measurement mask and static covariates) into the patient state $z_t$. \textbf{d}, Readouts. The joint competing-risks forecast gives, for each of four vital signs, the probabilities that a crossing of the training-set mean, live discharge from the unit or death comes first, 1 to 48 hours ahead. Clinical-endpoint probes read the state for four clinical endpoints (in the panel, ICU death is death in the unit and kidney injury is acute kidney injury; Methods). The counterfactual rollout generates the next hour under a chosen action and was evaluated only on a historical benchmark (Fig.~\ref{fig:benchmark}a). State synchronization re-reads newly recorded hours with frozen weights. \textbf{e}, Causal query, the intended use of a response module. Predicted courses under do($A=1$) and do($A=0$) are compared for the same history. In this paper, the twin's rollout is not used for individual effects: on washout panels of real records, effects estimated by readers and treatment-effect learners are scored against the known individual effects. The panels share the generator's code, and their realism audit is weak evidence (Figs.~\ref{fig:generation}c, \ref{fig:benchmark}b and \ref{fig:map}--\ref{fig:rh}; Results). The scatter in \textbf{e} is schematic.}
\label{fig:overview}
\end{figure}

For a machine-learning model, the patient is represented by the patient record: an irregularly sampled multivariate time series of vital signs, laboratory values and treatments that grows by one hour at a time. A twin therefore needs two capabilities. The first is continuous forecasting: the twin keeps a patient state, a learned summary of the record so far, updates it as each new hour arrives without retraining, and forecasts upcoming events from it. The second is treatment-response prediction: forecasting the patient's course under alternative treatments, whose contrast is, in causal terms, a conditional treatment effect that changes over time\cite{prosperi2020causal,feuerriegel2024causalml,schulam2017reliable}. Both must be learned from routine records, and two features of these records make this hard. First, what is measured and when depends on clinical practice and on the patient's condition\cite{agniel2018biases}; in 400 stays of the eICU Collaborative Research Database (eICU-CRD)\cite{pollard2018eicu} studied here, the recorded heart rate changes in 57\% of hours and creatinine, a laboratory marker of kidney function, in only 4\%. Second, a record shows only the outcome under the treatment that was given, and because the sickest patients are the ones treated, a beneficial treatment can look harmful, a bias known as confounding by indication (Fig.~\ref{fig:overview}a)\cite{gottesman2019guidelines,uffen2021retrospective}.

Machine learning has advanced each capability separately. For prediction, foundation and sequence models of health records learn one representation from many patients and forecast many outcomes, and they are judged by their accuracy across tasks and hospitals\cite{burger2025foundation,steinberg2023motor,moor2023foundation,wornow2023ehrshot,rajkomar2018scalable}. Relational models capture dependencies among irregularly sampled intensive care variables\cite{wang2026garlic}, and landmark models update a prediction at set times as data accrue\cite{vanhouwelingen2007dynamic}. For treatment effects, counterfactual sequence models estimate how outcomes would evolve under alternative treatment sequences\cite{lim2018rmsn,bica2020crn,melnychuk2022causal,li2021gnet,xiong2024gtransformer,seedat2022continuous,hess2024bncde}; g-computation models among them, such as G-Net\cite{li2021gnet}, simulate every recorded variable forward under a chosen treatment sequence. Because true effects are never observed in real data, these and other causal methods are scored on simulators in which the effects are known\cite{hill2011bayesian,hahn2019acic,geng2017prediction,oberst2019counterfactual,li2025karmats}. Intensive care twins have also been built from expert knowledge: a twin of early sepsis with expert-specified causal structure was checked prospectively against observed responses to single interventions, although not against known individual effects\cite{lal2020sepsis}.

These evaluations leave open two questions that must be answered before a twin can be tested at the bedside. The first is whether a twin's forecasts improve because it reads the patient's new data. Accuracy at fixed time points cannot show this, and neither can a gain in accuracy later in the stay, because a later forecast is closer to the events it predicts and can be more accurate without using anything new about the patient. A convincing test compares forecasts of the same future window made at the same lead time with and without the new hours, includes trivial references that need no learning, such as the patient's last measured value, and is fixed before the results are seen\cite{nosek2018preregistration,hofman2023preregistration}. Verification of any kind is rarely reported: in a scoping review of 149 studies of human digital twins, 2 mentioned verification, validation and uncertainty quantification\cite{tudor2025digitaltwins}. The second question is whether routine records hold enough information to learn individual treatment responses at all. A simulator can answer it only if it reproduces which values are recorded and when, because an effect enters the record only in hours in which the affected variable is measured\cite{agniel2018biases,pullenayegum2016longitudinal,vanderschueren2023informative}. Yet the intensive care benchmark used for the Causal Transformer writes a simulated outcome at every hour\cite{melnychuk2022causal}, generators that let the user set the effects have no time dimension\cite{parikh2022credence,demanela2024frugal}, and the realism of simulators is seldom audited\cite{curth2021benchmarking,neal2021realcause}.

Here we propose ICare-DT, a digital twin of intensive care patients built for continuous forecasting, and we test whether routine records support both capabilities (Fig.~\ref{fig:overview}). ICare-DT is trained on eICU-CRD (Table~\ref{tab:cohorts}). A causally masked Transformer maps the record up to each hour to a latent patient state. Every hour, the twin runs the same frozen network on the record so far and so assimilates the patient's new measurements without retraining, which we call synchronization. From the state, one head per vital sign forecasts which of three events comes first within 48 hours: a crossing of the vital sign's training-set mean, live discharge from the unit or death. Because a crossing marks an excursion from the cohort average rather than a clinical event, we also probed the state on four clinical endpoints: death in the unit, hypotension (low blood pressure), acute kidney injury (a rise in creatinine) and hypoxemia (low blood oxygen saturation).

To test forecasting, we compared each synchronized forecast with the twin's own hour-6 forecast of the same window, fed the model control inputs that keep the lead time but withhold the patient's new hours or replace them with another patient's, and used the patient's last measured value and gradient boosting as references. To test treatment-response prediction, we built a semi-synthetic generator that writes responses of known size onto real eICU-CRD records while reproducing how these records are measured, and audited its realism against a threshold fixed in advance. With the same code we built washout panels: real records with simulated dosing, a drug-free washout and then sustained treatment or none, so that each patient's effect is known. On these panels we tested standard treatment-effect learners and our candidate response module, ICare-RH; the twin's own counterfactual rollout, which generates the next hour under a chosen action, was not tested there. Registered analyses followed protocols committed to version control before they ran; these are internal registrations, not external preregistrations (Methods). They include a registered read of the Medical Information Mart for Intensive Care (MIMIC-IV)\cite{johnson2023mimic}, a single scoring of its previously analyzed test split.

The first capability is supported on eICU-CRD. On its test set, synchronized forecasts beat the twin's own hour-6 forecasts of the same window in all 12 settings (four vital signs, three forecast windows), by 0.073 to 0.153 in the area under the receiver operating characteristic curve (AUROC; Fig.~\ref{fig:sync}c). On the validation split, control inputs traced the gain to the patient's own measurements, though not to information beyond the target's last measured value (Fig.~\ref{fig:sync}b,d); on the same split, however, a gradient-boosted model reading the same hours was the stronger forecaster, so we make no claim of state-of-the-art accuracy.
Judged as a forecast, the synchronized one nevertheless beat the patient's last measured value in 11 of the 12 settings on eICU-CRD, in a disclosed comparison that was not registered. In the registered read of the previously analyzed test split of MIMIC-IV, the model, applied without refitting, kept the gain in 8 of the 9 settings that could be scored there ($+0.025$ to $+0.114$ AUROC), meeting the registered transfer criterion; against the patient's last measured value it was better in 1 of 9 settings and worse in 2, not meeting the registered secondary criterion (Table~\ref{tab:gates}). With all models refitted within MIMIC-IV, adding a twin state fine-tuned on the clinical endpoints to gradient boosting gave a significant gain for acute kidney injury only ($+0.0035$ AUROC).

For the second capability, the generator passed its realism audit, a classifier test that scores 0.5 when generated and real records cannot be told apart, once it reproduced how values are recorded (audit AUC 0.572 against a pass threshold of 0.75; Figs.~\ref{fig:generation} and \ref{fig:benchmark}). On the washout panels, which reuse the generator's code with separate parameters, ridge regressions trained with the true effects as labels (readers, which cannot be deployed) recovered the injected responses at real measurement cadence on heart rate but not on creatinine, the two responding variables tested, and only when reading the hours in which the value changed (Fig.~\ref{fig:map}). By design, the benchmark asks whether a model can carry what each patient's earlier responses reveal to a regime that the data never show. None of 16 standard treatment-effect learners, including the Causal Transformer and the g-computation model G-Net (Causal Transformer code), outperformed predicting zero effect at the decision query, a treatment regime that no patient had received. Neither registered hour inside the dosing period (30 or 42) met the registered overlap criteria for a query inside the treatment support. Unlike these standard learners, a learner given the generator's response structure (the form of the injected response, the distribution of patient traits and how outcomes are recorded) recovered the effects from the same factual data (Fig.~\ref{fig:estimators}). ICare-RH recovered effects in simulation but failed label-free placebo checks, run before any true effect was read, and was therefore never scored on real records (Fig.~\ref{fig:rh}).

A twin can therefore already keep a patient state whose forecasts improve as the patient's own measurements arrive, whereas predicting an individual patient's response to treatment remains unproven. Our results suggest that a response module should learn the response structure from factual data, and we will release the generator and washout-panel code, with the registered protocols, as a testbed for such modules.

\section*{Results}

We first describe ICare-DT and test continuous forecasting on eICU-CRD and MIMIC-IV. We then build a benchmark with known treatment responses and ask whether individual responses can be recovered from routine records.

\subsection*{ICare-DT keeps an hourly patient state}

ICare-DT reads each stay's record on an hourly grid of 161 channels (41 physiological, 120 for treatments such as drugs, fluids and ventilator settings), a mask of the hours in which each physiological variable was measured and four static covariates (Fig.~\ref{fig:overview}c; Methods). A causally masked Transformer maps the record up to hour $t$ to a 64-dimensional patient state $z_t$. The tracked vital signs are heart rate, respiratory rate and two oxygen-saturation channels, \spo{} and \sao{}; in eICU-CRD, the \sao{} channel combines monitor saturation readings with blood-gas values (Methods). Crossings are upward for the rates and downward for the saturations (Fig.~\ref{fig:overview}d). Five models, differing only in random seed, were trained on the 140,473 stays of the person-disjoint eICU-CRD training split (Table~\ref{tab:cohorts}). 

\subsection*{Hour-48 forecasts discriminate threshold crossings and death on eICU-CRD}

\begin{figure}[tbp]
\centering
\includegraphics[width=\textwidth]{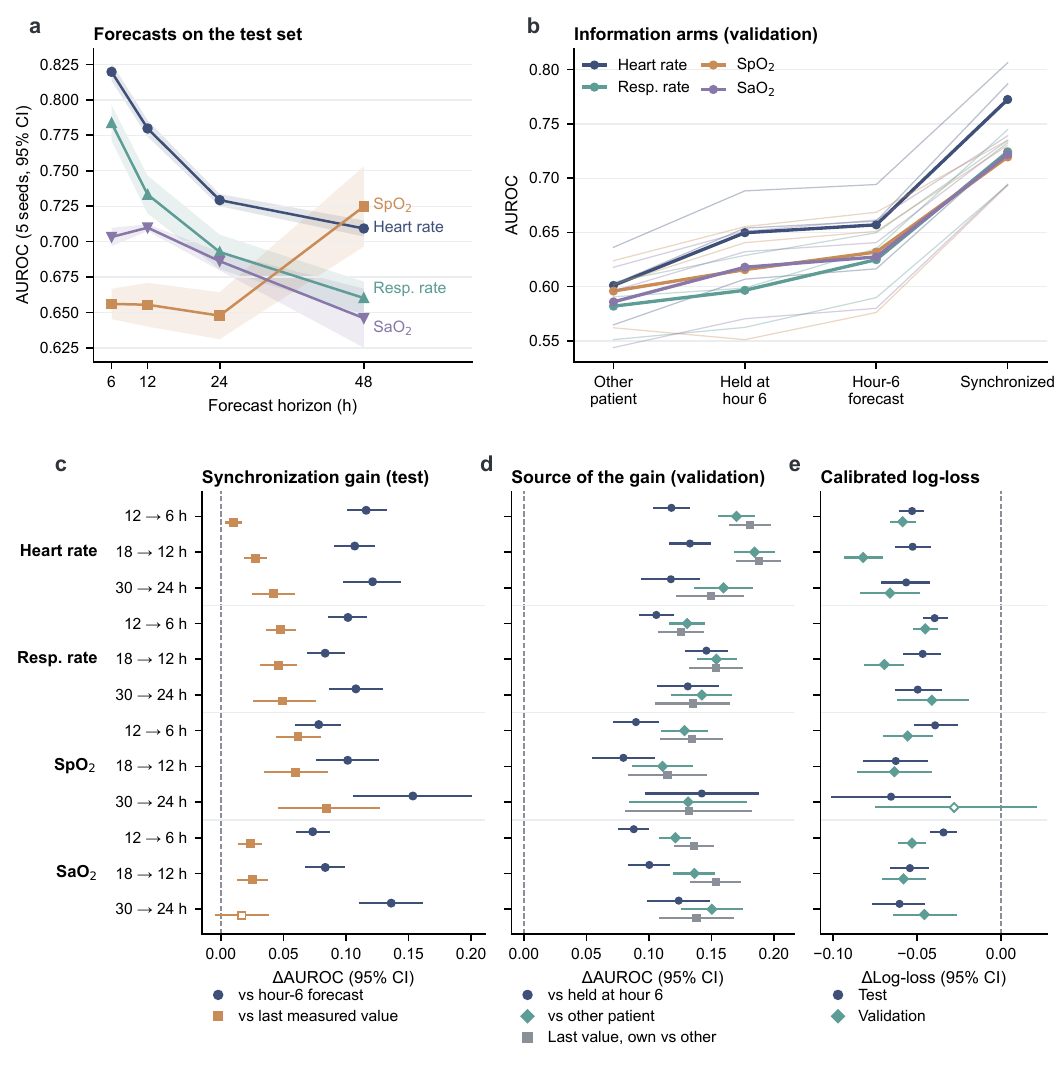}
\caption{\textbf{Synchronized forecasts beat the twin's own hour-6 forecasts.} All panels show eICU-CRD (MIMIC-IV: Table~\ref{tab:gates} and Supplementary Note~7). \textbf{a}, Forecasts on the test set: AUROC of the hour-48 forecasts of a crossing of each vital sign's training-set mean, by horizon (227 to 1,449 stays per setting; mean of five seeds with 95\% $t$ intervals). \textbf{b}, Information arms (validation), the control inputs of the text: AUROC of the same frozen model when hours 7 to $L$ are read as recorded (Synchronized), held at their hour-6 values (Held at hour 6) or taken from the other patient nearest in hour-6 state (Other patient), and of the model's own forecast made at hour 6 (Hour-6 forecast). Thin lines show settings; thick lines, means per vital sign. \textbf{c}, Synchronization gain (test): AUROC gain of the synchronized forecast over the hour-6 forecast (circles; registered) and over the patient's last measured value (squares; disclosed analysis). Rows give landmark $L$ and horizon $H$ as $L\to H$. \textbf{d}, Source of the gain (validation): AUROC gain of the synchronized forecast over Held at hour 6 (circles) and over Other patient (diamonds), and the advantage of the patient's own last measured value over the other patient's, without a model (squares). \textbf{e}, Calibrated log-loss of the synchronized forecast minus that of the hour-6 forecast (Test) or of Other patient (Validation). Negative values favor the synchronized forecast. Probabilities were clipped before taking logs, so only the sign is interpreted. In \textbf{c}--\textbf{e}, filled markers survive Holm correction within their family, and bars are 95\% bootstrap intervals (test: 2,000 stay-level replicates; validation: 10,000 whole-person replicates). In \textbf{b}--\textbf{e}, stays had their target measured every hour through the forecast window (592 to 6,648 stays per setting on the test set, 551 to 7,707 on validation). Both splits had been read in earlier analyses (Supplementary Note~2). AUROC, area under the receiver operating characteristic curve.}
\label{fig:sync}
\end{figure}

We first checked that the twin's forecasts are informative when issued at a fixed time in the stay. In a registered read of the eICU-CRD test set (30,099 stays of 20,886 patients; Supplementary Note~6), forecasts issued at hour 48 discriminated threshold crossings with AUROCs of 0.646 to 0.820 across the four vital signs and horizons of 6 to 48 hours (five-seed means; 227 to 1,449 stays per setting; Fig.~\ref{fig:sync}a). These stays were still in the unit at hour 48, measured and not yet past the threshold, and were followed until the outcome was known. Without recalibration, the mean forecast was 91\% to 105\% of the observed rate at horizons of 6 and 12 hours and 68\% to 89\% at 24 and 48 hours (Methods).

The registered plan specified a descriptive 48-hour death forecast, the largest of the four heads' death probabilities (Methods). Among the 11,498 stays still in the unit after hour 48, it discriminated death in the unit within 48 hours with an AUROC of 0.798 (95\% interval across seeds 0.768 to 0.828).
Its mean forecast was far below the observed rate (0.0023 against 0.026), in part by construction, because each target-specific head counts a death only if it precedes that target's crossing.

\subsection*{Synchronizing the twin with the patient's record improves its forecasts}

A twin should forecast better as it reads more of the patient's record. We tested synchronization in 12 settings, which pair the four vital signs with a landmark $L$ and horizon $H$ of (12, 6), (18, 12) or (30, 24) hours. The synchronized forecast reads the record up to hour $L$ and forecasts a crossing in the next $H$ hours; the hour-6 forecast is the same frozen model's forecast of that window from the first six hours of the stay. With model and window fixed, the gain measures the value of reading more of the record, which adds measurements but also shortens the lead time. A stay entered a setting if its target was measured every hour through the window (592 to 6,648 stays per setting; 42.9\% of test stays entered at least one; Methods).

In the registered analysis of the test set, synchronization improved on the hour-6 forecast in all 12 settings, by 0.073 to 0.153 AUROC, and every setting survived Holm correction over the 12 settings (Fig.~\ref{fig:sync}c and Supplementary Table~S6). Calibrated log-loss and Brier score also improved in all 12 settings (Fig.~\ref{fig:sync}e; Brier score in the source data). In a disclosed comparison (reported in full but not registered), the synchronized forecast also beat the patient's last measured value of the target, a persistence baseline, in 11 of 12 settings after Holm correction (median difference 0.044 AUROC, range 0.010 to 0.085; Fig.~\ref{fig:sync}c).

On the validation split, however, a multivariate gradient-boosted model reading the same hours (a fixed, untuned CatBoost recipe\cite{prokhorenkova2018catboost}) was the stronger forecaster of threshold crossings. The twin's forecasting heads, read from one checkpoint, trailed it by a median of 0.012 AUROC and were worse in 3 of 12 settings after Holm correction over the 120 contrasts of that comparison (Table~\ref{tab:gates}; Methods). Adding the twin's state to the boosted model raised its AUROC point estimate in 10 of 12 settings (median 0.004), but no AUROC gain survived correction.

\subsection*{Control inputs trace the gain to the patient's own measurements}

Because a later forecast is also closer to its window, the gain could come from the shorter lead time rather than from the new data. To separate the two, the same frozen model read hours 7 to $L$ of the validation records in three versions (Fig.~\ref{fig:sync}b): as recorded (synchronized); held at their hour-6 values and marked as unmeasured, which keeps the lead time but adds no information; or taken from the other patient whose hour-6 state was closest and spliced onto this patient's first six hours, which adds hours that are not the patient's own.

The synchronized forecast beat the held input in all 12 settings, by 0.080 to 0.146 AUROC, after Holm correction over the 36 primary contrasts (three per setting; Fig.~\ref{fig:sync}d). Part of this margin is a handicap of the held input: although it reads the same six hours as the hour-6 forecast, it fell below that forecast in all 12 settings, by 0.006 to 0.036 AUROC (significant in 5 after the same correction). The heads were never trained on records held constant, so the gain over this input is not purely a gain from new data (Proposition~1; Methods).

The synchronized forecast also beat the other-patient input in all 12 settings after the same correction, by 0.111 to 0.185 AUROC, which met the registered personalization condition (Table~\ref{tab:gates}). Splicing could itself degrade the forecast, but the synchronized forecast beat the other patient's own forecast, made without a splice, by a similar margin (0.119 to 0.188 AUROC; Methods). The condition is necessary for personalization but not sufficient, because vital signs track their recent values: without any model, the patient's own last measured value beat the other patient's by a similar margin (0.115 to 0.188 AUROC). A difference-in-differences contrast, prespecified as descriptive, asked whether the model's advantage over the other-patient input exceeds this model-free advantage. After Holm correction it did so in no setting and fell significantly short in two (Fig.~\ref{fig:sync}d and Supplementary Table~S6), although the contrast need not be zero even when the new hours inform the forecast only through the last measured value (Methods). Information beyond the target's last measured value is therefore not shown (Proposition~2). As a forecast, however, the synchronized one beat the last measured value in 11 of 12 test settings.

\subsection*{Transfer to MIMIC-IV without refitting}

All analyses so far used eICU-CRD, on which the twin was trained. To test whether the gain holds in a second database, we applied the five frozen checkpoints, without refitting, to the person-disjoint test split of MIMIC-IV (14,211 stays) in one registered read. This split had been read before, including zero-shot transfers of earlier versions of the eICU-CRD-trained twin, fine-tunes of the same pretrained model and outcomes that overlap ours (Methods; Supplementary Note~7). The \sao{} target was not scored, because in MIMIC-IV that channel holds sparse blood-gas values. This left 9 settings, with thresholds at the MIMIC-IV training means and a \spo{} target recorded differently from that of eICU-CRD (Methods).

The synchronized forecast outperformed the hour-6 forecast in 8 of 9 settings after Holm correction (estimates $+0.025$ to $+0.114$ AUROC; not significant in \spo{} at (30, 24)). This meets the registered transfer criterion of at least 5 of 9 settings with none significantly worse before correction, but unlike on eICU-CRD the gain was not significant in every setting. Against the patient's last measured value, the synchronized forecast was better in 1 of 9 settings after Holm correction and, unlike on eICU-CRD, worse in 2: \spo{} at (12, 6) and heart rate at (30, 24) (median difference $-0.011$ AUROC, range $-0.084$ to $+0.034$). The registered secondary criterion was therefore not met (Table~\ref{tab:gates}). The control inputs were not repeated on MIMIC-IV, so the attribution of the gain to the patient's newly recorded measurements rests on the eICU-CRD validation split.

As disclosed secondary outcomes with no registered criterion, the read also scored the frozen hour-48 forecasts of death and of leaving the unit. Because the harmonized records keep a hospital-death flag without its time, the death forecast was scored against hospital death, with AUROCs of 0.687 for death in a hospitalization whose unit stay ended within the next 48 hours and 0.726 for death at any time in the hospitalization; the AUROC for leaving the unit within 48 hours was 0.765. Both mean forecasts fell below the observed rates, in part by construction (Methods; Table~\ref{tab:gates}).

\subsection*{A fine-tuned twin state approaches gradient boosting on eICU-CRD and adds to it for acute kidney injury}

Beyond its own forecasts, a twin's state could serve as an input to risk models of clinical endpoints. Two registered analyses run during model development, the endpoint probes, asked whether the twin's state predicts the four clinical endpoints within 24 hours of landmarks at 12, 24 and 48 hours, and whether it adds information to the boosting recipe above. Probe P0 read the frozen state, and probe P1 trained the encoder on the endpoints (Methods). They were scored on a tuning set of the eICU-CRD training split that the pretrained encoder had seen in training, which favors the twin's states. Used alone, each representation trailed boosting: an encoder trained on the endpoints from random initialization by 0.004 to 0.021 AUROC, with every interval below zero, one fine-tuned from the pretrained twin by 0.010 to 0.029 and the frozen state by 0.020 to 0.086. The hybrid, which adds the fine-tuned state to the boosting features, improved AUROC by 0.0042 for hypotension and 0.0075 for acute kidney injury, with unadjusted intervals above zero (Supplementary Fig.~S3). Neither probe met its registered success criterion (P1 missed it narrowly; Table~\ref{tab:gates}), and the probes did not show a benefit of pretraining (Methods).

We refitted the boosting comparator, both encoders and the hybrid within the MIMIC-IV training split (66,007 stays) with the recipes above, with five seeds for the encoders and early stopping on 10\% of training patients, and the MIMIC-IV test split was read once. Death in the unit was not scored as an endpoint, because the harmonized records lack its time (Methods). Adding the fine-tuned state to the boosting features improved discrimination of acute kidney injury by $+0.0035$ (95\% CI $+0.0007$ to $+0.0065$; one-sided Holm-adjusted $P=0.018$), which was significant under the registered test, but not significantly that of hypotension ($+0.0010$; $-0.0012$ to $+0.0033$; $P=0.18$). The encoder trained from random initialization trailed boosting by 0.0109 to 0.0294 AUROC across the three endpoints, and because no non-inferiority margin was registered, these deficits were not tested against one; the fine-tuned encoder was below the randomly initialized one for hypotension and acute kidney injury (intervals below zero; Methods).

\subsection*{Reproducing the measurement process lets the generator pass its audit}

\begin{figure}[tbp]
\centering
\includegraphics[width=0.98\textwidth]{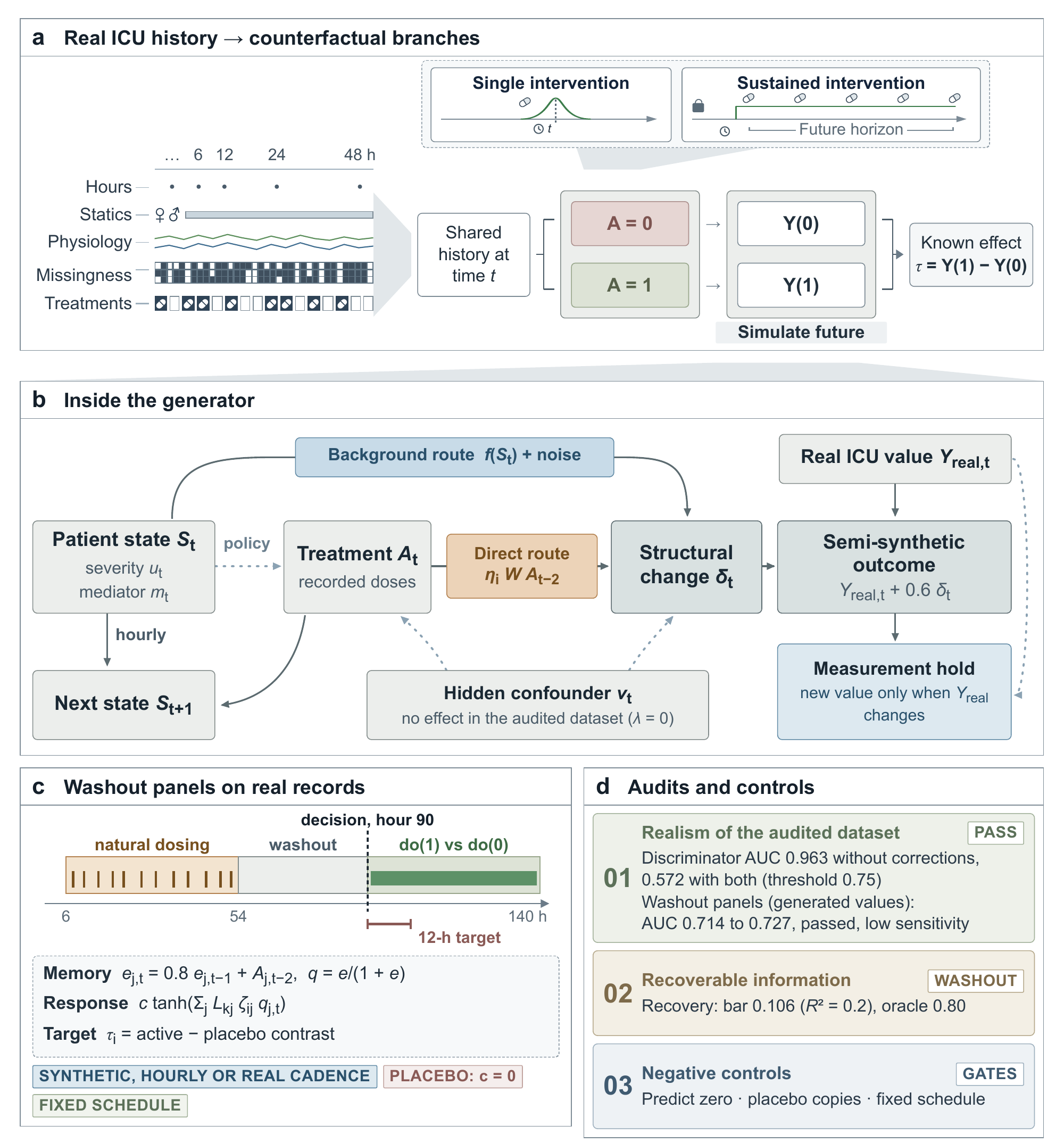}
\caption{\textbf{Semi-synthetic data on real intensive care records.} \textbf{a}, A real eICU-CRD history branches into simulated treatment arms, so the effect $\tau=Y(1)-Y(0)$ is known. \textbf{b}, A latent severity $u_t$ and mediator $m_t$ evolve hourly under the recorded treatments. The structural change $\delta_t$ combines a background route with a direct treatment route lagged by 2 hours and scaled by a per-stay responsiveness $\eta_i$. The generated outcome, the real value plus $0.6\,\delta_t$, is recorded only in hours in which the real value changes (measurement hold). Dotted arrows mark links switched off in the factual record (the treatment policy acts only in the counterfactual arms, and the hidden confounder has weight $\lambda=0$) or the real change times used by the hold. The generator's latent state is labeled $S_t$ in the panel and is distinct from the twin's state $z_t$. \textbf{c}, Washout panels, which carry every individual-effect result. Real windows supply the outcomes, and dosing is simulated: natural dosing, a washout, then do(1) versus do(0) from hour 90. A bounded response to a decaying memory of past doses, with patient-specific traits $\zeta_{ij}$, is injected at amplitude $c$. The target $\tau_i$ is the 12-hour contrast between the arms in the active frame minus that in a placebo frame with $c=0$. \textbf{d}, Audits and controls: the realism audit of the generated dataset (Fig.~\ref{fig:benchmark}b); the recovery bar ($R^2=0.2$, or 0.106 on the scale $1-\mathrm{RMSE}/\mathrm{SD}(\tau)$ of Figs.~\ref{fig:map} and \ref{fig:estimators}) against an oracle reference with a recovery of 0.80 (Fig.~\ref{fig:map}d); and the negative controls used as gates. The panels share the generator's code and measurement hold but use a separate frozen parameterization, and their realism audit is weak evidence (Fig.~\ref{fig:benchmark}b; Results). RMSE, root mean squared error.}
\label{fig:generation}
\end{figure}

\begin{figure}[tbp]
\centering
\includegraphics[width=\textwidth]{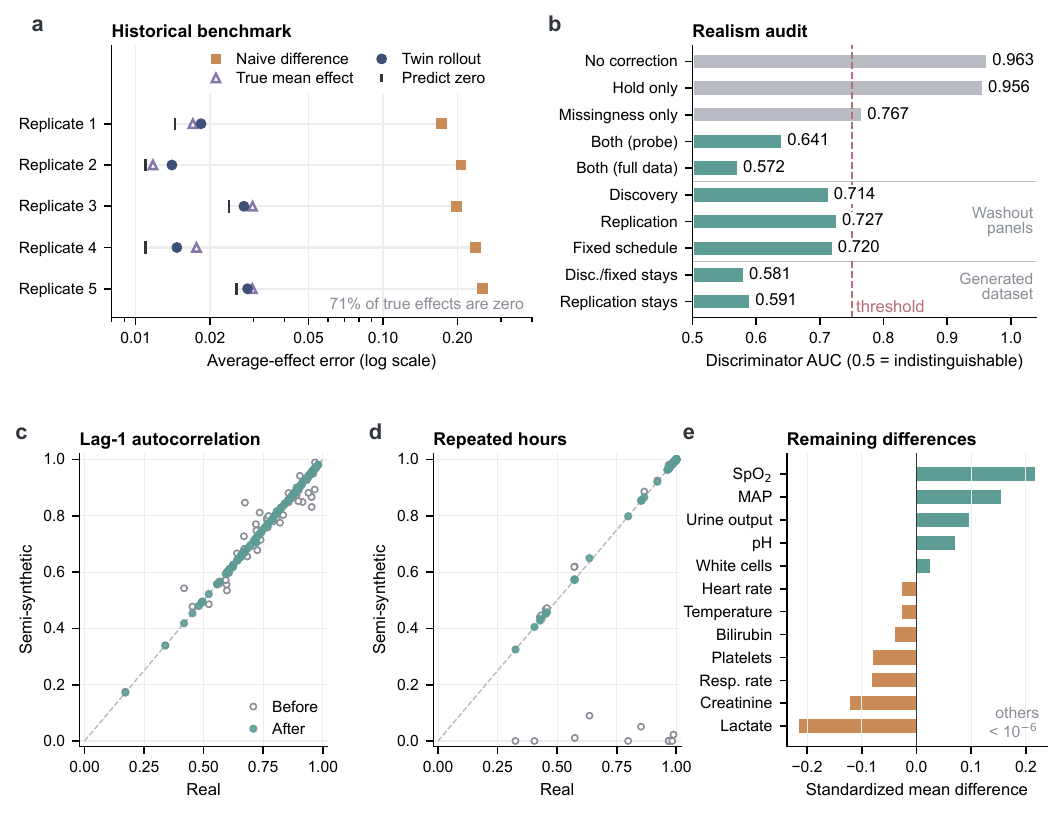}
\caption{\textbf{Reproducing the measurement process lets the generator pass its realism audit.} \textbf{a}, Error in the average treatment effect on the historical benchmark, in five replicates (log scale). The twin rollout, from an earlier model version with a clock defect that shifted its generated physiology by at least one hour (Methods), has a far lower error than the naive treated-versus-untreated difference but a higher error than predicting zero in all five replicates, because the true average effect is zero in 71\% of the outcome, treatment and horizon combinations. The true-mean-effect arm assigns the true mean effect to every combination. \textbf{b}, Audit AUC of the locked realism audit, a person-grouped ridge discriminator on 644 summaries of 161 channels in 400 matched patients (0.5, indistinguishable); teal bars pass its threshold of 0.75. Generator versions are shown on 3,000-patient probes (no correction, hold only, missingness correction only, both) and, with both corrections, on the full dataset of 200,825 stays. The last bars apply the same audit to the 140-hour windows of the three washout panels (discovery, replication, fixed schedule), each panel's generated outcome values, set to missing wherever the real record is missing, with the real record for every other channel, and to the generated dataset on the same stays and windows (Methods). On these windows the audit did not flag measurement noise of up to 0.15 reference standard deviations added to the real records, and as stored the panels were separable from real records (Supplementary Table~S29). \textbf{c},\textbf{d}, Lag-1 autocorrelation (\textbf{c}) and share of hours that repeat the last recorded value (\textbf{d}) per channel, real against semi-synthetic records, for an earlier generator version that wrote a new value every hour (before; audit AUC 0.9999) and for the audited dataset (after). \textbf{e}, Standardized mean differences (semi-synthetic minus real) that remain in the audited dataset. Only the 12 written outcome channels differ; all others match to within $10^{-6}$. AUC, area under the receiver operating characteristic curve.}
\label{fig:benchmark}
\end{figure}

We now turn to the second capability, predicting how a patient would respond to treatment. A response module, the part of a twin that makes this prediction, must be tested on records in which each patient's effect is known, which real records never reveal. The twin's counterfactual rollout, which generates the next hour under a chosen action, had been evaluated only on the semi-synthetic benchmark of earlier work, which we call the historical benchmark (Supplementary Fig.~S2), and that benchmark could not show that the twin recovers effects. On it, an earlier twin's rollout estimated average effects with a much lower mean absolute error than a naive treated-versus-untreated comparison (0.021 against 0.213), but it was less accurate than predicting zero effect (0.017) in all five replicates, because the true average effect was zero in 71\% of the benchmark's outcome, treatment and horizon combinations (Fig.~\ref{fig:benchmark}a; Supplementary Fig.~S2; Supplementary Note~1). We therefore built a semi-synthetic generator that keeps each real eICU-CRD record and writes treatment responses onto 12 outcome channels, so that outcomes under both treatments are known (Fig.~\ref{fig:generation}a,b; Methods). The analyses that follow do not score the twin's rollout: they ask how much of each patient's effect the records reveal and whether learners recover it from factual data.

We checked realism with a classifier two-sample test\cite{lopezpaz2017revisiting,alaa2022faithful}: a ridge logistic discriminator was trained to tell generated from real records of 400 matched patients from 644 summaries per patient, such as each channel's mean and lag-1 autocorrelation. Its AUROC, the audit AUC, is 0.5 when the two cannot be told apart. A generator version passed at an audit AUC of at most 0.75, a threshold locked in advance, if its standardized-mean-difference, missingness and label checks were also met (Methods).

Without corrections, the generator failed on a probe of 3,000 patients (audit AUC 0.963; Fig.~\ref{fig:benchmark}b). An earlier version (audit AUC 0.9999) differed from real records mainly in when values were recorded: in the audit's 400 real patients, creatinine repeated its previous value in 97\% of hours, whereas that version wrote a new value every hour (Fig.~\ref{fig:benchmark}c,d). Two parameter-free corrections reproduce this measurement process. The measurement hold records a new generated value only when the real value changed, and the missingness correction restores the gaps that real records leave in the treatment channels. Neither passed alone (audit AUC 0.956 and 0.767); together they gave 0.641 on the probe and 0.572 on all 200,825 generated stays, with every other check met (Table~\ref{tab:gates}). Because the corrections changed only how values are recorded and no generator parameter was tuned, the earlier failures had mainly reflected the measurement process; only the 12 written channels still differ from the real record (Fig.~\ref{fig:benchmark}e). The audit AUC bounds the distance between real and generated summaries only from below, so a pass cannot show that they agree in every respect (Proposition~3; Methods).

\subsection*{Washout panels carry individual responses of known size}

The generated dataset's interventions last a single hour, so the individual-effect analyses use washout panels built with the same code, which sustain the intervention and define each patient's effect against a copy with the injected response switched off (Fig.~\ref{fig:generation}c; Methods). Each panel holds 140-hour windows of 400 real eICU-CRD records with simulated dosing of one treatment: a pattern that varies between patients in window hours 6 to 53 (natural dosing), a washout until hour 89 and, from hour 90, sustained dosing in one counterfactual arm, do(1), and none in the other, do(0). No dose occurs in the 36-hour washout before the decision, so no patient received the queried regime and it lies outside the treatment support by design (Proposition~6). The benchmark therefore asks whether a model can carry what each patient's earlier responses reveal to a regime that the data never show. The injected response depends on decaying past exposure and on a patient-specific trait and is capped at an amplitude in reference standard deviations, a fixed scale per outcome. The effect $\tau_i$ is the 12-hour mean difference between the arms caused by this response. Because the same trait shapes the responses recorded before the decision, the pre-decision record (window hours 0 to 89, stay hours 24 to 113) can carry information about $\tau_i$, whereas measurement times come from the real record and do not depend on the effect (Supplementary Notes~4 and 5). The discovery panel holds the 400 stays of the exploratory analyses below, the replication panel 400 stays, 395 of them new, and the fixed-schedule panel instead doses every discovery patient in the same hours (window hours 6 to 24).

Applied to each panel's 140-hour window, with the panel's generated outcome values set to missing wherever the real record is missing and the real record for every other channel, the locked audit passed all three panels (audit AUC 0.714, 0.727 and 0.720 in the discovery, replication and fixed-schedule panels; threshold 0.75; Fig.~\ref{fig:benchmark}b; Methods), and it also passed them on window hours 0 to 89, which the estimators read (audit AUC 0.713 to 0.738). On the 140-hour windows, however, it did not flag measurement noise of up to 0.15 reference standard deviations added to the real records and did not flag a uniform heart-rate shift of 15 beats per minute (Supplementary Table~S29). As stored, with filled cells and simulated dosing, all three panels were separable from real records (audit AUC 0.968 to 0.978). The pass is therefore weak evidence and does not show that the panels resemble real records.

\subsection*{Recovery of injected responses depends on measurement cadence}

Before testing estimators that see only factual data (the treatments given and the outcomes observed), we asked how much the pre-decision record reveals about each patient's effect. We measured it with readers: ridge regressions fitted with the true per-patient effect as the label, which do not use the twin and cannot be deployed. Their cross-validated $R^2$ estimates a lower bound on the effect information in their features (Proposition~6); the figures plot the equivalent recovery, $1-\mathrm{RMSE}/\mathrm{SD}(\tau)=1-\sqrt{1-R^2}$. The history-blind reader uses the raw pre-decision values, with no model of how past doses accumulate. The occupancy-contrast reader projects each channel on a detrended, saturating summary of the patient's recent doses (occupancy), and the cadence-aware reader uses only the hours in which the recorded value changes (the inferred measurement times); both take their exposure family from the generator (Methods). A reader clears the registered bar of $R^2=0.2$ (recovery 0.106) when its bootstrap probability of lying below the bar is under 0.05. It separates when it clears the bar and, if the history-blind reader clears it too, exceeds that reader with a bootstrap interval of the difference above zero.

\begin{figure}[tbp]
\centering
\includegraphics[width=\textwidth]{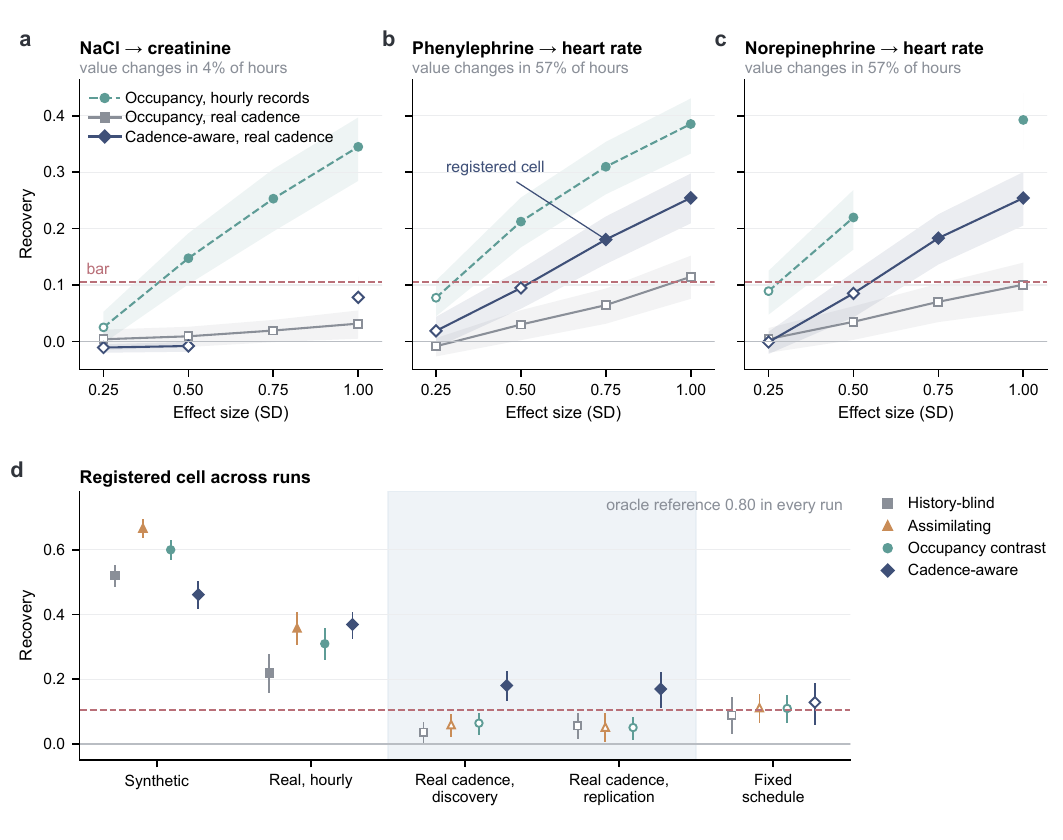}
\caption{\textbf{Recovery of injected responses depends on measurement cadence and on how the record is read.} \textbf{a}--\textbf{c}, Recovery of the individual memory-route effect from the pre-decision record against effect size (amplitude cap in reference standard deviations, SD) for NaCl acting on creatinine (\textbf{a}), phenylephrine on heart rate (\textbf{b}) and norepinephrine on heart rate (\textbf{c}), on the same 400 real eICU-CRD records. Each panel gives the share of hours in which the responding variable's recorded value changes. Of the records shown, only the three real-cadence washout panels of the registered cell (\textbf{d}) were realism-audited, and that audit is weak evidence (Fig.~\ref{fig:benchmark}b). Recovery is $1-\mathrm{RMSE}/\mathrm{SD}(\tau)$, and the dashed bar of 0.106 corresponds to $R^2=0.2$. Readers are ridge models fitted with the true per-patient effect, so they are not deployable estimators (Methods). Lines show the occupancy-contrast reader on hourly records and at real cadence, and the cadence-aware reader at real cadence. Bands are 95\% whole-person bootstrap intervals, and filled markers clear the bar. Missing points: the cadence-aware reader at 0.75 SD for NaCl and hourly records at 0.75 SD for norepinephrine. \textbf{d}, The registered cell, phenylephrine acting on heart rate at 0.75 SD, in five registered runs: synthetic records, real records observed every hour, real records at real cadence in the discovery and replication panels (both shaded) and a fixed schedule. The history-blind reader uses the raw pre-decision record. The assimilating reader normalizes the occupancy contrast by the energy of the patient's occupancy profile and adds exposure summaries and channel means; despite its name, it does not use the twin. An oracle reference that knows each patient's trait reaches a recovery of 0.80 ($R^2=0.96$) in every run. RMSE, root mean squared error.}
\label{fig:map}
\end{figure}

We mapped recovery at several amplitudes for three treatment-response pairs on the same 400 stays (Fig.~\ref{fig:map}a--c). Heart rate, the target of the injected phenylephrine and norepinephrine responses, changes its recorded value in 57\% of hours; creatinine, the target of the injected saline (NaCl) response, in 4\%. On hourly records, with a new generated value every hour, the occupancy-contrast reader cleared the bar in all three pairs from half a standard deviation upward ($R^2$ 0.57 to 0.63 at one standard deviation). At real cadence, with a new value only when the real one changed, it cleared the bar in no pair at any amplitude; its highest $R^2$, 0.22, had an interval reaching below the bar. Of the mapped readers, only the cadence-aware reader cleared it at real cadence, on heart rate for both drugs at 0.75 and 1 standard deviation ($R^2=0.44$ at one standard deviation; Fig.~\ref{fig:map}b,c). No reader cleared it on creatinine at real cadence at any amplitude tested (Fig.~\ref{fig:map}a; Supplementary Table~S2).

We then registered one cell of this map, one pair at one amplitude: phenylephrine acting on heart rate at 0.75 reference standard deviations, which caps each patient's shift at 11.25 beats per minute. It was the smallest amplitude at which the cadence-aware reader had separated in an exploratory sweep on the discovery stays. At real cadence, the reader separated in the discovery panel ($R^2=0.33$) and in the replication panel ($R^2=0.31$), exceeding the history-blind reader in both (Fig.~\ref{fig:map}d; Supplementary Table~S3; Supplementary Note~1). The replication panel reuses the discovery panel's person-indexed generator draws, traits included, so it tests recovery of nearly identical effects (correlation 0.9998) from largely new records. 
Because the discovery run reproduces the exploratory run, the new evidence comes from the replication and fixed-schedule panels. On the fixed schedule, without differences in exposure between patients, the reader did not separate ($R^2=0.24$, interval reaching below the bar), as the registered claim anticipated (Methods).

\subsection*{None of 16 standard learners beats predicting zero effect outside the treatment support}

\begin{figure}[tbp]
\centering
\includegraphics[width=\textwidth]{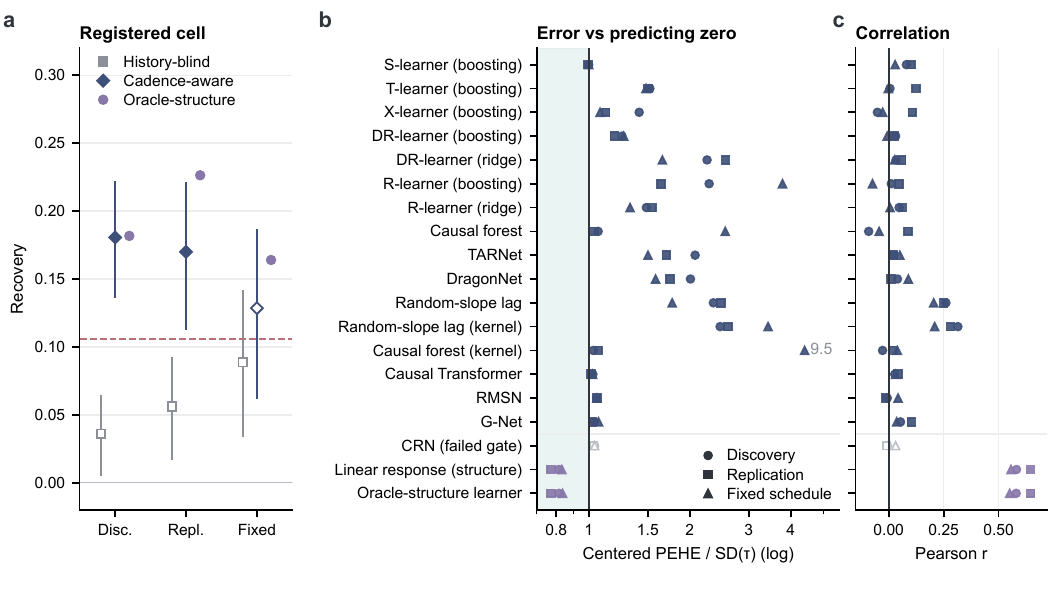}
\caption{\textbf{No standard learner beats predicting zero effect in the registered cell.} \textbf{a}, Recovery ($1-\mathrm{RMSE}/\mathrm{SD}(\tau)$) in the registered cell by the history-blind and cadence-aware readers, which are fitted with the true effect (95\% intervals; filled when clearing the dashed bar of 0.106, equal to $R^2=0.2$), and by the oracle-structure learner, which uses factual data only but is given the generator's response structure. \textbf{b}, Centered PEHE divided by that of predicting zero, which equals the standard deviation of the true effect (point estimates, log scale; a value beyond the axis is printed at its edge). Shown are 13 static learners, two sequence models (Causal Transformer and RMSN), the g-computation model G-Net (Causal Transformer code), CRN (excluded after failing its synthetic gate), a structure-informed linear response model and the oracle-structure learner; ``kernel'' marks kernel-informed variants. No standard learner's 95\% bootstrap interval of the difference from predicting zero lay wholly below zero. \textbf{c}, Correlation of each learner's predictions with the true individual effects. The two random-slope lag models correlated 0.20 to 0.31 with the true effects, but in the discovery and replication panels their predictions spread 2.4 to 2.7 times as widely as the true effects (Supplementary Table~S16). Registered comparisons with the oracle-structure learner gave 95 wins, no losses and 7 unresolved comparisons (six of them with the excluded CRN) among 102 comparisons, with Holm correction within three families (family-wise error 0.15). Every win was over a comparator that did not itself beat zero. The decision query lies outside the treatment support: no patient received the query regime, so every estimator must extrapolate to it (Methods). No query inside the support met the registered overlap criteria (Methods). Disc., discovery; Repl., replication (400 stays, 395 of them new, carrying nearly identical injected effects); Fixed, fixed schedule; CATE, conditional average treatment effect; PEHE, precision in estimation of heterogeneous effects; RMSE, root mean squared error; RMSN, recurrent marginal structural network; CRN, counterfactual recurrent network.}
\label{fig:estimators}
\end{figure}

Because readers see the true effects, we asked whether estimators that see only factual data extract the same information. Three registered benchmarks on the registered cell (Fig.~\ref{fig:estimators}) evaluated 13 static learners of the conditional average treatment effect (CATE; Methods), three sequence models, the Causal Transformer\cite{melnychuk2022causal}, recurrent marginal structural networks (RMSN)\cite{lim2018rmsn} and the counterfactual recurrent network (CRN)\cite{bica2020crn}, and a g-computation sequence model, which simulates every recorded variable forward under the queried treatments: G-Net\cite{li2021gnet}, as implemented with the Causal Transformer code. CRN failed its synthetic positive control with a large effect and was excluded; every scored learner passed it. G-Transformer\cite{xiong2024gtransformer}, a second g-computation sequence model, was not run, because we found no released code (Methods).

Most static learners were fitted to factual 12-hour treatment windows starting in window hours 6 to 42, before the washout; every model was then queried for dosing from hour 90 (the decision query). We scored learners by centered PEHE (precision in estimation of heterogeneous effects), the standard deviation across patients of the error in the predicted effect. It ignores a constant offset, so predicting zero, or any constant, scores the standard deviation of the true effects, and a learner beats this reference only if it captures differences between patients. None of the 16 scored learners outperformed predicting zero in any panel: no 95\% bootstrap interval of a learner's centered PEHE minus that of predicting zero lay wholly below zero (Fig.~\ref{fig:estimators}b and Supplementary Table~S4). Most learners' predictions were essentially uncorrelated with the true effects ($|r|\le0.124$; Fig.~\ref{fig:estimators}c), so no rescaling could lower their centered PEHE below that of predicting zero by more than 1\% (Proposition~6). The two random-slope lag models correlated 0.20 to 0.31 with the true effects, but in the discovery and replication panels their predictions spread 2.4 to 2.7 times as widely as the true effects (Fig.~\ref{fig:estimators}c; Supplementary Table~S16).

Because the decision query lay outside the treatment support, every estimator must extrapolate to it through assumptions (Proposition~6). Neither registered hour inside the dosing period (30 or 42) met the registered overlap criteria on the recorded treatments (Methods), so no control inside the support was scored, and the benchmark cannot distinguish a failure of the learners from the lack of treatment variation at the query.

Given the response structure, the factual record does hold information about each patient's effect. The oracle-structure learner sees factual data only but is given the generator's trait prior, bounded response map and measurement convention. It reached $R^2$ of 0.33, 0.40 and 0.30 in the discovery, replication and fixed-schedule panels, as high as the cadence-aware reader on the discovery panel (Fig.~\ref{fig:estimators}a), and its predictions correlated 0.55 to 0.65 with the true effects. It was better than the standard learners in 95 of 102 registered comparisons after Holm correction within the static (78 comparisons), sequence (18) and g-computation (6) families, with no losses. Every win was over a comparator that did not itself beat zero. A linear response model given the same structure matched it on both metrics, so the advantage does not require the bounded response map.

\subsection*{ICare-RH recovers simulated responses but fails its placebo gate}

\begin{figure}[tbp]
\centering
\includegraphics[width=\textwidth]{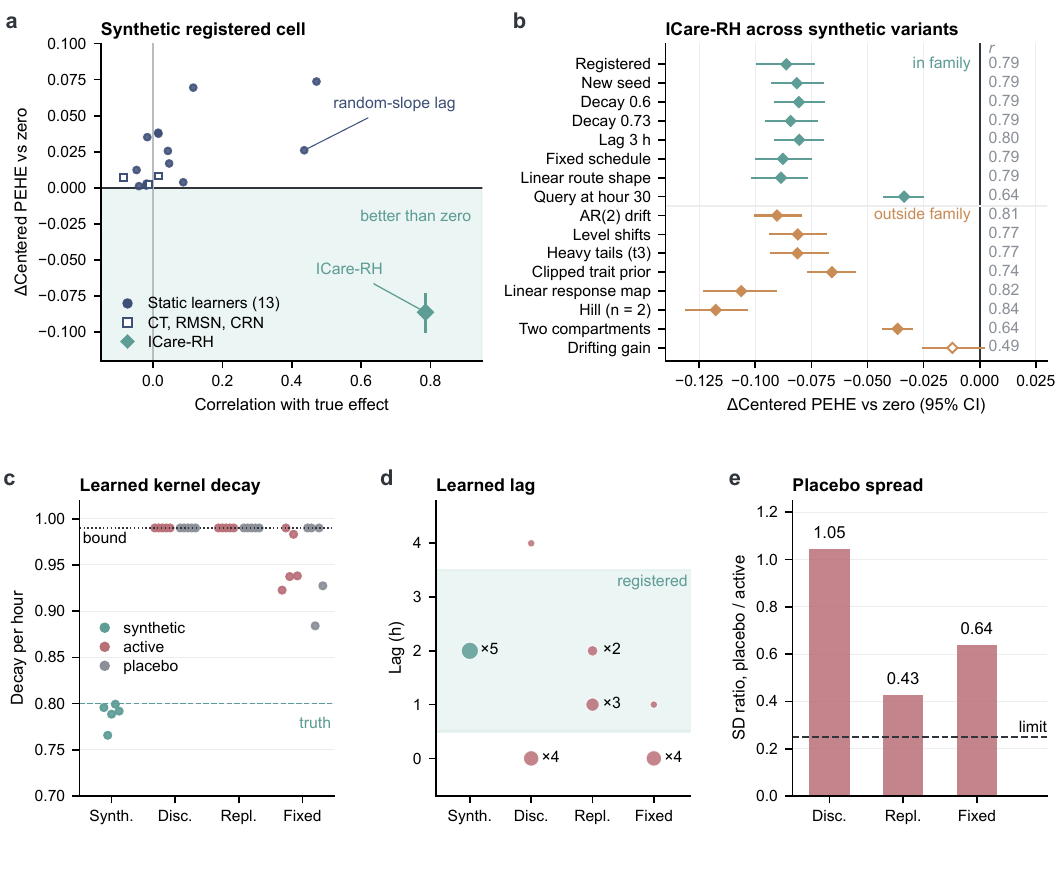}
\caption{\textbf{ICare-RH recovers simulated responses but fails its placebo gate.} ICare-RH is a hierarchical random-slope response model fitted to factual data (Methods). \textbf{a}, On the synthetic registered cell (simulated patients from a generator matched to the registered cell; Methods), correlation with the true effect against centered PEHE minus that of predicting zero, for ICare-RH (95\% interval), 13 static learners and three sequence models (Causal Transformer, CT; RMSN; CRN, excluded from the registered benchmark after failing its synthetic positive control). This comparison is descriptive and spans separately registered analyses. \textbf{b}, ICare-RH across synthetic variants inside and outside its model family (95\% intervals, filled when the interval lies below zero; correlations $r$ at right). \textbf{c}--\textbf{e}, Label-free placebo gate on the washout panels, run before any target was read. \textbf{c}, Learned kernel decay per fold on synthetic records (true decay 0.8, dashed) and on the active and placebo (zero-effect) copies of the three washout panels; the parameter's upper bound of 0.99 is dotted. The decay is a diagnostic outside the gate's clauses. \textbf{d}, Learned response lag per fold on synthetic and active records, with the registered range of 1 to 3 hours shaded. \textbf{e}, Spread of predicted effects on placebo records relative to active records, against the registered limit of 0.25. ICare-RH was not scored on real records. Synth., synthetic; Disc., discovery; Repl., replication; Fixed, fixed schedule; PEHE, precision in estimation of heterogeneous effects; RMSN, recurrent marginal structural network; CRN, counterfactual recurrent network.}
\label{fig:rh}
\end{figure}

Because the oracle-structure learner and the linear response model are given the generator's structure, neither is deployable. We built ICare-RH, a candidate response module for ICare-DT that is fitted to factual data alone and takes no numerical constant from the generator, although its design drew on the oracle-structure learner. The registered version does not yet read the ICare-DT state (Methods). Each patient has a random response slope and a mean-reverting baseline, and the lag and decay of an exposure kernel are fitted by marginal likelihood. On 400 simulated patients from a generator matched to the registered cell, ICare-RH correlated 0.786 with the true effects and lowered centered PEHE below that of predicting zero by 0.086 (95\% CI 0.073 to 0.100; Fig.~\ref{fig:rh}a), passing all four of its synthetic gates. Outside its model family, it beat predicting zero in seven of eight variants and failed only when the response gain drifted (Fig.~\ref{fig:rh}b; Methods).

On the washout panels, ICare-RH failed its registered placebo gate and was therefore never scored on real records (Fig.~\ref{fig:rh}c--e). The gate's first checks are label-free and ran before any target was read, on each panel and on a placebo copy without the injected effect, on which predicted effects should vary far less than on the active panel. Two of them failed. The learned response lag fell outside the required 1 to 3 hours in all five discovery folds, and on placebo records predicted effects spread 1.05, 0.43 and 0.64 times as widely as on active records in the discovery, replication and fixed-schedule panels, against a limit of 0.25. The kernel decay sat at its upper bound in every fold of both natural-dosing panels (discovery and replication), on placebo and active records alike, as it had on synthetic panels with baseline drift or level shifts but no injected response (Supplementary Note~5); at that bound, the exposure feature can absorb slow drift in the baseline. These data cannot tell whether a misspecified baseline, an uncaptured response, a non-global optimum or finite-sample variation caused the failure (Proposition~7).

\section*{Discussion}

Of the two capabilities that a patient's digital twin needs, routine intensive care records already support one: forecasts that improve as ICare-DT assimilates the patient's own measurements through synchronization. One is not yet supported: an estimator that sees only factual data and recovers individual effects without being given the response structure.
Injected individual responses were recovered from real records only by models that a deployed twin cannot use, those given the true effects or the generator's response structure, and at real measurement cadence only for a frequently measured variable read at its measurement times.

For forecasting, our evidence for the twin concerns assimilation: its forecasts improved as it read the patient's own new hours, although on eICU-CRD validation data a gradient-boosted model reading the same hours was the stronger forecaster.
Our two control inputs test where this gain comes from and let twin builders check two of the National Academies' three criteria\cite{nasem2024digital,tudor2025digitaltwins}: the record held at hour 6 tests dynamic updating, and another patient's hours test personalization (Fig.~\ref{fig:sync}b,d); the third criterion, informing decisions, was not tested. The twin forecast better from the patient's own hours than from another patient's, a necessary but not sufficient condition for personalization, because the patient's own last measured value held a similar advantage without any model; information beyond that value was not shown.
In the registered MIMIC-IV read, the frozen model met its transfer criterion for the synchronization gain but not the secondary criterion against the last measured value, which outperformed it in 2 of 9 settings after Holm correction (Results).
Evaluations of digital twins should therefore report such trivial references beside learned comparators.

A fine-tuned twin state can also complement a gradient-boosted risk model. With all models refitted within MIMIC-IV, adding the fine-tuned twin state to gradient boosting gave a significant gain for acute kidney injury only ($+0.0035$ AUROC). This complementarity suggests hybrid designs in which a twin's state feeds established risk models, although whether a gain of this size matters clinically was not assessed.

A complete twin must also answer questions about treatment. The authors of ICareFM, an intensive care foundation model, name the separation of treatment effects from physiological deterioration an open problem\cite{burger2025foundation}, and our results suggest how a twin that addresses it could be built and tested.
In such a twin, synchronization would supply forecasts that improve with the patient's own data, and a separate response module could be attached once it passes placebo checks and recovers injected effects at real measurement cadence. The twin's own rollout has not yet been tested this way.

For one injected response route in eICU-CRD, the recovery results of Fig.~\ref{fig:map} suggest what such a response module needs.
Plausible first targets are responses carried by frequently sampled vital signs, such as heart rate after vasoactive drugs, if the record is read at its measurement times (Proposition~5). The registered heart-rate amplitude, however, lies in the range of differences between phenylephrine and norepinephrine during spinal anesthesia\cite{chen2022norepinephrine,ngankee2015randomized}, whereas a randomized trial in septic shock found no such difference\cite{morelli2008phenylephrine}. No reader recovered a creatinine response at real cadence. On the panels, only learners given the generator's response structure recovered the effects from factual data, which suggests that a deployable module should learn that structure from the data; carrying it to a regime that no patient received still rests on assumptions (Proposition~6). It should also allow for slow drift in the patient's baseline, which these data cannot rule out as the cause of ICare-RH's placebo failure (Proposition~7). Finally, the panels copy measurement times from the real record, so sampling never responded to the injected effect. Clinicians often measure a variable more frequently after changing a treatment, and such treatment-responsive sampling\cite{vanderschueren2023informative} remains to be tested.

These conclusions rest on a benchmark built to reproduce when values are recorded. Our results, from one database, suggest that benchmarks of irregularly sampled health records should reproduce their measurement process and report a realism audit with a preset threshold.
The generated outcome values of the washout panels used here passed such an audit, but on their 140-hour windows it did not flag measurement noise of 0.15 reference standard deviations added to the real records, and as stored the panels were separable from real records, so the pass does not show that they resemble real records. Their dosing was simulated, and their stays were drawn without regard to the eICU-CRD split.

Several limitations qualify these conclusions. All results come from data already used during model development: the eICU-CRD validation and test sets had been read before, the endpoint probes' tuning set was scored twice, and several analysis choices followed earlier results (Supplementary Note~2).
All data come from hospitals in the United States. Apart from one registered read of MIMIC-IV for synchronization and the clinical-endpoint probes, with hospital death and leaving the unit as disclosed secondary outcomes, the analyses used eICU-CRD. The MIMIC-IV test split had been read before, including zero-shot transfers of earlier versions of the twin, fine-tunes of the same pretrained model and overlapping outcomes such as hospital death and length of stay, and we knew those results when we designed the read (Supplementary Note~7). Because the harmonized records keep no death time, the death forecast, which targets death in the unit before each head's crossing, was scored against hospital death. The frozen model was applied to MIMIC-IV without adaptation, although several inputs are recorded at different frequencies there, above all the \sao{} channel, which holds monitor saturation readings and blood-gas values in eICU-CRD but blood-gas values only in MIMIC-IV. The control inputs and the treatment-response analyses were not repeated on MIMIC-IV.
The forecast targets are crossings of each vital sign's training-set mean, which mark excursions and are not clinical events. The synchronization gain was scored in stays with hourly measurement of the target, together 42.9\% of test stays. In descriptive subgroups of the hour-48 forecasts, Hispanic patients were evaluable in only five settings, where their median AUROC was 0.58, against 0.71 and 0.70 for White and Black patients in those settings (Methods). The gradient-boosting comparator was a fixed CatBoost recipe and was never tuned. The washout panels inject simulated responses, so they establish no real treatment effect, and because their treatment is a binary hourly exposure, they do not cover dose--response effects\cite{gogl2025dosesurv}. The decision query lies outside the treatment support, and no query inside it met the registered overlap criteria, so the benchmark tests extrapolation under structural assumptions as well as effect estimation. G-Net was run as implemented with the Causal Transformer code, with declared changes (Supplementary Note~9), a configuration published for other datasets and limited tuning, and its bootstrap intervals hold the fitted models fixed. G-Transformer was not run, because we found no released code. The propositions were derived after the results and rest on idealizations (Methods).

Within these limits, assimilating the patient's own newly recorded measurements already improves ICare-DT's forecasts over those it made from the stay's first six hours.
Only models given the true effects or the response structure were shown to recover injected individual responses from eICU-CRD records.
The next step is a response module that needs neither: one that learns the response structure from factual data, passes placebo checks and recovers injected responses at real measurement cadence before it is tested against real treatment decisions, for example by emulating a randomized target trial\cite{hernan2016using,hernan2022target}. Before clinical use, the twin's forecasts would also need clinically defined targets, recalibration and prospective evaluation in the units where it is deployed. To support this work, we will release the generator and washout-panel code as a testbed for response modules, with the registered protocols and analysis records\cite{nosek2018preregistration,collins2024tripodai,lekadir2025futureai}.

\section*{Methods}

\subsection*{Data sources and cohorts}

We used three critical care databases that are available to credentialed users through PhysioNet\cite{goldberger2000physionet}: the eICU Collaborative Research Database (eICU-CRD)\cite{pollard2018eicu}, MIMIC-IV\cite{johnson2023mimic} and the CareVue era of MIMIC-III\cite{johnson2016mimic}. They were harmonized with the BlendedICU pipeline\cite{oliver2023introducing}, which our code base extends to MIMIC-III. The experimental cohort of 323,677 stays comprised 200,825 eICU-CRD stays of 139,362 patients from 208 hospitals, 94,458 MIMIC-IV stays of 65,366 patients and 28,394 MIMIC-III CareVue stays of 23,692 patients (Table~\ref{tab:cohorts}). The MIMIC-III cohort was restricted to CareVue patients who do not appear in MIMIC-IV. Before this restriction, it held 61,532 stays of 46,476 patients.
All analyses use eICU-CRD except the registered MIMIC-IV read of the two prediction results and its disclosed secondary outcomes.
MIMIC-III enters this paper only through earlier transfer analyses, which were not registered and are disclosed in Supplementary Note~8.

Records were placed on an hourly grid with 161 time-varying channels and four static covariates (age, sex, height and weight). Of the channels, 41 are physiological variables: 11 vital signs, 2 respiratory mechanics variables, 23 laboratory values, urine output and four Glasgow Coma Scale channels. The other 120 are treatment channels: drug, infusion and fluid exposures, and five ventilator settings. Medications were encoded as hourly exposures. A dose given over an interval was spread over the hours it overlapped, and a bolus was assigned to its hour. Because BlendedICU forward-fills values before export, its observation mask records availability after forward filling and does not mark measurement. From BlendedICU's hourly data before forward filling, we therefore built a measurement record that marks the hours in which each of the 41 physiological variables was measured. The model uses this measurement record as the observation mask of the physiological variables, and BlendedICU's availability mask as that of the treatment channels. For the four tracked vital signs, the measurement record also defines the forecast origins, labels and censoring (Model). The semi-synthetic generator and the washout panels instead infer measurement times from changes in the recorded value (Semi-synthetic generator and realism audit). Normalization statistics, and with them the crossing thresholds, were computed on training data only. They came from the eICU-CRD training split for the eICU-CRD analyses and from the MIMIC-IV training split for the MIMIC-IV read.

eICU-CRD was split by person into training (140,473 stays of 97,553 patients), validation (30,213 stays of 20,904 patients) and test (30,139 stays of 20,905 patients) splits (inclusion flow in Supplementary Table~S18). We removed from the test split all 40 stays of the 19 patients whose records had been opened in an earlier analysis. This left a test set of 30,099 stays of 20,886 patients. For the clinical-endpoint probes, a salted hash of the person identifier divided the training split further into a fitting set (119,330 stays of 82,946 patients) and a tuning set (21,143 stays of 14,607 patients). In the harmonized data, MIMIC-IV is split by person. The registered read used its training split (66,007 stays) and its test split (14,211 stays of 9,805 patients), and no MIMIC-IV validation data entered it.

Sex was taken as recorded in each database. It was a static covariate of the model and also defined descriptive subgroups. Race and ethnicity are not part of the harmonized records, and the model never used them. We took ethnicity from the eICU-CRD patient table, as recorded by the contributing hospitals, and used it only for descriptive subgroup analyses of the eICU-CRD test set. Each patient was assigned the most frequent entry across their stays. The entries were grouped as White (recorded as Caucasian), Black (recorded as African American), Hispanic, Asian, Native American and other or unknown (Supplementary Table~S7).

\subsection*{Model}

ICare-DT reads the hourly record of an intensive care stay: its 161 channels, their observation mask and the four static covariates (Fig.~\ref{fig:overview}b,c). An encoder maps each hour's physiology, treatments, mask and static covariates to a 64-dimensional embedding. It is a two-layer multilayer perceptron with sigmoid linear unit activations, and it adds learned random Fourier features of absolute time. A causally masked Transformer\cite{vaswani2017attention} then gives a 64-dimensional patient state $z_t$ that depends only on the record before the decision hour. The Transformer has two layers, four heads, feed-forward width 256, Gaussian error linear unit activations and dropout 0.1. Forecasts are made for future bins that end 1, 6, 12, 24 and 48 hours ahead. For each tracked target $k$ and bin $j$, a target-specific two-layer head outputs four logits. Their softmax gives the probabilities of no event, threshold crossing, live discharge from the unit and death.

The four tracked targets were heart rate, respiratory rate and the two oxygen-saturation channels of the harmonized data, \spo{} and \sao{}. In eICU-CRD, the \spo{} channel holds nurse-charted pulse oximetry, and the \sao{} channel combines the bedside monitor's periodic saturation readings with blood-gas values. In MIMIC-IV, the \spo{} channel holds hourly bedside pulse oximetry, and the \sao{} channel holds blood-gas values only. A threshold crossing is the first crossing of the standardized training mean ($z=0$) in a patient not already past it. It is upward for heart and respiratory rate and downward for the saturations. Such a crossing marks an excursion of the vital sign and does not by itself define a clinical event. For oxygen saturation, the clinically defined analogue is the hypoxemia endpoint (Clinical endpoints).

Forecasts start from origins. An origin for target $k$ is an hour, before the patient left the unit, in which the target was measured. The target must not have met the crossing criterion in that hour or in any earlier measured hour. Every forecast is therefore conditional on this status. From an origin, the bin in which a crossing is first detected is labeled crossing, and earlier bins are labeled no event. Without a crossing, a bin is labeled no event only if the target was measured in every hour of it. Live discharge and death are taken from the unit discharge time and status recorded in eICU-CRD. Either is labeled in its bin when it falls within 48 hours and within the retained record, provided that all earlier bins are complete. Otherwise follow-up is censored at the first hour in which the target was not measured or at the end of the record. A crossing detected in the hour of discharge or death is left unresolved, and bins after the last labeled bin are masked. The loss is the origin-weighted categorical negative log-likelihood of these labels, summed within an origin and averaged over origins. The cumulative incidence of crossing is $F_{k,J}=\sum_{j\le J}S_{k,j-1}p^{\mathrm{cross}}_{k,j}$ with $S_{k,j}=S_{k,j-1}p^{\mathrm{none}}_{k,j}$. Discharge and death therefore act as competing events. They rule out a later crossing and so remove probability mass from its cumulative incidence.

Models were trained on the person-disjoint eICU-CRD training split (140,473 stays) with seeds 13, 17, 23, 29 and 31. Training used AdamW with learning rate $4\times10^{-4}$, batch size 256, weight decay $10^{-4}$, gradient clipping at 1, at most 12 epochs and patience 3. Checkpoints were selected on validation negative log-likelihood balanced across origins and targets, with ties broken by the lower seed. Seed 29 (epoch 11) had the lowest validation loss and seed 23 the second lowest, and all five checkpoints were kept. No test data were used for training, early stopping, calibration or selection.

The twin has three operations (Fig.~\ref{fig:overview}d,e). Synchronization re-reads the patient's growing record with frozen weights to update the state. The readouts predict from the patient state: they are the joint competing-risks forecast (Forecast evaluation) and the clinical-endpoint probes (Clinical endpoints). The counterfactual rollout generates the next hours under a chosen action through a transition adapter. It was evaluated only on a historical benchmark (Historical rollout). The individual-effect analyses do not use the rollout. In these analyses, readers fitted with the true effects as labels estimate a lower bound on the individual-effect information in the pre-decision record (Washout panels and recoverability map). Standard learners and a separately built response model, ICare-RH, were tested on whether they recover the effects from factual data (CATE learners and oracle-structure references; ICare-RH and its placebo gate).

\subsection*{Forecast evaluation}

The competing-risks forecasts were scored on the eICU-CRD test set in a single registered read. Its analysis plan was committed before the read (Table~\ref{tab:gates}; Supplementary Table~S1 and Supplementary Note~6). It fixed the landmark, the 17 forecast settings (four vital signs at four horizons, and death), the metrics and the rule that every setting is reported. Each of the five checkpoints was applied with frozen weights at hour 48 of the stay, and nothing was fitted, recalibrated or selected on test data. For target $k$ and horizon $H\in\{6,12,24,48\}$ hours, the forecast is the cumulative incidence of crossing within $H$ hours of an origin at hour 48 ($F_{k,J}$ for the bin ending at $H$; Model). A stay entered a setting if hour 48 was an origin for the target and its outcome within $H$ hours was resolved. An outcome was resolved as positive when a crossing was detected. It was resolved as negative when the target was measured in every hour of the horizon without a crossing, or when the patient left the unit alive or died before any crossing. We excluded stays without a valid unit discharge record, stays censored by an unmeasured hour or by the end of the record, and stays with a crossing in the hour of discharge or death. This left 227 to 1,449 stays of 220 to 1,349 patients per setting, with 61 to 481 crossings (Supplementary Table~S19). The forecasts therefore describe patients who were still in the unit, measured and not past the threshold at hour 48, and who were then measured until the outcome was known. Without recalibration, the mean forecast was 91\% to 105\% of the observed rate at 6 and 12 hours and 68\% to 89\% at 24 and 48 hours (Supplementary Table~S19). In descriptive subgroups, the median AUROC was 0.69 to 0.77 across age, sex and unit type (Supplementary Table~S7). Hispanic patients were evaluable in only five settings, where their median was 0.58, against 0.71 and 0.70 for White and Black patients in those settings.

The model has no separate mortality head. Its death forecast, fixed in the plan as a descriptive summary, is the largest of the four target-specific death cumulative incidences at 48 hours. Each head's death class counts a death only if it precedes that head's target crossing. This summary therefore targets death in the unit before a crossing, whereas the outcome counts every death in the unit in its window. The death forecast was evaluated in the 11,498 stays of 9,506 patients still in the unit after hour 48 with a valid discharge record. Of these stays, 294 ended in death in the unit within the following 48 hours. This population includes stays whose target had already been measured beyond its threshold by hour 48, so some heads were read outside the hours on which they were trained. A descriptive decision-curve analysis of the crossing and death forecasts is in Supplementary Table~S8.

For each seed and setting we computed AUROC, the area under the precision-recall curve, log-loss, Brier score, the mean forecast, the calibration intercept and slope, and a ten-bin reliability table. The calibration intercept and slope come from a logistic regression of the outcome on the logit of the forecast. Five-seed means carry Student $t$ intervals with 4 degrees of freedom, which describe variation between training seeds only. Patient-level intervals came from 2,000 bootstrap replicates per seed and setting, each resampling patients with all their stays. A separate verification job reproduced all 85 seed-by-setting results exactly from the sealed prediction files (Supplementary Note~6).

The same frozen forecasts were scored on MIMIC-IV as disclosed secondary outcomes of the registered read (Registered MIMIC-IV read).
On the MIMIC-IV test split, the five checkpoints were applied without refitting at hour 48 to stays whose record reached hour 48 and whose unit stay lasted beyond it. The death forecast was the summary above. The forecast of leaving the unit within $H$ hours ($H=6$, 12, 24 or 48) applied the same summary to both exits. It was the largest, over the four target-specific heads, of the cumulative incidence of leaving the unit, alive or dead, before the head's target crossed its threshold. The death forecast therefore targets death in the unit before the head's crossing, and the exit forecast leaving the unit before the head's crossing. Each head was trained only at hours when its target was measured and had not yet crossed. Where a head's target had already been measured beyond its threshold by hour 48, that head was therefore read outside its training hours, as in the death summary on eICU-CRD. MIMIC-IV records the time of in-hospital death. The harmonized records used here, however, keep only a flag for death in the hospital, without its time, and no export of that time was added for this read. The death forecast was therefore scored against two outcomes. The first was death in a hospitalization whose unit stay ended within 48 hours of the landmark. It includes every death in the unit in that window, and also stays discharged alive from the unit in that window whose patient died later in the hospital. The second was death at any time in the hospitalization. Leaving the unit was scored from the recorded unit length of stay, rounded up to whole hours. Each head counts a death or a unit exit only if it precedes the head's crossing. The two death outcomes, in contrast, count every death in their windows, including deaths after a live unit exit, and the exit outcome counts every unit exit in its window. A mean forecast below the observed rate therefore arises in part by construction, as for the death forecast on eICU-CRD. The metrics were those above. Intervals are 95\% whole-person bootstrap intervals of the five-seed mean (10,000 replicates), and the Supplementary Information also gives Student $t$ intervals across seeds. These secondary outcomes were descriptive and had no registered criterion.
They were scored in the 6,951 stays of 5,396 patients still in the unit after hour 48. The death forecast had an AUROC of 0.687 (95\% whole-person bootstrap interval 0.657 to 0.718) for death in a hospitalization whose unit stay ended within the next 48 hours (324 stays), an outcome that includes every death in the unit in that window, and 0.726 (0.708 to 0.744) for death at any time in the hospitalization (972 stays). Its mean forecast, 0.0039, was below the observed rates of 0.047 and 0.14, in part by construction, because each target-specific head counts a death only if it precedes that target's crossing, and both outcomes also count deaths after a live unit exit. For the exit forecast, which targets leaving the unit before a vital-sign threshold crossing, the AUROC for leaving the unit within 48 hours was 0.765 (0.754 to 0.776; 0.765 to 0.808 for horizons of 6 to 48 hours), and its mean forecast, 0.27, was below the observed rate of 0.51, in part by construction, because each target-specific head counts a unit exit only if it precedes that target's crossing. Hospital death and length of stay had been scored on this split before, hospital death also by tabular models trained on MIMIC-IV, which reached AUROCs of 0.876 to 0.885 among stays of at least 48 hours with inputs on a corrected time grid (Supplementary Note~7). These are context only, and no comparison was made.

\subsection*{State synchronization and control inputs}

The synchronization analysis used the settings $(L,H)\in\{(12,6),(18,12),(30,24)\}$. The synchronized forecast (S2) read hours 1 to $L$ and predicted a crossing in $(L,L+H]$. The hour-6 forecast (S1) read hours 1 to 6 and predicted the same window as $[F_6(2H)-F_6(H)]/S_6(H)$. Here $F_6$ and $S_6$ are the cumulative incidence of crossing and the probability of no event of any class, read at hour 6. Because $L=6+H$, this forecast conditions on no event of any class by hour $L$. The roster of each setting contained the stays that reached hour 30 and had the target measured in every hour through the label window. Rosters held 592 to 6,648 stays per setting on the test set, and 42.9\% of test stays appeared in at least one setting. The estimate was the five-seed mean AUROC difference. The registered inference used 2,000 stay-level bootstrap replicates and Holm correction within the 12 settings\cite{holm1979simple} (Statistics and reproducibility). A re-run without the hour-30 condition (592 to 7,711 stays) gave the same result, with 12 of 12 settings surviving Holm correction and gains of 0.072 to 0.153 AUROC. Calibrated log-loss and Brier score used isotonic maps fitted on validation data. Their magnitudes depend on a probability clip, so we interpret only their direction.

The last-value comparator ranked stays by the last measured value of the target at the landmark, in the direction of the crossing. A disclosed post hoc read scored a second hour-6 forecast as an alternative comparator: the forecast $F_6(H)$ of a crossing within the $H$ hours after hour 6. This earlier-window forecast exceeded S1 by 0.021 and 0.050 AUROC in two saturation settings and by 0.002 in one other, none surviving Holm correction. The synchronized forecast exceeded the better of the two hour-6 forecasts in all 12 settings (point estimates 0.073 to 0.121 AUROC).

The control inputs were tested on the validation split. Each setting's roster kept the records that reached its own landmark, rather than hour 30, and whose label was resolved, with stays censored by an unmeasured hour dropped (551 to 7,707 stays per setting). We call the four forecasts compared there the information arms (Fig.~\ref{fig:sync}b; isolation arms in the protocols and the Supplementary Information). They are the synchronized forecast (S2), the hour-6 forecast (S1) and the forecasts from the two control inputs, labeled ``Held at hour 6'' and ``Other patient'' in Fig.~\ref{fig:sync}b. Both control inputs changed only hours 7 to $L$ of the record that the frozen model read. In the information-constant arm (IC), the input held at hour 6, these hours repeated the hour-6 values, and the observation mask was set to zero. In the donor arm (SH), the other-patient input, these hours were taken from a donor and spliced onto the patient's first six hours. The donor is the other person whose hour-6 state in the frozen model is closest to the patient's. Two sensitivity versions changed only the observation mask of the held input. IC-A repeated the hour-6 mask of the non-target variables and marked the target channels as unmeasured. IC-O did the same and also marked the forecast target as measured at hour $L$. A sensitivity version of the donor arm (SH-A) used the same donor. Wherever the donor's value was missing, it kept each variable that the patient had available at hour 6, at its hour-6 value. A splice-free comparison used the donor's own forecast.

Contrasts used 10,000 whole-person bootstrap replicates, shared across arms and seeds, and add-one $P$ values. Holm correction was applied over the 36 primary contrasts: S2 minus IC, IC minus S1 and S2 minus SH, each in 12 settings. The registered decision rules required at least 6 of 12 settings to survive correction, with no setting significantly negative before correction (unadjusted $P<0.05$). Donor contrasts were also tested under a two-way bootstrap that resampled the patients and their donors separately. Log-loss and Brier score on validation used isotonic maps cross-fitted over two person folds. With these maps, the synchronized forecast had a lower log-loss than S1, IC and SH in all 12 settings. In each of these three comparisons, the difference survived Holm correction in 11 of 12 settings. Supplementary Note~5 compares ten hour-6 and donor arms. They are S1, BR (S1 with a training-fitted intercept per setting), IC, IC-A, IC-O, SH and SH-A, and three donor-only scores: the donor's own forecast, last measured value and number of measured hours.

The synchronized forecast exceeded IC-A by 0.090 to 0.150 and IC-O by 0.089 to 0.144 AUROC in all 12 settings, each after Holm correction within its own 12 settings. It exceeded the donor's own forecast by 0.119 to 0.188 AUROC. Under the two-way bootstrap, its advantage over the donor arm survived Holm correction in all 12 settings. The donor arm itself did worse than the hour-6 forecast in all 12 settings, by 0.014 to 0.058 AUROC (point estimates outside the registered families). In \spo{} (18, 12), part of the advantage over the donor may reflect the measurement pattern rather than the measured values. In that setting, the number of measured hours also separated the patient's record from the donor's. A difference-in-differences contrast, prespecified as descriptive, was computed on the AUROC scale. It compared the advantage over the donor arm with the corresponding advantage of the patient's own last measured value over the donor's. Suppose that the new hours inform the forecast only through the target's last measured value (a current-value restriction). This contrast is then not zero in general, so zero is not its null value. The sign of its point estimates also depended on the scale on which it was computed (Proposition~2; Supplementary Note~5).

The multivariate gradient-boosted comparison ran under a prospective protocol (Table~\ref{tab:gates}). It fitted the fixed CatBoost recipe of the clinical-endpoint probes (Clinical endpoints) on the training split. The boosted model and the twin forecasts from the seed-29 checkpoint were scored on the same validation rows in the 12 settings. The protocol's Holm family held 120 contrasts, ten per setting: AUROC, log-loss and Brier score of the twin against this model and against a boosted model of the target channel's own history, the same three metrics for the multivariate against the target-history model, and the AUROC of the twin against the last measured value. A follow-up analysis on the same rows added the twin's state to the gradient-boosted model. This raised the AUROC point estimate in 10 of 12 settings (median 0.004), but no AUROC gain survived Holm correction over its 36 contrasts. Among the log-loss gains, only that for respiratory rate at (18, 12) survived (Holm-adjusted $P=0.043$).

The twin's weights were kept frozen during synchronization because, in an earlier model version, event-specific online adaptation of the forecasting heads had not improved forecasts (Supplementary Fig.~S1). That evaluation used 1,895 test patients and five events at four horizons, in a configuration with one isolated head per event and in a joint configuration. A win required both AUROC and cross-fitted calibrated log-loss to improve after Holm correction over 20 settings. No setting met this criterion, and one setting in each configuration was significantly worse after correction. Updates were attempted at all 37,253 scheduled update times of each configuration and seed, and the campaign's rollback rule reverted 31\% to 37\% of them. These results are a descriptive re-read of a completed campaign that used an earlier readout. Per-patient weight adaptation was set aside on this evidence before the analyses reported here.

\subsection*{Clinical endpoints}

Two registered development probes, P0 and P1 (Supplementary Fig.~S3), predicted four clinical endpoints from the patient state, each within 24 hours of landmarks at 12, 24 and 48 hours. The endpoints were death in the unit, hypotension, acute kidney injury and hypoxemia. Hypotension was a mean arterial pressure below 65 mmHg in two consecutive measured hours, using invasive pressure when it was measured in that hour and non-invasive pressure otherwise. Acute kidney injury was creatinine stage 1 or higher of the Kidney Disease: Improving Global Outcomes (KDIGO) criteria, computed from in-unit values. Stage 1 was met by a rise of at least 26.5\,$\mu$mol\,l$^{-1}$ over the lowest value of the previous 48 hours. It was also met by a value of at least 1.5 times the lowest value of the previous 7 days. A stay was at risk of acute kidney injury only with a creatinine value before the landmark and a first creatinine below 353.6\,$\mu$mol\,l$^{-1}$. Hypoxemia was \spo{} below 90\% in two consecutive measured hours. A landmark row was positive if the first detected onset fell in the 24-hour window and negative if the record covered every in-unit hour of the window. Rows with unknown status were excluded (133 to 1,936 per endpoint). The tuning set contained 555, 3,136, 1,762 and 587 events of the four endpoints, in the order above, among 40,050, 26,318, 27,402 and 37,401 landmark rows.

The comparator was CatBoost\cite{prokhorenkova2018catboost} with 800 trees, depth 6, learning rate 0.03 and random seed 13, fitted without tuning. Its 1,136 features summarize the record: seven statistics of each channel plus the static covariates. Probe P0 attached a logistic model or a multilayer perceptron to the frozen 64-dimensional state of seeds 29 and 23, the two checkpoints with the lowest validation loss. It also added the same state to the comparator's features. Probe P1 was designed after P0 had read the tuning labels. It trained the encoder on the four endpoints, starting from the pretrained weights or from random initialization. A learned landmark embedding was added to the state, and a two-layer head read the endpoints. P1 also added the fine-tuned state of the pretrained encoder to the comparator's features, and we call the resulting model the hybrid. Encoders were trained with AdamW at constant learning rates (weight decay $10^{-4}$, batch size 256 stays, gradient clipping at 1, at most 15 epochs). Training stopped early on a 10\% inner split of persons, with a patience of two epochs. Fine-tuning from the pretrained weights used learning rates of $10^{-4}$ for the encoder and $10^{-3}$ for the head, and training from random initialization used $5\times10^{-4}$ for all weights. The comparison between the two encoders therefore confounds initialization with learning rate. Each model trained in about 30 seconds on one graphics processing unit (GPU). The pretrained encoder had been trained on every tuning-set patient, which favors the frozen and fine-tuned states. AUROC contrasts used 2,000 whole-person bootstrap replicates over 14,122 patients shared by all arms. No family-wise correction was applied to these development probes, whose gates are given in Table~\ref{tab:gates}. The same recipes were refitted within MIMIC-IV for the registered read.

\subsection*{Registered MIMIC-IV read}

The two prediction results, synchronization and the endpoint probes, were read once on the person-disjoint test split of MIMIC-IV under a protocol committed and tagged before the read. The protocol fixed the analyses, the decision rules and the sentences that report each outcome, and every outcome is reported (Supplementary Note~7). The test manifest listed the stays that the read could use. It contained every MIMIC-IV stay of the test split with a harmonized time series (14,211 stays of 9,805 patients), with no filter on length of stay, context or outcome. It was built from identity columns alone and had to match the stay count, the permitted range of the patient count and the digests fixed in the protocol. This test split had been read before. The protocol discloses 34 earlier entries, 27 of them scientific reads (Supplementary Note~7 reproduces the list). These include zero-shot transfers of earlier versions of the eICU-CRD-trained twin, fine-tunes of the same pretrained model and outcomes that overlap ours: the blood-pressure rule that defines our hypotension endpoint, an earlier acute kidney injury definition, hypoxemia, heart rate above 130 per minute, \spo{} below 90\%, and future heart and respiratory rate. We knew their results when designing the read. The quantiles used to clip and fill the harmonized records had also been computed over all splits, including the test covariates but no labels.

\emph{Family A: synchronization with frozen weights.} The five eICU-CRD checkpoints were applied without refitting, and no weight, intercept, calibration map or threshold was fitted. Inputs, including the continuous dose channels, were standardized with MIMIC-IV training statistics, so the crossing thresholds were the MIMIC-IV training means. The domain embedding was switched off, and no channel was remapped. In particular, MIMIC-IV's hourly pulse oximetry was not copied into the arterial-saturation input, which in eICU-CRD carried the dense monitor saturation. Several inputs of the frozen model are recorded at different frequencies in MIMIC-IV. In the training data, arterial oxygen saturation is observed in 1.2\% of MIMIC-IV hours against 49.2\% of eICU-CRD hours. Heart rate and non-invasive mean arterial pressure are observed less often in MIMIC-IV (MIMIC-IV to eICU-CRD ratios 0.79 and 0.62), and most laboratory values more often (ratios 1.4 to 4.4). The \sao{} target was excluded before the read, because in MIMIC-IV that channel holds sparse blood-gas values and at most 10 training stays were evaluable in any setting. This left 9 settings, for heart rate, respiratory rate and \spo{} at $(L,H)=(12,6)$, $(18,12)$ and $(30,24)$. The MIMIC-IV \spo{} settings use hourly bedside pulse oximetry, unlike the nurse-charted pulse oximetry of the eICU-CRD \spo{} target. Each MIMIC-IV \spo{} setting can therefore be set beside either eICU-CRD saturation setting. Supplementary Note~7 shows both pairings, and neither enters a decision.

The arms (S2, S1 and the patient's last measured value), the labels and the hour-30 roster rule were those of the eICU-CRD analysis. The unit exit was the recorded unit length of stay rounded up to whole hours. Leaving the unit, alive or dead, counts as negative in the labels of families A and B, so these labels do not depend on the hospital-death flag. In both databases, the harmonized records can contain hours after the recorded unit exit. An endpoint-defining channel was measured after the recorded unit exit in 75.2\% of MIMIC-IV training stays, against 58.5\% of the eICU-CRD fitting set. Labels stop at the unit exit, but a stay can still enter a roster when rows recorded after the exit carry its record to hour 30.

The primary contrast (A1) was S2 minus S1, and the registered secondary contrast (A2) was S2 minus the last measured value. Each estimate was a five-seed mean AUROC difference. Family A resampled whole patients with 10,000 replicates, used two-sided add-one $P$ values and applied Holm correction over the 9 settings, separately for A1 and A2. The eICU-CRD analysis, by contrast, had resampled stays with 2,000 replicates, used $P$ values without the add-one rule and corrected over 12 settings. Three registered sensitivities were run on MIMIC-IV. One used the stay-level inference of the eICU-CRD analysis, one dropped the hour-30 roster condition and one set the thresholds at the eICU-CRD training means. A setting survived when its Holm-adjusted $P$ value was below 0.05 and its estimate was positive, with most replicates on the side of the estimate. It was significantly negative when its estimate was negative, its unadjusted $P$ value was below 0.05 and most replicates lay below zero. The registered transfer criterion was met when at least 5 of 9 settings survived and none was significantly negative before correction. The outcome was classed as contradicted when any setting was significantly negative after Holm correction, and A2 received the same categories. Inference conditions on the five frozen checkpoints. The control inputs and calibrated metrics were not computed on MIMIC-IV. The disclosed secondary outcomes used the same checkpoints at hour 48 (Forecast evaluation).

\emph{Family B: endpoint probes refitted within MIMIC-IV.} The arms of the endpoint probes were refitted within the MIMIC-IV training split (66,007 stays) with the recipes as run on eICU-CRD and with early stopping on 10\% of training patients. There were four arms: the boosting comparator (one fit, tree seed 13), two encoders and the hybrid. One encoder was trained from random initialization and the other fine-tuned from the pretrained twin, each with five seeds (13, 17, 23, 29 and 31). The hybrid combined the boosting features with the fine-tuned state. The endpoints were hypotension and acute kidney injury (primary) and hypoxemia (secondary), defined as on eICU-CRD and scored at landmarks of 12, 24 and 48 hours with a 24-hour horizon. Death in the unit was not scored because the harmonized records lack its time, and the training loss therefore pooled three endpoints. The hypotension endpoint is the hypotension rule that earlier reads had scored on this split, apart from its at-risk rule, clock and readout. Acute kidney injury and hypoxemia as defined here had not been scored there, although an earlier kidney-injury definition and an earlier hypoxemia definition had been. Checks of physical units and invalid values in the endpoint channels ran on the training and test splits before any family-B label was written. The primary analysis was a one-sided test of the AUROC difference between the hybrid and boosting, pooled over landmarks. It used add-one $P$ values from 10,000 whole-person bootstrap replicates over the 9,623 test patients whose record reached hour 12, with Holm correction over the two primary endpoints.
For the endpoint probes on MIMIC-IV, intervals are two-sided 95\% percentile intervals and the endpoint test is one-sided, so a gain significant at the second Holm step can have a lower interval bound at or below zero.
Hypoxemia carried an unadjusted one-sided $P$ value. The difference between the encoder trained from random initialization and boosting was reported with intervals, without a test or a non-inferiority margin. Two seed analyses were registered as sensitivities and enter no test.
When boosting was refitted under five random tree seeds, its AUROC spanned a range of 0.0008 for hypotension and 0.0004 for acute kidney injury. When each hybrid and boosting used the same tree seed (paired seeds), the hybrid's gain over boosting was $+0.0010$ ($-0.0012$ to $+0.0032$) for hypotension and $+0.0034$ ($+0.0006$ to $+0.0062$) for acute kidney injury.
For hypoxemia, a registered secondary endpoint, the hybrid's change was $+0.0009$ ($-0.0034$ to $+0.0054$; unadjusted one-sided $P=0.35$). The encoder trained from random initialization differed from boosting by $-0.0109$ ($-0.0145$ to $-0.0072$) for hypotension, $-0.0294$ ($-0.0367$ to $-0.0226$) for acute kidney injury and $-0.0197$ ($-0.0275$ to $-0.0118$) for hypoxemia; no non-inferiority margin was registered. The encoder fine-tuned from the pretrained twin was below the one trained from random initialization for hypotension and acute kidney injury, with intervals below zero, and differed from it by less than 0.0001 for hypoxemia (Supplementary Note~7).

\emph{Separation of predictions from labels.} The analysis code enforced the separation. The prediction stages saw the test records only through a view that refused every outcome field and showed hours 1 to 30 (family A) or 1 to 48 (family B and the secondary outcomes). Outside this view, the data loader still held the previously read hospital-death and length-of-stay columns in memory. The prediction processes also held later hours that fall inside label windows, and these hours were unused by construction. A fail-closed test, run on the real models with synthetic inputs, verified that every prediction issued at a landmark stayed bit-identical when all later hours were replaced at random. Predictions were frozen and hashed before any label was built, and the scorer opened the labels once. A verification stage recomputed every value with the same scoring code, which shows that the scoring is deterministic. A separate coding-agent pass with read-only access re-applied the decision rules to the shipped contrasts (Use of large language models and coding agents). Supplementary Note~7 gives the power analysis written before the read, the deviations listed in the protocol and those that arose after it.

\subsection*{Historical rollout}

The historical benchmark of earlier work on this model (Fig.~\ref{fig:benchmark}a) had five replicates, each drawn afresh from its data-generating model. Each replicate held 60 cells, one for each combination of outcome, treatment and horizon. Each cell held 19 to 30 treatment decisions, of which 1 to 10 were treated (Supplementary Note~1). Supplementary Fig.~S2 gives stress analyses of the benchmark. The action-conditioned rollout scored on it had a clock defect. It wrote each transition sample into the row of the action that conditioned it, although the transition adapter had been trained to predict physiology one hour after that action. Generated physiology therefore led its time index by at least one hour, and by more where the adapter carried persistence through generated physiology. In deterministic toy tests (Supplementary Note~3), the defect inflated the mean response over the benchmark's 12-hour window by factors of 1.00 to 1.38. Its effect on the historical readouts was not estimated. We report those readouts as run, and no positive finding of this paper rests on them.

\subsection*{Semi-synthetic generator and realism audit}

For each eICU-CRD stay, the generator takes the first 336 hourly rows of 12 outcome channels, standardized with the cohort normalization statistics (Fig.~\ref{fig:generation}b; implementation details and constants in Supplementary Note~4 and Supplementary Tables~S9 and S10). The channels are invasive mean arterial pressure, \spo{}, lactate, heart rate, respiratory rate, urine output, creatinine, pH, white cell count, platelets, bilirubin and temperature. A latent severity $u$ and a four-dimensional mediator $m$ evolve hourly,
\begin{equation}
u_{t+1}=0.85\,u_t+0.10\,\bar m_t+0.05\,\bar a_t+\xi^u_t,\qquad m_{t+1}=A_m m_t+b\,u_t+B^\top a_t+\xi^m_t,
\end{equation}
with noise standard deviation 0.25 and states clipped at $\pm8$. The states are driven by the indicators $a_t$ of recorded doses of five treatments: phenylephrine, NaCl, norepinephrine, dextrose and inspired oxygen fraction. The generator therefore responds to whether a dose was recorded and ignores its size. The recorded doses are written back at single precision. The generated outcome is $y_{tk}=y^{\mathrm{real}}_{tk}+0.6\,\delta_{tk}$, with
\begin{equation}
\delta_{tk}=0.02\,\big(w_k^\top[u_t,m_t]+\varepsilon_{tk}\big)+0.8\,\eta_i\,(W a_{t-2})_k,
\end{equation}
which combines a small latent background with a direct treatment route lagged by 2 hours. The sparse, sign-aligned weights $W$ are drawn once, the structural change $\delta$ is clipped at 6 standard deviations, and $y$ is clipped at 8 and to physiological bounds. The per-stay responsiveness $\eta_i$ is drawn from $1+N(0,0.5^2)$, independently of the record, and clipped to $[1/3,3]$. Individual responses therefore differ between stays, but the generated dataset carries no covariate-dependent effect heterogeneity. The generator's hidden confounder is simulated without effect ($\lambda=0$). For 674 test-split stays long enough for the 48-hour horizon, counterfactual outcomes were rolled out for 48 hours from the factual latent state at 20 random decision hours. Each treatment was set to 0 or 1 at the decision hour, later treatments were drawn from the generator's propensity policy, and no common random numbers were shared across arms. The counterfactual truth is the minimum and maximum of the recorded outcome over 6, 12, 24 and 48 hours. Generated with the two corrections described next, this dataset passed the locked realism audit. The individual-effect analyses do not use this dataset. They use the washout panels (Washout panels and recoverability map). These sustain the intervention for 48 hours after a washout, share random numbers across counterfactual arms and define each patient's effect against a placebo frame.

Two parameter-free corrections were added to the generator as options that are off by default. The measurement hold records a new value only in hours in which the standardized real value differs from the previous hour's value. In other hours it repeats the previous recorded value, as the forward-filled real record does. The hold applies to factual and counterfactual paths alike, so the counterfactual truth is the effect on the recorded value. The missingness correction restores the real missingness of the treatment channels and is permitted only when factual treatments replay the recorded doses. All other channels keep their real values and missingness.

The locked realism audit compares 400 matched patients on 161 channels with a person-grouped, five-fold ridge logistic discriminator. Its 644 features are per-patient means, standard deviations, missingness and lag-1 autocorrelations. The audit requires an AUC of at most 0.75 together with standardized mean difference, missingness and label checks. The passing dataset contains all 200,825 eICU-CRD stays, generated without error. The two corrections were designed after diagnosing earlier audit failures. The audit was read for each generator version shown in Fig.~\ref{fig:benchmark}b. The audit AUC bounds the distance between the real and generated summaries from below and never from above (Proposition~3). On the 3,000-patient probe with both corrections, a flexible classifier separated the real and generated summaries more than the audit did (out-of-fold AUC 0.735, against the audit's 0.641; Supplementary Note~5). This classifier was histogram gradient boosting in one configuration, chosen after the audit.
The same audit was applied to the three washout panels of the registered cell, under a protocol registered before it was run. It was run after all the panels' other analyses, except the in-support control and the g-computation comparator, whose registered runs followed it (Supplementary Note~2). Each audited record combined the panel's generated outcome values with the real record for every other channel. The generated values were set to missing wherever the real record is missing, and both were cut to hours 24 to 163 of the stay. Its treatment channels are therefore the real ones, while the outcome values respond to simulated dosing, a mismatch that the audit's per-channel summaries cannot detect (Proposition~3(b), Supplementary Note~5). The summary-difference checks cannot detect a difference confined to the 12 written channels, so the discriminator is the operative check. Every audit, control and interval is in Supplementary Note~9 and Supplementary Table~S29. The panels passed (audit AUC 0.714 to 0.727), but the audit did not flag measurement noise of 0.15 reference standard deviations added to the real records.
For comparison, the semi-synthetic dataset that passed the audit gave audit AUC 0.581 on the same windows of the discovery and fixed-schedule stays and 0.591 on those of the replication stays.

\subsection*{Washout panels and recoverability map}

Each washout panel contains 400 eICU-CRD windows (Fig.~\ref{fig:generation}c). The windows cover hours 24 to 163 of stays with at least 164 hourly rows, selected by a seeded permutation without regard to the eICU-CRD split. The real window supplies the 12 outcome channels, with missing values filled by fixed reference means. The panels use the generator's code with a separate frozen parameterization. Its background scale is 1, and its mediator persistence is 0.82 (spectral radius of the latent map 0.93, half-life 10 hours). It has no per-stay multiplier on the ordinary treatment routes and uses common random numbers across counterfactual arms. Outcomes are standardized to fixed reference values (85 and 15 beats per minute for heart rate).
The panels' generated outcome values passed the realism audit described above, which on their windows did not flag measurement noise of 0.15 reference standard deviations added to the real records.
The treatment columns are replaced by simulated dosing. In hours 6 to 53, one treatment follows a natural dosing pattern. It is taken from a factual rollout of the generator's propensity policy on the same window and topped up to at least eight dosed hours. Hours 54 to 89 are a washout in which this treatment is not given. From hour 90, it is held for 48 hours at 1 in one counterfactual arm and at 0 in the other. The other four treatments follow the propensity policy in both arms. The injected memory-route effect on outcome $k$ is
\begin{equation}
s_{tk}=c\,\tanh\Big(\sum_{j}L_{kj}\,\zeta_{ij}\,q_{tj}\Big),\qquad q_{tj}=\frac{e_{tj}}{1+e_{tj}},\qquad e_{tj}=0.8\,e_{t-1,j}+a_{t-2,j}.
\end{equation}
Here $c$ is the amplitude cap in reference standard deviations, and $L_{kj}$ is the normalized loading of treatment $j$ on outcome $k$ ($-1$ for phenylephrine on heart rate). The patient's trait for treatment $j$ is $\zeta_{ij}=\min\{\max\{0.6\,l_j\xi_i+0.15\,\epsilon_{ij},-2/3\},2\}$, with a shared standard normal factor $\xi_i$, treatment loadings $l_j$ drawn once from $U(0.7,1.3)$ and independent standard normal $\epsilon_{ij}$. The shift is applied after the ordinary clips, is attenuated near a bound and does not feed back into the latent state. The target $\tau_i$ starts from the do(1) minus do(0) difference in the responding outcome (heart rate in the registered cell), averaged over the 12 hours after the decision, in the active frame. From it we subtract the same average in a placebo frame with $c=0$ and all random draws shared. This subtraction cancels the ordinary treatment routes. The target uses this outcome at every hour, without the measurement hold.

Synthetic records, real windows observed every hour and real windows at their real cadence were all read with the same target. At real cadence, the generated record repeats its previous value in every hour in which the real value did not change. Its measurement times therefore come from the real record and do not depend on the injected effect. The discovery panel consists of the 400 stays used for the recoverability map. A fixed-schedule control on the same stays doses every patient in hours 6 to 24. The replication panel is a second sample of 400 stays, 395 of them new, so its pre-decision records and natural dosing differ from those of the discovery panel. Because it keeps the person-indexed generator draws, including the traits, its true effects are nearly identical to those of the discovery panel (correlation 0.9998). 

Readers are ridge regressions, fitted in five person-grouped cross-validation folds with the true per-patient effect as the label. They differ in the features they take from the pre-decision record. The history-blind reader uses the raw record. The occupancy-contrast reader uses the inner product of each channel's record with the patient's detrended occupancy profile, computed as $q$ in the injected effect but with the reader's lag and decay. The assimilating reader normalizes this inner product by the profile's energy and adds exposure summaries and channel means. Like the other readers, it is a ridge regression and does not use the twin or its synchronization. The cadence-aware reader sums, over consecutive measurement times of each channel, the change in the recorded value times the change in the occupancy profile. Hours without a new value therefore contribute nothing. The occupancy-based readers use the generator's exposure family. They choose its lag (0 to 3 hours) and decay (five values), together with the ridge penalty, by an inner four-fold cross-validation on the training patients. The grid contains the generator's values. The oracle reference uses each patient's true trait. Recovery is $1-\mathrm{RMSE}/\mathrm{SD}(\tau)$, where RMSE is the root mean squared error of the reader's prediction. Its intervals come from 2,000 whole-person bootstrap replicates. The figures show recovery, and the text reports the equivalent $R^2=1-(1-\mathrm{recovery})^2$. A reader clears the registered bar of $R^2=0.2$, which is $1-\sqrt{0.8}=0.106$ on the recovery scale, when its bootstrap probability of falling below the bar is under 0.05. It separates when it clears the bar and, if the history-blind reader also clears it, exceeds that reader with a bootstrap interval of the difference above zero. The second condition stops a reader from being credited with information that the history-blind reader already extracts from the raw record. Under a schedule shared by all patients, readers that apply one affine map to every record, such as the occupancy-contrast and assimilating readers, cannot exceed the history-blind reader in population $R^2$ (Proposition~5). The cadence-aware reader is not such a reader, because it still sees each patient's own measurement times. The proposition therefore does not explain why it did not separate on the fixed schedule.

The cadence-aware reader was also compared with six learned controls and with two constant predictions, zero and the training mean. This comparison followed a development protocol written before the run and was not registered (provenance in Supplementary Table~S1). The learned controls were ridge models fitted in the same way. Their features were the same actions together with summaries of the admission hours, of the six hours before the decision or of the full pre-decision record. Each comparison was made in the discovery, replication and fixed-schedule panels on two metrics, and Holm correction was applied over the resulting 48 contrasts. The first metric was the precision in estimation of heterogeneous effects (PEHE), the root mean squared error of the predicted individual effects. The second was centered PEHE (defined under CATE learners and oracle-structure references). After Holm correction, the cadence-aware reader won 32 of the 36 contrasts with the learned controls and all 12 with the two constant predictions. It lost none, and the four unresolved contrasts were all on the fixed schedule.

The registered amplitude, selected after an exploratory sweep (Registration and provenance), was 0.75 standard deviations for phenylephrine acting on heart rate. On the panels' reference scale of 15 beats per minute, it caps each patient's shift at 11.25 beats per minute. The median shift is 3.8 beats per minute at sustained steady-state exposure and 4.5 at full occupancy, which the kinetics never reach. The realized target has a standard deviation of 3.5 beats per minute. Shifts of this size lie in the range of the heart-rate differences between norepinephrine and phenylephrine in randomized comparisons during spinal anesthesia for cesarean delivery\cite{chen2022norepinephrine,ngankee2015randomized}. Those differences were measured in patients with intact baroreflexes. In septic shock, however, a randomized intensive care trial found no such difference\cite{morelli2008phenylephrine}.

\subsection*{CATE learners and oracle-structure references}

Static learners were S-, T-, X-, DR- and R-learners with histogram gradient boosting or ridge base models\cite{kunzel2019metalearners,kennedy2023towards,nie2021quasi} and causal forests with and without kernel-informed features\cite{wager2018estimation,athey2019generalized}. They also included TARNet\cite{shalit2017estimating}, DragonNet\cite{shi2019adapting} and random-slope distributed-lag models with and without the known kernel\cite{laird1982random}. Sequence models were the Causal Transformer\cite{melnychuk2022causal}, RMSN\cite{lim2018rmsn} and CRN\cite{bica2020crn}, run from their reference implementations. The g-computation sequence model was G-Net\cite{li2021gnet}, as implemented with the Causal Transformer code; it simulates every recorded variable forward under the queried treatments and averages 100 Monte Carlo trajectories with the same random draws in both arms. Besides the positive control, it had to recover a known average effect under time-varying confounding before it was scored. G-Transformer\cite{xiong2024gtransformer} was not run: no released code was found in the public sources searched on 8 October 2026 (Supplementary Note~9).
Hyperparameter grids and a synthetic positive control with an effect standard deviation near 1 were fixed in the protocols. A learner that failed its positive control was excluded. The primary metrics were centered PEHE and the correlation with the true effect. Centered PEHE is the standard deviation of the per-person error, and it equals $\mathrm{SD}(\tau)$ for a zero prediction. Each learner was compared with the oracle-structure learner on both metrics in three panels. The comparisons used 10,000 whole-person bootstrap replicates and Holm correction within each family (78 static, 18 sequence and 6 g-computation comparisons); statements that combine the three families carry a family-wise error bound of 0.15 (Supplementary Note~1). A comparison was declared a win or a loss only if it survived correction and was otherwise unresolved, which never implies equivalence. G-Net was also compared with predicting zero (Holm over 2 comparisons, in the discovery and replication panels); a gain counted only if it survived correction in both panels, the model ranked patients in both, and the gain was at least 2\% of the standard deviation of the true effects. The decision query is sustained dosing from hour 90, after a 36-hour washout in which no patient was dosed. No patient received the query regime, so its propensity is zero, and every estimator must extrapolate to it through assumptions (Proposition~6). The g-computation model and the control inside the support were added after the results at hour 90 were known, at the request of an internal review, under a protocol registered before they were run; the true effects at hour 90 had been read by earlier analyses (Supplementary Note~9), and the candidate hours (30 and 42) were set after label-free diagnostics of the Causal Transformer's predictions at hour 42 had been seen. A control inside the treatment support was registered at hours 30 and 42, each to be scored if it met overlap criteria on the recorded treatments (at least 20 patients per panel receiving each regime; at least half of the query patients had at least 10 treated and 10 untreated training windows within three hours with the same number of recent doses); neither met them.

The oracle-structure learner is a particle-filter response model (8,192 particles). It fits response weights and population-level nuisance parameters from factual data, but it knows the generator's trait prior, bounded response map and measurement convention. The structure-informed linear response model weights the same particles with a linear memory increment in place of the bounded map. It was scored as a descriptive reference. Label-free diagnostics were descriptive and could not gate or select. One of them, a person-grouped four-fold domain classifier, separated the query from the training units, the factual 12-hour treatment windows (AUROC above 0.999). At most three treated training units per fold fell in the query's recency cell (dosed earlier, no dose in the preceding 12 hours). 
An earlier label-free diagnostic at hour 42, which did not meet the overlap criteria, without true effects, had shown the Causal Transformer's predicted effects varying as little across patients as at hour 90 (per fold, standard deviation 0.03 to 0.09 at hour 42 and 0.03 to 0.09 at hour 90). 

\subsection*{ICare-RH and its placebo gate}

ICare-RH is a hierarchical random-slope response model. Its design drew on the structure of the oracle-structure learner and was fixed on synthetic data before the model was fitted to any real panel. The model describes each channel $k$ at its measurement times as $y_{ik}(t)=x_t^\top\beta_k+\gamma_k b_i\,\phi(e_i(t))+a_{ik}+o_{ik}(t)+\varepsilon_{ik}(t)$. Here $e_t=\lambda e_{t-1}+A_{t-\ell}$ is a learned exposure kernel with lag $\ell$, and $\phi(e)=e/(\kappa+e)$ is its saturation. The patient's response slope is $b_i\sim N(m(s_i),1)$, whose prior mean $m$ may depend on conditioning features $s_i$, and $\gamma_k$ is its loading on channel $k$. The term $a_{ik}$ is a random intercept, $o_{ik}$ a mean-reverting Ornstein--Uhlenbeck baseline evaluated at measurement times, and $x_t^\top\beta_k$ a 12-lag common route with a learned exponential tail. The slope is integrated out analytically, and the remaining parameters are fitted by type-II marginal likelihood, with no constant taken from the generator. The individual effect is predicted from the posterior mean of $b_i$ given the patient's own records before the query. In the registered arm the prior mean is a constant, so the arm uses no ICare-DT state.

ICare-RH passed all four of its synthetic gates. They were scored on 400 simulated patients in five folds, from a generator matched to the registered cell (true effect standard deviation 0.227, query at hour 90). Its correlation with the true effect was 0.786 against a required minimum of 0.75, and its centered PEHE was 0.141 against a maximum of 0.15. On a null scenario, its predicted effects had a spread of 0.021 against a maximum of 0.05. It recovered the exposure kernel in four of five folds, the minimum required. Fifteen variants inside and outside the model family were scored descriptively (Fig.~\ref{fig:rh}b). Outside its model family, ICare-RH beat predicting zero in seven of eight variants (correlation 0.64 to 0.84) and failed only when the response gain drifted (correlation 0.49; Supplementary Table~S5).

On real panels, the registered placebo gate applied label-free checks before any target was read. Its lag clause required a response lag within 1 to 3 hours in at least three of five folds of each natural-dosing panel. This clause also counted the folds in which a slope-variance test gave $P<0.05$, but the count did not gate. Before any real panel was read, the test had already rejected in 199 of 200 folds on 40 synthetic null panels with baseline drift or level shifts. This rate is consistent with unmodeled baseline variance along the exposure direction (Proposition~7). The gate's spread clauses concerned the zero-effect copies of every panel. On these copies, the spread of predicted effects had to be at most 0.05 standard deviations and at most 25\% of the spread on active records. A final clause was to be read only after these checks. It required the 95\% whole-person bootstrap interval of centered PEHE minus that of predicting zero to lie wholly below zero in both natural-dosing panels.

\subsection*{Theoretical analysis of the evaluation}

Seven propositions state what each evaluation design can and cannot identify. They were derived after the results, and no eICU-CRD analysis was chosen, registered or stopped because of them. Three of them largely restate standard results. Proposition~1 rests on the Neyman--Pearson lemma and the refinement of proper scores. Proposition~3 rests on bounds of an AUC by distances between distributions, and Proposition~6 on the decomposition of mean squared error by correlation and spread. Propositions~5 and 7 build on standard results of generalized least squares and of the profile likelihood of a one-way variance component. Supplementary Note~5 gives the statements, assumptions, proofs and numerical checks. It also names the standard part of each proposition with its source and states where the idealized models and the record disagree. Throughout, moments are population moments, and AUROC gives half credit to ties.

\emph{Proposition 1.} Let $\mathcal F_t$ be the record through hour $t$ and $p_{\mathcal G}$ the probability of the label on the roster given information $\mathcal G$ (the ideal forecast). By standard results, $p_{\mathcal G}$ has a receiver operating characteristic curve on or above that of every score computed from $\mathcal G$\cite{neyman1933problem,mcintosh2002combining}, and refining $\mathcal G$ cannot worsen its Brier score or log-loss\cite{degroot1983comparison,gneiting2007strictly}. For a fixed label, the ideal forecasts $p_{\mathcal F_t}$ form a martingale in $t$, the property that martingale-consistent self-supervised learning imposes on learned predictions\cite{gogl2026martingale}. Because $L=6+H$ and roster stays are at risk at hour $L$, the ideal forecast $S1^\ast$ given the hour-6 arm's input and conditioning event equals $p_{\mathcal F_6}$ and differs from the ideal synchronized forecast $S2^\ast$ only in information. Hours from a donor whose record is conditionally independent of the patient's label and later hours, given the first six hours, add no information. For AUROC and proper scores, the learned contrast S2 $-$ S1 equals the ideal gain minus the shortfall of S2 from $S2^\ast$ plus that of S1 from $S1^\ast$. It is therefore at most the ideal gain exactly when the hour-6 read falls short by no more than the synchronized read. The proposition uses only one property of the twin: its hour-$t$ state is a function of the record through hour $t$.

\emph{Proposition 2.} Let $V$ and $V'$ be the patient's and the donor's last measured target values at hour $L$. In a latent-threshold Gaussian model, where each AUROC maps to a latent correlation $\rho$, consider a forecaster that reads hours 7 to $L$ only through $V$, with index $U_6+\beta_VV$ ($U_6$ affine in the first six hours, $\beta_V>0$). Under exact matching, its latent difference-in-differences is $(w_V-1)(\rho_{\mathrm{own}}-\rho_{\mathrm{donor}})$, where $w_V:=\beta_V\mathrm{SD}(V)/\mathrm{SD}(U_6+\beta_VV)$ is unobserved. For the ideal forecaster, two inequalities on the four latent correlations hold in every such model. Quadruples that satisfy them are attainable under the current-state hypothesis $H_{\mathrm{CS}}$ (label independent of hours 7 to $L$ given the first six hours and $V$), and those that satisfy them strictly are attainable also without it. The four AUROCs therefore cannot reject $H_{\mathrm{CS}}$, or confirm it when both inequalities are strict. A violation would falsify the idealized model rather than $H_{\mathrm{CS}}$. Within that model, the label-free correlation of $V$ and $V'$ identifies $H_{\mathrm{CS}}$.

\emph{Proposition 3.} For a classifier two-sample test\cite{lopezpaz2017revisiting}, $|\mathrm{AUC}-\tfrac12|$ is at most $\mathrm{TV}-\mathrm{TV}^2/2$, and hence at most the total-variation distance TV between the real and generated laws of the 644 summaries. The first bound is sharp, so the audit bounds that distance from below and never from above. The population audit ($L_2$-penalized logistic regression with an unpenalized intercept, fixed standardization) has zero slope exactly when the mean summaries agree, and its AUC is then $\tfrac12$ whatever other differences exist. A mean difference makes the slope nonzero and, under an equal-covariance Gaussian idealization, makes the AUC exceed $\tfrac12$, although in general it need not.

\emph{Proposition 4.} With attenuation inactive, the washout target is $\tau_i=c\,T_i$ and the held record is exactly $Z^0_i+c\,U_i$, where the zero-effect record $Z^0_i$, the shift per unit amplitude $U_i$, $T_i$ and the design (doses and measurement times) do not depend on the amplitude $c$. For a reader affine in the record at fixed kinetics, under an orthogonality condition, the population $R^2$ gain over the reader's design-only features satisfies $1/\mathrm{gain}\ge C_0+C_1/c^2$, with equality when one statistic carries the signal. Here $1/C_0$ is the ceiling as $c\to\infty$, $C_1$ sets the gain at small amplitudes, and a change of sampling multiplies the amplitude needed for a small gain by $(C_1^{\mathrm{new}}/C_1^{\mathrm{old}})^{1/2}$\cite{mentre1997optimal}. For one responding channel in the linear regime with a random-walk baseline, the held record's Fisher information about the trait is proportional to $J:=\sum(\Delta q)^2/g$, summed over consecutive measurement times $d<d'$ with $\Delta q:=q(d')-q(d)$, $g:=d'-d$ and $q$ the generator's occupancy profile\cite{aitsahalia2003effects}. For hourly records the corresponding quantity is $Q^\circ:=\sum_t(q(t)-q(t-1))^2$, and the retention $\chi:=J/Q^\circ$ lies in $[0,1]$.

\emph{Proposition 5.} Among linear statistics of one channel that are invariant to a nuisance mean, the generalized least-squares weight is efficient\cite{aitken1936least}. Under white noise with a free quadratic trend, it gives, on hourly records, the detrended level contrast of the occupancy-contrast reader\cite{frisch1933partial}. Under a random-walk baseline, the gap-weighted sum $\sum\Delta Z\,\Delta q/g$ is efficient among level-invariant linear statistics. The implemented cadence-aware statistic $\sum\Delta Z\,\Delta q$ omits the gap weights. It is therefore efficient on hourly records, where every gap is one hour, but in general not on held records. Under a schedule shared by all patients, the assimilating and occupancy-contrast readers apply one affine map to every record and cannot exceed the history-blind reader in population $R^2$.

\emph{Proposition 6.} For a predictor $\hat\tau$ with correlation $r$ with $\tau$, the standard decomposition\cite{murphy1988skill}
\[
\mathrm{cPEHE}^2/\mathrm{Var}(\tau)=(1-r^2)+\{\mathrm{SD}(\hat\tau)/\mathrm{SD}(\tau)-r\}^2
\]
implies that $\hat\tau$ has lower centered PEHE than predicting zero exactly when $0<\mathrm{SD}(\hat\tau)<2r\,\mathrm{SD}(\tau)$. No function of data $X$ has squared correlation with $\tau$ above $R^2_\ast(X):=\mathrm{Var}\,\mathbb E[\tau\mid X]/\mathrm{Var}(\tau)$, which the conditional mean attains\cite{vegetabile2021distinction}. A supervised reader's held-out $R^2$ therefore estimates a lower bound on this information content and never bounds estimators from above. No dose occurs at or after hour 54, so the query regime has propensity zero, and the factual law fixes neither the conditional mean effect nor any predictor's accuracy; every estimator must extrapolate through assumptions.
Suppose that the trait that sets the pre-decision shift also sets the query effect, that the design and zero-effect record are independent of the trait, and that the response structure, including the query map, and the zero-effect law are known. Then the conditional mean effect is a known function of the pre-decision record. In large samples and with equal exposure information, an empirical-Bayes random slope\cite{robinson1991blup} whose working model understates the baseline variance along the exposure direction has a spread ratio above its correlation.

\emph{Proposition 7.} For ICare-RH at a fixed kernel, loading direction and baseline (Ornstein--Uhlenbeck working covariance at the measurement times\cite{jones1991unequally}), let $\mathcal A_i$ and $\mathcal B_i$ be the exposure information and the exposure projection of the baseline residual, $\hat\mu:=\sum_i\mathcal B_i/\sum_i\mathcal A_i$, and $\omega:=\sum_i(\mathcal B_i-\hat\mu\mathcal A_i)^2/\sum_i\mathcal A_i$ the overdispersion along the exposure direction. With the prior mean profiled out, the negative marginal log-likelihood of the slope variance $v:=\sum_k\gamma_k^2$ ($b_i$ has unit prior variance) is $P_n(v)=P_n(0)+\tfrac v2(\sum_i\mathcal A_i)(1-\omega)+O(v^2)$, as for a one-way variance component\cite{harville1977maximum}. For $\omega\ne1$, a small positive $v$ therefore improves the fit exactly when $\omega>1$. With equal information and no mean response along the exposure direction, the exact profile likelihood prefers in large samples, of two kernels, the one whose population $\omega$ is larger and exceeds 1. When $\omega$ stays above 1, the fixed-baseline likelihood ratio for $v$ diverges instead of following its 50:50 $\bar\chi^2$ reference\cite{self1987asymptotic}. A decay of 0.99, the parameter's upper bound, keeps at least 43\% of the end-of-dosing exposure at hour 89, so in the random-walk limit of the working covariance the exposure profile is not orthogonal to linear drift. If the baseline is correct along the exposure direction and a globally optimized fit captures a response of the form the model assumes, then under regularity conditions the two spread clauses of the placebo gate pass with probability tending to one. The lag clause and the comparison with predicting zero are not covered.

\emph{Idealizations.} The propositions use simplified models that do not describe real records. They assume the linear regime of the bounded map, one responding channel, baselines whose jump variance grows with the gap and measurement times independent of the baseline, whereas in the panels the baseline and the measurement times both come from the real record. Proposition 1 sets aside that the checkpoints were selected on the validation split, and Proposition 2 assumes exact matching. Proposition 3 fixes the standardization that the audit refits within folds. Propositions 4 and 5 analyze population least squares at fixed kinetics, whereas the readers use cross-validated ridge regression over a kernel grid. The retention values use simulated draw patterns, because the panels' draw masks were not shipped. Propositions 6 and 7 assume equal exposure information. Proposition 7 describes exact minimizers at a fixed baseline, whereas the implementation uses local optimizers and refits the baseline.

\subsection*{Statistics and reproducibility}

Unless stated otherwise, intervals are two-sided 95\% percentile bootstrap intervals\cite{efron1993introduction}. The seed-level intervals in Fig.~\ref{fig:sync}a and the replicate-level intervals in Supplementary Fig.~S2 are $t$ intervals. Replicate draws were shared across the arms of each analysis, so contrasts are paired. Resampling was over whole patients, except for the registered eICU-CRD test-set synchronization contrast and the disclosed last-value comparison, which resampled stays (2,000 replicates). The synchronization rosters hold only 1.02 to 1.12 stays per patient. A person-clustered bootstrap of the registered synchronization family, run afterwards as a sensitivity analysis, therefore changed little. It gave identical point estimates, widened the intervals by a median of 0.4\% and at most 2.5\%, and left all 12 settings significant after Holm correction.

Two-sided $P$ values use the add-one rule where stated. Where the observed statistic never crossed zero in 2,000 replicates, we report $P<0.001$ rather than zero, and $P<0.012$ after Holm correction over 12 settings. This was the case in every setting of the registered synchronization contrast on the eICU-CRD test set. Multiplicity was controlled by the Holm procedure within families fixed in the protocols\cite{holm1979simple}, and no correction was applied across families. No family-wise correction was applied to the registered development probes of clinical endpoints on eICU-CRD, so their intervals are unadjusted. For the benchmark families, a contrast is a win or a loss only when it survives correction with its interval wholly on one side of zero, and it is otherwise unresolved. Every learned treatment-effect estimator was also compared with a constant prediction of zero effect. A win over a comparator that did not itself beat zero is labeled as such. Subgroup and decision-curve analyses of the joint forecasts were descriptive. They excluded subgroups with fewer than 50 evaluable rows or a single class, and their intervals resample patients. The inference of the MIMIC-IV read is described above (Registered MIMIC-IV read).

\subsection*{Registration and provenance}

Each registered analysis followed a protocol that fixed the analysis and any decision rule and was committed and pushed to version control before the run. Some protocols also fixed the sentences that report their outcomes. Where this paper rewords such a sentence for readability, every number, outcome and caveat is unchanged, and the registered wording is in the protocols' result notes. Such registrations in version control are not external preregistrations\cite{nosek2018preregistration,hofman2023preregistration}. The protocols of the endpoint probes, the placebo gate, the MIMIC-IV read, the realism audit of the washout panels and the in-support control with the g-computation comparator also stated known biases or limitations and a forecast of the outcome. Those of the placebo gate, the MIMIC-IV read, the realism audit of the washout panels and the in-support control with the g-computation comparator were also tagged in version control. Automatic checks in the job-submission scripts and analysis stages backed these protocols. The submission scripts refused to submit a job from code with uncommitted changes or from a commit not yet pushed to the remote repository. They also refused any job whose registry entry did not match or whose pinned code or inputs had changed. The fitting stages could not open evaluation labels. A scoring stage opened the labels after the predictions had been sealed (frozen and hashed) and left a marker that blocks a second read by the same analysis. Each receipt, the record that a run writes, carries hashes of the run's code, inputs and outputs. The receipts of the endpoint probes and the placebo gate also carry the SHA-256 hash of their own canonical content. Table~\ref{tab:gates} lists the principal analyses with their status and outcome, including those whose decision rules were not met. Supplementary Table~S1 gives the provenance of each reported analysis.

Several analyses reported here reused splits that had been read before, or involved choices made after an earlier read. Much of the eICU-CRD test set had been scored by models trained with an earlier objective (Supplementary Note~6). The roster rule of the registered synchronization contrast was set after an earlier read of that test set, and the last-value comparison was run after the synchronization result. The probability clip of the calibrated log-loss was changed after the test read, so we interpret that metric by its direction only. The control inputs were tested on a validation split that had been read for checkpoint selection, calibration maps and an earlier reader. The endpoint probes P0 and P1 shared a tuning set, and P1 was designed after P0 had read its labels. The registered cell was chosen after an exploratory amplitude sweep on the 400 discovery stays, and its discovery run reproduces the exploratory run on those stays. The new evidence for this cell therefore lies in the replication panel, which carries nearly identical injected effects, and in the fixed-schedule control. ICare-RH's design was fixed before it was fitted to a real panel, but the panels had already been read by the reader and learner analyses. The in-support control and the g-computation comparator were registered after the results at hour 90 were known, and the control's candidate hours were set after a label-free diagnostic at hour 42 had been seen. The MIMIC-IV test split had been read before the registered read (Registered MIMIC-IV read). The numerical checks of the theory, run after the results, read stored eICU-CRD test-set predictions and labels. Supplementary Note~2 lists these reads and choices for eICU-CRD, Supplementary Note~7 those for MIMIC-IV and Supplementary Note~9 those for the in-support control and the g-computation comparator.

Earlier reads of evaluation splits that we could reconstruct are disclosed in a test-read registry, which states its coverage limits. The registry holds 46 plan rows, two of them for the MIMIC-IV read, two for the realism audit of the washout panels and one for the in-support control and the g-computation comparator, and 62 disclosure rows. Each plan row records a test-read plan or an analysis run under one, and each disclosure row records an earlier read of an evaluation split. A search ledger of 2,586 rows records the campaigns, runs and configurations of the project up to 26 September 2026. Of its 606 campaign and run rows, 280 were registered before their outcome and 189 were exploratory. A compute ledger records at least 17,683 central processing unit thread-hours and 362 GPU-hours (NVIDIA L40S) between 1 August and 26 September 2026. Both ledgers closed before eight analyses were run, so these analyses are reported here but not recorded in the ledgers (Supplementary Note~2). They are the endpoint probes, the control inputs, ICare-RH (its synthetic gates, 15 variants and placebo gate), the post hoc checks of Supplementary Note~5, the MIMIC-IV read, the realism audit of the washout panels, the in-support control and the g-computation comparator.

\subsection*{Software and compute}

Analyses used Python 3.10, PyTorch 2.5.1, NumPy 1.26.4, pandas 2.3.3, SciPy 1.15.3, scikit-learn 1.7.2, CatBoost 1.2.10, LightGBM 4.6.0 and XGBoost 3.2.0. Causal forests used R 4.3.3 with grf 2.6.1, and sequence baselines used their reference implementations under PyTorch 2.10.0 and PyTorch Lightning 1.4.5. Jobs ran on one shared node with an AMD EPYC 9555P processor and three NVIDIA L40S GPUs under Slurm.

\subsection*{Data availability}

eICU-CRD, MIMIC-IV and MIMIC-III are available to credentialed users through PhysioNet under their data use agreements. The data use agreements do not allow the authors to redistribute patient-level derived data, which credentialed users can regenerate with the preparation and split scripts that we will release (Code availability). Source data are provided with this paper and contain the aggregate tables behind every figure and behind Supplementary Tables~S2--S8 and S20--S27, including all aggregate outputs of the registered MIMIC-IV read. Tables~\ref{tab:cohorts} and \ref{tab:gates} and the other supplementary tables take their values from the reports and aggregate tables named in their captions, in Supplementary Table~S1 and in the Supplementary Notes.

\subsection*{Code availability}

The data-preparation and split scripts, the model, generator, washout-panel, benchmark, scoring and figure code, the registered protocols and receipts, the test-read registry, the search and compute ledgers and the aggregate source tables will be released in a public repository of sanitized files that contains no patient-level data. They are available to editors and referees on request during review. 

\subsection*{Use of large language models and coding agents}

Coding agents (Claude Code, Anthropic) were used to assist with the experimental implementation and writing polishing. The propositions of Supplementary Note~5 were derived after the results. Y.L. developed them, and coding agents assisted with their derivations, proofs, and numerical checks. Every registered gate followed a protocol committed before its run. Every registered gate result was recomputed from its shipped tables by a separate agent pass with its own context. For the MIMIC-IV read, that pass used the same model family as the agent that ran the read, so it was independent in context only. Scientific decisions were made by the authors, who take full responsibility for the content. No language model is an author, and no model output was used as data.

\section*{Ethics}

This study used de-identified research databases that are available to credentialed researchers. The institutional review board of Beth Israel Deaconess Medical Center approved the creation and sharing of MIMIC-IV with a waiver of informed consent. The institutional review boards of Beth Israel Deaconess Medical Center and the Massachusetts Institute of Technology approved those of MIMIC-III, also with a waiver of informed consent. eICU-CRD was released after certification of de-identification under the safe-harbor standard of the US Health Insurance Portability and Accountability Act, and its use is exempt from institutional review\cite{pollard2018eicu}. All analyses complied with the credentialing and data use requirements of each database.

\section*{Acknowledgements}

This study was partially funded by the JST Moonshot R\&D Program (grant JPMJMS2034-18). The funder had no role in the design, analysis or writing. We thank the MIT Laboratory for Computational Physiology for making the MIMIC and eICU databases available and the contributors of the BlendedICU project for a standardized data pipeline.

\section*{Author contributions}

Y.L. conceived the study and designed the methodology. Y.L. and Y.Q. implemented the models, conducted experiments and analyzed the results. Y.L. developed the theoretical analysis, with coding agents assisting in derivations, proofs and numerical checks (Methods). Y.K. contributed to data analysis, visualization design and manuscript revision. D.P.-G. provided critical suggestions on the methodology and the interpretation of results. Y.L. wrote the manuscript, with drafting assistance from coding agents described in Methods. All authors contributed to the manuscript revision and approved the final version.

\section*{Competing interests}

The authors declare no competing interests.

{\small
\bibliographystyle{naturemag-doi}
\bibliography{references}
}

\clearpage
\section*{Tables}
\makeatletter\setlength{\@fptop}{0pt}\makeatother

\begin{table}[h]
\centering
\small
\caption{\textbf{Cohorts.} Stays and patients in the three harmonized databases, in the person-disjoint eICU-CRD splits and in the semi-synthetic data built on eICU-CRD records. All analyses use eICU-CRD except the registered MIMIC-IV read of the two prediction results and its disclosed secondary outcomes. Earlier transfer analyses on MIMIC-III were not registered and are disclosed in Supplementary Note~8. ``Read before'' marks splits that earlier analyses had already read (Supplementary Notes~2 and~7). In the realism audit, a discriminator tries to tell generated from real records. Its area under the receiver operating characteristic curve, the audit AUC, is 0.5 when it cannot tell them apart and must be at most 0.75 for the audit to pass (Methods).}
\label{tab:cohorts}
\begin{tabularx}{\textwidth}{>{\raggedright\arraybackslash}X r r >{\raggedright\arraybackslash}p{7.0cm}}
\toprule
Cohort or split & Stays & Patients & Role in this study \\
\midrule
eICU-CRD (208 hospitals) & 200,825 & 139,362 & Model training and all analyses except the MIMIC-IV read \\
\quad Training split & 140,473 & 97,553 & Model training; clinical-endpoint probes (Supplementary Fig.~S3) \\
\quad\quad Fitting set & 119,330 & 82,946 & Fitting of the endpoint probes \\
\quad\quad Tuning set & 21,143 & 14,607 & Scoring of the endpoint probes \\
\quad Validation split & 30,213 & 20,904 & Checkpoint selection; control inputs and gradient-boosted comparison (read before) \\
\quad Test set & 30,099 & 20,886 & Hour-48 forecasts, synchronization and last-value comparison (read before) \\
Semi-synthetic dataset & 200,825 & 139,362 & Generator applied to every eICU-CRD stay; passed the locked realism audit (audit AUC 0.572); counterfactual truth for 674 test-split stays \\
Washout panels & 400 per panel & -- & Individual-effect analyses on real eICU-CRD windows with simulated dosing and injected responses (discovery, replication and fixed-schedule panels) \\
MIMIC-IV & 94,458 & 65,366 & Registered read of the prediction results, with hospital death and leaving the unit as disclosed secondary outcomes (person-disjoint test split of 14,211 stays; read before, including earlier zero-shot transfers and overlapping outcomes) \\
MIMIC-III CareVue & 28,394 & 23,692 & CareVue-era stays of patients not in MIMIC-IV; used only in earlier transfer analyses, disclosed and not registered (Supplementary Note~8) \\
\midrule
Experimental cohort & 323,677 & -- & Three harmonized databases \\
\bottomrule
\end{tabularx}
\end{table}

\begin{table}[h]
\centering
\small
\caption{\textbf{Analyses and outcomes.} The principal analyses with their status, data, decision rule and outcome, including those whose decision rule was not met. The status in parentheses says whether and how each analysis was fixed before its run. Registered analyses followed a protocol committed to version control before their run (Methods). Prospective-protocol analyses followed a development protocol written before their run but were not registered. Disclosed, exploratory and post hoc analyses were not registered. The realism audit of the generated dataset applied a locked threshold. Supplementary Table~S1 gives the provenance of each analysis. Earlier reads of the same data and post hoc choices are listed in Supplementary Note~2 for eICU-CRD, in Supplementary Note~7 for MIMIC-IV and in Supplementary Note~9 for the in-support control and the g-computation comparator. ``Holm over $n$'' denotes Holm correction within a family of $n$ tests. Differences without a stated unit are in AUROC. Table~2 continues in three further parts. AUROC, area under the receiver operating characteristic curve.}
\label{tab:gates}
\begin{tabularx}{\textwidth}{>{\raggedright\arraybackslash}p{2.7cm} >{\raggedright\arraybackslash}p{2.0cm} >{\raggedright\arraybackslash}X >{\raggedright\arraybackslash}p{4.7cm}}
\toprule
Analysis (status) & Data & Decision rule & Outcome \\
\midrule
\multicolumn{4}{l}{\textit{Forecasting}} \\
\addlinespace[2pt]
Hour-48 forecasts (registered; death forecast descriptive) & eICU-CRD test set, hour 48 & All 17 settings and five seeds reported; nothing fitted on test data & AUROC 0.646 to 0.820 (threshold crossings, 16 settings), 0.798 (death in the unit within 48 h); mean death forecast far below the observed rate, in part by construction \\
MIMIC-IV, death and unit-exit forecasts (disclosed secondaries) & MIMIC-IV test set, hour 48 & Frozen eICU-CRD weights; hospital death and leaving the unit; all 6 settings reported; no criterion & AUROC 0.687 (hospital death, unit exit within 48 h), 0.726 (hospital death), 0.765 (leaving the unit within 48 h); calibration in the text \\
Synchronization (registered) & eICU-CRD test set & Synchronized vs hour-6 forecast in stays with the target measured every hour through the label window (592 to 6,648 stays per setting), Holm over 12 settings & 12/12 settings, +0.073 to +0.153 AUROC \\
Last-value comparison (disclosed) & eICU-CRD test set & Synchronized vs the patient's last measured value on the same stays, Holm over 12 settings & 11/12 settings, median +0.044 AUROC (range +0.010 to +0.085) \\
Gradient-boosted comparator (prospective protocol) & eICU-CRD validation split & Twin forecasting heads (seed-29 checkpoint) vs multivariate CatBoost on the same rows, Holm over 120 contrasts; CatBoost with the twin state added vs CatBoost alone, Holm over 36 contrasts & Heads: median $-$0.012 AUROC, worse in 3/12 settings after correction; state added: median +0.004 AUROC, no AUROC gain after correction \\
Control inputs (registered) & eICU-CRD validation split & Synchronized vs input held at hour 6 (assimilation) and vs other-patient input (necessary condition for personalization), Holm over 36 primary contrasts; each condition requires $\ge$6/12 settings after correction and none worse at unadjusted $P<0.05$ & Both conditions met in 12/12 settings (+0.080 to +0.146 and +0.111 to +0.185 AUROC); held input below the hour-6 forecast in 12/12 (significant in 5); information beyond the patient's last value not shown \\
MIMIC-IV, synchronization (registered) & MIMIC-IV test set & Synchronized vs hour-6 forecast, frozen eICU-CRD weights, Holm over 9 settings; transfer if at least 5 of 9 with none significantly worse before correction & Met: 8/9 settings, +0.025 to +0.114 \\
MIMIC-IV, last-value comparison (registered secondary) & MIMIC-IV test set & Synchronized vs patient's last value, Holm over 9 & Not met: worse in 2/9 settings after correction, better in 1/9 \\
\bottomrule
\end{tabularx}
\end{table}

\begin{table}[h]
\ContinuedFloat
\centering
\small
\caption{\textbf{Analyses and outcomes (continued).} Boosting is a fixed CatBoost recipe on 1,136 summary features of the record, and the hybrid adds the fine-tuned twin state to these features. Differences from boosting are in AUROC. On eICU-CRD, the endpoint probes (Supplementary Fig.~S3) were scored on a tuning set that the pretrained encoder had seen in training, and their intervals are unadjusted. AUROC, area under the receiver operating characteristic curve.}
\begin{tabularx}{\textwidth}{>{\raggedright\arraybackslash}p{2.7cm} >{\raggedright\arraybackslash}p{2.0cm} >{\raggedright\arraybackslash}X >{\raggedright\arraybackslash}p{4.7cm}}
\toprule
Analysis (status) & Data & Decision rule & Outcome \\
\midrule
\multicolumn{4}{l}{\textit{Clinical-endpoint probes}} \\
\addlinespace[2pt]
Endpoint probe P0, frozen state (registered) & eICU-CRD tuning set & Frozen state $\ge$ boosting on $\ge$2/4 endpoints, or within 0.02 of it on $\ge$3/4, or a gain with interval above 0 from adding the state to boosting on $\ge$2/4 & Not met: state 0.020 to 0.086 below boosting; gain from adding it to boosting $\le$ +0.0014, no interval above 0 \\
Endpoint probe P1, encoders trained on the endpoints (registered) & eICU-CRD tuning set & Better encoder within 0.01 of boosting on $\ge$3/4 endpoints, or hybrid gain $\ge$ +0.005 with interval above 0 on $\ge$2/4 & Not met (near threshold): hybrid +0.0042 (hypotension) and +0.0075 (acute kidney injury), intervals above 0; better encoder (from random initialization) 0.004 to 0.021 below boosting \\
MIMIC-IV, endpoint probes (registered) & MIMIC-IV test set & Hybrid vs boosting on hypotension and acute kidney injury, refitted within MIMIC-IV, one-sided, Holm over 2 & Acute kidney injury only: +0.0035; hypotension +0.0010 \\
\bottomrule
\end{tabularx}
\end{table}

\begin{table}[h]
\ContinuedFloat
\centering
\small
\caption{\textbf{Analyses and outcomes (continued).} Readers are ridge regressions fitted with the true per-patient effect as the label; they are not deployable. AUC, area under the receiver operating characteristic curve; SD, standard deviation.}
\begin{tabularx}{\textwidth}{>{\raggedright\arraybackslash}p{2.7cm} >{\raggedright\arraybackslash}p{2.2cm} >{\raggedright\arraybackslash}X >{\raggedright\arraybackslash}p{5.4cm}}
\toprule
Analysis (status) & Data & Decision rule & Outcome \\
\midrule
\multicolumn{4}{l}{\textit{Semi-synthetic data and individual treatment effects}} \\
\addlinespace[2pt]
Realism audit (locked) & Semi-synthetic dataset & Discriminator AUC $\le$ 0.75 and all summary checks & Pass: AUC 0.572 \\
Washout-panel realism audit (registered, after all but two of the panels' other analyses) & Three washout panels & Locked audit per panel: AUC $\le$ 0.75 and all summary checks & Pass in 3/3: AUC 0.714 to 0.727, but the audit did not flag added measurement noise of 0.15 reference SD (weak evidence); other sensitivity controls not detected or not run \\
Recoverability map (exploratory) & 400 eICU-CRD stays, three treatment-response pairs & A reader clears the bar ($R^2=0.2$) and, if the history-blind reader also clears it, beats that reader & At real cadence, of the mapped readers only the cadence-aware reader cleared the bar, and so separated: on heart rate (value changes in 57\% of hours) for both drugs at amplitudes of 0.75 and 1 SD ($R^2=0.44$ at 1 SD). At real cadence, none cleared it on creatinine (value changes in 4\% of hours) \\
Registered cell, cadence-aware reader (registered after exploration) & Three washout panels & Clears the bar ($R^2=0.2$) and, if the history-blind reader also clears it, beats that reader & Separated in the discovery panel ($R^2=0.33$) and in a replication panel of 400 stays, 395 of them new, carrying nearly identical injected effects ($R^2=0.31$); did not separate on the fixed schedule ($R^2=0.24$, interval reaching below the bar) \\
\bottomrule
\end{tabularx}
\end{table}

\begin{table}[h]
\ContinuedFloat
\centering
\small
\caption{\textbf{Analyses and outcomes (continued).} The oracle-structure learner uses factual data only but is given the generator's response structure; it is not deployable. A learner beats zero in a panel when the 95\% bootstrap interval of its centered PEHE minus that of predicting zero lies wholly below zero. Comparisons that do not survive Holm correction are counted as unresolved. CATE, conditional average treatment effect; CRN, counterfactual recurrent network; PEHE, precision in estimation of heterogeneous effects; RMSN, recurrent marginal structural network.}
\begin{tabularx}{\textwidth}{>{\raggedright\arraybackslash}p{2.7cm} >{\raggedright\arraybackslash}p{2.2cm} >{\raggedright\arraybackslash}X >{\raggedright\arraybackslash}p{5.4cm}}
\toprule
Analysis (status) & Data & Decision rule & Outcome \\
\midrule
\multicolumn{4}{l}{\textit{Semi-synthetic data and individual treatment effects (continued)}} \\
\addlinespace[2pt]
Static CATE learners (registered) & Registered cell (decision query outside the support) & 13 learners vs the oracle-structure learner, Holm over 78 comparisons & Oracle-structure learner: 77 wins, 0 losses, 1 unresolved; no learner beats zero \\
Sequence CATE models (registered) & Registered cell (decision query outside the support) & Causal Transformer, RMSN and CRN vs the oracle-structure learner, Holm over 18 comparisons & Oracle-structure learner: 12 wins, 0 losses, 6 unresolved (the comparisons with the excluded CRN); no model beats zero \\
G-computation model (registered) & Registered cell & Holm over 6 comparisons vs oracle-structure learner; Holm over 2 vs predicting zero & Oracle-structure learner: 6 wins, 0 losses, 0 unresolved; G-Net (Causal Transformer code) does not beat zero \\
In-support query control (registered) & Two panels, hours 30 and 42 & Each learner vs predicting zero, Holm over 64; vs oracle-structure learner, Holm over 128 & No hour met the overlap criteria \\
ICare-RH placebo gate (registered) & Three washout panels and their zero-effect copies & Label-free lag and placebo-spread rules, checked before any target was read & Not met at the label-free stage; no target read, so ICare-RH was not scored on real records \\
Theory checks (post hoc, descriptive) & Stored tables of previously read splits; simulation & No criterion; run after the results & Reported in Supplementary Note~5, including a flexible classifier on the realism probe and an hour-6 read of eICU-CRD test-set predictions and labels \\
\bottomrule
\end{tabularx}
\end{table}

\end{document}


\maketitle
\tableofcontents
\clearpage

\section*{Supplementary Figures}
\addcontentsline{toc}{section}{Supplementary Figures}

\begin{figure}[h]
\centering
\includegraphics[width=\textwidth]{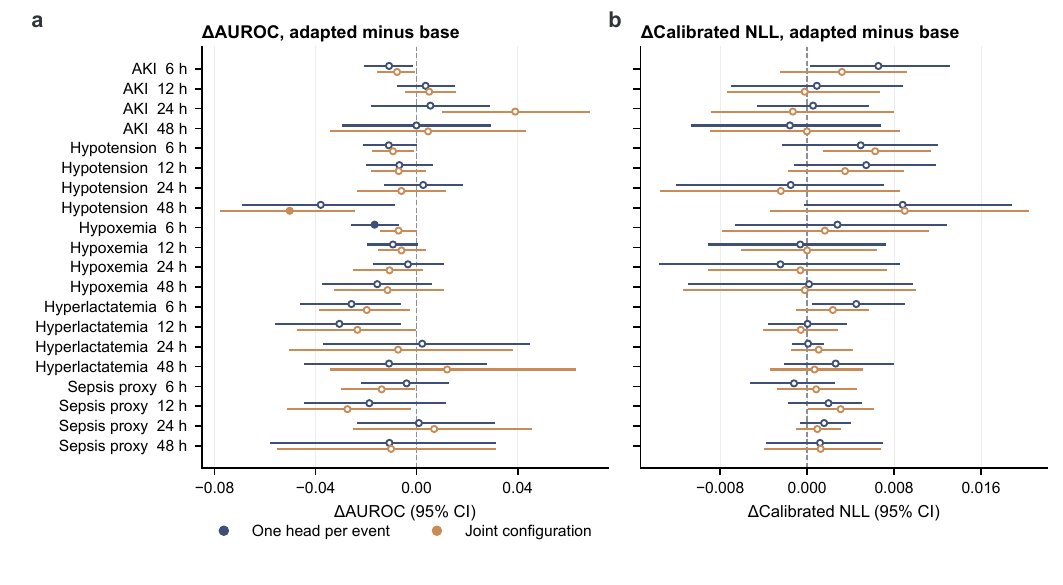}
\caption{\textbf{Per-patient online adaptation of the forecasting heads did not improve on frozen weights in an earlier model version.} Each marker shows the adapted model minus the base (frozen) model for one of five events (acute kidney injury, hypotension, hypoxemia, hyperlactatemia and a sepsis proxy) at a horizon of 6, 12, 24 or 48 hours on the eICU-CRD test split (1,895 patients; three seeds). The heads were adapted either as one isolated head per event or in a joint configuration. \textbf{a}, AUROC difference. \textbf{b}, Cross-fitted calibrated log-loss difference. A win required both metrics to improve after Holm correction over 20 settings, and no setting met this criterion. One setting in each configuration was significantly worse after correction, hypoxemia at 6 hours with isolated heads (AUROC $-0.017$) and hypotension at 48 hours in the joint configuration (AUROC $-0.050$). Filled markers survive correction. These descriptive re-reads of a completed campaign, which used an earlier readout, motivated the frozen-weight design of the twin.}
\end{figure}

\begin{figure}[h]
\centering
\includegraphics[width=\textwidth]{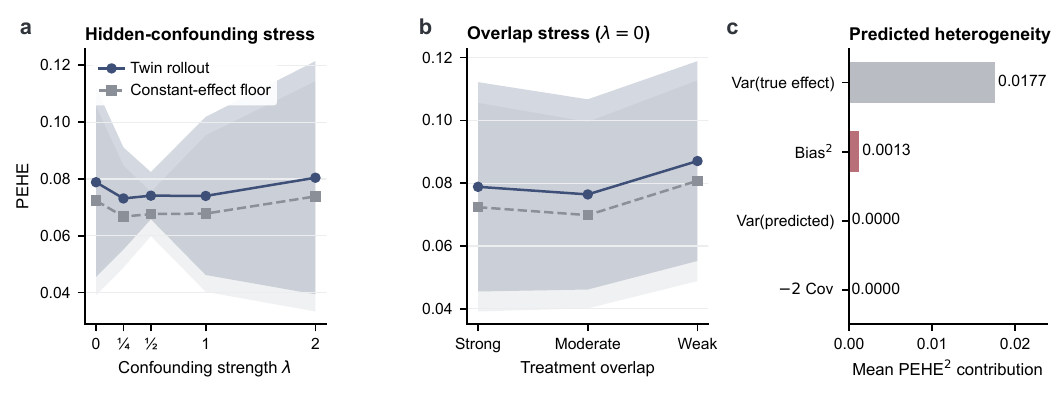}
\caption{\textbf{Stress analyses of the historical rollout benchmark.} \textbf{a}, Individual-effect error (PEHE) of the twin rollout and of a constant-effect floor, which assigns the true average effect to every patient, by strength of hidden confounding ($\lambda=0$, 0.25, 0.5, 1 and 2) at strong overlap. \textbf{b}, The same at $\lambda=0$ for strong, moderate and weak overlap (treatment-policy logit scale $\gamma=0.5$, 2 and 4). In \textbf{a} and \textbf{b}, values are means over five data-generating replicates, each averaged over three model seeds, with 95\% $t$ intervals across replicates. The rollout's mean error was above that of the floor at every confounding strength and overlap. \textbf{c}, Decomposition $\mathrm{PEHE}^2=\mathrm{Var}(\hat\tau)+\mathrm{Var}(\tau)-2\,\mathrm{Cov}(\hat\tau,\tau)+\mathrm{bias}^2$ of the historical model at $\lambda=0$, averaged over the 60 outcome, treatment and horizon cells, five replicates and three model seeds. The model predicted almost no heterogeneity. The variance of its predicted individual effects was about 0.15\% of that of the true effects, and their covariance with the true effects was near zero. The rollout ran with the clock defect described in Supplementary Note~3 and is reported as run.}
\end{figure}

\begin{figure}[htbp]
\centering
\includegraphics[width=\textwidth]{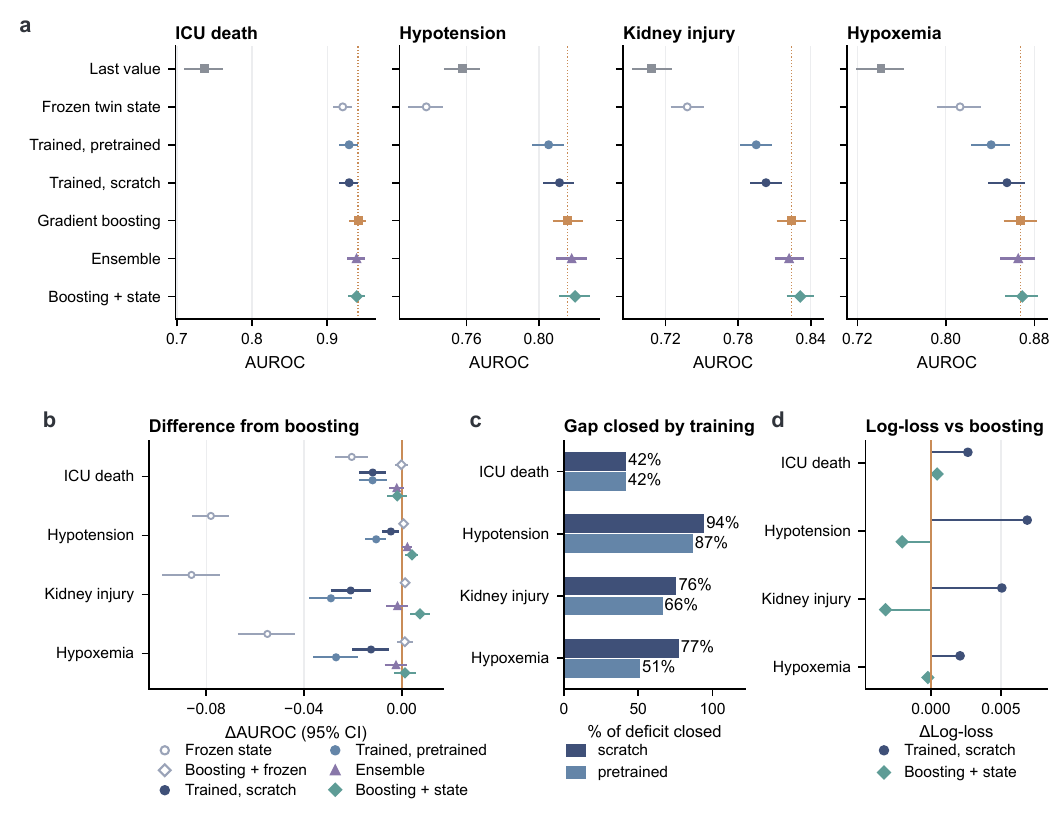}
\caption{\textbf{A fine-tuned twin state approaches gradient boosting and adds complementary signal on eICU-CRD.} Registered development probes P0 and P1 on the tuning set of the eICU-CRD training split (Supplementary Note~6). The pretrained encoder had seen every stay of the tuning set's 14,607 patients, and the 14,122 patients with at least one landmark row were scored. \textbf{a}, AUROC with 95\% whole-person bootstrap intervals for death in the unit, hypotension (mean arterial pressure below 65 mmHg in two consecutive measured hours), acute kidney injury (KDIGO creatinine criterion) and hypoxemia (\spo{} below 90\% in two consecutive measured hours) within 24 hours of landmarks at 12, 24 and 48 hours. The reference arms are logistic regression on the last measured value of the endpoint variable and the boosting comparator, a fixed CatBoost recipe on 1,136 summary features of the record (dotted line). The frozen twin state is read by a multilayer perceptron, and the encoder trained on the endpoints is either fine-tuned from the pretrained twin (pretrained) or trained from random initialization (scratch). Boosting and the fine-tuned encoder are also combined, in a mean-logit ensemble and in the hybrid (Boosting + state), which adds the fine-tuned state of the pretrained encoder to the boosting features. \textbf{b}, Differences from gradient boosting, with hollow markers for the frozen state. \textbf{c}, Share of the frozen-state deficit (frozen state minus boosting) closed by training the encoder on the endpoints. \textbf{d}, Log-loss difference from gradient boosting for the encoder trained from random initialization and for the hybrid (point estimates; negative is better). Two model seeds were averaged. Intervals are unadjusted. Neither probe met its registered gate, and P1 missed it narrowly (Table~2 of the main text). The registered MIMIC-IV read of this comparison is reported in the main text and Supplementary Note~7.}
\label{fig:endpoints}
\end{figure}

\clearpage
\section*{Supplementary Tables}
\addcontentsline{toc}{section}{Supplementary Tables}

\begin{table}[h]
\centering
\footnotesize
\renewcommand{\arraystretch}{0.98}
\caption{\textbf{Provenance of the reported analyses.} Each row traces an analysis to its protocol and result reports and to the scheduler jobs whose receipts record the run. Statuses, given in parentheses, are registered (protocol committed to version control before the run), prospective protocol (development protocol written before the run and not registered), disclosed (run and reported without registration), exploratory, descriptive or locked audit. Protocol and result reports are in \texttt{reports/revision\_v3/}, and aggregate tables are in \texttt{source\_tables/revision\_v3/}.}
\begin{tabularx}{\textwidth}{>{\raggedright\arraybackslash}p{3.6cm} >{\raggedright\arraybackslash}X >{\raggedright\arraybackslash}p{3.3cm}}
\toprule
Analysis & Protocol and result & Jobs \\
\midrule
Joint model training & Corrected competing-event model, five seeds; validation-only checkpoint selection & (before protocol registration began); selection 5255 \\
Joint forecasts, test set (registered) & \texttt{PHASE1\_TEST\_ANALYSIS\_PLAN} (locked before the read) and all-results receipt; Supplementary Note~6 & 5276, 5293, 5312 (verification) \\
Subgroups and decision curves, test set (descriptive) & \texttt{JOINT\_FACTUAL\_STRATIFIED\_20260918}; ethnicity re-read; Supplementary Tables~S7 and S8 & 10545; 10697 \\
Further single-seed reads under the same plan (registered) & Earlier synchronization contrast (superseded) and frozen-state probe versus LightGBM; Supplementary Note~6 & 5264, 5275; 5265 \\
Synchronization, test set (registered) & State-synchronization v2 lock and release review; re-run without the hour-30 condition & 10707, 10708; 10755, 10937 \\
Last-value comparison (disclosed) & Landmark last-value baseline & 10754 \\
Isolation arms, validation (registered) & \texttt{STATE\_SYNC\_ISOLATION\_PROTOCOL\_20260927} / \texttt{RESULT\_20260928} & 13006, 13007 \\
Gradient-boosted comparator and the twin state added to it, vital signs, validation (prospective protocol) & \texttt{LANDMARK\_TRAJECTORY\_DEVELOPMENT\_PROTOCOL\_20260920} and \texttt{LANDMARK\_STATE\_FUSION\_PROTOCOL\_20260920} & 11179--11181; 11208--11210 \\
Online adaptation of the forecasting heads (descriptive re-read) & Rescore of the completed targeted-head campaign; Supplementary Fig.~S1 & 2956--2973 \\
Endpoint probe P0 (registered) & \texttt{ROUTE\_B\_P0\_PROBE\_PROTOCOL\_20260927} / \texttt{RESULT\_20260928}; Supplementary Note~6 & 12997, 13001--13003 \\
Endpoint probe P1 (registered) & \texttt{ROUTE\_B\_P1\_PROBE\_PROTOCOL\_20260928} / \texttt{RESULT\_20260929}; Supplementary Note~6 & 13029--13032 \\
Realism audit (locked audit) & Hold probe and realism gate reports & 10568--10581 \\
Washout-panel realism audit (registered) & \texttt{WASHOUT\_PANEL\_REALISM\_AUDIT\_PROTOCOL\_20261008} / \texttt{RESULT\_20261014}; Supplementary Table~S29 & 13464--13469 \\
Recoverability map (exploratory) & Twin identifiability map; amplitude calibration & 10615--10689 \\
Registered washout cell (registered after exploration) & Registered phenylephrine cell & 10690--10694 \\
Matched learned comparators (prospective protocol) & Washout matched comparators & 11227--11230 \\
Static CATE learners (registered) & \texttt{WASHOUT\_STANDARD\_CATE\_PROTOCOL\_20260923} and amendment & 12770--12774 \\
Sequence CATE models (registered) & \texttt{WASHOUT\_SEQUENCE\_CATE\_PROTOCOL\_20260924} and amendments & 12939--12947 \\
G-computation sequence model G-Net (registered) & \texttt{WASHOUT\_Q5Q6\_PROTOCOL\_20261008} / \texttt{RESULT\_20261022}; Supplementary Note~9 & 13479--13486, 13541--13546, 13772--13781 \\
In-support query control (registered) & \texttt{WASHOUT\_Q5Q6\_PROTOCOL\_20261008} / \texttt{RESULT\_20261022}; Supplementary Table~S30 & 13479--13486, 13541--13546, 13772--13781 \\
Oracle-structure learner F (prospective protocol) & Washout factual-response result & 11287, 11288 \\
ICare-RH synthetic gates and variants (synthetic data, before any real-panel fit) & \texttt{WASHOUT\_RESPONSE\_HYBRID\_IMPLEMENTATION\_20260927}; Supplementary Table~S5 & (local runs; no cluster job) \\
ICare-RH placebo gate G7 (registered) & \texttt{WASHOUT\_RESPONSE\_HYBRID\_PROBE\_PROTOCOL\_20260929} / G7 result & 13044 (failed before any read), 13046 \\
Historical benchmark re-emission (disclosure test) & \texttt{A3\_FOUR\_ARM\_REEMISSION\_20260926} & (recomputed from shipped tables) \\
Theory checks (post hoc, descriptive) & Supplementary Note~5; scripts and outputs in \texttt{reports/revision\_v3/theory\_checks/} & (local re-reads and simulation) \\
Hour-6 read of the test panel (post hoc, disclosed) & Supplementary Note~5, Section 5.1; per-stay predictions and labels of the synchronization runs & 10707, 10708, 10755, 10937 (re-read) \\
MIMIC-IV read: synchronization, endpoint probes and disclosed secondary outcomes (registered) & \texttt{MIMIC\_IV\_READ\_PROTOCOL\_20260930} / \texttt{RESULT\_20261002} and release review; Supplementary Note~7 & 13123 (manifest); 13163--13173 \\
Earlier MIMIC-III reads (disclosed) & Test-read registry rows TRB-M4-01, -05 to -16 and -18 to -25; Supplementary Note~8 & As in the registry \\
\bottomrule
\end{tabularx}
\end{table}

\begin{table}[h]
\centering
\small
\caption{\textbf{Recoverability map.} Recovery of the injected individual memory-route effect (12-hour target) by readers fitted with the true effect, for three treatment-response pairs on the same 400 real eICU-CRD records (Fig.~5a--c of the main text). On hourly records, the occupancy-contrast reader cleared the bar in all three pairs at 0.5 SD and above, whereas at real cadence only the cadence-aware reader cleared it, on heart rate at 0.75 and 1 SD. Recovery is $1-\mathrm{RMSE}/\mathrm{SD}(\tau)$. The main text reports $R^2=1-(1-\mathrm{recovery})^2$, and the registered bar of 0.106 corresponds to $R^2=0.2$. Size: amplitude cap in reference standard deviations (SD). Occupancy: occupancy-contrast reader. Hourly: every hour observed; real: real measurement cadence. Oracle: oracle reference that knows each patient's trait. Dashes: not run. $^{*}$Clears the bar (bootstrap probability of falling below it under 0.05). The table has no hourly column for the cadence-aware reader. On hourly records it recovered 0.453, 0.392 and 0.342 at 1 SD for phenylephrine, norepinephrine and NaCl (jobs 10679, 10683 and 10675), and 0.369 for phenylephrine at 0.75 SD (job 10691). For phenylephrine at 1 SD, its hourly fits selected lag 2 and decay 0.95 in all five folds, and its fits at real cadence selected lag 2 and decay 0.8 (job 10678).}
\begin{adjustbox}{max width=\textwidth}\begin{tabular}{l r r r r r r}
\toprule
Cell & Size (SD) & Occupancy, hourly & History-blind, real & Occupancy, real & Cadence-aware, real & Oracle \\
\midrule
\rows{supp_tables/s2_map.tex}
\bottomrule
\end{tabular}\end{adjustbox}
\end{table}

\begin{table}[h]
\centering
\small
\caption{\textbf{Registered cell across the five registered runs.} The registered cell is phenylephrine acting on heart rate at 0.75 SD with the 12-hour memory-route target, scored on 400 stays per run with 2,000 whole-person bootstrap replicates (Fig.~5d of the main text). The oracle reference cleared the bar in every run, and every reader cleared it on synthetic and hourly records. At real cadence only the cadence-aware reader did so, in both the discovery and the replication panel, and no reader did on the fixed schedule. The replication run used a second sample of 400 stays, 395 of them new, carrying nearly identical injected effects (Supplementary Note~4). Recovered: $1-\mathrm{RMSE}/\mathrm{SD}(\tau)$. The main text reports $R^2=1-(1-\mathrm{recovery})^2$, and the bar of 0.106 corresponds to $R^2=0.2$. Gap: difference from the history-blind reader. Clears bar: whether the bootstrap probability of falling below 0.106 was under 0.05. Readers are defined in Methods, and the assimilating reader, despite its name, does not use the twin. Dosing-window means are channel means over four windows set by the first and last dose (the window-mean reader of Supplementary Note~5, Definition~6).}
\begin{adjustbox}{max width=\textwidth}\begin{tabular}{l l l l l}
\toprule
Run & Reader & Recovered [95\% CI] & Gap over history-blind [95\% CI] & Clears bar \\
\midrule
\rows{supp_tables/s3_registered.tex}
\bottomrule
\end{tabular}\end{adjustbox}
\end{table}

\begin{table}[h]
\centering
\small
\caption{\textbf{Every treatment-effect estimator on the registered cell.} Centered PEHE (lower is better; predicting zero equals the SD of the true effect) and correlation with the true individual effect, by panel.}
\begin{adjustbox}{max width=\textwidth}\begin{tabular}{l r r r r r r}
\toprule
& \multicolumn{2}{c}{Discovery} & \multicolumn{2}{c}{Replication} & \multicolumn{2}{c}{Fixed schedule} \\
Estimator & cPEHE & $r$ & cPEHE & $r$ & cPEHE & $r$ \\
\midrule
\rows{supp_tables/s4_learners.tex}
\bottomrule
\end{tabular}\end{adjustbox}
\end{table}

\begin{table}[h]
\centering
\small
\caption{\textbf{ICare-RH recovers individual effects in simulation.} ICare-RH was fitted to factual data only, with no constant taken from the generator, on 400 simulated patients in five folds per scenario. Its registered arm, shown here, uses no ICare-DT state. The registered scenario is matched to the registered cell, and the other 15 scenarios, inside and outside the model family, were scored descriptively (Fig.~7b of the main text). ICare-RH beat predicting zero in every scenario except a drifting response gain, where the interval of the difference included zero. In the registered scenario, its per-patient 80\% and 95\% intervals covered 0.77 and 0.94 of the true effects. SD of truth: standard deviation of the true effects. Difference: centered PEHE minus that of predicting zero, with 95\% bootstrap interval.}
\begin{adjustbox}{max width=\textwidth}\begin{tabular}{l l r r r l}
\toprule
Scenario & Family & SD of truth & Correlation & cPEHE & Difference from zero \\
\midrule
\rows{supp_tables/s5_icare_rh.tex}
\bottomrule
\end{tabular}\end{adjustbox}
\end{table}

\begin{table}[h]
\centering
\small
\caption{\textbf{Synchronization contrasts by setting (AUROC differences).} The synchronized forecast outperformed the model's hour-6 forecast, the information-constant arm and the donor arm in all 12 settings, and the patient's last measured value in 11 (Fig.~2c,d of the main text). The four ``vs'' columns give its AUROC gain over each comparator. On the test set, the comparators are the model's hour-6 forecast (registered) and the patient's last measured value (disclosed). On the validation split, the registered comparators are the information-constant arm and the donor arm, which replaces the patient's hours from hour 7 to the landmark with those of the nearest other patient in the frozen hour-6 state. The last column, prespecified as descriptive, is the difference-in-differences between the advantage over the donor arm and the advantage of the patient's own last measured value over the donor's, on the validation split. It was significantly positive in no setting and significantly negative in two, so this contrast did not show information beyond the patient's last measured value. $n$: test-set stays. $^{*}$Significant after Holm correction, over the 12 settings for the two test-set columns and the difference-in-differences, and over the 36 primary contrasts of the isolation arms for the information-constant and donor columns.}
\begin{adjustbox}{max width=\textwidth}\begin{tabular}{l r r r r r r}
\toprule
Setting & $n$ & vs hour-6 forecast & vs last value & vs information constant & vs donor & Beyond last value \\
\midrule
\rows{supp_tables/s6_sync.tex}
\bottomrule
\end{tabular}\end{adjustbox}
\end{table}

\begin{table}[h]
\centering
\small
\caption{\textbf{Discrimination of the joint forecasts by subgroup on the eICU-CRD test set.} Descriptive analysis (job 10545). Median AUROCs over settings ranged from 0.690 to 0.770 across the age, sex and unit-type subgroups. The settings are the crossings of four vital signs at four horizons and the 48-hour death forecast, all issued at hour 48 of the stay (Supplementary Note~6), and each setting's AUROC in a subgroup is the median over five seeds. The columns give the number of settings with at least 50 evaluable rows and both classes, the median over these settings and their range. The last column gives the AUROC of the death forecast for death in the unit within 48 hours, for seed 13, with a 95\% person-cluster bootstrap interval. The lowest single setting, with few stays, was 0.460 (\spo{} at 6 hours, 65 to 74 years, 64 stays). Discrimination of death was higher below 65 years (0.857 to 0.890) than at 65 years and older (0.771 to 0.803). Race, ethnicity and insurance are absent from the harmonized records, and the model did not use ethnicity. A descriptive re-read that joined ethnicity from the source eICU-CRD patient table (job 10697) gave median AUROCs over settings of 0.709 for White (17 settings), 0.712 for Black (17 settings) and 0.583 for Hispanic patients. Hispanic patients were evaluable in only five settings (SaO\textsubscript{2} at four horizons and the death forecast; range 0.487 to 0.736), in which the medians were 0.709 for White and 0.700 for Black patients. Death AUROCs for seed 13 were 0.812 [0.782, 0.841] for White, 0.833 [0.745, 0.907] for Black and 0.761 [0.473, 0.974] for Hispanic patients. The other groups had too few evaluable stays for most settings, and eICU-CRD does not record insurance.}
\begin{adjustbox}{max width=\textwidth}\begin{tabular}{l l r r l l}
\toprule
Stratum & Value & Settings & Median AUROC & Range over settings & ICU death, 48 h \\
\midrule
\rows{supp_tables/s7_subgroups.tex}
\bottomrule
\end{tabular}\end{adjustbox}
\end{table}

\begin{table}[h]
\centering
\small
\caption{\textbf{Decision curves of the joint forecasts on the eICU-CRD test set.} Descriptive decision-curve analysis (job 10545) at the 50 alert thresholds $t$ from 0.01 to 0.50, with net benefit $\mathrm{TP}/N-\mathrm{FP}/N\cdot t/(1-t)$ taken as the median over five seeds. The table gives the lowest and highest thresholds at which the forecast's net benefit exceeded both that of alerting on every stay and that of alerting on none. Count is the number of such thresholds, which need not be contiguous. Up to 24 hours ahead, the crossing forecasts of heart rate, respiratory rate and SaO\textsubscript{2} had net benefit at most thresholds. For crossings at long horizons, where prevalence is high, alerting on every stay is nearly optimal, and fewer thresholds show net benefit. The mean death forecast was far below the observed rate (0.0023 against 0.026), in part by construction, because each target-specific head counts a death only if it precedes that target's crossing. At thresholds from 0.03 to 0.15, the death forecast flagged at most 0.72 stays per 100 in any seed, so its net benefit in this range rests on very few alerts (Supplementary Note~6).}
\begin{adjustbox}{max width=\textwidth}\begin{tabular}{l r r l l}
\toprule
Forecast & Horizon (h) & Prevalence & Thresholds with net benefit & Count \\
\midrule
\rows{supp_tables/s8_net_benefit.tex}
\bottomrule
\end{tabular}\end{adjustbox}
\end{table}

\begin{table}[h]
\centering
\small
\caption{\textbf{Direct-route weights $W$ of the generator.} Each entry is the change of the standardized outcome (row) per treated hour of a treatment (column), before the route scale of 0.8, the per-stay responsiveness $\eta_i$ and the multiplier of 0.6 are applied. The weights are sparse, with 28 of the 60 entries at zero, and each nonzero weight moves its outcome in the direction that the generator treats as favorable (Supplementary Note~4). They were drawn once by the global random stream (seed 13) and are shared by the generated dataset and the washout panels, whose memory-route loadings are $L_{kj}=W_{kj}/\max_{k'}|W_{k'j}|$. Values were recomputed from the shipped generator configuration (\texttt{npj\_ai\_generator\_constants\_v1}).}
\label{tab:s9}
\begin{tabular}{lrrrrr}
\toprule
Outcome & Phenylephrine & NaCl & Norepinephrine & Dextrose & FiO\textsubscript{2} \\
\midrule
\rows{supp_tables/s9_direct_route.tex}
\bottomrule
\end{tabular}
\end{table}

\begin{table}[h]
\centering
\small
\caption{\textbf{The two constructions that use the generator.} The generated dataset, which passed the realism audit, and the washout panels share the latent structural model, the background weights $w_k$ and the direct-route weights $W$ (Supplementary Note~4).}
\label{tab:s10}
\begin{tabularx}{\textwidth}{>{\raggedright\arraybackslash}p{3.3cm} >{\raggedright\arraybackslash}X >{\raggedright\arraybackslash}X}
\toprule
 & Generated dataset & Washout panels \\
\midrule
Records & All 200,825 eICU-CRD stays, first 336 hours & 400 eICU-CRD windows per panel, hours 24 to 163 \\
Channels & 12 outcome channels written; all 161 channels kept & 12 outcome channels read \\
Standardization & Cohort normalization statistics & Fixed reference values (heart rate 85 and 15 beats per minute) \\
Missing cells & Mean-filled for simulation; real missingness restored & Filled with reference means; not restored \\
Treatments & Recorded doses (indicator drives the model) & Simulated binary dosing of one treatment \\
Background scale & 0.02 & 1 \\
Mediator persistence & 1.0 (spectral radius 1.03, not mean-reverting) & 0.82 (spectral radius 0.93, half-life 10 hours) \\
Responsiveness $\eta_i$ & $1+N(0,0.5^2)$, clipped to $[1/3,3]$ & 1 \\
Traits and memory route & Off & Patient traits; bounded memory route with cap $c$ \\
Hidden confounder & Simulated, no effect ($\lambda=0$) & No effect ($\lambda=0$) \\
Counterfactuals & One hour at 20 decision hours; independent draws across arms & Sustained for 48 hours from hour 90; common random numbers \\
Truth or target & Minimum and maximum of the recorded outcome over 6 to 48 hours & 12-hour mean memory-route contrast on the unheld path \\
Measurement hold & On & On (real cadence) or off (hourly) \\
Realism audit & Passed (AUC 0.572) & Passed (AUC 0.714 to 0.727); low sensitivity on these windows \\
\bottomrule
\end{tabularx}
\end{table}

\clearpage
\section*{Supplementary Note 1: Registered wording}
\addcontentsline{toc}{section}{Supplementary Note 1: Registered wording}

Several registered analyses fixed in advance the sentences that may be used to report them. We quote them here verbatim. Key to internal terms: F is the oracle-structure learner; the decision-90 query is the treatment decision at hour 90; \texttt{rs\_dlm\_mech} and \texttt{cf\_grf\_mech} are the kernel-informed random-slope model and causal forest; UNRESOLVED marks a comparison that did not survive Holm correction; a ``twin'' in the map sentence is a supervised reader.

\paragraph{Static CATE learners (registered headline).} ``no standard static learner extracted usable heterogeneity at the decision-90 query; learner failure at this signal level; positivity not separated. F not outperformed by static standard CATE learners on this cell; UNRESOLVED never means equivalence. F's remaining structural knowledge (exact trait prior and shrinkage, bounded map, natural non-target policy prior), together with its Huber likelihood, adds accuracy beyond kernel-informed standard learners; F beat a comparator that did not beat zero (\texttt{rs\_dlm\_mech}, \texttt{cf\_grf\_mech})''

\paragraph{Sequence CATE models (registered headline).} ``no Causal Transformer or RMSN extracted usable heterogeneity at the decision-90 query; sequence-model failure at this signal level; positivity not separated (CRN failed the synthetic gate). F not outperformed by Causal Transformer and RMSN on this cell; UNRESOLVED never means equivalence''. The only permitted combined statement about the decision query is the conjunction of the three headlines (static, sequence and g-computation), with a family-wise error bound of 0.15; the control inside the support is reported separately, with a family-wise error bound of 0.10 when its two families are cited together.

\paragraph{G-computation sequence model (registered headline).} ``F not outperformed by G-Net (Causal Transformer code) on this cell; UNRESOLVED never means equivalence''

\paragraph{In-support query control (registered sentence).} ``Neither registered hour inside the dosing period (30 or 42) met the registered overlap criteria on the recorded treatments (Methods), so no control inside the support was scored, and the benchmark cannot distinguish a failure of the learners from the lack of treatment variation at the query.''

\paragraph{Historical benchmark.} ``On a semi-synthetic benchmark with a known causal model (five data-generating replicates of 60 outcome, treatment and horizon cells, each cell with 19 to 30 decisions of which 1 to 10 were treated), the model's average-treatment-effect estimates had lower absolute error than a naive treated-versus-untreated comparison (0.021 versus 0.213) but higher error than always predicting no effect (0.021 versus 0.017, worse in all five replicates), because 71\% of the true cell effects were exactly zero; this benchmark therefore does not show that the model recovers treatment effects.''

\paragraph{Identifiability map.} ``on real eICU records at their real cadence, for a treatment whose response carrier the record samples hourly and at an amplitude calibrated from clinical effect sizes, a twin that reads the record at its own draw times recovers 17 to 18 percent of the individual memory-route effect against a bar of 10.6, beats a history-blind ridge by 0.11 to 0.14 with lower bounds above +0.05, replicates on patients the first sample did not use, and has no advantage on a fixed schedule, where there is no patient-specific history to align to.'' The readers are fitted with simulator effect labels; the claim concerns information content and is not a claim about a deployable factual-only estimator. The replication in this sentence uses a second sample of 400 stays, 5 of them shared with the first, and reuses the person-indexed generator draws, including the traits. Its true effects are therefore nearly identical to those of the first sample (correlation 0.9998), so it tests recovery from different records of nearly the same effects (Supplementary Notes~2 and~4).

\paragraph{Personalization on validation.} The registered sentence states that the protocol's personalization condition was met on validation (12 of 12 cells after Holm, none significantly negative) and must carry three clauses: the matched patient's hours were spliced onto the patient's own first six hours, so a forecast degraded by the splice would also produce the advantage; the advantage ``was not shown to exceed that of the patient's own last measured target value over the donor's under the protocol's count rule (0 of 12 cells after Holm within that family)''; and in one cell (\spo{}, 18 to 12 hours) the number of measured hours separates the patient's record from the donor's. The splice-free comparison reported in the main text addresses the first clause.

\section*{Supplementary Note 2: Prior reads, selection and deviations}
\addcontentsline{toc}{section}{Supplementary Note 2: Prior reads, selection and deviations}

This note records which data had been read before each analysis and which analysis choices were made after earlier results were known. Because of these reads and choices, the main text treats all of its results as development evidence. Supplementary Note~7 gives the corresponding detail for the MIMIC-IV read.

\paragraph{Prior reads of evaluation splits.} The eICU-CRD test set had been read in earlier analyses of the synchronization contrast, and the per-cell rosters were designed after such a read; the validation split had also been read before the isolation arms. The MIMIC-IV person-disjoint test split (14,211 stays) was scored in earlier campaigns for mortality, length of stay, legacy onset events and threshold events, before its registered read (Supplementary Note~7). The MIMIC-III test split (4,244 stays) was read in 21 earlier campaigns (Supplementary Note~8). All of these reads are disclosed in a test-read registry (46 plan rows and 62 disclosure rows).

\paragraph{Selection of the registered cell.} The new evidence for the registered cell comes from the replication panel and the fixed-schedule control, because the cell was chosen on the discovery stays. We registered the amplitude of 0.75 SD after an exploratory amplitude sweep on the 400 discovery stays, in which 0.75 SD was the smallest amplitude at which the cadence-aware reader separated. That reader had itself been added after the readers then in use did not separate at real cadence. The registered discovery run is bit-identical to the exploratory threshold run on the same stays. The replication panel, a second sample of 400 stays of which 395 are new, reuses the person-indexed generator draws of the discovery run, including the traits. It therefore carries nearly identical injected effects (correlation 0.9998; Supplementary Note~4) but differs from the discovery panel in real background physiology, cadence and natural dosing. Norepinephrine on heart rate was run as a sensitivity cell and was not registered. Because the amplitude and the reader were fixed after outcome-bearing runs, the search ledger classes the registered-cell runs as exploratory.

\paragraph{Synchronization analyses.} These analyses reused splits that had been read before, and some of their choices were made after an earlier read. The roster rule of the registered synchronization contrast was set after an earlier read of the eICU-CRD test set. The probability clip of the calibrated log-loss was changed after the test read, so we interpret the log-loss by its direction only. The last-value comparison was run after the synchronization result and is reported as a disclosed, unregistered analysis. The isolation arms ran on a validation split that had already been read for checkpoint selection, calibration maps and earlier landmark readouts.

\paragraph{Theory and post hoc checks.} The propositions of Supplementary Note~5 and their numerical checks were derived and run after the results, and no eICU-CRD analysis was chosen, registered or stopped because of them. The checks made no new prediction of the twin or of any benchmarked model, and they read washout-panel targets only as the per-patient targets that the registered benchmark had already scored. Two groups of checks were new analyses of previously read data. The first read the shipped per-stay predictions, last measured values and labels of the synchronization runs on the eICU-CRD test set. It compared the hour-6 arm with the model's hour-6 forecast of the earlier window and computed tie rates and prevalences of the last measured values. As an unreported diagnostic, it also fitted cross-fitted logistic combinations of the existing predictions on those labels. These reads are recorded as one disclosure row of the test-read registry (TRB-EI-18). The second group fitted classifiers to the summaries of the realism probe, including a flexible classifier in one configuration chosen after the audit.

\paragraph{Washout panels.} {\emergencystretch=2em The individual-effect analyses use washout panels of 400 real eICU-CRD windows whose outcome channels are real and whose treatment columns are replaced by simulated dosing; missing cells are filled with fixed reference means before the measurement hold is applied. The panels were realism-audited under a registered protocol (\texttt{reports/\allowbreak{}revision\_\allowbreak{}v3/\allowbreak{}WASHOUT\_\allowbreak{}PANEL\_\allowbreak{}REALISM\_\allowbreak{}AUDIT\_\allowbreak{}PROTOCOL\_\allowbreak{}20261011\_\allowbreak{}V2.md}, tag \texttt{washout-realism-audit-protocol-v2}) after their other analyses had been run, including the exploratory construction campaign from which the registered cell was chosen. The registered runs of the in-support control and the g-computation comparator followed the audit. Every audit of that protocol is in Supplementary Table~S29, and two earlier realism reads of an earlier washout design on synthetic tapes are disclosed there. The panels' stays were drawn without regard to the eICU-CRD split, so 117 of the 795 distinct stays belong to the test split. The realism audit of the generated dataset does not cover the panels; that dataset carries no covariate-dependent effect heterogeneity and is not the substrate of the individual-effect results (Supplementary Note~4).\par}

\paragraph{ICare-RH's placebo gate (G7).} ICare-RH's design was fixed before it was fitted to any real panel, although the reader and learner analyses had already read the washout panels. The first dispatch of G7 stopped when its native test suite failed, before the analysis driver ran, and nothing was read. One test had required a recorded comparison of two scorers to be bitwise identical, which the cluster's linear algebra library did not reproduce. A second frozen version, logged in the protocol, changed this test to a relative tolerance of $10^{-12}$, and the gate then ran once.

\paragraph{Clinical-endpoint probes.} The two probes, P0 and P1, used the same tuning set, which was therefore read twice, and P1 was designed after P0 had read its labels. The pretrained encoder had also been trained on every tuning-set patient, so any further recipe would have to be selected on an inner split and read once on the validation split. The gradient-boosting comparator was a fixed recipe, and the planned set of tuned comparators was not built after P0 stopped. Supplementary Note~6 and Supplementary Fig.~S3 report the results of both probes.

\paragraph{Search and compute.} A search ledger of 2,586 rows records every campaign, run and configuration of the project: of 606 campaign and run rows, 28 were positive, 110 negative, 17 null, 24 mixed, 245 descriptive, 72 failed, 23 not run, 69 withdrawn and 18 unrecorded; 280 were registered before their outcome and 189 were exploratory. The compute ledger records at least 17,683 allocated CPU-thread-hours and 362 GPU-hours between 1 August and 26 September 2026; earlier months are not metered. Both ledgers stop on 26 September 2026 and do not contain the endpoint probes, the isolation arms, the placebo gate, the post hoc checks of Supplementary Note~5, the registered MIMIC-IV read or the registered analyses of the washout panels added before submission; the protocols of the probes and of G7 estimated small costs (P0 about 1 to 1.5 GPU-hours and 30 to 80 CPU core-hours; G7 about 6 CPU-hours), and the compute of the analyses of the washout panels is given in Supplementary Note~9.

\section*{Supplementary Note 3: Rollout clock defect}
\addcontentsline{toc}{section}{Supplementary Note 3: Rollout clock defect}

A deterministic test suite (21 tests) found a clock defect in the action-conditioned rollouts of an earlier model version. The historical rollout benchmark (Fig.~4a of the main text and Supplementary Fig.~S2) ran with this defect and is reported as run. No positive finding of the paper rests on its readouts, and the individual-effect analyses on the washout panels do not use these rollouts. All three rollouts (the joint competing-event rollout, the transition-adapter rollout and a dynamics-head rollout) wrote each generated transition into the row of the action that conditioned it. The transition adapter had been trained to map the state through hour $t-1$ and the action at $t$ to physiology at $t+1$, so every generated row held a forecast for a later hour than its label. The error is a pure one-row shift where persistence is carried only by an exposure computed from the actions. Where persistence is carried through generated physiology, the error compounds, and a single dose is counted twice before decaying. Over the 12-hour window of the registered washout target, with sustained dosing, toy models inflated the mean response by factors of 1.00 to 1.38 (1.20 for an observed channel with autocorrelation 0.8), whereas a correctly clocked rollout read over the window one hour earlier would under-read it. The effect of the defect on the historical readouts was not estimated.

\section*{Supplementary Note 4: Semi-synthetic data generation}
\addcontentsline{toc}{section}{Supplementary Note 4: Semi-synthetic data generation}

This note gives the equations and constants of the semi-synthetic generator (Fig.~3 of the main text) for the two constructions that use it. Part A describes the generated dataset, which has treatment responses written onto every eICU-CRD stay and passed the realism audit. Part B describes the washout panels, which carry every individual-effect result. The two constructions share the latent structural model, the background weights and the direct-route weights, but they differ in several parameters and in the standardization, treatment policy, random-number regime and target (Supplementary Table~S10). The generated dataset was produced at the code version whose hash is recorded in its shipped configuration. The washout panels ran a later version of the same module, whose added options are off by default and inactive in the generated dataset. We recomputed the constants below from the shipped configuration and the generator code without reading patient data (\texttt{paper\_npj\_ai/make\_generator\_constants.py}) and read the panel facts from the shipped receipts of the registered cell.

\subsection*{A. The generated dataset}

\paragraph{Inputs and standardization.} Each stay's hourly record is cut to its first 336 rows. The generator writes 12 outcome channels (invasive mean arterial pressure, \spo{} by pulse oximetry, lactate, heart rate, respiratory rate, urine output, creatinine, pH, white cell count, platelets, bilirubin and temperature) and reads five treatment channels (phenylephrine, NaCl, norepinephrine, dextrose and inspired oxygen fraction). All other channels are copied from the real record. Missing cells are filled with the channel mean before simulation, and the unfilled record is kept to restore missingness afterward. Outcome $k$ is standardized with the cohort normalization statistics, $y^{0}_{tk}=(x_{tk}-\mu_k)/\sigma_k$, and clipped to $\pm6$ and to physiological bounds expressed in the same units (for example 20 to 250 beats per minute for heart rate and 20 to 200 mmHg for mean arterial pressure). A mean-filled cell therefore has $y^0=0$.

\paragraph{Random numbers and fixed constants.} Each stay has its own random stream, seeded from 13 and a hash of the stay identifier, so its draws do not depend on the other stays. A global stream seeded with 13 draws the background weights $w_k$ (templates per channel family with small jitter) and the direct-route matrix $W$ once. Each entry of $W$ is drawn from $N(0,0.35^2)$, set to zero with probability 0.35 and sign-aligned so that treatment moves each outcome in the direction that the generator treats as favorable (upwards for mean arterial pressure, \spo{}, urine output and pH, downwards otherwise). The realized $W$ has 28 zero entries of 60, and temperature has no direct route (Supplementary Table~S9). The mediator transition $A_m$ has diagonal entries 0.85 to 0.88 and off-diagonal entries 0.02 to 0.05, and severity enters the mediator through $b_u=(0.15,0.12,0.10,0.12)$. The treatment-to-mediator rows are fixed for phenylephrine, NaCl and norepinephrine and drawn per stay from $N(0,0.08^2)$ for dextrose and inspired oxygen fraction. Each stay also draws the propensity weights and intercepts that drive the counterfactual policy, from $N(0,0.6^2)$ with a positive first column and from $N(-2,0.3^2)$, respectively. Its responsiveness $\eta_i=\min\{\max\{1+N(0,0.5^2),1/3\},3\}$ is drawn independently of its record, and about 9\% of stays are expected at the lower clip.

\paragraph{Latent dynamics.} The severity $u$ and the mediator $m$ start from the mean standardized outcome of the first hour (clipped to $\pm2$) and evolve as
\[
u_{t+1}=0.85\,u_t+0.10\,\bar m_t+0.05\,\bar a_t+\xi^u_t,\qquad m_{t+1}=A_m m_t+b_u u_t+B^\top a_t+\xi^m_t,
\]
with $\xi\sim N(0,0.25^2)$ and clipping at $\pm8$. The generated dataset uses the legacy mediator persistence (the multiplier on $A_m$) of 1.0, under which the joint linear map of $(u,m)$ has spectral radius 1.03 (doubling time 25 hours). Its latent state is therefore not mean-reverting and is bounded only by the clip, but its influence on the outcomes is scaled by 0.02. A hidden confounder $v_{t+1}=0.9\,v_t+N(0,0.25^2)$ is simulated but enters neither the policy nor the outcomes ($\lambda=0$). Patient traits are off, so the bounded memory route of the washout panels is absent.

\paragraph{Factual treatments.} The factual path replays the record. Treatment $j$ is on in hour $t$ ($a_{tj}=1$) when it has a recorded dose above $10^{-9}$ in that hour, and only this indicator enters the dynamics. The recorded dose is written back at single precision, and a rounded dose not exceeding $10^{-9}$ is set to 0.

\paragraph{Outcome.} The structural change and the generated outcome are
\[
\delta_{tk}=\mathrm{clip}_{\pm6}\big\{0.02\,(w_k^\top[u_t,m_t]+\varepsilon_{tk})+0.8\,\eta_i\,(W a_{t-2})_k\big\},\qquad y_{tk}=\mathrm{clip}\big\{y^0_{tk}+0.6\,\delta_{tk}\big\},
\]
with $\varepsilon\sim N(0,0.25^2)$ and $y$ clipped to $\pm8$ and to the physiological bounds, then back-transformed as $\mu_k+\sigma_k y_{tk}$ and clipped to the same bounds in raw units. Per treated hour the direct route moves a standardized outcome by $0.6\times0.8\,\eta_iW_{kj}$, for example by $-0.39\,\eta_i$ standard deviations on creatinine per hour of NaCl and $-0.11\,\eta_i$ on heart rate per hour of phenylephrine.

\paragraph{Measurement hold.} The measurement hold records a new value only in hours in which the real value changed. Because BlendedICU forward-fills, an unchanged value is taken to mark an hour without a new measurement. Hour $t>0$ of channel $k$ therefore counts as not re-measured when its standardized real value equals that of the previous hour exactly, and in such hours the recorded value repeats the previous recorded value. Only the recorded value is held. The latent state, the noise draws and the unheld outcome are computed every hour exactly as without the hold, so the random-number stream is unchanged. The first hour always receives a fresh value, and runs of missing (mean-filled) cells are held and masked afterward. With the hold off, the generator reproduces earlier datasets bit for bit.

\paragraph{Missingness correction.} The missingness correction restores the real missingness of the treatment channels. After simulation, every cell that is missing in the real record is set to missing in the generated record, and the correction extends this rule from the non-treatment channels to the five treatment channels. The 144 channels that the generator does not write are therefore the real record, the 12 outcome channels carry semi-synthetic values where the real channel is observed, and the treatment channels equal the real ones. The generator permits the correction only when factual treatments replay the recorded doses. Under a simulated policy the treatment values are generated, and a missing value would assert an absence that the real record never made. The generator's own synthetic measurement process is inactive.

\paragraph{Counterfactuals.} Counterfactuals were requested for the first 2,000 test-split stays. Of these, 674 had records longer than 51 hours, as the 48-hour horizon requires, and 1,326 were too short. For each eligible stay, 20 decision hours were drawn uniformly with replacement, and a 48-hour rollout started from the factual latent state for each decision hour, treatment and arm, giving 200 rollouts per stay. At the decision hour the chosen treatment was set to the arm's value (0 or 1), and the others were drawn from the propensity policy $p_t=\sigma(\mathrm{clip}_{\pm8}\{2W_p[u_t,m_t]+b\})$. Afterward all five treatments followed this policy rather than the recorded doses. The two arms used different random draws. The direct route used factual actions for lagged hours before the decision and counterfactual actions after it, and the hold used the real record's change times. The stored truth is the minimum and maximum of the recorded counterfactual outcome over 6, 12, 24 and 48 hours. Under the hold, this truth is the effect on the recorded value, so the do(1) minus do(0) contrast of a channel is zero until the channel's first real re-measurement after the decision.

\paragraph{Scale.} The audit found all 200,825 eICU-CRD stays in the generated dataset. The generator's final log counts 198,825 stays without a counterfactual request, 674 with counterfactual truth and 1,326 requested but too short for it, which sum to 200,825. The index file has 200,826 lines including its header.

\subsection*{B. The washout panels}

\paragraph{Records.} Each washout panel holds 400 real eICU-CRD windows. Candidate stays are the eICU-CRD stays present in both the time-series index and the label table, with no split filter. A seeded permutation selects the first 400 stays with at least 164 hourly rows, and the window is hours 24 to 163 of each stay. The discovery and fixed-schedule panels share one seed, and the replication panel uses another. From each window the 12 outcome channels are read, missing cells are filled with fixed reference means, values are kept strictly inside the physiological bounds, and the treatment columns are set to zero. Synthetic records replace the real window with smooth random curves (an offset, a slope, a sinusoid and small noise per channel). No two consecutive hours of these curves are equal, so the hold has no effect on them.

\paragraph{Parameterization.} The panels use a frozen parameter set (Supplementary Table~S10). The background scale is 1, and a mediator persistence of 0.82 (spectral radius 0.93, half-life 10 hours) makes the latent state mean-reverting. There is no per-stay multiplier ($\eta_i=1$), and patient traits act through a bounded memory route. The lag is 2 hours, counterfactual arms share common random numbers, and dosing is simulated and binary. All other structural constants, including $W$ and $w_k$, equal those of the generated dataset. Every person-level draw (traits, propensity weights and intercepts, noise and top-up) uses a stream keyed to the person index rather than to the stay.

\paragraph{Traits.} Each person carries one trait per treatment, which weights that treatment's contribution to the memory route. The trait of person $i$ for treatment $j$ is $\zeta_{ij}=\min\{\max\{0.6\,l_j\xi_i+0.15\,\epsilon_{ij},-2/3\},2\}$, with one shared standard normal factor $\xi_i$, loadings $l=(1.09,1.11,0.99,1.14,0.78)$ drawn once and independent standard normal $\epsilon_{ij}$. For phenylephrine in the discovery panel the trait has standard deviation 0.59, and 14\% of persons sit at the lower clip.

\paragraph{Dosing.} Each factual record runs through a lead-in, a dosing period and a washout before the decision at row 90. Rows 0 to 5 are the lead-in. Natural dosing in rows 6 to 53 is harvested from a separate factual rollout on the same window in which all five treatments follow the propensity policy, and only the scored treatment's column is kept. If fewer than eight rows are dosed, a contiguous undosed block, or failing that random undosed rows, tops the count up to eight. All other rows and treatments are zero in the factual record. In the discovery panel the scored treatment was given in 21.4 rows on average (standard deviation 14.7, range 8 to 48). Rows 54 to 89 are the washout, after which the exposure memory at the decision averages 0.001. The fixed-schedule control doses rows 6 to 24 for every person.

\paragraph{Memory route.} The memory route adds a bounded, patient-specific response that builds with recent doses and decays after them. For each treatment, the exposure memory is $e_{tj}=0.8\,e_{t-1,j}+a_{t-2,j}$ and its occupancy $q_{tj}=e_{tj}/(1+e_{tj})$, both computed without noise and without feedback into the latent state. The shift on outcome $k$ is $s_{tk}=c\tanh(\sum_jL_{kj}\zeta_{ij}q_{tj})$ with $L_{kj}=W_{kj}/\max_{k'}|W_{k'j}|$. On heart rate the loadings are $-1$ for phenylephrine, $-0.75$ for norepinephrine, $-0.26$ for inspired oxygen fraction and 0 for NaCl and dextrose. The shift is added after the ordinary clips and attenuated near a bound, so the outcome stays inside it. Under sustained dosing from an empty memory, occupancy by outcome hour $90+k$ is 0, 0.50, 0.64, 0.71, 0.75 and 0.77 for $k=1$ to 6 and 0.82 by $k=12$. It approaches 5/6 and never reaches full occupancy. The registered cell uses a cap of $c=0.75$ reference standard deviations, which is 11.25 beats per minute on heart rate. For the 400 discovery traits, the scored treatment's own shift has a median of 3.8 beats per minute at steady-state occupancy (95th percentile 8.7) and 4.5 at full occupancy. The recoverability map also used caps of 0.25, 0.5 and 1.

\paragraph{Intervention, frames and target.} The target isolates the memory-route effect of the scored treatment by comparing an active frame with a placebo frame. At row 90 the scored treatment is set to 1 or 0 and held for 48 hours. The other four treatments follow the propensity policy in both arms, and arm 1 replays arm 0's random stream. Each person is simulated in two frames on the same record and actions, an active frame with cap $c$ and a placebo frame with $c=0$. Because the memory route does not feed back into the state, every trace except the outcome is identical between the frames (no mismatches in any receipt). The placebo frame's factual record serves as the zero-effect copy used by the ICare-RH placebo gate. The target is the 12-hour mean of the do(1) minus do(0) difference in the unheld heart rate in the active frame, minus the same quantity in the placebo frame, in units of 15 beats per minute. The ordinary treatment routes are identical in the two frames and cancel in the target. They comprise the direct route of $0.6\times0.8\times W_{\mathrm{HR,phenylephrine}}=-0.11$ standard deviations per dosed hour, the mediated route and any divergence of the other treatments' policy draws. Their contribution, the do(1) minus do(0) contrast of the placebo frame, has a standard deviation across persons of 1.48 beats per minute in the discovery panel. The target itself has a standard deviation of 3.48 beats per minute in every real panel, with or without the hold, and a mean of $-0.18$ beats per minute in the discovery and replication panels. Because the policy also doses the other treatments after the decision, their memory enters the tanh, and the target is not a function of the scored trait alone. A predictor that is exact when the other treatments are held at zero correlates 0.977 with the target, and the oracle reference, which knows only the scored trait, reaches a recovery of 0.80 rather than 1. The value 0.977 comes from a simulation that is quoted in the registered protocol but whose code was not committed.

\paragraph{Cadence.} Measurement cadence changes what the readers see but not the target. At real cadence the generated record repeats its previous value in every hour whose real value did not change. Heart rate repeats in 43\% of the hours of the discovery panel and in 41\% of the replication panel. With hourly observation every hour receives a fresh value. The hold acts only on the pre-decision record of the active frame, which is what the readers see, and the target is computed from the unheld outcome.

\paragraph{Replication and fixed schedule.} The replication panel tests recovery of nearly identical injected effects from different records. It keeps the person indices, and hence the traits and person-level streams, but selects different stays (5 of 400 shared with the discovery panel) and re-harvests natural dosing on them. Its true effects, the targets defined above, correlate 0.9998 with those of the discovery panel. The fixed-schedule panel uses the discovery stays and persons with rows 6 to 24 dosed.

\input{\suppdir supplementary_note5}

\clearpage
\section*{Supplementary Note 6: Cohort flow, forecasts and clinical-endpoint probes on eICU-CRD}
\addcontentsline{toc}{section}{Supplementary Note 6: Cohort flow, forecasts and clinical-endpoint probes on eICU-CRD}

This note gives the details behind the eICU-CRD forecast and endpoint results of the main text. It traces the cohort to the stays scored in the registered read of the test set and then reports the calibration of the forecasts, the death forecast and the provenance of the read. The last part describes the two clinical-endpoint probes in full.

\paragraph{Cohort flow.} Supplementary Table~S18 follows the ICU stays from the three source databases to the populations evaluated in the forecast read. Harmonization removed 34 of the 200,859 eICU-CRD stays, and MIMIC-III was restricted to CareVue stays of patients who do not appear in MIMIC-IV. Each eICU-CRD patient, with all of their stays, belongs to one split, so the patients of the three splits sum to the 139,362 eICU-CRD patients of the cohort. Baseline characteristics were not tabulated by cohort. Discrimination by age, sex and unit type is given in Supplementary Table~S7, and by ethnicity in its legend.

\paragraph{Evaluated populations.} Each crossing setting was scored in a population selected by measurement rather than drawn at random from the test set. All forecasts were read at one landmark, hour 48 (Methods, Forecast evaluation). A stay entered a crossing setting only if its target was measured at hour 48 without having crossed and its outcome was then resolved by a crossing, by an exit from the unit or by measurement of the target in every hour of the horizon. The evaluated populations therefore shrink as the horizon lengthens and are smaller for \spo{} (227 to 352 stays) than for heart rate (801 to 1,449 stays). Longer horizons also have higher crossing rates, from 0.066 to 0.444 for heart rate and from 0.173 to 0.815 for \spo{}, and AUPRC, which depends on that rate, rises with them (Supplementary Table~S19).

\paragraph{Calibration.} The forecasts were not recalibrated on test data, and at short horizons their mean came close to the observed crossing rate. At 6 and 12 hours the mean forecast was 91\% to 105\% of the observed rate for every vital sign. At 24 and 48 hours it was lower, 68\% to 89\% of the observed rate, with the largest shortfall for heart rate at 48 hours (0.300 against 0.444). Calibration slopes of the crossing forecasts ranged from 0.65 to 1.27 across settings. The lowest, 0.65 to 0.73 for \spo{} at 6 to 24 hours and for respiratory rate at 48 hours, show that these forecasts were too extreme. Reliability tables for every seed and setting are in \texttt{joint\_factual\_metrics\_by\_seed.csv}, a source file of Supplementary Table~S19.

\paragraph{Death forecast.} The model has no separate mortality head, so the death forecast, fixed in the analysis plan as a descriptive setting, is the largest of the four target-specific death cumulative incidences at 48 hours. It discriminated death in the unit within 48 hours with an AUROC of 0.798 (95\% interval across seeds 0.768 to 0.828) and an AUPRC of 0.195 against an observed rate of 0.0256. Its mean forecast, 0.0023, was far below that rate, in part by construction, because each target-specific head counts a death only if it precedes that target's crossing. In each seed, all but two to five of the 11,498 forecasts were below 0.1, and the logistic calibration fit diverged. At an alert threshold of 0.03 the forecast flagged 41 to 83 stays across seeds (0.36 to 0.72 per 100), with net benefit of 0.0012 to 0.0026, and at 0.15 it flagged at most three. Its net benefit over alerting on every patient or on none (Supplementary Table~S8) therefore rests on very few alerts. Recalibration was not part of the registered read.

\paragraph{Registration and reproduction.} The read followed an analysis plan (digest \texttt{ecf365f4}) that was locked before the test set was opened for this model. The plan fixed the landmark, the settings, the metrics, both kinds of interval (across seeds and from a patient bootstrap within each seed) and the rule that every setting is reported. Before this read, much of the test set had been scored by models trained with an earlier objective, as recorded in the test-read registry (Supplementary Note~2). The read ran as jobs 5276 (predictions), 5293 (metrics) and 5312 (verification). The verifier bundled with job 5293 failed after the metrics had been computed, because it carried an outdated copy of the plan digest. Job 5312 then reproduced all 85 seed-by-setting results exactly from the sealed prediction files, without a new model run. The same plan registered two further reads of this test set with the seed-29 checkpoint. The first, an earlier version of the synchronization contrast (jobs 5264 and 5275), is superseded because its hour-6 control differed from the synchronized forecast in estimand and calibration and because its panel required survival to the last landmark. The second (job 5265) compared a logistic model on the frozen state at hour 48 with LightGBM on summaries of the same 48 hours, for respiratory-rate crossings within the next 48 hours. The log-loss difference (logistic model minus LightGBM) was $-0.013$ (95\% interval $-0.031$ to 0.004; 697 stays of 674 patients). The subgroup and decision-curve analyses (Supplementary Tables~S7 and S8, job 10545) re-read the same prediction files and are descriptive.

\paragraph{Clinical-endpoint probes.} A threshold crossing does not by itself define a clinical event, so two registered development probes, P0 and P1, asked whether the patient state also carries information on clinical endpoints (Supplementary Fig.~S3; Methods of the main text, Clinical endpoints). The probes scored four endpoints within 24 hours of landmarks at 12, 24 and 48 hours. These were death in the unit, hypotension (mean arterial pressure below 65 mmHg in two consecutive measured hours), acute kidney injury by the KDIGO creatinine criterion, and hypoxemia (\spo{} below 90\% in two consecutive measured hours). Both probes used the tuning set of the eICU-CRD training split. The pretrained encoder had seen every stay of the set's 14,607 patients in training, and the 14,122 patients with at least one landmark row were scored. On this set the boosting comparator, a fixed CatBoost recipe on 1,136 summary features of the record, reached AUROCs of 0.941, 0.816, 0.824 and 0.868 for the four endpoints in the order above (Supplementary Fig.~S3a).

\paragraph{Frozen state (probe P0).} The frozen state of the vital-sign model, which was trained on other targets, trailed boosting by 0.020 to 0.086 AUROC (Supplementary Fig.~S3b). The probe's protocol, written before the run, had forecast deficits of about 0.01 to 0.02. The deficit was smallest for death, which the forecasting heads model as a competing event. Added to the boosting features, the frozen state raised AUROC by at most 0.0014, and every interval included zero.

\paragraph{Trained encoders (probe P1).} Training the same encoder architecture from random initialization on the four endpoints closed 42\% to 94\% of the frozen-state deficit (Supplementary Fig.~S3c). This encoder still trailed boosting on every endpoint, by 0.004 to 0.021 AUROC, with every interval below zero. Adding the fine-tuned state of the pretrained encoder to the boosting features (the hybrid) improved discrimination of hypotension by 0.0042 and of acute kidney injury by 0.0075 (95\% CIs 0.0015 to 0.0067 and 0.0035 to 0.0116). The hybrid also lowered log-loss on these two endpoints, and its AUROC intervals for death and hypoxemia included zero (Supplementary Fig.~S3b,d). Used alone, the encoder fine-tuned from the pretrained twin was worse than the encoder from random initialization on hypotension, acute kidney injury and hypoxemia, by 0.006 to 0.014 AUROC. This comparison is confounded, because fine-tuning used a lower learning rate for the encoder.

\paragraph{Status of the probes.} Neither probe met its registered gate, and P1 missed it narrowly (Table~2 of the main text). No family-wise correction was applied, so the intervals are unadjusted. P1 was designed after P0 had read the tuning labels, and the pretrained encoder had been trained on every tuning-set patient (Supplementary Note~2). We therefore report the two P1 results above as development observations on eICU-CRD. The hybrid improved discrimination of hypotension and acute kidney injury, and the encoder trained from random initialization trailed boosting on every endpoint. The boosting comparator, both encoders and the hybrid were later refitted within the MIMIC-IV training split and read once on the previously analyzed MIMIC-IV test split under a registered protocol (main text and Supplementary Note~7).

\begin{table}[h]
\centering
\small
\caption{\textbf{Cohort inclusion flow.} ICU stays and patients from the source databases to the populations evaluated in the forecast read on the eICU-CRD test set; --, not tabulated. Counts are from \texttt{inclusion\_flow.csv}, \texttt{cohort\_characteristics.csv} and \texttt{joint\_factual\_metrics\_by\_seed.csv} in \texttt{source\_tables/revision\_v3/}; the training and validation patient counts are from the landmark event-time reports in \texttt{reports/revision\_v3/}.}
\label{tab:s18}
\begin{tabularx}{\textwidth}{>{\raggedright\arraybackslash}X r r >{\raggedright\arraybackslash}p{6.2cm}}
\toprule
Stage & Stays & Patients & Restriction or note \\
\midrule
Raw ICU stays & 356,849 & -- & eICU-CRD 200,859; MIMIC-III 61,532; MIMIC-IV 94,458 \\
Harmonized stays & 356,815 & -- & BlendedICU harmonization; 34 eICU-CRD stays removed \\
Experimental cohort & 323,677 & -- & eICU-CRD 200,825; MIMIC-IV 94,458; MIMIC-III 28,394 CareVue stays of 23,692 patients not in MIMIC-IV (from 61,532 stays of 46,476 patients) \\
\quad eICU-CRD (208 hospitals) & 200,825 & 139,362 & Split by patient \\
\quad\quad Training split & 140,473 & 97,553 & \\
\quad\quad Validation split & 30,213 & 20,904 & \\
\quad\quad Test split & 30,139 & 20,905 & \\
\quad\quad\quad Test set & 30,099 & 20,886 & All 40 stays of the 19 patients opened in an earlier analysis removed \\
\quad\quad\quad\quad Evaluable for a crossing & 227 to 1,449 & 220 to 1,349 & Per setting at hour 48: target measured and not crossed, outcome resolved \\
\quad\quad\quad\quad Evaluable for death & 11,498 & 9,506 & In the unit after hour 48 with a valid discharge record \\
\bottomrule
\end{tabularx}
\end{table}

\begin{table}[h]
\centering
\footnotesize
\caption{\textbf{Discrimination and calibration of the joint competing-risks forecasts on the eICU-CRD test set.} Forecasts issued at hour 48 by the five checkpoints in the registered read (Methods, Forecast evaluation). Stays, patients and events (threshold crossings, or deaths in the unit for the death forecast) describe the evaluable population of each setting and are identical across seeds. Metrics are five-seed means. The AUROC interval is a 95\% $t$ interval across seeds (4 degrees of freedom), and the next column gives the AUROC of seed 29, the checkpoint with the lowest validation loss, with a 95\% interval from 2,000 bootstrap replicates that resample patients. Log-loss and Brier score are those of the raw forecasts. Intercept and slope: logistic regression of the outcome on the logit of the forecast (0 and 1 under perfect calibration). The death forecast is the largest of the four target-specific death cumulative incidences at 48 hours; its calibration fit diverged in every seed (n.e., not estimable). Sources: \texttt{source\_tables/revision\_v3/joint\_factual\_seed\_summary.csv}, \texttt{joint\_factual\_metrics\_by\_seed.csv} and \texttt{joint\_factual\_person\_cluster\_bootstrap.csv}.}
\label{tab:s19}
\setlength{\tabcolsep}{4pt}
\begin{adjustbox}{max width=\textwidth}\begin{tabular}{l r r r r r r l l r r r r r}
\toprule
Forecast & $H$ (h) & Stays & Patients & Events & \shortstack{Observed\\rate} & \shortstack{Mean\\forecast} & \shortstack{AUROC\\{[95\% CI]}} & \shortstack{AUROC, seed 29\\{[95\% CI]}} & AUPRC & Log-loss & Brier & Intercept & Slope \\
\midrule
Heart rate & 6 & 1,449 & 1,349 & 95 & 0.066 & 0.061 & 0.820 [0.813, 0.826] & 0.823 [0.784, 0.860] & 0.253 & 0.200 & 0.055 & 0.12 & 1.01 \\
 & 12 & 1,309 & 1,224 & 154 & 0.118 & 0.111 & 0.780 [0.774, 0.786] & 0.781 [0.740, 0.818] & 0.333 & 0.308 & 0.091 & 0.10 & 1.01 \\
 & 24 & 1,029 & 967 & 252 & 0.245 & 0.198 & 0.729 [0.725, 0.734] & 0.730 [0.695, 0.764] & 0.458 & 0.503 & 0.165 & 0.19 & 0.90 \\
 & 48 & 801 & 758 & 356 & 0.444 & 0.300 & 0.709 [0.704, 0.715] & 0.708 [0.673, 0.744] & 0.655 & 0.674 & 0.237 & 0.65 & 0.91 \\
\midrule
Respiratory rate & 6 & 1,102 & 1,058 & 108 & 0.098 & 0.102 & 0.784 [0.771, 0.796] & 0.791 [0.745, 0.834] & 0.341 & 0.272 & 0.078 & 0.46 & 1.27 \\
 & 12 & 1,016 & 979 & 201 & 0.198 & 0.185 & 0.733 [0.720, 0.747] & 0.740 [0.702, 0.779] & 0.415 & 0.443 & 0.141 & 0.22 & 1.09 \\
 & 24 & 838 & 809 & 298 & 0.356 & 0.316 & 0.693 [0.681, 0.705] & 0.701 [0.664, 0.738] & 0.534 & 0.603 & 0.208 & 0.13 & 0.89 \\
 & 48 & 697 & 674 & 407 & 0.584 & 0.468 & 0.660 [0.649, 0.672] & 0.665 [0.621, 0.707] & 0.703 & 0.682 & 0.243 & 0.47 & 0.67 \\
\midrule
\spo{} & 6 & 352 & 339 & 61 & 0.173 & 0.182 & 0.656 [0.645, 0.667] & 0.657 [0.584, 0.726] & 0.286 & 0.448 & 0.140 & $-$0.54 & 0.65 \\
 & 12 & 307 & 296 & 105 & 0.342 & 0.313 & 0.656 [0.640, 0.671] & 0.649 [0.586, 0.713] & 0.492 & 0.618 & 0.215 & $-$0.05 & 0.73 \\
 & 24 & 257 & 250 & 150 & 0.584 & 0.502 & 0.648 [0.631, 0.664] & 0.645 [0.577, 0.717] & 0.690 & 0.675 & 0.240 & 0.36 & 0.68 \\
 & 48 & 227 & 220 & 185 & 0.815 & 0.688 & 0.725 [0.697, 0.753] & 0.725 [0.636, 0.810] & 0.912 & 0.480 & 0.154 & 0.63 & 1.27 \\
\midrule
SaO\textsubscript{2} & 6 & 962 & 910 & 143 & 0.149 & 0.139 & 0.703 [0.697, 0.710] & 0.710 [0.668, 0.752] & 0.334 & 0.385 & 0.117 & $-$0.02 & 0.93 \\
 & 12 & 897 & 851 & 235 & 0.262 & 0.248 & 0.710 [0.707, 0.712] & 0.709 [0.672, 0.744] & 0.452 & 0.523 & 0.174 & 0.13 & 1.04 \\
 & 24 & 788 & 751 & 375 & 0.476 & 0.410 & 0.686 [0.680, 0.693] & 0.686 [0.651, 0.724] & 0.642 & 0.650 & 0.229 & 0.30 & 1.00 \\
 & 48 & 701 & 673 & 481 & 0.686 & 0.588 & 0.646 [0.625, 0.667] & 0.646 [0.599, 0.691] & 0.784 & 0.620 & 0.216 & 0.54 & 0.75 \\
\midrule
Death in the unit & 48 & 11,498 & 9,506 & 294 & 0.0256 & 0.0023 & 0.798 [0.768, 0.828] & 0.808 [0.780, 0.835] & 0.195 & 0.138 & 0.025 & n.e. & n.e. \\
\bottomrule
\end{tabular}\end{adjustbox}
\end{table}

\clearpage
\captionsetup{hypcap=false}
\section*{Supplementary Note 7: The registered MIMIC-IV read}
\addcontentsline{toc}{section}{Supplementary Note 7: The registered MIMIC-IV read}

\paragraph{Scope and sources.} This note reports every outcome and every setting of the registered read of the person-disjoint MIMIC-IV test split, as the all-results rule of the protocol requires. The read had two registered families. Family A applied the five frozen eICU-CRD checkpoints to MIMIC-IV without refitting and asked whether synchronization still improved the forecasts. Family B refitted the clinical-endpoint probes within the MIMIC-IV training split and asked whether the fine-tuned twin state still added to gradient boosting. The frozen model's death and unit-exit forecasts were also scored at hour 48 as disclosed secondary outcomes, with no criterion. The test split had been read before. The protocol lists 34 earlier entries, 27 of them scientific reads of the split, including zero-shot transfers of earlier versions of the eICU-CRD-trained twin and outcomes that overlap those scored here (protocol section 2.3, reproduced later in this note). The read is therefore a registered re-read of a previously analyzed split under a rule fixed before the read. The protocol (\texttt{MIMIC\_\allowbreak{}IV\_\allowbreak{}READ\_\allowbreak{}PROTOCOL\_\allowbreak{}20260930\_\allowbreak{}V1.md}), the result note (\texttt{MIMIC\_\allowbreak{}IV\_\allowbreak{}READ\_\allowbreak{}RESULT\_\allowbreak{}20261002\_\allowbreak{}V1.md}) and the release review (\texttt{MIMIC\_\allowbreak{}IV\_\allowbreak{}READ\_\allowbreak{}RELEASE\_\allowbreak{}REVIEW\_\allowbreak{}20261002\_\allowbreak{}V1.md}) are in \texttt{reports/\allowbreak{}revision\_\allowbreak{}v3/\allowbreak{}}. The 41 shipped aggregate files are in \texttt{source\_\allowbreak{}tables/\allowbreak{}revision\_\allowbreak{}v3/\allowbreak{}mimic4\_\allowbreak{}read\_\allowbreak{}6f9c4a6\_\allowbreak{}13172\_\allowbreak{}v1/\allowbreak{}}. A script (\texttt{make\_\allowbreak{}si\_\allowbreak{}note7\_\allowbreak{}tables.py}) wrote every table of this note from these files and from the committed files named in the captions, and it checks the counts and outcomes stated in this text against them. The power and precision table, the disclosure table and the lists of deviations and limitations are converted from the text of the protocol. As the protocol requires (sections 3.9 and 13), the provenance comes first, then the mechanism records, then the outcomes.

\paragraph{Design in brief.} Family A used three landmark and horizon pairs, $(L,H)=(12,6)$, $(18,12)$ and $(30,24)$, and three targets, heart rate, respiratory rate and \spo{}, which gave 9 settings. The arterial-saturation target was not scored, because in MIMIC-IV that channel holds sparse blood-gas values. The synchronized forecast (S2) read hours 1 to $L$ and forecast a first crossing of the target's threshold in the next $H$ hours. The primary comparator (S1) was the forecast of the same window that the same frozen model made at hour 6. The registered secondary comparator was the patient's last measured value of the target at hour $L$ (LOCF). A descriptive comparator (S1-adm) was the hour-6 forecast of a crossing within $H$ hours of hour 6. Inputs were standardized with MIMIC-IV training statistics, so each threshold was the MIMIC-IV training mean of its target. Rosters required a record reaching hour 30. The estimate was the five-seed mean AUROC difference, with 10,000 whole-person bootstrap replicates, two-sided add-one $P$ values and Holm correction over the 9 settings, separately for S2 $-$ S1 (A1) and S2 $-$ LOCF (A2). A setting survived when its Holm-adjusted $P$ was below 0.05, its estimate was positive and most replicates lay on the side of its estimate. Each comparison received one of four outcome classes (protocol section 3.7): criterion met in all 9 settings; criterion met in 5 to 8 settings with none significantly worse before correction (\texttt{A\_MAJORITY}); criterion not met (\texttt{A\_NOT\_SHOWN}); and criterion not met with at least one setting worse after Holm correction (\texttt{A\_CONTRADICTED}). A2 used the same classes with the prefix \texttt{A2}.

Family B refitted four arms within the 66,007 stays of the MIMIC-IV training split. The comparator $b$ was the fixed CatBoost recipe on summary features of the record, fitted once with tree seed 13. The encoder $f$ was initialized from the pretrained twin and fine-tuned on the endpoints, and $g$ was the same architecture trained from random initialization, each with seeds 13, 17, 23, 29 and 31. The hybrid $h$ was the CatBoost recipe on the features of $b$ and the fine-tuned state of $f$. Endpoints were scored within 24 hours of landmarks at 12, 24 and 48 hours: hypotension (E2) and acute kidney injury by the KDIGO creatinine criterion (E3) as primary endpoints, and hypoxemia (E4) as a registered secondary endpoint. Death in the unit was not scored as an endpoint, because the harmonized MIMIC-IV records keep a hospital-death flag without its time. The primary test was one-sided for $h-b$, the five-seed mean of $h$ minus $b$, on E2 and E3, with 10,000 whole-person bootstrap replicates, add-one $P$ values and Holm correction over the two endpoints. Intervals are two-sided 95\% percentile intervals, so a gain significant at the second Holm step can have a lower interval bound at or below zero.

For the disclosed secondary outcomes, the frozen checkpoints were read at hour 48 in stays whose record reached hour 48 and whose unit stay lasted beyond it. The death forecast $P_D$ was the largest of the four target-specific death cumulative incidences at 48 hours. The exit forecast $P_X(H)$ was the largest, over the four heads, of the cumulative incidence of leaving the unit, alive or dead, before the head's target crossed its threshold, for $H=6$, 12, 24 and 48 hours. The outcomes were hospital death in a hospitalization whose unit stay ended within 48 hours of hour 48 (MORT-W), hospital death at any time in the hospitalization (MORT-H) and leaving the unit within $H$ hours of hour 48 (LOS-X$H$). Their intervals are 95\% whole-person bootstrap intervals of the five-seed mean (10,000 replicates) and Student $t$ intervals across the five seeds (4 degrees of freedom).

\paragraph{Provenance and execution.} The protocol was committed alone as commit P (\texttt{3ad11ead}, tag \texttt{mimic4-\allowbreak{}read-\allowbreak{}protocol-\allowbreak{}v1}). The scholar then gave a dated acknowledgment of it (decision item 20 of \texttt{reports/\allowbreak{}revision\_\allowbreak{}v3/\allowbreak{}ROUTE\_\allowbreak{}B\_\allowbreak{}DECISION\_\allowbreak{}20260926\_\allowbreak{}V1.md}). The code, tests and launchers were committed next as package commit A, a sequence of seven commits ending at \texttt{80c86ec7}. Manifest job 13123 ran from A and listed 14,211 test stays of 9,805 persons. No test person had a training or validation stay, and the two split columns agreed in every MIMIC-IV row. Commit C2 (\texttt{6f9c4a62}, tag \texttt{mimic4-\allowbreak{}read-\allowbreak{}freeze-\allowbreak{}v1}) then froze the protocol by pinning these manifest values. Registry commit B (\texttt{3c58d87d}) entered the two rows of the test-read registry and the run row before any read was submitted. The first chain of jobs from B (13138 to 13148) stayed pending in the queue and was cancelled before any stage started, so nothing was read. The read then ran from relaunch commit D1 (\texttt{0e8b093b}) and registry commit B$'$1 (\texttt{52b5eef7}) as run \texttt{52b5eef7a7f8\_\allowbreak{}20261002t120818z\_\allowbreak{}mimic4\_\allowbreak{}read} (Supplementary Table~\ref{tab:m4-mech}a). Every job and array task ended with state COMPLETED and exit code 0:0. Each stage was written to run its native tests before touching data and to stop on a failure. The stage logs were not shipped, so this rests on the exit codes and on the launcher. A shipping script written after the protocol (\texttt{scripts/\allowbreak{}revision\_\allowbreak{}v3\_\allowbreak{}mimic4\_\allowbreak{}read\_\allowbreak{}ship\_\allowbreak{}v1.py}, commit \texttt{a29db2a3}) copied the 41 allowlisted aggregate files after scanning them against every stay and person identifier of the test manifest and of the MIMIC-IV training split. No stay-level file left the cluster or was committed. A release review by a separate agent with read-only access re-applied the decision rules of protocol sections 3.6, 3.7, 4.4 and 4.5 to the shipped contrasts, recomputed the Holm adjustments with a maximum difference of 0.0 and found no correctness error. The reviewing agent belongs to the same model family as the agent that built and ran the read, so the review is independent in context only.

\paragraph{Release, freeze, labels and verification.} Stage R released the occurrence of measurements in the test split, without outcomes, predictions or metrics, under an authorization that binds the 34 disclosed earlier entries, 27 of them scientific reads of the split, for the full split of 14,211 stays of 9,805 persons. Stage FZ recorded the SHA-256 digest of every prediction file before any label existed (freeze digest beginning \texttt{2f480e66}). Stage LB verified those digests without parsing a prediction value and then built the labels; its receipt records neither counts nor metrics. No test stay was a rescued zero-measurement stay. Stage SC wrote one read-once marker and scored the labels. Stage VF read the labels under that marker, wrote no marker and reproduced 185,380 values of the scored outputs at a tolerance of $10^{-12}$. VF calls the same scoring function as SC, so its agreement shows that the scoring is deterministic; it does not test an independent implementation.

\subsection*{Mechanism records}

The protocol required three descriptive records to be reported before the outcomes. They have no decision role and are stated here as recorded.

\paragraph{Family A.} The roster of every setting held 0.986 to 1.055 times the number of stays projected from the MIMIC-IV training census, inside the registered bounds of 0.5 to 2 (record M-A1; Supplementary Table~\ref{tab:m4-mech}b). No prediction was exactly 0 or 1, and the target was unmeasured at hour 6 in 2.7\% to 6.3\% of roster stays (M-A4). Between 66.7\% and 84.8\% of roster stays had rows after the terminal clock (M-A5). This is the disclosed deviation D15: labels stop at the terminal clock, and a record can reach hour 30 through such rows. The per-seed AUROCs of S2, S1 and LOCF (M-A2) are given with the other per-seed values in Supplementary Table~\ref{tab:m4-a-seed}. S2 exceeded S1 in every seed of every setting. Record M-A3 sets the observed-hour fractions of 12 input channels in test hours 1 to 30 beside the training-census fractions over whole records, as registered (Supplementary Table~\ref{tab:m4-mech}c). The two windows differ, so these ratios do not estimate a shift between the test and training splits.

The input shift that the frozen model meets in MIMIC-IV was described before the read by the MIMIC-IV training census (job 13062), which compared the observed-hour fractions of the 41 physiological channels in seeded samples of MIMIC-IV and eICU-CRD training stays (Supplementary Table~\ref{tab:m4-mech}d; protocol deviation D16). The arterial-saturation channel, which in eICU-CRD combines the bedside monitor's periodic saturation readings with blood-gas values, was observed in 0.012 of MIMIC-IV hours against 0.492 of eICU-CRD hours. Heart rate and non-invasive blood pressure were observed less often in MIMIC-IV, and most laboratory values more often. No arm of the read varied the input mapping, so neither this record nor any outcome attributes a result to the shift.

\paragraph{Family B.} The numbers of test rows with known status and of events at each endpoint and landmark lay between 0.986 and 1.012 and between 0.919 and 1.114 times their projections from the training census (Supplementary Table~\ref{tab:m4-mech}e). The bootstrap population was the 9,623 test persons with a test row. The training fit sets of $b$ and $h$ are given in Supplementary Table~\ref{tab:m4-mech}f. The physical-unit guard passed on the training and test splits, with channel medians that differed by at most 0.5 units between the splits (Supplementary Table~\ref{tab:m4-mech}g). All ten encoder runs clipped the gradient of every batch, 212 of 212 batches in every epoch at a clip norm of 1.0 (Supplementary Table~\ref{tab:m4-mech}h). In the four eICU-CRD encoder runs of probe P1, every batch was clipped as well, 360 of 360 in every epoch (\texttt{source\_\allowbreak{}tables/\allowbreak{}revision\_\allowbreak{}v3/\allowbreak{}route\_\allowbreak{}b\_\allowbreak{}p1\_\allowbreak{}result\_\allowbreak{}13032\_\allowbreak{}v1/\allowbreak{}train/\allowbreak{}}). The pretrained encoder $f$ stopped at the 15-epoch cap with its best epoch at the cap for seeds 13, 17 and 23, and on the patience rule for seeds 29 and 31. The scratch encoder $g$ stopped on the patience rule for four seeds and at the cap for seed 13, whose best epoch was 14. In probe P1 on eICU-CRD, one of the two pretrained-encoder runs (seed 29) also ended at the cap. No training receipt records access to the test or validation split.

\paragraph{Disclosed secondary outcomes.} Of the 14,211 manifest stays, 6,951 stays of 5,396 persons entered the population (Supplementary Table~\ref{tab:m4-mech}i). The main-text Results describe them as stays still in the unit after hour 48. The registered entry rule also requires a record reaching hour 48, so it excludes 65 stays still in the unit whose record ends before hour 48; the main-text Methods (Forecast evaluation) state the rule. One third of the hospital deaths (MORT-H events) occurred in a hospitalization whose unit stay ended within 48 hours of hour 48 (MORT-W events). No exit forecast needed the sum correction of protocol section 3.11. The head that attains the maximum of the death forecast varies by seed (Supplementary Table~\ref{tab:m4-mech}j).

\begingroup
\medskip
\noindent\begin{minipage}{\textwidth}
\captionof{table}{\textbf{Execution and mechanism records of the registered MIMIC-IV read.} Panel a gives the jobs of the read. Panels b to j give the descriptive mechanism records that the protocol required before the outcomes: family A and the input shift that the frozen model meets (b to d), family B (e to h) and the disclosed secondary outcomes (i and j). Settings $(L,H)$: read to hour $L$, forecast the next $H$ hours. Files named without a path are in \texttt{source\_\allowbreak{}tables/\allowbreak{}revision\_\allowbreak{}v3/\allowbreak{}mimic4\_\allowbreak{}read\_\allowbreak{}6f9c4a6\_\allowbreak{}13172\_\allowbreak{}v1/}.}
\label{tab:m4-mech}
\pnl{a}{\textbf{Jobs of the read.} Run \texttt{52b5eef7a7f8\_\allowbreak{}20261002t120818z\_\allowbreak{}mimic4\_\allowbreak{}read}, launched from registry commit B$'$1 (\texttt{52b5eef7}); the manifest stage M ran earlier from package commit A. Stages with several tasks ran as Slurm job arrays, and their elapsed time is the range over tasks. State: Slurm accounting state and exit code. Source: \texttt{job\_\allowbreak{}log\_\allowbreak{}summary.json}.}
{\small
\begin{tabularx}{\textwidth}{l >{\raggedright\arraybackslash}X r r l l}
\toprule
Stage & Role & Job & Tasks & Elapsed (h:min:s) & State \\
\midrule
\rows{supp_tables/s2x_jobs.tex}
\bottomrule
\end{tabularx}\par}
\end{minipage}

\bigskip
\noindent\begin{minipage}{\textwidth}
\pnl{b}{\textbf{Family A: rosters, predictions and the terminal clock.} Roster stays against the projection from the MIMIC-IV training census (record M-A1; registered bounds 0.5 to 2); share of roster stays whose target was unmeasured at hour 6 (M-A4); share of roster stays with rows after the terminal clock (M-A5); share of prediction values equal to exactly 0 or 1 (M-A4). Source: \texttt{mechanism\_a.json}.}
\centering\small
\begin{adjustbox}{max width=\textwidth}\begin{tabular}{l r r r r r r}
\toprule
Setting & Roster stays & Projected stays & Ratio & \shortstack{Target unmeasured\\at hour 6 (\%)} & \shortstack{Rows after the\\terminal clock (\%)} & \shortstack{Predictions at\\0 or 1 (share)} \\
\midrule
\rows{supp_tables/s2x_a_mech.tex}
\bottomrule
\end{tabular}\end{adjustbox}
\end{minipage}

\bigskip
\noindent\begin{minipage}{\textwidth}
\pnl{c}{\textbf{Family A, record M-A3: observed-hour fractions.} Fraction of hours in which each input channel was observed in test hours 1 to 30, identical across seeds, beside the fraction over whole records in the MIMIC-IV training census, as registered. The two windows differ, so the ratio does not estimate a shift between the test and training splits. Source: \texttt{mechanism\_a.json}.}
\centering\small
\begin{tabular}{l r r r}
\toprule
Channel & Test, hours 1 to 30 & Training census, whole records & Ratio \\
\midrule
\rows{supp_tables/s2x_a_ma3.tex}
\bottomrule
\end{tabular}
\end{minipage}

\bigskip
\noindent\begin{minipage}{\textwidth}
\pnl{d}{\textbf{Input shift seen by the frozen model.} Fraction of hours in which each of the 41 physiological input channels was observed, and its median standardized value with each database standardized by its own training statistics, in seeded samples of MIMIC-IV and eICU-CRD training stays, computed before the read by the MIMIC-IV training census (job 13062). In eICU-CRD the \spo{} channel holds nurse-charted pulse oximetry and the SaO\textsubscript{2} channel combines the bedside monitor's periodic saturation readings with blood-gas values; in MIMIC-IV the \spo{} channel holds hourly bedside pulse oximetry and the SaO\textsubscript{2} channel holds blood-gas values. No channel was remapped for the frozen model (protocol choice C9). Source: \texttt{source\_\allowbreak{}tables/\allowbreak{}revision\_\allowbreak{}v3/\allowbreak{}mimic4\_\allowbreak{}train\_\allowbreak{}census\_\allowbreak{}32de53f\_\allowbreak{}13062\_\allowbreak{}v1/\allowbreak{}c3\_\allowbreak{}input\_\allowbreak{}shift\_\allowbreak{}channels.csv}.}
\centering\footnotesize
\begin{tabular}{l r r r r}
\toprule
Channel & \shortstack{Observed hours,\\MIMIC-IV} & \shortstack{Observed hours,\\eICU-CRD} & Ratio & \shortstack{Median $z$,\\MIMIC-IV / eICU-CRD} \\
\midrule
\rows{supp_tables/s2x_a_shift.tex}
\bottomrule
\end{tabular}
\end{minipage}

\bigskip
\noindent\begin{minipage}{\textwidth}
\pnl{e}{\textbf{Family B: test rows against the projection.} Test rows with known status and events per endpoint and landmark against the projection from the MIMIC-IV training census, and the persons contributing rows. Source: \texttt{mechanism\_b.json}.}
\centering\footnotesize
\begin{adjustbox}{max width=\textwidth}\begin{tabular}{l r r r r r r r r}
\toprule
Endpoint & Landmark (h) & Known rows & Projected & Ratio & Events & Projected & Ratio & Persons \\
\midrule
\rows{supp_tables/s2x_b_rows.tex}
\bottomrule
\end{tabular}\end{adjustbox}
\end{minipage}

\bigskip
\noindent\begin{minipage}{\textwidth}
\pnl{f}{\textbf{Family B: training fit sets.} Rows, persons and events of the training fit sets of $b$ and $h$, identical for every tree fit of an endpoint. Source: \texttt{fit\_\allowbreak{}receipt\_\allowbreak{}*.json}.}
\centering\footnotesize
\begin{tabular}{l r r r}
\toprule
Endpoint & Training fit rows & Persons & Events \\
\midrule
\rows{supp_tables/s2x_b_fit.tex}
\bottomrule
\end{tabular}
\end{minipage}

\bigskip
\noindent\begin{minipage}{\textwidth}
\pnl{g}{\textbf{Family B: physical-unit guard.} Median of each channel in physical units and fraction of invalid measured hours on the training and test splits (family B stops if more than 1\% of measured hours are invalid). Sources: \texttt{mechanism\_b.json} and \texttt{label\_\allowbreak{}receipt.json}.}
\centering\footnotesize
\begin{tabular}{l r r r r}
\toprule
Channel & Training median & Training invalid & Test median & Test invalid \\
\midrule
\rows{supp_tables/s2x_b_guard.tex}
\bottomrule
\end{tabular}
\end{minipage}

\bigskip
\noindent\begin{minipage}{\textwidth}
\pnl{h}{\textbf{Family B: training histories of the encoders.} Ten runs on the MIMIC-IV training split: $f$, initialized from the pretrained eICU-CRD twin, and $g$, random initialization, each with seeds 13, 17, 23, 29 and 31. AdamW, at most 15 epochs, early stopping with a patience of 2 epochs on the mean binary cross-entropy of the inner split; the rows by epoch give that value, and $^{*}$marks the best epoch, whose weights were kept. Cap, stopped at the 15-epoch cap; patience, stopped by the patience rule. The 54,102 fitting and 5,936 early-stopping stays are the 60,038 of the 66,007 training-split stays that have at least one landmark row at risk with known status for hypotension, acute kidney injury or hypoxemia (\texttt{train\_\allowbreak{}stays\_\allowbreak{}with\_\allowbreak{}labels} in each receipt). Sources: \texttt{mechanism\_b.json} and \texttt{train\_\allowbreak{}receipt\_\allowbreak{}*.json}.}
\centering\footnotesize
\begin{adjustbox}{max width=\textwidth}\begin{tabular}{l r r r r r r r r r r}
\toprule
 & \multicolumn{5}{c}{$f$ (pretrained), seed} & \multicolumn{5}{c}{$g$ (random initialization), seed} \\
\cmidrule(lr){2-6}\cmidrule(lr){7-11}
 & 13 & 17 & 23 & 29 & 31 & 13 & 17 & 23 & 29 & 31 \\
\midrule
\rows{supp_tables/s2x_b_train.tex}
\bottomrule
\end{tabular}\end{adjustbox}
\end{minipage}

\bigskip
\noindent\begin{minipage}{\textwidth}
\pnl{i}{\textbf{Disclosed secondary outcomes: population and events,} identical across seeds. Source: \texttt{mechanism\_\allowbreak{}sec.json}.}
\centering\footnotesize
\begin{tabular}{l r}
\toprule
Quantity & Value \\
\midrule
\rows{supp_tables/s2x_sec_pop.tex}
\bottomrule
\end{tabular}
\end{minipage}

\bigskip
\noindent\begin{minipage}{\textwidth}
\pnl{j}{\textbf{Disclosed secondary outcomes: records by seed.} Exit-forecast sums above 1 by at most $10^{-6}$ that were set to 1; percentiles of the death forecast $P_D$ and of the 48-hour exit forecast $P_X(48)$; and the share of stays in which each target-specific head attains the maximum that defines each forecast. Source: \texttt{mechanism\_\allowbreak{}sec.json}.}
\centering\footnotesize
\begin{adjustbox}{max width=\textwidth}\begin{tabular}{l r r r r r}
\toprule
 & \multicolumn{5}{c}{Seed} \\
\cmidrule(lr){2-6}
Quantity & 13 & 17 & 23 & 29 & 31 \\
\midrule
\rows{supp_tables/s2x_sec_seedmech.tex}
\bottomrule
\end{tabular}\end{adjustbox}
\end{minipage}
\par\endgroup

\begingroup
\footnotesize
\setlength{\LTcapwidth}{\textwidth}
\begin{longtable}{l l r r r r r r}
\caption{\textbf{Family A AUROCs by seed.} S2, synchronized forecast; S1, the model's hour-6 forecast of the same window (primary comparator); S1-adm, the hour-6 forecast of a crossing within $H$ hours of hour 6 (descriptive); LOCF, the patient's last measured value of the target at hour $L$, which does not depend on the seed. The last two columns are the per-seed differences whose five-seed means are the estimates of A1 and A2. Hour-30 rosters and MIMIC-IV training-mean thresholds. Source: \texttt{metrics\_a.csv}; the per-seed AUROCs of S2, S1 and LOCF equal record M-A2 of \texttt{mechanism\_a.json}.}
\label{tab:m4-a-seed}\\
\toprule
Setting & Seed & S2 & S1 & S1-adm & LOCF & S2 $-$ S1 & S2 $-$ LOCF \\
\midrule
\endfirsthead
\multicolumn{8}{l}{\textit{Supplementary Table~\thetable, continued}} \\
\toprule
Setting & Seed & S2 & S1 & S1-adm & LOCF & S2 $-$ S1 & S2 $-$ LOCF \\
\midrule
\endhead
\bottomrule
\endfoot
\rows{supp_tables/s2x_a_seed.tex}
\end{longtable}
\endgroup

\clearpage
\subsection*{Outcomes}

\paragraph{Summary.} The read returned \texttt{A\_MAJORITY} for family A, \texttt{A2\_CONTRADICTED} for its registered secondary comparison and \texttt{B\_E3\_ONLY} for family B, and the disclosed secondary outcomes were scored. No flag was raised, and no setting or endpoint was undefined. For this combination of outcomes, the interpretation guide fixed in protocol section 3.9 reads: ``the frozen model still gains from reading the newer hours in MIMIC-IV, but its advantage over the patient's last measured value is shown on eICU only.'' The isolation arms were not repeated on MIMIC-IV, so the attribution of the gain to the patient's newly recorded measurements rests on the eICU-CRD validation split.

\paragraph{Family A.} The synchronized forecast outperformed the hour-6 forecast in 8 of 9 settings after Holm correction (estimates $+0.025$ to $+0.114$ AUROC; not significant in \spo{} (30, 24)). This meets the registered transfer criterion of at least 5 of 9 settings with none significantly worse before correction, but unlike on eICU-CRD the gain was not significant in every setting (Supplementary Table~\ref{tab:m4-a}a). In \spo{} (30, 24) the estimate was $+0.029$ (95\% interval $-0.021$ to $+0.077$; Holm-adjusted $P=0.26$), and no setting was significantly worse before or after correction. Eight of the nine estimates were smaller than the eICU-CRD estimate for the same setting; heart rate (18, 12) was the exception. Against the patient's last measured value, the synchronized forecast was better in 1 of 9 settings after Holm correction, respiratory rate (12, 6), and worse in 2, \spo{} (12, 6) and heart rate (30, 24), with a median difference of $-0.011$ (range $-0.084$ to $+0.034$ AUROC). The three registered sensitivities gave the same outcome classes with the same numbers of surviving settings: per-cell rosters, thresholds at the eICU-CRD training means and the stay-level bootstrap of the eICU-CRD analysis (Supplementary Table~\ref{tab:m4-a-sens}). Under the eICU-CRD-mean thresholds, \spo{} (12, 6) remained worse than the last measured value after Holm correction, and heart rate (30, 24) was worse before correction only. Because the stay-level bootstrap gave the same outcome classes as the primary analysis, the protocol's rule that a differing class be reported in the main-text Results did not apply. In raw probabilities, S2 had lower log-loss and Brier score than S1 in 8 of 9 settings and higher values in \spo{} (12, 6) (Supplementary Table~\ref{tab:m4-a}b). Nothing was calibrated on MIMIC-IV, whereas the eICU-CRD comparison in the main text uses log-loss and Brier score after isotonic calibration on eICU-CRD validation data, so the two are not on the same footing.

\begingroup
\medskip
\noindent\begin{minipage}{\textwidth}
\captionof{table}{\textbf{Family A on the MIMIC-IV test split: the registered contrasts and the descriptive contrasts.} Settings $(L,H)$: read to hour $L$, forecast the next $H$ hours. S2, synchronized forecast; S1, the model's hour-6 forecast of the same window; S1-adm, the hour-6 forecast of a crossing within $H$ hours of hour 6; LOCF, the patient's last measured value of the target at hour $L$. Files are in \texttt{source\_\allowbreak{}tables/\allowbreak{}revision\_\allowbreak{}v3/\allowbreak{}mimic4\_\allowbreak{}read\_\allowbreak{}6f9c4a6\_\allowbreak{}13172\_\allowbreak{}v1/}.}
\label{tab:m4-a}
\pnl{a}{\textbf{Synchronization (A1) and last-value comparison (A2).} Five-seed mean AUROC differences with 95\% intervals from 10,000 whole-person bootstrap replicates (stream seed 20260930), two-sided add-one $P$ values and Holm correction over the 9 settings within each family; the smallest attainable Holm-adjusted $P$ is about 0.0018. $^{*}$Survives (Holm-adjusted $P<0.05$, positive and direction-concordant). $^{\dagger}$Worse after Holm correction. Stays, persons and events: hour-30 roster. The last two columns give, for A1 and A2, the eICU-CRD test-set estimates of the same model head and, for \spo{}, of the SaO\textsubscript{2} head, whose eICU-CRD channel combines the bedside monitor's periodic saturation readings with blood-gas values (protocol section 3.2); neither pairing enters a decision. Source: \texttt{contrasts\_a.csv}.}
\centering\small
\begin{adjustbox}{max width=\textwidth}\begin{tabular}{l r r r l r l r l l}
\toprule
Setting & Stays & Persons & Events & \shortstack{A1: S2 $-$ S1\\{[95\% CI]}} & \shortstack{Holm\\$P$} & \shortstack{A2: S2 $-$ LOCF\\{[95\% CI]}} & \shortstack{Holm\\$P$} & \shortstack{eICU-CRD, same\\head: A1 / A2} & \shortstack{eICU-CRD, SaO\textsubscript{2}\\head: A1 / A2} \\
\midrule
\rows{supp_tables/s2x_a_primary.tex}
\bottomrule
\end{tabular}\end{adjustbox}
\end{minipage}

\bigskip
\noindent\begin{minipage}{\textwidth}
\pnl{b}{\textbf{Descriptive contrasts.} Threshold of the target in physical units (MIMIC-IV training mean; beats or breaths per minute, \spo{} in percent) and prevalence of crossings on the roster; S2 minus S1-adm; raw log-loss (probabilities clipped at $10^{-3}$) and raw Brier score of S2 minus S1; and the seed spread (largest minus smallest per-seed difference) of A1 and A2. Five-seed means with 95\% whole-person bootstrap intervals and no correction for multiple comparisons. Nothing was calibrated on MIMIC-IV. Sources: \texttt{contrasts\_a.csv} and \texttt{metrics\_a.csv}.}
\centering\footnotesize
\begin{adjustbox}{max width=\textwidth}\begin{tabular}{l r r l l l r r}
\toprule
Setting & Threshold & Prevalence & S2 $-$ S1-adm & Log-loss, S2 $-$ S1 & Brier, S2 $-$ S1 & \shortstack{Seed spread,\\A1} & \shortstack{Seed spread,\\A2} \\
\midrule
\rows{supp_tables/s2x_a_desc.tex}
\bottomrule
\end{tabular}\end{adjustbox}
\end{minipage}
\par\endgroup

\begingroup
\bigskip
\noindent\begin{minipage}{\textwidth}
\captionof{table}{\textbf{Family A: registered sensitivities.} Per-cell rosters (entry when the record reaches hour $L$); thresholds at the eICU-CRD training means applied to raw values; and the stay-level bootstrap of the eICU-CRD analysis, with $P$ values without the add-one correction, so that a value of 0 means that no replicate fell on the other side of zero. Each block ends with its outcome classes under the rules of protocol section 3.7; the sensitivities change no outcome. $^{*}$Survives; $^{\dagger}$worse after Holm correction; $^{\ddagger}$worse before correction only. Sources: \texttt{contrasts\_a.csv} and \texttt{outcomes.json}.}
\label{tab:m4-a-sens}
\centering\footnotesize
\begin{adjustbox}{max width=\textwidth}\begin{tabular}{l r r r l r l r}
\toprule
Setting & Stays & Persons & Events & A1: S2 $-$ S1 [95\% CI] & Holm $P$ & A2: S2 $-$ LOCF [95\% CI] & Holm $P$ \\
\midrule
\rows{supp_tables/s2x_a_sens.tex}
\bottomrule
\end{tabular}\end{adjustbox}
\end{minipage}
\par\endgroup

\paragraph{Family B.} Adding the fine-tuned state to the boosting features improved discrimination of acute kidney injury by $+0.0035$ (95\% CI $+0.0007$ to $+0.0065$; one-sided Holm-adjusted $P = 0.018$), which was significant under the registered test, but not significantly that of hypotension ($+0.0010$; $-0.0012$ to $+0.0033$; $P = 0.18$; Supplementary Table~\ref{tab:m4-b}a). No endpoint carried the flag for a hybrid worse than boosting. For hypoxemia, a registered secondary endpoint, the change was $+0.0009$ ($-0.0034$ to $+0.0054$; unadjusted one-sided $P = 0.35$). The test conditions on one tree seed of the comparator. Refitted with four further tree seeds, the comparator's AUROC varied by 0.0008 for hypotension, 0.0004 for acute kidney injury and 0.0014 for hypoxemia (Supplementary Table~\ref{tab:m4-b}b). With paired tree seeds, the hybrid's gain was $+0.0010$ ($-0.0012$ to $+0.0032$) for hypotension and $+0.0034$ ($+0.0006$ to $+0.0062$) for acute kidney injury. These seed analyses were registered as sensitivities and enter no test. The encoder trained from random initialization differed from boosting by $-0.0109$, $-0.0294$ and $-0.0197$ AUROC for hypotension, acute kidney injury and hypoxemia, and no non-inferiority margin was registered. The pretrained encoder was below the scratch encoder for hypotension and acute kidney injury, with intervals below zero, and differed from it by less than 0.0001 for hypoxemia ($f-g$). Per landmark, the hybrid's gain over boosting for acute kidney injury was $+0.0016$ at 12 hours, $+0.0030$ at 24 hours and $+0.0103$ at 48 hours (Supplementary Table~\ref{tab:m4-b}c). These are point differences of the five-seed means, without intervals. Every per-seed gain for acute kidney injury was positive, from $+0.0024$ to $+0.0045$ (Supplementary Table~\ref{tab:m4-b}d).

\begingroup
\medskip
\noindent\begin{minipage}{\textwidth}
\captionof{table}{\textbf{Family B on the MIMIC-IV test split.} Arms refitted within the MIMIC-IV training split: $b$, fixed CatBoost recipe (tree seed 13 unless stated); $f$, encoder initialized from the pretrained eICU-CRD twin and fine-tuned on the endpoints; $g$, the same architecture from random initialization; $h$, CatBoost on the features of $b$ and the fine-tuned state of $f$. Five-seed means of $f$, $g$ and $h$ (seeds 13, 17, 23, 29 and 31) unless stated. Intervals are 95\% two-sided percentile intervals from 10,000 whole-person bootstrap replicates (stream seed 20261001). Log-loss: raw probabilities, lower is better. Files are in \texttt{source\_\allowbreak{}tables/\allowbreak{}revision\_\allowbreak{}v3/\allowbreak{}mimic4\_\allowbreak{}read\_\allowbreak{}6f9c4a6\_\allowbreak{}13172\_\allowbreak{}v1/}.}
\label{tab:m4-b}
\pnl{a}{\textbf{AUROC contrasts.} The primary test was one-sided for $h-b$ on hypotension and acute kidney injury, with add-one $P$ values and Holm correction over the two endpoints; $^{*}$rejected. Hypoxemia was a registered secondary endpoint with an unadjusted one-sided $P$. The last column gives the eICU-CRD estimates of probe P1, which used seeds 23 and 29 and the tuning set of the eICU-CRD training split. Sources: \texttt{contrasts\_b.csv} and \texttt{mechanism\_b.json}.}
\centering\footnotesize
\begin{adjustbox}{max width=\textwidth}\begin{tabular}{l l r r l l}
\toprule
Contrast & AUROC difference [95\% CI] & One-sided $P$ & Holm $P$ & Log-loss difference [95\% CI] & eICU-CRD, probe P1 [95\% CI] \\
\midrule
\rows{supp_tables/s2x_b_contr.tex}
\bottomrule
\end{tabular}\end{adjustbox}
\end{minipage}

\bigskip
\noindent\begin{minipage}{\textwidth}
\pnl{b}{\textbf{AUROC and log-loss of every fitted arm.} AUROC with 95\% whole-person bootstrap interval, and raw log-loss, pooled over the three landmarks. The comparator $b$ was refitted with tree seeds 17, 23, 29 and 31 as a registered sensitivity, and $h$ was also fitted with the tree seed equal to its encoder seed (paired tree seeds). The means of seeds 23 and 29 follow the two-seed construction of probes P0 and P1 on eICU-CRD. Source: \texttt{metrics\_b.csv}.}
\centering\footnotesize
\begin{adjustbox}{max width=\textwidth}\begin{tabular}{l l r l r l r}
\toprule
 & \multicolumn{2}{c}{Hypotension (E2)} & \multicolumn{2}{c}{Acute kidney injury (E3)} & \multicolumn{2}{c}{Hypoxemia (E4, secondary)} \\
\cmidrule(lr){2-3}\cmidrule(lr){4-5}\cmidrule(lr){6-7}
Arm & AUROC [95\% CI] & Log-loss & AUROC [95\% CI] & Log-loss & AUROC [95\% CI] & Log-loss \\
\midrule
\rows{supp_tables/s2x_b_arms.tex}
\bottomrule
\end{tabular}\end{adjustbox}
\end{minipage}

\bigskip
\noindent\begin{minipage}{\textwidth}
\pnl{c}{\textbf{AUROC by landmark.} Pooled AUROC, AUROC at the landmarks of 12, 24 and 48 hours, and pooled raw log-loss with its 95\% whole-person interval. In the $h-b$ rows, the pooled columns give the registered contrast and its log-loss difference, and the landmark columns give point differences of the five-seed mean of $h$ and $b$, without intervals (descriptive). Sources: \texttt{metrics\_b.csv} and \texttt{contrasts\_b.csv}.}
\centering\footnotesize
\begin{adjustbox}{max width=\textwidth}\begin{tabular}{l r r r r l}
\toprule
Arm & Pooled AUROC & 12 h & 24 h & 48 h & Pooled log-loss [95\% CI] \\
\midrule
\rows{supp_tables/s2x_b_land.tex}
\bottomrule
\end{tabular}\end{adjustbox}
\end{minipage}

\bigskip
\noindent\begin{minipage}{\textwidth}
\pnl{d}{\textbf{Contrasts by encoder seed.} The arm of one seed minus $b$ (tree seed 13), with 95\% whole-person bootstrap intervals; descriptive. Source: \texttt{contrasts\_b.csv}.}
\centering\footnotesize
\begin{adjustbox}{max width=\textwidth}\begin{tabular}{l r l l l}
\toprule
Endpoint & Seed & $h_s-b$ [95\% CI] & $g_s-b$ [95\% CI] & $f_s-b$ [95\% CI] \\
\midrule
\rows{supp_tables/s2x_b_seed.tex}
\bottomrule
\end{tabular}\end{adjustbox}
\end{minipage}
\par\endgroup

\clearpage
\paragraph{Disclosed secondary outcomes.} The death forecast discriminated hospital death with unit exit within 48 hours (MORT-W) with an AUROC of 0.687 (95\% whole-person interval 0.657 to 0.718) and hospital death at any time in the hospitalization (MORT-H) with 0.726 (0.708 to 0.744). The exit forecast discriminated leaving the unit within 6 to 48 hours with AUROCs of 0.765 to 0.808 (Supplementary Table~\ref{tab:m4-sec}a). Every AUROC interval lay above 0.5. Every mean forecast lay below its observed rate, and every whole-person interval of the difference lay below zero. These shortfalls arise in part by construction: each target-specific head counts a death or a unit exit only if it precedes that target's crossing, whereas the outcomes count every death or exit in their windows, and both death outcomes also count deaths after a live unit exit. The shipped files do not separate the parts. Supplementary Table~\ref{tab:m4-sec}b gives every metric for every seed, with the five-seed means and their Student $t$ intervals. The logistic calibration intercept and slope of MORT-W and MORT-H are recorded as \texttt{NON\_\allowbreak{}ESTIMABLE\_\allowbreak{}SEPARATION} in every seed and as \texttt{NON\_\allowbreak{}ESTIMABLE\_\allowbreak{}IN\_\allowbreak{}ONE\_\allowbreak{}OR\_\allowbreak{}MORE\_\allowbreak{}SEEDS} for the five-seed mean. Why these fits failed has not been examined, and we state the codes without interpretation. Leaving out the arterial-saturation head, whose input channel is nearly empty in MIMIC-IV, changed no five-seed mean AUROC by more than 0.002 (Supplementary Table~\ref{tab:m4-sec}c). The reliability tables of every seed are in Supplementary Table~\ref{tab:m4-sec}d.

\paragraph{Earlier reads of the same outcomes.} Hospital death and length of stay had been scored on this split in many earlier reads (protocol section 2.3, reproduced later in this note). As the protocol requires, the committed results at a landmark of 48 hours of the two corrected-clock reads TRB-M4-20 and TRB-M4-21, whose models were trained on MIMIC-IV, are given beside this read (Supplementary Table~\ref{tab:m4-sec}e,f). TRB-M4-20 scored hospital death with nine tabular models among stays with an ICU length of stay of at least 48 hours on a corrected clock. This read scored the same outcome with the frozen eICU-CRD model on the legacy clock, among stays whose terminal clock lies after hour 48 and whose record reaches hour 48 (protocol deviation D29). The size of this population is given above. Protocol deviation D29 also stated that the scorer would count its overlap with the TRB-M4-20 stays, and the shipped aggregates contain no such count. TRB-M4-21 regressed the remaining ICU length of stay and reports mean absolute errors in hours. The unit-exit outcomes of this read are thresholds of the remaining stay, so they cannot be compared with those errors (protocol deviation D30). The earlier results are context, and no comparison between the reads was made or tested.

\begingroup
\medskip
\noindent\begin{minipage}{\textwidth}
\captionof{table}{\textbf{Disclosed secondary outcomes of the MIMIC-IV read and earlier reads of the same outcomes.} Panels a to d: frozen eICU-CRD checkpoints read at hour 48 in 6,951 stays of 5,396 persons; death forecast $P_D$ for the two death outcomes and exit forecast $P_X(H)$ for leaving the unit within $H$ hours of hour 48. No criterion was registered. Files of panels a to d are in \texttt{source\_\allowbreak{}tables/\allowbreak{}revision\_\allowbreak{}v3/\allowbreak{}mimic4\_\allowbreak{}read\_\allowbreak{}6f9c4a6\_\allowbreak{}13172\_\allowbreak{}v1/}. Panels e and f: committed results at a landmark of 48 hours of two earlier disclosed reads of the MIMIC-IV test split by models trained on MIMIC-IV (corrected clock; stays with an ICU length of stay of at least 48 hours). Their populations, clocks and models differ from those of this read, and no comparison is made or tested.}
\label{tab:m4-sec}
\pnl{a}{\textbf{Discrimination and mean forecast.} AUROC: five-seed mean with 95\% whole-person bootstrap interval (10,000 replicates of the five-seed mean) and 95\% Student $t$ interval across seeds (4 degrees of freedom). Mean forecast and AUPRC: five-seed means with $t$ intervals. The difference between the mean forecast and the observed rate has a whole-person interval. Sources: \texttt{metrics\_sec.csv} and \texttt{outcomes.json}.}
\centering\footnotesize
\begin{adjustbox}{max width=\textwidth}\begin{tabular}{l r r r l l l l l}
\toprule
Outcome & Events & \shortstack{Observed\\rate} & AUROC & \shortstack{Whole-person\\95\% interval} & \shortstack{$t$ interval\\across seeds} & \shortstack{Mean forecast\\{[$t$ interval]}} & \shortstack{Mean forecast minus\\observed rate [95\% CI]} & \shortstack{AUPRC\\{[$t$ interval]}} \\
\midrule
\rows{supp_tables/s2x_sec_main.tex}
\bottomrule
\end{tabular}\end{adjustbox}
\end{minipage}

\bigskip
\noindent\begin{minipage}{\textwidth}
\pnl{b}{\textbf{Every metric by seed.} Per-seed AUROC, AUPRC, log-loss (probabilities clipped at $10^{-7}$), Brier score, mean forecast, and the intercept and slope of a logistic regression of the outcome on the logit of the forecast (0 and 1 under perfect calibration), with five-seed means and 95\% Student $t$ intervals across seeds. $^{\mathrm{a}}$Calibration status recorded as \texttt{NON\_\allowbreak{}ESTIMABLE\_\allowbreak{}SEPARATION}. $^{\mathrm{b}}$Recorded as \texttt{NON\_\allowbreak{}ESTIMABLE\_\allowbreak{}IN\_\allowbreak{}ONE\_\allowbreak{}OR\_\allowbreak{}MORE\_\allowbreak{}SEEDS}. The codes are stated without interpretation. Source: \texttt{metrics\_sec.csv}.}
\centering\scriptsize
\setlength{\tabcolsep}{3pt}
\begin{tabular}{l l l l l l l l}
\toprule
 & AUROC & AUPRC & Log-loss & Brier & Mean forecast & Intercept & Slope \\
\midrule
\rows{supp_tables/s2x_sec_seed.tex}
\bottomrule
\end{tabular}
\end{minipage}

\bigskip
\noindent\begin{minipage}{\textwidth}
\pnl{c}{\textbf{Three-head sensitivity.} Both summaries recomputed over the heart-rate, respiratory-rate and \spo{} heads, leaving out the arterial-saturation head because its input channel is nearly empty in MIMIC-IV (registered sensitivity, descriptive). Five-seed means with 95\% Student $t$ intervals; the second column repeats the four-head AUROC. $^{\mathrm{b}}$As in panel b. Source: \texttt{metrics\_sec.csv}.}
\centering\footnotesize
\begin{adjustbox}{max width=\textwidth}\begin{tabular}{l r l l l l l}
\toprule
Outcome & \shortstack{AUROC,\\four heads} & \shortstack{AUROC,\\three heads} & AUPRC & Mean forecast & Intercept & Slope \\
\midrule
\rows{supp_tables/s2x_sec_three.tex}
\bottomrule
\end{tabular}\end{adjustbox}
\end{minipage}

\bigskip
\noindent\begin{minipage}{\textwidth}
\pnl{d}{\textbf{Reliability tables.} Ten equal-width bins of the four-head forecast in each seed; bins empty in every seed are omitted. Each cell gives the number of stays, the mean forecast and the observed rate; --, empty bin. The reliability tables of the three-head sensitivity are in the source file. Source: \texttt{metrics\_sec.csv}.}
\centering\scriptsize
\begin{adjustbox}{max width=\textwidth}\begin{tabular}{l l l l l l l}
\toprule
 & & \multicolumn{5}{c}{Seed: stays; mean forecast; observed rate} \\
\cmidrule(lr){3-7}
Outcome & Forecast bin & 13 & 17 & 23 & 29 & 31 \\
\midrule
\rows{supp_tables/s2x_sec_rel.tex}
\bottomrule
\end{tabular}\end{adjustbox}
\end{minipage}

\bigskip
\noindent\begin{minipage}{\textwidth}
\pnl{e}{\textbf{Earlier read TRB-M4-20: hospital death.} Nine tabular models; raw probabilities. Calibration-in-the-large: mean prediction minus outcome rate. The population, clock and models differ from those of this read (protocol deviation D29). Source: \texttt{source\_\allowbreak{}tables/\allowbreak{}revision\_\allowbreak{}v3/\allowbreak{}g3/\allowbreak{}g3\_\allowbreak{}mortality\_\allowbreak{}heldout\_\allowbreak{}scoreboard\_\allowbreak{}v1.csv}, rows \texttt{mimic4}, context 48, raw surface (report \texttt{reports/\allowbreak{}revision\_\allowbreak{}v3/\allowbreak{}G3\_\allowbreak{}MORTALITY\_\allowbreak{}HELDOUT\_\allowbreak{}FRONTIER\_\allowbreak{}V1.md}).}
\centering\footnotesize
\begin{adjustbox}{max width=\textwidth}\begin{tabular}{>{\raggedright\arraybackslash}p{5.2cm} r r r r r r r}
\toprule
Model & AUROC & AUPRC & Log-loss & Brier & Intercept & Slope & \shortstack{Calibration-\\in-the-large} \\
\midrule
\rows{supp_tables/s2x_trb20.tex}
\bottomrule
\end{tabular}\end{adjustbox}
\end{minipage}

\bigskip
\noindent\begin{minipage}{\textwidth}
\pnl{f}{\textbf{Earlier read TRB-M4-21: remaining ICU length of stay.} Mean absolute error, in hours, of the remaining ICU length of stay predicted by 13 models. This read scored binary unit-exit outcomes instead, which are not comparable with these errors (protocol deviation D30). Source: \texttt{source\_\allowbreak{}tables/\allowbreak{}revision\_\allowbreak{}v3/\allowbreak{}g3/\allowbreak{}g3\_\allowbreak{}los\_\allowbreak{}heldout\_\allowbreak{}frontier\_\allowbreak{}v1/\allowbreak{}scoreboard.csv}, rows \texttt{mimic4}, context 48 (report \texttt{reports/\allowbreak{}revision\_\allowbreak{}v3/\allowbreak{}G3\_\allowbreak{}LOS\_\allowbreak{}HELDOUT\_\allowbreak{}FRONTIER\_\allowbreak{}V1.md}).}
\centering\footnotesize
\begin{tabular}{>{\raggedright\arraybackslash}p{9cm} r r}
\toprule
Model & MAE (h) & Features \\
\midrule
\rows{supp_tables/s2x_trb21.tex}
\bottomrule
\end{tabular}
\end{minipage}
\par\endgroup

\clearpage
\subsection*{Power and precision, computed before the read}

The protocol computed the power of each registered outcome before the read (section 5); the scripts and their outputs are committed in \texttt{reports/\allowbreak{}revision\_\allowbreak{}v3/\allowbreak{}mimic4\_\allowbreak{}read\_\allowbreak{}power\_\allowbreak{}v1/\allowbreak{}}. Standard errors on eICU-CRD were taken from the widths of the registered intervals. They were rescaled by the event and non-event counts projected for the MIMIC-IV test split and, for family A, by an approximate design effect for repeat stays of one person. True effects were the eICU-CRD estimates multiplied by a retained fraction. The simulation drew 200,000 runs with standard errors 1 or 1.5 times the projection (family B also 1.25 times) and applied the registered Holm procedures and outcome rules. Supplementary Table~\ref{tab:m4-power}a gives the inputs for family A and Supplementary Table~\ref{tab:m4-power}b its results, including a heterogeneous case in which heart and respiratory rate keep their eICU-CRD gains and the three \spo{} settings share a common true gain. Supplementary Table~\ref{tab:m4-power}c,d gives the inputs and results for family B. The disclosed secondary outcomes had no criterion and therefore no power analysis; protocol section 5.5 stated their expected precision (Supplementary Table~\ref{tab:m4-power}e). The protocol also recorded the author's forecast of each outcome before the read (section 6).

\begingroup
\medskip
\noindent\begin{minipage}{\textwidth}
\captionof{table}{\textbf{Power and precision of the registered MIMIC-IV read, computed before the read (protocol section 5).} Standard errors (SE) on eICU-CRD were derived from the widths of the registered intervals and projected to the MIMIC-IV test split. Panels a, b, d and e are converted from the protocol; panel c is read from the committed power inputs.}
\label{tab:m4-power}
\pnl{a}{\textbf{Family A inputs (protocol section 5.2).} eICU-CRD estimates with their SE, and the SE projected for the MIMIC-IV test split.}
\centering\footnotesize
\begin{tabular}{l l r l r}
\toprule
\rows{supp_tables/s2x_pow_a_in.tex}
\bottomrule
\end{tabular}
\end{minipage}

\bigskip
\noindent\begin{minipage}{\textwidth}
\pnl{b}{\textbf{Family A results (protocol section 5.3).} Each cell gives the probabilities of the four outcome classes in the order: criterion met in all 9 settings / met in 5 to 8 settings with none significantly worse before correction / not met / not met with at least one setting worse after correction. Columns: multiple of the projected SE and correlation $\rho$ between settings. Rows of the first two blocks: fraction of the eICU-CRD gain retained in every setting. Rows of the last block (heterogeneous case): the common true gain of the three \spo{} settings, with heart and respiratory rate at their full eICU-CRD gains; after the semicolon, the probability of a failed outcome with at least 6 surviving settings. 200,000 simulated runs.}
\centering\footnotesize
\begin{adjustbox}{max width=\textwidth}\begin{tabular}{l l l l l}
\toprule
\rows{supp_tables/s2x_pow_a.tex}
\bottomrule
\end{tabular}\end{adjustbox}
\end{minipage}

\bigskip
\noindent\begin{minipage}{\textwidth}
\pnl{c}{\textbf{Family B inputs (protocol section 5.4).} eICU-CRD events in the tuning rows of probe P1, events projected for the MIMIC-IV test split, and the eICU-CRD estimates of $h-b$ and $g-b$ with their SE and the projected SE. Source: \texttt{reports/\allowbreak{}revision\_\allowbreak{}v3/\allowbreak{}mimic4\_\allowbreak{}read\_\allowbreak{}power\_\allowbreak{}v1/\allowbreak{}family\_\allowbreak{}b\_\allowbreak{}inputs.csv}.}
\centering\footnotesize
\begin{adjustbox}{max width=\textwidth}\begin{tabular}{l r r l r l r}
\toprule
Endpoint & \shortstack{eICU-CRD\\events} & \shortstack{Projected\\events} & \shortstack{eICU-CRD $h-b$\\(SE)} & \shortstack{Projected\\SE} & \shortstack{eICU-CRD $g-b$\\(SE)} & \shortstack{Projected\\SE} \\
\midrule
\rows{supp_tables/s2x_pow_b_in.tex}
\bottomrule
\end{tabular}\end{adjustbox}
\end{minipage}

\bigskip
\noindent\begin{minipage}{\textwidth}
\pnl{d}{\textbf{Family B results (protocol section 5.4).} Probabilities of rejecting hypotension / rejecting acute kidney injury / rejecting both (\texttt{B\_BOTH}) / rejecting neither (\texttt{B\_NONE}), with correlation 0.3 between the endpoints, by the fraction of the eICU-CRD gain retained and the multiple of the projected SE.}
\centering\footnotesize
\begin{tabular}{l l l l}
\toprule
\rows{supp_tables/s2x_pow_b.tex}
\bottomrule
\end{tabular}
\end{minipage}

\bigskip
\noindent\begin{minipage}{\textwidth}
\pnl{e}{\textbf{Expected precision of the disclosed secondary outcomes (protocol section 5.5).} Hanley-McNeil standard error of an AUROC and binomial standard error of a rate, both multiplied by a design effect for repeat stays of one person, computed before the read from the committed counts of TRB-M4-20; 95\% half-width, 1.96 standard errors.}
\centering\footnotesize
\begin{adjustbox}{max width=\textwidth}\begin{tabular}{l l l l}
\toprule
\rows{supp_tables/s2x_prec.tex}
\bottomrule
\end{tabular}\end{adjustbox}
\end{minipage}
\par\endgroup

\clearpage
\subsection*{Earlier reads of the MIMIC-IV test split (protocol section 2.3)}

Protocol section 2.3 is reproduced here as written before the read. Its list (Supplementary Table~\ref{tab:m4-disclosure}) is the exact set of 34 identifiers that the release authorization of stage R bound. Section numbers in the reproduced text refer to the protocol. Registry line numbers in the reproduced text (written \texttt{:66}, \texttt{:142-\allowbreak{}158} and similar) refer to \texttt{reports/\allowbreak{}revision\_\allowbreak{}v3/\allowbreak{}TEST\_\allowbreak{}READ\_\allowbreak{}REGISTRY.md} at commit P (\texttt{3ad11ead}). Registry commit B (\texttt{3c58d87d}) later added two rows above the disclosure section of that file. From commit B on, every cited line from \texttt{:66} onward is therefore two higher (TRB-M4-01 is at line 68 and TRB-M4-20 at line 87), and \texttt{:14} and \texttt{:24} are unchanged.

\begingroup\sloppy
\rows{supp_tables/s2x_disclosure_intro.tex}
\endgroup

\begingroup
\scriptsize
\setlength{\LTcapwidth}{\textwidth}
\setlength{\tabcolsep}{4pt}
\begin{longtable}{>{\raggedright\arraybackslash}p{1.6cm} >{\raggedright\arraybackslash}p{2.9cm} >{\raggedright\arraybackslash}p{2.6cm} >{\raggedright\arraybackslash}p{1.9cm} >{\raggedright\arraybackslash}p{6.0cm}}
\caption{\textbf{Earlier reads of the MIMIC-IV test split (protocol section 2.3).} The canonical disclosure list as written before the read. Record: line of \texttt{reports/\allowbreak{}revision\_\allowbreak{}v3/\allowbreak{}TEST\_\allowbreak{}READ\_\allowbreak{}REGISTRY.md} at commit P (written \texttt{:66}; two higher from registry commit B on) or another record. Converted from the protocol.}
\label{tab:m4-disclosure}\\
\toprule
ID & Record & Jobs & Date & What was read on the MIMIC-IV test split \\
\midrule
\endfirsthead
\multicolumn{5}{l}{\textit{Supplementary Table~\thetable, continued}} \\
\toprule
ID & Record & Jobs & Date & What was read on the MIMIC-IV test split \\
\midrule
\endhead
\bottomrule
\endfoot
\rows{supp_tables/s2x_disclosure.tex}
\end{longtable}
\endgroup

\begingroup\sloppy
\rows{supp_tables/s2x_disclosure_after.tex}
\endgroup

\subsection*{Deviations}

Protocol section 8 (deviations and disclosures) is reproduced in full below, as written before the read. The protocol gives it as a table of 32 rows, D1 to D32. Each item below gives the three cells of one row: the earlier decision, plan or analysis (From), what the read does (This read) and the consequence stated in the protocol (Effect). In it, C1 to C12 are the design choices of protocol section 0.3, items refer to the scholar's decision record (\texttt{reports/\allowbreak{}revision\_\allowbreak{}v3/\allowbreak{}ROUTE\_\allowbreak{}B\_\allowbreak{}DECISION\_\allowbreak{}20260926\_\allowbreak{}V1.md}), and other numbers refer to protocol sections.

\begingroup\sloppy\small
\rows{supp_tables/s2x_deviations_list.tex}
\endgroup

The following deviations and corrections are not in protocol section 8, whose text has not changed since commit P. Items 1 to 8 are listed in section 4 of the result note. Items 2, 3, 5 and 7 arose after commit P and before the freeze at C2, and items 1, 4 and 6 arose after the freeze. Item 8 concerns the design of the verification stage. Item 9 arose after the read.

\begingroup\sloppy
\begin{enumerate}
\item \textit{Relaunch.} The first chain of jobs (13138 to 13148) was cancelled before any stage started. The read ran under route 1 of protocol section 7.7 from relaunch commit D1 (\texttt{0e8b093b}), which changed only the dispatcher's resource block. The memory requests of stages BX, R, BP, LB, SC and VF fell from 64 GB to 48 GB, against measured peaks of 0.9 to 12.3 GB in the analogous earlier jobs.
\item \textit{Job steps.} Package commits A6 and A7 made the dispatcher run the clone and checkout (A6) and its remote checks and submissions (A7) inside Slurm job steps, because the login session's disk writes were throttled. This changed only where the commands ran.
\item \textit{Fixed width of 48 hours.} Family B built the encoder states at a fixed width of 48 hours instead of the variable width of probe P1, which moves predictions by about $10^{-7}$. The scholar acknowledged this as a deviation from protocol section 2.1 (decision item 21).
\item \textit{Dispatcher tests after D1.} On the tree after D1, \texttt{scripts/\allowbreak{}test\_\allowbreak{}revision\_\allowbreak{}v3\_\allowbreak{}mimic4\_\allowbreak{}read\_\allowbreak{}dispatch\_\allowbreak{}v1.py} fails 11 tests (9 failures and 2 class-setup errors). All of them pin the literal 64 GB values of the registered stage table, which D1 changed by design. Route 1 allows no other change, so the tests were not edited. The read's own stages were written to run their native tests first and to stop on a failure. Every stage ended with exit code 0:0. Their logs were not shipped, so the passing of these tests rests on the exit codes and the launcher (Provenance and execution, above). The release and common modules pass their 92 tests on the back-filled tree.
\item \textit{Reused code without a run of its tests.} Stage BF reuses \texttt{fit\_\allowbreak{}fixed\_\allowbreak{}catboost} of \texttt{revision\_\allowbreak{}v3\_\allowbreak{}landmark\_\allowbreak{}trajectory\_\allowbreak{}core\_\allowbreak{}v1}, whose tests no stage runs because the cluster environment lacks pytest (validation record, item 38).
\item \textit{Shipping.} The shipping script that copied the aggregates was written after the protocol (Provenance and execution, above).
\item \textit{Validation record.} The validation record pinned at C2 (\texttt{reports/\allowbreak{}revision\_\allowbreak{}v3/\allowbreak{}MIMIC\_\allowbreak{}IV\_\allowbreak{}READ\_\allowbreak{}VALIDATION\_\allowbreak{}RECORD\_\allowbreak{}V1.md}) contains two stale statements, which are corrected here because the pinned record is not edited. Its line 13 says that no cluster job was run, whereas its items 54 and 55 cite the infrastructure-test jobs 13104, 13105 and 13118, which read no data. Its item 1 says that the package pin names A5, whereas the pin is A7 (\texttt{80c86ec7}).
\item \textit{Verification.} Stage VF calls the same scoring function as stage SC (\texttt{score\_all}). The verification therefore shows that the scoring is deterministic; it does not test an independent implementation.
\item \textit{Revision of the manuscript.} Decision item 22 of the decision record (\texttt{reports/\allowbreak{}revision\_\allowbreak{}v3/\allowbreak{}ROUTE\_\allowbreak{}B\_\allowbreak{}DECISION\_\allowbreak{}20260926\_\allowbreak{}V1.md}) records the scholar's decisions on the revision of the manuscript after the lab's editorial review, taken on 2 and 7 October 2026 with the read's outcomes known. It also records the departures from protocol section 9 that these decisions cause. The paper has no contribution list, so the registered family-B contribution clause has no host. In a later revision of the manuscript for clarity and length, decided by the scholar on 20 October 2026 (for the Abstract, on 6 November 2026), the registered sentences of this read were reworded for readability and some were moved, for example from Results to Methods; every number, outcome and caveat is unchanged, and the registered wording is given in the read's \texttt{outcomes.json} (\texttt{source\_\allowbreak{}data/\allowbreak{}Supplementary\_\allowbreak{}Tables\_\allowbreak{}S20\_\allowbreak{}S27/\allowbreak{}mimic4\_\allowbreak{}read\_\allowbreak{}6f9c4a6\_\allowbreak{}13172\_\allowbreak{}v1/\allowbreak{}outcomes.json}), which the read's result note (\texttt{MIMIC\_\allowbreak{}IV\_\allowbreak{}READ\_\allowbreak{}RESULT\_\allowbreak{}20261002\_\allowbreak{}V1.md}) names.
\end{enumerate}
\endgroup

\subsection*{Limitations stated in advance (protocol section 10)}

Protocol section 10 is reproduced in full as written before the read. In it, plain section numbers refer to the protocol, C6 and C12 to the design choices of protocol section 0.3, and D16, D29 and D30 to items of protocol section 8 (Deviations, above). TRB-M4, REG-C01 and PREPROC identifiers refer to rows of Supplementary Table~\ref{tab:m4-disclosure}, and decision items to the scholar's decision record (\texttt{reports/\allowbreak{}revision\_\allowbreak{}v3/\allowbreak{}ROUTE\_\allowbreak{}B\_\allowbreak{}DECISION\_\allowbreak{}20260926\_\allowbreak{}V1.md}).

\begingroup\sloppy
\rows{supp_tables/s2x_limitations.tex}
\endgroup

\clearpage
\section*{Supplementary Note 8: Earlier transfer analyses on MIMIC-III}
\addcontentsline{toc}{section}{Supplementary Note 8: Earlier transfer analyses on MIMIC-III}

\paragraph{Scope and status.} No result of the main text uses MIMIC-III, but its test split was read in 21 earlier campaigns between May and September 2026. This note puts those reads on record and reports how earlier versions of the model transferred to this second database. Each read is disclosed in the test-read registry (rows TRB-M4-01, TRB-M4-05 to TRB-M4-16 and TRB-M4-18 to TRB-M4-25; Supplementary Table~\ref{tab:m3reads}). None had a registry row before it took place, so all 21 are reported as disclosed and not registered. Three of them (TRB-M4-19 to TRB-M4-21) followed rules fixed in their own records before the read. Because those rules were not entered in the registry, we treat the three reads like the others. Twelve of the reads concerned models or encoders trained on eICU-CRD and applied to MIMIC-III with or without fine-tuning. The other nine concerned models trained within MIMIC-III, a semi-synthetic dataset built on it, or early runs recorded only as output directories. No listed read applied the model of the main text to MIMIC-III. We report numbers only for reads whose aggregate tables are committed to the project repository, which will be released with the code (Code availability of the main text), and list the other reads without results.

\paragraph{Cohort and split.} The MIMIC-III cohort comprises 28,394 CareVue-era stays of 23,692 patients who do not appear in MIMIC-IV (Supplementary Table~S18). They were split by person into a training split of 20,007 stays of 16,661 patients, a validation split, and a test split of 4,244 stays of 3,531 patients. The earlier campaigns selected the test split through a legacy split column. For MIMIC-III this column equals the person-level column by construction, although no direct recount was made.

\paragraph{Models and readouts.} Apart from the withdrawn comparison of TRB-M4-05 (below), the reported transfer cells used A0\_dyn, an earlier version of the model with time-to-event, competing-risk and dynamics heads. It was trained on eICU-CRD with three seeds (13, 17 and 23) under a legacy marginal-status objective. None of these cells used the corrected competing-risks model of the main text. The cells were scored with the legacy readout on the legacy endpoints. These included the harmonized mortality label (hospital death) and length of stay at landmarks of 6, 12, 24 and 48 hours, and three threshold events (systolic blood pressure below 90 mmHg, \spo{} below 90\% and heart rate above 130 beats per minute). The endpoints also included incident onsets of acute kidney injury, hypotension, hypoxemia, high lactate and a sepsis proxy within 24 hours of the same landmarks. Since 30 August 2026 the project's evaluator has refused to score cumulative risk from models trained with this objective, so the cells run in September used the June 2026 evaluator (commit fc1e613). Their numbers are therefore comparable with each other but not with any result of the main text.

\paragraph{Zero-shot transfer and fine-tuning.} The models trained on eICU-CRD discriminated the mortality label on MIMIC-III without any fine-tuning, and their discrimination improved as they were fine-tuned on growing shares of the MIMIC-III training split. Applied to the test split without fine-tuning (zero-shot transfer; TRB-M4-22), the three A0\_dyn models had mortality AUROCs of 0.762 to 0.808 over seeds and landmarks. Fine-tuned on 1\% of the MIMIC-III training split (TRB-M4-08), they reached a mean AUROC of 0.794 over seeds and landmarks, and fine-tuned on all of it, 0.864. Mean discrimination rose at every step of this fine-tuning ladder (1, 5, 10, 25, 50 and 100\% of the training split) for mortality, for the threshold events (AUROC 0.809 to 0.872) and for the incident onsets (0.653 to 0.702). Over the same steps, however, the expected calibration error of the mortality forecast rose from 0.059 to 0.078.

\paragraph{Per-patient weight adaptation.} Each A0\_dyn transfer cell also scored the same model with its weights updated online for each patient. The main text kept the twin's weights frozen after a separate eICU-CRD evaluation, in which online adaptation of the forecasting heads of an earlier model version did not improve forecasts (Supplementary Fig.~S1). In the reported transfer cells on MIMIC-III, no improvement from the updates survived Holm correction either. Mechanism records show that the updates ran at every step, with no rollback and no non-finite update, in the zero-shot cells and in the crossed cells described below. On the fine-tuning ladder, none of the 216 differences between the adapted and the unadapted model on MIMIC-III and MIMIC-IV survived Holm correction within the campaign, in either direction. Without correction, 32 of the 108 differences on MIMIC-III had intervals across seeds that excluded zero. Eleven of them favored the adapted model, including threshold-event discrimination at 1\% of the training data (AUROC $+0.0028$; 95\% interval across seeds 0.0017 to 0.0040) and the onset log-loss, Brier score or calibration error at fractions from 1\% to 25\%. Twenty-one favored the unadapted model, among them the mean absolute error of length of stay from 10\% of the training data on ($+0.15$ to $+0.22$ hours), the Brier score of the mortality forecast from 10\% on ($+0.0004$ to $+0.0006$) and threshold-event discrimination at 25\% (AUROC $-0.0011$). In the zero-shot cells, none of the 36 differences on the two databases survived Holm correction. On MIMIC-III, three differences had intervals across seeds that excluded zero without correction, and all three favored the adapted model (threshold-event AUROC $+0.0043$ and AUPRC $+0.0117$, onset log-loss $-0.1497$). In the crossed cells (TRB-M4-23), a KL-anchored objective replaced the dynamics objective of these updates on models fine-tuned on 1\% and 100\% of the training split. The onset log-loss difference then changed sign, from $-0.1499$ and $-0.0889$ under the dynamics objective at the same fractions of the fine-tuning ladder to $+0.0477$ and $+0.0803$, with intervals including zero. The expected calibration error of the threshold events rose by 0.0090 at both fractions, and both rises survived Holm correction over the 36 differences of the MIMIC-III crossed cells.

\paragraph{Pretraining against random initialization.} TRB-M4-05 compared a model pretrained on eICU-CRD, with its head fine-tuned on MIMIC-III, against the same model trained from random initialization. The pretrained model discriminated the mortality label with AUROC 0.874 to 0.888 at the four landmarks. We withdraw the comparison because the model trained from random initialization failed to train: its AUROC was 0.572 at 24 hours and 0.197 at 48 hours on MIMIC-III, and at or below chance in five of the 12 mortality cells across the three databases.

\begingroup\sloppy
\paragraph{Tabular predictors trained within MIMIC-III.} Two corrected-clock reads (TRB-M4-20 and TRB-M4-21) scored gradient-boosted and other tabular models fitted on the training data of each database. Neither read involved an ICare-DT model. At 48 hours, the MIMIC-III test split held 2,289 stays of 1,930 patients with an ICU length of stay of at least 48 hours, and 353 of these stays had a positive mortality label (rate 0.154). Nine tabular models discriminated the mortality label with AUROC 0.878 to 0.882. For remaining length of stay, the read's candidate (\texttt{frozen\_\allowbreak{}oof\_\allowbreak{}simplex\_\allowbreak{}cell\_\allowbreak{}median\_\allowbreak{}calibrated\_\allowbreak{}v1}) was a weighted blend of gradient-boosted models whose weights were fixed on the training data before the test read. Its mean absolute error was 129.1 hours, against 177.2 hours for the training median. Five of the other 12 models of the read had lower errors, from 128.4 to 128.8 hours. The 48-hour MIMIC-IV cells of both reads are given in Supplementary Note~7 (Supplementary Table~\ref{tab:m4-sec}e,f).\par
\endgroup

\paragraph{Interpretation.} The transfer cells describe how earlier versions of the model, trained on eICU-CRD, behaved on a second database. Their discrimination of mortality, threshold events and incident onsets rose with the amount of MIMIC-III training data used for fine-tuning, while the calibration of the mortality forecast worsened. Per-patient weight adaptation gave no difference that survived Holm correction within its campaign, apart from a calibration harm under one objective. The reads do not test the model of the main text, its synchronization with frozen weights or any registered claim, and their endpoints and readout differ from those of the main text. The test split was read by many models, and four reads informed later choices of scope or comparators (TRB-M4-06, TRB-M4-10, TRB-M4-12 and TRB-M4-19). For these reasons, and because each transfer cell rests on three seeds, we draw no conclusion about MIMIC-III from these reads.

\begin{table}[h]
\centering
\small
\caption{\textbf{Earlier reads of the MIMIC-III test split.} Rows of the test-read registry (\texttt{reports/revision\_v3/TEST\_READ\_REGISTRY.md}) whose read included MIMIC-III test data, with dates and contents as recorded there. All are disclosed, and none had a row in this registry before its read. A0\_dyn and A0\_dyn\_thr are earlier model versions, trained on eICU-CRD except in TRB-M4-11, where A0\_dyn was trained on MIMIC-III. In the Results column, ``Note~8'' marks results reported in this note from committed aggregate tables, ``Internal notes'' marks results recorded only in the project's working notes (\texttt{icare\_dt/RESULTS.md}, \texttt{icare\_dt/CAMPAIGN.md}) and not reported, and ``Outputs only'' marks reads recorded only as output directories.}
\label{tab:m3reads}
\begin{tabularx}{\textwidth}{>{\raggedright\arraybackslash}p{1.7cm} >{\raggedright\arraybackslash}p{2.3cm} >{\raggedright\arraybackslash}X >{\raggedright\arraybackslash}p{1.9cm}}
\toprule
Registry row & Date (2026) & Analysis & Results \\
\midrule
TRB-M4-01 & 23 to 27 May & Baselines and early model variants: mortality, length of stay and legacy onsets & Outputs only \\
TRB-M4-05 & 7 to 14 June & eICU-pretrained model with its head fine-tuned on each database, against the same model trained from random initialization; conventional baselines; mortality and length of stay at 6 to 48 hours & Note~8 (withdrawn) \\
TRB-M4-06 & 15 June & Outcome-regression head of an earlier model version on each database; future values of the regression targets & Internal notes \\
TRB-M4-07 & June & Frozen eICU-trained A0\_dyn applied without fine-tuning, with and without online weight adaptation & Internal notes \\
TRB-M4-08 & June & A0\_dyn fine-tuned on 1\% to 100\% of the MIMIC-III training split, with and without online weight adaptation & Note~8 \\
TRB-M4-09 & 3 July & XGBoost, LSTM, GRU-D and Transformer trained, calibrated and tested within each database; 28 tasks & Internal notes \\
TRB-M4-10 & July & The same baselines and LightGBM and CatBoost trained on eICU-CRD and tested on MIMIC & Internal notes \\
TRB-M4-11 & July & A0\_dyn trained on MIMIC-III, with and without online weight adaptation & Internal notes \\
TRB-M4-12 & July & LightGBM and CatBoost trained within each database; six-model aggregate & Internal notes \\
TRB-M4-13 & July & Frozen eICU-trained A0\_dyn\_thr (threshold head) applied without fine-tuning, with and without online weight adaptation & Internal notes \\
TRB-M4-14 & July & Re-read of transfer predictions by risk, error, stay length and pre-onset strata & Internal notes \\
TRB-M4-15 & 10 July & Re-read of the TRB-M4-13 predictions by stratum & Internal notes \\
TRB-M4-16 & 10 July & The TRB-M4-13 models with a threshold-calibration adaptation; closed negative & Internal notes \\
TRB-M4-18 & From 11 July & LightGBM on the frozen states of three eICU-trained encoders; mortality, length of stay, lactate and sepsis onset at 24 hours & Internal notes \\
TRB-M4-19 & 21 to 23 July & Shared low-rank readout on the same frozen encoders and tasks; selected on validation, and its test result closed a follow-up & Internal notes \\
TRB-M4-20 & 26 August & Tabular predictors of the mortality label among stays with a length of stay of at least 6, 12, 24 or 48 hours (corrected clock); no ICare-DT model & Note~8 \\
TRB-M4-21 & 26 August & Tabular predictors of remaining ICU length of stay (corrected clock); no ICare-DT model & Note~8 \\
TRB-M4-22 & 19 September & Frozen eICU-trained A0\_dyn applied without fine-tuning (three seeds), with and without online weight adaptation; June 2026 evaluator & Note~8 \\
TRB-M4-23 & 19 to 20 September & A0\_dyn fine-tuned on 1\% and 100\% of the MIMIC-III training split, adapted online with a KL-anchored objective; June 2026 evaluator & Note~8 \\
TRB-M4-24 & 23 to 25 May & Online personalization configurations D1 and D3 of an early model version, run per database & Outputs only \\
TRB-M4-25 & 22 to 23 May & Counterfactual evaluations on a semi-synthetic dataset built on MIMIC-III & Outputs only \\
\bottomrule
\end{tabularx}
\end{table}

\clearpage
\section*{\raggedright Supplementary Note 9: Registered analyses of the washout panels added before submission}
\addcontentsline{toc}{section}{Supplementary Note 9: Registered analyses of the washout panels added before submission}

\subsection*{A. Realism audit of the washout panels}

\paragraph{Realism audit of the washout panels.} Placebo copies, in which the memory-route response is switched off but the simulated dosing and its direct and mediated routes remain, gave audit AUC 0.712 to 0.723 (largest difference from the panels 0.004, in the replication panel; 95\% interval of that paired difference $-0.005$ to 0.009), and the audit separated the panels from their own placebo copies with AUC 0.508 to 0.510, above its label-swap null band in the discovery, replication and fixed-schedule panels. Paired comparisons of the same stays detected systematic differences in the written channels in every panel (Hotelling's test, P < 0.05). In the panels, the lag-1 autocorrelation of 1 to 2 of the 12 outcome channels fell more than 0.02 below that of the real records. Near the threshold the verdict depends on sampling and on the assignment of stays to folds: at an AUC of 0.75 with 400 stays per group the Hanley-McNeil standard error is about 0.017, and Supplementary Table~S29 gives each panel's bootstrap interval and the range over 100 random fold assignments.

\paragraph{Wording.} The protocol fixed the sentences that report this audit. In the revision of the manuscript for clarity and length (20 October 2026; for the Abstract, 6 November 2026), some of these sentences were reworded for readability, shortened in figure captions or moved, for example from Results to Methods; every number, outcome and caveat is kept in the paper, and the registered wording is given in section 7 of the audit's result note (\texttt{WASHOUT\_\allowbreak{}PANEL\_\allowbreak{}REALISM\_\allowbreak{}AUDIT\_\allowbreak{}RESULT\_\allowbreak{}20261014\_\allowbreak{}V1.md}).

\paragraph{Compute.} The registered run (\texttt{06b23b75ed33\_\allowbreak{}20261014t202905z\_\allowbreak{}washout\_\allowbreak{}realism\_\allowbreak{}audit}) executed its six stages as six chained Slurm jobs, 13464 to 13469, one per stage, on 14 October 2026; they were submitted by a dispatch job that cloned the pinned commit, which the shipped accounting summary does not cover. The scheduler's accounting, as summarized in the shipped file \texttt{job\_\allowbreak{}log\_\allowbreak{}summary.json}, records elapsed times of 1\,min 45\,s for S0 (job 13464; authentication, pairing, channel contract and census), 6\,min 4\,s for S1 (job 13465; construction of the 26 record sets and the plumbing check G1), 1\,min 24\,s for S2 (job 13466; the three primary audits), 5\,min 51\,s for S3 (job 13467; the other 21 audits and the plateau diagnostics), 2\,min 33\,s for S4 (job 13468; the descriptive refits, with 16 parallel workers) and 1\,min 24\,s for S5 (job 13469; rendering, identifier scan and shipping). Each stage also ran its native test suite before its entry point (result note, section 2.1). Jobs 13464 to 13468 ended in state COMPLETED with exit code 0:0. The summary was written by S5 while S5 was still running, so S5's row records the state RUNNING and the elapsed time at that moment; S5's stage receipt records its completion. The stage receipts record completion times from 20:32:43 (S0) to 20:50:01 UTC (S5). Of the scheduler's accounting, the shipped summary carries elapsed time, state and exit code only; it records no CPU time, so none is reported here. No stage stopped the run.

\ifnum\value{table}=28\relax\else
  \typeout{WARNING (Supplementary Note 9): table counter was \the\value{table} before Supplementary Table S29; set to 28.}
  \setcounter{table}{28}
\fi
\begingroup
\setlength{\tabcolsep}{4pt}
\medskip
\noindent\begin{minipage}{\textwidth}
\captionof{table}{\textbf{Realism audits of the washout panels.} Every audit of the registered protocol \texttt{WASHOUT\_\allowbreak{}PANEL\_\allowbreak{}REALISM\_\allowbreak{}AUDIT\_\allowbreak{}PROTOCOL\_\allowbreak{}20261008\_\allowbreak{}V1.md} as amended by \texttt{WASHOUT\_\allowbreak{}PANEL\_\allowbreak{}REALISM\_\allowbreak{}AUDIT\_\allowbreak{}PROTOCOL\_\allowbreak{}20261011\_\allowbreak{}V2.md} (tag \texttt{washout-\allowbreak{}realism-\allowbreak{}audit-\allowbreak{}protocol-\allowbreak{}v2}), with its descriptive outputs. The panels were audited after their other analyses had been run, including the exploratory construction campaign from which the registered cell was chosen; the registered runs of the in-support control and the g-computation comparator followed the audit. Each audit applies the locked realism audit unchanged to two sets of 400 records cut to the same window, hours 24 to 163 of the stay (hours 24 to 113 for A1r): the 644 summaries of 161 channels (per-stay mean, standard deviation, missing fraction and lag-1 autocorrelation), a five-fold ridge logistic discriminator grouped by stay, and five checks, audit AUC at most 0.75, median absolute standardized mean difference ($|$SMD$|$) at most 0.25, its 90th percentile at most 0.50, median absolute missingness gap at most 0.05 and label match at least 0.999. The label tables of the constructed record sets carried no outcome label: they held the stay identifier, source database, age, sex and split, so the label check matches exactly by construction. Precision stated before the run (protocol section 7.3): at an audit AUC of 0.75 with 400 real and 400 generated records the Hanley-McNeil standard error is about 0.017; it treats the two groups as independent, and the paired design and the identical coordinates change the true value. On the probe of the generated dataset the pair-bootstrap standard deviation was 0.0129, 30 random pair-to-fold assignments moved the audit from 0.630 to 0.646, and the shipped 0.641 was reproduced only as 0.630 in another environment (Supplementary Note~5, Proposition~3). The audit assigns stays to folds deterministically from the sorted stay identifiers, without shuffling. A panel whose audit AUC lies within about 0.03 of 0.75 can therefore pass or fail by sampling and by the fold assignment; the locked rule still decides; panel b measures both sources, and its NEAR\_THRESHOLD flag reports the case. Panel a lists every audit; panels b to e give the registered descriptive outputs and the plateau diagnostic, none of which has a decision role; panel f gives values that are not audit values; two earlier realism reads of an earlier washout design are disclosed below the panels. The protocol, the result note (\texttt{WASHOUT\_\allowbreak{}PANEL\_\allowbreak{}REALISM\_\allowbreak{}AUDIT\_\allowbreak{}RESULT\_\allowbreak{}20261014\_\allowbreak{}V1.md}) and its release review (\texttt{WASHOUT\_\allowbreak{}PANEL\_\allowbreak{}REALISM\_\allowbreak{}AUDIT\_\allowbreak{}RELEASE\_\allowbreak{}REVIEW\_\allowbreak{}20261014\_\allowbreak{}V1.md}) are in \texttt{reports/\allowbreak{}revision\_\allowbreak{}v3/}. Files named without a path are in \texttt{source\_\allowbreak{}tables/\allowbreak{}revision\_\allowbreak{}v3/\allowbreak{}washout\_\allowbreak{}realism\_\allowbreak{}audit\_\allowbreak{}d68e077\_\allowbreak{}13469\_\allowbreak{}v1/}.}
\label{tab:q4-realism}
\end{minipage}

\bigskip
\noindent\begin{minipage}{\textwidth}
\pnl{a}{\textbf{Every audit.} The row above each group states its construction. G1 compares the discovery roster's real windows with an exact copy of themselves and is exact by construction. G2 is the discovery panel generated without the measurement hold, so that every hour receives a fresh value, under A1; it and CN15 are the sensitivity gates (detected: audit AUC above 0.75). CN05 to CN15 add independent noise to the real windows at each real measurement of each of the 12 outcome channels, held until the next one; 0.15 reference SD equals the panels' own noise term. CS025 to CS10 raise heart rate by 0.25, 0.5 and 1 reference SD on every finite cell. A1, the primary audit, combines the panel's generated outcome values, set to missing wherever the real record is missing, with the real record for every other channel, the treatment channels included. A1r is A1 on window hours 0 to 89, which the estimators read. A2 is the panel as stored, with filled cells and the simulated dosing as its treatment channels, completed with the real record's other channels. A3 is A1 built from the placebo copy, in which the memory-route response is switched off and the simulated dosing and its other routes remain; AP takes the placebo copy's A3 record as real against the panel's A1 record. B1 is the generated dataset that passed the audit, on the panel stays and windows, one audit per roster. The fixed-schedule panel uses the discovery roster. Rows: hours per stay in the window. Pass: all five checks. Every report was written and valid (binding and construction checks passed), and no audit was left not run. The median and 90th-percentile $|$SMD$|$ and the median missingness gap are taken over the 161 channels and cannot register a difference confined to the 12 written channels, so the discriminator is the operative check. Category \texttt{Q4\_\allowbreak{}PASS\_\allowbreak{}INSENSITIVE}: the three A1 audits passed, G2 was detected and CN15 was not. Sources: \texttt{outcomes.json} (\texttt{table\_\allowbreak{}s29}), \texttt{audits/*/realism\_\allowbreak{}report.json} and \texttt{calibration.csv}.}
\centering\footnotesize
\begin{tabular}{l l r r r l r r r r l}
\toprule
Audit & Panel or roster & Stays & Rows & \shortstack{Audit\\AUC} & Validity & \shortstack{Median\\$|$SMD$|$} & \shortstack{90th pct.\\$|$SMD$|$} & \shortstack{Median\\missing gap} & \shortstack{Label\\match} & Pass \\
\midrule
\multicolumn{11}{@{}l}{\textit{G1, plumbing check: the real windows against an exact copy}} \\
G1 & discovery roster & 400 & 140 & 0.500 & valid & 0 & 0 & 0 & 1 & yes \\
\addlinespace[3pt]
\multicolumn{11}{@{}l}{\textit{G2, gate: the discovery panel written at every hour, under A1}} \\
G2 & discovery roster & 400 & 140 & 0.959 & valid & 0 & 0 & 0 & 1 & no (detected) \\
\addlinespace[3pt]
\multicolumn{11}{@{}l}{\textit{CN05, CN10, CN15: measurement-time noise of 0.05, 0.10 and 0.15 reference SD (CN15 a gate)}} \\
CN05 & discovery roster & 400 & 140 & 0.513 & valid & 0 & 0 & 0 & 1 & yes \\
CN10 & discovery roster & 400 & 140 & 0.555 & valid & 0 & 0 & 0 & 1 & yes \\
CN15 & discovery roster & 400 & 140 & 0.609 & valid & 0 & 0 & 0 & 1 & yes (not detected) \\
\addlinespace[3pt]
\multicolumn{11}{@{}l}{\textit{CS025, CS05, CS10: heart rate raised by 3.75, 7.5 and 15 beats per minute}} \\
CS025 & discovery roster & 400 & 140 & 0.550 & valid & 0 & 0 & 0 & 1 & yes \\
CS05 & discovery roster & 400 & 140 & 0.599 & valid & 0 & 0 & 0 & 1 & yes \\
CS10 & discovery roster & 400 & 140 & 0.699 & valid & 0 & 0 & 0 & 1 & yes \\
\addlinespace[3pt]
\multicolumn{11}{@{}l}{\textit{A1, primary: generated outcome values, real missing cells, real other channels}} \\
A1 & discovery & 400 & 140 & 0.714 & valid & 0 & 0 & 0 & 1 & yes \\
A1 & replication & 400 & 140 & 0.727 & valid & 0 & 0 & 0 & 1 & yes \\
A1 & fixed schedule & 400 & 140 & 0.720 & valid & 0 & 0 & 0 & 1 & yes \\
\addlinespace[3pt]
\multicolumn{11}{@{}l}{\textit{A1r: A1 on window hours 0 to 89}} \\
A1r & discovery & 400 & 90 & 0.737 & valid & 0 & 0 & 0 & 1 & yes \\
A1r & replication & 400 & 90 & 0.713 & valid & 0 & 0 & 0 & 1 & yes \\
A1r & fixed schedule & 400 & 90 & 0.738 & valid & 0 & 0 & 0 & 1 & yes \\
\addlinespace[3pt]
\multicolumn{11}{@{}l}{\textit{A2: the panel as stored}} \\
A2 & discovery & 400 & 140 & 0.970 & valid & 0 & 0.014 & 0 & 1 & no \\
A2 & replication & 400 & 140 & 0.978 & valid & 0 & 0.021 & 0 & 1 & no \\
A2 & fixed schedule & 400 & 140 & 0.968 & valid & 0 & 0.019 & 0 & 1 & no \\
\addlinespace[3pt]
\multicolumn{11}{@{}l}{\textit{A3: the placebo copy under A1}} \\
A3 & discovery & 400 & 140 & 0.712 & valid & 0 & 0 & 0 & 1 & yes \\
A3 & replication & 400 & 140 & 0.723 & valid & 0 & 0 & 0 & 1 & yes \\
A3 & fixed schedule & 400 & 140 & 0.722 & valid & 0 & 0 & 0 & 1 & yes \\
\addlinespace[3pt]
\multicolumn{11}{@{}l}{\textit{AP: the placebo copy as real against the panel}} \\
AP & discovery & 400 & 140 & 0.509 & valid & 0 & 0 & 0 & 1 & yes \\
AP & replication & 400 & 140 & 0.508 & valid & 0 & 0 & 0 & 1 & yes \\
AP & fixed schedule & 400 & 140 & 0.510 & valid & 0 & 0 & 0 & 1 & yes \\
\addlinespace[3pt]
\multicolumn{11}{@{}l}{\textit{B1: the generated dataset on the panel stays}} \\
B1 & discovery roster & 400 & 140 & 0.581 & valid & 0 & $1.9\times10^{-11}$ & 0 & 1 & yes \\
B1 & replication roster & 400 & 140 & 0.591 & valid & 0 & $1.5\times10^{-11}$ & 0 & 1 & yes \\
\bottomrule
\end{tabular}
\end{minipage}

\bigskip
\noindent\begin{minipage}{\textwidth}
\pnl{b}{\textbf{Precision of the A1 audits (descriptive).} Fixed-score pair bootstrap: draws of the 400 pairs with both records of a pair together and the out-of-fold scores held fixed. Refit pair bootstrap: draws of the 400 pairs with the copies of a stay kept in one fold, refitted per draw. Fold band: random assignments of the 400 stays to five folds of 80, refitted per assignment; its minimum and maximum are given to four decimals so that no value is rounded onto 0.75. Percentile intervals. NEAR\_THRESHOLD is set when the refit interval or the fold band's central range contains 0.75. In the discovery and fixed-schedule panels the registered fold assignment gave a lower AUC than every random assignment, and in the fixed-schedule panel at least one random assignment gave an AUC above 0.75. Sources: \texttt{intervals.csv} and \texttt{outcomes.json}.}
\centering\footnotesize
\begin{tabular}{l r r r}
\toprule
 & Discovery & Replication & Fixed schedule \\
\midrule
A1 audit AUC (registered folds) & 0.714 & 0.727 & 0.720 \\
Fixed-score pair bootstrap, 95\% interval (2,000 draws) & 0.692 to 0.739 & 0.705 to 0.749 & 0.699 to 0.744 \\
Refit pair bootstrap, 95\% interval (200 draws) & 0.678 to 0.737 & 0.675 to 0.732 & 0.680 to 0.739 \\
Fold band, central 95\% range (100 assignments) & 0.718 to 0.742 & 0.721 to 0.742 & 0.726 to 0.745 \\
Fold band, minimum to maximum & 0.7148 to 0.7477 & 0.7195 to 0.7496 & 0.7247 to 0.7541 \\
NEAR\_THRESHOLD flag & not set & not set & not set \\
\bottomrule
\end{tabular}
\end{minipage}

\bigskip
\noindent\begin{minipage}{\textwidth}
\pnl{c}{\textbf{Label-swap nulls, placebo copies and paired tests (descriptive).} Label-swap null: 200 swaps of the two labels within pairs, refitted on the registered folds, for each A1 and each AP audit. A1 $-$ A3: the point difference with the percentile 95\% interval of a refit pair bootstrap of 200 draws, the same draws for both audits. Hotelling's $T^2$ on the paired differences (panel minus real) of the 36 means, standard deviations and lag-1 autocorrelations of the 12 written channels in the A1 records of the 400 stays, with $q$ coordinates kept and $F$ on $q$ and $400-q$ degrees of freedom. Sources: \texttt{nulls.csv}, \texttt{intervals.csv} and \texttt{hotelling.csv}.}
\centering\footnotesize
\begin{tabular}{l r r r}
\toprule
 & Discovery & Replication & Fixed schedule \\
\midrule
A1 audit AUC & 0.714 & 0.727 & 0.720 \\
A1 label-swap null, mean (SD) & 0.500 (0.012) & 0.499 (0.011) & 0.500 (0.013) \\
A1 label-swap null, 95th percentile & 0.518 & 0.517 & 0.520 \\
\addlinespace[2pt]
A3 audit AUC & 0.712 & 0.723 & 0.722 \\
A1 $-$ A3 [paired 95\% interval] & 0.002 [$-$0.006, 0.009] & 0.004 [$-$0.005, 0.009] & $-$0.001 [$-$0.004, 0.009] \\
\addlinespace[2pt]
AP audit AUC & 0.509 & 0.508 & 0.510 \\
AP label-swap null, mean (SD) & 0.500 (0.004) & 0.500 (0.003) & 0.500 (0.004) \\
AP label-swap null, 95th percentile & 0.506 & 0.505 & 0.506 \\
\addlinespace[2pt]
Hotelling: coordinates kept, $q$ of 36 & 36 & 36 & 36 \\
Hotelling: $F(36, 364)$ & 44.6 & 49.9 & 56.3 \\
Hotelling: $P$ & $2.0\times10^{-111}$ & $1.3\times10^{-118}$ & $1.7\times10^{-126}$ \\
\bottomrule
\end{tabular}
\end{minipage}

\bigskip
\noindent\begin{minipage}{\textwidth}
\pnl{d}{\textbf{Plateau diagnostic on the A1 records.} Outcome channels whose lag-1 autocorrelation in the panel falls more than 0.02 below that of the real records. Sources: \texttt{plateau/\allowbreak{}PL\_*/\allowbreak{}receipt.json} and \texttt{plateau/\allowbreak{}PL\_*/\allowbreak{}channel\_\allowbreak{}persistence\_\allowbreak{}summary.csv}.}
\centering\footnotesize
\begin{tabularx}{\textwidth}{l >{\raggedright\arraybackslash}X >{\raggedright\arraybackslash}X >{\raggedright\arraybackslash}X}
\toprule
 & Discovery & Replication & Fixed schedule \\
\midrule
Outcome channels with a lag-1 deficit, of 12 & 2 & 1 & 2 \\
Channel: lag-1 gap, panel minus real & invasive mean arterial pressure: $-$0.038; heart rate: $-$0.034 & heart rate: $-$0.038 & invasive mean arterial pressure: $-$0.032; heart rate: $-$0.028 \\
\bottomrule
\end{tabularx}
\end{minipage}

\bigskip
\noindent\begin{minipage}{\textwidth}
\pnl{e}{\textbf{Decomposition (descriptive).} The locked pipeline refitted on subsets of the 644 summaries of A1 and A2: single summary families (161 summaries each) and channel groups (the 12 written outcome channels, 48 summaries; the 5 treatment channels, 20; the 144 other channels, 576). All 644 is the audit AUC; the refit on all 644 summaries reproduced it with a gap of 0 for each of the 12 A1, A2, A3 and AP audits. Refitted on the 48 summaries of the written channels alone, the pipeline gives an AUC above 0.75 for A1 in every panel; in the full A1 audit, 36 of these 48 summaries can differ within a pair (the 12 restored missing fractions are equal), and they stand beside the 596 summaries of the 149 copied channels, which are identical in each pair, so 608 of the 644 summaries carry no difference. The same file gives, per written channel and summary, the paired difference, its paired $t$ and the audit's SMD (not shown). $^{\dagger}$Given to four decimals because it rounds to 1.000. Sources: \texttt{decomposition.csv} and \texttt{stage\_\allowbreak{}receipts/\allowbreak{}S4.json}.}
\centering\footnotesize
\begin{tabular}{l l r r r r r r r r}
\toprule
 & & & \multicolumn{4}{c}{Summary family} & \multicolumn{3}{c}{Channel group} \\
\cmidrule(lr){4-7}\cmidrule(lr){8-10}
Audit & Panel & All 644 & Means & SDs & \shortstack{Missing\\fractions} & Lag-1 & \shortstack{12 written\\outcome} & \shortstack{5\\treatment} & \shortstack{144\\other} \\
\midrule
A1 & discovery & 0.714 & 0.612 & 0.690 & 0.500 & 0.602 & 0.781 & 0.500 & 0.500 \\
 & replication & 0.727 & 0.627 & 0.680 & 0.500 & 0.610 & 0.781 & 0.500 & 0.500 \\
 & fixed schedule & 0.720 & 0.614 & 0.694 & 0.500 & 0.602 & 0.789 & 0.500 & 0.500 \\
\addlinespace[2pt]
A2 & discovery & 0.970 & 0.889 & 0.825 & 0.998 & 0.650 & 0.990 & 0.994 & 0.500 \\
 & replication & 0.978 & 0.915 & 0.878 & 0.996 & 0.683 & 0.993 & 0.998 & 0.500 \\
 & fixed schedule & 0.968 & 0.889 & 0.828 & 0.998 & 0.646 & 0.990 & 0.9995$^{\dagger}$ & 0.500 \\
\bottomrule
\end{tabular}
\end{minipage}

\bigskip
\noindent\begin{minipage}{\textwidth}
\pnl{f}{\textbf{Not audit values.} A flexible classifier, scikit-learn's histogram gradient boosting (200 iterations, random state 13, other defaults), on the same 644 summaries and folds as each A1 audit, with its native handling of missing values. These AUCs are not audit values and enter no decision (the A1 audit AUCs are in panel a). Source: \texttt{flexible.csv}.}
\centering\footnotesize
\begin{tabular}{l r}
\toprule
Panel & Flexible classifier AUC (not an audit value) \\
\midrule
Discovery & 0.922 \\
Replication & 0.909 \\
Fixed schedule & 0.925 \\
\bottomrule
\end{tabular}
\end{minipage}

\bigskip
\noindent{\footnotesize \textbf{Earlier realism reads (disclosure).} Two realism reads of an earlier washout design exist; they are not audits of these panels and are not listed above. Both used synthetic source tapes (smooth curves, not eICU-CRD records) and an earlier version of the construction, NaCl scored on creatinine. In the first, the locked discriminator recipe separated the synthetic source tape from the generated record with natural dosing, far above the threshold of 0.75, and did not separate the generated record without the injected response from the record with it. In the second, a discriminator trained to tell the record with the injected response from its copy without it stayed below 0.75. Source: protocol section 2.2.\par}
\par\endgroup

\subsection*{B. The in-support control and the g-computation comparator}

\paragraph{Registration and earlier reads.} Both analyses were registered in one protocol (\texttt{WASHOUT\_\allowbreak{}Q5Q6\_\allowbreak{}PROTOCOL\_\allowbreak{}20261008\_\allowbreak{}V1.md}, amended by \texttt{WASHOUT\_\allowbreak{}Q5Q6\_\allowbreak{}PROTOCOL\_\allowbreak{}20261011\_\allowbreak{}V2.md}, tag \texttt{washout-\allowbreak{}q5q6-\allowbreak{}protocol-\allowbreak{}v2}) after the results at hour 90 were known, at the request of an internal review; the scored hours and every pin were frozen (tag \texttt{washout-\allowbreak{}q5q6-\allowbreak{}freeze-\allowbreak{}v1}) before any clinical outcome was computed. The true effects at hour 90 had been read by earlier analyses of the three panels: the exploratory construction campaign from which the registered cell was chosen (disclosure TRB-EI-12 of the test-read registry), the identifiability map, the target restoration, the oracle-structure learner and the response controls, and the static and sequence benchmarks; the ICare-RH placebo gate stopped before any target read. The production run of the map reproduces one of the exploration checks byte for byte (jobs 10684 and 10690), so the two are one measurement. The panels' source stays include 50, 67 and 50 stays of the ICare-DT test split. Before the protocol was written, the analysts knew every learner's result at hour 90, and they knew that the vendored G-Net is numerically fragile on the intensive care cohort, from an earlier operational run on training and validation data and the repairs it required. They also knew a label-free diagnostic at hour 42: the Causal Transformer's predicted effects varied as little across patients at hour 42 as at hour 90 (per fold, standard deviation of the seed-mean prediction 0.033 to 0.086 at hour 42 and 0.031 to 0.086 at hour 90; Spearman correlation between its predictions at the two hours 0.31 to 0.83). A second label-free diagnostic, the Spearman correlation between the static learners' predictions at hours 42 and 90 (per fold 0.08 to 0.91; pooled over folds 0.27 to 0.84; undefined in 8 folds of the S-learner), was recorded on the cluster and available to them; whether anyone read it is not recorded. Both candidate hours, 30 and 42, were therefore registered and censused without an order. Neither analysis is confirmatory.

\paragraph{Wording.} The protocol fixed the sentences that report these analyses, and they stand in the paper as registered, except the heading of the Results section on standard learners, which was shortened for readability without a change of meaning; the registered heading is ``None of 16 standard treatment-effect learners outperforms predicting zero effect at a query outside the treatment support'' (section 7 of the result note, \texttt{WASHOUT\_\allowbreak{}Q5Q6\_\allowbreak{}RESULT\_\allowbreak{}20261022\_\allowbreak{}V1.md}).

\paragraph{G-Transformer.} A second g-computation sequence model, G-Transformer (Xiong et al., 2024), was not run. On 8 October 2026 no released code was found in the sources searched: the proceedings page, the first and latest arXiv versions, the PubMed Central manuscript, the corresponding author's laboratory page, the proceedings text, three GitHub repository searches and five web searches. The first arXiv version had promised code upon publication. A public repository named G-transformer implements a different method and shows no license on its landing page. The dated record is \texttt{WASHOUT\_\allowbreak{}Q5Q6\_\allowbreak{}GTRANSFORMER\_\allowbreak{}CODE\_\allowbreak{}SEARCH\_\allowbreak{}20261008\_\allowbreak{}V1.md}; code released later is not used.

\paragraph{G-Net as run.} G-Net (Li et al., 2021) was run from the Causal Transformer repository's reimplementation (\texttt{third\_\allowbreak{}party/\allowbreak{}CausalTransformer/\allowbreak{}src/\allowbreak{}models/\allowbreak{}gnet.py}), with the changes of earlier project work, each listed as a deviation with a test that it is inert or declared, among them a recurrent cell that clamps only finite states, a residual bank restricted to holdout trajectories with at least one transition, and a masked loss algebraically identical to the original. Three corrections were registered for these panels: losses on measured hours only, as for the other sequence models; residuals set to zero where a channel was not measured, so that the simulation injects no artefact of the measurement hold; and one seeded person-level holdout split of 10\%, shared by every input, so that G-Net was fitted on 90\% of the persons the other sequence models used. Three candidates were compared in each fold on a factual criterion, the error of 12-step simulated heart-rate trajectories from hours 6 to 72 on validation persons: the published configuration with early stopping (P0-ES), the same with the learning rate one step lower on the authors' grid (P1-ES), and the authors' fixed schedule (P0-F, 200 optimizer steps on this subset). The selected candidate was refitted with three seeds, and its predicted effects were averaged over seeds. The effect at a query was obtained by simulating every recorded variable forward under each arm, averaging 100 Monte Carlo trajectories whose random draws were the same in both arms, a choice that favours the model.

\paragraph{Synthetic gates.} Before any clinical read (campaign S, jobs 13479 to 13486), G-Net passed both registered gates on synthetic tapes without escalation. In the positive control, with an effect standard deviation near 1, its predictions correlated 0.60 with the true effects (gate 0.5). In a g-formula test under time-varying confounding, it estimated the average effect as 1.022 against a true 1.078 (relative error $-$5.2\%, tolerance 15\%), where a naive outcome regression on the same data was biased by 102\% of the true effect. On the registered cell's synthetic scenarios at hours 90 and 30 it was neither informative nor failed. The native test suites passed in a sandbox that hid the data roots.

\paragraph{The census and the in-support control.} The control was to score the same fitted learners at sustained dosing for 12 hours from hour 30 or from hour 42, inside the dosing period, with true effects computed by the generator. An hour was to be scored only if, before any of its effects was computed, it met two criteria in both the discovery and the replication panel: at least 20 query patients received each regime (POS-1), and at least half had an estimated propensity between 0.05 and 0.95 (POS-2). At hour 90, where no patient is dosed, the propensity criterion passed in the replication panel (median fraction 0.725), so it could not separate supported from unsupported queries and was replaced, as registered, by a criterion on the recorded treatments alone (POS-2$'$): the median over the five folds of the fraction of query patients whose bin of recent doses (0, 1 to 3, 4 to 8 or 9 to 12 in the preceding 12 hours) held at least 10 treated and 10 untreated training windows starting within three hours of the hour. POS-2$'$ was 0 at hour 90 in both panels. POS-1 held at both candidate hours, but POS-2$'$ failed at hour 30 in the discovery panel and at hour 42 in both panels (Supplementary Table~S30), so no in-support hour was scored: no in-support effect was computed and no learner was scored inside the support. The synthetic checks of the in-support code passed. Before the freeze, the static learners, the Causal Transformer, RMSN, the oracle-structure learner and the reference arms were refitted or replayed at hour 90 in the three panels, and all 19 arms reproduced their registered predictions exactly, without reading an evaluation target.

\paragraph{G-Net at hour 90.} In this run the true effects at hour 90 were opened once, after G-Net's predictions were sealed. The oracle-structure learner was better than G-Net in all six registered comparisons (Holm correction over 6; adjusted $p=0.0012$ each), each over a comparator that did not itself beat zero; the registered headline is in Supplementary Note~1. Against predicting zero (Holm correction over 2), G-Net's centered PEHE exceeded that of predicting zero by 0.0091 in the discovery panel (95\% interval 0.0016 to 0.0165; worse than predicting zero after correction, adjusted $p=0.033$) and by 0.0058 in the replication panel ($-$0.0020 to 0.0140; unresolved), and, descriptively, by 0.0161 in the fixed-schedule panel (0.0067 to 0.0256). Its correlations with the true effects were 0.05, 0.10 and 0.04 (Supplementary Table~S4; only the replication interval, 0.009 to 0.197, excluded zero), and its predictions spread 0.34, 0.35 and 0.42 times as widely as the true effects (\texttt{metrics.csv} of the aggregates).

\paragraph{Mechanism record.} All 15 fits (three panels, five folds) completed without a declared failure. In every fold P1-ES had the lowest selection criterion, and no selection fit was stopped by the epoch cap; the refits ran 172 to 231 epochs, one optimizer step each, against 200 steps in the authors' schedule, and each fit took 850 to 926 seconds. Every fit decoded 48,000 of 48,000 Monte Carlo trajectories as finite, and the median per-patient Monte Carlo standard error of the predicted effect, 0.0011 to 0.0025, was small against the spread of the predictions (standard deviation 0.079 to 0.097). Each refit seed alone was further above predicting zero than the average over seeds.

\paragraph{Precision and what the null means.} Computed before any data, with 400 patients per panel: a null of this kind is evidence against a calibrated correlation of about 0.27 or more in a panel (80\% power) and says little about correlations below about 0.2; a gain under the registered rule (Holm correction over 2, ranking in both panels and at least 2\% of the standard deviation of the true effects) needed a calibrated correlation of about 0.22. G-Net's correlation intervals ended at 0.148, 0.197 and 0.135. By Proposition~6, with correlations of at most 0.104 no rescaling of its predictions could have lowered its centered PEHE below that of predicting zero by more than 0.54\% of the true effects' standard deviation, so its spread alone does not account for the null. Because no query inside the support met the overlap criteria, the benchmark cannot distinguish a failure of the learners from the lack of treatment variation at the decision query. The result concerns this cell, a query that no patient received, and this implementation; the three panels are dependent (their true effects correlate 0.999 row by row), so agreement between the discovery and replication panels is not an independent replication.

\paragraph{Forecast.} The protocol recorded a forecast before any data, as its author's judgment: that G-Net would most likely not beat predicting zero at hour 90 and might need the registered escalation to pass its positive control (it did not need it); and, inside the support, that most likely no learner would beat predicting zero, an outcome that presupposed a scored hour, whereas the census ended the control earlier.

\paragraph{Compute.} From the scheduler's accounting (elapsed time, and elapsed time times allocated processor cores, each run with its dispatch job): the synthetic campaign (jobs 13470 and 13479 to 13486) 4.3 task-hours and 31.8 allocated CPU-hours; the census and the replays (13534 and 13541 to 13546) 18.8 and 150.9; the clinical run (13761 and 13772 to 13781) 3.7 and 26.1, almost all in the 15 G-Net fits of about 15 minutes each; 26.8 task-hours and 208.8 allocated CPU-hours in all.

\paragraph{Disclosures.} A first dispatch of the synthetic campaign was cancelled before it started, after a check before start found defects that were fixed by an amendment of the protocol (V2) before any registered run. A non-registered test job of 11 October 2026 ran the packages' tests outside the sandbox, and one test of the realism-audit package executed that package's first stage on the panels' archives (disclosure TRB-EI-19); no code of this protocol read panel content in it. The cluster's account association lapsed more than once during the runs, which delayed jobs but changed no input. The result note and its release review (\texttt{WASHOUT\_\allowbreak{}Q5Q6\_\allowbreak{}RESULT\_\allowbreak{}20261022\_\allowbreak{}V1.md}, \texttt{WASHOUT\_\allowbreak{}Q5Q6\_\allowbreak{}RELEASE\_\allowbreak{}REVIEW\_\allowbreak{}20261022\_\allowbreak{}V1.md}) are in \texttt{reports/\allowbreak{}revision\_\allowbreak{}v3/}, and the aggregates in \texttt{source\_\allowbreak{}tables/\allowbreak{}revision\_\allowbreak{}v3/\allowbreak{}washout\_\allowbreak{}q5q6\_\allowbreak{}56dc33f\_\allowbreak{}13776\_\allowbreak{}v1/}.

\ifnum\value{table}=29\relax\else
  \typeout{WARNING (Supplementary Note 9): table counter was \the\value{table} before Supplementary Table S30; set to 29.}
  \setcounter{table}{29}
\fi
\begingroup
\setlength{\tabcolsep}{3.5pt}
\medskip
\noindent\begin{minipage}{\textwidth}
\captionof{table}{\textbf{The in-support control.} Label-free census of the registered candidate hours 30 and 42, inside the dosing period, and of hour 90, the decision query (negative control), on the factual records of the three panels, pinned before any in-support effect could be computed (registered protocol \texttt{WASHOUT\_\allowbreak{}Q5Q6\_\allowbreak{}PROTOCOL\_\allowbreak{}20261008\_\allowbreak{}V1.md} as amended by \texttt{WASHOUT\_\allowbreak{}Q5Q6\_\allowbreak{}PROTOCOL\_\allowbreak{}20261011\_\allowbreak{}V2.md}). $W=1$ and $W=0$: query patients, of 400 per panel, whose factual 12-hour window from the hour was fully dosed or undosed (partial: the rest). POS-1: at least 20 patients with each of $W=1$ and $W=0$. POS-2: median over the five folds of the fraction of query patients with an estimated propensity between 0.05 and 0.95. POS-2$'$: median over the five folds of the fraction of query patients whose bin of recent doses held at least 10 treated and 10 untreated training windows starting within three hours of the hour; it replaced POS-2 because POS-2 passed at hour 90 in the replication panel, where no patient is treated. An hour was to be scored if POS-1 and POS-2$'$ held in both the discovery and the replication panel (threshold 0.50); none did, so no in-support effect was computed and the table has no learner row. Domain classifiers (descriptive): person-grouped classifiers separating the query patients' covariates from training units pooled over hours 6 to 42, from treated or untreated units only, and from units starting within three hours of the hour (undefined at hour 90, where no unit starts within 87 to 93); ranges over the five folds. The fixed-schedule panel is reported but does not enter the gate. Treated training units in the query's recency cell at hour 90 (dosed earlier, no dose in the preceding 12 hours), per fold: 1, 1, 0, 1, 1 (discovery), 3, 3, 2, 1, 3 (replication) and 0, 0, 0, 0, 0 (fixed schedule). Sources: \texttt{census.json} and \texttt{selection.json} in \texttt{source\_\allowbreak{}tables/\allowbreak{}revision\_\allowbreak{}v3/\allowbreak{}washout\_\allowbreak{}insupport\_\allowbreak{}prep\_\allowbreak{}v1/}.}
\label{tab:q5-insupport}
\end{minipage}

\medskip
\noindent\begin{minipage}{\textwidth}
\centering\footnotesize
\begin{tabular}{l r r r r l r r}
\toprule
Panel & Hour & $W=1$ & $W=0$ & Partial & POS-1 & POS-2 & POS-2$'$ \\
\midrule
Discovery & 30 & 84 & 35 & 281 & yes & 0.4375 & 0.2750 \\
Discovery & 42 & 108 & 50 & 242 & yes & 0.4000 & 0.2875 \\
Discovery & 90 & 0 & 400 & 0 & no & 0.1875 & 0.0000 \\
\addlinespace[3pt]
Replication & 30 & 80 & 39 & 281 & yes & 0.5875 & 0.6375 \\
Replication & 42 & 108 & 49 & 243 & yes & 0.5250 & 0.3000 \\
Replication & 90 & 0 & 400 & 0 & no & 0.7250 & 0.0000 \\
\addlinespace[3pt]
Fixed schedule & 30 & 0 & 400 & 0 & no & 0.0000 & 0.0000 \\
Fixed schedule & 42 & 0 & 400 & 0 & no & 0.0000 & 0.0000 \\
Fixed schedule & 90 & 0 & 400 & 0 & no & 0.0000 & 0.0000 \\
\bottomrule
\end{tabular}

\medskip
\begin{tabular}{l r l l l l}
\toprule
Panel & Hour & \shortstack{Pooled\\AUC} & \shortstack{Against treated\\units, AUC} & \shortstack{Against untreated\\units, AUC} & \shortstack{Same hour\\AUC} \\
\midrule
Discovery & 30 & 0.755 to 0.821 & 0.824 to 0.905 & 0.788 to 0.902 & 0.634 to 0.715 \\
Discovery & 42 & 0.807 to 0.845 & 0.835 to 0.915 & 0.843 to 0.899 & 0.536 to 0.729 \\
Discovery & 90 & 0.9996 to 0.9999 & 0.9966 to 0.9999 & 0.995 to 0.998 & -- \\
\addlinespace[3pt]
Replication & 30 & 0.727 to 0.814 & 0.822 to 0.888 & 0.787 to 0.854 & 0.626 to 0.751 \\
Replication & 42 & 0.799 to 0.882 & 0.791 to 0.887 & 0.854 to 0.916 & 0.581 to 0.698 \\
Replication & 90 & 0.99974 to 0.99997 & 0.999 to 1.000 & 0.980 to 0.999 & -- \\
\addlinespace[3pt]
Fixed schedule & 30 & 0.986 to 0.988 & 1.000 & 0.970 to 0.976 & 0.917 to 0.941 \\
Fixed schedule & 42 & 0.985 to 0.987 & 1.000 & 0.969 to 0.975 & 0.864 to 0.886 \\
Fixed schedule & 90 & 1.000 & 1.000 & 1.000 & -- \\
\bottomrule
\end{tabular}
\end{minipage}
\par\endgroup

%% file: supplementary/supplementary_note5.tex
\clearpage
\section*{Supplementary Note 5: Theory and proofs}
\addcontentsline{toc}{section}{Supplementary Note 5: Theory and proofs}
\begingroup\sloppy

This note sets out what each evaluation design in the paper can and cannot identify, so that the main results can be read at the strength their designs support. It gives the definitions, formal statements, assumptions, proofs and numerical checks of the seven propositions summarized in the main text's Methods (``Theoretical analysis of the evaluation''). Sections 5.1 to 5.7 take one proposition each, in the order of the Results. They cover synchronization (Proposition 1), the difference-in-differences and the current-value hypotheses (Proposition 2), the realism audit (Proposition 3), recoverability and cadence (Proposition 4), the efficient reader (Proposition 5), treatment-effect learners and transport (Proposition 6), and ICare-RH's placebo gate (Proposition 7). Section 5.8 states what the theory does not show.

\paragraph{Status.} The seven propositions were derived after the results. The theory was not used to choose, register or stop any eICU-CRD analysis, and no part of it was registered in advance, in version control or externally. Each proposition lists its assumptions and says where the implemented models, code or data depart from the idealization it uses. The numerical checks are post hoc and descriptive. They re-read shipped tables of previously read splits, reconstructed inputs from shipped files or used simulation, and none made a new prediction of the twin or of any benchmarked model. The checks of Section 5.4 re-read only the per-patient washout targets that the registered benchmark had already scored. Two groups of checks are new analyses of previously read data, and both are disclosed in Supplementary Note~2. The first group read the per-stay predictions, last measured values and labels of the previously read eICU-CRD test panels. With them it formed the new contrast of two existing hour-6 reads reported in Section 5.1 and computed the label prevalences and the tie rates of the last measured values (Section 5.2). As an unreported diagnostic, it also fitted cross-fitted logistic combinations of the existing predictions on those labels. These reads are recorded as one disclosure row of the test-read registry. The second group fitted classifiers to the summaries of the realism probe, including a flexible classifier in one configuration chosen after the audit (Section 5.3). Where a check depends on an idealization, it says so. For example, checks that need the panels' draw masks, which were not shipped, use simulated draw patterns. The check scripts and their outputs will be released with the code, in \texttt{reports/revision\_v3/theory\_checks/}.

\paragraph{Standard and specific parts.} Several parts of the propositions restate standard results for the quantities of this evaluation: the Neyman--Pearson lemma and the refinement decompositions of proper scores (Section 5.1), bounds of an AUC by distances between distributions and equal-covariance Gaussian discrimination (Section 5.3), generalized least squares and trend partialling (Section 5.5), the decomposition of mean squared error by correlation and spread (Section 5.6), and the profile likelihood of a one-way variance component (Section 5.7). Each section begins by naming its standard parts and their sources, so that the reader can see which parts are specific to this evaluation.

\paragraph{Conventions.} $\operatorname{Var}$, $\operatorname{Cov}$, $\operatorname{corr}$ and $\operatorname{SD}$ are population moments; shipped tables report their sample versions with population standard deviations. AUROC gives half credit to ties. $\mathbf 1\{\cdot\}$ is the indicator and $\Phi$ the standard normal distribution function. For a positive definite matrix $\Omega$, $\langle u,v\rangle_\Omega:=u^\top\Omega^{-1}v$ and $\cos^2_\Omega(u,v):=\langle u,v\rangle_\Omega^2/(\langle u,u\rangle_\Omega\langle v,v\rangle_\Omega)$. Assumption labels such as (A1) are local to the section in which they appear, and a symbol introduced inside a proof is local to that proof. Code references give file names and line numbers at commit 82587202 of the analysis repository; apart from the check scripts added in \texttt{reports/revision\_v3/theory\_checks/} and the files added for the registered analyses run after that commit (the realism audit of the washout panels, the in-support control and the g-computation comparator), the only code files changed between that commit and the commit of the submitted manuscript are the files added or changed for the registered MIMIC-IV read (among them \texttt{icare\_dt/measurement\_occurrence.py} and \texttt{icare\_dt/terminal\_semantics.py}) and the scripts that build the paper's figures, tables, source data and cover letter (\texttt{icare\_dt/figures/npj\_ai\_evaluation\_figures.py} and \texttt{paper\_npj\_ai/make\_*.py}). The trait $\zeta_{ij}$, ICare-RH's channel loadings $\gamma_k$ and the lag $\ell$ of an exposure kernel are those of Methods.

\paragraph{Symbols with more than one use.} Some symbols carry more than one meaning, each confined to the sections named.
\begin{itemize}
\item $L$: landmark hour (5.1, 5.2). $L_{kj}$: loading of treatment $j$ on channel $k$ (5.4 to 5.7); it always carries two subscripts, and $\Lambda_k:=L_{kj^\ast}$.
\item $R$: the roster event, with roster law $\mathbb P_R$ (5.1, 5.2). $P_R$, $P_G$, $f_R$, $f_G$: laws and summaries of real and generated records (5.3).
\item $\eta$: the latent index of the threshold model (5.2 only). The generator's per-stay responsiveness $\eta_i$ (Supplementary Note~4) does not occur in this note.
\item $u$: spread ratio of a predictor (5.6, 5.7). The generator's latent severity $u_t$ (Supplementary Note~4) occurs only as $u_0$, its shipped per-patient value at row 0, from which the natural doses were reconstructed (5.4 to 5.6).
\item $\mathcal F_6$: the information in the first six hours (5.1, 5.2); the letter $W$ is not used for it. $W$ is the generator's direct-route matrix (Supplementary Note~4) and enters this note only through $L_{kj}$.
\item $\lambda$: decay of an exposure kernel (5.4 to 5.7); penalty of the audit's logistic regression (5.3).
\item $j$: treatment index in $\Gamma_{ik}$, $\zeta_{ij}$ and $L_{kj}$; index of consecutive draws in $d_j$, $\Delta_j$ and $g_j$ (5.4, 5.5, 5.7). The two uses never occur in one formula.
\item $\Pi$: residual projector on a quadratic trend (5.4, 5.5). $\Pi_\zeta$: posterior law of the trait (5.6). $\Pi_{\mathrm{pool}}$: donor pool (5.1).
\item $v$: weight vector of a linear statistic (5.5); slope variance of a random-slope model (5.6, 5.7).
\item $b$: trait-free baseline $b_{ik}$ (5.4, 5.5); response slope $b_i$, $b^\ast_i$ (5.6, 5.7).
\item $\delta$: mean difference of the audit summaries (5.3); shortfalls $\delta_2$, $\delta_a$ of learned reads (5.1); drift variance $v_\delta$ (5.7).
\item $r$: a record read by the forecaster, with parts $r_{1:6}$ and $r_{7:L}$ (5.2); correlation of a predictor with $\tau$ (5.6, 5.7), as in the main text; residual vectors $r_i$ and $r_{ik}$ always carry a patient subscript.
\item $g_j$: gap between consecutive draws (5.4, 5.5, 5.7). The linear-drift direction is $\mathbf t$ (5.7).
\item $U_6$: affine index of the first six hours (5.2). $U_{ik}$: shift of the held record per unit amplitude (5.4).
\item $K$: exposure kernel. $d_{\mathrm{KS}}(s)$: Kolmogorov distance between the two laws of a score (5.3). K1, K2, K3: clauses of the placebo gate (below).
\end{itemize}

\subsection*{Notation and shared definitions}

\paragraph{Forecasting (Propositions 1 and 2).} A stay is indexed by $i$ and its hours by $t=1,2,\dots$. $X_t$ is the hour-$t$ model input (161 standardized values and the observation mask) and $x^s$ the static inputs. $\mathcal F_t:=\sigma(x^s,X_1,\dots,X_t)$ is the information in the record through hour $t$. The frozen twin state $z_t$ is a function of $(x^s,X_1,\dots,X_t)$, and hence $\mathcal F_t$-measurable. The encoder is causally masked, but the propositions do not rely on the mask, because every read is computed from the record truncated at its origin (Section 5.1, assumption (A0)). This measurability is the only property of the implemented encoder that the propositions use. A setting is $(L,H)\in\{(12,6),(18,12),(30,24)\}$, and in every setting $L=6+H$. For a target $k$, $T_{\mathrm{ev}}$ is the hour of the first event of any class (threshold crossing, live discharge, death), $Y:=\mathbf 1\{\text{first crossing in }(L,L+H]\}$, $R$ is the event that the stay is on the setting's roster and $\pi:=P(Y=1\mid R)$. The arms are those of the main text and Methods: S2, the synchronized forecast $F_L(H)$; S1, the hour-6 forecast $[F_6(2H)-F_6(H)]/S_6(H)$ of the same window; BR and BR2, S1 and S2 with a training-fitted intercept per setting on the logit scale; IC, IC-A and IC-O, the information-constant arm and its two sensitivity versions; SH, the donor arm (the patient's hours 1 to 6 followed by the matched donor's hours 7 to $L$), and SH-A, its availability-preserving version; own and donor, the patient's and the donor's last measured target values; and three donor-only scores (the donor's own forecast, last value and number of measured hours). The ten hour-6 and donor arms are S1, BR, IC, IC-A, IC-O, SH, SH-A and the three donor-only scores. $V$ is the direction-signed last measured value of target $k$ at hour $L$ and $V'$ the matched donor's value.

\paragraph{Definition 1 (ideal forecast, information-set value, donor independence).} For a $\sigma$-field $\mathcal G$, $p_{\mathcal G}:=P(Y=1\mid\mathcal G,R)$ is the ideal forecast given $\mathcal G$. The implemented S1 conditions on no event of any class by hour $L$, because $L=6+H$ and the ratio $[F_6(2H)-F_6(H)]/S_6(H)$ divides by the all-cause event-free probability. Every read at origin $L$ conditions on the patient being at risk at $L$, because the heads are supervised only at at-risk origins. The information-set value of an arm is $p_{\mathcal G_a}$, where $\mathcal G_a$ is generated by the arm's input and the event its read conditions on; $S1^\ast$ and $S2^\ast$ are the information-set values of S1 and S2, and Proposition 1(b) shows that $S1^\ast=p_{\mathcal F_6}$. Donor independence (D) states that the donor record is conditionally independent of the patient's label and hours 7 to $L$ given the patient's first six hours. The implementation supports it, since donors are matched on hour-6 encoder states only, never on labels and never from the same person. It remains an assumption, because the standardization constants are roster-wide and the checkpoints were selected on validation data (Proposition 1, (D)). Exact matching, under which the donor's hours 7 to $L$ have the conditional law of the patient's given the first six hours, is an idealization of nearest-neighbor matching in the 64-dimensional hour-6 state. The twin is a Transformer trained on a discriminative competing-risks likelihood and is not a Bayes filter, so treating its forecast as an ideal forecast is also an idealization.

\paragraph{Definition 2 (current-value hypotheses).} A forecaster obeys the current-value restriction (CVR) if its forecast at origin $L$ is a function of the first six hours and $V$ only. The data obey the current-state hypothesis $H_{\mathrm{CS}}$ if $Y\perp X_{7:L}\mid(\mathcal F_6,V,R)$, and current-value sufficiency (CVS) if $Y\perp\mathcal F_L\mid(V,R)$. Personalization beyond the current state is the contrast $\mathrm{AUROC}(p_{\mathcal F_L})-\mathrm{AUROC}(p_{\mathcal F_6\vee\sigma(V)})$, which no registered arm estimates. In the latent-threshold Gaussian model (LT), a standardized latent index $\eta$ and the record are jointly Gaussian, $Y=\mathbf 1\{\eta>t_\pi\}$ with $t_\pi:=\Phi^{-1}(1-\pi)$, and each arm is a strictly increasing function of a linear index. Then $\mathrm{AUROC}_a=A_\pi(\rho_a)$, where $\rho_a$ is the correlation of arm $a$'s index with $\eta$ and $A_\pi$ is a strictly increasing bijection of $[-1,1]$ onto $[0,1]$ (Proposition 2, Lemma 1).

\paragraph{Definition 3 (classifier two-sample test).} For a score $s$ of the 644 audit summaries $f$, $\mathrm{AUC}(s):=P(s(f_G)>s(f_R))+\tfrac12P(s(f_G)=s(f_R))$ for independent generated and real summaries. The audit statistic is the pooled out-of-fold AUC of an $L_2$-penalized logistic regression ($C=0.1$, unpenalized intercept) over five person-grouped folds of 400 matched pairs. Its population target is the AUC of the minimizer of the penalized population logistic risk (Proposition 3).

\paragraph{Washout panels (Propositions 4 to 7).} Rows $t=0,\dots,89$, one per hour, precede the decision row $t_0=90$. The horizon is $h=12$, and the post-decision step $s=0,\dots,11$ refers to row $91+s$. Outcome channels are $k=1,\dots,12$ ($k^\ast$ is heart rate in the registered cell, in units of 15 beats per minute) and treatments $j=1,\dots,5$ ($j^\ast$ is the scored treatment). $A_i(t)\in\{0,1\}$ is the dose of $j^\ast$. Under natural dosing (NATMIN), doses fall within rows 6 to 53 and are topped up to at least eight dosed rows. Under the fixed schedule (FIX19), rows 6 to 24 are dosed. Rows 54 to 89 are a washout. A kernel $K=(\ell,\lambda,\kappa)$ with lag $\ell$, decay $\lambda$ and saturation $\kappa$ defines the memory $e_K(t)=\lambda e_K(t-1)+A(t-\ell)$, $e_K(0)=0$, and the exposure profile $p_K(t):=\varphi_\kappa(e_K(t))$ with $\varphi_\kappa(e):=e/(\kappa+e)$. The generator kernel is $K_0=(2,0.8,1)$ and $q_i:=p_{K_0}$ is patient $i$'s occupancy. $\zeta_{ij}$ is the trait of patient $i$ for treatment $j$ and $\zeta_i:=\zeta_{ij^\ast}\in[-2/3,2]$, with mean $\mu_\zeta$ and variance $\sigma^2_\zeta$. $L_{kj}$ is the loading of treatment $j$ on channel $k$ and $\Lambda_k:=L_{kj^\ast}$ ($\Lambda_{k^\ast}=-1$ for phenylephrine on heart rate), and $c$ is the amplitude cap. The latent standardized outcome is $y_{ik}(t)=b_{ik}(t)+c\,\alpha_{ik}(t)\tanh\Gamma_{ik}(t)$ with $\Gamma_{ik}(t):=\sum_jL_{kj}\zeta_{ij}q_{ij}(t)$. Here $b_{ik}$ collects everything that involves no trait, namely the generator background, the ordinary treatment routes and the source tape (the stay's real record), while $\alpha_{ik}\in[0,1]$ is the attenuation near a physiological bound. The draw set $\mathcal D_{ik}$ consists of row 0 and the rows at which the source value changes, ordered $d_0=0<d_1<\dots<d_n$. The last draw at or before $t$ is $d_{ik}(t)$, and the held record is $Z_{ik}(t):=y_{ik}(d_{ik}(t))$. The design is $\mathcal S_i:=(A_i,\mathcal D_{i1},\dots,\mathcal D_{i,12})$. Over consecutive draws, $\Delta_jf:=f(d_{j+1})-f(d_j)$ and $g_j:=d_{j+1}-d_j$. The exposure energies are $Q^\circ_i:=\sum_{t=1}^{89}(q_i(t)-q_i(t-1))^2$ (hourly) and $Q^{(m)}_{ik}:=\sum_jg_j^m(\Delta_jq_i)^2$ for $m\in\{-1,0,1\}$ (sampled), with $J_{ik}:=Q^{(-1)}_{ik}$. The cadence-aware statistic is $\mathcal C_{ik}:=\sum_j\Delta_jZ_{ik}\,\Delta_jq_i$ (with the companion feature $Q^{(0)}_{ik}$), and the plain occupancy contrast is $\mathcal O_{ik}:=\sum_t(\Pi q_i)(t)Z_{ik}(t)$, with $\Pi$ the residual projector on a quadratic trend over rows 0 to 89. Recovery is $\mathrm{rec}:=1-\mathrm{RMSE}/\mathrm{SD}(\tau)$ and $R^2=1-(1-\mathrm{rec})^2$. The bar $\mathrm{rec}=1-\sqrt{0.8}=0.106$ corresponds to $R^2=0.2$. For a predictor $\hat\tau$, $r:=\operatorname{corr}(\hat\tau,\tau)$, $u:=\operatorname{SD}(\hat\tau)/\operatorname{SD}(\tau)$ (spread ratio) and $\mathrm{cPEHE}:=\operatorname{SD}(\hat\tau-\tau)$. For a $\sigma$-field or random vector $X$, $R^2_\ast(X):=\operatorname{Var}\mathbb E[\tau\mid X]/\operatorname{Var}(\tau)$ is the information content of $X$ about $\tau$.

\paragraph{Definition 4 (washout target).} $\tau_i:=h^{-1}\sum_{s=0}^{h-1}[Y^A_{is}(\bar1)-Y^A_{is}(\bar0)-Y^0_{is}(\bar1)+Y^0_{is}(\bar0)]$, a frame-by-regime interaction on the latent (every-hour) heart-rate path at row $91+s$. Frame A has cap $c$ and frame 0 has cap 0. Both are rolled out from the same factual trace with the same seed, so they share every state, action and noise path. Regime $\bar1$ gives $j^\ast$ on rows 90 to 101 and regime $\bar0$ withholds it. The other four treatments follow the generator's policy with common random numbers. The target is computed from the unheld path, whereas the readers see the held record.

\paragraph{Definition 5 (design exogeneity).} The design $\mathcal S_i$ is independent of the traits. Natural doses are harvested from a policy run on the same real window, which depends on the stay's tape and latent state but not on the trait. The draw sets are fixed by the source tape, and the trait enters only the injected shift, which does not feed back into the latent state, the doses or the draw sets. Independence of the doses from the trait was checked by reconstructing all 800 NATMIN dose columns of the discovery and replication panels from shipped inputs (dosed-row counts equal the shipped table for 400 of 400 patients in each panel). The design is not independent of the baseline $b$, because the doses and the draw sets both depend on the stay's tape. Any statement that treats them as independent is an idealization and says so.

\paragraph{Definition 6 (linear readers and information content).} A reader is a feature map that is affine in the held record for a fixed design. The history-blind, window-mean, plain, assimilating and cadence-aware readers are of this kind. The window-mean reader takes channel means over four windows set by the first and last dose, and the plain reader is the occupancy-contrast reader of the main text. The oracle reference, which uses the true trait, is not a reader in this sense. Design-only features depend on $\mathcal S_i$ alone. The population $R^2$ of a reader is that of population least squares on its features with the kinetics fixed. The shipped out-of-fold ridge regression with nested cross-validation estimates it with a finite-sample loss. Every predictor that is a function of the record has $R^2\le R^2_\ast(\text{record})$ (Proposition 4(c)(i), Proposition 6(b2)). A supervised reader's out-of-fold $R^2$ therefore estimates a lower bound on the information content of the record and does not bound from above what another estimator using the same record can reach.

\paragraph{Definition 7 (retention and aliasing).} Under a random-walk baseline, the retention $\chi_{ik}:=J_{ik}/Q^\circ_i\in[0,1]$ is the share of the hourly record's Fisher information about the trait that the held record of channel $k$ keeps. An interval $(d_j,d_{j+1}]$ is aliased when the exposure has the same value at its two draws although it moved between them. Under the generator's binary dosing no interval is exactly aliased (Proposition 4(d)(iii)), but nearly aliased intervals occur, and $\chi_{ik}$ measures their effect.

\paragraph{Definition 8 (beating zero).} The constant prediction 0 has centered PEHE $\operatorname{SD}(\tau)$, and a predictor beats predicting zero when $\mathrm{cPEHE}<\operatorname{SD}(\tau)$. This is a population criterion. In the benchmark, a learner was counted as outperforming predicting zero only if the 95\% bootstrap interval of its centered PEHE minus that of predicting zero lay wholly below zero.

\paragraph{Definition 9 (transport).} Under NATMIN and FIX19 no dose of $j^\ast$ occurs at or after row 54, so the query regime, sustained dosing from row 90, has propensity zero for every patient. Treated training windows in the query's recency cell of Methods (dosed earlier, no dose in the preceding 12 hours) are factual windows of other regimes, not instances of the query regime. Structural transport assumes T1 (invariance: the same $\zeta_i$ that sets the pre-decision shift sets the query effect), T2 (exogeneity: Definition 5) and T3 (known structure: kernel, bounded map, loadings, trait law and the law of the zero-effect record given the design are known or identified); see Proposition 6. Neither registered hour inside the dosing period (30 or 42) met the overlap criteria of Methods.

\paragraph{Definition 10 (overdispersion along the exposure direction).} For one channel (or a fixed loading direction), a kernel $K$ and a panel copy, let $\mathcal A_i:=\langle p_K,p_K\rangle_{\Omega_i}$ and $\mathcal B_i:=\langle p_K,r_i\rangle_{\Omega_i}$ be the exposure information and the exposure projection of the baseline residual $r_i$ at the draw rows under ICare-RH's working covariance $\Omega_i$. The overdispersion along the exposure direction is $\omega(K):=\sum_i(\mathcal B_i-\hat\mu\mathcal A_i)^2/\sum_i\mathcal A_i$ with $\hat\mu:=\sum_i\mathcal B_i/\sum_i\mathcal A_i$. If the residual has conditional covariance $\Sigma_i$, then on a mean-zero copy $\mathbb E[\mathcal B_i^2\mid\mathcal S_i]=\mathcal A_i\,\omega_i(K)$ with $\omega_i(K):=p_K^\top\Omega_i^{-1}\Sigma_i\Omega_i^{-1}p_K/p_K^\top\Omega_i^{-1}p_K$, which equals 1 when the working covariance is correct along $p_K$. $\omega_0$ denotes the population value on the zero-effect copy.

\paragraph{Definition 11 (pseudo-true parameter).} The pseudo-true parameter of a working model is the parameter value that maximizes the working model's expected log-likelihood under the true law.

\paragraph{Definition 12 (clauses of ICare-RH's placebo gate).} K1 is the lag clause. The analysis stops if, in either natural-dosing panel, the profile likelihood favors a lag outside 1 to 3 hours in at least three of five outer folds. The count of folds in which a slope-variance test has $P<0.05$ was reported descriptively and did not gate. K2 is the pair of spread clauses on the zero-effect copies of every panel: the standard deviation $\sigma_0$ of the predicted effects must be at most 0.05 (absolute clause) and at most 25\% of the standard deviation $\sigma_A$ on the active copy (ratio clause). K3, which was to be read only after K1 and K2 and was never reached, required the 95\% whole-person bootstrap interval of centered PEHE minus that of predicting zero to lie wholly below zero in both natural-dosing panels. Proposition 7(e) concerns K2 only.

\subsection*{5.1 Synchronization is information refinement}
\input{\suppdir supp_theory/p1}

\subsection*{5.2 The difference-in-differences and current-value hypotheses}
\input{\suppdir supp_theory/p2}

\subsection*{5.3 The realism audit as a two-sample test}
\input{\suppdir supp_theory/p3}

\subsection*{5.4 Recoverability, amplitude and retention}
\input{\suppdir supp_theory/p4}

\subsection*{5.5 Which reader is efficient}
\input{\suppdir supp_theory/p5}

\subsection*{5.6 Calibration geometry and transport}
\input{\suppdir supp_theory/p6}

\subsection*{5.7 Interpreting ICare-RH's placebo failure}
\input{\suppdir supp_theory/p7}

\subsection*{5.8 What the theory does not show}

\begin{itemize}
\item \emph{The twin is not shown to be an ideal forecaster or a Bayes filter.} Propositions 1 and 2 use only that the twin's state at hour $t$ is a function of the record through hour $t$. The twin's arms are not those of an ideal forecaster in the idealized class of Proposition 2 (Table~\ref{tab:P2quad}, last column).
\item \emph{The hour-6 arm is not shown to be the model's best hour-6 read.} In a disclosed post hoc read of the eICU-CRD test panel, the model's hour-6 forecast of the earlier window exceeded S1 by 0.021 and 0.050 AUROC in the two saturation settings at $(30,24)$, although neither excess survived Holm correction over the 12 settings (Section 5.1). In those two settings, therefore, S2 $-$ S1 cannot be assumed to be at most the ideal information gain.
\item \emph{How much of the twin's gain passes through the last measured value is not determined.} Exact CVR is false a priori for a model that reads every channel. The four AUROCs of the difference-in-differences cannot reject $H_{\mathrm{CS}}$, and they cannot confirm it when the two attainability conditions of Proposition 2(b) hold strictly. The label-free quantities that would bear on it, $\operatorname{corr}(V,V')$ and the own-value splice of Proposition 2(c), were not computed, because the per-stay values of the validation isolation run are not in the shipped aggregates and the splice needs a new model read. Personalization beyond the patient's current state is not shown.
\item \emph{Supervised readers are not upper bounds.} Readers fitted with the true effects as labels estimate lower bounds on the information content of their feature class (Definition 6); an estimator using the same record can exceed them.
\item \emph{The realism audits bound distances from below only.} The audit of the generated dataset bounds the distance between the real and generated summaries from below and certifies no upper bound (Proposition 3(a)); the audit of the washout panels' generated outcome values is a separate application of the same instrument, carries the same limits, and cannot see that those values respond to simulated rather than recorded dosing (Methods).
\item \emph{The cause of ICare-RH's placebo failure is not located.} Proposition 7(e) covers only the spread clauses K2, not the lag clause K1 or the final clause K3. The absolute clause passed in all three panels, and the ratio clause failed in all three. A failure of the ratio clause is consistent with a baseline misspecified along the exposure direction, with a response the fit did not capture (the spread of predicted effects on the discovery panel's active copy was 0.0068), with a non-global optimum or with finite-sample variation. The data do not decide between these explanations. The kernel preference of Proposition 7(a)(vi) requires equal information, no mean response along the exposure direction and exact minimizers at a fixed baseline, none of which the implementation guarantees. The slow component that the likelihood absorbs on real records is also not identified.
\item \emph{The response and baseline models are idealizations.} The linear regime, a single carrier, and white-noise, AR(1) or random-walk baselines are simplified models that do not describe real records. Three disagreements between the theory's magnitudes and the record are reported but not explained. In the hourly norepinephrine cell the readers tied, although the single-carrier model predicts a cadence-aware advantage. On heart rate, sampling of the scored channel alone accounts for a factor of about 1.4 in the plain reader's loss at real cadence, against observed factors of 5.9 and 7.2 in $C_1$, and for NaCl on creatinine a twelve-channel model under-predicts that loss. Finally, Proposition 5(e) does not cover the cadence-aware reader's lack of separation on the fixed schedule.
\item \emph{Open items.} Four questions remain open. The correlation $\operatorname{corr}(V,V')$ and the own-value splice test would bear on the current-state hypothesis (Proposition 2). A per-channel refit of the registered readers would test the attribution of the plain reader's loss at real cadence (Propositions 4 and 5). The advantage of a 1/gap-weighted reader depends on how jump variance grows with the gap, which the shipped aggregates cannot check (Proposition 5). ICare-RH's lag on real records and the slow component its likelihood absorbs are not yet known (Proposition 7).
\end{itemize}
\endgroup

%% file: supplementary/supp_theory/p1.tex
\paragraph{Proposition 1 (Synchronization is information refinement, and the isolation arms add model error to it).}
\emph{Standard and specific parts.} Part (a) collects standard facts: the ROC dominance of the ideal forecast is the Neyman--Pearson lemma applied within a sub-$\sigma$-field (Lemma~2 below), and the Brier and log-loss identities are the refinement decompositions of proper scoring rules (DeGroot and Fienberg; Gneiting and Raftery). Part (d) is an algebraic identity whose sign constraints come from (a). What is specific to this evaluation is the identification, in (b) and (c), of the information-set values of the implemented arms, including the exact Brier cost of a splice in (c)(iv).

\emph{Setting.} Fix a target $k$ and a setting $(L,H)\in\{(12,6),(18,12),(30,24)\}$, so that $L=6+H$. Let $R$ be the event that a stay is on the roster of this target and setting, $\mathbb P_R:=\mathbb P(\,\cdot\mid R)$ with expectation $\mathbb E_R$, and $Y\in\{0,1\}$ the label (first crossing of target $k$ in $(L,L+H]$), with $\pi:=\mathbb P_R(Y=1)\in(0,1)$. Let $x^{s}$ be the static inputs (including the domain indicator), $X_t$ the hour-$t$ model input (161 standardized values and the observation mask), $\omega_t:=(x^{s},X_1,\dots,X_t)$, $\mathcal F_t:=\sigma(\omega_t)$ and $X_{7:L}:=(X_7,\dots,X_L)$, so that $\mathcal F_L=\mathcal F_6\vee\sigma(X_{7:L})$. $T_{\mathrm{ev}}$ is the hour of the first event of any class for target $k$ (measured threshold crossing, discharge, death). For a $\sigma$-field $\mathcal G$, $p_{\mathcal G}:=\mathbb P_R(Y=1\mid\mathcal G)$ is the ideal forecast given $\mathcal G$, and
\[
S1^{*}:=p_{\mathcal F_6\vee\sigma(\{T_{\mathrm{ev}}>L\})},\qquad S2^{*}:=p_{\mathcal F_L}.
\]
\emph{Arms} (as implemented, with trained weights, intercepts, orientation sign and matching rule held fixed). $\mathrm{S2}=\psi(\omega_L)$ is the frozen model's onset cumulative incidence at bin $H$ read from hours $1$ to $L$; $\psi(\omega_6,x_{7:L})$ denotes the same read on the input whose statics and hours $1$ to $6$ are $\omega_6$ and whose hours $7$ to $L$ are $x_{7:L}$. $\mathrm{S1}=[F_6(2H)-F_6(H)]/S_6(H)$, clipped to $[0,1]$ and read from $\omega_6$ alone. For $a\in\{\mathrm{IC},\mathrm{IC\text{-}A},\mathrm{IC\text{-}O}\}$, $a=\psi(\omega_6,\kappa_a(\omega_6))$, where $\kappa_{\mathrm{IC}}$ repeats the hour-6 values in every channel with every mask entry $0$; $\kappa_{\mathrm{IC\text{-}A}}$ repeats the hour-6 values and the hour-6 non-target mask, with target-channel occurrence $0$; $\kappa_{\mathrm{IC\text{-}O}}$ equals $\kappa_{\mathrm{IC\text{-}A}}$ except that the occurrence of target $k$ at hour $L$ is $1$. $\omega'_L$ is the record (statics and hours $1$ to $L$) of the donor that the seed's matching rule assigns to the stay, a random element under $\mathbb P_R$ (see (D)), and $X'_{7:L}$ are its hours $7$ to $L$; donors are roster stays and therefore at risk at $L$. $\mathrm{SH}=\psi(\omega_6,X'_{7:L})$ and $\mathrm{SH\text{-}A}=\psi(\omega_6,\kappa_{\mathrm{SH\text{-}A}}(X_6,X'_{7:L}))$, where $\kappa_{\mathrm{SH\text{-}A}}$ keeps a non-target channel available, with its hour-6 value, wherever the recipient had it at hour 6 and the donor's row is unavailable. The three \emph{donor-only scores} read the donor's record alone: the donor's own $\mathrm{S2}=\psi(\omega'_L)$, the donor's direction-signed last target value $V'$ at $L$, and the donor's number of measured target hours in $7,\dots,L$ times a fixed sign. $\mathrm{BR}=g_b(\mathrm{S1})$ and $\mathrm{BR2}=g_{b'}(\mathrm{S2})$ with $g_b(p):=\operatorname{expit}\bigl(\operatorname{logit}(\min\{\max\{p,\epsilon\},1-\epsilon\})+b\bigr)$, $\epsilon=10^{-6}$, $b,b'$ fixed. Write $\mathcal A_6:=\{\mathrm{S1},\mathrm{BR},\mathrm{IC},\mathrm{IC\text{-}A},\mathrm{IC\text{-}O}\}$ (hour-6 arms), $\mathcal A_{\mathrm{SH}}:=\{\mathrm{SH},\mathrm{SH\text{-}A}\}$ (spliced arms) and $\mathcal A_{\mathrm{don}}$ for the three donor-only scores.
The \emph{information-set value} of an arm $a$ is $p_{\mathcal G_a}$ with $\mathcal G_a:=\sigma(\iota_a)\vee\sigma(E_a)$, where $\iota_a$ is the input the arm reads (for $\mathrm{BR}$ and $\mathrm{BR2}$ that of $\mathrm{S1}$ and $\mathrm{S2}$; for the donor-only scores $\iota_a:=\omega'_L$) and $E_a$ is the event its read conditions on: $\{T_{\mathrm{ev}}>L\}$ for $\mathrm{S1}$ and $\mathrm{BR}$ through the ratio form; the recipient being at risk at $L$ for the recipient's reads at origin $L$ ($\mathrm{S2}$, $\mathrm{BR2}$, the IC-type and spliced arms); the donor being at risk at $L$ for the donor's own $\mathrm{S2}$; and the sure event for $V'$ and the measured-hours score.

\emph{Metrics.} For a score $s$ (a real random variable) and $\alpha\in[0,1]$, $\mathrm{ROC}_s(\alpha)$ is the true-positive rate, at false-positive rate exactly $\alpha$, of the test that calls a stay positive when $s$ exceeds a threshold, ties at the threshold being broken by independent randomization (Lemma~1 of the proof). $\mathrm{AUROC}(s):=\mathbb P(s_1>s_0)+\tfrac12\mathbb P(s_1=s_0)$, with $s_1,s_0$ independent draws from the laws of $s$ given $Y=1$ and $Y=0$ under $\mathbb P_R$. For a forecast $q\in[0,1]$, $\mathrm{Brier}(q):=\mathbb E_R(Y-q)^2$ and $\mathrm{LogLoss}(q):=\mathbb E_R[-Y\log q-(1-Y)\log(1-q)]\in[0,\infty]$. For $\sigma$-fields $\mathcal F\subseteq\mathcal G$, $\mathrm{MI}(Y;\mathcal G\mid\mathcal F):=\mathbb E_R\,\mathrm{KL}\bigl(\mathrm{Ber}(p_{\mathcal G})\,\|\,\mathrm{Ber}(p_{\mathcal F})\bigr)$ in nats.

\smallskip
\noindent\textbf{(a) Dominance.} Let $\mathcal F\subseteq\mathcal G$. For every $\mathcal G$-measurable score $s$ and every $\alpha\in[0,1]$, $\mathrm{ROC}_s(\alpha)\le\mathrm{ROC}_{p_{\mathcal G}}(\alpha)$; hence $\mathrm{AUROC}(s)\le\mathrm{AUROC}(p_{\mathcal G})$. For every $\mathcal G$-measurable forecast $q$,
\[
\mathrm{Brier}(q)-\mathrm{Brier}(p_{\mathcal G})=\mathbb E_R(q-p_{\mathcal G})^2,\qquad
\mathrm{LogLoss}(q)-\mathrm{LogLoss}(p_{\mathcal G})=\mathbb E_R\,\mathrm{KL}\bigl(\mathrm{Ber}(p_{\mathcal G})\,\|\,\mathrm{Ber}(q)\bigr).
\]
In particular $\mathrm{Brier}(p_{\mathcal F})-\mathrm{Brier}(p_{\mathcal G})=\mathbb E_R(p_{\mathcal G}-p_{\mathcal F})^2\ge0$ and $\mathrm{LogLoss}(p_{\mathcal F})-\mathrm{LogLoss}(p_{\mathcal G})=\mathrm{MI}(Y;\mathcal G\mid\mathcal F)\in[0,\log 2]$, and each of the two vanishes if and only if $p_{\mathcal F}=p_{\mathcal G}$ almost surely.

\smallskip
\noindent\textbf{(b) Hour-6 arms; lead time is not a separate factor.} Under the roster rule (A1), $S1^{*}=p_{\mathcal F_6}$ almost surely. The information-set value of each arm in $\mathcal A_6$ is $S1^{*}$, and that of $\mathrm{S2}$ and $\mathrm{BR2}$ is $S2^{*}$. Consequently
\[
\mathrm{LogLoss}(S1^{*})-\mathrm{LogLoss}(S2^{*})=\mathrm{MI}(Y;\mathcal F_L\mid \mathcal F_6)\ge0,\qquad
\mathrm{Brier}(S1^{*})-\mathrm{Brier}(S2^{*})=\mathbb E_R(S2^{*}-S1^{*})^2\ge0,
\]
$\mathrm{ROC}_{S1^{*}}\le\mathrm{ROC}_{S2^{*}}$ pointwise and $\mathrm{AUROC}(S2^{*})\ge\mathrm{AUROC}(S1^{*})$. The two proper-score differences vanish if and only if $Y\perp X_{7:L}\mid \mathcal F_6$ under $\mathbb P_R$, in which case $S2^{*}=S1^{*}$. The ideal synchronized and hour-6 forecasts of the same window, under the same at-risk conditioning, therefore differ only through their information sets.

\smallskip
\noindent\textbf{(c) Another patient's hours.} Assume donor independence (D).
\begin{enumerate}
\item[(i)] $p_{\mathcal F_6\vee\sigma(\omega'_L)}=S1^{*}$ almost surely. Every $\mathcal F_6\vee\sigma(\omega'_L)$-measurable score $s$, in particular every arm in $\mathcal A_{\mathrm{SH}}\cup\mathcal A_{\mathrm{don}}$, satisfies $\mathrm{ROC}_s\le\mathrm{ROC}_{S1^{*}}$ pointwise and $\mathrm{AUROC}(s)\le\mathrm{AUROC}(S1^{*})$.
\item[(ii)] Every $\mathcal F_6\vee\sigma(\omega'_L)$-measurable forecast $q$ satisfies $\mathrm{Brier}(q)=\mathrm{Brier}(S1^{*})+\mathbb E_R(q-S1^{*})^2$.
\item[(iii)] The information-set value of $\mathrm{SH}$ and of $\mathrm{SH\text{-}A}$ is $S1^{*}$. The information-set value of each donor-only score is
\[
S1^{*}_{\mathrm{don}}:=p_{\sigma(\omega'_L)}=\mathbb E_R[S1^{*}\mid\omega'_L],
\]
with $\mathrm{AUROC}(S1^{*}_{\mathrm{don}})\le\mathrm{AUROC}(S1^{*})$, $\mathrm{Brier}(S1^{*}_{\mathrm{don}})-\mathrm{Brier}(S1^{*})=\mathbb E_R(S1^{*}-S1^{*}_{\mathrm{don}})^2$ and $\mathrm{LogLoss}(S1^{*}_{\mathrm{don}})-\mathrm{LogLoss}(S1^{*})=\mathrm{MI}(Y;\mathcal F_6\vee\sigma(\omega'_L)\mid\sigma(\omega'_L))$; the two proper-score gaps vanish if and only if $S1^{*}$ is almost surely equal to a function of $\omega'_L$.
\item[(iv)] If in addition exact matching (EM) holds, then for the implemented map $\psi$
\[
\mathrm{Brier}(\mathrm{SH})-\mathrm{Brier}(S1^{*})=\mathbb E_R(\mathrm{S2}-S1^{*})^2,
\]
and for the ideal map ($\psi(\omega_L)=S2^{*}$, spliced forecast $\mathrm{SH}^{*}:=\psi(\omega_6,X'_{7:L})$) one has $\mathrm{Brier}(\mathrm{SH}^{*})-\mathrm{Brier}(S1^{*})=\mathrm{Brier}(S1^{*})-\mathrm{Brier}(S2^{*})$: on the Brier score the ideal donor contrast $\mathrm{Brier}(\mathrm{SH}^{*})-\mathrm{Brier}(S2^{*})$ is exactly twice the ideal information gain $\mathrm{Brier}(S1^{*})-\mathrm{Brier}(S2^{*})$.
\end{enumerate}

\smallskip
\noindent\textbf{(d) Decomposition for the learned model.} For every real-valued metric $M$ of an arm's scores on the roster (population or sample, per seed or averaged over seeds, and in each bootstrap replicate that weights all arms alike) and every arm $a$,
\[
M(\mathrm{S2})-M(a)=\bigl[M(\mathrm{S2})-M(\mathrm{S1})\bigr]+\bigl[M(\mathrm{S1})-M(a)\bigr].
\]
From here on $M$ is the \emph{population} $\mathrm{AUROC}$, $-\mathrm{Brier}$ or $-\mathrm{LogLoss}$ under $\mathbb P_R$ ($-\mathrm{LogLoss}$ only when the log-losses involved are finite; for the two proper scores only arms that are forecasts, which excludes $V'$ and the measured-hours score). Let $\mathcal A:=\mathcal A_6$ without (D) and $\mathcal A:=\mathcal A_6\cup\mathcal A_{\mathrm{SH}}\cup\mathcal A_{\mathrm{don}}$ under (D), and define
\[
I^{*}:=M(S2^{*})-M(S1^{*}),\qquad \delta_2:=M(S2^{*})-M(\mathrm{S2}),\qquad \delta_a:=M(S1^{*})-M(a)\ \ (a\in\mathcal A).
\]
\begin{enumerate}
\item[(d1)] $I^{*}\ge0$, $\delta_2\ge0$ and $\delta_a\ge0$ for $a\in\mathcal A$, and
\[
M(\mathrm{S2})-M(\mathrm{S1})=I^{*}-\delta_2+\delta_{\mathrm{S1}},\qquad M(\mathrm{S1})-M(a)=\delta_a-\delta_{\mathrm{S1}}.
\]
\item[(d2)] For $a\in\mathcal A_6$ and, under (D), $a\in\mathcal A_{\mathrm{SH}}$, the arms $\mathrm{S1}$ and $a$ have the same information-set value $S1^{*}$: the second bracket compares two uses of the same information and vanishes when every arm is replaced by its information-set value. For a donor-only score $a$ (under (D)),
\[
\delta_a=\varepsilon+\delta'_a,\qquad \varepsilon:=M(S1^{*})-M(S1^{*}_{\mathrm{don}})\ge0,\qquad \delta'_a:=M(S1^{*}_{\mathrm{don}})-M(a)\ge0;
\]
$\varepsilon$ is the loss from not reading the recipient's own statics and hours $1$ to $6$, the second bracket equals $\varepsilon$ when every arm is replaced by its information-set value, and for the two proper scores $\varepsilon=0$ if and only if $S1^{*}$ is almost surely a function of $\omega'_L$.
\item[(d3)] For every score $b$ that is $\mathcal F_6$-measurable or, under (D), $\mathcal F_6\vee\sigma(\omega'_L)$-measurable (a forecast, for the proper scores), $M(b)\le M(S1^{*})$; hence $\delta_{\mathrm{S1}}\ge M(b)-M(\mathrm{S1})$ and $M(\mathrm{S2})-M(S1^{*})\le M(\mathrm{S2})-M(b)$. For every $\mathcal F_L$-measurable score (forecast) $b$, $\delta_2\ge M(b)-M(\mathrm{S2})$. In particular, for every nonempty $\mathcal B\subseteq\mathcal A$,
\[
M(\mathrm{S2})-M(S1^{*})\;\le\;M(\mathrm{S2})-\max_{a\in\mathcal B}M(a),
\]
so among the tested arms the smallest contrast $M(\mathrm{S2})-M(a)$ is the tightest upper bound they give on the advantage of the learned synchronized forecast over the ideal hour-6 forecast. It is not a bound on $I^{*}$. For sample metrics, the smallest estimated contrast is a selected estimate of that bound and is biased downwards.
\item[(d4)] $M(\mathrm{S2})-M(a)\le I^{*}$ if and only if $\delta_a\le\delta_2$. In particular the learned contrast $M(\mathrm{S2})-M(\mathrm{S1})$ is at most the ideal information gain $I^{*}$ exactly when the hour-6 read falls short of the ideal hour-6 forecast by no more than the synchronized read falls short of the ideal synchronized forecast.
\item[(d5)] $\mathrm{AUROC}(\mathrm{BR})=\mathrm{AUROC}(\mathrm{S1})$ and $\mathrm{AUROC}(\mathrm{BR2})=\mathrm{AUROC}(\mathrm{S2})$ whenever $g_b$ (respectively $g_{b'}$) is injective on a set carrying the values of the score (a set of probability one for the population metric, the set of observed values for the sample metric); $g_b$ is injective on a set $A\subseteq[0,1]$ if and only if $A$ has at most one point in $[0,\epsilon]$ and at most one in $[1-\epsilon,1]$.
\end{enumerate}

\paragraph{Assumptions.}
\begin{description}
\item[(A0) Roster law and fixed maps.] $\mathbb P(R)>0$ and $0<\pi<1$. All statements concern the population law $\mathbb P_R$ of one stay drawn from the roster population, together with its donor record $\omega'_L$ as described in (D). Held fixed: the five trained checkpoints; the TRAIN intercepts $b,b'$ of $\mathrm{BR}$ and $\mathrm{BR2}$; the TRAIN orientation sign of the measured-hours score; and, for each seed, the matching rule of \texttt{match\_donors} (the nearest stay of a different person in that seed's standardized hour-6 encoder state, among candidates whose record reaches $L$, exact ties going to the smallest stay identifier) together with its standardization constants. Every learned arm is then a fixed measurable function of the input listed in the Setting. No property of the encoder mask is needed for this, because every read is computed from the truncated prefix alone (the producer passes only the first $6$ or $L$ rows, statics and domain to the model). Nothing treats the twin as a Bayes filter: learned arms are arbitrary measurable functions of their inputs, and only the ideal forecasts $p_{\mathcal G}$ are conditional probabilities. Ties are randomized, matching the half-credit AUROC of the code. Two dependences are set aside. Each checkpoint is the epoch with the lowest validation log-loss (\texttt{r2\_joint\_train\_validate.py} L148--L198; selection status \texttt{VALIDATION\_SELECTED\_NO\_TEST\_ACCESS}), so on the validation split the learned maps depend on the stays' labels through that choice; and the standardization constants are roster-wide moments. Both affect the reading of the validation sample values, not the population statements.
\item[(A1) Roster rule (code fact).] Every roster stay is at risk at origin $L$: target $k$ is measured at hour $L$ and not past its threshold there, no earlier measured value crossed the threshold, and $L$ precedes the terminal (discharge or death) hour; in addition its label window is resolved. Hence $R\subseteq\{T_{\mathrm{ev}}>L\}$ and $R\subseteq\{\text{at risk at }L\}$. (\texttt{terminal\_semantics.py} L409--L414, L755--L835, L958--L975, L1073--L1079; \texttt{revision\_v3\_state\_sync\_producer\_v2.py} L187--L191.)
\item[(A2) Input structure (code fact).] The IC-type transforms replace hours $7$ to $L$ by fixed functions of hour $6$ and constants and leave statics, domain and hours $1$ to $6$ unchanged; SH and SH-A replace them by functions of the donor's hours $7$ to $L$ and the recipient's hour $6$; the donor-only scores are functions of $\omega'_L$; $\mathrm{S1}$ is a function of the hour-6 prefix read, clipped to $[0,1]$; $\mathrm{BR},\mathrm{BR2}$ are $g_b\circ\mathrm{S1}$, $g_{b'}\circ\mathrm{S2}$. (\texttt{revision\_v3\_state\_sync\_isolation\_v1.py} L526--L638, L983--L1045, L1118--L1150; \texttt{revision\_v3\_state\_sync\_cells\_v2.py} L172--L194.)
\item[(D) Donor independence] (used in (c), for $\mathcal A_{\mathrm{SH}}\cup\mathcal A_{\mathrm{don}}$ in (d), and for the $\mathcal F_6\vee\sigma(\omega'_L)$ case of (d3)). Under $\mathbb P_R$, the donor record $\omega'_L$ is conditionally independent of $(Y,X_{7:L})$ given $\mathcal F_6$; the proofs use only $Y\perp\omega'_L\mid \mathcal F_6$. By Lemma~4 of the proof, (D) holds whenever $\omega'_L=r(\omega_6,\Pi_{\mathrm{pool}})$ for a fixed measurable map $r$ and a donor pool $\Pi_{\mathrm{pool}}$ that is independent of the recipient's $(\omega_L,Y)$ under $\mathbb P_R$. The construction has this form if persons are independent, the seed's encoder and the standardization constants are fixed (A0), and $\Pi_{\mathrm{pool}}$ is the list of roster stays of persons other than the recipient's, with their records: \texttt{match\_donors} uses hour-6 encoder states, person identifiers and record lengths, no label enters it, and it never returns a stay of the recipient's person; the pool is the evaluable roster, whose membership depends only on each candidate's own record. Two departures make (D) an assumption that the construction supports rather than a code fact: the standardization constants are roster-wide means and standard deviations (\texttt{standardize\_states}, L681--L690), to which the recipient's person contributes with weight of order $1/|R|$ ($|R|$ between 551 and 7{,}707), and the checkpoints that define the encoder state were selected with validation data (A0).
\item[(EM) Exact matching] (idealization; used only in (c)(iv)). Under $\mathbb P_R$, $(\omega_6,X'_{7:L})$ has the same law as $(\omega_6,X_{7:L})$. The implementation takes the nearest different-person neighbor in the standardized 64-dimensional hour-6 state from a finite pool; (EM) would follow from exact state matches together with sufficiency of that state for the law of hours $7$ to $L$, neither of which holds exactly.
\item[(RI) Targets of the learned reads] (interpretive; used in no proof). Let $R_6:=R\cap\{\text{target }k\text{ measured at hour }6\}$, and call a head population-optimal when it maximizes the population training likelihood over measurable functions of the prefix. (RI) states that the population-optimal hour-$L$ read equals $S2^{*}$ almost surely on $R$ and the population-optimal hour-6 ratio read equals $S1^{*}$ almost surely on $R_6$. It holds, for example, if (i) a label $\tilde Y$ and a first-event time are defined for every stay at risk at the origin in question, for instance from the gap-free hourly path, with $\tilde Y=Y$ on $R$; (ii) training draws persons from the same population and censoring by measurement gaps is noninformative, so that the population-optimal head at origin $t$ returns the conditional law of the event process given $\omega_t$ among stays at risk at $t$; and (iii) $\tilde Y\perp\mathbf 1_R\mid\mathcal F_L$ among stays at risk at $L$, and $\tilde Y\perp\mathbf 1_R\mid \mathcal F_6$ among stays at risk at $6$ with $T_{\mathrm{ev}}>L$. It is an assumption because the roster requires the label window to be resolved. On $R\setminus R_6$ origin $6$ is not a supervised origin type, the population optimum does not determine the hour-6 read, and (RI) says nothing there. (RI) serves only to read $\delta_2$, and the part of $\delta_{\mathrm{S1}}$ that arises on $R_6$, as model error of the heads rather than as a mismatch between their targets and $S1^{*}$, $S2^{*}$.
\end{description}

\paragraph{Proof.}
\noindent\textbf{Conventions.} All probabilities and expectations are under $\mathbb P_R$. For $y\in\{0,1\}$ and a bounded or nonnegative random variable $Z$, $\mathbb E_yZ:=\mathbb E_R[Z\mathbf 1\{Y=y\}]/\mathbb P_R(Y=y)$. If $Z$ is bounded, or $[0,\infty]$-valued, and $\mathcal G$-measurable, the tower property (with monotone convergence in the second case and $0\cdot\infty:=0$) gives
\begin{equation}\label{eq:P1tower}
\mathbb E_R[Z\mathbf 1\{Y=1\}]=\mathbb E_R[Z\,p_{\mathcal G}],\qquad \mathbb E_R[Z\mathbf 1\{Y=0\}]=\mathbb E_R[Z(1-p_{\mathcal G})].
\end{equation}
For $\sigma$-fields $\mathcal H\subseteq\mathcal G$ the tower property also gives $p_{\mathcal H}=\mathbb E_R[p_{\mathcal G}\mid\mathcal H]$ almost surely. A test function is a $[0,1]$-valued random variable $\varphi$, the probability of calling the stay positive; its false-positive rate is $\mathbb E_0\varphi$ and its true-positive rate $\mathbb E_1\varphi$. Write $h(p):=-p\log p-(1-p)\log(1-p)$ and $\mathrm{KL}(p\|q):=p\log(p/q)+(1-p)\log((1-p)/(1-q))$ for the Bernoulli divergence, with $0\log(0/x):=0$ for $x\ge0$ and $a\log(a/0):=\infty$ for $a>0$. With these conventions, for all $p,q\in[0,1]$,
\begin{equation}\label{eq:P1ll}
-p\log q-(1-p)\log(1-q)=h(p)+\mathrm{KL}(p\|q)\in[0,\infty],
\end{equation}
$0\le h(p)\le\log2$, and $\mathrm{KL}(p\|q)\ge0$ with equality if and only if $p=q$ (Gibbs' inequality).

\medskip
\noindent\textbf{Lemma 1 (randomized threshold tests).} Let $s$ be a score with conditional laws $\mu_y:=\mathbb P_R(s\in\cdot\mid Y=y)$, let $U\sim\mathrm{Unif}(0,1)$ be independent of $(Y,s)$ (on a product extension of the probability space), and set $V_s:=\mu_0((s,\infty))+U\mu_0(\{s\})$. For $\alpha\in(0,1)$ let $c_\alpha:=\inf\{x\in\mathbb R:\mu_0((x,\infty))\le\alpha\}$, and let $\gamma_\alpha:=(\alpha-\mu_0((c_\alpha,\infty)))/\mu_0(\{c_\alpha\})$ if $\mu_0(\{c_\alpha\})>0$ and $\gamma_\alpha:=1$ otherwise. Then:
\begin{enumerate}
\item[(i)] $c_\alpha\in\mathbb R$ and $\mu_0((c_\alpha,\infty))\le\alpha\le\mu_0([c_\alpha,\infty))$, so $\gamma_\alpha\in[0,1]$;
\item[(ii)] almost surely $\mathbf 1\{V_s\le\alpha\}=\mathbf 1\{s>c_\alpha\}+\mathbf 1\{s=c_\alpha\}\mathbf 1\{U\le\gamma_\alpha\}$;
\item[(iii)] given $Y=0$, $V_s$ is uniform on $[0,1]$;
\item[(iv)] $\varphi^s_\alpha:=\mathbf 1\{s>c_\alpha\}+\gamma_\alpha\mathbf 1\{s=c_\alpha\}=\mathbb P(V_s\le\alpha\mid s,Y)$ is a $\sigma(s)$-measurable test function with $\mathbb E_0\varphi^s_\alpha=\alpha$ and $\mathbb E_1\varphi^s_\alpha=\mathrm{ROC}_s(\alpha)$, where we define $\mathrm{ROC}_s(\alpha):=\mathbb P_R(V_s\le\alpha\mid Y=1)$ for $\alpha\in[0,1]$;
\item[(v)] $\int_0^1\mathrm{ROC}_s(\alpha)\,d\alpha=\mathrm{AUROC}(s)$.
\end{enumerate}
By (ii) and (iv), $\mathrm{ROC}_s(\alpha)$ is the true-positive rate at exact false-positive rate $\alpha$ of the upper-threshold test of $s$ with randomization at the threshold; as $\alpha$ runs through $[\mu_0((c,\infty)),\mu_0([c,\infty))]$ for a tied value $c$, the rejected fraction of $\{s=c\}$ grows linearly, which is the linear interpolation across ties that corresponds to half credit for ties. This is the ROC curve used in the statement.

\emph{Proof of Lemma 1.} (i) $G_0(x):=\mu_0((x,\infty))$ is nonincreasing and right-continuous with limits $1$ at $-\infty$ and $0$ at $+\infty$. As $0<\alpha<1$, the set $\{x:G_0(x)\le\alpha\}$ is nonempty and bounded below, so $c_\alpha$ is finite; right-continuity gives $G_0(c_\alpha)\le\alpha$, and $G_0(x)>\alpha$ for $x<c_\alpha$ gives $\mu_0([c_\alpha,\infty))=\lim_{x\uparrow c_\alpha}G_0(x)\ge\alpha$.
(ii) If $s>c_\alpha$, then $V_s\le\mu_0([s,\infty))\le\mu_0((c_\alpha,\infty))\le\alpha$. If $s<c_\alpha$, then $V_s\ge G_0(s)>\alpha$. If $s=c_\alpha$, then $V_s\le\alpha$ if and only if $U\mu_0(\{c_\alpha\})\le\alpha-G_0(c_\alpha)$; when $\mu_0(\{c_\alpha\})>0$ this is $U\le\gamma_\alpha$, and when $\mu_0(\{c_\alpha\})=0$, (i) gives $G_0(c_\alpha)=\alpha$, so $V_s=\alpha$ and the event holds, as $\gamma_\alpha=1$ prescribes.
(iii) By (ii) and the independence of $U$, $\mathbb P_R(V_s\le\alpha\mid Y=0)=\mu_0((c_\alpha,\infty))+\gamma_\alpha\mu_0(\{c_\alpha\})=\alpha$ for $\alpha\in(0,1)$, in both cases of the definition of $\gamma_\alpha$. Since $0\le V_s\le\mu_0([s,\infty))\le1$, $V_s$ given $Y=0$ is uniform on $[0,1]$.
(iv) Immediate from (ii), because $U$ is independent of $(s,Y)$; then $\mathbb E_y\varphi^s_\alpha=\mathbb P_R(V_s\le\alpha\mid Y=y)$, which equals $\alpha$ for $y=0$ by (iii).
(v) By Tonelli, $\int_0^1\mathbb P_R(V_s\le\alpha\mid Y=1)\,d\alpha=\mathbb E_1\int_0^1\mathbf 1\{V_s\le\alpha\}\,d\alpha=1-\mathbb E_1V_s$. Given $Y=1$, $(s,U)$ has law $\mu_1\otimes\mathrm{Unif}(0,1)$, so with $s_0\sim\mu_0$ and $s_1\sim\mu_1$ independent, $\mathbb E_1V_s=\int[\mu_0((x,\infty))+\tfrac12\mu_0(\{x\})]\,\mu_1(dx)=\mathbb P(s_0>s_1)+\tfrac12\mathbb P(s_0=s_1)$. Hence the area is $1-\mathbb P(s_0>s_1)-\tfrac12\mathbb P(s_0=s_1)=\mathbb P(s_1>s_0)+\tfrac12\mathbb P(s_1=s_0)=\mathrm{AUROC}(s)$. \hfill$\square$

\medskip
\noindent\textbf{Lemma 2 (Neyman--Pearson on a sub-$\sigma$-field).} Let $\mathcal G$ be a $\sigma$-field, $t\in[0,1)$, and $\varphi^{*}$ a $\mathcal G$-measurable test function with $\varphi^{*}=1$ on $\{p_{\mathcal G}>t\}$, $\varphi^{*}=0$ on $\{p_{\mathcal G}<t\}$ and $\mathbb E_0\varphi^{*}=\alpha$. Then $\mathbb E_1\varphi\le\mathbb E_1\varphi^{*}$ for every $\mathcal G$-measurable test function $\varphi$ with $\mathbb E_0\varphi\le\alpha$.

\emph{Proof.} Pointwise $(\varphi^{*}-\varphi)(p_{\mathcal G}-t)\ge0$, and $p_{\mathcal G}-t=(1-t)p_{\mathcal G}-t(1-p_{\mathcal G})$. Taking expectations and applying \eqref{eq:P1tower} to the bounded $\mathcal G$-measurable $Z=\varphi^{*}-\varphi$,
\[
0\le(1-t)\,\mathbb E_R[(\varphi^{*}-\varphi)p_{\mathcal G}]-t\,\mathbb E_R[(\varphi^{*}-\varphi)(1-p_{\mathcal G})]=(1-t)\pi(\mathbb E_1\varphi^{*}-\mathbb E_1\varphi)-t(1-\pi)(\mathbb E_0\varphi^{*}-\mathbb E_0\varphi).
\]
The last term is $\le0$ because $t\ge0$ and $\mathbb E_0\varphi^{*}=\alpha\ge\mathbb E_0\varphi$. Hence $(1-t)\pi(\mathbb E_1\varphi^{*}-\mathbb E_1\varphi)\ge0$, and $(1-t)\pi>0$. \hfill$\square$

\medskip
\noindent\textbf{Lemma 3 (conditionally irrelevant information).} Let $\mathcal G,\mathcal H$ be $\sigma$-fields with $Y\perp\mathcal H\mid\mathcal G$ under $\mathbb P_R$. Then $p_{\mathcal G\vee\mathcal H}=p_{\mathcal G}$ almost surely. This applies in particular to $\mathcal H=\sigma(E)$ for an event $E$ with $\mathbb P_R(E)=1$, since a $\mathbb P_R$-trivial $\sigma$-field is independent of every random variable.

\emph{Proof.} For $A\in\mathcal G$ and $B\in\mathcal H$, conditional independence gives $\mathbb E_R[Y\mathbf 1_B\mid\mathcal G]=p_{\mathcal G}\,\mathbb P_R(B\mid\mathcal G)$, so $\mathbb E_R[Y\mathbf 1_{A\cap B}]=\mathbb E_R[\mathbf 1_Ap_{\mathcal G}\mathbb P_R(B\mid\mathcal G)]=\mathbb E_R[\mathbf 1_Ap_{\mathcal G}\mathbf 1_B]$, the last step by the tower property since $\mathbf 1_Ap_{\mathcal G}$ is $\mathcal G$-measurable. The sets $A\cap B$ form a $\pi$-system generating $\mathcal G\vee\mathcal H$, and the class of $C\in\mathcal G\vee\mathcal H$ with $\mathbb E_R[Y\mathbf 1_C]=\mathbb E_R[p_{\mathcal G}\mathbf 1_C]$ is a $\lambda$-system (it contains $\Omega$ and is closed under proper differences and increasing unions, by linearity and monotone convergence). By Dynkin's $\pi$--$\lambda$ theorem it is all of $\mathcal G\vee\mathcal H$. As $p_{\mathcal G}$ is $\mathcal G\vee\mathcal H$-measurable, it is a version of $p_{\mathcal G\vee\mathcal H}$. \hfill$\square$

\medskip
\noindent\textbf{Lemma 4 (a donor drawn from an independent pool).} Let $\Pi_{\mathrm{pool}}$ be a random element of a measurable space that is independent of $(\omega_L,Y)$ under $\mathbb P_R$, and let $\omega'_L=r(\omega_6,\Pi_{\mathrm{pool}})$ for a measurable map $r$. Then $(Y,X_{7:L})\perp\omega'_L\mid \mathcal F_6$ under $\mathbb P_R$; that is, (D) holds.

\emph{Proof.} Write $\Xi:=(Y,X_{7:L})$, so that $(\omega_6,\Xi)$ is a function of $(\omega_L,Y)$ and $\Pi_{\mathrm{pool}}$ is independent of $(\omega_6,\Xi)$. Let $f,g,k$ be bounded measurable functions and put $\Gamma(u):=\mathbb E_R\,g(r(u,\Pi_{\mathrm{pool}}))$, which is measurable in $u$ by Fubini's theorem. Because the joint law of $((\omega_6,\Xi),\Pi_{\mathrm{pool}})$ is the product of its marginals, Fubini's theorem gives
\[
\mathbb E_R\bigl[k(\omega_6)f(\Xi)g(r(\omega_6,\Pi_{\mathrm{pool}}))\bigr]=\mathbb E_R\bigl[k(\omega_6)f(\Xi)\Gamma(\omega_6)\bigr]=\mathbb E_R\bigl[k(\omega_6)\Gamma(\omega_6)\,\mathbb E_R[f(\Xi)\mid \mathcal F_6]\bigr].
\]
As $k$ is arbitrary and $\mathcal F_6=\sigma(\omega_6)$, $\mathbb E_R[f(\Xi)g(\omega'_L)\mid \mathcal F_6]=\Gamma(\omega_6)\,\mathbb E_R[f(\Xi)\mid \mathcal F_6]$ almost surely. With $f\equiv1$ this gives $\mathbb E_R[g(\omega'_L)\mid \mathcal F_6]=\Gamma(\omega_6)$, hence $\mathbb E_R[f(\Xi)g(\omega'_L)\mid \mathcal F_6]=\mathbb E_R[f(\Xi)\mid \mathcal F_6]\,\mathbb E_R[g(\omega'_L)\mid \mathcal F_6]$. Taking $f,g$ to be indicators gives the conditional independence. \hfill$\square$

\medskip
\noindent\textbf{Proof of (a).} \emph{ROC dominance.} Let $s$ be $\mathcal G$-measurable and write $p:=p_{\mathcal G}$, a $\mathcal G$-measurable score with values in $[0,1]$; let $\mu^p_0$ be its law given $Y=0$.
Case $\alpha\in(0,1)$. By Lemma~1(iv), $\varphi^s_\alpha$ is $\sigma(s)$-measurable, hence $\mathcal G$-measurable, with $\mathbb E_0\varphi^s_\alpha=\alpha$ and $\mathbb E_1\varphi^s_\alpha=\mathrm{ROC}_s(\alpha)$. Applied to $p$, Lemma~1 gives $\varphi^p_\alpha=\mathbf 1\{p>c\}+\gamma\mathbf 1\{p=c\}$ with $c:=c_\alpha(p)$ and $\mathbb E_0\varphi^p_\alpha=\alpha$. We have $c\in[0,1)$: for $x<0$, $\mu^p_0((x,\infty))=1>\alpha$, so $c\ge0$; and by \eqref{eq:P1tower}, $\mu^p_0(\{1\})=\mathbb E_R[(1-p)\mathbf 1\{p=1\}]/(1-\pi)=0$, so $\mu^p_0([1,\infty))=0<\alpha\le\mu^p_0([c,\infty))$ (Lemma~1(i)), which forces $c<1$. Thus $\varphi^p_\alpha$ satisfies the hypotheses of Lemma~2 with $t=c$, and $\mathrm{ROC}_s(\alpha)=\mathbb E_1\varphi^s_\alpha\le\mathbb E_1\varphi^p_\alpha=\mathrm{ROC}_p(\alpha)$.
Case $\alpha=1$. $V\le1$ always, so $\mathrm{ROC}_s(1)=\mathrm{ROC}_p(1)=1$.
Case $\alpha=0$. Since $U>0$ almost surely, $\mathbb P(V_s\le0\mid s,Y)=\mathbf 1\{\mu_0([s,\infty))=0\}=:\varphi^s_0$, a $\mathcal G$-measurable test function with $\mathbb E_0\varphi^s_0=0$ by Lemma~1(iii). Any $\mathcal G$-measurable test function $\varphi$ with $\mathbb E_0\varphi=0$ has $\mathbb E_R[\varphi(1-p)]=0$ by \eqref{eq:P1tower}, so $\varphi=0$ almost surely on $\{p<1\}$ and $\mathbb E_1\varphi=\mathbb E_R[\varphi p]/\pi\le\mathbb E_R[p\mathbf 1\{p=1\}]/\pi=\mathbb P_R(p=1\mid Y=1)$. For $s=p$, $\mu^p_0([1,\infty))=0$ shows $\{p=1\}\subseteq\{\mu^p_0([p,\infty))=0\}$, hence $\mathrm{ROC}_p(0)\ge\mathbb P_R(p=1\mid Y=1)\ge\mathrm{ROC}_s(0)$.
Integrating over $\alpha$ and using Lemma~1(v) gives $\mathrm{AUROC}(s)\le\mathrm{AUROC}(p_{\mathcal G})$.

\emph{Brier score.} For a $\mathcal G$-measurable forecast $q$, $\mathbb E_R(Y-q)^2=\mathbb E_R(Y-p_{\mathcal G})^2+\mathbb E_R(p_{\mathcal G}-q)^2+2\mathbb E_R[(Y-p_{\mathcal G})(p_{\mathcal G}-q)]$, and the cross term vanishes by \eqref{eq:P1tower} with the bounded $\mathcal G$-measurable $Z=p_{\mathcal G}-q$.

\emph{Log-loss.} The variables $-\log q$ and $-\log(1-q)$ are $\mathcal G$-measurable with values in $[0,\infty]$, so by \eqref{eq:P1tower} and \eqref{eq:P1ll}
\[
\mathrm{LogLoss}(q)=\mathbb E_R\bigl[-p_{\mathcal G}\log q-(1-p_{\mathcal G})\log(1-q)\bigr]=\mathbb E_R h(p_{\mathcal G})+\mathbb E_R\,\mathrm{KL}(p_{\mathcal G}\|q).
\]
With $q=p_{\mathcal G}$ this gives $\mathrm{LogLoss}(p_{\mathcal G})=\mathbb E_Rh(p_{\mathcal G})\le\log2<\infty$, and subtracting gives the stated identity in $[0,\infty]$.

\emph{Nested information sets.} If $\mathcal F\subseteq\mathcal G$, then $q=p_{\mathcal F}$ is $\mathcal G$-measurable, so $\mathrm{Brier}(p_{\mathcal F})-\mathrm{Brier}(p_{\mathcal G})=\mathbb E_R(p_{\mathcal G}-p_{\mathcal F})^2$ and $\mathrm{LogLoss}(p_{\mathcal F})-\mathrm{LogLoss}(p_{\mathcal G})=\mathbb E_R\mathrm{KL}(p_{\mathcal G}\|p_{\mathcal F})=\mathrm{MI}(Y;\mathcal G\mid\mathcal F)$. The log-loss identity applied to $\mathcal F$ with $q=p_{\mathcal F}$ gives $\mathrm{LogLoss}(p_{\mathcal F})=\mathbb E_Rh(p_{\mathcal F})$, so $\mathrm{MI}(Y;\mathcal G\mid\mathcal F)=\mathbb E_Rh(p_{\mathcal F})-\mathbb E_Rh(p_{\mathcal G})\in[0,\log2]$. Both differences are expectations of nonnegative quantities that vanish exactly where $p_{\mathcal G}=p_{\mathcal F}$ (Gibbs' inequality for the divergence), so each is zero if and only if $p_{\mathcal F}=p_{\mathcal G}$ almost surely. This proves (a).

\medskip
\noindent\textbf{Proof of (b).} \emph{Step 1.} By (A1), $R\subseteq\{T_{\mathrm{ev}}>L\}$, so $\mathbb P_R(T_{\mathrm{ev}}>L)=1$ and Lemma~3 gives $S1^{*}=p_{\mathcal F_6\vee\sigma(\{T_{\mathrm{ev}}>L\})}=p_{\mathcal F_6}$ almost surely.

\emph{Step 2 (information sets).} For $\mathrm{S1}$ and $\mathrm{BR}$ the input is $\iota=\omega_6$, so $\sigma(\iota)=\mathcal F_6$. For $a\in\{\mathrm{IC},\mathrm{IC\text{-}A},\mathrm{IC\text{-}O}\}$, (A2) gives $\iota_a=(\omega_6,\kappa_a(\omega_6))$ with $\kappa_a$ a fixed measurable map (the transforms listed in the Setting; the occurrence value $1$ written by $\kappa_{\mathrm{IC\text{-}O}}$ is a constant). Then $\sigma(\iota_a)\subseteq \mathcal F_6$ because $\iota_a$ is a measurable function of $\omega_6$, and $\mathcal F_6\subseteq\sigma(\iota_a)$ because $\omega_6$ is a coordinate of $\iota_a$; so $\sigma(\iota_a)=\mathcal F_6$. The conditioning event $E_a$ is $\{T_{\mathrm{ev}}>L\}$ for $\mathrm{S1}$ and $\mathrm{BR}$ and the recipient being at risk at $L$ for the reads at origin $L$; by (A1) both contain $R$, so $\mathbb P_R(E_a)=1$ and Lemma~3 gives $p_{\mathcal G_a}=p_{\sigma(\iota_a)}=p_{\mathcal F_6}=S1^{*}$. For $\mathrm{S2}$ and $\mathrm{BR2}$, $\sigma(\iota)=\sigma(\omega_L)=\mathcal F_L$ and the same argument gives $p_{\mathcal G}=S2^{*}$.

\emph{Step 3.} Apply (a) with $\mathcal F=\mathcal F_6\subseteq\mathcal G=\mathcal F_L$ and use $S1^{*}=p_{\mathcal F_6}$: this gives the log-loss and Brier identities, the pointwise ROC dominance of $S2^{*}$ over the $\mathcal F_6$-measurable (hence $\mathcal F_L$-measurable) score $S1^{*}$, and the AUROC inequality. By (a) the two proper-score differences vanish if and only if $p_{\mathcal F_L}=p_{\mathcal F_6}$ almost surely, that is $\mathbb P_R(Y=1\mid \mathcal F_6,X_{7:L})=\mathbb P_R(Y=1\mid \mathcal F_6)$, which for binary $Y$ is $Y\perp X_{7:L}\mid \mathcal F_6$. In that case $S2^{*}=S1^{*}$. Both $S1^{*}$ and $S2^{*}$ are forecasts of the same event $\{Y=1\}$ under the same law $\mathbb P_R$, so every difference in any metric is a functional of the joint law of $(Y,\mathcal F_6,X_{7:L})$ under $\mathbb P_R$ alone; the forecast origin enters only through the information set. For the implemented $\mathrm{S1}$ this matching of events is what the design $L=6+H$ secures: from origin $6$ the window $(L,L+H]$ is the bin interval $(H,2H]$, and dividing by the all-cause event-free probability $S_6(H)$ conditions on $\{T_{\mathrm{ev}}>L\}$.

\medskip
\noindent\textbf{Proof of (c).} Put $\mathcal G_D:=\mathcal F_6\vee\sigma(\omega'_L)$.
(i) By (D), $Y\perp\sigma(\omega'_L)\mid \mathcal F_6$ under $\mathbb P_R$, so Lemma~3 and Step~1 of (b) give $p_{\mathcal G_D}=p_{\mathcal F_6}=S1^{*}$. Each arm in $\mathcal A_{\mathrm{SH}}\cup\mathcal A_{\mathrm{don}}$ is $\mathcal G_D$-measurable: $\mathrm{SH}=\psi(\omega_6,X'_{7:L})$; $\mathrm{SH\text{-}A}=\psi(\omega_6,\kappa_{\mathrm{SH\text{-}A}}(X_6,X'_{7:L}))$ with $X_6$ a coordinate of $\omega_6$; the donor's own forecast $\psi(\omega'_L)$, its last value $V'$ and its oriented count of measured hours are functions of $\omega'_L$ (the orientation sign is fixed under (A0)). Part (i) is then (a) with $\mathcal G=\mathcal G_D$, whose ideal forecast is $S1^{*}$.
(ii) This is the Brier identity of (a) for $\mathcal G=\mathcal G_D$.
(iii) For $a\in\mathcal A_{\mathrm{SH}}$ the conditioning event (the recipient at risk at $L$) has $\mathbb P_R$-probability one by (A1), so Lemma~3 gives $p_{\mathcal G_a}=p_{\sigma(\iota_a)}$. The input $\iota_a$ is $(\omega_6,X'_{7:L})$ for $\mathrm{SH}$ and $(\omega_6,\kappa_{\mathrm{SH\text{-}A}}(X_6,X'_{7:L}))$ for $\mathrm{SH\text{-}A}$. In both cases $\omega_6$ is a coordinate of $\iota_a$, so $\mathcal F_6\subseteq\sigma(\iota_a)$, and $\iota_a$ is a measurable function of $(\omega_6,\omega'_L)$, so $\sigma(\iota_a)\subseteq\mathcal G_D$. By the tower property and (i), $p_{\sigma(\iota_a)}=\mathbb E_R[p_{\mathcal G_D}\mid\iota_a]=\mathbb E_R[S1^{*}\mid\iota_a]=S1^{*}$, the last equality because $S1^{*}=p_{\mathcal F_6}$ is $\mathcal F_6$-measurable and $\mathcal F_6\subseteq\sigma(\iota_a)$.
For a donor-only score, $\iota_a=\omega'_L$ and $E_a$ is either the sure event or the donor being at risk at $L$, which has $\mathbb P_R$-probability one because donors are roster stays. Lemma~3 gives $p_{\mathcal G_a}=p_{\sigma(\omega'_L)}$, and since $\sigma(\omega'_L)\subseteq\mathcal G_D$ the tower property and (i) give $p_{\sigma(\omega'_L)}=\mathbb E_R[S1^{*}\mid\omega'_L]=S1^{*}_{\mathrm{don}}$. Now apply (a) with $\mathcal F=\sigma(\omega'_L)\subseteq\mathcal G=\mathcal G_D$, for which $p_{\mathcal F}=S1^{*}_{\mathrm{don}}$ and $p_{\mathcal G}=S1^{*}$: this gives $\mathrm{AUROC}(S1^{*}_{\mathrm{don}})\le\mathrm{AUROC}(S1^{*})$ ($S1^{*}_{\mathrm{don}}$ is $\mathcal G_D$-measurable), the Brier gap $\mathbb E_R(S1^{*}-S1^{*}_{\mathrm{don}})^2$, the log-loss gap $\mathrm{MI}(Y;\mathcal G_D\mid\sigma(\omega'_L))$, and that both gaps vanish if and only if $S1^{*}=S1^{*}_{\mathrm{don}}$ almost surely. The last condition holds if and only if $S1^{*}$ is almost surely equal to a $\sigma(\omega'_L)$-measurable variable $f$: if it holds, take $f=S1^{*}_{\mathrm{don}}$; conversely, if $S1^{*}=f$ almost surely then $S1^{*}_{\mathrm{don}}=\mathbb E_R[f\mid\omega'_L]=f=S1^{*}$ almost surely. By Doob--Dynkin, $\sigma(\omega'_L)$-measurable variables are the measurable functions of $\omega'_L$.
(iv) Under (EM), write $S1^{*}=f_1(\omega_6)$ with $f_1$ measurable (Doob--Dynkin). Because $(\omega_6,X'_{7:L})$ and $(\omega_6,X_{7:L})$ have the same law and $(u,v)\mapsto(\psi(u,v)-f_1(u))^2$ is measurable,
\[
\mathbb E_R(\mathrm{SH}-S1^{*})^2=\mathbb E_R\bigl(\psi(\omega_6,X'_{7:L})-f_1(\omega_6)\bigr)^2=\mathbb E_R\bigl(\psi(\omega_6,X_{7:L})-f_1(\omega_6)\bigr)^2=\mathbb E_R(\mathrm{S2}-S1^{*})^2,
\]
and (ii) with $q=\mathrm{SH}$, a $\mathcal G_D$-measurable forecast, gives $\mathrm{Brier}(\mathrm{SH})-\mathrm{Brier}(S1^{*})=\mathbb E_R(\mathrm{S2}-S1^{*})^2$. For the ideal map take $\psi$ to be a version of $p_{\mathcal F_L}$ as a function of $\omega_L$; two versions agree almost everywhere for the law of $\omega_L$, which under (EM) is also the law of $(\omega_6,X'_{7:L})$, so $\mathrm{SH}^{*}$ does not depend on the version. Then $\mathbb E_R(S2^{*}-S1^{*})^2=\mathrm{Brier}(S1^{*})-\mathrm{Brier}(S2^{*})$ by (b), which gives $\mathrm{Brier}(\mathrm{SH}^{*})-\mathrm{Brier}(S1^{*})=\mathrm{Brier}(S1^{*})-\mathrm{Brier}(S2^{*})$ and hence $\mathrm{Brier}(\mathrm{SH}^{*})-\mathrm{Brier}(S2^{*})=2[\mathrm{Brier}(S1^{*})-\mathrm{Brier}(S2^{*})]$.

\medskip
\noindent\textbf{Proof of (d).} \emph{Identity.} Add and subtract $M(\mathrm{S1})$. For seed averages the identity holds because averaging is linear; in a bootstrap replicate that gives every arm the same stay weights (the whole-person bootstrap of the isolation run shares its weights across arms and seeds) each $M$ is computed on the same weighted sample, so the identity holds replicate by replicate. (The two-way bootstrap of donor contrasts uses different weights, so for those contrasts the identity concerns the point estimates.) The remaining claims concern the population metrics under $\mathbb P_R$; for sample metrics the signs below can fail through sampling error.

\emph{(d1).} $I^{*}\ge0$ by (b). $\mathrm{S2}=\psi(\omega_L)$ is $\mathcal F_L$-measurable by (A0), so (a) with $\mathcal G=\mathcal F_L$ gives $\mathrm{AUROC}(\mathrm{S2})\le\mathrm{AUROC}(S2^{*})$, $\mathrm{Brier}(\mathrm{S2})\ge\mathrm{Brier}(S2^{*})$ and $\mathrm{LogLoss}(\mathrm{S2})\ge\mathrm{LogLoss}(S2^{*})$, that is $\delta_2\ge0$ for each of the three metrics. Each $a\in\mathcal A_6$ is $\mathcal F_6$-measurable by Step~2 of (b), and $p_{\mathcal F_6}=S1^{*}$ by Step~1, so (a) with $\mathcal G=\mathcal F_6$ gives $\delta_a\ge0$. Under (D) each $a\in\mathcal A_{\mathrm{SH}}\cup\mathcal A_{\mathrm{don}}$ is $\mathcal G_D$-measurable with $p_{\mathcal G_D}=S1^{*}$ by (c)(i), so (a) with $\mathcal G=\mathcal G_D$ gives $\delta_a\ge0$. By definition $M(\mathrm{S2})=M(S2^{*})-\delta_2$ and $M(a)=M(S1^{*})-\delta_a$, so $M(\mathrm{S2})-M(\mathrm{S1})=I^{*}-\delta_2+\delta_{\mathrm{S1}}$ and $M(\mathrm{S1})-M(a)=\delta_a-\delta_{\mathrm{S1}}$; more generally $M(\mathrm{S2})-M(a)=I^{*}-\delta_2+\delta_a$.

\emph{(d2).} The information-set value of $\mathrm{S1}$ is $S1^{*}$ by (b). For $a\in\mathcal A_6$ it is $S1^{*}$ by (b), and for $a\in\mathcal A_{\mathrm{SH}}$ it is $S1^{*}$ by (c)(iii). Replacing $\mathrm{S1}$ and $a$ by their information-set values therefore turns the second bracket into $M(S1^{*})-M(S1^{*})=0$. For a donor-only score $a$ the information-set value is $S1^{*}_{\mathrm{don}}$ by (c)(iii), so the same replacement turns the second bracket into $M(S1^{*})-M(S1^{*}_{\mathrm{don}})=\varepsilon$, and $\varepsilon\ge0$ by (c)(iii). The split $\delta_a=\varepsilon+\delta'_a$ is algebra. As $a$ is $\sigma(\omega'_L)$-measurable (A2) and $p_{\sigma(\omega'_L)}=S1^{*}_{\mathrm{don}}$, (a) with $\mathcal G=\sigma(\omega'_L)$ gives $\delta'_a\ge0$ (for the proper scores $a$ is the donor's own forecast). For the proper scores $\varepsilon$ is the Brier or log-loss gap of (c)(iii), which vanishes if and only if $S1^{*}$ is almost surely a function of $\omega'_L$.

\emph{(d3).} If $b$ is $\mathcal F_6$-measurable, (a) with $\mathcal G=\mathcal F_6$ gives $M(b)\le M(p_{\mathcal F_6})=M(S1^{*})$; if $b$ is $\mathcal G_D$-measurable and (D) holds, (a) with $\mathcal G=\mathcal G_D$ gives the same through (c)(i). Hence $\delta_{\mathrm{S1}}=M(S1^{*})-M(\mathrm{S1})\ge M(b)-M(\mathrm{S1})$ and $M(\mathrm{S2})-M(S1^{*})\le M(\mathrm{S2})-M(b)$. If $b$ is $\mathcal F_L$-measurable, (a) with $\mathcal G=\mathcal F_L$ gives $M(b)\le M(S2^{*})$, so $\delta_2\ge M(b)-M(\mathrm{S2})$. Every $a\in\mathcal A$ is a score of the first kind by (d1), so for nonempty $\mathcal B\subseteq\mathcal A$ one has $\max_{a\in\mathcal B}M(a)\le M(S1^{*})$, which is the displayed bound; the smallest of the bounds $M(\mathrm{S2})-M(a)$, $a\in\mathcal B$, is the one attained at the maximizing arm.

\emph{(d4).} By (d1), $M(\mathrm{S2})-M(a)=I^{*}-\delta_2+\delta_a$, which is $\le I^{*}$ if and only if $\delta_a\le\delta_2$. With $a=\mathrm{S1}$ this is the stated condition.

\emph{(d5).} Let $g$ be nondecreasing and injective on a set $A$ that carries the values of $s$ (with probability one for the population metric; the observed values for the sample metric, whose AUROC is that of the empirical law). For $s_0,s_1$ with values in $A$, $\{g(s_1)>g(s_0)\}=\{s_1>s_0\}$ and $\{g(s_1)=g(s_0)\}=\{s_1=s_0\}$, so $\mathrm{AUROC}(g\circ s)=\mathrm{AUROC}(s)$. The map $g_b$ is the composition of $p\mapsto\min\{\max\{p,\epsilon\},1-\epsilon\}$, which is nondecreasing, constant on $[0,\epsilon]$ and on $[1-\epsilon,1]$ and strictly increasing on $[\epsilon,1-\epsilon]$, with the strictly increasing maps $\operatorname{logit}$, $z\mapsto z+b$ and $\operatorname{expit}$. It is therefore nondecreasing, and injective on a set $A\subseteq[0,1]$ if and only if $A$ has at most one point in $[0,\epsilon]$ and at most one in $[1-\epsilon,1]$. The same holds for $g_{b'}$ and $\mathrm{S2}$. This completes the proof.
\hfill$\square$

\paragraph{Reading of part (d) for the implemented arms.} The following facts about the code locate the shortfalls $\delta_a$; they are descriptions and are not used in the proof. The heads are supervised only at at-risk origins, where the target is measured (A1). $\mathrm{IC}$ and $\mathrm{IC\text{-}A}$ present the synchronized map with target occurrence $0$ at hour $L$, an origin type never supervised; $\mathrm{IC}$ also sets every non-target availability indicator to $0$ on hours $7$ to $L$, although that indicator records availability after forward filling and the isolation module itself treats a fall from $1$ to $0$ as atypical of real records. $\mathrm{IC\text{-}O}$ is read on the supervised origin type, but its input asserts that the target was measured at $L$ with its hour-6 value and that nothing else was newly measured. The ideal synchronized map would return a version of $p_{\mathcal F_L}$ evaluated at that input; such inputs may lie outside the support of $\omega_L$, where the value is not determined by the law of the data, and in any case it need not equal $S1^{*}$. For $\mathrm{SH}$ and $\mathrm{SH\text{-}A}$, (c)(iv) shows that under (EM) a loss relative to $S1^{*}$ arises even for an ideal forecaster on spliced records that are distributed exactly as real records: it is the cost of treating another patient's hours as the patient's own, and it does not require the spliced input to be unusual. Nearest-neighbor matching and the junction at hour $7$ can add a distributional component for the learned model. The hour-6 read $\mathrm{S1}$ is taken for every roster stay, including stays whose target is not measured at hour 6, for which origin 6 is not a supervised type; this part of $\delta_{\mathrm{S1}}$ has no model-error reading under (RI). For the IC-type arms and $\mathrm{SH}$, $\mathrm{SH\text{-}A}$, $\mathrm{S1}-a$ compares two uses of the same information set; for the donor-only scores it also contains $\varepsilon$, the loss from not reading the recipient's own statics and hours $1$ to $6$. Finally, (d3) gives lower bounds on $\delta_{\mathrm{S1}}$ and $\delta_2$ from any further hour-6 or hour-$L$ score, but none of the arms bounds either shortfall from above, so the arms do not decide whether $\delta_{\mathrm{S1}}\le\delta_2$.

\paragraph{Numerical check.} Table~\ref{tab:P1iso} re-reads the validation isolation run (job 13007). The identity of (d) holds with residual $0$ in all 12 settings. $\mathrm{S1}$ and $\mathrm{BR}$, whose AUROCs are equal by construction, are the best of the ten hour-6 and donor arms ($\mathrm{S1}$, $\mathrm{BR}$, $\mathrm{IC}$, $\mathrm{IC\text{-}A}$, $\mathrm{IC\text{-}O}$, $\mathrm{SH}$, $\mathrm{SH\text{-}A}$ and the three donor-only scores) in 12 of 12 settings, by $0.001$ to $0.034$ over the next arm, so $\mathrm{S2}-\mathrm{S1}$ ($+0.066$ to $+0.126$) is also the smallest of the upper bounds of (d3) that these arms give; $|\mathrm{AUROC}(\mathrm{BR})-\mathrm{AUROC}(\mathrm{S1})|=|\mathrm{AUROC}(\mathrm{BR2})-\mathrm{AUROC}(\mathrm{S2})|=0$ in all 60 seed-by-setting cells. $\mathrm{S1}-\mathrm{SH}$ is positive in 12 of 12 settings and in 59 of 60 seed-by-setting cells; it belongs to no registered Holm family, so these are point estimates, and since (c) orders $\mathrm{SH}$ only against $S1^{*}$, this ordering of two learned arms is consistent with (c) without being implied by it. The donor measured-hours score has AUROC $0.495$ to $0.517$; for the donor-only scores $\mathrm{S1}-a$ contains the information term $\varepsilon$ of (d2), which the arms do not separate from $\delta'_a$. On the cross-fitted (isotonic) scale $\mathrm{IC}$ is worse than $\mathrm{S1}$ on the Brier score in 12 of 12 settings (Holm-significant in 5) and on log-loss in 10 of 12 (Holm-significant in 2, both in that direction); the shipped log-loss clips probabilities to $[10^{-3},1-10^{-3}]$, so the identity with $\mathrm{MI}$ in (a) does not apply to these values and only their signs are read. Post hoc and unregistered (a new analysis of the per-stay predictions and labels of the previously read test panel, disclosed in Supplementary Note~2 and the test-read registry; no new predictions were made): on the test panel of jobs 10707/10708, the model's hour-6 cumulative incidence $F_6(H)$, which is $\mathcal F_6$-measurable and forecasts the earlier window $(6,L]$, exceeds $\mathrm{S1}$ in AUROC at arterial oxygen saturation $(30,24)$ by $0.021$ (95\% whole-person bootstrap interval $0.004$ to $0.039$) and at pulse-oximetry saturation $(30,24)$ by $0.050$ ($0.010$ to $0.089$); it exceeds $\mathrm{S1}$ in one further setting, by $0.002$, and falls below it in the other nine. Neither of the two larger excesses survives Holm correction over the 12 settings (adjusted one-sided $P=0.066$ and $0.1265$). By (d3) these are estimates of lower bounds on $\delta_{\mathrm{S1}}$, equal to 16\% and 33\% of $\mathrm{S2}-\mathrm{S1}$ in those two settings, so the condition $\delta_{\mathrm{S1}}\le\delta_2$ of (d4) cannot be taken for granted there. $\mathrm{S2}-\max(\mathrm{S1},F_6(H))$ is $+0.073$ to $+0.121$ and positive in 12 of 12 settings.

\begin{table}[ht]
\centering\small
\begin{tabular}{llrrrrrrrrr}
\hline
Target & $(L,H)$ & $n$ & S2 & S1 & S2$-$S1 & S1$-$IC & S1$-$IC-A & S1$-$IC-O & S1$-$SH & S2$-$IC\\
\hline
SaO$_2$ & (12, 6) & 6206 & 0.739 & 0.661 & 0.077 & 0.010$^{*}$ & 0.013 & 0.012 & 0.044 & 0.088\\
SaO$_2$ & (18, 12) & 3579 & 0.733 & 0.641 & 0.092 & 0.008 & 0.011 & 0.008 & 0.045 & 0.100\\
SaO$_2$ & (30, 24) & 1650 & 0.694 & 0.580 & 0.114 & 0.010 & 0.019 & 0.015 & 0.036 & 0.124\\
SpO$_2$ & (12, 6) & 2582 & 0.730 & 0.651 & 0.079 & 0.010 & 0.028 & 0.019 & 0.049 & 0.090\\
SpO$_2$ & (18, 12) & 1395 & 0.735 & 0.669 & 0.066 & 0.013 & 0.038 & 0.023 & 0.045 & 0.080\\
SpO$_2$ & (30, 24) & 551 & 0.693 & 0.576 & 0.117 & 0.025 & 0.033 & 0.024 & 0.014 & 0.142\\
Heart rate & (12, 6) & 7707 & 0.806 & 0.694 & 0.112 & 0.006$^{*}$ & 0.005 & 0.005 & 0.058 & 0.118\\
Heart rate & (18, 12) & 4799 & 0.787 & 0.660 & 0.126 & 0.006 & 0.001 & 0.001 & 0.058 & 0.133\\
Heart rate & (30, 24) & 2229 & 0.724 & 0.616 & 0.108 & 0.010 & 0.009 & 0.009 & 0.052 & 0.118\\
Respiratory rate & (12, 6) & 7127 & 0.735 & 0.650 & 0.085 & 0.021$^{*}$ & 0.019 & 0.019 & 0.046 & 0.106\\
Respiratory rate & (18, 12) & 4301 & 0.745 & 0.635 & 0.110 & 0.036$^{*}$ & 0.034 & 0.034 & 0.044 & 0.146\\
Respiratory rate & (30, 24) & 1899 & 0.694 & 0.590 & 0.104 & 0.027$^{*}$ & 0.029 & 0.029 & 0.039 & 0.131\\
\hline
\end{tabular}
\caption{\textbf{Isolation arms on the validation split (Proposition 1).} Job 13007 (\texttt{source\_tables/revision\_v3/state\_sync\_isolation\_13007\_v1/score/isolation\_metrics.csv}); raw scale, AUROC averaged over the five per-seed values; $n$ is the roster size. All contrasts are positive in all 12 settings. Entries are rounded independently, so a row need not add exactly; unrounded, $\mathrm{S2}-\mathrm{IC}=(\mathrm{S2}-\mathrm{S1})+(\mathrm{S1}-\mathrm{IC})$ with residual $0$. $\mathrm{S2}-\mathrm{S1}$ is Holm-significant in 12 of 12 settings (a replication family of the protocol, not independent of the test-set result; correction within its 12 settings); an asterisk marks the 5 settings in which $\mathrm{IC}-\mathrm{S1}$ is Holm-significantly negative in the primary family of 36 contrasts. $\mathrm{S1}-\mathrm{SH}$ is not a registered contrast and is reported as a point estimate. By Proposition~1(d), $\mathrm{S2}-\mathrm{S1}$ is the learned contrast $I^{*}-\delta_2+\delta_{\mathrm{S1}}$ and $\mathrm{S1}-\mathrm{IC}=\delta_{\mathrm{IC}}-\delta_{\mathrm{S1}}$ compares two reads of the same information. These are sample values on a split used for checkpoint selection; the proposition concerns population quantities.}
\label{tab:P1iso}
\end{table}

%% file: supplementary/supp_theory/p2.tex
\paragraph{Proposition 2 (The four AUROCs of the difference-in-differences do not identify the current-value hypothesis).}
\emph{Standard and specific parts.} Lemmas 0 to 3 of the proof are elementary Gaussian facts. The statements (a) to (c) concern the difference-in-differences of this evaluation and are specific to it.

\noindent\emph{Setting.} Fix a target $k$ and a setting $(L,H)$. All laws are conditional on the roster event $R$, and $\pi:=P(Y=1\mid R)\in(0,1)$. A record $r=(r_{1:6},r_{7:L})$ is an input that the frozen forecaster $\mathcal M$ reads at origin $L$; $r_{1:6}$ denotes its statics, domain indicator and hours $1$ to $6$, $r_{7:L}$ its hours $7$ to $L$ (values and masks), and $v(r)$ its direction-signed target value at hour $L$. For a roster stay with record $r$ let $V:=v(r)$. Its donor is a roster stay of another person whose target is measured at hour $L$, and $V'$ is the donor's value there. The spliced record $r^{\mathrm{SH}}$ keeps $r_{1:6}$ and takes hours $7$ to $L$ from the donor, so $v(r^{\mathrm{SH}})=V'$; the own-value splice $r^{\mathrm{OV}}$ is $r^{\mathrm{SH}}$ with its hour-$L$ target value set to $V$ (both targets are measured at $L$, so the masks agree). The arms are $\mathrm{S2}:=\mathcal M(r)$, $\mathrm{SH}:=\mathcal M(r^{\mathrm{SH}})$, $\mathrm{OV}:=\mathcal M(r^{\mathrm{OV}})$, $\mathrm{LV}_{\mathrm{own}}:=V$ and $\mathrm{LV}_{\mathrm{donor}}:=V'$, each scored against the recipient's label $Y$ by the AUROC with half credit for ties; OV is proposed here and was not run. The analysis protocol fixes in advance the following descriptive contrast, Holm-corrected within the 12 settings; by the protocol's own rules it cannot change a decision outcome and acts only on the choice of reported wording:
\[
\mathrm{DiD}_{\mathrm{AUROC}}:=(\mathrm{AUROC}_{\mathrm{S2}}-\mathrm{AUROC}_{\mathrm{SH}})-(\mathrm{AUROC}_{\mathrm{own}}-\mathrm{AUROC}_{\mathrm{donor}}).
\]

\noindent\emph{Latent-threshold Gaussian model (LT).} There are random vectors $w\in\mathbb R^{p}$ (statics and hours $1$ to $6$) and $x\in\mathbb R^{q}$ (hours $7$ to $L$) with $\operatorname{Cov}((w,x))$ nonsingular, $\mathcal F_6=\sigma(w)$, $\mathcal F_L=\sigma(w,x)$ and $V=x_1$, and a latent index $\eta$ (a symbol of this section only), such that $(w,x,\eta)$ is jointly Gaussian, $E\eta=0$, $\operatorname{Var}\eta=1$, $\operatorname{Var}(\eta\mid w,x)>0$ and $Y=1\{\eta>t_\pi\}$ with $t_\pi:=\Phi^{-1}(1-\pi)$. A record is identified with a point $(w,x)$ of $\mathbb R^{p+q}$, with $r_{1:6}=w$, $r_{7:L}=x$ and $v(r)=x_1$. $A_\pi(\rho)$ is the AUROC for $Y$ of any strictly increasing function of a variable $J$ such that $(J,\eta)$ is bivariate Gaussian with $\operatorname{Var}J>0$ and $\operatorname{corr}(J,\eta)=\rho$ (Gaussianity of $J$ alone is not enough); $A_\pi$ is a continuous strictly increasing bijection of $[-1,1]$ onto $[0,1]$ (Lemma 1). The latent correlation of arm $a$ is $\rho_a:=A_\pi^{-1}(\mathrm{AUROC}_a)$; write $\rho_{\mathrm{own}}$, $\rho_{\mathrm{donor}}$ for the two last-value arms, $\Delta_{\mathrm{LV}}:=\rho_{\mathrm{own}}-\rho_{\mathrm{donor}}$ and $\mathrm{DiD}_\rho:=(\rho_{\mathrm{S2}}-\rho_{\mathrm{SH}})-\Delta_{\mathrm{LV}}$. Put $m_V:=E[V\mid w]$, $e_V:=V-m_V$, $m_\eta:=E[\eta\mid w]$, $\theta_V:=\operatorname{Cov}(\eta,e_V)/\operatorname{Var}(e_V)$ and $\iota^2:=\operatorname{Var}(e_V)/\operatorname{Var}(V)$. Let $\mu$ be the affine function on $\mathbb R^{p+q}$ with $\mu(w,x)=E[\eta\mid w,x]$ almost surely; it is unique because $\operatorname{Cov}((w,x))$ is nonsingular. Put $\sigma^2_{\eta\mid L}:=\operatorname{Var}(\eta\mid w,x)$.

\noindent\emph{Exact matching (EM).} The donor's hours $7$ to $L$ are represented by $x'\in\mathbb R^{q}$ with $V'=x'_1$, and conditionally on $(w,x,\eta)$ the law of $x'$ is the conditional law of $x$ given $w$. The spliced record is then $(w,x')$, and the own-value record is $(w,x'')$ with $x''_1:=x_1$ and $x''_j:=x'_j$ for $j\ge2$.

\noindent\emph{Hypotheses and forecasters.} $\mathcal M$ obeys the current-value restriction (CVR) if $\mathcal M(r)=g(r_{1:6},v(r))$ for some function $g$ and every record $r$ in a domain that contains the natural, spliced and own-value records. In LT, $\mathcal M$ obeys linear CVR if moreover $g(r_{1:6},v)=\Psi(U_6+\beta_V v)$ with $\Psi$ strictly increasing, $U_6$ an affine function of $w$ and $\beta_V>0$; then $I:=U_6+\beta_V V$, $I_{\mathrm{SH}}:=U_6+\beta_V V'$ and $w_V:=\beta_V\operatorname{SD}(V)/\operatorname{SD}(I)$. The data obey the current-state hypothesis $H_{\mathrm{CS}}$ if $Y\perp x\mid(w,V)$, and current-value sufficiency (CVS) if $Y\perp(w,x)\mid V$. The ideal forecaster is $\mathcal M^*(r):=\Phi\{(\mu(r)-t_\pi)/\sigma_{\eta\mid L}\}$; it is defined at every record and is a version of $P(Y=1\mid\mathcal F_L,R)$. $\mathrm{S1}^*:=P(Y=1\mid w,R)$ is the ideal hour-6 forecast and $\mathrm{CS}^*:=P(Y=1\mid w,V,R)$ the ideal current-state forecast, with latent correlations $\rho_{\mathrm{S1}^*}$ and $\rho_{\mathrm{CS}^*}$. By Definition 2, $\mathrm{AUROC}(\mathcal M^*)-\mathrm{AUROC}(\mathrm{CS}^*)$ is the personalization beyond the current state.

\noindent Parts (a)(i) to (a)(iv) and (b) assume LT and $\Delta_{\mathrm{LV}}>0$. Part (a)(v), except its last sentence, and part (c) need neither.

\medskip
\noindent\textbf{(a) The null value of the difference-in-differences.}

\noindent(i) Under EM, if $\mathcal M$ obeys linear CVR, then $\rho_{\mathrm{S2}}-\rho_{\mathrm{SH}}=w_V\,\Delta_{\mathrm{LV}}$, so $\mathrm{DiD}_\rho=(w_V-1)\Delta_{\mathrm{LV}}$. $\mathrm{DiD}_\rho$ is zero if and only if $\operatorname{Var}(U_6)+2\beta_V\operatorname{Cov}(U_6,V)=0$, negative if and only if this quantity is positive, and positive if and only if it is negative, which requires $\operatorname{Cov}(U_6,V)<0$.

\noindent(ii) Under EM and $H_{\mathrm{CS}}$, the ideal forecaster obeys linear CVR at every record, with $U_6=m_\eta-\theta_V m_V$ and $\beta_V=\theta_V>0$. Its $\mathrm{DiD}_\rho$ is zero if and only if $\operatorname{Var}(m_\eta)=\theta_V^2\operatorname{Var}(m_V)$, and negative if and only if $\operatorname{Var}(m_\eta)>\theta_V^2\operatorname{Var}(m_V)$.

\noindent(iii) Any donor assignment. Without EM, if $\mathcal M$ obeys linear CVR, the indices $I,I_{\mathrm{SH}},V,V'$ have positive variances, and $(I_{\mathrm{SH}},\eta)$ and $(V',\eta)$ are bivariate Gaussian, then
\[
\rho_{\mathrm{S2}}-s_{\mathrm{SH}}\,\rho_{\mathrm{SH}}=w_V\,(\rho_{\mathrm{own}}-s_{\mathrm{LV}}\,\rho_{\mathrm{donor}}),\qquad s_{\mathrm{SH}}:=\frac{\operatorname{SD}(I_{\mathrm{SH}})}{\operatorname{SD}(I)},\quad s_{\mathrm{LV}}:=\frac{\operatorname{SD}(V')}{\operatorname{SD}(V)},
\]
so that $\mathrm{DiD}_\rho=(w_V-1)\Delta_{\mathrm{LV}}+(s_{\mathrm{SH}}-1)\rho_{\mathrm{SH}}-w_V(s_{\mathrm{LV}}-1)\rho_{\mathrm{donor}}$. Under EM, $s_{\mathrm{SH}}=s_{\mathrm{LV}}=1$. If instead, conditionally on $(w,x,\eta)$, $x'$ has the conditional law of $x$ given $z(w)$ for a linear map $z$ (exact matching on a linear summary of the first six hours, a Gaussian stand-in for the frozen hour-6 state $z_6$ on which donors are matched), then $(I_{\mathrm{SH}},\eta)$ and $(V',\eta)$ are bivariate Gaussian, $s_{\mathrm{LV}}=1$, and $s_{\mathrm{SH}}^2=1+2\beta_V\operatorname{Cov}(U_6,E[V\mid z(w)]-m_V)/\operatorname{Var}(I)$, which equals $1$ if and only if $\operatorname{Cov}(U_6,E[V\mid z(w)]-m_V)=0$.

\noindent(iv) Scale. $\mathrm{DiD}_{\mathrm{AUROC}}=A_\pi(\rho_{\mathrm{S2}})-A_\pi(\rho_{\mathrm{SH}})-A_\pi(\rho_{\mathrm{own}})+A_\pi(\rho_{\mathrm{donor}})$, and $\mathrm{DiD}_\rho=0$ does not imply $\mathrm{DiD}_{\mathrm{AUROC}}=0$: the proof exhibits an LT model with EM and $H_{\mathrm{CS}}$ whose ideal forecaster has $\mathrm{DiD}_\rho=0$ and $\mathrm{DiD}_{\mathrm{AUROC}}=+0.0023$ at $\pi=0.25$.

\noindent(v) Current-value sufficiency. Suppose CVS holds and $\Psi_0(v):=P(Y=1\mid V=v,R)$ is strictly increasing. Then $r\mapsto\Psi_0(v(r))$ is a version of the ideal forecaster, and for it $\mathrm{AUROC}_{\mathrm{S2}}=\mathrm{AUROC}_{\mathrm{own}}$ and $\mathrm{AUROC}_{\mathrm{SH}}=\mathrm{AUROC}_{\mathrm{donor}}$ exactly; hence $\phi(\mathrm{AUROC}_{\mathrm{S2}})-\phi(\mathrm{AUROC}_{\mathrm{SH}})-\phi(\mathrm{AUROC}_{\mathrm{own}})+\phi(\mathrm{AUROC}_{\mathrm{donor}})=0$ for every function $\phi$, for every donor assignment, and for the empirical AUROCs on every finite roster. Neither Gaussianity nor matching is used. The same holds for every forecaster $r\mapsto\tilde\Psi(v(r))$ with $\tilde\Psi$ strictly increasing. In LT under CVS (where $\operatorname{Var}V>0$ because $\operatorname{Cov}((w,x))$ is nonsingular), $\Psi_0$ is strictly increasing if and only if $\rho_{\mathrm{own}}>0$, and under EM this follows from $\Delta_{\mathrm{LV}}>0$.

\medskip
\noindent\textbf{(b) Non-identification.} Assume EM and consider the ideal forecaster. Let $Q=(\rho_{\mathrm{S2}},\rho_{\mathrm{SH}},\rho_{\mathrm{own}},\rho_{\mathrm{donor}})$ with $0<\rho_{\mathrm{SH}}<\rho_{\mathrm{S2}}<1$ and $\Delta_{\mathrm{LV}}>0$. Call $Q$ attainable by a class of LT models if some model of the class, with EM, gives $Q$ as the latent correlations of the four arms of its ideal forecaster.

\noindent(i) In every LT model with EM and $\operatorname{Var}\mu>0$: $\rho_{\mathrm{SH}}\rho_{\mathrm{S2}}=\operatorname{Var}(m_\eta)=\rho_{\mathrm{S1}^*}^2$, hence $\rho_{\mathrm{SH}}\le\rho_{\mathrm{S1}^*}\le\rho_{\mathrm{S2}}$, and $\rho_{\mathrm{S2}}(\rho_{\mathrm{S2}}-\rho_{\mathrm{SH}})=\operatorname{Var}(\mu-m_\eta)$.

\noindent(ii) $Q$ is attainable under $H_{\mathrm{CS}}$ if and only if
\[
\text{(R1)}\quad \Delta_{\mathrm{LV}}\le[\rho_{\mathrm{S2}}(\rho_{\mathrm{S2}}-\rho_{\mathrm{SH}})]^{1/2}
\quad\text{and}\quad
\text{(R2)}\quad \rho_{\mathrm{donor}}^2\le\rho_{\mathrm{SH}}\rho_{\mathrm{S2}}\Bigl[1-\frac{\Delta_{\mathrm{LV}}^2}{\rho_{\mathrm{S2}}(\rho_{\mathrm{S2}}-\rho_{\mathrm{SH}})}\Bigr].
\]
(R2 implies R1.)

\noindent(iii) R1 and R2 hold in every LT model with EM, whether or not $H_{\mathrm{CS}}$ holds, and $Q$ is attainable by a model that violates $H_{\mathrm{CS}}$ if and only if R1 and R2 hold with strict inequality.

\noindent(iv) Consequently every attainable $Q$ is attainable under $H_{\mathrm{CS}}$, and every $Q$ that satisfies R1 and R2 strictly is attainable both under $H_{\mathrm{CS}}$ and under its failure. For known $\pi$ the four AUROCs and $Q$ determine each other, so no function of the four AUROCs, including a difference-in-differences on any scale, can reject $H_{\mathrm{CS}}$ at an attainable quadruple, or confirm it at a quadruple that satisfies R1 and R2 strictly. The same holds when the AUROC of the ideal hour-6 forecast $\mathrm{S1}^*$ is added, because $\rho_{\mathrm{S1}^*}=(\rho_{\mathrm{SH}}\rho_{\mathrm{S2}})^{1/2}$ is a function of $Q$ by (i). For the ideal forecaster in LT, CVR holds if and only if $H_{\mathrm{CS}}$ holds, so the same is true of CVR; and because the two models that attain such a $Q$ carry one forecaster that obeys CVR and one that does not, the conclusion persists for every class of forecasters that contains the ideal ones. Part (b) does not model the measured-hours arms, the second pre-specified difference-in-differences built from them, the donor's own forecast, or the learned same-information arms, and says nothing about them.

\noindent(v) Size of a departure, and a label-free completion. In every LT model with EM whose ideal forecaster has a quadruple $Q$ as above, put $\sigma^2_{\mathrm{new}}:=\rho_{\mathrm{S2}}(\rho_{\mathrm{S2}}-\rho_{\mathrm{SH}})$, $\iota_0^2:=\Delta_{\mathrm{LV}}^2/\sigma^2_{\mathrm{new}}$ and $\iota_{\max}^2:=1-\rho_{\mathrm{donor}}^2/(\rho_{\mathrm{SH}}\rho_{\mathrm{S2}})$, and let $\varsigma:=\operatorname{corr}^2(\mu-m_\eta,e_V)$, the share of the new-hour latent variance $\operatorname{Var}(\mu-m_\eta)$ carried by the current value (through its part $e_V$ not predicted by hours $1$ to $6$); $1-\varsigma$ is the share that lies outside it. Then:
\begin{enumerate}
\item[(1)] $\iota^2=1-\operatorname{corr}(V,V')$, a quantity computable without labels;
\item[(2)] $\iota_0\le\iota\le\iota_{\max}$ and $\varsigma=\iota_0^2/\iota^2\in[\iota_0^2/\iota_{\max}^2,\,1]$; $H_{\mathrm{CS}}$ holds if and only if $\varsigma=1$, if and only if $\operatorname{corr}(V,V')=1-\iota_0^2$;
\item[(3)] $\rho_{\mathrm{CS}^*}^2=\rho_{\mathrm{SH}}\rho_{\mathrm{S2}}+\varsigma\,\sigma^2_{\mathrm{new}}$, hence $\rho_{\mathrm{S2}}^2-\rho_{\mathrm{CS}^*}^2=(1-\varsigma)\sigma^2_{\mathrm{new}}\le(1-\iota_0^2/\iota_{\max}^2)\sigma^2_{\mathrm{new}}$ and
\[
\mathrm{AUROC}(\mathcal M^*)-\mathrm{AUROC}(\mathrm{CS}^*)\le A_\pi(\rho_{\mathrm{S2}})-A_\pi\bigl((\rho_{\mathrm{SH}}\rho_{\mathrm{S2}}+\sigma^2_{\mathrm{new}}\iota_0^2/\iota_{\max}^2)^{1/2}\bigr);
\]
\item[(4)] every $\varsigma\in[\iota_0^2/\iota_{\max}^2,1]$ occurs in an LT model with EM whose ideal forecaster has quadruple $Q$, so the bounds in (2) and (3) are attained.
\end{enumerate}
Within this idealized class the four AUROCs therefore bound the size of a departure from $H_{\mathrm{CS}}$ without detecting one, and adding $\operatorname{corr}(V,V')$ identifies $H_{\mathrm{CS}}$. This identification uses EM: (5) if EM is replaced by exact matching on a linear summary $z(w)$ as in (a)(iii), then $\operatorname{corr}(V,V')=\operatorname{Var}(E[V\mid z(w)])/\operatorname{Var}(V)\le1-\iota^2$, with equality if and only if $E[V\mid z(w)]=m_V$ almost surely, so $1-\operatorname{corr}(V,V')$ overstates $\iota^2$; the quadruple, and with it $\iota_0$, is changed by the same layer, and the direction of the net error of the completed test is not determined here.

\medskip
\noindent\textbf{(c) A label-free test of the current-value restriction.} This part uses neither LT, EM nor labels, except for the AUROCs in (ii).

\noindent(i) If $\mathcal M$ obeys CVR, then $\mathrm{OV}=\mathrm{S2}$ for every stay, and $\mathrm{SH}=\mathrm{S2}$ for every recipient-donor pair with $V'=V$. Conversely, for records $r,\tilde r$ let the own-value splice $r^{\mathrm{OV}}(\tilde r)$ be the record whose first part is $r_{1:6}$ and whose second part is $\tilde r_{7:L}$ with the hour-$L$ target value replaced by $v(r)$. If $\mathcal N$ is a set of records whose target is measured at hour $L$ that contains $r^{\mathrm{OV}}(\tilde r)$ for all $r,\tilde r\in\mathcal N$, and $\mathcal M(r)=\mathcal M(r^{\mathrm{OV}}(\tilde r))$ for all $r,\tilde r\in\mathcal N$, then $\mathcal M$ obeys CVR on $\mathcal N$. A single stay with $\mathrm{OV}\neq\mathrm{S2}$, or a single pair with $V'=V$ and $\mathrm{SH}\neq\mathrm{S2}$, shows that $\mathcal M$ does not obey CVR.

\noindent(ii) Quantitative form. For every forecaster, $\mathrm{S2}-\mathrm{SH}=(\mathrm{S2}-\mathrm{OV})+(\mathrm{OV}-\mathrm{SH})$, where OV and SH are read from records that differ only in the hour-$L$ target value, and S2 and OV from records that differ only in the other entries of hours $7$ to $L$. The label-free ratio $\delta_{\mathrm{OV}}:=E[(\mathrm{S2}-\mathrm{OV})^2]/E[(\mathrm{S2}-\mathrm{SH})^2]$ (defined when the denominator is positive; it is not bounded by $1$ in general) is zero under CVR. For every $\varepsilon>0$,
\[
|\mathrm{AUROC}_{\mathrm{S2}}-\mathrm{AUROC}_{\mathrm{OV}}|\le P(|\mathrm{S2}-\mathrm{OV}|>\varepsilon\mid Y=1)+P(|\mathrm{S2}-\mathrm{OV}|>\varepsilon\mid Y=0)+P(|\mathrm{S2}_1-\mathrm{S2}_0|\le2\varepsilon),
\]
with $\mathrm{S2}_1\sim\mathcal L(\mathrm{S2}\mid Y=1)$ and $\mathrm{S2}_0\sim\mathcal L(\mathrm{S2}\mid Y=0)$ independent; the same bound holds for empirical AUROCs with the empirical class-conditional laws. A forecaster with $|\mathrm{S2}-\mathrm{OV}|\le\varepsilon$ for all but a small fraction of the stays of each class therefore has $\mathrm{AUROC}_{\mathrm{OV}}$ close to $\mathrm{AUROC}_{\mathrm{S2}}$ when few pairs of positive and negative stays have S2 within $2\varepsilon$ of each other.

\paragraph{Assumptions.}
\begin{enumerate}
\item[(A0)] \emph{Roster and splice as implemented.} All laws are conditional on the roster event $R$ of the setting, with $\pi=P(Y=1\mid R)\in(0,1)$. Every roster stay and every donor has its target measured at hour $L$: the roster admits only at-risk origins, which require a measurement at $L$, donors are drawn from the same roster, and the last-value score refuses a reading that is not taken at $L$ (none of the 40,702 last-value inputs of the test-set last-value comparison, the only split for which per-stay last values are shipped, was taken at another hour). Hence $V$ and $V'$ exist and $v(r^{\mathrm{SH}})=V'$. The splice replaces hours $7$ to $L$ (values and masks) and leaves statics, domain and hours $1$ to $6$ unchanged. The arms are scored on the same recipients and labels, so $\pi$ is common to them.
\item[(A1)] \emph{Latent-threshold Gaussian model (LT)}, used in (a)(i) to (a)(iv), the last sentence of (a)(v), and (b): $(w,x,\eta)$ jointly Gaussian with $\operatorname{Cov}((w,x))$ nonsingular, $\operatorname{Var}\eta=1$, $\operatorname{Var}(\eta\mid w,x)>0$, $Y=1\{\eta>t_\pi\}$, $V=x_1$, and each arm a strictly increasing function of a linear index. Nonsingularity makes $\mu$, $m_V$ and $m_\eta$ unique affine functions defined at every record, which is what lets the ideal forecaster be evaluated on own-value records, whose values need not occur among natural records of a degenerate Gaussian model. LT is an idealization. On the roster $V$ is truncated at the crossing threshold (an at-risk origin is not past the threshold at $L$ and has no earlier observed crossing); some targets take few distinct values, so $V$ has ties (independent-pair tie probability 0.29 to 0.34 for pulse-oximetry saturation, 0.09 to 0.12 for arterial saturation, at most 0.005 for heart and respiratory rate); the real record contains exact dependencies (masks, carried-forward values) that a nonsingular covariance excludes; the label is a first crossing in $(L,L+H]$ that competes with discharge and death; and the learned forecaster (a Transformer with a competing-risks head) is not known to be an increasing function of a linear index. The current-value share $w_V$ is defined only for linear-index forecasters.
\item[(A2)] \emph{Exact matching (EM)}, used in (a)(i), (a)(ii), the last clause of (a)(v) and (b): conditionally on $(w,x,\eta)$ the donor's $x'$ has the conditional law of $x$ given $w$. EM combines donor independence with equality of conditional laws. Donor independence holds by construction provided stays of different persons are independent: donors are chosen from hour-6 encoder states only, never from labels, and never from the recipient's person. Equality of conditional laws is idealized in two layers. First, the implementation takes the nearest neighbor, with reuse, where EM draws from the exact conditional law. Second, it matches on a summary of $w$, the standardized 64-dimensional frozen hour-6 state $z_6$, and exact matching on a summary equals EM only if the summary carries all of the information in $w$ about $x$. Part (a)(iii) removes EM from the identity of (a)(i) and shows that exact matching on a linear summary keeps $s_{\mathrm{LV}}=1$ but not $s_{\mathrm{SH}}=1$; part (b) is proved under EM and under neither of the two implemented layers, and (b)(v)(5) shows that the summary layer alone already biases $\operatorname{corr}(V,V')$. Size, by simulation only (Gaussian worlds that reproduce four observed quadruples, with a 64-dimensional state whose geometry was calibrated to the shipped donor summaries): nearest-neighbor matching moved the ratio $(\rho_{\mathrm{S2}}-\rho_{\mathrm{SH}})/\Delta_{\mathrm{LV}}$ away from $w_V$ by $0.006$ to $0.025$ (Monte Carlo standard errors $0.001$ to $0.004$) and lowered $\rho_{\mathrm{SH}}$ by $0.005$ to $0.040$ relative to EM. With the roster truncation and the ties of (A1) added, singly or together, the ratio differed from $w_V$ by $-0.027$ to $+0.067$ under nearest-neighbor matching and by $-0.023$ to $+0.048$ under exact matching within the truncated roster. In these worlds the null value of (a)(i) therefore holds only to about $0.02$ on the latent scale, comparable to several observed latent differences-in-differences.
\item[(A3)] \emph{CVR on records.} In (a)(iii) and (c) the restriction must hold on the spliced and own-value records themselves, which need not lie in the support of natural records. Under EM spliced records have the law of natural records, so (a)(i) and the necessity parts of (b) need CVR only almost surely on natural records; the ideal forecaster of (a)(ii) and (b)(iv) obeys CVR at every record because $\mu$ is unique.
\item[(A4)] $\Delta_{\mathrm{LV}}>0$. On validation (job 13007) the latent $\Delta_{\mathrm{LV}}$ lies between 0.252 and 0.381, and the AUROC-scale difference between 0.115 and 0.188, in all 12 settings.
\item[(A5)] In (a)(v): $v\mapsto P(Y=1\mid V=v,R)$ strictly increasing. In LT under CVS this is equivalent to $\rho_{\mathrm{own}}>0$ and follows from (A4) under EM.
\item[(A6)] \emph{Ideal forecaster} in (a)(ii), (a)(iv), (a)(v) and (b): the forecast $P(Y=1\mid\mathcal F_L,R)$ on the roster. Treating the twin's forecast as this object is an idealization that the record does not support. Under LT, EM and roster ignorability (Definition 1), a forecaster that is ideal at origins $6$ and $L$ has hour-$6$ latent correlation $(\rho_{\mathrm{SH}}\rho_{\mathrm{S2}})^{1/2}$ by (b)(i); computed from the twin's S2 and SH, this value exceeds the latent correlation of the twin's own hour-6 arm S1 in 12 of 12 settings, by $0.003$ to $0.071$ (sampling uncertainty not propagated). So the twin is not ideal at both origins, or LT or EM fails; the record does not say which, although the nearest-neighbor layer, which lowered $\rho_{\mathrm{SH}}$ in the simulations cited under (A2), would push the computed value down rather than up. Part (b)(iv) states what survives for learned forecasters; the bounds of (b)(v) hold only for the ideal forecaster. The only implemented property used elsewhere is that the forecast at origin $L$ is a function of the record read at $L$.
\end{enumerate}

\paragraph{Proof.}
\noindent Throughout, $\operatorname{Var}$, $\operatorname{Cov}$, $\operatorname{corr}$ and $\operatorname{SD}$ are population moments of the roster law, $\mathcal L(\cdot)$ denotes a law and $\perp$ denotes (conditional) independence. For a score $S$ and independent draws $S_1\sim\mathcal L(S\mid Y=1)$ and $S_0\sim\mathcal L(S\mid Y=0)$, $\mathrm{AUROC}(S):=E\,\mathrm{hc}(S_1,S_0)$ with $\mathrm{hc}(a,b):=1\{a>b\}+\tfrac12 1\{a=b\}$; the empirical AUROC is the same functional of the empirical class-conditional laws. In LT, conditional expectations given $w$ or $(w,x)$ are taken in their affine versions, which are unique because $\operatorname{Cov}((w,x))$ is nonsingular; the same nonsingularity gives $\operatorname{Var}V>0$ and $\operatorname{Var}(e_V)=\operatorname{Var}(x_1\mid w)>0$.

\paragraph{Lemma 0 (invariance).} If $\Psi$ is strictly increasing on the range of $S$, then $\mathrm{AUROC}(\Psi(S))=\mathrm{AUROC}(S)$, in the population and empirically.

\noindent\emph{Proof.} Pointwise, $\{\Psi(S_1)>\Psi(S_0)\}=\{S_1>S_0\}$ and $\{\Psi(S_1)=\Psi(S_0)\}=\{S_1=S_0\}$. $\square$

\paragraph{Lemma 1 (AUROC link).} Let $(J,\eta)$ be bivariate Gaussian with $\operatorname{Var}J>0$ and $\rho:=\operatorname{corr}(J,\eta)$, and let $\Psi$ be strictly increasing. Then $\mathrm{AUROC}(\Psi(J))=A_\pi(\rho)$, where $A_\pi(1)=1$, $A_\pi(-1)=0$ and, for $|\rho|<1$,
\[
A_\pi(\rho)=E\,\Phi\!\left(\frac{\rho\,(\eta_1-\eta_0)}{\sqrt{2(1-\rho^2)}}\right),
\]
with $\eta_1\sim\mathcal L(\eta\mid\eta>t_\pi)$ and $\eta_0\sim\mathcal L(\eta\mid\eta\le t_\pi)$ independent. $A_\pi$ is continuous and strictly increasing on $[-1,1]$ with $A_\pi(0)=\tfrac12$, and a constant score has AUROC $\tfrac12=A_\pi(0)$. Hence $\rho_a=\operatorname{corr}(J_a,\eta)$ for every arm whose score is a strictly increasing function of a variable $J_a$ with $(J_a,\eta)$ bivariate Gaussian and $\operatorname{Var}J_a>0$.

\noindent\emph{Proof.} By Lemma 0, and because an increasing affine map preserves the AUROC, we may take $J=\rho\eta+(1-\rho^2)^{1/2}\varepsilon$ with $\varepsilon\sim N(0,1)$ independent of $\eta$; this representation uses the joint Gaussianity of $(J,\eta)$. Let $(\eta_1,\varepsilon_1)$ and $(\eta_0,\varepsilon_0)$ be independent, $\eta_y\sim\mathcal L(\eta\mid Y=y)$, $\varepsilon_y\sim N(0,1)$ independent of everything else; then $J_y:=\rho\eta_y+(1-\rho^2)^{1/2}\varepsilon_y$ has the law of $J$ given $Y=y$. If $|\rho|<1$, $\varepsilon_0-\varepsilon_1\sim N(0,2)$ is independent of $(\eta_1,\eta_0)$, so $J_1-J_0$ has a density, $P(J_1=J_0)=0$, and conditioning on $(\eta_1,\eta_0)$ gives $P(J_1>J_0)=E\,\Phi\{\rho(\eta_1-\eta_0)/\sqrt{2(1-\rho^2)}\}$. If $\rho=\pm1$, $J_y=\pm\eta_y$ and $\eta_1>t_\pi\ge\eta_0$ almost surely, so the AUROC is $1$ or $0$. Since $D:=\eta_1-\eta_0>0$ almost surely, $k\mapsto\Phi(kD)$ is strictly increasing, and $\rho\mapsto\rho/\sqrt{2(1-\rho^2)}$ is a strictly increasing bijection of $(-1,1)$ onto $\mathbb R$; so $A_\pi$ is strictly increasing on $(-1,1)$ with values in $(0,1)$, continuous there by dominated convergence, and tends to $1$ and $0$ as $\rho\to1$ and $\rho\to-1$, again by dominated convergence. $A_\pi(0)=\Phi(0)=\tfrac12$. A constant score has $P(S_1=S_0)=1$. $\square$

\noindent\emph{Note.} Gaussianity of $J$ alone does not suffice. $J:=\eta\,(1\{|\eta|<1.5\}-1\{|\eta|\ge1.5\})$ is standard normal by symmetry, and $\operatorname{corr}(J,\eta)=E[\eta^2;|\eta|<1.5]-E[\eta^2;|\eta|\ge1.5]=-0.044$; at $\pi=0.25$ its AUROC is about $0.67$ (two Monte Carlo runs of $2\times10^6$ draws gave $0.6671$ and $0.6679$), while $A_{0.25}(-0.044)=0.479$.

\paragraph{Lemma 2 (conditional independence in LT).} Assume LT, and let $\mathcal G\subseteq\mathcal F_L$ be generated by finitely many affine functions of $(w,x)$. Then $Y\perp(w,x)\mid\mathcal G$ if and only if $\mu=E[\eta\mid\mathcal G]$ almost surely. In particular, CVS holds if and only if $\mu=E[\eta\mid V]$, and $H_{\mathrm{CS}}$ holds if and only if $\mu=E[\eta\mid w,V]=m_\eta+\theta_Ve_V$.

\noindent\emph{Proof.} Let $M_{\mathcal G}:=E[\eta\mid\mathcal G]$ and $D:=\mu-M_{\mathcal G}$. In the jointly Gaussian family both conditional expectations are affine in $(w,x)$, so $D$ is Gaussian with mean zero and uncorrelated with the generators of $\mathcal G$; being jointly Gaussian with them, it is independent of $\mathcal G$. Conditionally on $(w,x)$, $\eta\sim N(\mu,\sigma^2_{\eta\mid L})$, so $P(Y=1\mid w,x)=\Phi\{(M_{\mathcal G}+D-t_\pi)/\sigma_{\eta\mid L}\}$. As $Y$ is binary, $Y\perp(w,x)\mid\mathcal G$ holds if and only if $P(Y=1\mid w,x)$ is almost surely equal to a $\mathcal G$-measurable variable (which is then $P(Y=1\mid\mathcal G)$). If $D=0$ almost surely this holds. Conversely, if $P(Y=1\mid w,x)=Z$ almost surely with $Z$ $\mathcal G$-measurable, then $Z\in(0,1)$ and $D=\sigma_{\eta\mid L}\Phi^{-1}(Z)+t_\pi-M_{\mathcal G}$ almost surely, so $D$ is $\mathcal G$-measurable and independent of $\mathcal G$, hence almost surely constant, and $ED=0$ gives $D=0$. For $\mathcal G=\sigma(w,V)=\sigma(w,e_V)$: $e_V$ is Gaussian and uncorrelated with $w$, hence independent of $w$, and the projection of $\eta$ on $\operatorname{span}\{1,w,e_V\}$ is $m_\eta+\{\operatorname{Cov}(\eta,e_V)/\operatorname{Var}e_V\}e_V$. $\square$

\paragraph{Lemma 3 (splicing under exact matching).} Assume LT and EM, and put $\xi:=x-E[x\mid w]$ and $\xi':=x'-E[x\mid w]$. Then $\xi\perp w$; $\xi'$ is independent of $(w,x,\eta)$ and has the law of $\xi$; $(w,x,\eta,x')$ is jointly Gaussian; and $(w,x')\overset{d}{=}(w,x)$. For an affine $J=c_0+c^\top w+b^\top x$ with $\operatorname{Var}J>0$ and its splice $J':=c_0+c^\top w+b^\top x'$,
\[
\operatorname{Var}J'=\operatorname{Var}J,\qquad \operatorname{Cov}(J',\eta)=\operatorname{Cov}(E[J\mid w],\eta),\qquad \operatorname{corr}(J,\eta)-\operatorname{corr}(J',\eta)=\frac{\operatorname{Cov}(b^\top\xi,\eta)}{\operatorname{SD}(J)}.
\]

\noindent\emph{Proof.} $\xi$ is Gaussian and uncorrelated with $w$, hence independent of $w$, and $\operatorname{Cov}\xi$ is the constant conditional covariance of $x$ given $w$. By EM, conditionally on $(w,x,\eta)$, $x'\sim N(E[x\mid w],\operatorname{Cov}\xi)$, so $\xi'\sim N(0,\operatorname{Cov}\xi)$ conditionally on $(w,x,\eta)$; thus $\xi'$ is independent of $(w,x,\eta)$ with the law of $\xi$, and $(w,x,\eta,\xi')$, hence $(w,x,\eta,x')$, is jointly Gaussian. Since $\xi$ and $\xi'$ are both independent of $w$ with the same law, $(w,E[x\mid w]+\xi')\overset{d}{=}(w,E[x\mid w]+\xi)$. Write $J=E[J\mid w]+b^\top\xi$ and $J'=E[J\mid w]+b^\top\xi'$. Independence gives $\operatorname{Var}J'=\operatorname{Var}E[J\mid w]+\operatorname{Var}(b^\top\xi)=\operatorname{Var}J$ and $\operatorname{Cov}(J',\eta)=\operatorname{Cov}(E[J\mid w],\eta)$. Subtracting from $\operatorname{Cov}(J,\eta)=\operatorname{Cov}(E[J\mid w],\eta)+\operatorname{Cov}(b^\top\xi,\eta)$ and dividing by $\operatorname{SD}(J)=\operatorname{SD}(J')$ (with $\operatorname{SD}\eta=1$) gives the last identity. $\square$

\paragraph{Proof of (a)(i).} By Lemma 3, $(w,x')\overset{d}{=}(w,x)$, so a spliced record has the law of a natural record and an almost-sure version of CVR applies to it. Linear CVR gives $\mathrm{S2}=\Psi(I)$ and $\mathrm{SH}=\Psi(I_{\mathrm{SH}})$, where $I_{\mathrm{SH}}=U_6+\beta_Vx'_1$ is the splice of $I=U_6+\beta_Vx_1$; the last-value scores are $V$ and its splice $V'$. Lemma 3 with $J=V$ and Lemma 1 give $\operatorname{SD}(V')=\operatorname{SD}(V)$ and
\begin{equation*}
\Delta_{\mathrm{LV}}=\operatorname{corr}(V,\eta)-\operatorname{corr}(V',\eta)=\frac{\operatorname{Cov}(e_V,\eta)}{\operatorname{SD}(V)}.\tag{P2.1}
\end{equation*}
Since $\Delta_{\mathrm{LV}}>0$, $\operatorname{Cov}(e_V,\eta)>0$. As $I-E[I\mid w]=\beta_Ve_V$ with $e_V\perp w$, $\operatorname{Var}I=\operatorname{Var}(U_6+\beta_Vm_V)+\beta_V^2\operatorname{Var}e_V>0$, and Lemma 3 with $J=I$, Lemma 1 and (P2.1) give
\[
\rho_{\mathrm{S2}}-\rho_{\mathrm{SH}}=\frac{\beta_V\operatorname{Cov}(e_V,\eta)}{\operatorname{SD}(I)}=\frac{\beta_V\operatorname{SD}(V)}{\operatorname{SD}(I)}\,\Delta_{\mathrm{LV}}=w_V\,\Delta_{\mathrm{LV}}.
\]
Hence $\mathrm{DiD}_\rho=(w_V-1)\Delta_{\mathrm{LV}}$, which has the sign of $w_V-1$. Because $\beta_V>0$ and $\operatorname{Var}I=\operatorname{Var}U_6+2\beta_V\operatorname{Cov}(U_6,V)+\beta_V^2\operatorname{Var}V$, $w_V=\beta_V\operatorname{SD}(V)/\operatorname{SD}(I)$ is smaller than, equal to or larger than $1$ according as $\operatorname{Var}U_6+2\beta_V\operatorname{Cov}(U_6,V)$ is positive, zero or negative. The last case requires $\operatorname{Cov}(U_6,V)<0$ because $\operatorname{Var}U_6\ge0$. $\square$

\paragraph{Proof of (a)(ii).} By (P2.1), $\theta_V=\operatorname{Cov}(\eta,e_V)/\operatorname{Var}e_V>0$. By Lemma 2, $H_{\mathrm{CS}}$ gives $\mu=m_\eta+\theta_Ve_V=(m_\eta-\theta_Vm_V)+\theta_VV$ almost surely. Both sides are affine functions of $(w,x)$, and two affine functions that agree almost surely under a law with nonsingular covariance agree at every point of $\mathbb R^{p+q}$; so $\mu(w,x)=U_6(w)+\theta_Vx_1$ for every record, with $U_6:=m_\eta-\theta_Vm_V$ affine in $w$. Hence $\mathcal M^*(r)=\Psi(U_6(r_{1:6})+\theta_V v(r))$ at every record, natural, spliced or own-value, with $\Psi(\upsilon):=\Phi\{(\upsilon-t_\pi)/\sigma_{\eta\mid L}\}$ strictly increasing: the ideal forecaster obeys linear CVR with $\beta_V=\theta_V$. Because $e_V\perp w$, $\operatorname{Cov}(U_6,V)=\operatorname{Cov}(U_6,m_V)$, and
\begin{align*}
\operatorname{Var}U_6+2\theta_V\operatorname{Cov}(U_6,m_V)&=\operatorname{Var}m_\eta-2\theta_V\operatorname{Cov}(m_\eta,m_V)+\theta_V^2\operatorname{Var}m_V+2\theta_V\operatorname{Cov}(m_\eta,m_V)-2\theta_V^2\operatorname{Var}m_V\\
&=\operatorname{Var}m_\eta-\theta_V^2\operatorname{Var}m_V.
\end{align*}
Part (i) gives the conclusion. $\square$

\paragraph{Proof of (a)(iii).} CVR holds on the spliced records, so $I-I_{\mathrm{SH}}=\beta_V(V-V')$ for every stay and
\[
\operatorname{Cov}(I,\eta)-\operatorname{Cov}(I_{\mathrm{SH}},\eta)=\beta_V\{\operatorname{Cov}(V,\eta)-\operatorname{Cov}(V',\eta)\}.
\]
Lemma 1 applies to each of the four pairs, and $\operatorname{SD}\eta=1$, so $\operatorname{Cov}(I,\eta)=\rho_{\mathrm{S2}}\operatorname{SD}(I)$, $\operatorname{Cov}(I_{\mathrm{SH}},\eta)=\rho_{\mathrm{SH}}\operatorname{SD}(I_{\mathrm{SH}})$, $\operatorname{Cov}(V,\eta)=\rho_{\mathrm{own}}\operatorname{SD}(V)$ and $\operatorname{Cov}(V',\eta)=\rho_{\mathrm{donor}}\operatorname{SD}(V')$. Substituting and dividing by $\operatorname{SD}(I)$ gives the identity. Writing $\rho_{\mathrm{S2}}-\rho_{\mathrm{SH}}=(\rho_{\mathrm{S2}}-s_{\mathrm{SH}}\rho_{\mathrm{SH}})+(s_{\mathrm{SH}}-1)\rho_{\mathrm{SH}}$ and $w_V(\rho_{\mathrm{own}}-s_{\mathrm{LV}}\rho_{\mathrm{donor}})=w_V\Delta_{\mathrm{LV}}-w_V(s_{\mathrm{LV}}-1)\rho_{\mathrm{donor}}$, and subtracting $\Delta_{\mathrm{LV}}$, gives the expression for $\mathrm{DiD}_\rho$. No property of the donor assignment beyond the Gaussianity of the two pairs is used. Under EM, Lemma 3 with $J=I$ and $J=V$ gives $\operatorname{SD}(I_{\mathrm{SH}})=\operatorname{SD}(I)$ and $\operatorname{SD}(V')=\operatorname{SD}(V)$, which recovers (i).

Now let $z$ be linear and, conditionally on $(w,x,\eta)$, $x'\sim\mathcal L(x\mid z(w))$. As $(z(w),x)$ is jointly Gaussian, $\mathcal L(x\mid z(w))=N(E[x\mid z(w)],\Sigma_z)$ with $E[x\mid z(w)]$ affine in $w$ and $\Sigma_z$ constant; so $x'=E[x\mid z(w)]+\xi''$ with $\xi''\sim N(0,\Sigma_z)$ independent of $(w,x,\eta)$, and $(w,x,\eta,x')$ is jointly Gaussian. Hence $(I_{\mathrm{SH}},\eta)$ and $(V',\eta)$ are bivariate Gaussian, $(z(w),x')\overset{d}{=}(z(w),x)$, and in particular $V'\overset{d}{=}V$, so $s_{\mathrm{LV}}=1$ and $\operatorname{Var}V'=\operatorname{Var}V>0$; also $\operatorname{Var}I_{\mathrm{SH}}\ge\beta_V^2\operatorname{Var}\xi''_1=\beta_V^2\operatorname{Var}(V\mid z(w))\ge\beta_V^2\operatorname{Var}(e_V)>0$, so the hypotheses of (iii) hold. Put $m^z_V:=E[V\mid z(w)]$. As $U_6$ is a function of $w$ and $E[V'\mid w]=m^z_V$, $\operatorname{Cov}(U_6,V')=\operatorname{Cov}(U_6,m^z_V)$, while $\operatorname{Cov}(U_6,V)=\operatorname{Cov}(U_6,m_V)$. Therefore $\operatorname{Var}I_{\mathrm{SH}}-\operatorname{Var}I=2\beta_V\operatorname{Cov}(U_6,m^z_V-m_V)$, and dividing by $\operatorname{Var}I>0$ gives the formula for $s_{\mathrm{SH}}^2$. $\square$

\paragraph{Proof of (a)(iv).} The display restates $\rho_a=A_\pi^{-1}(\mathrm{AUROC}_a)$. For the example let $w=(w_1,w_2)$, $\varepsilon_V$ and $\varepsilon_0$ be independent standard normal, $m_V:=2^{-1/2}w_1$, $e_V:=2^{-1/2}\varepsilon_V$, $V:=m_V+e_V$, $x:=V$, $m:=0.18^{1/2}(\tfrac12w_1+\tfrac{\sqrt3}{2}w_2)$ and $\eta:=m+0.6\,e_V+0.8\,\varepsilon_0$, with donors drawn as in EM. $\operatorname{Cov}((w_1,w_2,V))$ is nonsingular (its determinant is $\tfrac12$). Then $\operatorname{Var}V=\operatorname{Var}\eta=1$, $m_\eta=m$, and $\mu=m+0.6\,e_V$ is $\sigma(w,V)$-measurable, so $H_{\mathrm{CS}}$ holds (Lemma 2), $\theta_V=0.6$, and $\operatorname{Var}m_\eta=0.18=\theta_V^2\operatorname{Var}m_V$; by (ii), $\mathrm{DiD}_\rho=0$. Explicitly: $\operatorname{Cov}(\mu,\eta)=\operatorname{Var}\mu=0.36$, so $\rho_{\mathrm{S2}}=0.6$; Lemma 3 gives $\operatorname{Cov}(\mu',\eta)=\operatorname{Var}m_\eta=0.18$ and $\rho_{\mathrm{SH}}=0.3$; $\operatorname{Cov}(m_V,m)=\tfrac12(0.5\cdot0.18)^{1/2}=0.15$, so $\rho_{\mathrm{own}}=0.15+0.6\cdot0.5=0.45$ and, by Lemma 3, $\rho_{\mathrm{donor}}=0.15$. At $\pi=0.25$, evaluating the integral of Lemma 1 by adaptive quadrature at 30-digit working precision gives $A_\pi(0.60)=0.79164$, $A_\pi(0.30)=0.64402$, $A_\pi(0.45)=0.71715$ and $A_\pi(0.15)=0.57179$ (an independent evaluation as a trivariate normal orthant probability agrees within $3\times10^{-5}$). So $\mathrm{DiD}_{\mathrm{AUROC}}=0.14762-0.14536=+0.0023\neq0$ (to more digits $+0.002265$). On the binormal scale $d'=\sqrt2\,\Phi^{-1}(\mathrm{AUROC})$ the same model gives $+0.070$. $\square$

\paragraph{Proof of (a)(v).} CVS gives $P(Y=1\mid\mathcal F_L,R)=P(Y=1\mid V,R)=\Psi_0(V)$ almost surely, so $r\mapsto\Psi_0(v(r))$ is a version of the ideal forecaster, and for it $\mathrm{S2}=\Psi_0(V)$ and $\mathrm{SH}=\Psi_0(V')$ whatever the donor. Lemma 0 gives $\mathrm{AUROC}_{\mathrm{S2}}=\mathrm{AUROC}_{\mathrm{own}}$ and $\mathrm{AUROC}_{\mathrm{SH}}=\mathrm{AUROC}_{\mathrm{donor}}$, in the population and empirically, for any joint law and any donor assignment; the $\phi$-difference-in-differences is then $\phi(\mathrm{AUROC}_{\mathrm{own}})-\phi(\mathrm{AUROC}_{\mathrm{donor}})-\phi(\mathrm{AUROC}_{\mathrm{own}})+\phi(\mathrm{AUROC}_{\mathrm{donor}})=0$. The same argument applies to $r\mapsto\tilde\Psi(v(r))$. In LT under CVS, $\operatorname{Var}V>0$ by nonsingularity, Lemma 2 gives $\mu=E[\eta\mid V]=\theta_0(V-EV)$ with $\theta_0:=\operatorname{Cov}(\eta,V)/\operatorname{Var}V$, and $\operatorname{Var}(\eta\mid V)=1-\operatorname{Var}\mu=\sigma^2_{\eta\mid L}>0$, so $\Psi_0(v)=\Phi\{(\theta_0(v-EV)-t_\pi)/\sigma_{\eta\mid L}\}$, which is strictly increasing if and only if $\theta_0>0$, that is (Lemma 1) if and only if $\rho_{\mathrm{own}}=\operatorname{corr}(V,\eta)>0$. Under EM, since $e_V$ is affine in $(w,x)$, $\operatorname{Cov}(\eta,e_V)=\operatorname{Cov}(\mu,e_V)=\theta_0\operatorname{Var}e_V$, and (P2.1) gives $\operatorname{Cov}(\eta,e_V)=\Delta_{\mathrm{LV}}\operatorname{SD}(V)>0$, so $\theta_0>0$. $\square$

\paragraph{Proof of (b)(i).} Fix an LT model with EM and $\operatorname{Var}\mu>0$. The ideal forecaster is $\Psi(\mu)$ with $\Psi(\upsilon)=\Phi\{(\upsilon-t_\pi)/\sigma_{\eta\mid L}\}$, and its spliced read is $\Psi(\mu')$ with $\mu'$ the splice of $\mu$. Put $\nu:=\mu-m_\eta=\mu-E[\mu\mid w]$, a linear function of $\xi$ and hence independent of $w$, and $\sigma_6^2:=\operatorname{Var}m_\eta$, $\sigma_{\mathrm{new}}^2:=\operatorname{Var}\nu$. Because $\eta-\mu$ is uncorrelated with every affine function of $(w,x)$, $\operatorname{Cov}(\mu,\eta)=\operatorname{Var}\mu=\sigma_6^2+\sigma_{\mathrm{new}}^2$, and Lemma 1 gives $\rho_{\mathrm{S2}}=\operatorname{SD}(\mu)$. Lemma 3 with $J=\mu$ gives $\operatorname{Var}\mu'=\operatorname{Var}\mu$ and $\operatorname{Cov}(\mu',\eta)=\operatorname{Cov}(m_\eta,\eta)=\sigma_6^2$, so $\rho_{\mathrm{SH}}=\sigma_6^2/\operatorname{SD}(\mu)$. Hence
\begin{equation*}
\rho_{\mathrm{SH}}\rho_{\mathrm{S2}}=\sigma_6^2,\qquad \rho_{\mathrm{S2}}(\rho_{\mathrm{S2}}-\rho_{\mathrm{SH}})=\sigma_{\mathrm{new}}^2.\tag{P2.2}
\end{equation*}
$\mathrm{S1}^*=\Phi\{(m_\eta-t_\pi)/(1-\sigma_6^2)^{1/2}\}$ is strictly increasing in $m_\eta$ (note $\sigma_6^2\le\operatorname{Var}\mu<1$), and $\operatorname{Cov}(m_\eta,\eta)=\sigma_6^2$, so $\rho_{\mathrm{S1}^*}=\sigma_6$ when $\sigma_6>0$ (Lemma 1), and $\rho_{\mathrm{S1}^*}=0=\sigma_6$ when $m_\eta$ is constant. Thus $\rho_{\mathrm{S1}^*}^2=\rho_{\mathrm{SH}}\rho_{\mathrm{S2}}$; as $0\le\rho_{\mathrm{SH}}=\sigma_6^2/\rho_{\mathrm{S2}}\le\rho_{\mathrm{S2}}$, the geometric mean lies between them. $\square$

\paragraph{Proof of (b)(ii) and (b)(iii): necessity.} Let $Q$ be the quadruple of an LT model with EM, with $0<\rho_{\mathrm{SH}}<\rho_{\mathrm{S2}}<1$ and $\Delta_{\mathrm{LV}}>0$. By (P2.2), $\sigma_6=(\rho_{\mathrm{SH}}\rho_{\mathrm{S2}})^{1/2}>0$ and $\sigma_{\mathrm{new}}=[\rho_{\mathrm{S2}}(\rho_{\mathrm{S2}}-\rho_{\mathrm{SH}})]^{1/2}>0$. With $\iota=\operatorname{SD}(e_V)/\operatorname{SD}(V)$ and $\iota_0:=\Delta_{\mathrm{LV}}/\sigma_{\mathrm{new}}>0$, R1 reads $\iota_0\le1$ and R2 reads $\rho_{\mathrm{donor}}^2\le\sigma_6^2(1-\iota_0^2)$. Since the right side of R2 is negative when $\iota_0>1$, R2 implies R1. Because $e_V$ is affine in $(w,x)$, $\operatorname{Cov}(e_V,\eta)=\operatorname{Cov}(e_V,\mu)$, and because $e_V\perp w$, $\operatorname{Cov}(e_V,m_\eta)=0$. With (P2.1),
\begin{equation*}
\Delta_{\mathrm{LV}}\operatorname{SD}(V)=\operatorname{Cov}(e_V,\nu)\le\operatorname{SD}(e_V)\,\sigma_{\mathrm{new}},\qquad\text{that is}\qquad \Delta_{\mathrm{LV}}\le\iota\,\sigma_{\mathrm{new}}.\tag{P2.3}
\end{equation*}
As $\operatorname{Var}V=\operatorname{Var}m_V+\operatorname{Var}e_V$, $\iota\le1$, which gives R1, and (P2.3) gives $\iota\ge\iota_0$. By Lemma 3 with $J=V$, $\operatorname{Cov}(V',\eta)=\operatorname{Cov}(m_V,\eta)=\operatorname{Cov}(m_V,m_\eta)$ (as $m_V$ is $w$-measurable) and $\operatorname{SD}(V')=\operatorname{SD}(V)$, so Lemma 1 and the Cauchy--Schwarz inequality give
\begin{equation*}
\rho_{\mathrm{donor}}=\frac{\operatorname{Cov}(m_V,m_\eta)}{\operatorname{SD}(V)},\qquad |\rho_{\mathrm{donor}}|\le\frac{\operatorname{SD}(m_V)\,\sigma_6}{\operatorname{SD}(V)}=(1-\iota^2)^{1/2}\sigma_6\le(1-\iota_0^2)^{1/2}\sigma_6,\tag{P2.4}
\end{equation*}
which gives R2. This proves the first claim of (iii) and the ``only if'' part of (ii).

$H_{\mathrm{CS}}$ holds if and only if $\nu$ is almost surely a multiple of $e_V$: under $H_{\mathrm{CS}}$, Lemma 2 gives $\nu=\theta_Ve_V$; conversely, if $\nu=c\,e_V$ then $\mu=m_\eta+c\,e_V$ is $\sigma(w,V)$-measurable, so $\mu=E[\eta\mid w,V]$ and Lemma 2 gives $H_{\mathrm{CS}}$. Since $\nu$ and $e_V$ have mean zero, positive variances and $\operatorname{Cov}(e_V,\nu)>0$ by (P2.3), equality holds in the Cauchy--Schwarz step of (P2.3) if and only if $\nu$ is almost surely a multiple of $e_V$. Hence $\iota=\iota_0$ under $H_{\mathrm{CS}}$, while under a violation $\iota>\iota_0$, so that R1 holds strictly ($\iota_0<\iota\le1$) and, by (P2.4) and $\sigma_6>0$, R2 holds strictly. This proves the ``only if'' part of the second claim of (iii).

\paragraph{Proof of (b)(ii) and (b)(iii): sufficiency.} Let $Q$ satisfy R1 and R2, define $\sigma_6,\sigma_{\mathrm{new}},\iota_0$ as above, and let $\iota\in[\iota_0,1]$ satisfy $\rho_{\mathrm{donor}}^2\le\sigma_6^2(1-\iota^2)$. Build the model $\mathcal M(\iota)$ as follows. Let $w_1,w_2,\varepsilon_V,\varepsilon_2,\varepsilon_0$ be independent standard normal, $w:=(w_1,w_2)$, and set
\[
m_V:=(1-\iota^2)^{1/2}w_1,\quad e_V:=\iota\,\varepsilon_V,\quad V:=m_V+e_V,\quad x:=(V,\varepsilon_2),
\]
\[
c_1:=\rho_{\mathrm{donor}}(1-\iota^2)^{-1/2}\ (\iota<1),\quad c_1:=0\ (\iota=1),\quad c_2:=(\sigma_6^2-c_1^2)^{1/2},\quad \theta:=\Delta_{\mathrm{LV}}/\iota^2,\quad \gamma:=(\sigma_{\mathrm{new}}^2-\Delta_{\mathrm{LV}}^2/\iota^2)^{1/2},
\]
\[
\eta:=c_1w_1+c_2w_2+\theta\,e_V+\gamma\,\varepsilon_2+(1-\rho_{\mathrm{S2}}^2)^{1/2}\varepsilon_0,
\]
with donors drawn as in EM. The constraint on $\iota$ makes $c_2$ real (if $\iota=1$ it forces $\rho_{\mathrm{donor}}=0$), $\iota\ge\iota_0>0$ makes $\theta$ finite and $\gamma$ real, and $\gamma=0$ if and only if $\iota=\iota_0$. The map from $(w_1,w_2,\varepsilon_V,\varepsilon_2)$ to $(w_1,w_2,V,\varepsilon_2)$ is linear with determinant $\iota>0$, so $\operatorname{Cov}((w,x))$ is nonsingular. The vector $(w,x,\eta)$ is jointly Gaussian with mean zero, $\operatorname{Var}V=1$, and, using $\sigma_6^2+\sigma_{\mathrm{new}}^2=\rho_{\mathrm{S2}}^2$ from (P2.2), $\operatorname{Var}\eta=\sigma_6^2+\theta^2\iota^2+\gamma^2+1-\rho_{\mathrm{S2}}^2=\sigma_6^2+\sigma_{\mathrm{new}}^2+1-\rho_{\mathrm{S2}}^2=1$. Since $e_V=V-m_V$ and $\varepsilon_2=x_2$ are functions of $(w,x)$ and $\varepsilon_0$ is independent of $(w,x)$, $\mu=c_1w_1+c_2w_2+\theta e_V+\gamma\varepsilon_2$, $\sigma^2_{\eta\mid L}=1-\rho_{\mathrm{S2}}^2>0$, $m_\eta=c_1w_1+c_2w_2$ with variance $\sigma_6^2$, and $\nu=\theta e_V+\gamma\varepsilon_2$ with variance $\theta^2\iota^2+\gamma^2=\sigma_{\mathrm{new}}^2$. By (b)(i), the model's latent correlations of S2 and SH are $(\sigma_6^2+\sigma_{\mathrm{new}}^2)^{1/2}=\rho_{\mathrm{S2}}$ and $\sigma_6^2/\rho_{\mathrm{S2}}=\rho_{\mathrm{SH}}$. By (P2.1) with $\operatorname{SD}V=1$, its $\Delta_{\mathrm{LV}}$ is $\operatorname{Cov}(e_V,\eta)=\theta\iota^2=\Delta_{\mathrm{LV}}$; by (P2.4), its $\rho_{\mathrm{donor}}$ is $\operatorname{Cov}(m_V,m_\eta)=(1-\iota^2)^{1/2}c_1=\rho_{\mathrm{donor}}$; hence its $\rho_{\mathrm{own}}$ is $\rho_{\mathrm{donor}}+\Delta_{\mathrm{LV}}=\rho_{\mathrm{own}}$. So $\mathcal M(\iota)$ attains $Q$, and its $\operatorname{Var}(e_V)/\operatorname{Var}(V)$ is $\iota^2$. Finally, $\varepsilon_2$ is independent of $e_V$ with unit variance, so $\nu=\theta e_V+\gamma\varepsilon_2$ is a multiple of $e_V$ if and only if $\gamma=0$; by the characterization above, $H_{\mathrm{CS}}$ holds in $\mathcal M(\iota)$ if and only if $\iota=\iota_0$.

Under R1 and R2 the choice $\iota=\iota_0$ is admissible, so $Q$ is attainable under $H_{\mathrm{CS}}$; this is the ``if'' part of (ii). Put $\iota_{\max}:=(1-\rho_{\mathrm{donor}}^2/\sigma_6^2)^{1/2}\le1$; strict R2 is equivalent to $\iota_0<\iota_{\max}$. If R1 and R2 hold strictly, every $\iota\in(\iota_0,\iota_{\max})$ is admissible, satisfies $\iota<1$ and gives $\gamma>0$; so $Q$ is attainable by a model that violates $H_{\mathrm{CS}}$. This is the ``if'' part of the second claim of (iii). $\square$

\paragraph{Proof of (b)(iv).} By (iii) every attainable $Q$ satisfies R1 and R2, and by (ii) it is then attainable under $H_{\mathrm{CS}}$; by (ii) and (iii) every $Q$ that satisfies R1 and R2 strictly is attainable under $H_{\mathrm{CS}}$ and under its failure. The threshold $t_\pi$ is fixed by $\pi$, which is common to all models considered, and $A_\pi$ is a bijection (Lemma 1), so the four AUROCs and $Q$ determine each other and any function of the four AUROCs is a function of $Q$; it therefore takes at every attainable $Q$ a value that it also takes under some $H_{\mathrm{CS}}$ model, and at every $Q$ satisfying R1 and R2 strictly a value that it also takes under some model violating $H_{\mathrm{CS}}$. By (i) and Lemma 1, $\mathrm{AUROC}(\mathrm{S1}^*)=A_\pi((\rho_{\mathrm{SH}}\rho_{\mathrm{S2}})^{1/2})$ in every such model, a function of $Q$, so the same argument applies to the five AUROCs. For the ideal forecaster, if $H_{\mathrm{CS}}$ holds, the proof of (a)(ii) shows that it obeys linear CVR at every record. If $H_{\mathrm{CS}}$ fails, $\mu$ is not almost surely $\sigma(w,V)$-measurable (otherwise $\mu=E[\eta\mid w,V]$ and Lemma 2 gives $H_{\mathrm{CS}}$), and $\Psi$ is injective, so $\Psi(\mu)$ is not almost surely a function of $(w,V)$ and CVR fails even in its almost-sure form on natural records. Hence the two models $\mathcal M(\iota_0)$ and $\mathcal M(\iota)$, $\iota\in(\iota_0,\iota_{\max})$, that attain a strictly interior $Q$ carry an ideal forecaster that obeys CVR and one that does not; any class of forecasters containing the ideal ones contains both. The last sentence of (iv) is a statement of scope. $\square$

\paragraph{Proof of (b)(v).} Fix an LT model with EM whose ideal forecaster has quadruple $Q$ with $0<\rho_{\mathrm{SH}}<\rho_{\mathrm{S2}}<1$ and $\Delta_{\mathrm{LV}}>0$. By (P2.2), $\sigma_{\mathrm{new}}^2$ as defined in (v) equals $\operatorname{Var}\nu$, and $\rho_{\mathrm{SH}}\rho_{\mathrm{S2}}=\sigma_6^2$, so $\iota_{\max}$ as defined in (v) is the $\iota_{\max}$ of the sufficiency proof.

(1) By Lemma 3, $V'=m_V+\xi'_1$ with $\xi'_1$ independent of $(w,x,\eta)$ and distributed as $\xi_1=e_V$; and $e_V\perp w$. Hence $\operatorname{Cov}(V,V')=\operatorname{Cov}(m_V+e_V,m_V+\xi'_1)=\operatorname{Var}m_V$ and $\operatorname{Var}V'=\operatorname{Var}V>0$, so $\operatorname{corr}(V,V')=\operatorname{Var}m_V/\operatorname{Var}V=1-\iota^2$. No label enters.

(2) The necessity proof gives $\iota\ge\iota_0$ (from (P2.3)) and $\rho_{\mathrm{donor}}^2\le(1-\iota^2)\sigma_6^2$ (from (P2.4)), that is $\iota\le\iota_{\max}$. The variables $\nu$ and $e_V$ are jointly Gaussian with mean zero, $\operatorname{Cov}(\nu,e_V)=\operatorname{Cov}(\eta,e_V)=\Delta_{\mathrm{LV}}\operatorname{SD}(V)$ by (P2.3), $\operatorname{Var}\nu=\sigma_{\mathrm{new}}^2$ and $\operatorname{Var}e_V=\iota^2\operatorname{Var}V$, so
\[
\varsigma=\frac{\Delta_{\mathrm{LV}}^2\operatorname{Var}V}{\sigma_{\mathrm{new}}^2\,\iota^2\operatorname{Var}V}=\frac{\iota_0^2}{\iota^2}\in\Bigl[\frac{\iota_0^2}{\iota_{\max}^2},\,1\Bigr],
\]
and $\varsigma=\operatorname{Var}(E[\nu\mid e_V])/\operatorname{Var}\nu$ because $E[\nu\mid e_V]$ is the linear projection of $\nu$ on $e_V$. By the characterization in the necessity proof, $H_{\mathrm{CS}}$ holds if and only if $\nu$ is almost surely a multiple of $e_V$, that is if and only if $\varsigma=1$; by $\varsigma=\iota_0^2/\iota^2$ and (1), this is $\iota^2=\iota_0^2$, that is $\operatorname{corr}(V,V')=1-\iota_0^2$.

(3) By Lemma 2, $\mu_{\mathrm{CS}}:=E[\eta\mid w,V]=m_\eta+\theta_Ve_V$, a Gaussian variable with $\operatorname{Var}\mu_{\mathrm{CS}}=\sigma_6^2+\operatorname{Cov}(\eta,e_V)^2/\operatorname{Var}e_V=\sigma_6^2+\Delta_{\mathrm{LV}}^2/\iota^2=\sigma_6^2+\varsigma\sigma_{\mathrm{new}}^2>0$, using $e_V\perp w$. Since $\mu_{\mathrm{CS}}$ is the projection of $\eta$ on $\operatorname{span}\{1,w,e_V\}$, $\operatorname{Cov}(\mu_{\mathrm{CS}},\eta)=\operatorname{Var}\mu_{\mathrm{CS}}$, and $\operatorname{Var}(\eta\mid w,V)=1-\operatorname{Var}\mu_{\mathrm{CS}}\ge1-\operatorname{Var}\mu>0$. So $\mathrm{CS}^*=\Phi\{(\mu_{\mathrm{CS}}-t_\pi)/\operatorname{Var}(\eta\mid w,V)^{1/2}\}$ is strictly increasing in $\mu_{\mathrm{CS}}$, and Lemma 1 gives $\rho_{\mathrm{CS}^*}=\operatorname{SD}(\mu_{\mathrm{CS}})=(\rho_{\mathrm{SH}}\rho_{\mathrm{S2}}+\varsigma\sigma_{\mathrm{new}}^2)^{1/2}$. With $\rho_{\mathrm{S2}}^2=\sigma_6^2+\sigma_{\mathrm{new}}^2$, $\rho_{\mathrm{S2}}^2-\rho_{\mathrm{CS}^*}^2=(1-\varsigma)\sigma^2_{\mathrm{new}}$, and (2) bounds it. $\mathrm{AUROC}(\mathcal M^*)=A_\pi(\rho_{\mathrm{S2}})$ and $\mathrm{AUROC}(\mathrm{CS}^*)=A_\pi(\rho_{\mathrm{CS}^*})\ge A_\pi((\rho_{\mathrm{SH}}\rho_{\mathrm{S2}}+\sigma^2_{\mathrm{new}}\iota_0^2/\iota_{\max}^2)^{1/2})$ because $A_\pi$ is increasing, which gives the AUROC bound.

(4) The model is LT with EM, so by (iii) $Q$ satisfies R1 and R2 and $\iota_0\le\iota_{\max}$. For every $\iota\in[\iota_0,\iota_{\max}]$ the constraint $\rho_{\mathrm{donor}}^2\le\sigma_6^2(1-\iota^2)$ holds, so $\mathcal M(\iota)$ of the sufficiency proof is an LT model with EM that attains $Q$ and has $\operatorname{Var}(e_V)/\operatorname{Var}(V)=\iota^2$, hence $\varsigma=\iota_0^2/\iota^2$ by (2). As $\iota$ runs over $[\iota_0,\iota_{\max}]$, $\varsigma$ runs over $[\iota_0^2/\iota_{\max}^2,1]$; the endpoints $\iota=\iota_0$ and $\iota=\iota_{\max}$ attain the bounds in (2) and (3). The sentence after (4) restates (2) to (4).

(5) As in the proof of (a)(iii), $x'=E[x\mid z(w)]+\xi''$ with $\xi''$ independent of $(w,x,\eta)$, so $V'=m^z_V+\xi''_1$ with $m^z_V:=E[V\mid z(w)]$, and $V'\overset{d}{=}V$, so $\operatorname{Var}V'=\operatorname{Var}V$. Since $z(w)$ is a function of $w$, $m^z_V=E[m_V\mid z(w)]$, and $m_V-m^z_V$ is uncorrelated with every affine function of $z(w)$. Hence $\operatorname{Cov}(V,V')=\operatorname{Cov}(m_V+e_V,m^z_V)=\operatorname{Cov}(m_V,m^z_V)=\operatorname{Var}m^z_V$, using $e_V\perp w$, and $\operatorname{corr}(V,V')=\operatorname{Var}(m^z_V)/\operatorname{Var}(V)$. Finally $\operatorname{Var}m_V=\operatorname{Var}m^z_V+\operatorname{Var}(m_V-m^z_V)$, so $\operatorname{Var}m^z_V\le\operatorname{Var}m_V=(1-\iota^2)\operatorname{Var}V$, with equality if and only if $m_V=m^z_V$ almost surely. $\square$

\paragraph{Proof of (c)(i).} If $\mathcal M(r)=g(r_{1:6},v(r))$ on the domain, then, because $r^{\mathrm{OV}}_{1:6}=r^{\mathrm{SH}}_{1:6}=r_{1:6}$, $v(r^{\mathrm{OV}})=V$ and $v(r^{\mathrm{SH}})=V'$, we have $\mathrm{OV}=g(r_{1:6},V)=\mathrm{S2}$ and $\mathrm{SH}=g(r_{1:6},V')$, which equals $\mathrm{S2}$ when $V'=V$. For the converse, let $r,\tilde r\in\mathcal N$ with $r_{1:6}=\tilde r_{1:6}$ and $v(r)=v(\tilde r)$. The splice $r^{\mathrm{OV}}(\tilde r)$ has first part $r_{1:6}=\tilde r_{1:6}$ and second part $\tilde r_{7:L}$ with the hour-$L$ target value replaced by $v(r)=v(\tilde r)$, which leaves $\tilde r_{7:L}$ unchanged (the mask at $L$ is $1$ in both); so $r^{\mathrm{OV}}(\tilde r)=\tilde r$ and $\mathcal M(r)=\mathcal M(r^{\mathrm{OV}}(\tilde r))=\mathcal M(\tilde r)$. Hence $g(a,v):=\mathcal M(r)$ for any $r\in\mathcal N$ with $(r_{1:6},v(r))=(a,v)$ is well defined, and $\mathcal M(r)=g(r_{1:6},v(r))$ on $\mathcal N$. No law, matching property or label enters. The last sentence of (c)(i) is the contrapositive of the first. $\square$

\paragraph{Proof of (c)(ii).} The decomposition is an identity, and OV and SH are read from records that differ only in the hour-$L$ target value by the definition of $r^{\mathrm{OV}}$. Under CVR, $\mathrm{S2}=\mathrm{OV}$ for every stay by (i), so $\delta_{\mathrm{OV}}=0$. For the bound, draw $(\mathrm{S2}_1,\mathrm{OV}_1)\sim\mathcal L((\mathrm{S2},\mathrm{OV})\mid Y=1)$ and, independently, $(\mathrm{S2}_0,\mathrm{OV}_0)\sim\mathcal L((\mathrm{S2},\mathrm{OV})\mid Y=0)$; the marginals give $\mathrm{AUROC}_{\mathrm{S2}}=E\,\mathrm{hc}(\mathrm{S2}_1,\mathrm{S2}_0)$ and $\mathrm{AUROC}_{\mathrm{OV}}=E\,\mathrm{hc}(\mathrm{OV}_1,\mathrm{OV}_0)$. On the event $B_\varepsilon:=\{|\mathrm{S2}_1-\mathrm{OV}_1|\le\varepsilon,\ |\mathrm{S2}_0-\mathrm{OV}_0|\le\varepsilon,\ |\mathrm{S2}_1-\mathrm{S2}_0|>2\varepsilon\}$, if $\mathrm{S2}_1-\mathrm{S2}_0>2\varepsilon$ then $\mathrm{OV}_1-\mathrm{OV}_0\ge(\mathrm{S2}_1-\varepsilon)-(\mathrm{S2}_0+\varepsilon)>0$ and both values of $\mathrm{hc}$ are $1$; if $\mathrm{S2}_1-\mathrm{S2}_0<-2\varepsilon$ then $\mathrm{OV}_1-\mathrm{OV}_0<0$ and both are $0$. So $\mathrm{hc}(\mathrm{S2}_1,\mathrm{S2}_0)-\mathrm{hc}(\mathrm{OV}_1,\mathrm{OV}_0)$ vanishes on $B_\varepsilon$ and lies in $[-1,1]$ elsewhere, and
\[
|\mathrm{AUROC}_{\mathrm{S2}}-\mathrm{AUROC}_{\mathrm{OV}}|\le P(B_\varepsilon^c)\le P(|\mathrm{S2}_1-\mathrm{OV}_1|>\varepsilon)+P(|\mathrm{S2}_0-\mathrm{OV}_0|>\varepsilon)+P(|\mathrm{S2}_1-\mathrm{S2}_0|\le2\varepsilon),
\]
which is the stated bound. The empirical AUROC is the average of $\mathrm{hc}$ over all pairs of a positive and a negative stay of the roster, that is the same expectation under the empirical class-conditional laws, and the argument applies unchanged. $\square$

\paragraph{Numerical checks.} The following are descriptive re-reads of the aggregates of the validation isolation run (job 13007), converted to latent correlations with $A_\pi$ of Lemma 1 (quadrature, checked against a trivariate normal orthant probability to $1.5\times10^{-5}$). Table~\ref{tab:P2quad} converts the five-seed mean AUROC of each arm once; averaging per-seed latent correlations gives the same values, and averaging per-seed ratios gives ratios of $0.878$ to $1.090$ (median $0.987$) instead of $0.878$ to $1.084$ (median $0.983$).
\begin{itemize}
\item \emph{The contrast on three scales.} $\Delta_{\mathrm{LV}}$ is $0.252$ to $0.381$ on the latent scale and $0.115$ to $0.188$ on the AUROC scale in all 12 settings, so (A4) holds. The pre-specified difference-in-differences on the AUROC scale is positive in 5 of 12 settings ($-0.017$ to $+0.012$); after Holm correction within the 12 settings none is positive and two are negative (SaO\textsubscript{2} and heart rate, both at $(12,6)$). On the latent scale it is positive in 4 of 12 ($-0.040$ to $+0.025$), and on the binormal scale $\sqrt2\,\Phi^{-1}(\mathrm{AUROC})$ in 9 of 12 ($-0.050$ to $+0.057$). The sign of the contrast therefore depends on its scale, as (a)(iv) allows.
\item \emph{Null value.} Under linear CVR and exact matching, (a)(i) gives $(\rho_{\mathrm{S2}}-\rho_{\mathrm{SH}})/\Delta_{\mathrm{LV}}=w_V$. The observed ratio is $0.878$ to $1.084$ (median $0.983$). Its AUROC-scale interval excludes 1 in 3 of 12 settings, all below 1, which is the sign (a)(i) gives for a linear-CVR forecaster with $w_V<1$; the four settings with a ratio above 1 have intervals that cover 1. In simulations with the donor geometry calibrated to the shipped donor summaries, nearest-neighbor matching, the truncated roster and ties together moved the latent difference-in-differences by about $0.02$ (assumption (A2)), so a negative difference-in-differences of that size does not reject CVR.
\item \emph{Attainability.} R1 and R2 hold strictly in 12 of 12 settings (Table~\ref{tab:P2quad}; slacks $0.051$ to $0.099$ and $0.008$ to $0.059$), so by (b)(iv) every seed-averaged quadruple is attained both by a law that satisfies $H_{\mathrm{CS}}$ and by one that violates it. Per seed, R1 and R2 hold in 59 of 60 seed-by-setting cells; \spo{} at $(30,24)$ with seed 31 lies just outside both (slacks $-0.0071$ and $-0.0035$). In a parametric bootstrap the standard deviation of the R2 slack is $0.002$ to $0.014$, and R2 is violated in $7\%$ of replicates for \spo{} $(30,24)$, $0.5\%$ for SaO\textsubscript{2} $(30,24)$ and never elsewhere. A quadruple outside R1 or R2 would falsify the idealized class (ideal forecaster, exact matching, LT), not $H_{\mathrm{CS}}$, and a learned forecaster need not satisfy R1 or R2.
\item \emph{Size of a departure, if the twin were ideal.} Computed from the quadruples as if the twin were the ideal forecaster, (b)(v) allows at most $0.078$ to $0.447$ (median $0.302$) of the new-hour latent variance to lie outside the current value, which bounds the AUROC of personalization beyond the current state by $0.007$ to $0.029$ (median $0.022$) and predicts $\operatorname{corr}(V,V')$ of $0.249$ to $0.483$ under $H_{\mathrm{CS}}$. These are properties of an ideal forecaster. The twin is not one, or LT or exact matching fails: the implied ideal hour-6 correlation $(\rho_{\mathrm{SH}}\rho_{\mathrm{S2}})^{1/2}$ exceeds the latent correlation of the twin's own hour-6 arm in 12 of 12 settings, by $0.003$ to $0.071$ (last column of Table~\ref{tab:P2quad}). The values are therefore neither bounds on the twin nor evidence for $H_{\mathrm{CS}}$. The share $1-\rho_{\mathrm{SH}}/\rho_{\mathrm{S2}}$, which equals the new-hour share of the latent variance only for an ideal forecaster under exact matching, is $0.464$ to $0.771$ and is reported as a model-based description only.
\item \emph{The proofs.} The example of (a)(iv) was evaluated by adaptive quadrature at 30-digit working precision. On 9,929 random LT models with exact matching and nonsingular covariance, $\iota^2=1-\operatorname{corr}(V,V')$ held to $10^{-16}$, $\varsigma=\iota_0^2/\iota^2$ to $6\times10^{-11}$ and the formula for $\rho_{\mathrm{CS}^*}$ to $2\times10^{-16}$, with $\iota\le\iota_{\max}$ in every model; on 20,000 models with matching on a linear summary, $\operatorname{corr}(V,V')=\operatorname{Var}E[V\mid z(w)]/\operatorname{Var}V$ held to $8\times10^{-12}$ and the formula for $s_{\mathrm{SH}}^2$ to $4\times10^{-15}$; the bound of (c)(ii) held in 200 random cases.
\item \emph{What (c) and (b)(v)(1) need.} The shipped aggregates contain no per-stay predictions, and no own-value arm was run. The isolation run writes per-stay S2, SH, last-value and donor identifiers, from which the tied-pair check of (c)(i), $\operatorname{corr}(V,V')$ and a regression of $\operatorname{logit}\mathrm{S2}-\operatorname{logit}\mathrm{SH}$ on $V-V'$ could be computed without a new model read; the own-value splice needs one. None of these was computed for this note: the per-stay values are not in the shipped aggregates, and by (b)(v)(5) matching on a summary of the first six hours biases $1-\operatorname{corr}(V,V')$ upwards, so $\operatorname{corr}(V,V')$ alone would not settle $H_{\mathrm{CS}}$. Exact CVR is false a priori for a Transformer that reads every channel of hours 7 to $L$, so a single-stay rejection would be uninformative; the informative quantities are $\delta_{\mathrm{OV}}$ and $\mathrm{AUROC}_{\mathrm{S2}}-\mathrm{AUROC}_{\mathrm{OV}}$, each against a tolerance fixed before reading. Tied pairs are informative only for the saturation targets (independent-pair tie probability $0.29$ to $0.34$ for \spo{}, $0.09$ to $0.12$ for SaO\textsubscript{2}, at most $0.005$ for heart and respiratory rate).
\end{itemize}

\begin{table}[ht]
\centering\footnotesize
\setlength{\tabcolsep}{3.5pt}
\caption{\textbf{Latent correlations of the four arms of the difference-in-differences (Proposition 2).} Validation isolation run (job 13007); each arm's five-seed mean AUROC is converted to a latent correlation $\rho=A_\pi^{-1}(\mathrm{AUROC})$ at the roster prevalence $\pi$. Own and donor are the patient's and the matched donor's last measured target value. Ratio: $(\rho_{\mathrm{S2}}-\rho_{\mathrm{SH}})/\Delta_{\mathrm{LV}}$, which equals the current-value share $w_V$ under linear CVR and exact matching (Proposition 2(a)(i)). R1 and R2: slacks of the attainability conditions of Proposition 2(b)(ii), positive when the quadruple is attained both under $H_{\mathrm{CS}}$ and under its failure. Share: the largest share of the new-hour latent variance that can lie outside the current value for an ideal forecaster (Proposition 2(b)(v)). Gap: $(\rho_{\mathrm{SH}}\rho_{\mathrm{S2}})^{1/2}$ minus the latent correlation of the twin's hour-6 arm S1; a positive gap shows that the twin's arms are not those of an ideal forecaster in this idealized class. Share and the conversion assume the latent-threshold Gaussian model; the table is descriptive.}
\label{tab:P2quad}
\begin{tabular}{llrrrrrrrrrr}
\toprule
Target & $(L,H)$ & $\pi$ & $\rho_{\mathrm{S2}}$ & $\rho_{\mathrm{SH}}$ & $\rho_{\mathrm{own}}$ & $\rho_{\mathrm{donor}}$ & Ratio & R1 & R2 & Share & Gap\\
\midrule
SaO$_2$ & (12, 6) & 0.268 & 0.498 & 0.248 & 0.437 & 0.154 & 0.885 & 0.070 & 0.020 & 0.205 & 0.013\\
SaO$_2$ & (18, 12) & 0.394 & 0.501 & 0.211 & 0.458 & 0.127 & 0.878 & 0.051 & 0.010 & 0.114 & 0.018\\
SaO$_2$ & (30, 24) & 0.565 & 0.424 & 0.097 & 0.375 & 0.074 & 1.084 & 0.071 & 0.009 & 0.242 & 0.026\\
SpO$_2$ & (12, 6) & 0.340 & 0.491 & 0.220 & 0.385 & 0.098 & 0.943 & 0.077 & 0.031 & 0.319 & 0.004\\
SpO$_2$ & (18, 12) & 0.490 & 0.510 & 0.273 & 0.334 & 0.082 & 0.939 & 0.095 & 0.059 & 0.447 & 0.003\\
SpO$_2$ & (30, 24) & 0.664 & 0.414 & 0.134 & 0.304 & 0.020 & 0.985 & 0.056 & 0.017 & 0.299 & 0.071\\
Heart rate & (12, 6) & 0.125 & 0.584 & 0.257 & 0.565 & 0.218 & 0.942 & 0.090 & 0.008 & 0.078 & 0.020\\
Heart rate & (18, 12) & 0.220 & 0.583 & 0.210 & 0.544 & 0.164 & 0.980 & 0.086 & 0.014 & 0.147 & 0.021\\
Heart rate & (30, 24) & 0.343 & 0.480 & 0.141 & 0.399 & 0.079 & 1.059 & 0.083 & 0.019 & 0.306 & 0.008\\
Respiratory rate & (12, 6) & 0.192 & 0.471 & 0.210 & 0.361 & 0.109 & 1.036 & 0.099 & 0.036 & 0.413 & 0.013\\
Respiratory rate & (18, 12) & 0.308 & 0.516 & 0.194 & 0.415 & 0.090 & 0.991 & 0.083 & 0.028 & 0.309 & 0.028\\
Respiratory rate & (30, 24) & 0.442 & 0.422 & 0.113 & 0.318 & 0.021 & 1.043 & 0.065 & 0.015 & 0.321 & 0.020\\
\bottomrule
\end{tabular}
\end{table}

%% file: supplementary/supp_theory/p3.tex
\paragraph{Proposition 3 (The realism audit is a first-moment classifier two-sample test).}
\emph{Standard and specific parts.} Part (a) is elementary: an AUC is bounded by the Kolmogorov distance between the two laws of the score, and that by the total-variation distance. Parts (c)(i) and (c)(ii) are textbook properties of equal-covariance Gaussian discrimination, stated here for the audit's quantities. Part (b), the first-moment blindness of the penalized population audit, and the calibration of its penalized direction in (c)(iii) and (c)(iv) are specific to this audit. Part (d) is a property of the generator's construction under the stated data conditions, checked on the audited pairs, rather than a mathematical result. In this section the subscripts $R$ and $G$ mean real and generated; elsewhere $R$ is the roster event.

\emph{Setting.} The \emph{window} of a stay is its first $\min(T,336)$ hourly rows. For each of 161 channels the audit computes four summaries on the window: the mean and the population standard deviation of the channel's finite values (both undefined when there is none); the fraction of window rows whose value is not finite; and the lag-1 autocorrelation, the Pearson correlation of the pairs of consecutive finite values, which is defined only when the window has at least 3 rows, there are at least 3 such pairs, and both lagged series have population standard deviation at least $10^{-8}$. An undefined summary is replaced by the pooled median, and every coordinate is then centered and scaled by pooled constants. This gives the feature vector $f\in\mathbb R^{d}$, $d=644$, one measurable function of the window applied identically to real and generated records. Let $P_R$ and $P_G$ be the laws of $f$ for the real and for the generated record of an audited stay, let $f_R\sim P_R$ and $f_G\sim P_G$ be independent, and put $\delta:=\mathbb Ef_G-\mathbb Ef_R$ when it is defined. For $\pi\in(0,1)$ let $\bar P:=(1-\pi)P_R+\pi P_G$. Call the pair $(P_R,P_G)$ \emph{standardized} if $\bar P$ has mean $0$ and variance $0$ or $1$ in every coordinate. Under (A1), with $\pi=\tfrac12$, the laws of the audit features form a standardized pair, and the audit's standardization leaves a standardized pair unchanged; every construction below is a standardized pair. For a Borel score $s:\mathbb R^d\to\mathbb R$ let $S_R=s(f_R)$, $S_G=s(f_G)$ and
\[\mathrm{AUC}(s):=\mathbb P(S_G>S_R)+\tfrac12\,\mathbb P(S_G=S_R).\]
Let $F^s_R$ and $F^s_G$ be the distribution functions of $S_R$ and $S_G$, let $d_{\mathrm{KS}}(s):=\sup_x|F^s_R(x)-F^s_G(x)|$, and let $\mathrm{TV}(P,Q):=\sup_A|P(A)-Q(A)|$ over Borel sets $A$. Put $g(t):=t-t^2/2$ and $\sigma(t):=1/(1+e^{-t})$. For $\lambda\ge0$ let
\[J_\lambda(b,w):=(1-\pi)\,\mathbb E\log\bigl(1+e^{\,b+w^\top f_R}\bigr)+\pi\,\mathbb E\log\bigl(1+e^{-(b+w^\top f_G)}\bigr)+\tfrac{\lambda}{2}\|w\|^2,\qquad b\in\mathbb{R},\ w\in\mathbb{R}^d ,\]
the population risk of $L_2$-penalized logistic regression with an unpenalized intercept, $\pi$ being the share of generated rows. A minimizer $(b^*,w^*)$ defines the population audit score $s^*(f)=b^*+w^{*\top}f$ and the population audit value $\mathrm{AUC}(s^*)$. The implemented audit has $\pi=\tfrac12$ and $\lambda=1/64$ (A2).

\medskip
\textbf{(a) Soundness, and no certificate of closeness.} For every Borel score $s$,
\[|\mathrm{AUC}(s)-\tfrac12|\le g(d_{\mathrm{KS}}(s))\le d_{\mathrm{KS}}(s)\le\mathrm{TV}(P_R,P_G)\qquad\text{and}\qquad|\mathrm{AUC}(s)-\tfrac12|\le g\bigl(\mathrm{TV}(P_R,P_G)\bigr).\]
The bounds are sharp: for every $t\in[0,1]$ there is a standardized pair with $d_{\mathrm{KS}}(s)=\mathrm{TV}=t$ and $\mathrm{AUC}(s)=\tfrac12+g(t)$ for the score $s(f)=f_1$. A population AUC of $\tfrac12\pm\varepsilon$ ($0\le\varepsilon\le\tfrac12$) therefore certifies $\mathrm{TV}(P_R,P_G)\ge d_{\mathrm{KS}}(s)\ge 1-(1-2\varepsilon)^{1/2}\ge\varepsilon$, and the same lower bound holds for the laws of the underlying records. No AUC value certifies that the laws are close:
(i) for every affine score $s(f)=b+w^\top f$ with $w\neq0$ and every $a\in[0,1]$, there is a standardized pair of mutually singular laws ($\mathrm{TV}=1$) with $\mathrm{AUC}(s)=a$;
(ii) for every $a\in[\tfrac12,1)$, $\pi\in(0,1)$, $\lambda>0$ and $d\ge2$, there is a standardized pair of mutually singular laws with finite second moments for which $J_\lambda$ has a unique minimizer and $\mathrm{AUC}(s^*)=a$.

\textbf{(b) First-moment blindness.} Assume $\mathbb{E}\|f_R\|+\mathbb{E}\|f_G\|<\infty$.
(i) If $\lambda>0$, then $J_\lambda$ has a unique minimizer, and $w^*=0$ if and only if $\delta=0$. In that case $b^*=\operatorname{logit}\pi$, $s^*$ is constant and $\mathrm{AUC}(s^*)=\tfrac12$.
(ii) If $\lambda=0$ and $\delta=0$, then $(\operatorname{logit}\pi,0)$ minimizes $J_0$. Every minimizer has a score equal to $\operatorname{logit}\pi$ almost surely under $P_R$ and under $P_G$, so its AUC is $\tfrac12$. The minimizer is unique when the covariance of $f$ under $\bar P$ is nonsingular.
(iii) Alignment. If $\delta\neq0$, every minimizer of $J_\lambda$, for any $\lambda\ge0$, satisfies $\delta^\top w^*>0$. In particular $w^*\ne0$ and $\mathrm{AUC}(s^*)>0$.
(iv) Hence, for $\lambda>0$ (and for $\lambda=0$ whenever a minimizer exists), the population audit value departs from $\tfrac12$ only if $\delta\neq0$. It equals $\tfrac12$ whenever the 644 feature means agree, whatever other differences coexist with equal means. These include differences in the variances and dependence of the summaries, and differences in properties of the records that $f$ does not summarize (cross-channel dynamics, lags beyond 1, within-stay distributions beyond the standard deviation). The mean of an imputed coordinate combines the probability that the summary is undefined with its mean where it is defined.
(v) Strong penalty. If $\delta\ne0$, then $\lambda w^*\to\pi(1-\pi)\,\delta$ as $\lambda\to\infty$, so $w^*/\|w^*\|\to\delta/\|\delta\|$, and $\mathrm{AUC}(s^*)\to\mathrm{AUC}(\delta^\top f)$ whenever $\mathbb P\bigl(\delta^\top(f_G-f_R)=0\bigr)=0$.
(vi) A mean difference does not force $\mathrm{AUC}(s^*)>\tfrac12$: for every $a\in(0,\tfrac12]$, $\pi\in(0,1)$, $\lambda>0$ and $d\ge1$, there is a standardized pair with $\delta\ne0$ and $\mathrm{TV}(P_R,P_G)=1$ for which $\mathrm{AUC}(s^*)=a$.

\textbf{(c) Gaussian calibration.} Suppose $P_R=N(\mu_R,\Sigma)$ and $P_G=N(\mu_G,\Sigma)$ with $\Sigma\succ0$, and let $D_M:=(\delta^\top\Sigma^{-1}\delta)^{1/2}$.
(i) A linear score $w^\top f$ has $\mathrm{AUC}=\Phi\bigl(w^\top\delta/(2w^\top\Sigma w)^{1/2}\bigr)$, and $\tfrac12$ if $w=0$. The largest AUC of any score, linear or not, is $\Phi(D_M/\sqrt2)$, and it is attained by $w\propto\Sigma^{-1}\delta$.
(ii) $\mathrm{TV}(P_R,P_G)=2\Phi(D_M/2)-1$.
(iii) For $\lambda>0$ the minimizer satisfies $w^*=\kappa(\alpha\Sigma+\lambda I)^{-1}\delta$, where $\kappa=(1-\pi)\mathbb{E}\,\sigma(s^*(f_R))>0$ and $\alpha=(1-\pi)\mathbb{E}\,\sigma'(s^*(f_R))+\pi\,\mathbb{E}\,\sigma'(s^*(f_G))\in(0,\tfrac14]$. If $\delta\neq0$, then $\tfrac12<\mathrm{AUC}(s^*)\le\Phi(D_M/\sqrt2)$, with equality if and only if $\delta$ is an eigenvector of $\Sigma$. For $\lambda=0$ the minimizer is unique, $w^*=\Sigma^{-1}\delta$, and equality holds.
(iv) Consequently a population audit value $a\in[\tfrac12,1)$ implies $D_M\ge\sqrt2\,\Phi^{-1}(a)$ and $\mathrm{TV}\ge2\Phi\bigl(\Phi^{-1}(a)/\sqrt2\bigr)-1$. Equality holds exactly when the audit direction is efficient, that is, when $\lambda=0$ or $\delta$ is an eigenvector of $\Sigma$. For the audit values $0.5716$ (full dataset), $0.6409$ (3,000-patient probe) and the threshold $0.75$, these lower bounds are $D_M\ge0.255,\ 0.510,\ 0.953$ and $\mathrm{TV}\ge0.101,\ 0.201,\ 0.366$ (rounded down). The distribution-free bound of (a) gives $\mathrm{TV}\ge0.074,\ 0.152,\ 0.292$. A pass ($a\le0.75$) bounds $D_M$ by $0.954$ and $\mathrm{TV}$ by $0.367$ (rounded up) only when the audit direction is efficient. For the penalized audit no such bound holds: for every $a\in(\tfrac12,1)$, $\pi\in(0,1)$, $\lambda>0$ and $d\ge3$ there are standardized Gaussian pairs with a common covariance whose population audit value lies in $(\tfrac12,a]$, while $D_M$ is arbitrarily large and $\mathrm{TV}$ is arbitrarily close to $1$.
(v) If $\delta=(0,\delta_V)$ in a partition $f=(f_U,f_V)$, then $D_M^2=\delta_V^\top\Sigma_{V\mid U}^{-1}\delta_V\ge\delta_V^\top\Sigma_{VV}^{-1}\delta_V$, where $\Sigma_{V\mid U}:=\Sigma_{VV}-\Sigma_{VU}\Sigma_{UU}^{-1}\Sigma_{UV}$.

\textbf{(d) Scope by construction.} Assume (A4) and (A4$'$). Let $U$ be the 608 coordinates formed by the four summaries of the 149 channels the generator does not write (the 144 channels other than the 12 written outcome channels and the 5 treatment channels that drive the generator, together with those 5 treatment channels) together with the missing fractions of the 12 written outcome channels. Let $V$ be the remaining 36 coordinates: the mean, within-stay standard deviation and lag-1 autocorrelation of the 12 written channels. Let $\tilde f_G$ be the feature vector of the generated record after its 5 treatment columns are replaced by the real ones, computed with the same imputation and scaling constants; let $\tilde P_G$ be its law and $\tilde\delta:=\mathbb E\tilde f_G-\mathbb Ef_R$.
(i) In every matched pair $\tilde f_{G,U}=f_{R,U}$. Hence $P_R$ and $\tilde P_G$ have the same marginal law on $U$, $\tilde\delta_U=0$, every score that depends on $f$ only through $f_U$ has AUC exactly $\tfrac12$ between them, and the two laws can differ only through the conditional law of $f_V$ given $f_U$.
(ii) For the pair $(P_R,\tilde P_G)$, $\nabla_wJ_\lambda(\operatorname{logit}\pi,0)=-\pi(1-\pi)\tilde\delta$, which is supported on $V$. If $\tilde\delta_V\neq0$, then $\tilde P_G\neq P_R$, and for $\lambda>0$ (implemented $1/64$) the minimizer exists, is unique and satisfies $\tilde\delta_V^\top w^*_V>0$. In the Gaussian model of (c), $\mathrm{AUC}(s^*)>\tfrac12$ and $D_M^2=\tilde\delta_V^\top\Sigma_{V\mid U}^{-1}\tilde\delta_V$. Without such a model, $\tilde\delta_V\neq0$ does not by itself force $\mathrm{AUC}(s^*)>\tfrac12$ (by (b)(vi)).
(iii) The generated treatment values are the single-precision roundings of the real doses, with a rounded dose not exceeding $10^{-9}$ set to $0$. So $P_G$ can differ from $\tilde P_G$, but only in the 20 treatment coordinates.
The threshold $0.75$ is a tolerance on how well this audit detects the difference. By (a), passing it certifies neither $P_G=P_R$ nor any upper bound on $\mathrm{TV}(P_R,P_G)$. All statements concern the audited generated dataset and its real source. The washout panels were audited separately under a registered protocol with the same features, discriminator and threshold (Methods), with $P_R$ and $P_G$ redefined as the laws of the 644 summaries of hours 24 to 163 of stays drawn as the panel rosters were. Parts (a) to (c) apply to that audit unchanged. Part (d)(i) holds for it because every channel other than the 12 written outcome channels is copied from the real record and the written channels' missing cells are restored, although the copied treatment channels did not drive the generated values; the treatment-rounding exception of (d)(iii) does not arise. The panels passed (audit AUC 0.714 to 0.727), but the audit did not flag measurement noise of 0.15 reference standard deviations added to the real records.

\paragraph{Assumptions.}
\begin{description}
\item[(A1) Population feature map.] The median-imputation values and the centering and scaling constants are fixed at their population values under the pooled law with $\pi=\tfrac12$, and they act coordinatewise and identically on both groups. So $f=\Psi(X)$ for one measurable map $\Psi$ of the stay record $X$ restricted to its window, and $(P_R,P_G)$ is a standardized pair. $P_R$ and $P_G$ are the laws of $f$ for the real and the generated record of a stay drawn uniformly from the matched stays of the audited dataset; the audit's 400 stays are a uniform random subset of the 200,825 matched stays. In the implementation (\texttt{icare\_dt/audit\_semisynth\_realism.py}, l.~80--100 and 155--182), imputation and scaling are refitted on each pooled training fold. The population statement idealizes this.

\item[(A2) Population target of the audit.] The audit's statistic is the pooled out-of-fold AUC of \texttt{LogisticRegression(C=0.1)} with the default lbfgs solver, after \texttt{SimpleImputer(median)} and \texttt{StandardScaler}. The solver penalizes only the slope vector, so the intercept is unpenalized, and its objective is the mean log loss plus $\|w\|^2/(2Cn)$. Five person-grouped folds on 400 matched pairs (each person contributes exactly one real and one generated row) give $n_{\mathrm{train}}=640$, $\pi=\tfrac12$ exactly and $\lambda=1/(0.1\cdot640)=1/64$ in every training fold. The imputer removes a column undefined for every training row. In the population idealization such a coordinate is the constant $0$, which enters $J_\lambda$ only through the penalty and receives weight $0$ (Lemma 3), so the two conventions give the same score. On the probe, 76 of the 644 columns are undefined for all 800 rows (72 lag-1 autocorrelations, 2 means, 2 standard deviations), all in $U$. The population target is $\mathrm{AUC}(s^*)$, and parts (a) to (c) are population statements. The shipped number pools the out-of-fold scores of five fitted models over $400^2$ comparisons, including the 400 within-pair ones, with $d=644$ close to $n_{\mathrm{train}}=640$. So it matches $\mathrm{AUC}(s^*)$ only up to estimation error. The null band and the reproduction are given in the numerical checks.

\item[(A3) Moments.] $\mathbb{E}\|f_R\|+\mathbb{E}\|f_G\|<\infty$ wherever the risk $J_\lambda$ or $\delta$ is used. This holds for standardized features.

\item[(A4) Generator configuration, for (d).] The audited dataset is \texttt{blended\_data\_semisynth\_\allowbreak eicu\_complex\_v2\_seed13\_\allowbreak faithful\_v3\_a434a5c}. It was generated at commit a434a5c, and its configuration records generator SHA-256 \texttt{f29b4eea\ldots}, which equals that of \texttt{generate\_semisynth.py} at a434a5c. The configuration has \texttt{preserve\_observed\_missingness}, \texttt{hold\_carried\_forward\_outcomes} and \texttt{preserve\_treatment\_missingness} on, \texttt{factual\_treatment\_policy=observed\_dose}, \texttt{treatment\_off\_raw}$=0$ for all five treatments, \texttt{max\_len}$=336$ with a head window, and \texttt{measurement\_base\_rate}$=1$ with the other synthetic measurement parameters at $0$. The bounded-memory route is off (\texttt{trait\_response\_route=legacy}, cap $0$). The generator changes between a434a5c and the paper commit 82587202 affect only the bounded-memory route and optional counterfactual path outputs, so the factual path used here is unchanged. Line numbers in the proof refer to 82587202. The same configuration governs the 3,000-patient probe used for verification.

\item[(A4$'$) Data conditions, for (d).]
(1) Source identity. For every audited stay, the real frame the generator read (\texttt{preprocessed\_timeseries} under its data root, G l.~2489) equals, on the 161 audited columns and the window, the real frame the audit read (\texttt{preprocessed\_timeseries} under \texttt{real\_root}$=$\texttt{.../blended\_icu/blended\_data}). The generator's configuration does not record its data root, so this is an assumption.
(2) Every real cell of the 161 audited columns in the window is finite or missing; none is $\pm\infty$.
(3) Every finite real treatment value is nonnegative.
Conditions (1) to (3) are not verified over all 200,825 stays. They are consistent with the audited pairs. On the probe, the 144 other channels agree exactly, the treatment summaries agree to $1.1\times10^{-5}$ with identical missingness, and the written channels' missing fractions agree exactly. On the full dataset's audit summary, the missing-fraction difference is $0$ on all 161 channels and the 149 unwritten channels have $|\mathrm{SMD}|\le1.3\times10^{-9}$.

\item[(A5) Gaussian idealization, for (c).] The equal-covariance Gaussian model is a calibration device and does not describe the data. Several coordinates are fractions, are bounded, or carry atoms after imputation.

\item[(A6) Scope.] $P_R$ and $P_G$ refer to the audited generated dataset and its real eICU source only. The washout panels come from a separate parameterization, with simulated dosing, mean-filled cells and an injected memory route, and their separate audit (Methods) is not covered by (A1)'s sampling clause, (A4) or (A4$'$).
\end{description}

\paragraph{Proof.}

\paragraph{Conventions.} Real rows carry the label $y=0$ and generated rows the label $y=1$. $\bar{\mathbb E}$ denotes expectation under the mixture in which $y\sim\mathrm{Bernoulli}(\pi)$ and, given $y$, $f\sim P_R$ if $y=0$ and $f\sim P_G$ if $y=1$. The law of $f$ under this mixture is $\bar P$, and $P_R\ll\bar P$, $P_G\ll\bar P$ because $0<\pi<1$. With $\ell(y,t):=\log(1+e^{t})-yt$ we have $\ell(0,t)=\log(1+e^t)$ and $\ell(1,t)=\log(1+e^{-t})$, so
\[J_\lambda(b,w)=\bar{\mathbb E}\,\ell(y,b+w^\top f)+\tfrac\lambda2\|w\|^2 .\]
Moreover $\partial_t\ell(y,t)=\sigma(t)-y\in(-1,1)$ and $\partial_t^2\ell(y,t)=\sigma'(t)=\sigma(t)(1-\sigma(t))\in(0,\tfrac14]$, so $\ell(y,\cdot)$ is strictly convex. Also $\ell(0,t)\ge t$, $\ell(1,t)\ge-t$ and $0\le\ell(y,t)\le\log2+|t|$. Write $m_R:=\mathbb E\|f_R\|$, $m_G:=\mathbb E\|f_G\|$ and $\theta_0:=(\operatorname{logit}\pi,0)$. The function $g(t)=t-t^2/2$ is increasing on $[0,1]$ and satisfies $g(t)\le t$.

\emph{Standardizing a construction.} Applying one strictly increasing affine map to a coordinate in both groups leaves unchanged the TV, and the AUC and $d_{\mathrm{KS}}(s)$ of any score that is a strictly increasing function of that coordinate. Adjoining coordinates that have the same law in both groups and, within each group, are independent of the others also leaves the TV unchanged: $\mathrm{TV}(\mu\otimes\nu,\mu'\otimes\nu)=\mathrm{TV}(\mu,\mu')$. Indeed, sets $A\times\mathbb R^k$ give $\ge$, and for $\le$, $|(\mu\otimes\nu-\mu'\otimes\nu)(A)|=|\int(\mu-\mu')(A^z)\,\nu(dz)|\le\mathrm{TV}(\mu,\mu')$, with $A^z$ the section of $A$ at $z$. Each construction below is either standardized by these two operations or written directly in standardized form.

\paragraph{Lemma 1 (mid-distribution form of the AUC).} For a real random variable $S$ with distribution function $F$ put $M(x):=\mathbb P(S<x)+\tfrac12\mathbb P(S=x)=\tfrac12\{F(x)+F(x-)\}$. Let $M_R$ and $M_G$ be these functions for $S_R$ and $S_G$. Then
\[\mathrm{AUC}(s)=\mathbb E\,M_R(S_G)=1-\mathbb E\,M_G(S_R),\qquad\mathbb E\,M_R(S_R)=\mathbb E\,M_G(S_G)=\tfrac12 .\]
\emph{Proof.} By independence and Fubini, $\mathbb P(S_G>S_R\mid S_G)+\tfrac12\mathbb P(S_G=S_R\mid S_G)=M_R(S_G)$, and taking expectations gives the first identity. The events $\{S_G>S_R\}$, $\{S_G<S_R\}$ and $\{S_G=S_R\}$ partition the sample space, so $\mathrm{AUC}(s)=1-\{\mathbb P(S_R>S_G)+\tfrac12\mathbb P(S_R=S_G)\}=1-\mathbb E M_G(S_R)$. If $S'$ is an independent copy of $S$, exchangeability gives $\mathbb P(S'<S)=\mathbb P(S'>S)$, so $\mathbb E M(S)=\mathbb P(S'<S)+\tfrac12\mathbb P(S'=S)=\tfrac12$. $\square$

\paragraph{Lemma 2 (the population logistic risk).} Assume $m_R+m_G<\infty$ and $\lambda\ge0$. Then $J_\lambda$ is finite, convex and continuously differentiable on $\mathbb R^{1+d}$, with
\[\partial_bJ_\lambda=\bar{\mathbb E}[\sigma(s)-y],\qquad\nabla_wJ_\lambda=\bar{\mathbb E}[(\sigma(s)-y)f]+\lambda w,\qquad s=b+w^\top f .\]
Hence $(b,w)$ minimizes $J_\lambda$ if and only if both derivatives vanish there. At $\theta_0$,
\[\partial_bJ_\lambda(\theta_0)=0,\qquad\nabla_wJ_\lambda(\theta_0)=-\pi(1-\pi)\,\delta,\qquad J_\lambda(\theta_0)=H(\pi):=-\pi\log\pi-(1-\pi)\log(1-\pi).\]
If $\lambda>0$, $J_\lambda$ has exactly one minimizer.

\emph{Proof.} Finiteness follows from $0\le\ell\le\log2+|b|+\|w\|\,\|f\|$. The partial derivatives of $\ell(y,b+w^\top f)$ in $(b,w)$ are bounded by $1$ and by $\|f\|$, both integrable. Differentiation under $\bar{\mathbb E}$ is therefore justified by dominated convergence, and the same bound with the continuity of $\sigma$ makes the gradient continuous. $J_\lambda$ is convex because $\ell(y,\cdot)$ is convex and $s$ is affine in $(b,w)$. For a convex differentiable function on $\mathbb R^{1+d}$, a point is a minimizer exactly when the gradient vanishes. At $\theta_0$ the score is the constant $\operatorname{logit}\pi$ and $\sigma(s)=\pi$, so $\partial_bJ_\lambda(\theta_0)=\pi-\bar{\mathbb E}y=0$ and $\nabla_wJ_\lambda(\theta_0)=\bar{\mathbb E}[(\pi-y)f]=(1-\pi)\pi\,\mathbb Ef_R-\pi(1-\pi)\,\mathbb Ef_G=-\pi(1-\pi)\delta$; the value $H(\pi)$ follows by substitution.

Now let $\lambda>0$. From $\ell(0,t)\ge t$ and $\ell(1,t)\ge-t$,
\[J_\lambda(b,w)\ge\max\bigl\{\tfrac\lambda2\|w\|^2,\ (1-\pi)(b-\|w\|m_R),\ \pi(-b-\|w\|m_G)\bigr\}.\]
This tends to $\infty$ as $\|(b,w)\|\to\infty$: if $\|w\|$ stays bounded, then $|b|\to\infty$ and one of the last two terms diverges. A continuous coercive function attains its minimum. Let $(b_1,w_1)$ and $(b_2,w_2)$ be minimizers with value $J^*$. By convexity their midpoint is also a minimizer. If $w_1\neq w_2$, the penalty is strictly convex along the segment and the loss term is convex, so the midpoint would have value $<J^*$. Hence $w_1=w_2=:w$. The function $b\mapsto J_\lambda(b,w)$ has second derivative $\bar{\mathbb E}\,\sigma'(b+w^\top f)>0$ (the integrand is bounded by $\tfrac14$), so it is strictly convex and $b_1=b_2$. $\square$

\paragraph{Lemma 3 (coordinates without mean information).} Write $f=(f_A,f_N)$ and $w=(w_A,w_N)$, and suppose that within each group $f_N$ is independent of $f_A$ and has mean $0$, with $m_R+m_G<\infty$. Then
\[J_\lambda(b,w_A,w_N)\ge J_\lambda(b,w_A,0)+\tfrac\lambda2\|w_N\|^2 .\]
Consequently, if $\lambda>0$, every minimizer has $w^*_N=0$, and $(b^*,w^*_A)$ minimizes $J_\lambda(\cdot,\cdot,0)$, which is the risk computed from $f_A$ alone.

\emph{Proof.} Given $(y,f_A)$, independence within each group gives $\mathbb E[f_N\mid y,f_A]=\mathbb E[f_N\mid y]=0$. Conditional Jensen for the convex function $\ell(y,\cdot)$ gives $\bar{\mathbb E}[\ell(y,b+w_A^\top f_A+w_N^\top f_N)\mid y,f_A]\ge\ell(y,b+w_A^\top f_A)$. Take expectations and add the penalties, which split as $\tfrac\lambda2\|w_A\|^2+\tfrac\lambda2\|w_N\|^2$. If $\lambda>0$ and a minimizer had $w^*_N\neq0$, then $J_\lambda(b^*,w^*_A,0)<J_\lambda(b^*,w^*)$, a contradiction. $\square$

\paragraph{Proof of (a): the bounds.} Write $F_c:=F^s_c$.

\emph{Scores without atoms.} Suppose first that $S_R$ and $S_G$ have no atoms. Then $M_c=F_c$, and Lemma 1 gives both
\[\mathrm{AUC}(s)-\tfrac12=\mathbb E[F_R(S_G)-F_G(S_G)]\quad\text{and}\quad\mathrm{AUC}(s)-\tfrac12=\mathbb E[F_R(S_R)-F_G(S_R)].\]
Average the two. With $D:=F_R-F_G$ and $Q:=\tfrac12\{\mathcal L(S_R)+\mathcal L(S_G)\}$, this gives $\mathrm{AUC}(s)-\tfrac12=\int D\,dQ$. The distribution function of $Q$ is $F_Q=\tfrac12(F_R+F_G)$, and pointwise
\[|D|\le d_{\mathrm{KS}}(s),\qquad|D|\le\max(F_R,F_G)\le2F_Q,\qquad|D|\le\max(1-F_R,1-F_G)\le2(1-F_Q).\]
So $|\mathrm{AUC}(s)-\tfrac12|\le\int h(F_Q)\,dQ$ with $h(u):=\min\{d_{\mathrm{KS}}(s),2u,2(1-u)\}$. Since $Q$ has no atoms, $F_Q(Z)$ is uniform on $(0,1)$ for $Z\sim Q$. Hence
\[\int h(F_Q)\,dQ=\int_0^1h(u)\,du=\tfrac{d_{\mathrm{KS}}(s)^2}{4}+d_{\mathrm{KS}}(s)(1-d_{\mathrm{KS}}(s))+\tfrac{d_{\mathrm{KS}}(s)^2}{4}=g(d_{\mathrm{KS}}(s)).\]

\emph{General scores.} Let $U_R$ and $U_G$ be uniform on $(0,1)$, independent of each other and of $(f_R,f_G)$. For $\epsilon>0$ put $S_c^\epsilon:=S_c+\epsilon U_c$. These variables have no atoms, and their distribution functions $F_c^\epsilon(x)=\int_0^1F_c(x-\epsilon u)\,du$ satisfy $|F_R^\epsilon-F_G^\epsilon|\le d_{\mathrm{KS}}(s)$. By the atomless case, $|\mathbb P(S^\epsilon_G>S^\epsilon_R)-\tfrac12|\le g(d_{\mathrm{KS}}(s))$. As $\epsilon\downarrow0$, the indicator $1\{S^\epsilon_G>S^\epsilon_R\}$ converges to $1\{S_G>S_R\}$ on $\{S_G\neq S_R\}$. On $\{S_G=S_R\}$ it equals $1\{U_G>U_R\}$, which is independent of that event and has probability $\tfrac12$. By dominated convergence, $\mathbb P(S^\epsilon_G>S^\epsilon_R)\to\mathrm{AUC}(s)$. Hence $|\mathrm{AUC}(s)-\tfrac12|\le g(d_{\mathrm{KS}}(s))$ for every score.

\emph{The chain.} For every $x$, $F_R(x)-F_G(x)=P_R(A_x)-P_G(A_x)$ with the Borel set $A_x:=s^{-1}((-\infty,x])$, so $d_{\mathrm{KS}}(s)\le\mathrm{TV}(P_R,P_G)$. As $g$ is increasing on $[0,1]$ with $g(t)\le t$, we get $g(d_{\mathrm{KS}}(s))\le d_{\mathrm{KS}}(s)\le\mathrm{TV}$ and $g(d_{\mathrm{KS}}(s))\le g(\mathrm{TV})$.

\emph{Attainment.} Fix $t\in[0,1]$. Let a raw variable $r$ have law $(1-t)\,\mathrm U(0,1)+t\,\delta_{-1}$ in the real group and $(1-t)\,\mathrm U(0,1)+t\,\delta_{2}$ in the generated group. Let $f_1$ be the pooled standardization of $r$, an increasing affine map (the pooled variance of $r$ is positive), let $f_2,\dots,f_d$ be independent $N(0,1)$ in both groups, and take $s(f)=f_1$. The pair is standardized. Two uniform draws compare with value $\tfrac12$. Every comparison that involves the generated atom at $2$ or the real atom at $-1$ is won by the generated draw. Hence $\mathrm{AUC}=\tfrac12(1-t)^2+2t(1-t)+t^2=\tfrac12+g(t)$. In the scale of $r$, $F_R-F_G=t$ on $[-1,2)$ and $0$ elsewhere, so $d_{\mathrm{KS}}(s)=t$, and the two laws of $r$ differ by $t(\delta_{-1}-\delta_2)$, so their TV is $t$. By the conventions these values carry over to $f$.

\emph{Certificate.} If $|\mathrm{AUC}(s)-\tfrac12|=\varepsilon$, then $g(d_{\mathrm{KS}}(s))\ge\varepsilon$. With $r_\varepsilon:=(1-2\varepsilon)^{1/2}$, $g(1-r_\varepsilon)=(1-r_\varepsilon)(1+r_\varepsilon)/2=\varepsilon$. Since $g$ is increasing on $[0,1]$, $d_{\mathrm{KS}}(s)\ge1-r_\varepsilon$. Moreover $1-r_\varepsilon\ge\varepsilon$ because $(1-\varepsilon)^2\ge1-2\varepsilon$. For the records, $f=\Psi(X)$ with $\Psi$ measurable (A1), and $\{f\in A\}=\{X\in\Psi^{-1}(A)\}$. So $\mathrm{TV}(P_R,P_G)\le\mathrm{TV}(\mathcal L(X_R),\mathcal L(X_G))$. $\square$

\paragraph{Proof of (a)(i).} Let $s(f)=b+w^\top f$ with $w\neq0$ and let $a\in[0,1]$. Put $v_j:=1$ if $w_j\ge0$ and $v_j:=-1$ otherwise, so $v_j^2=1$ and $w^\top v=\|w\|_1>0$. Let $f=Xv$ for a scalar $X$ whose law in each group is given below. Then every coordinate $f_j=v_jX$ has pooled mean $v_j\bar{\mathbb E}X$ and pooled variance $\mathrm{Var}_{\bar P}(X)$, and $s(f)=b+\|w\|_1X$ is strictly increasing in $X$. In every case below $X$ takes finitely many values, with disjoint supports in the two groups, so the laws of $f$ have disjoint finite supports and $\mathrm{TV}=1$.
For $a\in(0,1)$, let $X=0$ in the real group, and in the generated group let $X$ take the values $t(1-a)$ and $-ta$ with probabilities $a$ and $1-a$, where $t:=(\pi a(1-a))^{-1/2}$. Then $\bar{\mathbb E}X=\pi t\{a(1-a)-(1-a)a\}=0$, $\mathrm{Var}_{\bar P}(X)=\pi t^2\{a(1-a)^2+(1-a)a^2\}=\pi t^2a(1-a)=1$, and $\mathrm{AUC}(s)=\mathbb P(X_G>0)=a$.
For $a=1$, let $X=-(\pi/(1-\pi))^{1/2}$ in the real group and $X=((1-\pi)/\pi)^{1/2}$ in the generated group. The pooled mean is $-(\pi(1-\pi))^{1/2}+(\pi(1-\pi))^{1/2}=0$, the pooled variance is $\pi+(1-\pi)=1$, and $\mathrm{AUC}(s)=1$. For $a=0$ exchange the two signs.
In every case the pair is standardized. $\square$

\paragraph{Proof of (b).}
(i) Let $\lambda>0$. By Lemma 2, $\nabla J_\lambda(\theta_0)=(0,-\pi(1-\pi)\delta)$. If $\delta=0$, then $\theta_0$ is a stationary point. It is therefore the minimizer, which is unique by Lemma 2, so $w^*=0$. Conversely, suppose $w^*=0$. Then $b^*$ minimizes $b\mapsto J_\lambda(b,0)=(1-\pi)\log(1+e^b)+\pi\log(1+e^{-b})$, whose derivative $\sigma(b)-\pi$ vanishes only at $\operatorname{logit}\pi$. So $(b^*,w^*)=\theta_0$, and $0=\nabla_wJ_\lambda(\theta_0)=-\pi(1-\pi)\delta$ forces $\delta=0$. With $w^*=0$ the score is the constant $\operatorname{logit}\pi$, so $S_G=S_R$ surely and $\mathrm{AUC}(s^*)=\tfrac12$.

(ii) Let $\lambda=0$ and $\delta=0$. By Lemma 2, $\theta_0$ is stationary, and hence a minimizer of the convex function $J_0$. Let $\theta_1=(b_1,w_1)$ be any minimizer, and put $\Delta:=(b_1-\operatorname{logit}\pi)+w_1^\top f$ and $\theta_t:=\theta_0+t(\theta_1-\theta_0)$. By convexity $J_0(\theta_t)=J^*$ for $t\in[0,1]$. Suppose $\bar P(\Delta\neq0)>0$. For $t\in(0,1)$, strict convexity of $\ell(y,\cdot)$ gives $\ell(y,\operatorname{logit}\pi+t\Delta)<(1-t)\ell(y,\operatorname{logit}\pi)+t\,\ell(y,\operatorname{logit}\pi+\Delta)$ on $\{\Delta\neq0\}$, with equality elsewhere. Taking expectations gives $J_0(\theta_t)<J^*$, a contradiction. Hence $b_1+w_1^\top f=\operatorname{logit}\pi$ holds $\bar P$-almost surely, and therefore almost surely under $P_R$ and under $P_G$. The score is almost surely constant and its AUC is $\tfrac12$. If $\mathrm{Cov}_{\bar P}(f)$ is nonsingular, an almost surely constant $w_1^\top f$ forces $w_1=0$, and then $b_1=\operatorname{logit}\pi$.

(iii) Let $\delta\ne0$, $\lambda\ge0$, and let $\theta^*=(b^*,w^*)$ be a minimizer. By Lemma 2, $\nabla J_\lambda(\theta_0)\ne0$, so $\theta_0$ is not a minimizer and $J_\lambda(\theta^*)<J_\lambda(\theta_0)$. Convexity and differentiability give
\[J_\lambda(\theta^*)\ge J_\lambda(\theta_0)+\nabla J_\lambda(\theta_0)^\top(\theta^*-\theta_0)=J_\lambda(\theta_0)-\pi(1-\pi)\,\delta^\top w^* .\]
Hence $\pi(1-\pi)\,\delta^\top w^*\ge J_\lambda(\theta_0)-J_\lambda(\theta^*)>0$. Then $\mathbb E[w^{*\top}(f_G-f_R)]=w^{*\top}\delta>0$, so $\mathbb P(w^{*\top}f_G>w^{*\top}f_R)>0$, and $\mathrm{AUC}(s^*)\ge\mathbb P(S_G>S_R)>0$.

(iv) The population audit value depends on the laws of the records only through $(P_R,P_G)$. By (i) and (ii) it equals $\tfrac12$ on the set $\{\delta=0\}$, which restricts the laws only through their means. For the imputation remark, suppose the raw summary $r_j$ is undefined with probability $\nu_{c,j}$ in group $c$, is imputed by $m_j$, and is standardized as $(r-c_j)/\tau_j$. Then $\mathbb Ef_{c,j}=\{(1-\nu_{c,j})\,\mathbb E(r_j\mid r_j\text{ defined},\,c)+\nu_{c,j}m_j-c_j\}/\tau_j$.

(v) Let $\theta_\lambda=(b_\lambda,w_\lambda)$ be the minimizer for $\lambda>0$. From $J_\lambda(\theta_\lambda)\le J_\lambda(\theta_0)=H(\pi)$ we get $\|w_\lambda\|^2\le2H(\pi)/\lambda\to0$. Once $\|w_\lambda\|\le1$, the lower bounds in the proof of Lemma 2 give $-H(\pi)/\pi-m_G\le b_\lambda\le H(\pi)/(1-\pi)+m_R$. Take any subsequence along which $b_\lambda\to\bar b$. Dominated convergence in $\bar{\mathbb E}[\sigma(b_\lambda+w_\lambda^\top f)-y]=0$ gives $\sigma(\bar b)=\pi$, so $b_\lambda\to\operatorname{logit}\pi$. Dominated convergence in the second first-order condition, with the integrand bounded by $\|f\|$, then gives
\[\lambda w_\lambda=\bar{\mathbb E}\bigl[(y-\sigma(b_\lambda+w_\lambda^\top f))f\bigr]\ \longrightarrow\ \bar{\mathbb E}[(y-\pi)f]=\pi(1-\pi)\,\delta .\]
If $\delta\ne0$, then $w_\lambda\ne0$ for large $\lambda$, and $u_\lambda:=w_\lambda/\|w_\lambda\|=\lambda w_\lambda/\|\lambda w_\lambda\|\to u:=\delta/\|\delta\|$. The AUC of $s^*$ equals that of $u_\lambda^\top f$, because the intercept and a positive factor do not change the ordering. With $\Delta:=f_G-f_R$, $\mathrm{AUC}(s^*)=\mathbb E[1\{u_\lambda^\top\Delta>0\}+\tfrac12 1\{u_\lambda^\top\Delta=0\}]$. On the event $\{u^\top\Delta\ne0\}$, $u_\lambda^\top\Delta\to u^\top\Delta$, so the integrand converges to $1\{u^\top\Delta>0\}$. If this event has probability one, dominated convergence gives $\mathrm{AUC}(s^*)\to\mathbb P(u^\top\Delta>0)=\mathrm{AUC}(\delta^\top f)$.

(vi) Let $a\in(0,\tfrac12]$, $\pi\in(0,1)$, $\lambda>0$, $d\ge1$ and $x_a:=2(1-a)/a$. Let a raw variable $r$ equal $0$ in the real group, and in the generated group equal $1$ with probability $1-a$ and $-x_a$ with probability $a$. Its pooled mean is $\bar r:=\pi\{(1-a)-ax_a\}=-\pi(1-a)$, and its pooled variance $v$ is positive because $r$ is not constant in the generated group. Put $f_1:=(r-\bar r)/v^{1/2}$, and let $f_2,\dots,f_d$ be independent $N(0,1)$ in both groups. The pair is standardized, $\delta_1=-(1-a)/v^{1/2}<0$ and $\delta_k=0$ for $k\ge2$, so $\delta\ne0$. The point $-\bar r/v^{1/2}$ carries all the mass of $f_1$ in the real group and none in the generated group, so $\mathrm{TV}=1$. By Lemma 3, $w^*_k=0$ for $k\ge2$. By (iii), $\delta_1w^*_1>0$, so $w^*_1<0$ and $s^*$ is strictly decreasing in $f_1$. Hence $\mathrm{AUC}(s^*)=\mathbb P(f_{G,1}<f_{R,1})+\tfrac12\mathbb P(f_{G,1}=f_{R,1})=\mathbb P(r_G=-x_a)=a$, for every $\lambda>0$. $\square$

\paragraph{Proof of (c).} We use Stein's lemma: if $X\sim N(\mu,\Sigma)$ and $h:\mathbb R^d\to\mathbb R$ is continuously differentiable with bounded gradient, then $\mathbb E[h(X)(X-\mu)]=\Sigma\,\mathbb E\nabla h(X)$.

(i) For $w$ with $w^\top\Sigma w>0$, independence gives $w^\top f_G-w^\top f_R\sim N(w^\top\delta,\,2w^\top\Sigma w)$. This law has no atoms, so $\mathrm{AUC}=\Phi\bigl(w^\top\delta/(2w^\top\Sigma w)^{1/2}\bigr)$. For $w=0$ the score is constant and $\mathrm{AUC}=\tfrac12=\Phi(0)$. By Cauchy--Schwarz, $w^\top\delta=(\Sigma^{1/2}w)^\top(\Sigma^{-1/2}\delta)\le(w^\top\Sigma w)^{1/2}D_M$, with equality if and only if $w=c\,\Sigma^{-1}\delta$ for some $c\ge0$. As $\Phi$ is increasing, the largest AUC over linear scores is $\Phi(D_M/\sqrt2)$.

Now let $s$ be any score. If $\delta=0$, then $P_R=P_G$ and Lemma 1 gives $\mathrm{AUC}(s)=\mathbb EM_R(S_R)=\tfrac12=\Phi(0)$. Let $\delta\neq0$. Let $U$ be uniform on $(0,1)$ and independent of $f$, and put $T:=F_R(s(f)-)+U\{F_R(s(f))-F_R(s(f)-)\}$. Under $P_R$, $T$ is uniform on $(0,1)$ (the randomized probability integral transform). Also $\mathbb E[T\mid f]=M_R(s(f))$. By Lemma 1, $\mathrm{AUC}(s)=\mathbb E\,T(f_G,U)=\int_0^1\mathbb P(T(f_G,U)>u)\,du$. The randomized test $1\{T>u\}$ has size $1-u$ under $P_R$. By the Neyman--Pearson lemma, its power under $P_G$ is at most the power $\beta(1-u)$ of the likelihood-ratio test of that size. The likelihood ratio $dP_G/dP_R$ is a strictly increasing function of $L(f):=\delta^\top\Sigma^{-1}f$, whose law has no atoms under either group. For the score $L$, the test $1\{T>u\}$ is therefore the likelihood-ratio test of size $1-u$. The same identity gives $\mathrm{AUC}(L)=\int_0^1\beta(1-u)\,du\ge\mathrm{AUC}(s)$, and $\mathrm{AUC}(L)=\Phi(D_M/\sqrt2)$ by the linear formula.

(ii) The affine bijection $x\mapsto\Sigma^{-1/2}(x-\mu_R)$ maps Borel sets onto Borel sets and so preserves $\mathrm{TV}$. It sends the two laws to $N(0,I)$ and $N(\Sigma^{-1/2}\delta,I)$, and a rotation, which also preserves $\mathrm{TV}$, sends $\Sigma^{-1/2}\delta$ to $D_Me_1$. The laws then have densities $\varphi(x_1)\rho(x_{2:d})$ and $\varphi(x_1-D_M)\rho(x_{2:d})$, with $\rho$ the standard normal density on $\mathbb R^{d-1}$. Using $\mathrm{TV}(P,Q)=\tfrac12\int|p-q|$ for densities, and the fact that $\varphi(x)>\varphi(x-D_M)$ if and only if $x<D_M/2$,
\[\mathrm{TV}=\tfrac12\int|\varphi(x_1)-\varphi(x_1-D_M)|\,dx_1=\int_{x_1<D_M/2}\{\varphi(x_1)-\varphi(x_1-D_M)\}\,dx_1=2\Phi(D_M/2)-1 .\]

(iii) Let $\lambda>0$. By Lemma 2 (Gaussian laws have all moments) there is a unique minimizer $(b^*,w^*)$; write $s^*=b^*+w^{*\top}f$. Apply Stein's lemma under $P_R$ to $h_R(x)=\sigma(b^*+w^{*\top}x)$, and under $P_G$ to $h_G(x)=\sigma(b^*+w^{*\top}x)-1$. Both have gradient $\sigma'(b^*+w^{*\top}x)\,w^*$, so
\[\mathbb E[h_c(f_c)f_c]=\mu_c\,\mathbb E\,h_c(f_c)+\Sigma w^*\,\mathbb E\,\sigma'(s^*(f_c)).\]
The first-order condition in $b$ reads $(1-\pi)\mathbb E\sigma(s^*(f_R))=\pi\,\mathbb E\{1-\sigma(s^*(f_G))\}=:\kappa>0$. Substituting into the condition for $w$,
\[0=(1-\pi)\mathbb E[h_R(f_R)f_R]+\pi\,\mathbb E[h_G(f_G)f_G]+\lambda w^*=\kappa\mu_R-\kappa\mu_G+(\alpha\Sigma+\lambda I)w^*,\]
so $w^*=\kappa(\alpha\Sigma+\lambda I)^{-1}\delta$, with $\alpha\in(0,\tfrac14]$ because $0<\sigma'\le\tfrac14$. If $\delta\neq0$, then $w^{*\top}\delta=\kappa\,\delta^\top(\alpha\Sigma+\lambda I)^{-1}\delta>0$, so by (i) $\tfrac12<\mathrm{AUC}(s^*)\le\Phi(D_M/\sqrt2)$. By (i), equality holds if and only if $w^*$ is a positive multiple of $\Sigma^{-1}\delta$. Put $c:=\lambda/\alpha>0$. The condition $(\Sigma+cI)^{-1}\delta=t\,\Sigma^{-1}\delta$ for some $t>0$ is equivalent to $(1-t)\delta=tc\,\Sigma^{-1}\delta$. Since $\delta\neq0$ and $c>0$, this forces $\Sigma^{-1}\delta$, and hence $\Sigma\delta$, to be a multiple of $\delta$. Conversely, if $\Sigma\delta=\nu\delta$, then $(\Sigma+cI)^{-1}\delta=\delta/(\nu+c)$ and $\Sigma^{-1}\delta=\delta/\nu$.

Now let $\lambda=0$. Under the mixture, $\mathbb P(y=1\mid f)=\sigma(\beta_0+\beta^\top f)$ with $\beta=\Sigma^{-1}\delta$ and $\beta_0=\operatorname{logit}\pi-\tfrac12(\mu_G^\top\Sigma^{-1}\mu_G-\mu_R^\top\Sigma^{-1}\mu_R)$. For any $(b,w)$, $J_0(b,w)-J_0(\beta_0,\beta)=\bar{\mathbb E}\,\mathrm{KL}\bigl(\mathrm{Bern}(\sigma(\beta_0+\beta^\top f))\,\|\,\mathrm{Bern}(\sigma(b+w^\top f))\bigr)\ge0$. Equality holds if and only if $b+w^\top f=\beta_0+\beta^\top f$ $\bar P$-almost surely, that is, $(b,w)=(\beta_0,\beta)$, because $\mathrm{Cov}_{\bar P}(f)=\Sigma+\pi(1-\pi)\delta\delta^\top\succ0$. So $w^*=\Sigma^{-1}\delta$ and equality holds in (i).

(iv) The audit score is linear, so by (iii) $a=\mathrm{AUC}(s^*)=\Phi(D_\lambda/\sqrt2)$ with $D_\lambda:=w^{*\top}\delta/(w^{*\top}\Sigma w^*)^{1/2}\le D_M$ (and $D_\lambda:=0$ when $\delta=0$). Hence $D_M\ge\sqrt2\,\Phi^{-1}(a)$, and by (ii), $\mathrm{TV}=2\Phi(D_M/2)-1\ge2\Phi(\Phi^{-1}(a)/\sqrt2)-1$. Both hold with equality exactly when $D_\lambda=D_M$, which by (iii) means $\lambda=0$ or $\delta$ an eigenvector of $\Sigma$. The numbers are $\sqrt2\,\Phi^{-1}(0.57163)=0.2553$, $\sqrt2\,\Phi^{-1}(0.64095)=0.5105$ and $\sqrt2\,\Phi^{-1}(0.75)=0.9539$, with corresponding $2\Phi(D/2)-1=0.1016,\ 0.2015,\ 0.3666$. The distribution-free values $1-(1-2\varepsilon)^{1/2}$ are $0.0744$, $0.1526$ and $0.2929$. Lower bounds are rounded down in the statement and the upper limits $0.954$ and $0.367$ are rounded up.

For the last claim, let $a\in(\tfrac12,1)$, $\pi\in(0,1)$, $\lambda>0$, $d\ge3$, $\rho\in[0,1)$ and $\eta>0$. Put $\sigma_\eta^2:=1+\pi(1-\pi)\eta^2$, $\tilde\delta:=\sigma_\eta^{-1}\eta\,(e_1-e_2+e_3)$ and
\[\tilde\Sigma:=\sigma_\eta^{-2}\bigl(e_1e_1^\top+e_2e_2^\top+\rho(e_1e_2^\top+e_2e_1^\top)+e_3e_3^\top\bigr)+\textstyle\sum_{k\ge4}e_ke_k^\top ,\]
and let $P_R=N(-\pi\tilde\delta,\tilde\Sigma)$ and $P_G=N((1-\pi)\tilde\delta,\tilde\Sigma)$. Then $\tilde\Sigma\succ0$ and the mean difference is $\tilde\delta$. The pooled mean is $0$. The pooled covariance $\tilde\Sigma+\pi(1-\pi)\tilde\delta\tilde\delta^\top$ has diagonal $\sigma_\eta^{-2}(1+\pi(1-\pi)\eta^2)=1$ in coordinates 1 to 3 and $1$ elsewhere, so the pair is standardized. The vectors $e_\pm:=(e_1\pm e_2)/\sqrt2$ and $e_3$ are eigenvectors of $\tilde\Sigma$ with eigenvalues $\nu_\pm=\sigma_\eta^{-2}(1\pm\rho)$ and $\nu_3=\sigma_\eta^{-2}$, and $\tilde\delta=\sigma_\eta^{-1}\eta(\sqrt2\,e_-+e_3)$. Hence
\[D_M^2=\frac{2\eta^2\sigma_\eta^{-2}}{\nu_-}+\frac{\eta^2\sigma_\eta^{-2}}{\nu_3}=\eta^2\Bigl(1+\frac{2}{1-\rho}\Bigr).\]
By (iii), $w^*\propto x:=(\tilde\Sigma+cI)^{-1}\tilde\delta$ with $c:=\lambda/\alpha\ge4\lambda$, so $D_\lambda^2=(x^\top\tilde\delta)^2/(x^\top\tilde\Sigma x)$. In the eigenbasis,
\[x^\top\tilde\delta=\frac{2\eta^2\sigma_\eta^{-2}}{\nu_-+c}+\frac{\eta^2\sigma_\eta^{-2}}{\nu_3+c}\le\frac{3\eta^2\sigma_\eta^{-2}}{c},\qquad x^\top\tilde\Sigma x\ge\frac{\nu_3\,\eta^2\sigma_\eta^{-2}}{(\nu_3+c)^2}.\]
Therefore, using $\nu_3=\sigma_\eta^{-2}\le1$,
\[D_\lambda^2\le\frac{9\eta^2(\nu_3+c)^2}{c^2}=9\eta^2\Bigl(1+\frac{\nu_3}{c}\Bigr)^2\le\eta^2B_\lambda^2,\qquad B_\lambda:=3\Bigl(1+\frac1{4\lambda}\Bigr),\]
uniformly in $\rho$ ($B_{1/64}=51$). Choose $\eta:=\sqrt2\,\Phi^{-1}(a)/B_\lambda$; $\sigma_\eta$ does not depend on $\rho$. Then for every $\rho$ the population audit value is $\Phi(D_\lambda/\sqrt2)\le\Phi(\eta B_\lambda/\sqrt2)=a$, and it exceeds $\tfrac12$ by (iii) because $\tilde\delta\ne0$. As $\rho\uparrow1$, $D_M\to\infty$ and $\mathrm{TV}=2\Phi(D_M/2)-1\to1$.

(v) By the block-inverse formula, $(\Sigma^{-1})_{VV}=\Sigma_{V\mid U}^{-1}$, so $\delta^\top\Sigma^{-1}\delta=\delta_V^\top\Sigma_{V\mid U}^{-1}\delta_V$ when $\delta_U=0$. Since $\Sigma_{VU}\Sigma_{UU}^{-1}\Sigma_{UV}\succeq0$, we have $0\prec\Sigma_{V\mid U}\preceq\Sigma_{VV}$, and inversion reverses the Loewner order on positive definite matrices. $\square$

\paragraph{Proof of (a)(ii).} Let $a\in[\tfrac12,1)$, $\pi\in(0,1)$, $\lambda>0$ and $d\ge2$, and put $q:=\Phi^{-1}(a)\ge0$, $s_0^2:=1/\{1+2\pi(1-\pi)q^2\}$ and $\eta:=\sqrt2\,q\,s_0$. Within each group let the coordinates be independent, with:
$f_1\sim N(-\pi\eta,s_0^2)$ in the real group and $N((1-\pi)\eta,s_0^2)$ in the generated group;
$f_2$ uniform on $\{-h_R,h_R\}$ in the real group and on $\{-h_G,h_G\}$ in the generated group, where $h_R:=2^{-1/2}$ and $h_G:=\{(1+\pi)/(2\pi)\}^{1/2}$;
$f_k\sim N(0,1)$ in both groups for $k\ge3$.
All pooled means are $0$. The pooled variances are $s_0^2+\pi(1-\pi)\eta^2=s_0^2\{1+2\pi(1-\pi)q^2\}=1$, $(1-\pi)h_R^2+\pi h_G^2=\tfrac{1-\pi}2+\tfrac{1+\pi}2=1$, and $1$. So the pair is standardized. Since $h_G^2-h_R^2=1/(2\pi)>0$, the set $\mathbb R\times\{-h_R,h_R\}\times\mathbb R^{d-2}$ has probability $1$ under $P_R$ and $0$ under $P_G$, so the laws are mutually singular. Both have finite second moments, and Lemma 2 gives a unique minimizer. Within each group $(f_2,\dots,f_d)$ is independent of $f_1$ and has mean $0$. So by Lemma 3, $w^*_k=0$ for $k\ge2$, and $(b^*,w^*_1)$ minimizes the risk of $f_1$ alone, a Gaussian pair with variance $s_0^2$ and mean difference $\eta$. If $q>0$, (c)(iii) with $d=1$ gives $w^*_1=\kappa\eta/(\alpha s_0^2+\lambda)>0$, and (c)(i) gives $\mathrm{AUC}(s^*)=\Phi(\eta/(\sqrt2 s_0))=\Phi(q)=a$. If $q=0$, then $\eta=0$, $w^*_1=0$ by (b)(i), and $\mathrm{AUC}(s^*)=\tfrac12=a$. $\square$

\paragraph{Proof of (d).} Line numbers refer to \texttt{icare\_dt/generate\_semisynth.py} (G) and \texttt{icare\_dt/audit\_semisynth\_realism.py} (A) at 82587202. By (A4), the factual path they describe is the one that produced the audited dataset. Fix an audited stay, and write $\mathrm{fl}(x)$ for the single-precision rounding of $x$.
\begin{enumerate}
\item \emph{Window.} The generator truncates the real frame to its first $\min(T,336)$ rows (G l.~2956--2958; \texttt{max\_len}$=336$, head window) and generates exactly these rows. The audit reads the first 336 rows of both frames (A l.~88). For the generated frame these are all its rows, and for the real frame they are the same $\min(T,336)$ rows. By (A4$'$)(1), the real frame is the one the generator read.
\item \emph{Other columns.} Missing cells of the real frame are filled with channel means (G l.~3000). \texttt{simulate\_patient} copies this frame (G l.~5078) and overwrites only the 12 outcome columns and the 5 treatment columns (G l.~5330--5333). It then maps $\pm\infty$ to missing and fills missing cells: forward and backward in non-treatment columns, and with $0$ in treatment columns (G l.~5335--5341). By (A4$'$)(2) no real cell is infinite. So every cell that is finite in the real frame keeps its real value in the 144 columns that are neither outcomes nor treatments.
\item \emph{Treatments.} Under \texttt{observed\_dose}, the real dose is converted to single precision (G l.~2989), declared on when that value is finite and exceeds the off level plus $10^{-9}$ (G l.~2993; off level $0$), and replayed as the rounded dose when on and as the off level otherwise (G l.~2996, 5210), in a single-precision array (G l.~5182). By (A4$'$)(3), a finite real dose $x\ge0$ therefore becomes $\mathrm{fl}(x)$ when $\mathrm{fl}(x)>10^{-9}$ and $0$ otherwise. A missing real dose becomes $0$ until step 5.
\item \emph{Written outcomes.} The written values are finite, being clipped to fixed bounds (G l.~5255, 5283). With the hold on, a written cell takes a new value only in hours in which the standardized real value changed (G l.~5275). This fixes the values of the written channels and does not affect their missingness.
\item \emph{Missingness.} With \texttt{measurement\_base\_rate}$=1$ and the other measurement parameters $0$, \texttt{\_apply\_measurement\_process} returns the frame unchanged (G l.~4806--4829; called at l.~3095). \texttt{restore\_observed\_missingness} (G l.~4373--4406, called at l.~3097) then makes every column missing exactly where the real frame is missing, including the treatment columns under \texttt{preserve\_treatment\_missingness}, and leaves the other cells unchanged.
\end{enumerate}
Hence, on the window: the 144 other columns equal the real ones cell by cell; the treatment columns are missing exactly where the real ones are and elsewhere carry the values of step 3; and each written column is finite exactly where the real column is finite (steps 4 and 5 with (A4$'$)(2)). After the treatment columns are replaced by the real ones, the generated frame equals the real frame on all 149 unwritten columns and has the real finiteness pattern on the 12 written columns. Each channel's four summaries depend only on that channel's column on the window (A l.~90--96), and its missing fraction depends only on the finiteness pattern. Imputation and standardization act coordinatewise with the same constants in both groups (A1), and $\tilde f_G$ uses these constants. So $\tilde f_{G,j}=f_{R,j}$ for every $j\in U$ in every pair.

(i) For a uniformly drawn audited stay, the coupling $(f_R,\tilde f_G)$ has $\tilde f_{G,U}=f_{R,U}$ surely. So the marginal laws on $U$ coincide and $\tilde\delta_U=0$. If $s$ depends on $f$ only through $f_U$, then $s(\tilde f_G)$ and $s(f_R)$ have the same law, and Lemma 1 gives $\mathrm{AUC}(s)=\mathbb EM_R(S_G)=\mathbb EM_R(S_R)=\tfrac12$. Disintegrating both laws with respect to the common marginal law of $f_U$ shows that they can differ only through the conditional law of $f_V$ given $f_U$.

(ii) The gradient identity is Lemma 2 at $\theta_0$, applied to $(P_R,\tilde P_G)$; it is supported on $V$ because $\tilde\delta_U=0$. If $\tilde\delta_V\neq0$, the means differ and so do the laws. For $\lambda>0$, Lemma 2 gives existence and uniqueness, and (b)(iii) gives $\tilde\delta^\top w^*=\tilde\delta_V^\top w^*_V>0$. In the Gaussian model of (c), (c)(iii) gives $\mathrm{AUC}(s^*)>\tfrac12$ and (c)(v) gives $D_M$. The number of finite consecutive pairs is the same in both members of a pair, so the definedness of a written channel's lag-1 autocorrelation can differ only through the condition that both lagged series have standard deviation at least $10^{-8}$. This affects only coordinates in $V$. The last sentence of (ii) is (b)(vi).

(iii) By steps 3 and 5, the generated frame and the frame with real treatment columns differ only in finite cells of the 5 treatment columns, by the rounding of step 3. Since the constants are shared, $f_G$ and $\tilde f_G$ differ only in the 20 coordinates of the treatment channels.

The closing sentences of (d) restate (a) and the scope (A6). $\square$

\emph{Verification on shipped summaries.} On the passing 3,000-patient probe (\texttt{g2\_hold\_probe\_ptm\_a434a5c\_20260918\_v1}, 400 pairs):
\begin{itemize}
\item the 144 other channels agree exactly on all four summaries;
\item the treatment summaries agree to at most $1.1\times10^{-5}$ absolute ($5.1\times10^{-8}$ relative), with FiO$_2$, norepinephrine and phenylephrine exactly equal and missingness identical;
\item the written channels' missing fractions agree exactly;
\item the definedness of the written lag-1 autocorrelation differs in 21 of 4,800 cells (creatinine 15, bilirubin 5, lactate 1);
\item the linear audit restricted to the coordinates in $U$ has out-of-fold AUC $0.5000$.
\end{itemize}
On the full dataset's audit summary (\texttt{g2\_realism\_full\_root\_a434a5c\_20260918\_v1}), the missing-fraction difference is $0$ on all 161 channels. Sixteen channels have a nonzero standardized mean difference: the 12 written channels (absolute values $0.025$ to $0.217$) and 4 treatment channels at the rounding scale (at most $1.3\times10^{-9}$; FiO$_2$ exactly $0$). The other 144 channels have difference $0$, except ventilator mode, whose difference is undefined because it has no finite value in any audited window.

\paragraph{Numerical checks.} The following are descriptive re-reads of the shipped audit outputs and of the probe's per-patient summaries (the full dataset ships only aggregate summaries, so nothing below can be checked at the value $0.572$ itself). Parts (a) to (c) were checked on the constructions of the proof: the attainment example of (a), the constructions of (a)(i), (a)(ii) and (b)(vi), and the three-coordinate construction of (c)(iv) reproduce their stated values exactly in population computations. For example, at $\lambda=1/64$, $\pi=\tfrac12$ and $a=0.75$, the construction of (c)(iv) with $\rho=0.9999$ has population audit value $0.673$ with $D_M=2.65$ and $\mathrm{TV}=0.814$, and with $\rho=1-10^{-8}$ it has $0.678$ with $D_M=264.5$ and $\mathrm{TV}=1.000$. Table~\ref{tab:P3calib} gives the calibration of (a) and (c)(iv) for the audit values of Fig.~4b of the main text.
\begin{itemize}
\item \emph{Null band and reproduction.} Within-pair label swaps on the probe's features give an out-of-fold AUC of $0.5011\pm0.0088$ (30 swaps, maximum 0.520) and $0.5005\pm0.0087$ (60 swaps). The band belongs to the probe's features, so its use for the full-dataset value is a proxy. In stored row order the audit pipeline reproduces $0.630$ against the shipped $0.641$, and 30 random pair-to-fold assignments give $0.630$ to $0.646$. The probe and the full dataset should not be treated as one audit population: their values differ by $0.069$, which is 3.8 times $\sqrt2$ times the pair-bootstrap standard deviation ($0.0129$), and the root sum of squared standardized mean differences over the 12 written-channel means is $0.62$ on the probe and $0.40$ on the full dataset.
\item \emph{Efficiency of the ridge direction.} On the probe, a smaller penalty ($C=0.01$; out-of-fold AUC $0.647$) and shrinkage linear discriminant analysis ($0.645$) exceed the audit's linear classifier on the same folds ($0.630$), and a debiased full-covariance estimate of $D_M$ on the coordinates in $V$ is $0.563$, against the $0.510$ implied by $0.641$. This is consistent with (c)(iv), which reads the audit value as a lower bound. On the full dataset the root sum of squared standardized mean differences over the 12 written-channel means is $0.404$; it equals $D_M$ only if those means were uncorrelated, so it is neither a bound nor a calibration.
\item \emph{A flexible classifier separates more.} Histogram gradient boosting (scikit-learn defaults, 200 iterations, the same person-grouped folds; one configuration chosen after the audit) reaches out-of-fold AUC $0.735$ on all 644 probe features, $0.730$ after the dose rounding is removed by copying the real treatment values, and $0.696$ on the 36 coordinates in $V$ alone (linear audit on $V$: $0.640$; label-swap null for boosting on $V$: $0.489\pm0.014$ over 12 swaps). By (a), a fitted score whose population AUC is near $0.73$ would certify $\mathrm{TV}\ge0.26$ for the probe's laws, against $0.152$ from $0.641$. This shows that passing the audit does not certify closeness in distribution. It is not a new audit value.
\item \emph{Scope of (d).} Boosting reaches $0.570$ on the 20 treatment features alone, which (d)(iii) permits through the dose rounding, and exactly $0.5000$ on the coordinates in $U$ once the real treatment values are copied in, as (d)(i) predicts; the linear audit gives $0.5000$ on $U$ either way.
\item \emph{What the linear signal is.} On the probe the summed squared standardized mean difference over $V$ is $0.382$ for the 12 means, $0.041$ for the within-stay standard deviations and $0.011$ for the lag-1 autocorrelations, and the linear audit on the 12 means alone gives $0.629$ against $0.640$ on all 36 coordinates. Paired $t$ statistics (generated minus real) reach $18.5$ for the respiratory-rate mean; among within-stay standard deviations the largest are \spo{} ($-10.6$), platelets ($+4.8$), temperature ($+4.0$) and invasive mean arterial pressure ($+3.1$); lag-1 differences are smaller (largest $|t|=2.94$, temperature). Hotelling's $T^2$ on the paired differences in $V$ gives $F(36,364)=10.7$. On the full dataset the within-stay standard deviation of the generated records is $33\%$ lower for lactate and $12\%$ higher for creatinine; all other written channels change by at most $5\%$, and lag-1 autocorrelations by at most $0.015$.
\end{itemize}

\begin{table}[ht]
\centering\small
\caption{\textbf{Calibration of realism-audit values (Proposition 3).} For an audit value $a$, the distribution-free lower bound on the total-variation distance between the real and generated laws of the 644 summaries, $1-(1-2|a-\tfrac12|)^{1/2}$ (Proposition 3(a)), and the lower bounds implied under an equal-covariance Gaussian idealization, $D_M\ge\sqrt2\,\Phi^{-1}(a)$ and $\mathrm{TV}\ge2\Phi(\Phi^{-1}(a)/\sqrt2)-1$ (Proposition 3(c)(iv)); the Gaussian bounds are attained only if the audit's penalized direction is efficient. All lower bounds are rounded down. No audit value gives an upper bound on the distance (Proposition 3(a)(i), (a)(ii) and (c)(iv)). Audit values of the generated dataset are those of Fig.~4b of the main text; probe values use 3,000-patient probes. The audit values are sample AUCs with sampling error (pair-bootstrap standard deviation about 0.013 on the probe), and the bounds are evaluated at them as if they were population values.}
\label{tab:P3calib}
\begin{tabular}{lrrrr}
\toprule
Generator version & Audit AUC & TV $\ge$ (any law) & $D_M\ge$ (Gaussian) & TV $\ge$ (Gaussian)\\
\midrule
No correction (probe) & 0.963 & 0.727 & 2.525 & 0.793\\
Measurement hold only (probe) & 0.956 & 0.704 & 2.417 & 0.773\\
Treatment missingness only (probe) & 0.767 & 0.317 & 1.029 & 0.393\\
Both corrections (probe) & 0.641 & 0.152 & 0.510 & 0.201\\
Both corrections (full dataset) & 0.572 & 0.074 & 0.255 & 0.101\\
Tolerance & 0.750 & 0.292 & 0.953 & 0.366\\
\bottomrule
\end{tabular}
\end{table}

%% file: supplementary/supp_theory/p4.tex
\paragraph{Proposition 4 (Recoverability: the target is trait recovery, amplitude enters through an exact law, and a held record keeps only what its draws see).}
\emph{Standard and specific parts.} Parts (a)(i) to (a)(iii) are properties of the generator code, (a)(iv) restates the definition of $R^2_\ast(\sigma(\zeta_i))$ for this target, and (e) follows at once from design exogeneity. Parts (b) to (d), the exact amplitude scaling of the record, the amplitude law of readers affine in the record and the retention of a held record, are specific to the washout panels.

\smallskip\noindent\emph{Setting.} A patient $i$ is an independent draw of the source tape, the generator's person-level random numbers (propensity parameters, factual and counterfactual noise, dose-harvest and top-up draws) and the trait vector $(\zeta_{ij})_{j=1}^{5}$. $\mathrm E$, $\mathrm{Var}$, $\mathrm{Cov}$, $\mathrm{corr}$ and $\mathrm{SD}$ are moments under this law. Rows $t=0,\dots,89$ precede the decision row $t_0=90$, and $h=12$. The scored treatment $j^\ast$ has binary doses $A_i(t)\in\{0,1\}$, with $A_i(t):=0$ for $t<0$. The generator's memory is $e_i(t)=\lambda_0e_i(t-1)+A_i(t-2)$ with $e_i(0)=0$ and $\lambda_0=0.8$, and its occupancy is $q_i(t):=\varphi_1(e_i(t))$, where $\varphi_\kappa(e):=e/(\kappa+e)$. Write $\zeta_i:=\zeta_{ij^\ast}$ (mean $\mu_\zeta$, variance $\sigma_\zeta^2>0$) and $\Lambda_k:=L_{kj^\ast}$. Let $k^\ast$ be the scored channel: heart rate in the registered phenylephrine cell; the benchmark also scores heart rate for norepinephrine and creatinine for NaCl. On channel $k$ at row $t$ the generator outputs, in standardized units,
\[
y_{ik}(t)=\tilde y_{ik}(t)+c\,\alpha_{ik}(t)\tanh\Gamma_{ik}(t),\qquad \Gamma_{ik}(t):=\textstyle\sum_{j=1}^{5}L_{kj}\,\zeta_{ij}\,q_{ij}(t).
\]
Here $\tilde y_{ik}(t)$ is the value after every clip that precedes the memory route, $q_{ij}$ is the occupancy of treatment $j$, $\alpha_{ik}(t)\in[0,1]$ is the attenuation near a physiological bound and $c>0$ is the amplitude cap.

The draw set is $\mathcal D_{ik}:=\{0\}\cup\{1\le t\le 89:\ y^0_{ik}(t)\neq y^0_{ik}(t-1)\}$, where $y^0$ is the standardized clipped source tape. Its elements are ordered $d_0=0<d_1<\dots<d_n$ with $n=n_{ik}$, and $d_{ik}(t):=\max\{d\in\mathcal D_{ik}:d\le t\}$. The held record is $Z_{ik}(t):=y_{ik}(d_{ik}(t))$ for $t\le 89$, and the hourly record is the case $\mathcal D_{ik}=\{0,\dots,89\}$. The design is $\mathcal S_i:=(A_i(0),\dots,A_i(89),\mathcal D_{i1},\dots,\mathcal D_{i,12})$. The record is $\mathcal R_i:=(Z_i,\mathcal S_i)$ with $Z_i:=(Z_{ik}(t))_{k\le 12,\,t\le 89}$.

In the counterfactual rollout from $t_0$ under regime $a\in\{0,1\}$, $j^\ast$ is fixed at $a$ at every step and the other treatments are drawn by the generator's policy with common random numbers. Let $e^a_{is,j}$ be the memory of treatment $j$ read by the outcome at row $t_0+s+1$, $q^a_{is,j}:=\varphi_1(e^a_{is,j})$ and $\Gamma^a_{is}:=\sum_jL_{k^\ast j}\zeta_{ij}q^a_{is,j}$. The target $\tau_i$ is the frame-by-regime interaction of Definition 4 on the latent path of channel $k^\ast$. We assume throughout that $\mathrm{Var}(T_i)>0$, with $T_i$ as in (a)(i).

For a predictor $\hat\tau$, $R^2(\hat\tau):=1-\mathrm E(\tau-\hat\tau)^2/\mathrm{Var}(\tau)$. The population $R^2$ of a feature vector is the $R^2$ of its population least-squares predictor with an intercept. $R^2_\ast(\mathcal G):=\mathrm{Var}\,\mathrm E[\tau\mid\mathcal G]/\mathrm{Var}(\tau)$ is the information content of $\mathcal G$.

A reader with kinetics $K=(\ell,\lambda,\kappa)$ uses the exposure profile $p_K(t):=\varphi_\kappa(e_K(t))$, where $e_K(t)=\lambda e_K(t-1)+A(t-\ell)$ and $e_K(0)=0$. At the generator's kinetics $K_0=(2,0.8,1)$, $p_{K_0}=q_i$ on rows $0,\dots,89$. Let $\mathbb A$ be the set of amplitudes $c>0$ at which (A1) and (R) hold.

The assumptions (G1) to (G4), (A1), (R), (K), (P), (H), (E), (L) and (RW) are listed separately, and each part names the ones it uses.

\medskip\noindent\textbf{(a) Target structure} [(G1), (G2), (A1) at the amplitude $c$].
\begin{enumerate}
\item[(i)] $\tau_i=c\,T_i$ with $T_i:=h^{-1}\sum_{s=0}^{h-1}\bigl[\tanh\Gamma^1_{is}-\tanh\Gamma^0_{is}\bigr]$. $T_i$ does not depend on $c$, and its $s=0$ summand is $0$.
\item[(ii)] $e^a_{is,j^\ast}=\lambda_0^{s+1}e_i(t_0)+a\,(1-\lambda_0^{s})/(1-\lambda_0)$. Hence $\tau_i=\tau^{\rm own}(\zeta_i,e_i(t_0))+\nu_i$, where
\[
\tau^{\rm own}(\zeta,e):=\frac ch\sum_{s=0}^{h-1}\Bigl[\tanh\bigl(\Lambda_{k^\ast}\zeta\,q^1_s(e)\bigr)-\tanh\bigl(\Lambda_{k^\ast}\zeta\,q^0_s(e)\bigr)\Bigr],\qquad q^a_s(e):=\varphi_1\Bigl(\lambda_0^{s+1}e+a\,\frac{1-\lambda_0^{s}}{1-\lambda_0}\Bigr).
\]
The map $\tau^{\rm own}$ is shared by all patients and dosing schedules. The remainder $\nu_i:=\tau_i-\tau^{\rm own}(\zeta_i,e_i(t_0))$ is $0$ whenever the counterfactual policy, under either regime, gives no treatment $j\neq j^\ast$ with $L_{k^\ast j}\neq 0$ at steps $0,\dots,h-2$.
\item[(iii)] Let $D(e):=h^{-1}\sum_{s=0}^{h-1}[q^1_s(e)-q^0_s(e)]$ be the mean own-occupancy increment. Then $D(0)=D^\ast:=h^{-1}\sum_{s=1}^{h-1}\varphi_1\bigl((1-\lambda_0^{s})/(1-\lambda_0)\bigr)=0.6842$ (to four decimals), and for every $e\ge 0$
\[
0\ \le\ D^\ast-D(e)\ \le\ \frac{\lambda_0(1-\lambda_0^{h})}{h(1-\lambda_0)}\,e\ \le\ 0.3105\,e .
\]
\item[(iv)] Among predictors that are functions of $\zeta_i$ alone, $\bar\tau(\zeta_i):=\mathrm E[\tau_i\mid\zeta_i]$ has the smallest mean squared error. Its $R^2$ is $R^2_{\rm trait}:=R^2(\bar\tau)=1-\mathrm E\,\mathrm{Var}(\tau\mid\zeta)/\mathrm{Var}(\tau)$, which is $R^2_\ast(\sigma(\zeta_i))$ by definition, and its recovery is $1-(1-R^2_{\rm trait})^{1/2}$. This bounds only $\sigma(\zeta_i)$-measurable predictors. The record $\mathcal R_i$ is not a function of $\zeta_i$, so for predictors built on the record $R^2_{\rm trait}$ is a reference level.
\end{enumerate}

\medskip\noindent\textbf{(b) Exact scaling of the record} [(G1), (G2), (G4), (A1)]. For every amplitude $c$ at which (A1) holds, every $k\le 12$ and every $t\le 89$,
\[
Z_{ik}(t;c)=Z^0_{ik}(t)+c\,U_{ik}(t),\qquad Z^0_{ik}(t):=\tilde y_{ik}(d_{ik}(t)),\qquad U_{ik}(t):=\tanh\bigl(\Lambda_k\zeta_i\,q_i(d_{ik}(t))\bigr),
\]
where $Z^0_i$ is the held record of the cap-0 run. $Z^0_i$, $U_i$, $T_i$ and $\mathcal S_i$ do not depend on $c$.

\medskip\noindent\textbf{(c) Amplitude law} [(b), (R), (K), (H)]. Fix the reader's kinetics $K$. Let its features be $(X_i(c),X^{\mathcal S}_i)$. Here $X_i(c)=a_{\mathcal S_i}+\Phi_{\mathcal S_i}Z_i(c)\in\mathbb R^p$, where the vector $a_{\mathcal S}$ and the matrix $\Phi_{\mathcal S}$ depend only on $K$ and on the design through its true draw sets; that is, the reader is affine in the record for fixed design. $X^{\mathcal S}_i=x^{\mathcal S}(\mathcal S_i)$ collects the reader's design-only features and a constant, with $X^{\mathcal S}_i=1$ if it has none.

For square-integrable $V$ let $\tilde V:=V-\mathrm{Proj}(V\mid X^{\mathcal S})$ be the residual of the population linear projection on $\mathrm{span}(X^{\mathcal S})$. Put $X^0:=a_{\mathcal S}+\Phi_{\mathcal S}Z^0$, $U_X:=\Phi_{\mathcal S}U$, $\Sigma_0:=\mathrm{Var}(\tilde X^0)$, $\Sigma_U:=\mathrm{Var}(\tilde U_X)$ and $\boldsymbol\theta:=\mathrm{Cov}(\tilde U_X,T)$. Let $R^2_{\mathcal S}$ be the population $R^2$ of $T$ on $X^{\mathcal S}$, and define
\[
C_0:=\frac{\mathrm{Var}(T)}{\boldsymbol\theta^\top\Sigma_U^{+}\boldsymbol\theta},\qquad C_1:=\frac{\mathrm{Var}(T)}{\boldsymbol\theta^\top\Sigma_0^{-1}\boldsymbol\theta},\qquad \mathcal L(c):=\frac{\boldsymbol\theta^\top(\Sigma_0/c^2+\Sigma_U)^{-1}\boldsymbol\theta}{\mathrm{Var}(T)}\quad(c>0),
\]
where $\Sigma_U^{+}$ is the Moore--Penrose inverse. These objects are built from $Z^0$, $U$, $T$ and the design alone, so none of them depends on $c$. The law $\mathcal L$ is defined at every $c>0$, whether or not $c\in\mathbb A$.
\begin{enumerate}
\item[(i)] For $c\in\mathbb A$, $\tilde X(c)=\tilde X^0+c\,\tilde U_X$, and the population $R^2$ of $\tau$ on $(X(c),X^{\mathcal S})$ is $R^2(c)=R^2_{\mathcal S}+\mathcal L(c)$. $R^2_{\mathcal S}$ does not depend on $c$. Every $\mathcal R$-measurable predictor, linear or not, has $R^2\le R^2_\ast(\sigma(\mathcal R))$; in particular $R^2(c)\le R^2_\ast(\sigma(\mathcal R))$.
\item[(ii)] Properties of the law. $1/C_0\le 1$ is the population $R^2$ of $T$ on $\tilde U_X$, and $C_1>0$. The function $H(x):=1/\mathcal L(x^{-1/2})$ is concave and strictly increasing on $(0,\infty)$, and
\[
H(x)\ \ge\ C_0+C_1x\quad(x>0),\qquad \lim_{x\downarrow 0}H(x)=C_0,\qquad \lim_{x\to\infty}H(x)/x=C_1 .
\]
Equivalently, $1/\mathcal L(c)\ge C_0+C_1/c^2$, $\mathcal L(c)\to 1/C_0$ as $c\to\infty$, and $\mathcal L(c)/c^2\to 1/C_1$ as $c\to 0$. Equality $H(x)=C_0+C_1x$ holds at one $x>0$, equivalently at every $x>0$, iff some $v\neq 0$ has both $\Sigma_0v$ and $\Sigma_Uv$ in $\mathrm{span}\{\boldsymbol\theta\}$. Such a $v$ is a nonzero multiple of $\Sigma_0^{-1}\boldsymbol\theta$, and the population least-squares coefficient on $\tilde X(c)$ is then proportional to $v$ at every amplitude; we say that one statistic carries the signal. In that case
\[
C_0=1/\mathrm{corr}^2(v^\top\tilde U_X,T),\qquad C_1=\mathrm{Var}(T)\,\mathrm{Var}(v^\top\tilde X^0)/\mathrm{Cov}(v^\top\tilde U_X,T)^2 .
\]
One statistic always carries the signal when $p=1$.
\item[(iii)] Consequences for the reader.
\begin{itemize}
\item For every $c\in\mathbb A$, $R^2(c)-R^2_{\mathcal S}=\mathcal L(c)<1/C_0$. The ceiling $1/C_0=\sup_{c>0}\mathcal L(c)$ is not attained. The limit $c\to\infty$ is a property of the law, which describes the reader only on $\mathbb A$.
\item The bar. When $R^2_{\mathcal S}=0$, the bar is a gain of $0.2$ (recovery $1-\sqrt{0.8}$). If $R^2(c)-R^2_{\mathcal S}\ge 0.2$ at some $c\in\mathbb A$, then $C_0<5$ and $c\ge c^\ast:=\bigl(C_1/(5-C_0)\bigr)^{1/2}$. When one statistic carries the signal, $\mathcal L(c^\ast)=0.2$, so $c^\ast$ is the reader's exact crossing amplitude if $c^\ast\in\mathbb A$. If $1/C_0\le 0.2$, the gain is below $0.2$ at every $c\in\mathbb A$. For the total $R^2$, replace $0.2$ by $0.2-R^2_{\mathcal S}$.
\item Low signal. If $(0,\bar c]\subseteq\mathbb A$ for some $\bar c>0$, then $(R^2(c)-R^2_{\mathcal S})/c^2\to 1/C_1$ as $c\downarrow 0$.
\item Change of sampling. Take two samplings of the same target, read at the same kinetics and each satisfying (H), with laws $\mathcal L_{\rm old}$, $\mathcal L_{\rm new}$ and constants $C_1^{\rm old}$, $C_1^{\rm new}$. For $\varrho\in(0,\min(1/C_0^{\rm old},1/C_0^{\rm new}))$, let $c_{\rm old}(\varrho)$ and $c_{\rm new}(\varrho)$ be the amplitudes at which the two laws equal $\varrho$. Then $c_{\rm new}(\varrho)/c_{\rm old}(\varrho)\to(C_1^{\rm new}/C_1^{\rm old})^{1/2}$ as $\varrho\downarrow 0$. If both sets $\mathbb A$ contain an interval $(0,\bar c]$, then for all small $\varrho$ these are the amplitudes at which the readers' gains equal $\varrho$.
\end{itemize}
\item[(iv)] Ceiling in the linear regime [(L), (G3), (E)]. Let $p=1$ and $U_X=\Lambda_k\zeta\,\Xi$ for a design-only scalar $\Xi$ with $\mathrm E\,\Xi\neq 0$. Write $T=k_T\zeta+\varepsilon$ with $k_T:=\mathrm{Cov}(T,\zeta)/\sigma_\zeta^2\neq 0$, and let $\mathrm{Cov}(\tilde U_X,\varepsilon)=0$ (assumption (E)). Put $\mathrm{CV}(\Xi):=\mathrm{SD}(\Xi)/|\mathrm E\,\Xi|$, $\tilde\Xi:=\Xi-\mathrm{Proj}(\Xi\mid X^{\mathcal S})$ and $\rho^2_{\rm lin}:=\mathrm{corr}^2(T,\zeta)=k_T^2\sigma_\zeta^2/\mathrm{Var}(T)$. Then
\[
C_0=\frac{1+\mathrm{CV}^2(\Xi)+(\mu_\zeta^2/\sigma_\zeta^2)\,\mathrm{Var}(\tilde\Xi)/(\mathrm E\,\Xi)^2}{\rho^2_{\rm lin}} .
\]
Two single-channel readers of channel $k$ are special cases; sums over $j$ run over that channel's draws.
\begin{itemize}
\item Cadence-aware reader at kinetics $K$: statistic $\mathcal C_{ik}(K):=\sum_j\Delta_jZ_{ik}\,\Delta_jp_K$, design-only feature $Q^{(0)}_{ik}(K):=\sum_j(\Delta_jp_K)^2$, and $\Xi=\Xi_{ik}(K):=\sum_j\Delta_jp_K\,\Delta_jq_i$. At $K=K_0$, $\Xi=Q^{(0)}_{ik}$ is itself a feature, so $\tilde\Xi=0$ and $C_0=(1+\mathrm{CV}^2(Q^{(0)}_{ik}))/\rho^2_{\rm lin}$. At other $K$ the general formula applies with $\tilde\Xi$ the residual of $\Xi_{ik}(K)$ on $(1,Q^{(0)}_{ik}(K))$; then $\mathrm{Var}(\tilde\Xi)>0$ unless $\Xi_{ik}(K)$ is almost surely an affine function of $Q^{(0)}_{ik}(K)$.
\item Plain reader at kinetics $K$: statistic $\mathcal O_{ik}(K):=\sum_{t=0}^{89}(\Pi p_K)(t)Z_{ik}(t)$, no design-only feature ($X^{\mathcal S}=1$), and $\Xi=M_{ik}(K):=\sum_t(\Pi p_K)(t)\,q_i(d_{ik}(t))$. At every $K$, $C_0=\bigl(1+(1+\mu_\zeta^2/\sigma_\zeta^2)\,\mathrm{CV}^2(M_{ik}(K))\bigr)/\rho^2_{\rm lin}$, and $M_{ik}(K_0)=M_{ik}$.
\end{itemize}
\item[(v)] Two-point constants [(ii)]. For $0<c_2<c_1$ put $\varrho_\ell:=\mathcal L(c_\ell)$. Let $\hat C_0+\hat C_1/c^2$ be the line in $1/c^2$ through the points $(c_1^{-2},1/\varrho_1)$ and $(c_2^{-2},1/\varrho_2)$. Then:
\begin{itemize}
\item $\hat C_0\ge C_0$ and $\hat C_1\ge C_1$;
\item the two-point prediction $1/(\hat C_0+\hat C_1/c^2)$ is at least $\mathcal L(c)$ for $c\in[c_2,c_1]$ and at most $\mathcal L(c)$ for $c\notin[c_2,c_1]$;
\item all of these relations hold with equality when one statistic carries the signal.
\end{itemize}
So the two-point ceiling $1/\hat C_0$ is at most the population ceiling $1/C_0$, and the two-point low-signal constant is at least $C_1$. When $c_1,c_2\in\mathbb A$, $\varrho_1$ and $\varrho_2$ are the reader's population gains.
\end{enumerate}

\medskip\noindent\textbf{(d) What a held record retains} [(G2), (G4), (A1), (L), (RW)]. Fix a channel $k$ with $\Lambda_k\neq 0$, condition on $\mathcal S_i$ and drop the indices $i,k$. By (A1), (L) and (RW), $y(t)=m(t)+b(t)+c\Lambda_k\zeta q(t)$ for $t\le 89$. Here $m$ is a known $\mathcal S$-measurable mean path ($m\equiv 0$ is allowed) and $b$ is the random walk of (RW). For $0\le j<n$ write $\Delta_jf:=f(d_{j+1})-f(d_j)$ and $g_j:=d_{j+1}-d_j$. Put
\begin{align*}
Q^\circ&:=\textstyle\sum_{t=1}^{89}(q(t)-q(t-1))^2, & J&:=\textstyle\sum_j(\Delta_jq)^2/g_j,\\
Q^{(0)}&:=\textstyle\sum_j(\Delta_jq)^2, & Q^{(1)}&:=\textstyle\sum_jg_j(\Delta_jq)^2,\\
\mathcal E_j&:=\textstyle\sum_{t=d_j+1}^{d_{j+1}}(q(t)-q(t-1))^2, & \mathcal E_{\rm tail}&:=\textstyle\sum_{t=d_n+1}^{89}(q(t)-q(t-1))^2,\\
\mathcal C^{\rm eff}&:=\textstyle\sum_j\Delta_j(Z-m)\,\Delta_jq/g_j .
\end{align*}
\begin{enumerate}
\item[(i)] The held record $(Z(t))_{t\le 89}$ and the draw values $(y(d_j))_{j\le n}$ generate the same $\sigma$-field.
\item[(ii)] $\mathcal C^{\rm eff}$ is sufficient for $\zeta$, and $\mathcal C^{\rm eff}\sim N(c\Lambda_k\zeta J,\ \sigma_b^2J)$. The Fisher information about $\zeta$ is $(c\Lambda_k/\sigma_b)^2J$; for the hourly record it is $(c\Lambda_k/\sigma_b)^2Q^\circ$.
\item[(iii)] Let $Q^\circ>0$. For $\mathcal E_j>0$ put $w_j:=\mathcal E_j/Q^\circ$ and $\psi_j:=(\Delta_jq)^2/(g_j\mathcal E_j)$. Then $\psi_j\in[0,1]$; $\psi_j=1$ iff $q$ is affine on $\{d_j,\dots,d_{j+1}\}$; $\psi_j=0$ iff $q(d_{j+1})=q(d_j)$; and
\[
\chi:=\frac{J}{Q^\circ}=\sum_{j:\,\mathcal E_j>0}w_j\psi_j,\qquad \sum_{j:\,\mathcal E_j>0}w_j=1-\frac{\mathcal E_{\rm tail}}{Q^\circ}.
\]
So $\chi\in[0,1]$. Interval $j$ contributes $\psi_j\mathcal E_j=(\Delta_jq)^2/g_j$, which depends only on the net change of exposure between its two draws, and exposure energy after the last draw is lost.

For the generator's profile (binary doses, $\lambda_0=4/5$) more holds. For $0\le t<t'\le 89$, $q(t)=q(t')$ iff $q(t')=0$ iff $A(u)=0$ for every $u\le t'-2$. Hence $\psi_j>0$ for every $j$ with $\mathcal E_j>0$, so no interval over which the exposure moves is fully aliased. This exact statement carries little practical weight: alignment factors as small as $1.6\times10^{-9}$ on heart rate and $1.1\times10^{-13}$ on creatinine occur (numerical checks), so near-total aliasing does occur, and the retention $\chi$ is the quantity that measures it. Moreover $J=0$ iff $Q^{(0)}=0$ iff $q(d_n)=0$ iff no dose is given on rows $0,\dots,d_n-2$.
\item[(iv)] Let $J>0$. The held record at amplitude $c$ and the hourly record at amplitude $c'$ have equal Fisher information about $\zeta$ iff $c'=c\,\chi^{1/2}$, and this is also the condition under which the standardized laws of their sufficient statistics $\mathcal C^{\rm eff}$ coincide for every $\zeta$. If $J=0$, the held record carries no information about $\zeta$ at any amplitude.
\item[(v)] Pooled over patients [(E)]. Let $\mathcal S$ be random and let (RW) hold given $\mathcal S$, with the baseline independent of $(\zeta,\varepsilon)$ given $\mathcal S$. Let $T=k_T\zeta+\varepsilon$ with $k_T\neq 0$ and $\mathcal S\perp(\zeta,\varepsilon)$, and let $\mathrm EJ>0$. On hourly records $J=Q^{(0)}=Q^{(1)}=Q^\circ$.
\begin{enumerate}
\item[(1)] Efficient statistic. The reader with features $(\mathcal C^{\rm eff},J)$ satisfies (H) and has $R^2_{\mathcal S}=0$,
\[
C_0=(1+\mathrm{CV}^2(J))/\rho^2_{\rm lin},\qquad C_1=\sigma_b^2/(\Lambda_k^2\sigma_\zeta^2\rho^2_{\rm lin}\,\mathrm EJ).
\]
By (c)(iii), its low-signal amplitude factor from hourly to held records is $(\mathrm EQ^\circ/\mathrm EJ)^{1/2}$. Without the mean correction, the reader uses $\mathcal C^{\rm raw}:=\sum_j\Delta_jZ\,\Delta_jq/g_j=\mathcal C^{\rm eff}+m_q$, where $m_q:=\sum_j\Delta_jm\,\Delta_jq/g_j$. The reader with features $(\mathcal C^{\rm raw},J)$ also satisfies (H), has the same $C_0$, and has
\[
C_1=\bigl(\sigma_b^2\,\mathrm EJ+\mathrm{Var}(\tilde m_q)\bigr)/\bigl(\Lambda_k^2\sigma_\zeta^2\rho^2_{\rm lin}(\mathrm EJ)^2\bigr),
\]
where $\tilde m_q$ is the residual of $m_q$ on $(1,J)$.
\item[(2)] Information content. If moreover $\zeta\sim N(\mu_\zeta,\sigma_\zeta^2)$, the channel's record $\mathcal R^{(k)}:=((Z(t))_{t\le 89},\mathcal S)$ has trait information content $\mathrm{Var}\,\mathrm E[\zeta\mid\mathcal R^{(k)}]/\sigma_\zeta^2=\mathrm E\bigl[\varsigma J/(1+\varsigma J)\bigr]$, where $\varsigma:=c^2\Lambda_k^2\sigma_\zeta^2/\sigma_b^2$. The amplitudes at which held and hourly records reach a common level $\varrho$ satisfy $c_{\rm held}(\varrho)/c_{\rm hourly}(\varrho)\to(\mathrm EQ^\circ/\mathrm EJ)^{1/2}$ as $\varrho\downarrow 0$.
\item[(3)] Implemented cadence-aware statistic at $K_0$ [(K)]. Let $\mathcal C^m:=\sum_j\Delta_j(Z-m)\,\Delta_jq$, which is $\mathcal C_{ik}$ when $m\equiv 0$. The reader with features $(\mathcal C^m,Q^{(0)})$ satisfies (H) and has $R^2_{\mathcal S}=0$,
\[
C_0=(1+\mathrm{CV}^2(Q^{(0)}))/\rho^2_{\rm lin},\qquad C_1=\sigma_b^2\,\mathrm EQ^{(1)}/(\Lambda_k^2\sigma_\zeta^2\rho^2_{\rm lin}(\mathrm EQ^{(0)})^2).
\]
Its low-signal amplitude factor from hourly to held records is $(\mathrm EQ^\circ\,\mathrm EQ^{(1)})^{1/2}/\mathrm EQ^{(0)}\ \ge\ (\mathrm EQ^\circ/\mathrm EJ)^{1/2}$.
\end{enumerate}
\end{enumerate}

\medskip\noindent\textbf{(e) Exposure is necessary} [(G1) to (G4); (A1) is not needed]. Let $\mathcal N_i:=\{q_i(d)=0$ for every $d\in\mathcal D_{ik}$ and every $k$ with $\Lambda_k\neq 0\}$. This event is determined by $\mathcal S_i$. Under binary dosing it is the event that $A_i(u)=0$ for every $u\le\max\mathcal D_{ik}-2$ and every $k$ with $\Lambda_k\neq 0$.

On $\mathcal N_i$, $Z_i=Z^0_i$, and for every bounded measurable $f$, $\mathrm E[f(\zeta_i)\mid\mathcal R_i]=\mathrm E f(\zeta_i)$ almost surely; in particular $\mathrm E[\bar\tau(\zeta_i)\mid\mathcal R_i]=\mathrm E\,\bar\tau(\zeta_i)$ there.

Suppose $P(\mathcal N_i)=1$, as in a panel with no pre-decision dose ($A_i\equiv 0$ on rows $0,\dots,89$, so that $q_i\equiv 0$ there). Then $Z_i=Z^0_i$ for every patient, and every $\mathcal R$-measurable predictor $\hat\tau$ of that panel's target has
\[
R^2(\hat\tau)\ \le\ R^2_\ast(\sigma(\mathcal R))\ \le\ \frac{\mathrm E\,\mathrm{Var}(\tau\mid\zeta)}{\mathrm{Var}(\tau)}=1-R^2_{\rm trait},
\]
with both moments computed for that panel. The target still depends on the trait: by (a), under (A1), its own-treatment part is $\tau^{\rm own}(\zeta_i,0)$.

\paragraph{Assumptions.}
\begin{description}
\item[(G1) Shared paths and no feedback (code fact).] The memory-route shift $c\,\alpha\tanh\Gamma$ is added after every state-dependent clip. The state transition, the policy and the memory recursion do not read it, and it draws no random numbers (\texttt{response\_memory\_outcome\_shift}, \texttt{apply\_response\_memory\_bounded\_shift}, \texttt{simulate\_patient} and \texttt{simulate\_counterfactuals} in \texttt{icare\_dt/generate\_semisynth.py}). The factual runs at cap $0$ and cap $c$ use the same tape, doses and seed. Both counterfactual frames start from the cap-0 factual trace with the same seed (\texttt{simulate\_person} in \texttt{icare\_dt/revision\_v3\_g2\_washout\_occupancy\_benchmark\_v1.py}, which checks that the two factual traces agree except for the recorded outcome). Doses are harvested with the generator's policy, which reads only the latent state, and are topped up with a seed fixed by the person index. Common random numbers across regimes are implemented but the proofs do not need them.
\item[(G2) Dosing and memory (code fact).] In the factual record only $j^\ast$ is dosed, and doses are binary. NATMIN gives the policy's natural doses on rows 6 to 53 and, if fewer than 8 rows are dosed, adds dosed rows to reach 8. FIX19 doses rows 6 to 24. Neither schedule doses at or after row 54, so $A_i(89)=0$. Factual memory: $e(t)=0.8\,e(t-1)+A(t-2)$ with $e(0)=0$ and saturation constant 1. Counterfactual memory: $e^a_{i,-1,j}=e_{ij}(t_0)$ and $e^a_{is,j}=0.8\,e^a_{i,s-1,j}+A_{ij}(t_0+s-1)$. Here $A_{ij}(t_0-1)$ is the factual dose and, for $s\ge 1$, $A_{ij}(t_0+s-1)$ is the action at step $s-1$ ($a$ for $j^\ast$, the policy draw otherwise). The outcome at row $t_0+s+1$ reads $e^a_{is,\cdot}$.
\item[(G3) Trait independence (code fact plus a modeling convention).] The traits come from a separate random stream and enter only the shift. We model distinct pseudo-random streams and the tape draw as independent random variables. Then the trait vector $(\zeta_{ij})_j$ is independent of the tape, the factual noise and the dose-harvest and top-up draws, hence of $(Z^0_i,\mathcal S_i)$ (Definition 5; all 800 NATMIN dose columns were reconstructed from the shipped $u_0$). Tapes are drawn without replacement from a finite list, so treating patients as independent draws is an idealization.
\item[(G4) Hold (code fact).] At a row outside $\mathcal D_{ik}$ the record repeats the previous recorded value, and row 0 is always recorded. $\mathcal D_{ik}$ is defined by an equality test on the standardized clipped source tape, so it depends on the tape only.
\item[(A1) Attenuation inactive (a property of an amplitude $c$; used by (a) to (d)).] $\alpha_{ik}(t)=1$ wherever the shift is nonzero, both on the pre-decision record and on the $h$ counterfactual rows of channel $k^\ast$ in frame A under both regimes, and the final raw-unit clip does not bind. It holds when at least $c$ standardized units of room remain in the direction of the shift.
\begin{itemize}
\item Heart rate: in the 240-person scaling check (A1) never failed, and the shipped per-patient targets are affine in $c$ to $1.6\times10^{-7}$ for phenylephrine and $2.4\times10^{-7}$ for norepinephrine.
\item Phenylephrine's other carriers: in the same check (A1) failed for 9 persons on white cells and 6 on platelets (maximum deviation 0.084).
\item NaCl on creatinine: the shipped targets are not affine in $c$ for 56 to 62 of 400 persons (maximum residual 0.028, root mean square 0.0019 to 0.0024, against a target SD of 0.248). So (a) to (c) hold only approximately for that pair.
\end{itemize}
Statements are in exact arithmetic; the generator computes in single precision.
\item[(R) Draw detection (a property of an amplitude $c$; used by (c)).] The reader takes as draws the rows at which the recorded value changes, row 0 included, and this set equals $\mathcal D_{ik}$. Then $a_{\mathcal S}$ and $\Phi_{\mathcal S}$ depend only on the design and on $K$. Consecutive draw values differ with probability one when no state-dependent clip binds at either draw, because $\tilde y$ contains independent Gaussian outcome noise at every row. Such a clip can bind at two consecutive draws on a channel whose values sit at a physiological bound; this is plausible for SpO$_2$, a carrier of NaCl's response. There the cap-0 values coincide while the cap-$c$ values can differ, so the detected set, and with it $\Phi_{\mathcal S}$, can depend on $c$. This bears only on the NaCl numbers and was not quantified.
\item[(K) Fixed kinetics (used by (c) and (d)(v)(3)).] The reader's kinetics $K$ are fixed in advance. The shipped readers choose $(\ell,\lambda)$ by cross-validation with $\kappa=1$.
\begin{itemize}
\item Cadence-aware reader on heart rate: it picked $(2,0.95)$ in all five folds on hourly records (job 10679) and $(2,0.8)$ in all five on held records (job 10678).
\item Plain reader, job 10679: it picked $(2,0.65)$ twice and $(2,0.8)$ three times.
\end{itemize}
Fitted constants compared across cadences can therefore mix kinetics.
\item[(P) Population readers (used by (c)).] (c) is a statement about population least squares with an intercept at fixed kinetics. The shipped readers use ridge regression with a nested cross-validated penalty and kinetics chosen from a grid, and their held-out $R^2$ estimates the population value with a finite-sample loss.

Take ordinary least squares with an intercept on $p$ features (design-only features included) and $n$ training rows, and suppose the features are Gaussian and the regression residual is independent of them with constant variance. Then the expected held-out loss relative to $\mathrm{Var}(y)$ is exactly $(1-R^2)(p+1-2/n)/(n-p-2)$. With $n_{\rm tr}=320$, $p=24$ and $R^2=0.2$ this is $0.068$. For the shipped ridge readers the expression is an approximation.
\item[(H) Orthogonality (used by (c)).] $\mathrm{Cov}(\tilde X^0,T)=0$, $\mathrm{Cov}(\tilde X^0,\tilde U_X)=0$, $\Sigma_0\succ 0$ and $\boldsymbol\theta\neq 0$.

Exact sufficient conditions under (G3):
\begin{itemize}
\item the second condition holds if $\Phi_{\mathcal S}\,\mathrm E[U\mid\mathcal S]\in\mathrm{span}(X^{\mathcal S})$ or $\mathrm E[X^0\mid\mathcal S]\in\mathrm{span}(X^{\mathcal S})$;
\item the first holds if $\mathrm E[T\mid Z^0,\mathcal S]\in\mathrm{span}(X^{\mathcal S})$.
\end{itemize}
In the generator's linear regime, $\mathrm E[U\mid\mathcal S]=\Lambda\,\mu_\zeta\,q(d(\cdot))$ channelwise, and $\mathrm E[T\mid Z^0,\mathcal S]$ is a combination of the trait means of the treatments that load the scored channel. So (H) holds up to terms proportional to trait means and to the $\tanh$ nonlinearity. For phenylephrine the population trait mean is about 0.057 and the SD about 0.58, so $\mu_\zeta^2/\sigma_\zeta^2\approx 0.01$ (panel estimates 0.012, 0.004 and 0.010 for phenylephrine, norepinephrine and NaCl).

Reader-specific remarks:
\begin{itemize}
\item Cadence-aware reader at $K_0$: the per-channel energies $Q^{(0)}_{ik}$ are features and contain $\Phi_{\mathcal S}\mathrm E[U\mid\mathcal S]$ in their span, which removes the first kind of term exactly. At other kinetics the term is removed only to the extent that $\Xi_{ik}(K)$ is affine in $Q^{(0)}_{ik}(K)$.
\item Plain reader: it has no design-only feature.
\end{itemize}
In the scaling check, response-free record features explained an adjusted $R^2$ of $-0.007$ of $T$.
\item[(E) Design-noise orthogonality (idealization; used by (c)(iv) and (d)(v)).] With $T=k_T\zeta+\varepsilon$ as in (c)(iv), $\mathrm{Cov}(\tilde U_X,\varepsilon)=0$. A sufficient condition is $\mathcal S\perp(\zeta,\varepsilon)$, which (d)(v) assumes. In the generator, $\zeta\perp\mathcal S$ holds by (G3), but $\mathcal S\perp\varepsilon$ holds only approximately. $\varepsilon$ contains the part of $\tau^{\rm own}$ that is nonlinear in $\zeta$, which depends on $e_i(t_0)$, a function of the doses. It also contains $\nu_i$, which depends on the counterfactual policy; that policy shares the person's propensity parameters and latent state with the harvest run that produced the doses, and it runs on the tape that produced the draw sets. Measured on a 400-person simulated panel (synthetic tapes, NATMIN, $c=0.75$):
\begin{itemize}
\item $\mathrm{Cov}(\tilde U_X,\varepsilon)/\boldsymbol\theta=-0.007$ (bootstrap SD 0.031) for $\Xi=Q^{(0)}$;
\item $\mathrm{Cov}(\tilde U_X,\varepsilon)/\boldsymbol\theta=-0.006$ (bootstrap SD 0.028) for $\Xi=M$;
\item $\mathrm{corr}(\varepsilon,\text{number of dosed rows})=-0.018$.
\end{itemize}
\item[(L) Linear regime (idealization; used by (c)(iv) and (d)).] $\tanh x$ is replaced by $x$ in the record's shift, so $U_{ik}(t)=\Lambda_k\zeta_i q_i(d_{ik}(t))$. The argument reaches $|\Lambda_k\zeta q|\le 2\cdot 5/6\approx 1.67$ at the trait ceiling, where $\tanh(1.67)=0.93$.
\item[(RW) Random-walk baseline (idealization; used by (d) only).] There is one carrier $k$. Given $\mathcal S$, its trait-free path is $m(t)+b(t)$, where $m$ is a known $\mathcal S$-measurable mean path and $b(t)=b(0)+\sum_{u=1}^{t}\xi_u$ with $\xi_u$ i.i.d. $N(0,\sigma_b^2)$, $\sigma_b>0$ known, independent of $b(0)$ and of the traits. The generator's trait-free action route is linear time-invariant in the doses when its clips are inactive, so it is such an $m$.

This idealizes four things:
\begin{itemize}
\item It treats one carrier, whereas the generator writes phenylephrine's response into heart rate, pH, white cells and platelets, and NaCl's into mean arterial pressure, SpO$_2$, lactate, respiratory rate, urine output, creatinine, platelets and bilirubin.
\item The real baseline is persistent but not a random walk: the detrended heart-rate lag-1 autocorrelation is 0.69. The white-noise baseline of Proposition 5 is the other bracket.
\item Draw sets are not independent of the baseline, since both come from the real tape.
\item (d)(v)(2) also treats the trait as Gaussian, whereas the generator's trait is a censored Gaussian with an atom of about 15\% at $-2/3$.
\end{itemize}
\end{description}

\paragraph{Proof.}

\noindent\textbf{Conventions.} Arguments are in exact arithmetic. Every generator variable used is bounded ($|T_i|<2$, $|y_{ik}(t)|\le 8+c$, $0\le q<5/6$ under binary dosing, gaps at most 89), and (RW) adds Gaussian variables. So all moments below exist. Statements conditional on $\mathcal S_i$ hold almost surely.

\medskip\noindent\textbf{Proof of (a).}

\emph{(i)} Fix $i$ and $a$. By (G1), the rollouts of frames A and 0 under regime $a$ start from the same factual trace with the same seed. The only operation that reads the cap is the shift, which nothing else reads and which draws no random numbers. By induction over the steps $s$, the two frames have the same latent state, the same actions of all five treatments and the same noise, and hence by (G2) the same memories $e^a_{is,\cdot}$ at every step. In particular, the pre-shift value $\tilde y^{\,a}_{is}$ of channel $k^\ast$ at row $t_0+s+1$ is common to the two frames.

Frame 0 has cap 0 and adds nothing. Frame A adds $c\,\alpha\tanh\Gamma^a_{is}$ with $\alpha=1$ by (A1), and by (A1) the raw-unit clip is inactive. On the latent path in standardized units, therefore, $y^{A,a}_{is}=\tilde y^{\,a}_{is}+c\tanh\Gamma^a_{is}$ and $y^{0,a}_{is}=\tilde y^{\,a}_{is}$. The summand of Definition 4 is
\[
\bigl(y^{A,1}_{is}-y^{A,0}_{is}\bigr)-\bigl(y^{0,1}_{is}-y^{0,0}_{is}\bigr)=c\,\bigl(\tanh\Gamma^1_{is}-\tanh\Gamma^0_{is}\bigr).
\]
Averaging over $s=0,\dots,h-1$ gives $\tau_i=cT_i$.

$T_i$ is a function of the traits, the loadings and the memories. The memories are functions of the factual doses and of the counterfactual actions, and the actions depend on the latent state and the policy draws, neither of which depends on $c$ by (G1). Hence $T_i$ does not depend on $c$.

At $s=0$, (G2) gives $e^a_{i0,j}=0.8\,e_{ij}(t_0)+A_{ij}(t_0-1)$. Both terms are factual and therefore equal for $a=0$ and $a=1$. Hence $\Gamma^1_{i0}=\Gamma^0_{i0}$, and the $s=0$ summand is $0$.

\emph{(ii)} Unrolling the recursion of (G2) from $e^a_{i,-1,j^\ast}=e_i(t_0)$ gives
\[
e^a_{is,j^\ast}=\lambda_0^{s+1}e_i(t_0)+\lambda_0^{s}A_i(t_0-1)+\sum_{m=1}^{s}\lambda_0^{s-m}a .
\]
By (G2) $A_i(89)=0$, and $\sum_{m=1}^{s}\lambda_0^{s-m}=(1-\lambda_0^s)/(1-\lambda_0)$, so $q^a_{is,j^\ast}=q^a_s(e_i(t_0))$.

For $j\neq j^\ast$ the factual memory and dose are $0$ (G2), so $e^a_{is,j}=\sum_{m=1}^{s}\lambda_0^{s-m}\tilde a^{a}_{i,m-1,j}$, where $\tilde a^a_{i,m,j}$ is the policy action at step $m$. Suppose $\tilde a^a_{i,m,j}=0$ for all $m\le h-2$, both regimes and every $j\neq j^\ast$ with $L_{k^\ast j}\neq 0$. Then $q^a_{is,j}=0$ for these $j$ and all $s\le h-1$, so $\Gamma^a_{is}=\Lambda_{k^\ast}\zeta_iq^a_s(e_i(t_0))$ and $cT_i=\tau^{\rm own}(\zeta_i,e_i(t_0))$, that is, $\nu_i=0$. The map $\tau^{\rm own}$ involves only $c$, $h$, $\lambda_0$, $\Lambda_{k^\ast}$ and $\varphi_1$, so it is the same for every patient and dosing schedule.

\emph{(iii)} Since $q^0_s(0)=\varphi_1(0)=0$ and $q^1_0(0)=\varphi_1(0)=0$, we have $D(0)=h^{-1}\sum_{s=1}^{h-1}\varphi_1\bigl((1-\lambda_0^s)/(1-\lambda_0)\bigr)$, which evaluates to $0.684222$.

Put $\upsilon_s:=(1-\lambda_0^s)/(1-\lambda_0)$ and $x_s:=\lambda_0^{s+1}e\ge 0$. Then $q^1_s(e)=\varphi_1(\upsilon_s+x_s)$ and $q^0_s(e)=\varphi_1(x_s)$, so, using $\varphi_1(0)=0$,
\[
D(e)-D^\ast=\frac1h\sum_{s=0}^{h-1}\bigl[\varphi_1(\upsilon_s+x_s)-\varphi_1(\upsilon_s)-\varphi_1(x_s)\bigr].
\]
$\varphi_1$ is concave on $[0,\infty)$ with $\varphi_1(0)=0$, hence subadditive, so every bracket is at most $0$. Since $\varphi_1$ is nondecreasing and $\varphi_1(x)\le x$, every bracket is at least $-\varphi_1(x_s)\ge -x_s$. Summing,
\[
0\le D^\ast-D(e)\le h^{-1}\sum_{s=0}^{h-1}\lambda_0^{s+1}e=\lambda_0(1-\lambda_0^{h})e/\bigl(h(1-\lambda_0)\bigr),
\]
and $0.8(1-0.8^{12})/2.4=0.310427\le 0.3105$.

\emph{(iv)} For measurable $f$ with $\mathrm E f(\zeta_i)^2<\infty$, the tower property removes the cross term in
\[
\mathrm E\bigl(\tau-f(\zeta)\bigr)^2=\mathrm E\,\mathrm{Var}(\tau\mid\zeta)+\mathrm E\bigl(\bar\tau(\zeta)-f(\zeta)\bigr)^2 ,
\]
which is minimized exactly when $f=\bar\tau$ almost surely. Hence $R^2(\bar\tau)=1-\mathrm E\,\mathrm{Var}(\tau\mid\zeta)/\mathrm{Var}(\tau)$. Since $\mathrm E[\tau-\bar\tau(\zeta)]=0$, the root mean squared error of $\bar\tau$ is $(\mathrm E\,\mathrm{Var}(\tau\mid\zeta))^{1/2}$ and its recovery is $1-(1-R^2_{\rm trait})^{1/2}$. The argument ranges over $\sigma(\zeta_i)$-measurable $f$ only.

\medskip\noindent\textbf{Proof of (b).} Fix $c$ at which (A1) holds. By (G2) only $j^\ast$ is dosed in the factual record, so $q_{ij}\equiv 0$ for $j\neq j^\ast$ and $\Gamma_{ik}(t)=\Lambda_k\zeta_iq_i(t)$. By (G1) the factual runs at caps $0$ and $c$ share $\tilde y_{ik}(t)$ at every row. By (A1), $y_{ik}(t;c)=\tilde y_{ik}(t)+c\tanh(\Lambda_k\zeta_iq_i(t))$, while the cap-0 run has $y_{ik}(t;0)=\tilde y_{ik}(t)$.

At a draw row the recorded raw value is $\mu_k+\sigma_ky_{ik}(t)$, unclipped by (A1), and by (G4) every other row repeats the previous recorded value. Hence the standardized held record is $Z_{ik}(t;c)=y_{ik}(d_{ik}(t);c)$, and the identity follows by substitution.

Finally, $\tilde y$ depends on the tape, the noise and the doses, and $q_i$ depends on the doses. The doses are harvested by the policy, which reads only the latent state (G1), and are topped up with a seed fixed by the person index. $\mathcal D_{ik}$ depends on the tape only (G4). None of these depends on $c$, and $T_i$ does not by (a).

\medskip\noindent\textbf{Proof of (c).} Two lemmas are used.

\emph{Lemma 1 (partial $R^2$).} Let $X^{\mathcal S}$ contain a constant, let $Y$ be square integrable with $\mathrm{Var}(Y)>0$, and let $X$ be square integrable with $\mathrm{Var}(\tilde X)\succ 0$. Then the population $R^2$ of $Y$ on $(X,X^{\mathcal S})$ equals $R^2(Y\mid X^{\mathcal S})+\mathrm{Cov}(\tilde X,Y)^\top\mathrm{Var}(\tilde X)^{-1}\mathrm{Cov}(\tilde X,Y)/\mathrm{Var}(Y)$.

\emph{Proof.} In $L^2$, $\mathrm{span}(X,X^{\mathcal S})=\mathrm{span}(X^{\mathcal S})\oplus\mathrm{span}(\tilde X)$ with orthogonal summands, because every component of $\tilde X$ is orthogonal to $\mathrm{span}(X^{\mathcal S})$. The projection of $Y$ on an orthogonal sum is the sum of the projections, so the explained variances add. The projection of $Y$ on $\mathrm{span}(\tilde X)$ is $\beta^\top\tilde X$ with $\beta=\mathrm{Var}(\tilde X)^{-1}\mathrm E[\tilde XY]$, and $\mathrm E[\tilde XY]=\mathrm{Cov}(\tilde X,Y)$ since $\mathrm E\tilde X=0$. $\square$

\emph{Lemma 2 (constrained minimum).} Let $\mathbf N\succeq 0$ be symmetric and $\boldsymbol\theta\in\mathrm{range}(\mathbf N)\setminus\{0\}$. Then $\min\{v^\top\mathbf Nv:\ v^\top\boldsymbol\theta=1\}=1/(\boldsymbol\theta^\top\mathbf N^{+}\boldsymbol\theta)>0$, and $v$ attains the minimum iff $v^\top\boldsymbol\theta=1$ and $\mathbf Nv\in\mathrm{span}\{\boldsymbol\theta\}$. If $\mathbf N\succ 0$, the minimizer is unique: $v=\mathbf N^{-1}\boldsymbol\theta/(\boldsymbol\theta^\top\mathbf N^{-1}\boldsymbol\theta)$.

\emph{Proof.} Write $\boldsymbol\theta=\mathbf Nw$. Then $w^\top\mathbf Nw=w^\top\mathbf N\mathbf N^{+}\mathbf Nw=\boldsymbol\theta^\top\mathbf N^+\boldsymbol\theta$, and this is positive, since $w^\top\mathbf Nw=0$ would give $\mathbf Nw=0$. For $v^\top\boldsymbol\theta=1$, the Cauchy--Schwarz inequality for the semi-inner product $u^\top\mathbf Nv$ gives $1=(v^\top\mathbf Nw)^2\le(v^\top\mathbf Nv)(w^\top\mathbf Nw)$. Equality holds iff $\mathbf N^{1/2}v$ and $\mathbf N^{1/2}w$ are proportional, that is, iff $\mathbf Nv\in\mathrm{span}\{\mathbf Nw\}=\mathrm{span}\{\boldsymbol\theta\}$. The vector $v=\mathbf N^+\boldsymbol\theta/(\boldsymbol\theta^\top\mathbf N^+\boldsymbol\theta)$ satisfies $v^\top\boldsymbol\theta=1$ and $\mathbf Nv=\boldsymbol\theta/(\boldsymbol\theta^\top\mathbf N^+\boldsymbol\theta)$, so the bound is attained. If $\mathbf N\succ 0$, the conditions $\mathbf Nv=\beta\boldsymbol\theta$ and $v^\top\boldsymbol\theta=1$ determine $v$. $\square$

\emph{(i)} Let $c\in\mathbb A$. By (R) the reader's detected draw sets are the $\mathcal D_{ik}$, so its features are $a_{\mathcal S}+\Phi_{\mathcal S}Z(c)$ and $X^{\mathcal S}$ with the objects of the statement. By (A1) and (b), $X(c)=X^0+c\,U_X$.

Since $\tau=cT$ with $c>0$ fixed, the $R^2$ of $\tau$ equals that of $T$ on the same features. The residual map $V\mapsto\tilde V$ is linear and $X^{\mathcal S}$ does not depend on $c$, so $\tilde X(c)=\tilde X^0+c\,\tilde U_X$. By (H), $\mathrm{Cov}(\tilde X(c),T)=c\,\boldsymbol\theta$ and $\mathrm{Var}(\tilde X(c))=\Sigma_0+c^2\Sigma_U\succ 0$. Lemma 1 gives
\[
R^2(c)=R^2_{\mathcal S}+c^2\boldsymbol\theta^\top(\Sigma_0+c^2\Sigma_U)^{-1}\boldsymbol\theta/\mathrm{Var}(T)=R^2_{\mathcal S}+\mathcal L(c).
\]
$R^2_{\mathcal S}$ involves only $T$ and $X^{\mathcal S}$, which do not depend on $c$.

For any $\mathcal R$-measurable $\hat\tau$ with finite variance, $\mathrm E(\tau-\hat\tau)^2\ge\mathrm E(\tau-\mathrm E[\tau\mid\mathcal R])^2=\mathrm{Var}(\tau)-\mathrm{Var}\,\mathrm E[\tau\mid\mathcal R]$. Hence $R^2(\hat\tau)\le R^2_\ast(\sigma(\mathcal R))$. The population least-squares predictor on $(X(c),X^{\mathcal S})$ is $\mathcal R$-measurable.

\emph{(ii)} First, $\boldsymbol\theta\in\mathrm{range}(\Sigma_U)$. If $\Sigma_Uu=0$, then $\mathrm{Var}(u^\top\tilde U_X)=0$, so $u^\top\tilde U_X$ is almost surely constant and $u^\top\boldsymbol\theta=\mathrm{Cov}(u^\top\tilde U_X,T)=0$; thus $\boldsymbol\theta\perp\mathrm{null}(\Sigma_U)$.

Directions with $v^\top\boldsymbol\theta=0$ have zero correlation with $T$. Hence, by Lemma 2, the population $R^2$ of $T$ on $\tilde U_X$ is
\[
\sup_v\mathrm{corr}^2(v^\top\tilde U_X,T)=\sup_{v^\top\boldsymbol\theta=1}\frac{1}{\mathrm{Var}(T)\,v^\top\Sigma_Uv}=\frac{\boldsymbol\theta^\top\Sigma_U^+\boldsymbol\theta}{\mathrm{Var}(T)}=\frac{1}{C_0},
\]
and it is at most 1. $C_1>0$ because $\Sigma_0\succ 0$ and $\boldsymbol\theta\neq 0$.

For $x>0$, $H(x)=\mathrm{Var}(T)/\bigl(\boldsymbol\theta^\top(x\Sigma_0+\Sigma_U)^{-1}\boldsymbol\theta\bigr)$. Since $x\Sigma_0+\Sigma_U\succ 0$, Lemma 2 gives
\[
H(x)=\mathrm{Var}(T)\min_{v^\top\boldsymbol\theta=1}\bigl[x\,v^\top\Sigma_0v+v^\top\Sigma_Uv\bigr].
\]
For fixed $v$ the bracket is affine in $x$ with slope $v^\top\Sigma_0v>0$. A pointwise minimum of affine functions is concave, and here it is nondecreasing. It is strictly increasing: for $0<x_1<x_2$ and $v_2$ the minimizer at $x_2$, $H(x_1)\le\mathrm{Var}(T)[x_1v_2^\top\Sigma_0v_2+v_2^\top\Sigma_Uv_2]<H(x_2)$.

The minimum of a sum over a common set is at least the sum of the minima. Lemma 2, applied to $\Sigma_0$ and to $\Sigma_U$, therefore gives
\[
H(x)\ \ge\ \mathrm{Var}(T)\Bigl[\frac{x}{\boldsymbol\theta^\top\Sigma_0^{-1}\boldsymbol\theta}+\frac{1}{\boldsymbol\theta^\top\Sigma_U^{+}\boldsymbol\theta}\Bigr]=C_0+C_1x .
\]
For the limits, let $v_U:=\Sigma_U^+\boldsymbol\theta/(\boldsymbol\theta^\top\Sigma_U^+\boldsymbol\theta)$ and $v_0:=\Sigma_0^{-1}\boldsymbol\theta/(\boldsymbol\theta^\top\Sigma_0^{-1}\boldsymbol\theta)$ be the minimizers of the two terms. Inserting them gives
\[
H(x)\le C_0+x\,\mathrm{Var}(T)\,v_U^\top\Sigma_0v_U,\qquad H(x)\le C_1x+\mathrm{Var}(T)\,v_0^\top\Sigma_Uv_0 .
\]
Together with the lower bound, $H(x)\to C_0$ as $x\downarrow 0$ and $H(x)/x\to C_1$ as $x\to\infty$. With $x=c^{-2}$, these are $\mathcal L(c)\to 1/C_0$ as $c\to\infty$ and $\mathcal L(c)/c^2=x/H(x)\to 1/C_1$ as $c\to 0$.

Equality. Over a common constraint set, $\min(f+g')=\min f+\min g'$ iff some point minimizes $f$ and $g'$ simultaneously. (If $v^\ast$ minimizes $f+g'$ and equality holds, then $f(v^\ast)-\min f$ and $g'(v^\ast)-\min g'$ are nonnegative and sum to $0$.) For $x>0$, $x\,v^\top\Sigma_0v$ is minimized on $\{v^\top\boldsymbol\theta=1\}$ only at $v_0$. By Lemma 2, $v^\top\Sigma_Uv$ is minimized exactly by the $v$ with $v^\top\boldsymbol\theta=1$ and $\Sigma_Uv\in\mathrm{span}\{\boldsymbol\theta\}$. So equality at some $x>0$ holds iff $\Sigma_Uv_0\in\mathrm{span}\{\boldsymbol\theta\}$, a condition free of $x$.

This condition is equivalent to the stated one. If $\Sigma_Uv_0\in\mathrm{span}\{\boldsymbol\theta\}$, then $v=v_0$ is nonzero (because $\boldsymbol\theta\neq 0$) and has $\Sigma_0v_0=\boldsymbol\theta/(\boldsymbol\theta^\top\Sigma_0^{-1}\boldsymbol\theta)$. Conversely, let $v\neq 0$ with $\Sigma_0v=\beta\boldsymbol\theta$ and $\Sigma_Uv\in\mathrm{span}\{\boldsymbol\theta\}$. Then $\beta\neq 0$ because $\Sigma_0\succ 0$, so $v=\beta\,\Sigma_0^{-1}\boldsymbol\theta$ is a nonzero multiple of $v_0$, and $\Sigma_Uv_0\in\mathrm{span}\{\boldsymbol\theta\}$.

The population least-squares coefficient on $\tilde X(c)$ is $(\Sigma_0+c^2\Sigma_U)^{-1}c\boldsymbol\theta=c^{-1}(x\Sigma_0+\Sigma_U)^{-1}\boldsymbol\theta$ with $x=c^{-2}$. By Lemma 2 applied to $x\Sigma_0+\Sigma_U\succ 0$, this is proportional to the unique minimizer at $x$. Under the condition, that minimizer is $v_0$ at every $x$.

Now take $v=v_0$, so that $\mathrm{Cov}(v^\top\tilde U_X,T)=v^\top\boldsymbol\theta=1$. Then $C_1=\mathrm{Var}(T)\,v^\top\Sigma_0v=\mathrm{Var}(T)\mathrm{Var}(v^\top\tilde X^0)/\mathrm{Cov}(v^\top\tilde U_X,T)^2$. Under the condition $v_0$ also minimizes $v^\top\Sigma_Uv$, so $C_0=\mathrm{Var}(T)\,v^\top\Sigma_Uv=1/\mathrm{corr}^2(v^\top\tilde U_X,T)$. Both ratios are invariant to rescaling $v$. For $p=1$ every $v\neq 0$ satisfies the condition.

\emph{(iii)} By (i), for $c\in\mathbb A$ the gain is $\mathcal L(c)=1/H(c^{-2})\le 1/(C_0+C_1/c^2)<1/C_0$. Since $\mathcal L(c)\to 1/C_0$ as $c\to\infty$ by (ii), $\sup_{c>0}\mathcal L(c)=1/C_0$, and the strict inequality shows that the supremum is not attained.

If the gain at $c\in\mathbb A$ is at least $0.2$, then $5\ge H(c^{-2})\ge C_0+C_1/c^2$. Since $C_1/c^2>0$, this forces $C_0<5$ and $c^2\ge C_1/(5-C_0)$. In the one-statistic case $H(x)=C_0+C_1x$, so $\mathcal L(c^\ast)=1/(C_0+(5-C_0))=0.2$. Because $\mathcal L$ is strictly increasing in $c$, $\mathcal L(c)<0.2$ for $c<c^\ast$; hence, if $c^\ast\in\mathbb A$, it is the reader's crossing amplitude. If $1/C_0\le 0.2$, then $\mathcal L(c)<1/C_0\le 0.2$ at every $c$. The total $R^2$ is $R^2_{\mathcal S}+\mathcal L(c)$ with $R^2_{\mathcal S}$ free of $c$, which gives the stated replacement of $0.2$. If $(0,\bar c]\subseteq\mathbb A$, the gain equals $\mathcal L$ on that interval, and $\mathcal L(c)/c^2\to 1/C_1$ by (ii).

Amplitude ratio. $H$ is concave, hence continuous on $(0,\infty)$, and strictly increasing. So $c\mapsto\mathcal L(c)=1/H(c^{-2})$ is continuous and strictly increasing, with limits $0$ as $c\to 0$ (because $\mathcal L(c)\le c^2/C_1$) and $1/C_0$ as $c\to\infty$. For $\varrho$ in the stated range each law therefore takes the value $\varrho$ at a unique $c(\varrho)$, and $c(\varrho)\to 0$ as $\varrho\downarrow 0$. By the low-signal limit, $\varrho=c(\varrho)^2(1+o(1))/C_1$, so $c(\varrho)^2=C_1\varrho(1+o(1))$. Applying this to both laws and taking the ratio gives $c_{\rm new}(\varrho)/c_{\rm old}(\varrho)\to(C_1^{\rm new}/C_1^{\rm old})^{1/2}$. If both sets $\mathbb A$ contain $(0,\bar c]$, then $c(\varrho)\le\bar c$ for all small $\varrho$, and there each reader's gain equals its law.

\emph{(iv)} Here $p=1$. For a design-only $w$, $\zeta\perp\mathcal S$ gives $\mathrm E[\zeta\Xi]=\mu_\zeta\mathrm E\,\Xi$ and $\mathrm{Cov}(\zeta\Xi,w)=\mu_\zeta\mathrm{Cov}(\Xi,w)$. The linear projection depends only on means and covariances, so $\mathrm{Proj}(\zeta\Xi\mid X^{\mathcal S})=\mu_\zeta\mathrm{Proj}(\Xi\mid X^{\mathcal S})$ and
\[
\tilde U_X=\Lambda_k\bigl[(\zeta-\mu_\zeta)\Xi+\mu_\zeta\tilde\Xi\bigr].
\]
$\tilde\Xi$ has mean $0$ because $X^{\mathcal S}$ contains a constant. Using $\zeta\perp\mathcal S$ and $\mathrm E(\zeta-\mu_\zeta)=0$, the cross term vanishes, and $\mathrm{Var}(\tilde U_X)=\Lambda_k^2\bigl[\sigma_\zeta^2\mathrm E\,\Xi^2+\mu_\zeta^2\mathrm{Var}(\tilde\Xi)\bigr]$.

For the covariance: $\mathrm{Cov}((\zeta-\mu_\zeta)\Xi,\zeta)=\mathrm E\,\Xi\cdot\mathrm E[(\zeta-\mu_\zeta)\zeta]=\sigma_\zeta^2\mathrm E\,\Xi$, and $\mathrm{Cov}(\tilde\Xi,\zeta)=\mathrm E\,\tilde\Xi\,\mathrm E\,\zeta=0$. By (E), $\mathrm{Cov}(\tilde U_X,\varepsilon)=0$. So $\boldsymbol\theta=\mathrm{Cov}(\tilde U_X,T)=\Lambda_kk_T\sigma_\zeta^2\mathrm E\,\Xi\neq 0$. With $\mathrm{Var}(T)=k_T^2\sigma_\zeta^2/\rho^2_{\rm lin}$ and $\mathrm E\,\Xi^2/(\mathrm E\,\Xi)^2=1+\mathrm{CV}^2(\Xi)$,
\[
C_0=\frac{\mathrm{Var}(T)\,\mathrm{Var}(\tilde U_X)}{\boldsymbol\theta^2}=\frac{\sigma_\zeta^2\mathrm E\,\Xi^2+\mu_\zeta^2\mathrm{Var}(\tilde\Xi)}{\rho^2_{\rm lin}\,\sigma_\zeta^2(\mathrm E\,\Xi)^2}=\frac{1+\mathrm{CV}^2(\Xi)+(\mu_\zeta^2/\sigma_\zeta^2)\mathrm{Var}(\tilde\Xi)/(\mathrm E\,\Xi)^2}{\rho^2_{\rm lin}} .
\]

The two readers. In the linear regime, $Z_{ik}(t)=Z^0_{ik}(t)+c\Lambda_k\zeta_iq_i(d_{ik}(t))$, so $\Delta_jZ_{ik}=\Delta_jZ^0_{ik}+c\Lambda_k\zeta_i\Delta_jq_i$.
\begin{itemize}
\item Cadence-aware: $\mathcal C_{ik}(K)=\sum_j\Delta_jZ^0_{ik}\Delta_jp_K+c\Lambda_k\zeta_i\Xi_{ik}(K)$. So $U_X=\Lambda_k\zeta_i\Xi_{ik}(K)$, and the reader's design-only features are $(1,Q^{(0)}_{ik}(K))$. At $K=K_0$, $p_{K_0}=q_i$ on rows $0,\dots,89$, so $\Xi_{ik}(K_0)=Q^{(0)}_{ik}(K_0)=Q^{(0)}_{ik}$ lies in $\mathrm{span}(X^{\mathcal S})$ and $\tilde\Xi=0$. At other $K$, $\mathrm{Var}(\tilde\Xi)=0$ iff $\Xi_{ik}(K)$ is almost surely an affine function of $Q^{(0)}_{ik}(K)$.
\item Plain: $\mathcal O_{ik}(K)=\sum_t(\Pi p_K)(t)Z^0_{ik}(t)+c\Lambda_k\zeta_iM_{ik}(K)$ and $X^{\mathcal S}=1$. So $\tilde\Xi=\Xi-\mathrm E\,\Xi$, and $\mathrm{Var}(\tilde\Xi)/(\mathrm E\,\Xi)^2=\mathrm{CV}^2(\Xi)$.
\end{itemize}

Finally, if $\mathcal S\perp(\zeta,\varepsilon)$, then $\mathrm E[(\zeta-\mu_\zeta)\Xi\varepsilon]=\mathrm E\,\Xi\cdot\mathrm{Cov}(\zeta,\varepsilon)=0$ (by the definition of $k_T$) and $\mathrm{Cov}(\tilde\Xi,\varepsilon)=0$. Hence $\mathrm{Cov}(\tilde U_X,\varepsilon)=0$.

\emph{(v)} By (ii), $H$ is concave and strictly increasing on $(0,\infty)$, $H(x)\to C_0$ as $x\downarrow 0$ and $H(x)/x\to C_1$ as $x\to\infty$. Put $x_\ell:=c_\ell^{-2}$, so $x_1<x_2$, and let $L(x):=\hat C_0+\hat C_1x$ be the line through $(x_1,H(x_1))$ and $(x_2,H(x_2))$; thus $\hat C_1=(H(x_2)-H(x_1))/(x_2-x_1)$. For a concave function and $u<v<w$, the chord slopes satisfy
\[
\frac{H(v)-H(u)}{v-u}\ge\frac{H(w)-H(u)}{w-u}\ge\frac{H(w)-H(v)}{w-v}.
\]
We apply this three times.
\begin{itemize}
\item For $x\in(0,x_1)$, take $(u,v,w)=(x,x_1,x_2)$: $(H(x_1)-H(x))/(x_1-x)\ge\hat C_1$, so $H(x)\le L(x)$. Letting $x\downarrow 0$ gives $C_0\le\hat C_0$.
\item For $y>x_2$, take $(u,v,w)=(x_1,x_2,y)$: $(H(y)-H(x_2))/(y-x_2)\le\hat C_1$, so $H(y)\le L(y)$. As $y\to\infty$ the left side tends to $C_1$, because $H(y)/y\to C_1$ and $y/(y-x_2)\to 1$; hence $C_1\le\hat C_1$.
\item For $x\in[x_1,x_2]$, take $(u,v,w)=(x_1,x,x_2)$: $(H(x)-H(x_1))/(x-x_1)\ge\hat C_1$, so $H(x)\ge L(x)$.
\end{itemize}
$L$ is positive on $(0,\infty)$: on $[x_1,x_2]$ it lies between $H(x_1)>0$ and $H(x_2)>0$, and outside that interval $L\ge H>0$. Taking reciprocals, $1/L(x)\ge\mathcal L(x^{-1/2})$ for $x\in[x_1,x_2]$, that is, for $c\in[c_2,c_1]$, and $1/L(x)\le\mathcal L(x^{-1/2})$ outside. In the one-statistic case $H$ is affine, so $L=H$ and every relation is an equality. When $c_1,c_2\in\mathbb A$, (i) identifies $\varrho_\ell$ with the reader's gains.

\medskip\noindent\textbf{Proof of (d).} Condition on $\mathcal S$ throughout. By (A1), (L) and (RW), $y(t)=m(t)+b(t)+c\Lambda\zeta q(t)$ for $t\le 89$, where $m$ is known and $\mathcal S$-measurable.

\emph{(i)} With $j(t):=\max\{j:d_j\le t\}$, $Z(t)=y(d_{j(t)})$ is a function of the draw values and $\mathcal S$; conversely $y(d_j)=Z(d_j)$. This part uses neither (L) nor (RW).

\emph{(ii)} Because $e(0)=0$, we have $q(0)=\varphi_1(0)=0$, so $Y_0-m(0)=b(0)$, where $Y_j:=y(d_j)$. The map from the draw values $(Y_0,\dots,Y_n)$ to $(Y_0,\Delta_0Y,\dots,\Delta_{n-1}Y)$ is a bijection, and
\[
\Delta_j(Y-m)=c\Lambda\zeta\,\Delta_jq+\sum_{u=d_j+1}^{d_{j+1}}\xi_u .
\]
The noise sums run over disjoint blocks, so they are independent $N(0,\sigma_b^2g_j)$ variables, independent of $b(0)$ and $\zeta$. With $\phi_v$ the $N(0,v)$ density, the joint density given $\zeta$ therefore factorizes as
\[
p_{b(0)}(Y_0-m(0))\prod_j\phi_{\sigma_b^2g_j}\bigl(\Delta_j(Y-m)-c\Lambda\zeta\Delta_jq\bigr).
\]
Its logarithm equals $(c\Lambda\zeta/\sigma_b^2)\,\mathcal C^{\rm eff}-(c\Lambda\zeta)^2J/(2\sigma_b^2)$ plus terms free of $\zeta$, where $\Delta_jY=\Delta_jZ$. By the factorization theorem $\mathcal C^{\rm eff}$ is sufficient. It is Gaussian with mean $c\Lambda\zeta\sum_j(\Delta_jq)^2/g_j=c\Lambda\zeta J$ and variance $\sum_j(\Delta_jq)^2\sigma_b^2g_j/g_j^2=\sigma_b^2J$. Minus the second derivative of the log-likelihood in $\zeta$ is $(c\Lambda)^2J/\sigma_b^2$. On the hourly record $d_j=j$, $g_j=1$ and $J=Q^\circ$.

For a general profile with $q(d_0)\neq 0$ and a diffuse level, write $b(0)=\beta_0+\xi_0$, with $\beta_0$ an unknown real parameter and $\xi_0\sim N(0,s_0^2)$. The information matrix for $(\zeta,\beta_0)$ has entries $I_{\beta\beta}=1/s_0^2$, $I_{\zeta\beta}=c\Lambda q(d_0)/s_0^2$ and $I_{\zeta\zeta}=(c\Lambda q(d_0))^2/s_0^2+(c\Lambda)^2J/\sigma_b^2$. So the efficient information $I_{\zeta\zeta}-I_{\zeta\beta}^2/I_{\beta\beta}$ is again $(c\Lambda/\sigma_b)^2J$.

\emph{(iii)} $\Delta_jq=\sum_{t=d_j+1}^{d_{j+1}}(q(t)-q(t-1))$ is a sum of $g_j$ terms. By the Cauchy--Schwarz inequality $(\Delta_jq)^2\le g_j\mathcal E_j$, with equality iff the $g_j$ hourly increments are equal, that is, iff $q$ is affine on the integers of $[d_j,d_{j+1}]$. Hence $\psi_j\in[0,1]$ with the stated equality cases, and $\psi_j=0$ iff $\Delta_jq=0$. If $\mathcal E_j=0$, then $\Delta_jq=0$ and the interval contributes to neither $J$ nor $Q^\circ$. The intervals $(d_j,d_{j+1}]$, $0\le j<n$, and $(d_n,89]$ partition $\{1,\dots,89\}$. So $Q^\circ=\sum_j\mathcal E_j+\mathcal E_{\rm tail}$ and $J=\sum_{j:\mathcal E_j>0}\psi_j\mathcal E_j$. Dividing by $Q^\circ$ gives the identity, and $\chi\le\sum_jw_j\le 1$.

Generator's profile. Let $A$ be binary and $\lambda_0=4/5$.
\begin{itemize}
\item Zero memory. For $t\ge 0$, $e(t)=\sum_{u=0}^{t-2}\lambda_0^{t-2-u}A(u)$ (an empty sum is $0$). This is a sum of nonnegative terms with positive weights, so $e(t)=0$ iff $A(u)=0$ for all $u\le t-2$. In particular $e(t')=0$ implies $e(t)=0$ for every $t\le t'$.
\item Equal memories are zero. Let $0\le t<t'$ with $e(t)=e(t')$. If $t'\le 1$, both are $0$. Otherwise let $M:=t'-2\ge 0$ and $\delta:=t'-t\ge 1$. Then
\[
0=e(t')-e(t)=\sum_{m=0}^{M}c_m\lambda_0^m,\qquad c_m:=A(t'-2-m)-A(t-2-m)\in\{-1,0,1\},
\]
with $A:=0$ at negative rows. Suppose some $c_m\neq 0$, and let $m^\ast$ be the largest such index. Multiplying by $5^M$ gives $\sum_{m\le m^\ast}c_m4^m5^{M-m}=0$. Every term with $m<m^\ast$ is divisible by $5^{M-m^\ast+1}$, so $5^{M-m^\ast+1}$ divides $c_{m^\ast}4^{m^\ast}5^{M-m^\ast}$; that is, $5$ divides $c_{m^\ast}4^{m^\ast}$. This is impossible, because $|c_{m^\ast}|=1$ and $5$ does not divide $4^{m^\ast}$. Hence every $c_m=0$, that is, $A(v)=A(v-\delta)$ for $0\le v\le t'-2$. For $v<\delta$ the right side is a dose at a negative row, hence $0$; by induction on $v$, $A(v)=0$ for all $0\le v\le t'-2$. So $e(t)=e(t')=0$.
\item Equivalences. $\varphi_1$ is strictly increasing, so $q(t)=q(t')$ iff $e(t)=e(t')$. Together with the two previous items this proves: $q(t)=q(t')$ iff $q(t')=0$ iff $A(u)=0$ for every $u\le t'-2$.
\item $\psi_j>0$ when $\mathcal E_j>0$. If $\mathcal E_j>0$, then $q$ is not constant on $\{d_j,\dots,d_{j+1}\}$, so $e(s)>0$ for some $s\le d_{j+1}$, and hence $e(d_{j+1})>0$ by the zero-memory item. By the equivalences, $q(d_j)\neq q(d_{j+1})$ and $\psi_j>0$.
\item When $J=0$. $J=0$ iff $\Delta_jq=0$ for every $j$ iff $Q^{(0)}=0$. By the equivalences each $\Delta_jq=0$ forces $q(d_{j+1})=0$, so $J=0$ implies $q(d_n)=0$; this also holds when $n=0$, since $q(d_0)=0$. Conversely, $q(d_n)=0$ gives $e(d_n)=0$, hence $q(d_j)=0$ for all $j\le n$, and $J=0$. Finally, $q(d_n)=0$ iff $A(u)=0$ for every $u\le d_n-2$.
\end{itemize}

\emph{(iv)} Let $J>0$. By (ii), $\mathcal C^{\rm eff}/(\sigma_bJ^{1/2})\sim N(c\Lambda\zeta J^{1/2}/\sigma_b,1)$ for the held record. For the hourly record at amplitude $c'$, $\mathcal C^{\rm eff}_{\rm hourly}/(\sigma_b(Q^\circ)^{1/2})\sim N(c'\Lambda\zeta(Q^\circ)^{1/2}/\sigma_b,1)$. Since $\Lambda\neq 0$, the two laws coincide for every $\zeta$ iff $cJ^{1/2}=c'(Q^\circ)^{1/2}$, that is, iff $c'=c\chi^{1/2}$; this is also the condition for equal Fisher information. If $J=0$, then $\Delta_jq=0$ for every $j$, the density in (ii) does not depend on $\zeta$, and the Fisher information is $0$ at every amplitude.

\emph{(v)} In the idealized model of (d), (A1) and (R) hold at every $c$, so $\mathbb A=(0,\infty)$ for every sampling.

(1) The statistic $\mathcal C^{\rm raw}=\mathcal C^{\rm eff}+m_q$ is linear in the held record for fixed design, and $J$ and $m_q$ are design-only. Its zero-effect part is $X^0=m_q+N$, where $N:=\sum_j\Delta_jb\,\Delta_jq/g_j$; its signal part is $U_X=\Lambda\zeta J$; and $X^{\mathcal S}=(1,J)$. Given $(\mathcal S,\zeta,\varepsilon)$, $N$ has mean $0$ and variance $\sigma_b^2J$. Hence $\mathrm EN=0$ and $\mathrm{Cov}(N,J)=0$, so $\mathrm{Proj}(N\mid 1,J)=0$ and $\tilde X^0=\tilde m_q+N$. The cross term $\mathrm E[\tilde m_qN]=\mathrm E[\tilde m_q\,\mathrm E[N\mid\mathcal S]]$ is $0$, so $\Sigma_0=\mathrm{Var}(\tilde m_q)+\sigma_b^2\mathrm EJ>0$.

The orthogonality conditions of (H) hold:
\begin{itemize}
\item $\mathrm E[NT]=\mathrm E\bigl[T\,\mathrm E[N\mid\mathcal S,\zeta,\varepsilon]\bigr]=0$.
\item $\tilde m_q$ is $\mathcal S$-measurable with mean $0$ and $\mathcal S\perp(\zeta,\varepsilon)$, so $\mathrm E[\tilde m_qT]=\mathrm E\,\tilde m_q\,\mathrm ET=0$. With the previous item, $\mathrm{Cov}(\tilde X^0,T)=0$.
\item By (c)(iv) with $\Xi=J\in\mathrm{span}(1,J)$, $\tilde U_X=\Lambda(\zeta-\mu_\zeta)J$, a function of $(\mathcal S,\zeta)$. Then $\mathrm E[N\tilde U_X]=0$ as in the first item, and $\mathrm E[\tilde m_q\Lambda(\zeta-\mu_\zeta)J]=\Lambda\,\mathrm E[\tilde m_qJ]\,\mathrm E(\zeta-\mu_\zeta)=0$. So $\mathrm{Cov}(\tilde X^0,\tilde U_X)=0$.
\end{itemize}
By (c)(iv), $\boldsymbol\theta=\Lambda k_T\sigma_\zeta^2\mathrm EJ\neq 0$ and $C_0=(1+\mathrm{CV}^2(J))/\rho^2_{\rm lin}$. By (c)(ii) with $p=1$, and using $\mathrm{Var}(T)=k_T^2\sigma_\zeta^2/\rho^2_{\rm lin}$,
\[
C_1=\frac{\mathrm{Var}(T)\,\Sigma_0}{\boldsymbol\theta^2}=\frac{\sigma_b^2\mathrm EJ+\mathrm{Var}(\tilde m_q)}{\Lambda^2\sigma_\zeta^2\rho^2_{\rm lin}(\mathrm EJ)^2}.
\]
$R^2_{\mathcal S}=0$ because $\mathrm{Cov}(J,T)=k_T\mathrm{Cov}(J,\zeta)+\mathrm{Cov}(J,\varepsilon)=0$. For $\mathcal C^{\rm eff}$ the term $m_q$ is absent, $\mathrm{Var}(\tilde m_q)$ is replaced by $0$, and $C_1=\sigma_b^2/(\Lambda^2\sigma_\zeta^2\rho^2_{\rm lin}\mathrm EJ)$. The hourly record is the case $J=Q^\circ$, so $C_1^{\rm held}/C_1^{\rm hourly}=\mathrm EQ^\circ/\mathrm EJ$ for the mean-corrected reader, and (c)(iii) gives the factor.

(2) Given $\mathcal S$: $\zeta\sim N(\mu_\zeta,\sigma_\zeta^2)$ is independent of the baseline; $Y_0-m(0)=b(0)$ is independent of $(\zeta,\Delta(Y-m))$; and $\mathcal C^{\rm eff}$, which is computable from $\mathcal R^{(k)}$ because $m$ is $\mathcal S$-measurable, is sufficient by (ii). The posterior of $\zeta$ given $\mathcal R^{(k)}$ is therefore its posterior given $\mathcal C^{\rm eff}$. That posterior is normal with variance $\sigma_\zeta^2/(1+\varsigma J)$, a function of $\mathcal S$ only. Hence
\[
\mathrm{Var}\,\mathrm E[\zeta\mid\mathcal R^{(k)}]/\sigma_\zeta^2=1-\mathrm E\,\mathrm{Var}(\zeta\mid\mathcal R^{(k)})/\sigma_\zeta^2=\mathrm E[\varsigma J/(1+\varsigma J)].
\]
For $X\in\{J,Q^\circ\}$ let $F_X(\varsigma):=\mathrm E[\varsigma X/(1+\varsigma X)]$. $F_X$ is continuous and $F_X(0)=0$. It is strictly increasing when $P(X>0)>0$, which follows from $\mathrm EJ>0$ and $Q^\circ\ge J$. By monotone convergence ($X$ is bounded), $F_X(\varsigma)/\varsigma=\mathrm E[X/(1+\varsigma X)]\uparrow\mathrm EX$ as $\varsigma\downarrow 0$. Hence $\varsigma_X(\varrho):=F_X^{-1}(\varrho)$ satisfies $\varsigma_X(\varrho)/\varrho\to 1/\mathrm EX$. Since $\varsigma\propto c^2$ with the same constant for both records,
\[
c_{\rm held}(\varrho)^2/c_{\rm hourly}(\varrho)^2=\varsigma_J(\varrho)/\varsigma_{Q^\circ}(\varrho)\to\mathrm EQ^\circ/\mathrm EJ .
\]

(3) $\mathcal C^m=N'+c\Lambda\zeta Q^{(0)}$ with $N':=\sum_j\Delta_jb\,\Delta_jq=\sum_j\Delta_jq\sum_{u=d_j+1}^{d_{j+1}}\xi_u$. Given $(\mathcal S,\zeta,\varepsilon)$, $N'$ has mean $0$ and variance $\sigma_b^2\sum_jg_j(\Delta_jq)^2=\sigma_b^2Q^{(1)}$. $Q^{(0)}>0$ iff some $\Delta_jq\neq 0$ iff $J>0$, so $\mathrm EJ>0$ gives $\mathrm EQ^{(0)}>0$.

The argument of (1), with $m\equiv 0$ and $(N',Q^{(0)})$ in place of $(N,J)$, gives:
\begin{itemize}
\item $\tilde X^0=N'$ and $\Sigma_0=\sigma_b^2\mathrm EQ^{(1)}$;
\item (H) holds, with $\tilde U_X=\Lambda(\zeta-\mu_\zeta)Q^{(0)}$ and $\boldsymbol\theta=\Lambda k_T\sigma_\zeta^2\mathrm EQ^{(0)}\neq 0$;
\item $R^2_{\mathcal S}=0$;
\item $C_0=(1+\mathrm{CV}^2(Q^{(0)}))/\rho^2_{\rm lin}$ by (c)(iv), with $\Xi=Q^{(0)}\in\mathrm{span}(1,Q^{(0)})$;
\item $C_1=\mathrm{Var}(T)\Sigma_0/\boldsymbol\theta^2=\sigma_b^2\mathrm EQ^{(1)}/(\Lambda^2\sigma_\zeta^2\rho^2_{\rm lin}(\mathrm EQ^{(0)})^2)$.
\end{itemize}
On hourly records $g_j=1$ and $Q^{(0)}=Q^{(1)}=Q^\circ$, so $C_1^{\rm held}/C_1^{\rm hourly}=\mathrm EQ^{(1)}\,\mathrm EQ^\circ/(\mathrm EQ^{(0)})^2$, and (c)(iii) gives the factor.

For the inequality, the Cauchy--Schwarz inequality per patient gives
\[
Q^{(0)}=\sum_j\bigl(g_j^{1/2}|\Delta_jq|\bigr)\bigl(g_j^{-1/2}|\Delta_jq|\bigr)\le(Q^{(1)}J)^{1/2},
\]
and the Cauchy--Schwarz inequality in $L^2$ then gives $\mathrm EQ^{(0)}\le\mathrm E[(Q^{(1)}J)^{1/2}]\le(\mathrm EQ^{(1)}\,\mathrm EJ)^{1/2}$. Hence $\mathrm EQ^{(1)}\mathrm EQ^\circ/(\mathrm EQ^{(0)})^2\ge\mathrm EQ^\circ/\mathrm EJ$.

\medskip\noindent\textbf{Proof of (e).} By (G2) only $j^\ast$ has nonzero factual memory, so the factual shift on channel $k$ at row $d$ is $c\,\alpha\tanh(\Lambda_k\zeta_iq_i(d))$. On $\mathcal N_i$ this is $0$ at every draw $d\in\mathcal D_{ik}$ of every channel, because $\Lambda_k=0$ or $q_i(d)=0$; this uses no property of $\alpha$. Hence $y_{ik}(d)=\tilde y_{ik}(d)$ at every draw, and by (G1) and (G4) $Z_{ik}(t)=\tilde y_{ik}(d_{ik}(t))=Z^0_{ik}(t)$ for all $k$ and $t\le 89$. So $\mathcal R_i\mathbf 1_{\mathcal N_i}=(Z^0_i,\mathcal S_i)\mathbf 1_{\mathcal N_i}$.

For bounded measurable $f$ and $\phi$, since $\mathcal N_i$ is determined by $\mathcal S_i$ and by (G3),
\[
\mathrm E\bigl[f(\zeta_i)\,\phi(\mathcal R_i)\mathbf 1_{\mathcal N_i}\bigr]=\mathrm E\bigl[f(\zeta_i)\,\phi(Z^0_i,\mathcal S_i)\mathbf 1_{\mathcal N_i}\bigr]=\mathrm Ef(\zeta_i)\ \mathrm E\bigl[\phi(\mathcal R_i)\mathbf 1_{\mathcal N_i}\bigr].
\]
Since $\mathbf 1_{\mathcal N_i}$ is $\sigma(\mathcal R_i)$-measurable, this says $\mathbf 1_{\mathcal N_i}\mathrm E[f(\zeta_i)\mid\mathcal R_i]=\mathbf 1_{\mathcal N_i}\mathrm Ef(\zeta_i)$ almost surely. With $f=\bar\tau$ (bounded) this is the second claim.

Characterization. Under binary dosing, $q_i(d)=0$ iff $e_i(d)=0$ iff $A_i(u)=0$ for every $u\le d-2$, by the zero-memory item in the proof of (d)(iii). Moreover $e_i(d')=0$ implies $e_i(d)=0$ for every $d\le d'$. So, for each $k$, $q_i$ vanishes at every draw of channel $k$ iff it vanishes at $\max\mathcal D_{ik}$, which gives the stated description of $\mathcal N_i$.

Now let $P(\mathcal N_i)=1$. Write $\tau=\bar\tau(\zeta)+\nu'$ with $\nu':=\tau-\bar\tau(\zeta)$, so that $\mathrm E\nu'^2=\mathrm E\,\mathrm{Var}(\tau\mid\zeta)$. Then $\mathrm E[\tau\mid\mathcal R]=\mathrm E\,\bar\tau(\zeta)+\mathrm E[\nu'\mid\mathcal R]$, and by Jensen's inequality $\mathrm{Var}\,\mathrm E[\tau\mid\mathcal R]=\mathrm{Var}\,\mathrm E[\nu'\mid\mathcal R]\le\mathrm E\nu'^2$. Combined with $R^2(\hat\tau)\le R^2_\ast(\sigma(\mathcal R))$ from the proof of (c)(i), this gives $R^2(\hat\tau)\le\mathrm E\,\mathrm{Var}(\tau\mid\zeta)/\mathrm{Var}(\tau)$, with every moment computed for the panel in question. With no dose on rows $0,\dots,89$, $e_i(t)=0$ for $t\le 90$, so $q_i\equiv 0$ on rows $0,\dots,89$, $e_i(t_0)=0$ and $A_i(89)=0$. Then (a)(ii), under (A1), gives the own-treatment part $\tau^{\rm own}(\zeta_i,0)$. $\blacksquare$

\paragraph{Numerical checks.} All checks below are descriptive re-reads of shipped panel outputs, reconstructions of the panels' doses from shipped inputs, or simulations; none reads a new label.
\begin{itemize}
\item \emph{Target structure (a).} A replica of the generator reproduces the shipped targets of the synthetic-record run (job 10692), which are the same targets as in the real-record discovery panels, to $8.5\times10^{-8}$. $D^\ast=0.6842$, and the $s=0$ increment is exactly $0$. In the registered phenylephrine cell $\mathrm{SD}(\tau)=0.2322$, $\mathrm{corr}(\tau,\zeta)=-0.964$ and the $R^2$ of $\tau$ on $\tau^{\rm own}$ is $0.941$; $\tau-\tau^{\rm own}$ correlates $+0.559$ with $\zeta$, which is why $\mathrm E[\tau\mid\zeta]$ predicts better than $\tau^{\rm own}$ alone. The cross-fitted $R^2$ of an estimate of $\mathrm E[\tau\mid\zeta]$ is $0.959$ to $0.960$ (recovery $0.797$ to $0.800$), and the shipped oracle rung scores $0.802$ to $0.805$; this is the reference level of 0.80 in the main text. The target standard deviations of the discovery, replication and fixed-schedule panels are $0.2322$, $0.2317$ and $0.2323$, which is consistent with (a)(ii)--(iii); those parts bound only the own-treatment part and do not predict the standard deviations. The largest retained memory at the decision is $2.03\times10^{-3}$ under NATMIN, so (a)(iii) bounds the change of the own-occupancy increment by $6.3\times10^{-4}$; under FIX19 the largest retained memory is $3.09\times10^{-6}$.
\item \emph{Exact scaling (b).} In a 240-person scaling check the held heart-rate record is affine in $c$ to $2.6\times10^{-6}$ and $T$ to $1.0\times10^{-6}$; the exceptions to (A1) are listed under (A1).
\item \emph{Amplitude law (c).} Two-point fits through the shipped recovery map at $c=0.5$ and $c=1$ predict the value at $c=0.75$ with mean absolute error $0.007$ (maximum $0.014$; 17 series) when the fitted values are first corrected for the expected finite-sample loss of (P), and $0.006$ (13 series) without the correction; the mean signed errors are $+0.004$ and $-0.004$. Extrapolated to $c=0.25$, the mean signed errors are $-0.019$ (corrected; 21 series) and $+0.014$ (uncorrected; 16 series). By (c)(v) the population law gives two-point predictions that are too high inside $[0.5,1]$ and too low outside; the corrected fits show these signs and the uncorrected fits do not. (The correction uses the expression $(1-R^2)p/(n-p-1)$, which is 4\% to 8\% smaller than the exact value in (P), so it slightly under-corrects.) In the registered cell the corrected fit predicts recovery $0.185$ at $c=0.75$ ($0.177$ uncorrected) against the observed $0.181$ (95\% interval $0.137$ to $0.221$). The chord crossing amplitude $c^\ast$ of (c)(iii) is $0.53$ from the uncorrected fit and $0.44$ from the corrected fit; the observed recovery at $c=0.5$ is $0.095$, below the bar of $0.106$. In 2 corrected and 7 uncorrected effect-scale fits the estimated $\hat C_0$ is below 1, which is impossible for the population constant and reflects estimation error. A series is one reader (window-mean, assimilating, plain or cadence-aware) on one treatment-response pair at one cadence. It enters a count when the map has recoveries at $c=0.5$, at $c=1$ and at the predicted amplitude, and when the two fitted $R^2$ gains, after the correction where it is used, both exceed $0.005$, so that the line in $1/c^2$ is defined. The correction raises every gain, so more series qualify with it (17 against 13 for interpolation, 21 against 16 for extrapolation), and not every reader was run at every amplitude (the cadence-aware reader on hourly records only at $c=1$, and at $c=0.75$ for phenylephrine). Interpolation inside $[0.5,1]$ is a weak test, because any smooth increasing curve passes close to its chord there; the size and sign of the extrapolation errors are the more informative check, and the uncorrected fits have the sign opposite to the one the law predicts.
\item \emph{Ceilings.} By (c)(v), two-point ceilings bound the population ceiling from below before sampling error. For the plain reader the corrected chord ceilings fall from $0.79$ (hourly) to $0.53$ (real cadence) for phenylephrine on heart rate, from $0.79$ to $0.36$ for norepinephrine on heart rate, and from $0.86$ to $0.13$ for NaCl on creatinine. The single-channel linear-regime ceiling of (c)(iv) at $K_0$ changes much less on heart rate: from $0.855$ to $0.788$ ($0.774$ at the panel's change rate) for phenylephrine and from $0.792$ to $0.731$ ($0.718$) for norepinephrine for the plain reader, and from $0.822$ to $0.825$ for the cadence-aware reader. The real-cadence plain chord ceilings on heart rate lie below this single-channel value but within sampling error (bootstrap probability that the chord ceiling is at least the theory value: $0.31$ for phenylephrine, $0.17$ for norepinephrine), so on heart rate the sampling acts mainly through $C_1$ for both readers. The theory predicts a ceiling collapse only for the plain reader on creatinine, where $1+(1+\mu_\zeta^2/\sigma_\zeta^2)\mathrm{CV}^2(M)$ is heavy-tailed across seeds (28 to 42 at the proxy change rates and 71 to 161 at the panel's rate), giving a single-channel ceiling of about $0.006$ to $0.03$, while the cadence-aware single-channel ceiling on creatinine is $0.54$ to $0.60$. The multi-channel creatinine chord ceiling ($0.13$ corrected, $0.32$ uncorrected) does not settle whether the population ceiling lies below the bar: the bootstrap probability that it is below $0.2$ is $0.75$ (corrected) and $0.56$ (uncorrected).
\item \emph{Amplitude factors.} The low-signal amplitude factors from hourly to real cadence estimated from the map are $2.68$, $2.44$ and $2.20$ for the plain reader on the three pairs, and $1.64$ (corrected; $1.90$ uncorrected) for the cadence-aware reader on phenylephrine and heart rate, with 90\% Monte Carlo ranges $[1.18,2.54]$ and $[1.34,3.10]$. The cadence-aware estimate mixes kinetics, because the hourly fits selected $(2,0.95)$ and the held fits $(2,0.8)$ (assumption (K)), so it does not test (c)(iii) cleanly. Table~\ref{tab:P4ret} gives the single-carrier theory values; the map's $1.64$ lies between the implemented statistic's white-noise and random-walk values.
\item \emph{Formula (c)(iv).} A Monte Carlo of the plain reader gives $C_0=2.254$ against $2.252$ from the formula with the $\mu_\zeta$ term and $1.962$ without it.
\item \emph{Retention (d).} Table~\ref{tab:P4ret} gives retention under the idealization (RW) with Bernoulli draw patterns at each real stay's own change rate (400 stays of the realism diagnostic), reported at these proxy rates and rescaled to the panels' own mean rates. The identity $\chi=\sum_jw_j\psi_j$ holds to $8\times10^{-16}$. The alignment factors of intervals with $\mathcal E_j>0$ are at least $1.6\times10^{-9}$ on heart rate and $1.1\times10^{-13}$ on creatinine, positive as (d)(iii) requires; the zero-memory lemma was also verified exhaustively over 8,190 short binary dose sequences and on 300 random 90-row sequences in exact rational arithmetic.
\item \emph{Exposure (e).} In a replica panel with no pre-decision dosing, $Z^A=Z^0$ exactly and a label-free transport predictor is constant (centered PEHE divided by $\mathrm{SD}(\tau)$ equals $1.000$), while the target still correlates $0.9986$ with the NATMIN target. The bound of (e) must be computed with that panel's own moments; the NATMIN value of $R^2_{\rm trait}$ is not a bound for it.
\end{itemize}

\begin{table}[ht]
\centering\small
\caption{\textbf{Retention of trait information and amplitude factors on a single carrier (Proposition 4(d)).} Idealized single-carrier random-walk baseline (assumption (RW)), generator kinetics, the panels' reconstructed natural doses (phenylephrine for heart rate, NaCl for creatinine) and Bernoulli draw patterns at each real stay's own change rate (proxy; mean rate 0.677 for heart rate and 0.046 for creatinine), then rescaled to the panels' own mean rates (0.572 and 0.037). Retention $\chi=J/Q^\circ$ is the share of the hourly record's Fisher information about the trait that the held record keeps. The low-signal amplitude factor is the ratio of the amplitudes that give held and hourly records the same small $R^2$: $(\mathrm EQ^\circ/\mathrm EJ)^{1/2}$ for the efficient statistic (Proposition 4(d)(v)(1)), and $(\mathrm EQ^\circ\,\mathrm EQ^{(1)})^{1/2}/\mathrm EQ^{(0)}$ for the implemented cadence-aware statistic under a random walk (Proposition 4(d)(v)(3)); the white-noise value for the implemented statistic is from Proposition 5. These are properties of the idealized model, not measurements of the panels.}
\label{tab:P4ret}
\begin{tabular}{lrrrr}
\toprule
 & \multicolumn{2}{c}{Heart rate} & \multicolumn{2}{c}{Creatinine}\\
 & proxy & panel rate & proxy & panel rate\\
\midrule
Median retention $\chi$ & 0.806 & 0.607 & 0.048 & 0.036\\
Mean retention $\chi$ & 0.702 & 0.610 & 0.070 & 0.058\\
Patients with $\chi<0.05$ (\%) & 2 & 2 & 51 & 59\\
Energy in intervals with $\psi_j<0.1$ (\%) & 10.8 & 13.7 & 72.1 & 70.6\\
Energy after the last draw (\%) & 0.5 & 0.6 & 12.8 & 17.3\\
Patients with $J=0$ (\%) & 0.4 & 0.4 & 8.6 & 12.4\\
Amplitude factor, efficient statistic & 1.20 & 1.29 & 4.09 & 4.50\\
Amplitude factor, implemented statistic, random walk & 1.84 & 1.97 & 5.25 & 5.71\\
Amplitude factor, implemented statistic, white noise & 0.97 & 0.96 & 1.15 & 1.22\\
\bottomrule
\end{tabular}
\end{table}

%% file: supplementary/supp_theory/p5.tex
\paragraph{Proposition 5 (Levels or increments: the baseline decides which reader is efficient).}
\emph{Standard and specific parts.} Parts (a)(i) and (a)(ii) are the Gauss--Markov theorem for generalized least squares with a nuisance mean (Aitken), and (b) is its white-noise case, in which partialling out the trend (Frisch and Waugh) gives the plain contrast. Parts (a)(iii) and (c) to (e) are specific to the pooled readers and the held records of this evaluation, and (d) is partly numerical. The weight vectors $v$ of this section are unrelated to the slope variance $v$ of Sections 5.6 and 5.7.

\textit{Setting.} Symbols not defined here are those of the notation and definitions at the start of this note. Assumptions N1 to N4 hold throughout; N5, N6 and N7 are used where stated. Readers use the generator kinetics $K_0=(2,0.8,1)$, so that $p_{K_0}=q_i$. For a set $\mathcal K$ of outcome channels, $y_i^{\mathcal K}$ stacks the latent paths of those channels on rows $t=0,\dots,89$, and $\bar q_i^{\mathcal K}$ stacks the vectors $\Lambda_kq_i$, $k\in\mathcal K$; by N1, $y_i^{\mathcal K}=b_i^{\mathcal K}+c\,\zeta_i\,\bar q_i^{\mathcal K}$. For one channel $k$ with loading $\Lambda:=\Lambda_k\neq0$ the superscript is dropped, so $y_i=b_i+c\Lambda\zeta_iq_i$ and $\bar q_i=\Lambda q_i$. The held record of channel $k$ is $Z_i=H_iy_i$, where the $90\times90$ hold matrix has $H_i(t,d_{ik}(t))=1$ and is zero elsewhere; on hourly records $H_i=I$. A linear statistic is $v_i^\top y_i^{\mathcal K}$ with a weight vector $v_i$ that is a measurable function of the design $\mathcal S_i$; a statistic of channel $k$'s held record has $v_i=H_i^\top w_i$, which is supported on the draw set $\mathcal D_{ik}$. Let $\mu_i:=\mathrm E[b_i^{\mathcal K}\mid\mathcal S_i]$ and let $\Sigma_i:=\operatorname{Cov}(b_i^{\mathcal K}\mid\mathcal S_i)$, including the covariances between channels, on the rows the statistic reads, positive definite there. Given $(\mathcal S_i,\zeta_i)$, $v_i^\top y_i^{\mathcal K}$ has mean $v_i^\top\mu_i+c\,\zeta_i\,v_i^\top\bar q_i^{\mathcal K}$ (its signal is $c\,\zeta_i\,v_i^\top\bar q_i^{\mathcal K}$) and variance $v_i^\top\Sigma_iv_i$.

\textit{Notation.} For a matrix $B$ of full column rank, possibly with no columns, $M_B:=\Sigma^{-1}-\Sigma^{-1}B(B^\top\Sigma^{-1}B)^{-1}B^\top\Sigma^{-1}$, and $M_B:=\Sigma^{-1}$ when $B$ has no columns. For the draws $d_0=0<d_1<\dots<d_n$ of channel $k$: $\Delta_jf:=f(d_{j+1})-f(d_j)$, $g_j:=d_{j+1}-d_j$, $Q^{\circ}:=\sum_{t=1}^{89}(q(t)-q(t-1))^2$, $Q^{(m)}:=\sum_{j=0}^{n-1}g_j^{m}(\Delta_jq)^2$ for $m\in\{-1,0,1\}$, and $J:=Q^{(-1)}$. $\nabla$ is the $89\times 90$ first-difference matrix, $(\nabla x)(t)=x(t)-x(t-1)$; $\mathbf t:=(0,1,\dots,89)^\top$ and $\mathbf t^2$ is its elementwise square. Plain statistic $\mathcal O:=(\Pi q)^\top Z$ with $\Pi:=I-B_2B_2^{+}$ and $B_2:=[\mathbf 1,\mathbf t,\mathbf t^2]$ (no companion feature); cadence-aware statistic $\mathcal C:=\sum_j\Delta_jZ\,\Delta_jq$ with companion feature $Q^{(0)}$; $1/$gap-weighted statistic $\mathcal G:=\sum_j\Delta_jZ\,\Delta_jq/g_j$ with companion feature $J$. A \emph{pooled linear reader} of the channels $\mathcal K$ is the population least-squares regression of $T:=\tau/c$ on $(1,F^{\mathcal S}_i,X_i)$, where $F^{\mathcal S}_i$ are design-only features and $X_i=V_i^\top y_i^{\mathcal K}\in\mathbb R^p$ with $V_i$ design-measurable and $\mathrm E[V^\top\Sigma V]$ nonsingular; $R^2(c)$ is its population $R^2$ at amplitude $c$ and $1/C_1:=\lim_{c\to0}R^2(c)/c^2$. Put $\Upsilon:=k_T^2\sigma_\zeta^4/\operatorname{Var}(T)$, with $k_T$ from N4 and $\mu_\zeta$, $\sigma_\zeta^2$ the mean and variance of the trait, and, for $p=1$ and $X=v^\top y^{\mathcal K}$,
\[
\mathcal I(v):=\frac{(\mathrm E[v^\top\bar q])^2}{\mathrm E[v^\top\Sigma v]+V_m(v)},
\]
where $V_m(v)$ is the residual variance of the design-measurable offset $v^\top\mu$ after linear regression on $(1,F^{\mathcal S})$. The amplitude factor of a reader from hourly to held records of the same baseline is $f:=(C_1^{\mathrm{held}}/C_1^{\mathrm{hourly}})^{1/2}$. Unconditional expectations $\mathrm E$ are over patients.

\begin{enumerate}
\item[(a)] \textit{Gauss--Markov bounds.}
(i) Per patient, baseline mean known: for $v\neq 0$, $(v^\top\bar q)^2/(v^\top\Sigma v)\le\bar q^\top\Sigma^{-1}\bar q$, with equality iff $v$ and $\Sigma^{-1}\bar q$ are linearly dependent. When $\bar q\neq0$ on the rows read this means $v=\alpha\Sigma^{-1}\bar q$ for some $\alpha\neq0$; when $\bar q=0$ there, both sides are $0$ for every $v$.
(ii) Per patient, $\mu\in\operatorname{span}(B)$ with free coefficients: the linear statistics whose law does not depend on the free coefficients are exactly those with $B^\top v=0$. For them $(v^\top\bar q)^2/(v^\top\Sigma v)\le\bar q^\top M_B\bar q$, with equality iff $v$ and $M_B\bar q$ are linearly dependent, that is, when $M_B\bar q\neq0$, iff $v=\alpha M_B\bar q$ with $\alpha\neq0$. For Gaussian $b$, $c^2\,\bar q^\top M_B\bar q$ is the Fisher information for $\zeta$ with the free coefficients unknown, and generalized least squares attains it. The per-patient efficiency of $v$ is $\mathrm{eff}(v):=(v^\top\bar q)^2/(v^\top\Sigma v\cdot\bar q^\top M_B\bar q)\in[0,1]$ when $\bar q^\top M_B\bar q>0$, and $\mathrm{eff}(v):=0$ otherwise.
(iii) Pooled: if $B_i^\top V_i=0$ almost surely, then $R^2(c)\le c^2\,\Upsilon\,\mathrm E[\bar q^\top M_B\bar q]$ at every amplitude $c$, where $\Sigma$ is the joint covariance of the channels read. For $p=1$, $1/C_1=\Upsilon\,\mathcal I(v)$ and $\mathcal I(v)\le\mathrm E[\bar q^\top M_B\bar q]$; if $\mathrm E[\bar q^\top M_B\bar q]=0$ both sides are $0$, and otherwise equality holds iff $V_m(v)=0$ and $v=\alpha M_B\bar q$ almost surely for one constant $\alpha\neq0$. When $V_m(v)=0$ and $\Pr(a\neq0)>0$, with $a:=v^\top\bar q$ and $s_v^2:=v^\top\Sigma v$,
\[
\frac{\mathcal I(v)}{\mathrm E[\bar q^\top M_B\bar q]}=\frac{\mathrm E[\mathrm{eff}(v)\,\bar q^\top M_B\bar q]}{\mathrm E[\bar q^\top M_B\bar q]}\cdot\frac{(\mathrm E a)^2}{\mathrm E[a^2/s_v^2]\,\mathrm E[s_v^2]},
\]
with $a^2/s_v^2:=0$ where $s_v=0$; the second factor is at most $1$, with equality iff $a/s_v^2$ is almost surely constant on $\{s_v>0\}$. If the nuisance basis is block-diagonal across channels, the per-patient bound $\bar q^{\mathcal K\top}M_B\bar q^{\mathcal K}$, and hence the pooled bound, is nondecreasing in the set $\mathcal K$ of channels read. These are bounds for pooled linear readers; they are not bounds on estimators.

\item[(b)] \textit{White noise plus a free quadratic trend (N5), hourly records of one channel.} Here $M_{B_2}=\sigma^{-2}\Pi$, so the plain statistic $(\Pi q)^\top y$ is efficient among trend-invariant linear statistics of the channel, and $(\Pi q)^\top y/\lVert\Pi q\rVert^2$ is the generalized least-squares estimate of $c\Lambda\zeta$. At kinetics $K$ the plain statistic $(\Pi p_K)^\top y$ has efficiency $(p_K^\top\Pi q)^2/(p_K^\top\Pi p_K\cdot q^\top\Pi q)$. The hourly increment statistic $\sum_{t=1}^{89}(\nabla y)(t)\,(\nabla q)(t)=(\nabla^\top\nabla q)^\top y$ is not efficient. For every dosing of the scored treatment with doses in $[0,1]$ (the implemented doses are binary) on rows 6 to 53 only and at least one nonzero dose, $(\mathbf t^2)^\top\nabla^\top\nabla q=177\,q(89)-2\sum_{s=1}^{88}q(s)<0$, so the law of the increment statistic depends on the free quadratic coefficient; and with the mean known its efficiency $(Q^\circ)^2/(\lVert\nabla^\top\nabla q\rVert^2\,\lVert q\rVert^2)$ is strictly below $1$.

\item[(c)] \textit{Random-walk baseline (N6), one channel.}
(i) The level-invariant linear statistics of the held record are exactly the linear functions of the draw-to-draw increments. Among them $\mathcal G$ is efficient, and on the draw rows $\bar q_{\mathcal D}^\top M_{\mathbf 1}\bar q_{\mathcal D}=\Lambda^2J/\sigma_b^2$ for every positive finite level variance.
(ii) $\mathcal C$ reads the held record only through $\Delta_jZ$, which equals the latent increment $\Delta_jy$; the implemented reader finds the draws as the rows where the recorded value changes, which recovers $\mathcal D_{ik}$ except for exact ties. The hold therefore acts on $\mathcal C$ only through the draw set. If $Q^{(0)}>0$, the per-patient efficiency of $\mathcal C$ is $(Q^{(0)})^2/(Q^{(1)}J)\le1$, with equality iff $g_j$ is constant on $\{j:\Delta_jq\neq0\}$; on hourly records $\mathcal C=\mathcal G$.
(iii) The pooled readers of the channel on $(\mathcal C,Q^{(0)})$ and on $(\mathcal G,J)$ satisfy $1/R^2(c)=C_0+C_1/c^2$ exactly, with $1/C_1=\Upsilon\Lambda^2(\mathrm EQ^{(0)})^2/(\sigma_b^2\,\mathrm EQ^{(1)})$ and $1/C_1=\Upsilon\Lambda^2\,\mathrm EJ/\sigma_b^2$ respectively. The latter is the largest $1/C_1$ of any pooled linear reader of this channel's held record whose statistics are level-invariant. From hourly records of the same baseline to held records,
\[
f_{\mathcal C}=\Big(\frac{\mathrm EQ^{(1)}\,\mathrm EQ^{\circ}}{(\mathrm EQ^{(0)})^2}\Big)^{1/2}\ \ge\ f_{\mathcal G}=\Big(\frac{\mathrm EQ^{\circ}}{\mathrm EJ}\Big)^{1/2}.
\]

\item[(d)] \textit{Pooled ordering and the AR(1) switch on hourly records.} For pooled linear readers with single statistics $v_1,v_2$ and $\mathcal I(v_2)>0$, $R_1^2(c)/R_2^2(c)\to\mathcal I(v_1)/\mathcal I(v_2)$ as $c\to0$; if $\mathcal I(v_1)>\mathcal I(v_2)$, reader 1 has the larger $R^2$ at every sufficiently small amplitude. Under N7 on hourly records of one channel let $\mathcal R(\vartheta):=\mathcal I(\Pi q)/\mathcal I(\nabla^\top\nabla q)$, plain over cadence-aware. Then $\mathcal R$ is continuous on $[0,1)$; $\mathcal R(\vartheta)\to\mathcal R_{\mathrm{RW}}$ as $\vartheta\uparrow1$, where $\mathcal R_{\mathrm{RW}}$ is its value under N6 with every row a draw; and $\mathcal R_{\mathrm{RW}}<1$ whenever $\Pr\{(\Pi q)(0)\neq0\}>0$. For the reconstructed discovery doses $\mathcal R(0)=9.22$ and $\mathcal R_{\mathrm{RW}}=0.110$, so $\mathcal R-1$ has a zero in $(0,1)$ and the more informative reader on hourly records changes from the plain to the cadence-aware reader as $\vartheta$ rises. Numerically, on grids of step $0.01$ and $0.001$, $\mathcal R$ is strictly decreasing and $\mathcal R-1$ changes sign once, between $0.642$ and $0.643$; bisection gives $\vartheta^*=0.642$ (population detrended lag-1 autocorrelation $0.583$), and at the calibrated persistence $0.754$ (N7) $\mathcal R=0.55$. For the shared FIX19 schedule the same computation gives $\mathcal R(0)=23.57$, $\mathcal R_{\mathrm{RW}}=0.254$, one sign change between $0.806$ and $0.807$, and $\mathcal R=1.4$ at persistence $0.754$. These statements concern hourly records; they do not determine the ordering on held records.

\item[(e)] \textit{Shared schedule.} Let $\mathbf Z_i\in\mathbb R^{90\times12}$ be the record of all twelve channels on rows 0 to 89 (hourly or held). If all patients receive the same doses, then at every fixed kinetics the features of the window-mean, assimilating and plain readers are one affine map of $\mathbf Z_i$, the same for all patients, on hourly and held records alike. Their population least-squares $R^2$ is therefore at most that of the history-blind reader on all 1,080 values of $\mathbf Z_i$, at every amplitude and without N1; with kinetics chosen from a finite grid the same holds for the best kinetics; and with a next-draw target, where the same draw-step indicator is appended to every rung including the history-blind one, the same holds. On hourly records without exact ties the cadence-aware features are also such a map. Under a shared schedule a population advantage of a record-based reader of the ladder (window-mean, assimilating, plain, cadence-aware) over the history-blind reader can therefore arise only from patient-specific draw sets, which exist only on held records. On hourly records under a shared schedule, weights that depend on the design only through the doses are the same for all patients; under N1 and N2, with $\Sigma:=\operatorname{Cov}(b^{\mathcal K})$ and a common nuisance basis $B$, the efficient such weight $M_B\bar q^{\mathcal K}$ is one fixed vector for every set of channels $\mathcal K$, and the efficient statistic is a fixed linear functional of $\operatorname{vec}(\mathbf Z_i)$.
\end{enumerate}

\paragraph{Assumptions.}
\begin{description}
\item[N1 (linear regime, fixed kinetics).] On rows $0,\dots,89$ the factual latent path of every outcome channel $k$ is exactly $y_{ik}(t)=b_{ik}(t)+c\,\alpha_{ik}(t)\tanh(\Lambda_k\zeta_iq_i(t))$, because only $j^*$ is dosed in the factual record, so that $\Gamma_{ik}=\Lambda_k\zeta_iq_i$. The linear regime sets $\alpha\equiv1$ and replaces $\tanh x$ by $x$; then $y_i^{\mathcal K}=b_i^{\mathcal K}+c\zeta_i\bar q_i^{\mathcal K}$ for every set of channels. Readers use the generator kinetics $K_0$, at which the implemented occupancy $e_t/(1+e_t)$ with $e_t=0.8e_{t-1}+A_{t-2}$ equals $q_i$. \emph{Idealizations:} on heart rate $|\Lambda\zeta q|$ reaches values near $5/3$ ($\zeta\le2$, $q<5/6$), where $\tanh$ is far from linear; the ladder chooses kinetics by nested cross-validation and fits a ridge, whereas population least squares at fixed kinetics is analyzed (Definition 6). Parts (b) to (d) concern statistics of one channel, whereas the implemented plain and cadence-aware readers read all twelve channels (12 and 24 features). Two features of the generator then change the numbers for the implemented readers, although not the statements (a)(iii) and (e), which cover several channels: the phenylephrine response is also written into pH, platelets and white cells (loadings about $0.75$, $-0.80$, $-0.22$), and the channels' baselines are correlated, because every channel's generator contribution contains the shared latent state and the real tapes are measured jointly (the generator part of each other channel correlates with that of heart rate at $0.24$ to $0.64$ in absolute value).
\item[N2 (exogeneity).] $\zeta_i$ is independent of $(\mathcal S_i,b_i^{\mathcal K})$ for every $\mathcal K$. The design is trait-free (Definition 5, verified by reconstructing all dose columns from shipped $u_0$); $b_i$ contains no trait because the bounded shift never feeds back into state, policy or memory, and the trait is drawn from a separate pseudo-random stream.
\item[N3 (second moments and nuisance).] Patients are independent and identically distributed, and $\Sigma_i$ is positive definite on the rows read. For per-patient statements the design is held fixed and $\mu_i$ is known up to $\operatorname{span}(B_i)$ with free coefficients. Pooled statements keep the offset variance $V_m$. In the generator $\mu_i$ contains the scored treatment's ordinary, trait-free action route, a time-varying function of the doses, so $V_m>0$ unless a reader's design-only features absorb it; the numerical AR(1) and random-walk comparisons omit it except where stated.
\item[N4 (target).] $T:=\tau/c$, free of $c$ by P4(a), satisfies $T=\mathrm{const}+k_T\zeta+\varepsilon$ with $k_T:=\operatorname{Cov}(T,\zeta)/\sigma_\zeta^2\neq0$, $\operatorname{Cov}(\zeta,\varepsilon)=0$, $\operatorname{Var}(T)>0$, and $(\zeta,\varepsilon)$ independent of $(\mathcal S,b^{\mathcal K})$. \emph{Idealization:} $\varepsilon$ contains the nonlinear part of the target and the policy-dosed cross-treatment term $\nu$, which depends on the latent state at the decision and hence weakly on the pre-decision baseline; the retained memory $e(t_0)\le2.03\times10^{-3}$ is a design quantity inside $T$.
\item[N5 (used in (b)).] Hourly rows $0,\dots,89$ of one channel; $b=B_2\beta+b^\circ$ with $\beta\in\mathbb R^3$ free and $\operatorname{Cov}(b^\circ\mid\mathcal S)=\sigma^2I$. This is the model under which the plain reader's projector is efficient; no shipped panel follows it exactly.
\item[N6 (used in (c)).] One channel. Given $\mathcal S$, the hourly increments $b(t)-b(t-1)$ are uncorrelated with mean $0$ and variance $\sigma_b^2$, and the level $b(0)$ is a free nuisance with any positive finite variance, uncorrelated with the increments (level-invariant statistics do not depend on it). The draw-to-draw increments are then uncorrelated with variance $\sigma_b^2g_j$ (Proposition~4(d)). The ``hourly record of the same baseline'' observes this $b$ at every row. \emph{Idealizations:} real heart rate is not a random walk (N7 calibrates its persistence at $0.754$); the ordinary treatment route gives the increments a nonzero mean; and the draw set is treated as independent of $b$, whereas in the shipped panels the draw set is the tape's change set, so the tape component of $b$ is constant between draws and the variance of its jumps need not grow with the gap. Statistics of the held record depend only on the draw values. The shipped real-hourly panel (hold off) observes a baseline whose tape component is piecewise constant between the tape's change times; only the generator's contribution moves every hour.
\item[N7 (used in (d)).] Hourly rows of one channel; $\Sigma(t,t')=\sigma_{\mathrm{AR}}^2\vartheta^{|t-t'|}/(1-\vartheta^2)$ with $\vartheta\in[0,1)$ and innovation variance $\sigma_{\mathrm{AR}}^2$, the same for all patients, and a constant mean, so $V_m=0$ for level-invariant statistics. The family interpolates between white noise ($\vartheta=0$) and a random walk ($\vartheta\uparrow1$). Calibration: the shipped pooled lag-1 autocorrelation of the quadratically detrended zero-effect heart-rate record on hourly real records is $0.687$ (real-cadence probe with NaCl dosing, in which heart rate carries no response; \texttt{g2\_washout\_real\_cadence\_probe\_20260918\_v1/gls.json}, entry \texttt{real\_hold0}, \texttt{projected}, \texttt{heart\_rate}, \texttt{lag1}). The population version of the same pooled estimator (ratio of expected numerator and denominator) equals $0.687$ under N7 at persistence $\vartheta=0.754$. The corresponding creatinine value $0.602$ gives $0.662$. No calibration is taken from held records, which are a zero-order hold of the latent path and not an hourly AR(1).
\item[Data for the numerical parts.] Doses: the 400 NATMIN dose columns of the discovery panel, reconstructed from the shipped $u_0$ (400 of 400 dosed-row counts equal the shipped table); FIX19 doses on rows 6 to 24. Draw sets for the held-record numbers: row 0 plus independent Bernoulli draws at each stay's own heart-rate change rate from the realism diagnostic's real eICU stays (382 stays with a rate; mean $0.677$, median $0.713$), a proxy for the panels' own draw masks, which are not shipped. Heart rate changes in 57\% of hours on the panel stays (main text), so every number is also given with the rates rescaled by a common factor to a mean of $0.58$.
\end{description}

\paragraph{Proof.}
\paragraph{Conventions.} For random vectors $X$ and $F$ with finite second moments, $L(X\mid1,F)$ is the population linear projection of $X$ on $(1,F)$ and $\tilde X:=X-L(X\mid1,F)$. Unconditional $\mathrm E$ averages over patients. All second moments below are finite: $0\le q<5/6$, every weight vector used is a bounded function of the design, and $\Sigma_i$ is bounded. When a statistic reads only a design-dependent set of rows (all rows, or the draw rows of each channel read), $v$ is supported on that set and $\bar q,\Sigma,B,M_B$ are restricted to it. The memory of the scored treatment is $e(t)=\sum_{s\le t-2}0.8^{\,t-2-s}A(s)$ with $A(s)=0$ for $s<0$, and $q=e/(1+e)$; in particular $e(0)=e(1)=0$, hence $q(0)=q(1)=0$, for every dosing.

\medskip\noindent\textbf{Lemma 1 (Gauss--Markov with a free nuisance).}
Let $\Sigma\in\mathbb R^{N\times N}$ be positive definite, $\bar q\in\mathbb R^N$, $B\in\mathbb R^{N\times n_B}$ of rank $n_B\ge0$, and $M_B$ as in the proposition.
(1) $B^\top M_B=0$, $M_B\Sigma M_B=M_B$ and $M_B\succeq0$.
(2) If $B^\top v=0$ then $v^\top\bar q=v^\top\Sigma M_B\bar q$.
(3) If $v\neq0$ and $B^\top v=0$ then $(v^\top\bar q)^2\le(v^\top\Sigma v)(\bar q^\top M_B\bar q)$, with equality iff $v$ and $M_B\bar q$ are linearly dependent. If $M_B\bar q=0$ both sides are $0$; if $M_B\bar q\neq0$, equality means $v=\alpha M_B\bar q$ with $\alpha\neq0$, and $B^\top M_B\bar q=0$. Hence $\bar q^\top M_B\bar q=\sup\{(v^\top\bar q)^2/(v^\top\Sigma v):v\neq0,\ B^\top v=0\}$, with the supremum of the empty set read as $0$.
(4) In the model $y=B\beta+\theta\bar q+e$ with $\mathrm Ee=0$ and $\operatorname{Cov}e=\Sigma$, the law of $v^\top y$ does not depend on $\beta\in\mathbb R^{n_B}$ iff $B^\top v=0$.
(5) If $e$ is Gaussian, the Fisher information for $\theta$ with $\beta$ unknown is $\bar q^\top M_B\bar q$, and, when it is positive, $\hat\theta:=\bar q^\top M_By/\bar q^\top M_B\bar q$ is unbiased with variance $1/\bar q^\top M_B\bar q$.

\emph{Proof.}
Let $P:=\Sigma^{-1/2}B(B^\top\Sigma^{-1}B)^{-1}B^\top\Sigma^{-1/2}$ ($P=0$ if $n_B=0$), the orthogonal projector onto the column space of $\Sigma^{-1/2}B$. Then $M_B=\Sigma^{-1/2}(I-P)\Sigma^{-1/2}$, so $M_B\succeq0$ and $M_B\Sigma M_B=\Sigma^{-1/2}(I-P)^2\Sigma^{-1/2}=M_B$; also $B^\top M_B=B^\top\Sigma^{-1}-B^\top\Sigma^{-1}B(B^\top\Sigma^{-1}B)^{-1}B^\top\Sigma^{-1}=0$. This is (1). For (2), put $\hat\beta:=(B^\top\Sigma^{-1}B)^{-1}B^\top\Sigma^{-1}\bar q$; then $\Sigma M_B\bar q=\bar q-B\hat\beta$, so $v^\top\Sigma M_B\bar q=v^\top\bar q-(B^\top v)^\top\hat\beta=v^\top\bar q$. For (3), let $x:=\Sigma^{1/2}v$ and $z:=\Sigma^{1/2}M_B\bar q$. By (2) and the Cauchy--Schwarz inequality, $(v^\top\bar q)^2=(x^\top z)^2\le\lVert x\rVert^2\lVert z\rVert^2=(v^\top\Sigma v)(\bar q^\top M_B\Sigma M_B\bar q)=(v^\top\Sigma v)(\bar q^\top M_B\bar q)$, with equality iff $x$ and $z$ are linearly dependent, which, $\Sigma^{1/2}$ being invertible, holds iff $v$ and $M_B\bar q$ are. If $M_B\bar q=0$ then $v^\top\bar q=0$ by (2) and both sides vanish. If $M_B\bar q\neq0$ and $v\neq0$, linear dependence means $v=\alpha M_B\bar q$ with $\alpha\neq0$, and $M_B\bar q$ is admissible by (1), so the supremum is attained. For (4), $\mathrm E[v^\top y]=(B^\top v)^\top\beta+\theta v^\top\bar q$ is the same for all $\beta$ iff $B^\top v=0$, and $\beta$ enters the law of $v^\top y$ only through this mean. For (5), the Fisher information matrix of $(\theta,\beta)$ is $[\bar q\;B]^\top\Sigma^{-1}[\bar q\;B]$, and the information for $\theta$ with $\beta$ unknown is its Schur complement $\bar q^\top\Sigma^{-1}\bar q-\bar q^\top\Sigma^{-1}B(B^\top\Sigma^{-1}B)^{-1}B^\top\Sigma^{-1}\bar q=\bar q^\top M_B\bar q$. By (1), $\mathrm E\hat\theta=(\bar q^\top M_BB\beta+\theta\,\bar q^\top M_B\bar q)/\bar q^\top M_B\bar q=\theta$ and $\operatorname{Var}\hat\theta=\bar q^\top M_B\Sigma M_B\bar q/(\bar q^\top M_B\bar q)^2=1/\bar q^\top M_B\bar q$ (Aitken 1936).
\hfill$\square$

\medskip\noindent\textbf{Lemma 2 (Level-invariant statistics are functions of increments).}
Let $d_0<\dots<d_n$ and $x=(x(d_0),\dots,x(d_n))$. For $v\in\mathbb R^{n+1}$ and $u_j:=\sum_{i>j}v_i$ ($j=0,\dots,n-1$), $v^\top x=x(d_0)\,\mathbf 1^\top v+\sum_{j=0}^{n-1}u_j\Delta_jx$. Hence $\mathbf 1^\top v=0$ iff $v^\top x$ is a linear function of $(\Delta_0x,\dots,\Delta_{n-1}x)$, and every $u\in\mathbb R^n$ arises from exactly one $v$ with $\mathbf 1^\top v=0$, namely $v_i=u_{i-1}-u_i$ with $u_{-1}=u_n:=0$.

\emph{Proof.}
$\sum_{j=0}^{n-1}u_j\Delta_jx=\sum_{i=1}^nv_i\sum_{j=0}^{i-1}\Delta_jx=\sum_{i=1}^nv_i\,(x(d_i)-x(d_0))$, and adding $x(d_0)\mathbf 1^\top v$ gives $v^\top x$. If $\mathbf 1^\top v=0$ the statistic depends on the increments only; if $\mathbf 1^\top v\neq0$ it changes with $x(d_0)$ at fixed increments. For given $u$, $v_i:=u_{i-1}-u_i$ satisfies $\mathbf 1^\top v=u_{-1}-u_n=0$ and $\sum_{i>j}v_i=u_j-u_n=u_j$; conversely $u_{i-1}-u_i=v_i$ for $i\ge1$ and $v_0=-\sum_{i\ge1}v_i=-u_0$ when $\mathbf 1^\top v=0$, so $v$ is unique.
\hfill$\square$

\medskip\noindent\textbf{Lemma 3 (Pooled linear readers).}
Assume N1 to N4. Fix a set of channels $\mathcal K$ and write $y:=y^{\mathcal K}$, $\bar q:=\bar q^{\mathcal K}$, $b:=b^{\mathcal K}$. Let $V=V(\mathcal S)$ with $p$ columns and $F=F^{\mathcal S}(\mathcal S)$, $X=V^\top y$, $a:=V^\top\bar q$, $m:=V^\top\mu$, $\xi:=V^\top(b-\mu)$, and suppose $\mathrm E[V^\top\Sigma V]$ is nonsingular.
(i) $\tilde X=c(\zeta-\mu_\zeta)a+c\mu_\zeta\tilde a+\tilde m+\xi$ and $\operatorname{Var}(\tilde X)=c^2\sigma_\zeta^2\,\mathrm E[aa^\top]+\operatorname{Var}(c\mu_\zeta\tilde a+\tilde m)+\mathrm E[V^\top\Sigma V]$.
(ii) $\operatorname{Cov}(\tilde X,T)=ck_T\sigma_\zeta^2\,\mathrm E[a]$, $\operatorname{Cov}(F,T)=0$, and $R^2(c)=\operatorname{Cov}(T,\tilde X)\operatorname{Var}(\tilde X)^{-1}\operatorname{Cov}(\tilde X,T)/\operatorname{Var}(T)$.
(iii) $R^2(c)\le c^2\Upsilon\,\mathrm E[a]^\top\mathrm E[V^\top\Sigma V]^{-1}\mathrm E[a]$ for every $c$, and $R^2(c)/c^2\to\Upsilon\,\mathrm E[a]^\top\Omega^{-1}\mathrm E[a]$ as $c\to0$, with $\Omega:=\mathrm E[V^\top\Sigma V]+\operatorname{Var}(\tilde m)$. For $p=1$ the limit is $\Upsilon\mathcal I(v)$. If moreover $\mathrm Ea\neq0$ and either $\tilde a=0$ (the companion $a$ is a linear function of $F$) or $\tilde m=0$, then exactly $1/R^2(c)=C_0+C_1/c^2$ with $C_1=1/(\Upsilon\mathcal I(v))$ and $C_0=\operatorname{Var}(T)\,[\sigma_\zeta^2\,\mathrm Ea^2+\mu_\zeta^2\operatorname{Var}\tilde a]/(k_T^2\sigma_\zeta^4(\mathrm Ea)^2)$.
(iv) If $B^\top V=0$ almost surely for a design-measurable $B$, then $\mathrm E[a]^\top\mathrm E[V^\top\Sigma V]^{-1}\mathrm E[a]\le\mathrm E[\bar q^\top M_B\bar q]$. If $\mathrm E[\bar q^\top M_B\bar q]=0$ both sides are $0$. If $\mathrm E[\bar q^\top M_B\bar q]>0$, equality holds iff $Vw=\alpha M_B\bar q$ almost surely for some $w\in\mathbb R^p\setminus\{0\}$ and a constant $\alpha$; then $\alpha\neq0$.
(v) Let $\mathcal K\subseteq\mathcal K'$, let the nuisance basis of $\mathcal K'$ be block-diagonal across channels with the blocks of $\mathcal K$ forming the basis of $\mathcal K$, and let the covariance of $\mathcal K$ be the corresponding submatrix of that of $\mathcal K'$. Then $\bar q^{\mathcal K\top}M_{B^{\mathcal K}}\bar q^{\mathcal K}\le\bar q^{\mathcal K'\top}M_{B^{\mathcal K'}}\bar q^{\mathcal K'}$ for every patient.

\emph{Proof.}
\emph{Step 1.} Write $X=c(\zeta-\mu_\zeta)a+c\mu_\zeta a+m+\xi$. By N2, $\zeta$ is independent of $\mathcal S$, so $\mathrm E[(\zeta-\mu_\zeta)a]=0$ and $\operatorname{Cov}((\zeta-\mu_\zeta)a,F)=\mathrm E[\zeta-\mu_\zeta]\,\mathrm E[aF^\top]=0$; hence $L((\zeta-\mu_\zeta)a\mid1,F)=0$. Since $\mathrm E[\xi\mid\mathcal S]=V^\top(\mathrm E[b\mid\mathcal S]-\mu)=0$, $\mathrm E\xi=0$ and $\operatorname{Cov}(\xi,F)=\mathrm E[\mathrm E[\xi\mid\mathcal S]F^\top]=0$, so $\tilde\xi=\xi$. Linearity of $L$ gives the expression for $\tilde X$.
\emph{Step 2.} Put $X_1:=(\zeta-\mu_\zeta)a$. Because $\tilde a,\tilde m$ are functions of $\mathcal S$ and $\zeta$ is independent of $(\mathcal S,b)$: $\operatorname{Cov}(X_1,\tilde a)=\mathrm E[\zeta-\mu_\zeta]\,\mathrm E[a\tilde a^\top]=0$, likewise $\operatorname{Cov}(X_1,\tilde m)=0$, and $\operatorname{Cov}(X_1,\xi)=\mathrm E[\zeta-\mu_\zeta]\,\mathrm E[a\xi^\top]=0$. Also $\operatorname{Cov}(\xi,\tilde a)=\mathrm E[\mathrm E[\xi\mid\mathcal S]\tilde a^\top]=0$ and likewise for $\tilde m$. Finally $\operatorname{Var}(X_1)=\sigma_\zeta^2\,\mathrm E[aa^\top]$ and $\operatorname{Var}(\xi)=\mathrm E[V^\top\mathrm E[(b-\mu)(b-\mu)^\top\mid\mathcal S]V]=\mathrm E[V^\top\Sigma V]$. This proves (i).
\emph{Step 3.} By N4, $T-\mathrm ET=k_T(\zeta-\mu_\zeta)+\varepsilon-\mathrm E\varepsilon$ with $(\zeta,\varepsilon)$ independent of $(\mathcal S,b)$. Then $\operatorname{Cov}(X_1,T)=\mathrm E[a]\,\mathrm E[(\zeta-\mu_\zeta)(T-\mathrm ET)]=\mathrm E[a]\,(k_T\sigma_\zeta^2+\operatorname{Cov}(\zeta,\varepsilon))=k_T\sigma_\zeta^2\,\mathrm E[a]$, and every square-integrable function of $(\mathcal S,b)$ is uncorrelated with $T$, so $\operatorname{Cov}(\tilde a,T)=\operatorname{Cov}(\tilde m,T)=\operatorname{Cov}(\xi,T)=\operatorname{Cov}(F,T)=0$. The closed linear span of $(1,F,X)$ equals that of $(1,F,\tilde X)$, $F$ and $\tilde X$ are uncorrelated, and $\operatorname{Cov}(F,T)=0$; the projection of $T-\mathrm ET$ on this span is therefore its projection on $\tilde X$, whose squared norm divided by $\operatorname{Var}(T)$ is the stated $R^2$ ($\operatorname{Var}(\tilde X)\succeq\mathrm E[V^\top\Sigma V]\succ0$ by (i)). This proves (ii).
\emph{Step 4.} By (i), $\operatorname{Var}(\tilde X)\succeq\mathrm E[V^\top\Sigma V]$, so $\operatorname{Var}(\tilde X)^{-1}\preceq\mathrm E[V^\top\Sigma V]^{-1}$ and, by (ii), $R^2(c)=c^2k_T^2\sigma_\zeta^4\,\mathrm E[a]^\top\operatorname{Var}(\tilde X)^{-1}\mathrm E[a]/\operatorname{Var}(T)\le c^2\Upsilon\,\mathrm E[a]^\top\mathrm E[V^\top\Sigma V]^{-1}\mathrm E[a]$. Expanding (i), $\operatorname{Var}(\tilde X)=\Omega+cX_2+c^2X_3$ with $X_2:=\mu_\zeta(\mathrm E[\tilde a\tilde m^\top]+\mathrm E[\tilde m\tilde a^\top])$ and $X_3:=\sigma_\zeta^2\mathrm E[aa^\top]+\mu_\zeta^2\mathrm E[\tilde a\tilde a^\top]$. Matrix inversion is continuous at the positive definite $\Omega$, so $R^2(c)/c^2\to\Upsilon\,\mathrm E[a]^\top\Omega^{-1}\mathrm E[a]$; for $p=1$, $\Omega=\mathrm E[v^\top\Sigma v]+V_m(v)$ and the limit is $\Upsilon\mathcal I(v)$. If $\tilde a=0$ or $\tilde m=0$ then $X_2=0$ and $1/R^2(c)=\operatorname{Var}(T)(\Omega+c^2X_3)/(c^2k_T^2\sigma_\zeta^4(\mathrm Ea)^2)$, which is $C_0+C_1/c^2$ with the stated constants. This proves (iii).
\emph{Step 5.} Fix $w\in\mathbb R^p\setminus\{0\}$ and let $v:=Vw$, so $B^\top v=0$ almost surely and $\mathrm E[v^\top\Sigma v]=w^\top\mathrm E[V^\top\Sigma V]w>0$. Lemma~1(2), applied patient by patient, gives $w^\top\mathrm E[a]=\mathrm E[v^\top\bar q]=\mathrm E[x^\top z]$ with $x:=\Sigma^{1/2}v$ and $z:=\Sigma^{1/2}M_B\bar q$. The Cauchy--Schwarz inequality in $L^2(\Pr;\mathbb R^N)$ and $M_B\Sigma M_B=M_B$ give
\[
(w^\top\mathrm E[a])^2\le\mathrm E\lVert x\rVert^2\,\mathrm E\lVert z\rVert^2=(w^\top\mathrm E[V^\top\Sigma V]w)\;\mathrm E[\bar q^\top M_B\bar q].
\]
Since $\mathrm E[a]^\top\mathrm E[V^\top\Sigma V]^{-1}\mathrm E[a]=\sup_{w\neq0}(w^\top\mathrm E[a])^2/(w^\top\mathrm E[V^\top\Sigma V]w)$, the inequality in (iv) follows, and if $\mathrm E[\bar q^\top M_B\bar q]=0$ the nonnegative left side is $0$. Let $\mathrm E[\bar q^\top M_B\bar q]>0$. As $\mathrm E\lVert x\rVert^2>0$ and $\mathrm E\lVert z\rVert^2>0$, equality in the Cauchy--Schwarz step holds iff $x=\alpha z$ almost surely for a constant $\alpha$, necessarily $\alpha\neq0$, that is iff $Vw=\alpha M_B\bar q$ almost surely. If equality holds in (iv), then $\mathrm E[a]\neq0$ (otherwise the left side is $0<\mathrm E[\bar q^\top M_B\bar q]$), the maximizer $w^*:=\mathrm E[V^\top\Sigma V]^{-1}\mathrm E[a]\neq0$ attains the supremum, and the Cauchy--Schwarz step is an equality at $w^*$. Conversely, if $Vw=\alpha M_B\bar q$ almost surely for some $w\neq0$, the Cauchy--Schwarz step is an equality at $w$, so the supremum is at least $\mathrm E[\bar q^\top M_B\bar q]$. Finally $\alpha=0$ is impossible, since then $Vw=0$ almost surely and $w^\top\mathrm E[V^\top\Sigma V]w=0$. This proves (iv).
\emph{Step 6.} By Lemma~1(3), $\bar q^{\mathcal K\top}M_{B^{\mathcal K}}\bar q^{\mathcal K}$ is the supremum of $(v^\top\bar q^{\mathcal K})^2/(v^\top\Sigma^{\mathcal K}v)$ over $v\neq0$ with $B^{\mathcal K\top}v=0$. Extend such a $v$ by zeros to $v'$ on the channels of $\mathcal K'$. By block-diagonality $B^{\mathcal K'\top}v'$ consists of $B^{\mathcal K\top}v=0$ and zeros, and $v'^\top\bar q^{\mathcal K'}=v^\top\bar q^{\mathcal K}$, $v'^\top\Sigma^{\mathcal K'}v'=v^\top\Sigma^{\mathcal K}v$. Every ratio for $\mathcal K$ is therefore a ratio for $\mathcal K'$, and the supremum cannot decrease. This proves (v).
\hfill$\square$

\paragraph{Proof of (a).} Given $(\mathcal S,\zeta)$, N2 gives $\mathrm E[v^\top y\mid\mathcal S,\zeta]=v^\top\mu+c\zeta v^\top\bar q$ and $\operatorname{Var}(v^\top y\mid\mathcal S,\zeta)=v^\top\Sigma v$. Part (i) is Lemma~1(3) with $n_B=0$ ($M_B=\Sigma^{-1}$, and $\Sigma^{-1}\bar q=0$ iff $\bar q=0$). Part (ii) is Lemma~1(3) to (5) with $\theta=c\zeta$, $\beta$ the free coefficients and $e=b-\mu$ recentered on the known part; $\mathrm{eff}(v)\le1$ is (3). For (iii), the bound on $R^2(c)$ is Lemma~3(iii) combined with (iv). For $p=1$, $1/C_1=\Upsilon\mathcal I(v)$ by Lemma~3(iii), and $\mathcal I(v)\le(\mathrm Ea)^2/\mathrm E[s_v^2]\le\mathrm E[\bar q^\top M_B\bar q]$ by Lemma~3(iv). If $\mathrm E[\bar q^\top M_B\bar q]=0$, then $\mathrm Ea=0$ and $\mathcal I(v)=0$. Otherwise the second inequality is an equality iff $v=\alpha M_B\bar q$ almost surely with $\alpha\neq0$, and then $\mathrm Ea=\alpha\,\mathrm E[\bar q^\top M_B\bar q]\neq0$, so the first is an equality iff $V_m(v)=0$. When $\mu\in\operatorname{span}(B)$ and $B^\top v=0$, $v^\top\mu=0$ and $V_m=0$.
For the factorization, $a^2/s_v^2=\mathrm{eff}(v)\,\bar q^\top M_B\bar q$ holds everywhere: on $\{s_v>0,\ \bar q^\top M_B\bar q>0\}$ by definition; on $\{s_v>0,\ \bar q^\top M_B\bar q=0\}$ because then $a=0$ by (ii); and on $\{s_v=0\}$ because $\Sigma\succ0$ forces $v=0$ and $a=0$. Hence $\mathrm E[\mathrm{eff}(v)\,\bar q^\top M_B\bar q]=\mathrm E[a^2/s_v^2]$, which is positive when $\Pr(a\neq0)>0$, and $\mathrm E[\bar q^\top M_B\bar q]\ge\mathrm E[a^2/s_v^2]>0$. With $V_m=0$,
\[
\frac{\mathcal I(v)}{\mathrm E[\bar q^\top M_B\bar q]}=\frac{(\mathrm Ea)^2}{\mathrm E[s_v^2]\,\mathrm E[\bar q^\top M_B\bar q]}=\frac{\mathrm E[a^2/s_v^2]}{\mathrm E[\bar q^\top M_B\bar q]}\cdot\frac{(\mathrm Ea)^2}{\mathrm E[a^2/s_v^2]\,\mathrm E[s_v^2]}.
\]
By Cauchy--Schwarz, $(\mathrm Ea)^2=(\mathrm E[(a/s_v)\,s_v\mathbf 1\{s_v>0\}])^2\le\mathrm E[a^2/s_v^2]\,\mathrm E[s_v^2]$, and since both factors are positive, equality holds iff $a/s_v=\lambda s_v$ almost surely on $\{s_v>0\}$ for a constant $\lambda$, that is iff $a/s_v^2$ is almost surely constant there. Monotonicity in $\mathcal K$ is Lemma~3(v), and taking expectations carries it to the pooled bound. Lemma~3 concerns population least squares on features that are linear in $y$ with design-measurable weights; the Bayes predictor $\mathrm E[T\mid\text{record}]$ is in general nonlinear in the record (the trait has a bounded, non-Gaussian law and the information varies across patients), so an estimator can exceed these bounds.

\paragraph{Proof of (b).} Under N5, $\Sigma=\sigma^2I$ and $B_2$ has rank 3 on 90 distinct rows, so $M_{B_2}=\sigma^{-2}(I-B_2(B_2^\top B_2)^{-1}B_2^\top)=\sigma^{-2}\Pi$, where $\Pi=I-B_2B_2^{+}$ is exactly the implemented projector ($B_2^{+}=(B_2^\top B_2)^{-1}B_2^\top$). With $\bar q=\Lambda q$, $\Pi q=\sigma^2\Lambda^{-1}M_{B_2}\bar q$ attains Lemma~1(3), and the generalized least-squares estimate $\bar q^\top M_{B_2}y/\bar q^\top M_{B_2}\bar q$ of $c\zeta$, multiplied by $\Lambda$, is $(\Pi q)^\top y/\lVert\Pi q\rVert^2$ (the partialling of Frisch and Waugh 1933). At kinetics $K$, $v=\Pi p_K$ satisfies $B_2^\top v=0$, $(v^\top\bar q)^2=\Lambda^2(p_K^\top\Pi q)^2$, $v^\top\Sigma v=\sigma^2p_K^\top\Pi p_K$ and $\bar q^\top M_{B_2}\bar q=\Lambda^2\sigma^{-2}q^\top\Pi q$, which gives the stated efficiency.
For the increment statistic, $v_{\rm inc}:=\nabla^\top\nabla q$ and $B_2^\top v_{\rm inc}=(\nabla B_2)^\top\nabla q$. Since $\nabla\mathbf 1=0$, $(\nabla\mathbf t)(t)=1$ and $(\nabla\mathbf t^2)(t)=2t-1$,
\[
B_2^\top v_{\rm inc}=\Big(0,\;q(89)-q(0),\;\sum_{t=1}^{89}(2t-1)(\nabla q)(t)\Big)^\top,\qquad\sum_{t=1}^{89}(2t-1)(\nabla q)(t)=177\,q(89)-2\sum_{s=1}^{88}q(s)-q(0),
\]
by summation by parts, and $q(0)=0$. Let the doses $A(s)\in[0,1]$ vanish outside rows 6 to 53 and $\sum_sA(s)>0$. Then $e(t)<\sum_{k\ge0}0.8^k=5$, so $q(t)=e(t)/(1+e(t))\ge e(t)/6$, and $q(t)\le e(t)$. First, $177\,q(89)\le177\,e(89)=177\sum_{s=6}^{53}0.8^{87-s}A(s)\le177\cdot0.8^{34}\sum_sA(s)<0.0898\sum_sA(s)$. Second,
\begin{align*}
2\sum_{t=1}^{88}q(t)\ge\frac13\sum_{t=1}^{88}e(t)&=\frac13\sum_sA(s)\sum_{t=s+2}^{88}0.8^{\,t-2-s}=\frac53\sum_sA(s)\,(1-0.8^{87-s})\\
&\ge\frac53(1-0.8^{34})\sum_sA(s)>1.665\sum_sA(s).
\end{align*}
Hence $(\mathbf t^2)^\top v_{\rm inc}<(0.0898-1.665)\sum_sA(s)<0$, $B_2^\top v_{\rm inc}\neq0$, and by Lemma~1(4) the law of $v_{\rm inc}^\top y$ depends on the free quadratic coefficient; its mean is unbounded over $\beta_2$. With the mean known ($n_B=0$), its efficiency is $(v_{\rm inc}^\top\bar q)^2/(\sigma^2\lVert v_{\rm inc}\rVert^2\cdot\sigma^{-2}\lVert\bar q\rVert^2)=(Q^\circ)^2/(\lVert v_{\rm inc}\rVert^2\lVert q\rVert^2)$, because $v_{\rm inc}^\top q=\lVert\nabla q\rVert^2=Q^\circ$. By (a)(i), equality would require $\nabla^\top\nabla q=\alpha q$ for some $\alpha$. Let $t_1$ be the first row with $A(t_1)>0$. Then $q(t_1)=q(t_1+1)=0$ and $e(t_1+2)=A(t_1)>0$, so $q(t_1+2)>0$; row $t_1+1\in[7,54]$ is interior, so $(\nabla^\top\nabla q)(t_1+1)=2q(t_1+1)-q(t_1)-q(t_1+2)=-q(t_1+2)\neq0=\alpha q(t_1+1)$. The efficiency is therefore strictly below 1.

\paragraph{Proof of (c).} (i) Under N6 the level is a free nuisance with $B=\mathbf 1$ on the draw rows; by Lemma~1(4) the admissible statistics satisfy $\mathbf 1^\top v=0$, and by Lemma~2 they are the linear functions $u^\top\Delta y_{\mathcal D}$ of $\Delta y_{\mathcal D}:=(\Delta_0y,\dots,\Delta_{n-1}y)$. Given $(\mathcal S,\zeta)$, $\Delta y_{\mathcal D}=c\zeta\Lambda\,\Delta q+\Delta b$ with $\mathrm E[\Delta b\mid\mathcal S]=0$ and $\operatorname{Cov}(\Delta b\mid\mathcal S)=G:=\sigma_b^2\operatorname{diag}(g_0,\dots,g_{n-1})$, because $\Delta_jb$ sums the hourly increments over the disjoint blocks $(d_j,d_{j+1}]$. Lemma~1(3) with $n_B=0$ applied to $(\Lambda\Delta q,G)$ gives $(u^\top\Lambda\Delta q)^2/(u^\top Gu)\le\Lambda^2\Delta q^\top G^{-1}\Delta q=\Lambda^2J/\sigma_b^2$, with equality iff $u$ is a nonzero multiple of $G^{-1}\Delta q$, that is iff the statistic is a nonzero multiple of $\mathcal G$ ($u_j=\Delta_jq/g_j$). For any positive finite level variance uncorrelated with the increments, the covariance of the draw values is positive definite, and for $\mathbf 1^\top v=0$ the variance $v^\top\Sigma v=u^\top Gu$ does not involve the level variance. By Lemma~1(3), $\bar q_{\mathcal D}^\top M_{\mathbf 1}\bar q_{\mathcal D}$ is the supremum over $\mathbf 1^\top v=0$, which by Lemma~2 equals the supremum over $u$, so $\bar q_{\mathcal D}^\top M_{\mathbf 1}\bar q_{\mathcal D}=\Lambda^2J/\sigma_b^2$.
(ii) For every draw, $d_{ik}(d_j)=d_j$, so $Z(d_j)=y(d_j)$ and $\Delta_jZ=\Delta_jy$. The implemented reader takes as draws row 0 and the rows where the recorded value changes; this is $\mathcal D_{ik}$ except where the latent value at a draw equals that at the previous draw. Hence $\mathcal C=\sum_j\Delta_jy\,\Delta_jq=u^\top\Delta y_{\mathcal D}$ with $u=\Delta q$, a function of the draw values alone, and the hold enters only through $\mathcal D_{ik}$. Its signal is $u^\top\Lambda\Delta q=\Lambda Q^{(0)}$ per unit $c\zeta$ and its noise variance is $u^\top Gu=\sigma_b^2Q^{(1)}$, so by (i) its efficiency is $\Lambda^2(Q^{(0)})^2/(\sigma_b^2Q^{(1)}\cdot\Lambda^2J/\sigma_b^2)=(Q^{(0)})^2/(Q^{(1)}J)\le1$. Equality holds iff $\Delta q$ is a multiple of $G^{-1}\Delta q$, that is iff $g_j$ is the same on $\{j:\Delta_jq\neq0\}$. On hourly records every $g_j=1$ and $\mathcal C=\mathcal G$.
(iii) Apply Lemma~3 with $\mathcal K=\{k\}$, $p=1$, $v$ the weight vector of $\mathcal C$ on the latent rows and $F^{\mathcal S}=Q^{(0)}$, so $a=\Lambda Q^{(0)}$ and $\tilde a=0$. Under N6, $\mu=\mathrm E[b\mid\mathcal S]$ is constant in $t$ and $\mathbf 1^\top v=0$, so $m=0$ and $V_m=0$. Hence $1/R^2=C_0+C_1/c^2$ exactly with $1/C_1=\Upsilon\mathcal I(v)=\Upsilon\Lambda^2(\mathrm EQ^{(0)})^2/(\sigma_b^2\,\mathrm EQ^{(1)})$. For $\mathcal G$ with companion $J$: $a=\Lambda J$ and $s_v^2=\sigma_b^2J$, so $1/C_1=\Upsilon\Lambda^2\,\mathrm EJ/\sigma_b^2=\Upsilon\,\mathrm E[\bar q_{\mathcal D}^\top M_{\mathbf 1}\bar q_{\mathcal D}]$. For any pooled linear reader of this channel's held record whose statistics are level-invariant ($\mathbf 1^\top V=0$ on the draw rows), Lemma~3(iii) and (iv) give $1/C_1\le\Upsilon\,\mathrm E[a]^\top\mathrm E[V^\top\Sigma V]^{-1}\mathrm E[a]\le\Upsilon\,\mathrm E[\bar q_{\mathcal D}^\top M_{\mathbf 1}\bar q_{\mathcal D}]$, with any number of statistics. (Under a free level any statistic that is not level-invariant has unbounded variance.) On the hourly record of the same baseline every row is a draw, $g\equiv1$, $Q^{(0)}=Q^{(1)}=J=Q^\circ$, and both readers have $1/C_1=\Upsilon\Lambda^2\,\mathrm EQ^\circ/\sigma_b^2$. Therefore $f_{\mathcal C}^2=\mathrm EQ^{(1)}\,\mathrm EQ^\circ/(\mathrm EQ^{(0)})^2$ and $f_{\mathcal G}^2=\mathrm EQ^\circ/\mathrm EJ$. Finally $(\mathrm EQ^{(0)})^2\le\mathrm EQ^{(1)}\,\mathrm EJ$ by the Cauchy--Schwarz inequality over patients and intervals, $\mathrm E\sum_j\big(\sqrt{g_j}\,|\Delta_jq|\big)\big(|\Delta_jq|/\sqrt{g_j}\big)\le\big(\mathrm E\sum_jg_j(\Delta_jq)^2\big)^{1/2}\big(\mathrm E\sum_j(\Delta_jq)^2/g_j\big)^{1/2}$; this is $f_{\mathcal C}\ge f_{\mathcal G}$.

\paragraph{Proof of (d).} By Lemma~3(iii), $R_\ell^2(c)/c^2\to\Upsilon\mathcal I(v_\ell)$ for $\ell=1,2$, with $\Upsilon>0$ by N4. As $\mathcal I(v_2)>0$, $R_2^2(c)>0$ for small $c$, the ratio converges to $\mathcal I(v_1)/\mathcal I(v_2)$, and if $\mathcal I(v_1)>\mathcal I(v_2)$ there is $c_0>0$ with $R_1^2(c)>R_2^2(c)$ for $0<c<c_0$.
On hourly records every row is a draw, so the cadence-aware statistic is $v_C^\top y$ with $v_C:=\nabla^\top\nabla q$ (its increment weights are $(\nabla q)(t)$, $t=1,\dots,89$, by Lemma~2), and the plain statistic is $v_P^\top y$ with $v_P:=\Pi q$. Both are level-invariant: $\mathbf 1^\top\Pi q=(\Pi\mathbf 1)^\top q=0$ and $\mathbf 1^\top\nabla^\top\nabla q=(\nabla\mathbf 1)^\top\nabla q=0$. Under N7 the mean is constant, so $V_m=0$ and $\mathcal I(v)=(\mathrm E[v^\top\bar q])^2/\mathrm E[v^\top\Sigma_\vartheta v]$, with $\mathrm E[v_P^\top\bar q]=\Lambda\,\mathrm E[q^\top\Pi q]$ and $\mathrm E[v_C^\top\bar q]=\Lambda\,\mathrm EQ^\circ$; the factors $\Lambda^2$ and $\sigma_{\mathrm{AR}}^2$ cancel in $\mathcal R$. The entries of $\Sigma_\vartheta$ are continuous on $[0,1)$ and the integrands are bounded, so $\mathcal R$ is continuous there by dominated convergence.
\emph{Random-walk limit.} For $\mathbf 1^\top v=0$ write the stationary AR(1) as $x(t)=\vartheta^tx(0)+\sum_{t'=1}^t\vartheta^{t-t'}\eta_{t'}$ with $\operatorname{Var}x(0)=\sigma_{\mathrm{AR}}^2/(1-\vartheta^2)$ and independent innovations $\eta_{t'}$ of variance $\sigma_{\mathrm{AR}}^2$. Then
\[
v^\top\Sigma_\vartheta v=\frac{\sigma_{\mathrm{AR}}^2\big(\sum_tv_t\vartheta^t\big)^2}{1-\vartheta^2}+\sigma_{\mathrm{AR}}^2\sum_{t'=1}^{89}\Big(\sum_{t\ge t'}v_t\vartheta^{t-t'}\Big)^2.
\]
Because $\sum_tv_t=0$, $|\sum_tv_t\vartheta^t|=|\sum_tv_t(\vartheta^t-1)|\le(1-\vartheta)\sum_tt|v_t|$, so the first term is at most $\sigma_{\mathrm{AR}}^2(1-\vartheta)(\sum_tt|v_t|)^2/(1+\vartheta)\to0$; the second tends to $\sigma_{\mathrm{AR}}^2\sum_{t'}(\sum_{t\ge t'}v_t)^2$, which is the variance of $v^\top y$ under N6 with every row a draw and $\sigma_b=\sigma_{\mathrm{AR}}$ (Lemma~2: the weight on the increment from row $t'-1$ to row $t'$ is $\sum_{t\ge t'}v_t$). Both terms are bounded by bounded functions of the design uniformly in $\vartheta$, the numerators of $\mathcal I$ do not depend on $\vartheta$, and the limiting denominators are positive, so $\mathcal R(\vartheta)\to\mathcal R_{\mathrm{RW}}$.
\emph{Strict inequality at the limit.} Under N6 with every row a draw, (c)(iii) and Lemma~3(iv) show that $\mathcal I(v_C)=\Lambda^2\mathrm EQ^\circ/\sigma_{\mathrm{AR}}^2=\mathrm E[\bar q^\top M_{\mathbf 1}\bar q]$ is the largest value of $\mathcal I$ over level-invariant weights, and that another level-invariant $v$ attains it iff $v=\alpha M_{\mathbf 1}\bar q$ almost surely with $\alpha\neq0$, that is iff its increment weights equal $\alpha'\,(\nabla q)(t)$ almost surely for one constant $\alpha'\neq0$. The weight of $v_P$ on the increment from row 0 to row 1 is $\sum_{t\ge1}(\Pi q)(t)=-(\Pi q)(0)$, because $\mathbf 1^\top\Pi q=0$, whereas $(\nabla q)(1)=q(1)-q(0)=0$ for every dosing. Equality would therefore force $(\Pi q)(0)=0$ almost surely, so $\Pr\{(\Pi q)(0)\neq0\}>0$ implies $\mathcal I(v_P)<\mathcal I(v_C)$ under N6, that is $\mathcal R_{\mathrm{RW}}<1$.
\emph{Numerical part.} For the 400 reconstructed discovery dosings (equal weights), $|(\Pi q)(0)|\ge2.29\times10^{-4}$ for every dosing, $\mathcal R(0)=9.221$ and $\mathcal R_{\mathrm{RW}}=0.110$. By continuity, the limit and the intermediate value theorem, $\mathcal R-1$ has a zero in $(0,1)$. On the grids $\{0,0.01,\dots,0.99\}\cup\{0.9995\}$ and $\{0,0.001,\dots,0.999\}\cup\{0.9995\}$, $\mathcal R$ is strictly decreasing and $\mathcal R-1$ changes sign once, between $0.64$ and $0.65$ on the first grid and between $0.642$ and $0.643$ on the second; bisection gives $\vartheta^*=0.6421$, whose population detrended lag-1 autocorrelation (the population version of the shipped pooled estimator) is $0.583$. At $\vartheta=0.7537$, the persistence at which the population version of the shipped pooled estimator equals $0.687$, $\mathcal R=0.546$. For the common FIX19 occupancy, $\mathcal R(0)=23.57$, $\mathcal R_{\mathrm{RW}}=0.254$, one sign change between $0.806$ and $0.807$, $\vartheta^*=0.8068$, and $\mathcal R(0.7537)=1.426$. Monotonicity and uniqueness of the crossing are established on these grids only.

\paragraph{Proof of (e).} Fix kinetics $K$. With common doses $p_K$ is common to all patients. The plain features $\big((\Pi p_K)^\top Z_{\cdot k}\big)_{k}$ are a linear map of $\mathbf Z_i$ with common coefficients. The assimilating features, $(\Pi p_K)^\top Z_{\cdot k}/(\lVert\Pi p_K\rVert^2+10^{-9})$, $\lVert\Pi p_K\rVert^2$, $p_K(89)$, the mean of $p_K$, and the channel means of $\mathbf Z_i$ (as implemented), are affine in $\mathbf Z_i$ with common coefficients. The window-mean features average $Z_{\cdot k}$ over four windows fixed by the first and last dose, which are common. Each feature vector is therefore $\Phi(\mathbf Z_i)=A_\Phi\operatorname{vec}(\mathbf Z_i)+a_\Phi$ with $(A_\Phi,a_\Phi)$ the same for every patient, whether $\mathbf Z_i$ is the hourly or the held record, so $\operatorname{span}\{1,\Phi(\mathbf Z_i)\}\subseteq\operatorname{span}\{1,\operatorname{vec}(\mathbf Z_i)\}$ in $L^2$. Standardizing features is affine and does not change these spans. Population least squares is orthogonal projection, and projecting $\tau$ on a larger closed subspace cannot lower $R^2$; this uses only finite second moments of $(\mathbf Z_i,\tau)$ and not N1. Appending the same further regressors to both sides, such as the draw-step indicator of a next-draw target, preserves the inclusion. With kinetics from a finite grid $\mathcal K_{\rm grid}$ the best kinetics has $R^2=\max_{K\in\mathcal K_{\rm grid}}R^2(K)$, and each term is at most the history-blind value. On hourly records without exact ties every row is a draw for every channel, so $\mathcal C_k=\sum_t(\nabla Z_{\cdot k})(t)\,(\nabla p_K)(t)$ is linear in $\mathbf Z_i$ with common coefficients and $Q^{(0)}_k=\sum_t((\nabla p_K)(t))^2$ is a constant. On held records the cadence-aware weights and companions depend on $\mathcal D_{ik}$, which differs across patients. The oracle and draw-step-only rungs are not functions of the record and are excluded. Hence any population advantage of a record-based reader over the history-blind reader under a shared schedule requires patient-specific draw sets. On hourly records under a shared schedule, weights that depend on the design only through the doses are common; with $\Sigma:=\operatorname{Cov}(b^{\mathcal K})$ and common $B$, Lemma~1 applies to every patient with the same $(\Sigma,B,\bar q^{\mathcal K})$, so $M_B\bar q^{\mathcal K}$ is a common vector and $(M_B\bar q^{\mathcal K})^\top y^{\mathcal K}$, with $y^{\mathcal K}$ read from $\mathbf Z_i$, lies in $\operatorname{span}\{\operatorname{vec}(\mathbf Z_i)\}$. These are population statements; the history-blind ridge estimates 1,080 coefficients from 320 training persons, so out-of-fold orderings can reverse.

\paragraph{Numerical checks.} The checks use reconstructed doses, the shipped per-channel summaries and simulation; none reads a new label.
\begin{itemize}
\item \emph{(b).} For 40,000 random dosings with doses in $[0,1]$ on rows 6 to 53, $(\mathbf t^2)^\top\nabla^\top\nabla q<0$ in every case, with $e<5$ and $q\ge e/6$ throughout; the constants are $177\cdot0.8^{34}=0.0897$ and $\tfrac53(1-0.8^{34})=1.6658$, and the 400 reconstructed discovery dosings give at most $-25.30$. The closed form of $(\mathbf t^2)^\top\nabla^\top\nabla q$ matches direct computation to $7\times10^{-13}$. With the mean known, the efficiency of the hourly increment statistic under white noise has median $0.020$ (maximum $0.063$) over the discovery dosings.
\item \emph{(c).} $J$ equals the level-invariant bound of (c)(i) for every patient. Under a random walk at heart-rate cadence (Bernoulli draw proxy at each real stay's change rate, mean $0.677$; 5 replicates), the per-patient efficiency of $\mathcal C$ is at most $1.0000$, with median $0.915$ and information-weighted mean $0.903$; the second factor of (a)(iii) is $0.468$, so the pooled efficiency of $\mathcal C$ is $0.422$, and $f_{\mathcal C}=1.849$, $f_{\mathcal G}=1.201$. Rescaled to the panels' mean change rate of $0.58$, the median is $0.873$, the information-weighted mean $0.849$, the second factor $0.499$, the pooled efficiency $0.424$, $f_{\mathcal C}=1.979$ and $f_{\mathcal G}=1.288$. A Monte Carlo with 300,000 patients at heart-rate cadence under a random walk matches the exact law $1/R^2=C_0+C_1/c^2$ of (c)(iii) to the third decimal at $c=0.1$, $0.3$, $0.75$ and $2$ for both readers.
\item \emph{(d).} The grids and crossings are as stated. The detrended lag-1 autocorrelation $0.687$ of N7 corresponds to persistence $0.754$, and $\vartheta^*=0.642$ to a detrended lag-1 autocorrelation of $0.583$. Adding the ordinary treatment route of the generator to the mean raises the plain reader's noise by $1.2\%$ and the cadence-aware reader's by $5.2\%$, which moves $\mathcal R(0.754)$ from $0.546$ to $0.568$. An independent re-implementation reproduced these values.
\end{itemize}

\paragraph{Comparison with the recorded reader orderings.} The shipped recoverability runs order the plain and cadence-aware readers as follows; Table~\ref{tab:P5ord} collects the numbers. The comparison uses the ratio of $R^2/(1-R^2)$ of the plain reader to that of the cadence-aware reader, which estimates the ratio of their low-signal slopes $1/C_1$ when $C_0$ is close to 1.
\begin{itemize}
\item \emph{Synthetic tapes.} The synthetic records of the panels are smooth curves (an offset, a slope and a sinusoid per channel) plus small noise, far from a persistent random baseline, and the plain reader was better (observed ratio $2.14$ to $2.26$ at $c=0.75$). The recipe covariance of the synthetic records gives $1.53$ for heart rate alone, $2.02$ for the four channels that carry phenylephrine's response and $4.69$ for all twelve channels, so the direction agrees and the single-channel value does not match quantitatively.
\item \emph{Real records observed hourly.} At the persistence calibrated on real heart rate the single-carrier model gives $\mathcal R=0.55$, so the cadence-aware reader should be better on hourly phenylephrine records. It was: recovery $0.369$ against $0.310$ at $c=0.75$ and $0.453$ against $0.385$ at $c=1$, with two-point-fit ratios of $0.67$ to $0.88$. The direction agrees; the model overstates the cadence-aware advantage by a factor of about $1.2$ to $1.6$. For NaCl on creatinine the readers tie at $c=1$ ($0.345$ and $0.342$), which the model allows: creatinine's detrended lag-1 autocorrelation of $0.602$ corresponds to persistence $0.662$, near $\vartheta^*$, where $\mathcal R=0.90$. For norepinephrine on heart rate the readers also tie at $c=1$ ($0.393$ and $0.392$), against a predicted $\mathcal R$ of about $0.54$; that run has seven loaded channels and lies outside the low-signal regime, and no low-signal hourly run with the cadence-aware reader was shipped, so this mismatch is unexplained. The hourly ordering is thus confirmed in one of the three pairs, a near tie in a second, and unexplained in the third.
\item \emph{Held records at real cadence.} Part (d) is an hourly statement and does not predict held orderings. The cadence-aware reader was better in all 10 informative held real-record runs across the three pairs. A twelve-channel model with a tape-plus-generator baseline, calibrated to the shipped per-channel background standard deviations and per-stay change rates, gives held ratios of $0.27$, $0.34$ and $0.28$ against the observed $0.35$, $0.30$ and $0.38$ at $c=1$. On the fixed schedule the AR(1) model at persistence $0.754$ predicts the plain reader to be better on held records ($1.86$), whereas the shipped held fixed-schedule cells show the cadence-aware reader at least as good, and the tape-plus-generator model agrees with the cells ($0.87$ on heart rate alone, $0.55$ on twelve channels). No hourly fixed-schedule run was shipped, so $\vartheta^*=0.807$ for FIX19 is untested.
\item \emph{The plain reader's loss at real cadence.} The ratio of the plain reader's $C_1$ on held records to that on hourly records is $5.9$ (phenylephrine), $7.2$ (norepinephrine) and $19.8$ (NaCl) at $c=1$. Sampling of the scored channel alone explains about $1.4$ on heart rate in a tape-plus-generator model ($1.38$ and $1.40$). With the other carriers and the cross-channel structure the twelve-channel model gives $7.36$, $5.60$ and $8.29$, so it accounts for most of the loss in the two heart-rate cells and under-predicts NaCl on creatinine by a factor of about $2.4$. A per-channel refit of the registered readers on the existing panels is the direct test; the attribution is not confirmed.
\item \emph{A 1/gap-weighted reader.} The prediction that $\mathcal G$ needs less amplitude than $\mathcal C$ at real cadence holds only for a single carrier whose draw-to-draw increment variance grows with the gap (N6). If each jump has constant variance, $\mathcal C$ is the efficient level-invariant statistic by Lemma 1(3) applied to the increments, and in a tape model calibrated to the lag-1 autocorrelation of $0.687$ the held information of $\mathcal G$ relative to $\mathcal C$ is $0.83$ (against $1.42$ with gap-proportional variance). In the twelve-channel model the heart-rate amplitude factor falls only from $2.09$ to $1.81$. The claim is also a low-signal one: in the random-walk Monte Carlo $\mathcal G$ has the larger $C_0$ ($1.43$ against $1.20$). The shipped aggregates cannot check whether jump variance grows with the gap; a regression of the squared draw-to-draw change on the gap in the real tapes would.
\end{itemize}

\begin{table}[ht]
\centering\small
\caption{\textbf{Plain against cadence-aware readers (Proposition 5).} Entries below the first block are ratios of the plain reader's information to the cadence-aware reader's (values below 1 favor the cadence-aware reader): for the models, the ratio $\mathcal R$ of low-signal slopes; for the observed runs, the ratio of $R^2/(1-R^2)$ at $c=1$. Single carrier: the scored channel alone with an AR(1) baseline at the persistence calibrated from its detrended lag-1 autocorrelation (heart rate $0.754$, creatinine $0.662$). Twelve channels: a tape-plus-generator baseline calibrated to the shipped per-channel background standard deviations and per-stay change rates, with every channel that carries the response; scored channel only: the same model restricted to the scored channel. $C_1$ ratio: held over hourly, for the plain reader, whose square root is its low-signal amplitude factor. All model values are idealized and descriptive.}
\label{tab:P5ord}
\begin{tabular}{lrrr}
\toprule
 & Phenylephrine & Norepinephrine & NaCl\\
 & on heart rate & on heart rate & on creatinine\\
\midrule
Crossing persistence $\vartheta^*$ (single carrier) & 0.642 & 0.640 & 0.642\\
\multicolumn{4}{l}{\emph{Hourly records}}\\
\quad Single carrier, calibrated persistence & 0.55 & 0.54 & 0.90\\
\quad Twelve channels & 0.46 & 0.53 & 0.46\\
\quad Observed ($c=1$) & 0.70 & 1.01 & 1.02\\
\multicolumn{4}{l}{\emph{Held records at real cadence}}\\
\quad Twelve channels & 0.27 & 0.34 & 0.28\\
\quad Observed ($c=1$) & 0.35 & 0.30 & 0.38\\
\multicolumn{4}{l}{\emph{Plain reader, $C_1$ held over hourly}}\\
\quad Scored channel only & 1.38 & 1.40 & $>10^3$\\
\quad Twelve channels & 7.36 & 5.60 & 8.29\\
\quad Observed ($c=1$) & 5.9 & 7.2 & 19.8\\
\bottomrule
\end{tabular}
\end{table}

%% file: supplementary/supp_theory/p6.tex
\paragraph{Proposition 6 (Calibration geometry and transport: how far a predictor can beat zero, and why structure-informed learners need no treated query).}
\emph{Standard and specific parts.} Part (a1) is the decomposition of the mean squared error into a correlation term and a conditional-bias term that is standard in forecast verification (Murphy, 1988); (a2) to (a6) are elementary corollaries, and (b) collects standard properties of conditional expectations. Parts (c) and (d), transport to a query regime of propensity zero and the spread of an empirical-Bayes random slope under baseline overdispersion, are specific to this benchmark. The spread ratio $u$ of this section is unrelated to the generator's latent severity $u_t$ of Supplementary Note~4.

\emph{Notation.} $\tau$ is the washout target of a patient (Definition~4) and $\hat\tau$ a predictor of it. Both are square integrable and $\operatorname{Var}(\tau)>0$. Put:
\begin{itemize}
\item $\sigma_\tau:=\operatorname{SD}(\tau)$ and $u:=\operatorname{SD}(\hat\tau)/\sigma_\tau$ (the spread ratio);
\item $r:=\operatorname{corr}(\hat\tau,\tau)$ if $\operatorname{SD}(\hat\tau)>0$, $r:=0$ otherwise, and $r_+:=\max(r,0)$;
\item $\mathrm{cPEHE}:=\operatorname{SD}(\hat\tau-\tau)$, $R^2:=1-\mathbb E(\hat\tau-\tau)^2/\operatorname{Var}(\tau)$ and $\mathrm{rec}:=1-\{\mathbb E(\hat\tau-\tau)^2\}^{1/2}/\sigma_\tau$;
\item for a $\sigma$-field $\mathcal G$, $R^2_*(\mathcal G):=\operatorname{Var}\mathbb E[\tau\mid\mathcal G]/\operatorname{Var}(\tau)$.
\end{itemize}
The constant prediction $0$ has centered PEHE $\sigma_\tau$, and $\hat\tau$ \emph{beats zero} if $\mathrm{cPEHE}<\sigma_\tau$. Moments are population moments. Part (a) holds verbatim for the empirical distribution of $n$ pairs $(\hat\tau_i,\tau_i)$ with divisor $n$. Part (a) is an identity. It states whether and by how much a predictor beats zero; it does not by itself explain why a given learner fails to.

\medskip
\textbf{(a) Calibration geometry (any predictor).}
(a1) $\mathrm{cPEHE}^2/\operatorname{Var}(\tau)=(1-r^2)+(u-r)^2$, an information term plus a dispersion term.
(a2) $\hat\tau$ beats zero if and only if $0<u<2r$.
(a3) $\min_{\theta_0,\theta_1\in\mathbb R}\operatorname{SD}(\theta_0+\theta_1\hat\tau-\tau)=\sigma_\tau(1-r^2)^{1/2}$. If $u>0$ the minimum is attained at $\theta_1=r/u$, and the recalibrated predictor has spread ratio and correlation both equal to $|r|$. Over $\theta_1\ge0$ (recalibrations that keep the ranking) the minimum is $\sigma_\tau(1-r_+^2)^{1/2}$. The dispersion term $(u-r)^2$ of (a1) vanishes exactly when $u=r$, which requires $r\ge0$.
(a4) $|\sigma_\tau-\mathrm{cPEHE}|\le\operatorname{SD}(\hat\tau)$. Hence the gain over zero satisfies $\sigma_\tau-\mathrm{cPEHE}\le\operatorname{SD}(\hat\tau)=u\,\sigma_\tau$, a bound computable without labels. Also $\mathrm{cPEHE}\ge|1-u|\,\sigma_\tau$.
(a5) $R^2=r^2-(u-r)^2-(\mathbb E\hat\tau-\mathbb E\tau)^2/\operatorname{Var}(\tau)\le r^2$, and $\mathrm{rec}=1-(1-R^2)^{1/2}$.
(a6) If $u>0$, the least-squares slope of $\tau$ on $\hat\tau$ equals $r/u$. Hence $u=r$ if and only if $\operatorname{Cov}(\tau-\hat\tau,\hat\tau)=0$.

\medskip
\textbf{(b) Conditional expectations.} Let $\mathcal G$ be a $\sigma$-field.
(b1) If $\hat\tau=\mathbb E[\tau\mid\mathcal G]$, then $\mathbb E\hat\tau=\mathbb E\tau$, $u=r$, $r^2=R^2=R^2_*(\mathcal G)$ and $\mathrm{rec}=1-(1-r^2)^{1/2}$.
(b2) Every square-integrable $\mathcal G$-measurable $f$ satisfies
\[
\operatorname{corr}(f,\tau)^2\le R^2_*(\mathcal G),\qquad R^2(f)\le R^2_*(\mathcal G),\qquad \mathrm{cPEHE}(f)\ge\sigma_\tau\{1-R^2_*(\mathcal G)\}^{1/2}.
\]
If $\mathcal G\subseteq\mathcal H$ then $R^2_*(\mathcal G)\le R^2_*(\mathcal H)$. Let $X_i$ be the data on patient $i$ that a predictor reads. Suppose $f$ is a function of $X_i$ and of data $T$ independent of $(X_i,\tau_i)$, for example a rule fitted on other, independent patients. Then the same bounds hold with $\mathcal G=\sigma(X_i)$. For the supervised readers and the oracle-structure learner, $X_i=O_i$ as defined in (c). For a learner that also reads the admission row, $X_i$ includes that row.
(b3) $\hat\tau=\mathbb E[\tau\mid\mathcal G]$ for some $\sigma$-field $\mathcal G$ if and only if $\mathbb E[\tau\mid\hat\tau]=\hat\tau$. The condition $u=r$ is necessary for this but not sufficient. Consequently, $u\neq r$ shows that $\hat\tau$ is not the conditional expectation of $\tau$ given any information.

\medskip
\textbf{(c) Transport.} For patient $i$ let:
\begin{itemize}
\item $\mathcal S_i$ be the design (doses of $j^*$ and draw sets);
\item $Z_i$ be the held pre-decision record (rows $0$ to $89$, 12 channels), and $O_i:=(\mathcal S_i,Z_i)$;
\item $Z^0_i$ be the zero-effect record (same panel row, cap $0$);
\item $\zeta_i$ be the trait, with law $\pi_\zeta$;
\item $O_i^+$ be all factual data on patient $i$: the admission covariates, rows $0$ to $89$ of every channel and treatment, the design, and any factual rows recorded after the decision.
\end{itemize}
Let $\mathcal R_i$ be the realized treatment regime on rows $90$ to $101$, and let $\bar1$ be the query regime, in which $j^*$ is given on each of these rows. Under NATMIN and FIX19, $A_i(t)=0$ for every $t\ge54$, so $P(\mathcal R_i=\bar1)=0$: the query regime has propensity zero.

(c1) \emph{Non-identification.} Let $\mathbf Y_i$ be the post-decision potential outcomes of patient $i$ under every regime in both frames. Let $\mathfrak P$ be the set of joint laws of $(O_i^+,\zeta_i,Z_i^0,\mathbf Y_i)$ that
\begin{itemize}
\item satisfy consistency (factual post-decision outcomes equal the frame-$A$ potential outcomes under $\mathcal R_i$), and
\item give $(O_i^+,\zeta_i,Z_i^0)$ its actual law.
\end{itemize}
Then:
\begin{itemize}
\item For every measurable $g$ with $\mathbb E|g(O_i^+)|<\infty$, some law in $\mathfrak P$ has $\mathbb E[\tau_i\mid O_i^+]=g(O_i^+)$ almost surely.
\item For every fixed measurable $f$ with $0<\operatorname{Var}f(O_i^+)<\infty$, every $\delta\ge0$ and every $\varrho\in[-1,1]$, some law in $\mathfrak P$ has $\mathrm{cPEHE}(f)=\delta$, and some has $\operatorname{corr}(f,\tau_i)=\varrho$.
\item Apply the construction to every patient with independent auxiliary draws. This keeps patients independent and identically distributed and leaves every patient's factual data unchanged. Since training uses factual data only, the accuracy statements hold for a learner's rule given any realized training set.
\end{itemize}
Hence the factual law, even together with T2 and T3(i)--(ii), determines neither $\mathbb E[\tau_i\mid O_i^+]$ nor the accuracy of any predictor of $\tau_i$. Every estimator relies on an assumption that links the query regime to the factual law, that is, it transports. In (c2) the assumptions that play this role are T1 and T3(iii).

(c2) \emph{Structural transport.} Assume T2 and T3, and put $m(z\mid s):=\int k(z\mid s,\zeta')\,\pi_\zeta(d\zeta')$. Define
\[
\Pi_\zeta(d\zeta\mid O_i):=\frac{k(Z_i\mid\mathcal S_i,\zeta)\,\pi_\zeta(d\zeta)}{m(Z_i\mid\mathcal S_i)}\quad\text{on }\{m(Z_i\mid\mathcal S_i)>0\},
\]
and $\Pi_\zeta(\cdot\mid O_i):=\pi_\zeta$ on the complement, which has probability zero. Then $\Pi_\zeta$ is a version of the conditional law of $\zeta_i$ given $O_i$, so
\[
\mathbb E[\bar\tau(\zeta_i,\mathcal S_i)\mid O_i]=\int\bar\tau(\zeta,\mathcal S_i)\,\Pi_\zeta(d\zeta\mid O_i).
\]
This is a known function of $O_i$ that uses no outcome observed under the query regime. The query regime enters only through the map $\bar\tau$, which T3(iii) assumes known. If T1 also holds, then $\mathbb E[\tau_i\mid O_i]=\int\bar\tau(\zeta,\mathcal S_i)\,\Pi_\zeta(d\zeta\mid O_i)$. Without T1, take $\bar\tau(\zeta,s):=\mathbb E[\tau_i\mid\zeta_i=\zeta,\mathcal S_i=s]$. Then
\[
\Big\|\,\mathbb E[\tau_i\mid O_i]-\int\bar\tau(\zeta,\mathcal S_i)\,\Pi_\zeta(d\zeta\mid O_i)\Big\|_{L^2}\le\{\mathbb E\operatorname{Var}(\tau_i\mid\zeta_i,\mathcal S_i)\}^{1/2}\le\{\mathbb E\operatorname{Var}(\tau_i\mid\zeta_i)\}^{1/2}.
\]

(c3) \emph{Exposure is necessary.} Suppose the design $s$ has $q_s(d)=0$ at every draw $d$ of every channel $k$ with $\Lambda_k\neq0$. Then $\Pi_\zeta(\cdot\mid O_i)=\pi_\zeta$ almost surely on $\{\mathcal S_i=s\}$, and the transported prediction is the constant $\int\bar\tau(\zeta,s)\,\pi_\zeta(d\zeta)$. This is the Bayesian form of Proposition~4(e).

(c4) \emph{Identification of the trait law.} If $\pi_\zeta$ is not given, the law of $O_i$ identifies it under two conditions:
\begin{itemize}
\item for some design $s$ with $P(\mathcal S_i=s)>0$, some channel $k$ has a draw $d$ with $\Lambda_kq_s(d)\neq0$ at which, almost surely on $\{\mathcal S_i=s\}$, the attenuation is inactive and no clip binds;
\item the characteristic function of $Z^0_{ik}(d)$ given $\mathcal S_i=s$ vanishes only on a Lebesgue-null set.
\end{itemize}
The second condition holds for every bounded zero-effect value, such as that of the implemented, clipped record. Its characteristic function extends to an entire function equal to $1$ at $0$, so its zeros are isolated.

\medskip
\textbf{(d) Empirical-Bayes random slopes under baseline overdispersion.} Fix one channel (or a fixed loading direction), a kernel and the working baseline. For patient $i$ let $\tilde p_i$ and $\tilde r_i$ be the exposure profile and the residual, both whitened by the working covariance. Put $\mathcal A:=\tilde p_i^\top\tilde p_i$ (the same for every patient), $\mathcal B_i:=\tilde p_i^\top\tilde r_i$ and $S_i:=\mathcal B_i/\mathcal A^{1/2}$.
\begin{itemize}
\item \emph{Truth.} The $S_i$ are independent and identically distributed with $S_i=\mathcal A^{1/2}b^*_i+\varepsilon_i$, where $\operatorname{Var}(b^*_i)=\sigma^2_*>0$, $\operatorname{Var}(\varepsilon_i)=\omega_0$ and $\operatorname{Cov}(b^*_i,\varepsilon_i)=0$. Put $\mathrm{SNR}:=\mathcal A\sigma_*^2$ and $V:=\mathrm{SNR}+\omega_0=\operatorname{Var}(S_i)$.
\item \emph{Working model.} $\tilde r_i=\tilde p_i\beta_i+e_i$, with $\beta_i\sim N(\mu,v)$ independent of $e_i\sim N(0,I)$. The pair $(\mu,v)\in\mathbb R\times[0,\infty)$ is fitted by maximum likelihood on $n$ patients, with all other parameters fixed.
\item \emph{Prediction and target.} $\hat\tau_i:=D\,\mathbb E_w[\beta_i\mid\tilde r_i]$ at the fitted values, where $\mathbb E_w$ is the working-model expectation, and $\tau_i=Db^*_i$ with a constant $D\neq0$.
\end{itemize}
(d1) As a function of $(\mu,v)$, the working likelihood depends on the data only through $S_1,\dots,S_n$. The estimates are $\mathcal A^{1/2}\hat\mu=\bar S$ and $1+\mathcal A\hat v=\max(\hat V_n,1)$, where $\hat V_n$ is the sample variance with divisor $n$. As $n\to\infty$, $1+\mathcal A\hat v\to\max(V,1)$ almost surely. Moreover $\mathbb E_w[\beta_i\mid\tilde r_i]=(\mu+v\mathcal B_i)/(1+\mathcal Av)$.

(d2) Take a patient independent of the fitting sample, with the fitted values held fixed, and put $\hat w:=\mathcal A^{1/2}\hat v/(1+\mathcal A\hat v)$. If $\hat v>0$, then exactly $r=(\mathrm{SNR}/V)^{1/2}$ and $u=\hat wV^{1/2}/\sigma_*$. If $\hat v=0$, the prediction is constant, $u=0$, and it does not beat zero. As $n\to\infty$ there are three cases.
\begin{itemize}
\item \emph{$V>1$.} Almost surely $u\to(V-1)/(\mathrm{SNR}\,V)^{1/2}$, so $u/r\to1+(\omega_0-1)/\mathrm{SNR}$. This limit equals the ratio of the limiting fitted shrinkage weight to the weight of the best linear predictor of $b^*_i$ from $S_i$. In the limit:
 \begin{itemize}
 \item $\hat\tau$ is calibrated ($u=r$) if and only if $\omega_0=1$;
 \item $\mathrm{cPEHE}^2/\operatorname{Var}(\tau)\to\{\omega_0\mathrm{SNR}+(\omega_0-1)^2\}/(\mathrm{SNR}\,V)$;
 \item the limiting predictor beats zero if and only if $\omega_0-1<\mathrm{SNR}$. Almost surely, the fitted predictor beats zero for all large $n$ when $\omega_0-1<\mathrm{SNR}$, and fails to for all large $n$ when $\omega_0-1>\mathrm{SNR}$.
 \end{itemize}
 The mean error tends to $D\,\mathbb E\varepsilon_i/\mathcal A^{1/2}$ and affects neither $r$ nor $u$.
\item \emph{$V<1$.} Almost surely $\hat v=0$ for all large $n$, so the prediction is constant and does not beat zero.
\item \emph{$V=1$.} Almost surely $u\to0$. On the event $\{\hat v>0\}$ the predictor beats zero for all large $n$, by a margin of at most $u\sigma_\tau\to0$ (a4). If $\mathbb ES_i^4<\infty$ and $|S_i-\mathbb ES_i|$ is not almost surely constant, this event has probability tending to $1/2$.
\end{itemize}

(d3) \emph{Extension.} Suppose instead that $\tau_i=\tau_0+D_\tau b^*_i+\nu_i$, where $\tau_0$ and $D_\tau$ are constants with $DD_\tau>0$, and $\nu_i$ is square integrable and uncorrelated with $b^*_i$ and $\varepsilon_i$. Let $\rho^2_{\rm lin}:=D_\tau^2\sigma_*^2/\operatorname{Var}(\tau_i)$. For a fitted $\hat v>0$ and a patient independent of the fitting sample, $r=\rho_{\rm lin}(\mathrm{SNR}/V)^{1/2}$ exactly. If $V>1$, then almost surely:
\begin{itemize}
\item $u\to\rho_{\rm lin}(D/D_\tau)(V-1)/(\mathrm{SNR}\,V)^{1/2}$;
\item $u/r\to(D/D_\tau)\{1+(\omega_0-1)/\mathrm{SNR}\}$;
\item the limiting predictor beats zero if and only if $(D/D_\tau)\{1+(\omega_0-1)/\mathrm{SNR}\}<2$.
\end{itemize}

\paragraph{Assumptions.}
\begin{itemize}
\item[(A0)] \emph{Moments.} $\tau,\hat\tau\in L^2$ and $\operatorname{Var}(\tau)>0$. Parts (a) and (b) are exact statements about population moments.
 \begin{itemize}
 \item Part (a) uses only the bilinearity of covariance. It therefore holds exactly for the empirical distribution of the $n$ scored patients with divisor $n$, which is the convention of the shipped metrics tables.
 \item The spread ratio $u$ is shipped as \texttt{prediction\_to\_target\_sd\_ratio} in the metrics files and is read from there.
 \item $u$ cannot be recovered reliably from rounded $(r,\mathrm{cPEHE})$ entries such as those of Supplementary Table S4. By (a1), $u=r\pm\{r^2-1+\mathrm{cPEHE}^2/\operatorname{Var}(\tau)\}^{1/2}$ has two roots, and the inversion is ill-conditioned when $u$ is close to $r$. On S4 the nearer root is off by up to $0.05$, and the choice of root is ambiguous.
 \end{itemize}
\item[(A1)] \emph{Learned predictors.} Given its training folds, a cross-fitted predictor is a function of the test patient's data. (b2) applies when the training data are independent of the test patient's $(X_i,\tau_i)$, as they are with independent patients. Out-of-fold metrics pool several fitted predictors, so they estimate these population quantities with sampling error.
\item[(C0)] \emph{Potential outcomes and dosing.} The target is the frame-by-regime interaction of Definition~4, built from post-decision potential outcomes. Factual post-decision outcomes, if recorded, obey consistency in frame $A$.
 \begin{itemize}
 \item \texttt{natmin\_actions} writes binary doses only on rows $6$ to $53$. It copies the natural column on rows $[6,54)$; the top-up block starts at $6+\mathrm{randint}(0,48-\mathrm{need})$ (exclusive upper end) and so ends by row $52$; the fallback sets idle rows in $[6,54)$. Only the column of $j^*$ is nonzero.
 \item FIX19 doses rows $6$ to $24$.
 \item Hence $A_i(t)=0$ for all $t\ge54$ and every patient of both designs, and $P(\mathcal R_i=\bar1)=0$.
 \item The supervised readers and \texttt{factual\_response} read $O_i$ only. The generic unit learners also read the archived admission row, and the sequence learners read static features. No learner reads a row at or after row $90$.
 \end{itemize}
\item[T1] \emph{Invariance.} $\mathbb E[\tau_i\mid\zeta_i,O_i]=\bar\tau(\zeta_i,\mathcal S_i)$ almost surely: the record carries no information about the query effect beyond the trait that also sets the pre-decision shift. In the generator this holds only approximately. There $\tau_i=\tau^{\rm own}(\zeta_i,e_i(t_0))+\nu_i$, where $\nu_i$ depends on the other four traits and on the post-decision state (Proposition~4(a)). (c2) bounds the resulting error.
\item[T2] \emph{Exogeneity.} $(\mathcal S_i,Z^0_i)$ is independent of $\zeta_i$. In code:
 \begin{itemize}
 \item the trait enters only the bounded shift, which feeds back into no state, policy or memory;
 \item doses are harvested from a policy run on the same real window, which depends on the stay's tape and latent state but not on the trait (verified by reconstructing all 800 NATMIN dose columns from the shipped $u_0$); the design therefore depends on the baseline, and T2 asserts only its independence from the trait;
 \item draw sets are fixed by the source tape.
 \end{itemize}
 The pseudo-random streams of one person are treated as independent draws, the usual idealization.
\item[T3] \emph{Known structure.}
 \begin{itemize}
 \item[(i)] $Z_i=M(Z^0_i,\mathcal S_i,\zeta_i)$ for a known measurable map $M$, and the conditional law $P^0(\cdot\mid s)$ of $Z^0_i$ given $\mathcal S_i=s$ is known. In the generator, $M$ adds the shift $c\,\alpha\tanh(\Lambda_k\zeta_iq_i(d))$ to the held zero-effect record at each draw $d$ of each channel; attenuation and raw clips are known functions of the pre-shift value; and in the factual record only $j^*$ is dosed. With T2, the conditional law of $Z_i$ given $(\mathcal S_i,\zeta_i)=(s,\zeta)$ is then a known kernel. It is assumed to have a jointly measurable density $k(z\mid s,\zeta)$ with respect to a $\sigma$-finite measure $\mu_s$ that does not depend on $\zeta$.
 \begin{itemize}
 \item The existence of such a density is an idealization. It holds when the zero-effect record has no atom at an effective bound at an exposed draw. It fails if, for example, a heart-rate value sits at the source clip ($-4.33$ z, 20 bpm) with positive probability at an exposed draw, because attenuation then moves that atom to a trait-dependent location.
 \item Without the density, the conditional law of $\zeta_i$ given $O_i$ still exists; only the ratio formula for $\Pi_\zeta$ needs the density.
 \item The design takes finitely many values (binary doses by row, and subsets of rows $0$ to $89$ as draw sets), so no measurability question arises in $s$.
 \end{itemize}
 \item[(ii)] The trait law $\pi_\zeta$ is known, or identified as in (c4).
 \item[(iii)] The query map $\bar\tau(\zeta,s)$ is known and $\mathbb E|\bar\tau(\zeta_i,\mathcal S_i)|<\infty$. This map is the only place where the query regime enters.
 \end{itemize}
 T3 is what the oracle-structure learner (\texttt{factual\_response}) is given: the generator's mechanism, kernel, bounded map and loadings, and a particle prior over traits and policy. It is not given $P^0$. It replaces the zero-effect likelihood by a Huber pseudo-likelihood ($\delta=1.345$) on factual increments, with nuisance scales fitted on training persons, so it approximates $\Pi_\zeta$.
\item[(C4)] \emph{For (c4).} At the chosen draw, almost surely on $\{\mathcal S_i=s\}$, the attenuation is inactive ($\alpha=1$) and no clip binds. Also $c>0$, $P^0(\cdot\mid s)$ is known, and its characteristic function at that coordinate is nonzero Lebesgue-almost everywhere; the last condition is automatic for a bounded record. The first condition holds only approximately in the generator, where the attenuation band near the physiologic bounds is reached with small positive probability.
\item[(D)] \emph{For (d).}
 \begin{itemize}
 \item[(D1)] Equal exposure information $\mathcal A$ across patients. In the panels it varies with dosing and draw pattern.
 \item[(D2)] A response linear in the slope along the working exposure direction, and a constant query contrast $D$. On the reconstructed real NATMIN doses the generator-kernel contrast is $0.6838$ to $0.6842$. A fitted kernel can have a different, patient-varying contrast, which (d3) and Proposition~7(d) treat.
 \item[(D3)] A Gaussian working model with unit noise along $\tilde p_i$ after whitening. All other parameters are held at their fitted values: kernel, baseline covariance, fixed effects, and, for \texttt{rs\_dlm} and \texttt{rs\_dlm\_mech}, the noise variance $\sigma^2$, which both profile in closed form in the shared \texttt{\_profile}. At an interior joint optimum the conditional maximizer in $\lambda$ at $\hat\sigma^2$ is the joint one, so this reading is consistent.
 \item[(D4)] Patients independent and identically distributed with finite second moments, in the population limit $n\to\infty$.
 \end{itemize}
 Only second moments of $(b^*_i,\varepsilon_i)$ enter; the truth need not be Gaussian. Code correspondence:
 \begin{itemize}
 \item \emph{ICare-RH stage 2 at a fixed kernel} (\texttt{\_stage2} and \texttt{posterior} in \texttt{revision\_v3\_washout\_response\_hybrid\_core\_v1.py}). The prior is $b_i\sim N(m,1)$ with loading $\gamma$, so $v=\gamma^2$ and $\mu=\gamma m$. The posterior mean is $(m+h_i)/(1+a_i)$, and the prior mean is profiled.
 \item \emph{\texttt{rs\_dlm\_mech}} (in \texttt{revision\_v3\_washout\_standard\_cate\_dlm\_v1.py}). Laird--Ware slope $\beta+b_i$ with $b_i\sim N(0,\sigma_b^2)$. Each person's intercept and linear trend are projected out, and $\lambda=\sigma_b^2/\sigma^2\in[0,100]$ is profiled. The BLUP is $\lambda e^\top(r-X\beta)/(1+\lambda e^\top e)$ and the prediction is $(\beta+\hat b_i)\,dq_i$.
 \item \emph{\texttt{rs\_dlm}} (same file). A random slope on the 12-lag dose sum; the prediction is a constant plus $6.5\,\hat b_i$, so its query scale differs from the target's.
 \item For both random-slope learners, $\tilde p_i=e_i/\hat\sigma$, $\mathcal A_i=e_i^\top e_i/\hat\sigma^2$ and $v=\sigma_b^2$, so (D1) is an idealization.
 \item For both, (d3) is the relevant idealized form. It also needs equal $\mathcal A$ and $\nu_i$ uncorrelated with $(b^*_i,\varepsilon_i)$, and neither holds exactly, since $\nu_i$ depends on the post-decision state.
 \item The upper bound $\lambda\le100$ is assumed not to bind.
 \end{itemize}
\end{itemize}

\paragraph{Proof.}

\paragraph{Conventions.} For $X\in L^2$ write $X^c:=X-\mathbb EX$ and $\|X\|:=(\mathbb EX^2)^{1/2}$, so $\operatorname{SD}(X)=\|X^c\|$ and $\operatorname{Cov}(X,Y)=\mathbb E[X^cY^c]$. Let $s:=\operatorname{SD}(\hat\tau)=u\sigma_\tau$. With the convention $r:=0$ when $s=0$, in every case
\begin{equation}\label{eq:P6cov}
\operatorname{Cov}(\hat\tau,\tau)=r\,s\,\sigma_\tau=r\,u\,\sigma_\tau^2 .
\end{equation}
Part (a) uses only (\ref{eq:P6cov}) and bilinearity, so it holds for any probability measure, in particular for the empirical distribution of $n$ pairs with divisor $n$.

\paragraph{Proof of (a).}
(a1) By (\ref{eq:P6cov}),
\[
\mathrm{cPEHE}^2=\operatorname{Var}(\hat\tau-\tau)=s^2-2\operatorname{Cov}(\hat\tau,\tau)+\sigma_\tau^2=\sigma_\tau^2(u^2-2ru+1),
\]
and $u^2-2ru+1=(1-r^2)+(u-r)^2$.

(a2) The zero prediction has $\mathrm{cPEHE}=\operatorname{SD}(-\tau)=\sigma_\tau$. By (a1), $\mathrm{cPEHE}<\sigma_\tau$ if and only if $u^2-2ru<0$, that is $u(u-2r)<0$. Since $u\ge0$, this holds if and only if $u>0$ and $u<2r$.

(a3) For $\theta_0,\theta_1\in\mathbb R$,
\[
Q(\theta_1):=\operatorname{Var}(\theta_0+\theta_1\hat\tau-\tau)=\theta_1^2s^2-2\theta_1rs\sigma_\tau+\sigma_\tau^2 ,
\]
which does not depend on $\theta_0$.
\begin{itemize}
\item If $s=0$, then $r=0$ and $Q\equiv\sigma_\tau^2=\sigma_\tau^2(1-r^2)$.
\item If $s>0$, $Q$ is a convex quadratic with minimizer $\theta_1^*=r\sigma_\tau/s=r/u$ and $Q(\theta_1^*)=\sigma_\tau^2(1-r^2)$. The recalibrated predictor $\tilde\tau:=\theta_0+\theta_1^*\hat\tau$ has $\operatorname{SD}(\tilde\tau)=|\theta_1^*|s=|r|\sigma_\tau$. When $r\neq0$, $\operatorname{corr}(\tilde\tau,\tau)=\operatorname{sign}(\theta_1^*)\,r=|r|$; when $r=0$ both equal $0$.
\end{itemize}
On $\theta_1\ge0$:
\begin{itemize}
\item if $r\ge0$, then $\theta_1^*\ge0$ and the minimum is unchanged;
\item if $r<0$, then $Q'(\theta_1)=2\theta_1s^2-2rs\sigma_\tau>0$ on $[0,\infty)$, and the minimum is $Q(0)=\sigma_\tau^2$.
\end{itemize}
In both cases the minimum equals $\sigma_\tau^2(1-r_+^2)$. Finally, the term $(u-r)^2$ in (a1) is zero exactly when $u=r$. Since $u\ge0$, this cannot happen when $r<0$.

(a4) We have $\mathrm{cPEHE}=\|\hat\tau^c-\tau^c\|$, $\sigma_\tau=\|\tau^c\|$ and $\operatorname{SD}(\hat\tau)=\|\hat\tau^c\|$. The triangle inequality in $L^2$ gives $|\,\|\tau^c\|-\|\hat\tau^c-\tau^c\|\,|\le\|\hat\tau^c\|$, and also $\|\hat\tau^c-\tau^c\|\ge|\,\|\hat\tau^c\|-\|\tau^c\|\,|=|u-1|\sigma_\tau$.

(a5) $\mathbb E(\hat\tau-\tau)^2=\operatorname{Var}(\hat\tau-\tau)+(\mathbb E\hat\tau-\mathbb E\tau)^2$. Dividing by $\operatorname{Var}(\tau)$ and using (a1) gives
\[
R^2=r^2-(u-r)^2-(\mathbb E\hat\tau-\mathbb E\tau)^2/\operatorname{Var}(\tau).
\]
Since $\{\mathbb E(\hat\tau-\tau)^2\}^{1/2}/\sigma_\tau=(1-R^2)^{1/2}$, we get $\mathrm{rec}=1-(1-R^2)^{1/2}$.

(a6) If $s>0$, the least-squares slope is $\operatorname{Cov}(\tau,\hat\tau)/\operatorname{Var}(\hat\tau)=rs\sigma_\tau/s^2=r/u$. Also $\operatorname{Cov}(\tau-\hat\tau,\hat\tau)=rs\sigma_\tau-s^2=s\sigma_\tau(r-u)$, which is zero if and only if $u=r$. \hfill$\square$

\paragraph{Proof of (b).}
Let $\hat\tau:=\mathbb E[\tau\mid\mathcal G]$; $\hat\tau\in L^2$ by the conditional Jensen inequality. For every $\mathcal G$-measurable $f\in L^2$, $\mathbb E[f\tau]=\mathbb E[\mathbb E[f\tau\mid\mathcal G]]=\mathbb E[f\hat\tau]$, and $f\equiv1$ gives $\mathbb E\hat\tau=\mathbb E\tau$. Hence
\begin{equation}\label{eq:P6orth}
\operatorname{Cov}(f,\tau)=\operatorname{Cov}(f,\hat\tau),\qquad \mathbb E[f(\tau-\hat\tau)]=0\quad\text{for every $\mathcal G$-measurable }f\in L^2 .
\end{equation}

(b1) Take $f=\hat\tau$ in (\ref{eq:P6orth}): $\operatorname{Cov}(\hat\tau,\tau)=\operatorname{Var}(\hat\tau)$, that is $ru\sigma_\tau^2=u^2\sigma_\tau^2$ by (\ref{eq:P6cov}), so $u(u-r)=0$. If $u>0$ then $u=r$; if $u=0$ then $r=0=u$ by convention. Hence $r^2=u^2=\operatorname{Var}(\hat\tau)/\operatorname{Var}(\tau)=R^2_*(\mathcal G)$. The mean error is zero, so (a5) gives $R^2=r^2$ and $\mathrm{rec}=1-(1-r^2)^{1/2}$.

(b2) Let $f\in L^2$ be $\mathcal G$-measurable.
\begin{itemize}
\item \emph{Correlation.} If $\operatorname{Var}f=0$, then $\operatorname{corr}(f,\tau)=0$ by convention. Otherwise, by (\ref{eq:P6orth}) and the Cauchy--Schwarz inequality, $|\operatorname{Cov}(f,\tau)|=|\operatorname{Cov}(f,\hat\tau)|\le\operatorname{SD}(f)\operatorname{SD}(\hat\tau)$, so $\operatorname{corr}(f,\tau)^2\le R^2_*(\mathcal G)$.
\item \emph{$R^2$.} $\hat\tau-f$ is $\mathcal G$-measurable, so by (\ref{eq:P6orth})
\[
\mathbb E(\tau-f)^2=\mathbb E(\tau-\hat\tau)^2+\mathbb E(\hat\tau-f)^2\ge\mathbb E(\tau-\hat\tau)^2 .
\]
Because $\mathbb E[\hat\tau(\tau-\hat\tau)]=0$ and $\mathbb E\tau=\mathbb E\hat\tau$, we have $\mathbb E(\tau-\hat\tau)^2=\mathbb E\tau^2-\mathbb E\hat\tau^2=\operatorname{Var}(\tau)-\operatorname{Var}(\hat\tau)$. Dividing by $\operatorname{Var}(\tau)$ gives $R^2(f)\le R^2_*(\mathcal G)$.
\item \emph{cPEHE.} $\tau-\hat\tau$ has mean zero and $f-\hat\tau$ is $\mathcal G$-measurable, so (\ref{eq:P6orth}) gives $\operatorname{Cov}(f-\hat\tau,\tau-\hat\tau)=0$. Hence
\[
\operatorname{Var}(f-\tau)=\operatorname{Var}(f-\hat\tau)+\operatorname{Var}(\tau-\hat\tau)\ge\operatorname{Var}(\tau)\{1-R^2_*(\mathcal G)\},
\]
which is the bound on $\mathrm{cPEHE}(f)$.
\item \emph{Monotonicity.} If $\mathcal G\subseteq\mathcal H$, then $\mathbb E[\tau\mid\mathcal G]=\mathbb E[\mathbb E[\tau\mid\mathcal H]\mid\mathcal G]$. By the conditional Jensen inequality, $\mathbb E(\mathbb E[\tau\mid\mathcal G])^2\le\mathbb E(\mathbb E[\tau\mid\mathcal H])^2$. Both have mean $\mathbb E\tau$, so $\operatorname{Var}\mathbb E[\tau\mid\mathcal G]\le\operatorname{Var}\mathbb E[\tau\mid\mathcal H]$.
\item \emph{Independent training data.} Let $T$ be independent of $(X_i,\tau_i)$. For bounded measurable $\varphi,\psi$,
\[
\mathbb E[\tau_i\varphi(X_i)\psi(T)]=\mathbb E[\tau_i\varphi(X_i)]\,\mathbb E\psi(T)=\mathbb E[\mathbb E[\tau_i\mid X_i]\varphi(X_i)]\,\mathbb E\psi(T)=\mathbb E[\mathbb E[\tau_i\mid X_i]\varphi(X_i)\psi(T)].
\]
The sets $\{X_i\in A\}\cap\{T\in B\}$ form a $\pi$-system generating $\sigma(X_i,T)$. By the $\pi$--$\lambda$ theorem, $\mathbb E[\tau_i\mid X_i,T]=\mathbb E[\tau_i\mid X_i]$, so $R^2_*(\sigma(X_i,T))=R^2_*(\sigma(X_i))$. A predictor $f(X_i,T)$ is $\sigma(X_i,T)$-measurable, so the bounds above apply with $\mathcal G=\sigma(X_i)$. Which data $X_i$ a given predictor reads is recorded in (C0).
\end{itemize}

(b3) If $\hat\tau=\mathbb E[\tau\mid\mathcal G]$, then $\sigma(\hat\tau)\subseteq\mathcal G$ and, by the tower property, $\mathbb E[\tau\mid\hat\tau]=\mathbb E[\mathbb E[\tau\mid\mathcal G]\mid\hat\tau]=\hat\tau$. Conversely, if $\mathbb E[\tau\mid\hat\tau]=\hat\tau$, take $\mathcal G=\sigma(\hat\tau)$. By (b1) every conditional expectation has $u=r$, so $u\neq r$ excludes every $\mathcal G$. The condition $u=r$ is not sufficient. Take $X\sim N(0,1)$, $\hat\tau:=X$ and $\tau:=X+X^2-1$. Then $\operatorname{Cov}(\tau,\hat\tau)=1=\operatorname{Var}(\hat\tau)$, so the slope in (a6) is $1$ and $u=r$ ($=3^{-1/2}$), while $\mathbb E[\tau\mid\hat\tau]=X+X^2-1\neq\hat\tau$. \hfill$\square$

\paragraph{Proof of (c1).}
\emph{Setup.} Let $P$ be the actual joint law. Let $\mathbf Y_i$ contain the potential outcomes on rows $91$ to $102$ for every post-decision regime and both frames. $\tau_i$ is the fixed linear combination of Definition~4; its frame-$A$, regime-$\bar1$ coordinates $Y^A_{is}(\bar1)$ enter with coefficient $1/h$ each. Every coordinate of $O_i^+$ is of one of two kinds:
\begin{itemize}
\item pre-decision (admission covariates, rows $0$ to $89$, the design), which involves no post-decision potential outcome; or
\item a factual post-decision outcome, which by consistency is a frame-$A$ coordinate of $\mathbf Y_i$ under $\mathcal R_i$.
\end{itemize}
Since $P(\mathcal R_i=\bar1)=0$ (C0), on a set of probability one no coordinate of $O_i^+$ involves any $Y^A_{is}(\bar1)$.

\emph{Construction.} Let $\Delta$ be any integrable variable measurable with respect to $(O_i^+,\mathbf Y_i,\xi)$. Here $\xi$ is a standard normal variable independent of everything else, obtained by a product extension of the probability space that leaves the law of $(O_i^+,\zeta_i,Z_i^0,\mathbf Y_i)$ unchanged. Define $\mathbf Y_i'$ from $\mathbf Y_i$ by
\[
Y'^A_{is}(\bar1):=Y^A_{is}(\bar1)+\Delta\quad(s=0,\dots,h-1),\qquad\text{all other coordinates unchanged.}
\]
The variables $(O_i^+,\zeta_i,Z^0_i)$ are not changed, so their law is the actual one. Consistency still holds on $\{\mathcal R_i\neq\bar1\}$, a set of probability one. The law of $(O_i^+,\zeta_i,Z_i^0,\mathbf Y_i')$ therefore lies in $\mathfrak P$, and the target it defines is $\tau_i'=\tau_i+\Delta$.

\emph{Conditional mean.} With $\Delta:=g(O_i^+)-\mathbb E[\tau_i\mid O_i^+]$, we get $\mathbb E[\tau_i'\mid O_i^+]=\mathbb E[\tau_i\mid O_i^+]+\Delta=g(O_i^+)$ almost surely.

\emph{Accuracy.}
\begin{itemize}
\item Take $\Delta:=f(O_i^+)+\delta\xi-\tau_i$. Then $\tau_i'=f+\delta\xi$, so $\mathrm{cPEHE}(f)=\operatorname{SD}(\delta\xi)=\delta$, and $\operatorname{Var}(\tau_i')=\operatorname{Var}f+\delta^2>0$.
\item Take $\Delta:=\varrho f^c/\operatorname{SD}(f)+(1-\varrho^2)^{1/2}\xi-\tau_i$. Then $\operatorname{Var}(\tau_i')=1$ and $\operatorname{Cov}(f,\tau_i')=\varrho\operatorname{SD}(f)$, so $\operatorname{corr}(f,\tau_i')=\varrho$.
\end{itemize}

\emph{Several patients and learned rules.} Suppose that under $P$ the patients are independent and identically distributed. Fix one measurable map $F$ and set $\Delta_i:=F(O_i^+,\mathbf Y_i,\xi_i)$ for every patient, with $\xi_1,\xi_2,\dots$ independent standard normals that are independent of everything else. The modified patient vectors are one fixed function of independent and identically distributed vectors, so they are again independent and identically distributed. Every patient's factual data are unchanged pointwise. A learner trained on factual data therefore returns the same rule under the modified law. Given a realized training set, that rule is a fixed function $f$. Using the $F$ of the previous step for this $f$ gives a law in $\mathfrak P$, independent and identically distributed across patients and with the actual factual law, under which a new patient has $\mathrm{cPEHE}(f)=\delta$ or $\operatorname{corr}(f,\tau_i)=\varrho$.

\emph{Assumptions.} T2 and T3(i)--(ii) restrict only the law of $(O_i,\zeta_i,Z^0_i)$, which the construction leaves unchanged. Every constructed law therefore satisfies them if the actual law does. T1 and T3(iii) restrict the query regime directly: they fix $\mathbb E[\tau_i\mid\zeta_i,O_i]$ as a known function. A construction with $\mathbb E[\Delta\mid\zeta_i,O_i]\neq0$ violates them.

\emph{Structural zero at $s=0$.} If one also requires the $s=0$ term of $\tau_i$ to vanish (Proposition~4(a)), modify only $Y^A_{is}(\bar1)$ for $s=1,\dots,h-1$, each by $h\Delta/(h-1)$. This again gives $\tau_i'=\tau_i+\Delta$, and nothing else changes.

\emph{Conclusion.} The factual law thus fixes neither $\mathbb E[\tau_i\mid O_i^+]$, nor $\mathrm{cPEHE}(f)$, nor $\operatorname{corr}(f,\tau_i)$. A guarantee for any estimator must restrict how the query-regime outcomes relate to the factual variables. The usual route by overlap, $\mathbb E[Y(\bar1)\mid X]=\mathbb E[Y\mid\mathcal R=\bar1,X]$, needs $P(\mathcal R=\bar1\mid X)>0$ and is unavailable. The zero-propensity fact is needed only when factual post-decision outcomes are part of the data. For data that contain no post-decision row, the construction applies directly. Since $\sigma(O_i)\subseteq\sigma(O_i^+)$, the non-identification of $\mathbb E[\tau_i\mid O_i]$ is the case where $g$ is a function of $O_i$. \hfill$\square$

\paragraph{Proof of (c2).}
\emph{Joint law.} Let $P_{\mathcal S}$ be the law of $\mathcal S_i$. It has finite support (T3(i)), so integrals over $s$ are finite sums over designs with $P(\mathcal S_i=s)>0$. By T2, $\zeta_i$ is independent of $\mathcal S_i$, and $Z_i^0$ is independent of $\zeta_i$ given $\mathcal S_i$, with conditional law $P^0(\cdot\mid s)$. By T3(i), given $(\mathcal S_i,\zeta_i)=(s,\zeta)$, the record $Z_i=M(Z_i^0,s,\zeta)$ has the law of $M(\cdot,s,\zeta)$ under $P^0(\cdot\mid s)$, with density $k(\cdot\mid s,\zeta)$ with respect to $\mu_s$. Hence the joint law of $(\mathcal S_i,\zeta_i,Z_i)$ is
\[
P_{\mathcal S}(ds)\,\pi_\zeta(d\zeta)\,k(z\mid s,\zeta)\,\mu_s(dz).
\]

\emph{Null sets.} Let $N:=\{(s,z):m(z\mid s)=0\}$. Then
\[
P(O_i\in N)=\sum_sP_{\mathcal S}(s)\int\mathbf 1_N(s,z)\,m(z\mid s)\,\mu_s(dz)=0,
\]
and on $N$, $k(z\mid s,\zeta)=0$ for $\pi_\zeta$-almost every $\zeta$. Now fix a jointly measurable $h(\zeta,s)$ with $\mathbb E|h(\zeta_i,\mathcal S_i)|<\infty$, and let
\[
N_h:=\Big\{(s,z):\int|h(\zeta,s)|k(z\mid s,\zeta)\pi_\zeta(d\zeta)=\infty\Big\}.
\]
By Tonelli's theorem,
\[
\mathbb E|h(\zeta_i,\mathcal S_i)|=\sum_sP_{\mathcal S}(s)\int\mu_s(dz)\int|h(\zeta,s)|k(z\mid s,\zeta)\pi_\zeta(d\zeta)<\infty,
\]
so $\mu_s(\{z:(s,z)\in N_h\})=0$ for every $s$ in the support, and $P(O_i\in N_h)=0$.

\emph{Posterior formula.} Set
\[
H(s,z):=\frac{\int h(\zeta,s)k(z\mid s,\zeta)\pi_\zeta(d\zeta)}{m(z\mid s)}\quad\text{off }N\cup N_h,
\]
and $H:=0$ on $N\cup N_h$. On $N$ the inner integral is $0=H\,m$, and $N_h$ is $\mu_s$-null. So for bounded measurable $\varphi$, Fubini's theorem gives
\begin{align*}
\mathbb E[h(\zeta_i,\mathcal S_i)\varphi(O_i)]&=\sum_sP_{\mathcal S}(s)\int\varphi(s,z)\Big[\int h(\zeta,s)k(z\mid s,\zeta)\pi_\zeta(d\zeta)\Big]\mu_s(dz)\\
&=\sum_sP_{\mathcal S}(s)\int\varphi(s,z)H(s,z)m(z\mid s)\mu_s(dz)=\mathbb E[H(O_i)\varphi(O_i)].
\end{align*}
Hence $H(O_i)=\mathbb E[h(\zeta_i,\mathcal S_i)\mid O_i]$ almost surely. Off the null set $N\cup N_h$, $H(O_i)=\int h(\zeta,\mathcal S_i)\,\Pi_\zeta(d\zeta\mid O_i)$. Taking $h=\mathbf 1_B(\zeta)$, which is bounded and has $N_h=\emptyset$, shows that $\Pi_\zeta$ is a version of the conditional law of $\zeta_i$ given $O_i$; on $N$, any choice, such as $\pi_\zeta$, is allowed. With $h=\bar\tau$ we obtain the displayed formula. $\Pi_\zeta$ is computed from $O_i$, $k$ and $\pi_\zeta$, none of which involves an outcome under the query regime. The query regime enters only through $\bar\tau$, which T3(iii) assumes known.

\emph{With T1.} The tower property ($\sigma(O_i)\subseteq\sigma(\zeta_i,O_i)$) gives
\[
\mathbb E[\tau_i\mid O_i]=\mathbb E[\mathbb E[\tau_i\mid\zeta_i,O_i]\mid O_i]=\mathbb E[\bar\tau(\zeta_i,\mathcal S_i)\mid O_i].
\]

\emph{Without T1.} Take $\bar\tau(\zeta,s):=\mathbb E[\tau_i\mid\zeta_i=\zeta,\mathcal S_i=s]$, with $\tau_i\in L^2$. Then
\[
\mathbb E[\tau_i\mid O_i]-\mathbb E[\bar\tau(\zeta_i,\mathcal S_i)\mid O_i]=\mathbb E[\tau_i-\bar\tau(\zeta_i,\mathcal S_i)\mid O_i].
\]
Conditional expectation is a contraction in $L^2$, so the norm of this difference is at most $\|\tau_i-\bar\tau(\zeta_i,\mathcal S_i)\|=\{\mathbb E\operatorname{Var}(\tau_i\mid\zeta_i,\mathcal S_i)\}^{1/2}$. This is in turn at most $\{\mathbb E\operatorname{Var}(\tau_i\mid\zeta_i)\}^{1/2}=\|\tau_i-\mathbb E[\tau_i\mid\zeta_i]\|$: $\mathbb E[\tau_i\mid\zeta_i,\mathcal S_i]$ is the $\sigma(\zeta_i,\mathcal S_i)$-measurable variable closest to $\tau_i$ in $L^2$, and $\mathbb E[\tau_i\mid\zeta_i]$ is one such variable. \hfill$\square$

\paragraph{Proof of (c3).}
In the factual record only $j^*$ is dosed (C0). So at a draw $d$ of channel $k$ the tanh argument is $\Gamma_{ik}(d)=\Lambda_k\zeta_iq_s(d)$. Under the hypothesis it is zero at every draw of every channel; for channels with $\Lambda_k=0$ this is trivial. The shift $c\,\alpha\tanh(0)=0$ at every draw, whatever the attenuation, and the hold copies draw values, so $M(z^0,s,\zeta)=z^0$ for every $\zeta$. (In code, a zero shift returns the base value unchanged.) Hence:
\begin{itemize}
\item $k(\cdot\mid s,\zeta)$ is the density of $P^0(\cdot\mid s)$ for every $\zeta$, so $k(z\mid s,\zeta)=m(z\mid s)$;
\item $\Pi_\zeta(d\zeta\mid O_i)=\pi_\zeta(d\zeta)$ on $\{\mathcal S_i=s\}$, both off $N$ and, by the convention of (c2), on $N$;
\item the prediction of (c2) is $\int\bar\tau(\zeta,s)\pi_\zeta(d\zeta)$. \hfill$\square$
\end{itemize}

\paragraph{Proof of (c4).}
Fix $s$ and $(k,d)$ as in the hypothesis, and write $a:=\Lambda_kq_s(d)\neq0$.
\begin{itemize}
\item Almost surely on $\{\mathcal S_i=s\}$, the attenuation is inactive and no clip binds. So there $Z_{ik}(d)=Z^0_{ik}(d)+W_i$, with $W_i:=c\tanh(a\zeta_i)$.
\item By T2, $Z^0_{ik}(d)$ and $\zeta_i$ are independent given $\mathcal S_i=s$, and $\zeta_i$ has law $\pi_\zeta$ given $\mathcal S_i=s$. Hence the conditional characteristic functions satisfy $\phi_{Z}(t)=\phi_0(t)\,\psi(t)$, with $\psi(t):=\int\exp\{itc\tanh(a\zeta)\}\pi_\zeta(d\zeta)$.
\item Because $P(\mathcal S_i=s)>0$, $\phi_Z$ is determined by the law of $O_i$, and $\phi_0$ is known by T3(i). So $\psi=\phi_Z/\phi_0$ is determined on $\{\phi_0\neq0\}$. This set has full Lebesgue measure and is therefore dense. $\psi$ is continuous, so it is determined everywhere, and by the uniqueness theorem the law of $W_i$ is determined.
\item The map $\zeta\mapsto c\tanh(a\zeta)$ is a continuous, strictly monotone bijection of $\mathbb R$ onto $(-c,c)$. So $\pi_\zeta$ is the image of the law of $W_i$ under its inverse.
\end{itemize}
It remains to show that a bounded record satisfies the characteristic-function condition. If $|Z^0_{ik}(d)|\le B$ almost surely given $\mathcal S_i=s$, then $\phi_0(t)=\mathbb E[e^{itZ^0_{ik}(d)}\mid\mathcal S_i=s]$ extends to $\phi_0(w)=\mathbb E[e^{iwZ^0_{ik}(d)}\mid\mathcal S_i=s]$ for complex $w$. By dominated differentiation, using $|e^{iwx}|\le e^{|w|B}$ for $|x|\le B$, this extension is entire. Since $\phi_0(0)=1$, it is not identically zero, so its zeros are isolated. Its real zeros are therefore countable and form a Lebesgue-null set. \hfill$\square$

\paragraph{Proof of (d).}
(d1) Fix $i$ and an orthogonal matrix $U_i$ whose first row is $\tilde p_i^\top/\mathcal A^{1/2}$. Then $U_i\tilde r_i=(S_i,W_i^\top)^\top$ with $S_i=\mathcal B_i/\mathcal A^{1/2}$. Under the working model,
\[
U_i\tilde r_i=(\mathcal A^{1/2}\beta_i,0,\dots,0)^\top+U_ie_i,\qquad U_ie_i\sim N(0,I).
\]
So $S_i=\mathcal A^{1/2}\beta_i+(U_ie_i)_1$, and $W_i=(U_ie_i)_{2:}\sim N(0,I)$ is independent of $(\beta_i,S_i)$ and free of $(\mu,v)$. The Jacobian of $U_i$ is one, so the working density of $\tilde r_i$ is $N(S_i;\mathcal A^{1/2}\mu,1+\mathcal Av)\cdot N(W_i;0,I)$. The working likelihood of $n$ independent patients is therefore $\prod_iN(S_i;\mathcal A^{1/2}\mu,1+\mathcal Av)$ times a factor free of $(\mu,v)$.

Write $\eta:=\mathcal A^{1/2}\mu$ and $x:=1+\mathcal Av\in[1,\infty)$. The log-likelihood is $-\frac12\sum_i\{(S_i-\eta)^2/x+\log x\}+\text{const}$.
\begin{itemize}
\item For every $x$ it is maximized over $\eta$ at $\bar S$.
\item The profile $-\frac n2\{\hat V_n/x+\log x\}$ has derivative $\frac n{2x^2}(\hat V_n-x)$. It increases for $x<\hat V_n$ and decreases for $x>\hat V_n$, so its maximizer on $[1,\infty)$ is $\max(\hat V_n,1)$.
\item The $S_i$ are independent and identically distributed with finite variance, so $\hat V_n\to V$ almost surely by the strong law of large numbers. Since $\max(\cdot,1)$ is continuous, $1+\mathcal A\hat v\to\max(V,1)$ almost surely.
\end{itemize}
Finally, $W_i$ is independent of $(\beta_i,S_i)$ under the working model, and $\operatorname{Cov}_w(\beta_i,S_i)=\mathcal A^{1/2}v$, $\operatorname{Var}_w(S_i)=1+\mathcal Av$. Gaussian regression gives
\[
\mathbb E_w[\beta_i\mid\tilde r_i]=\mathbb E_w[\beta_i\mid S_i]=\mu+\frac{\mathcal A^{1/2}v}{1+\mathcal Av}(S_i-\mathcal A^{1/2}\mu)=\frac{\mu+v\mathcal B_i}{1+\mathcal Av}.
\]
In ICare-RH's parametrization ($\beta_i=\gamma b_i$, $b_i\sim N(m,1)$, $v=\gamma^2$, $\mu=\gamma m$), this is $\gamma(m+h_i)/(1+a_i)$ with $h_i=\gamma\mathcal B_i$ and $a_i=\gamma^2\mathcal A$. For the Laird--Ware learners ($\tilde p_i=e_i/\hat\sigma$, $v=\sigma_b^2$, $\lambda=\sigma_b^2/\hat\sigma^2$), the deviation from $\mu$ is $\lambda e_i^\top(r_i-e_i\mu)/(1+\lambda e_i^\top e_i)$, which is the coded BLUP.

(d2) Fix $(\hat\mu,\hat v)$ and let patient $i$ be independent of the fitting sample. The moments below are over $(b^*_i,\varepsilon_i)$ with $(\hat\mu,\hat v)$ fixed. The prediction is
\[
\hat\tau_i=D\hat\mu+D\hat w(S_i-\mathcal A^{1/2}\hat\mu),
\]
so $\hat\tau_i^c=D\hat wS_i^c$ and $\tau_i^c=Db_i^{*c}$.
\begin{itemize}
\item If $\hat v=0$, then $\hat w=0$, the prediction is constant, $u=0$ and $\mathrm{cPEHE}=\sigma_\tau$.
\item If $\hat v>0$, then $\hat w>0$. Since $\operatorname{Cov}(S_i,b^*_i)=\mathcal A^{1/2}\sigma_*^2+\operatorname{Cov}(\varepsilon_i,b^*_i)=\mathcal A^{1/2}\sigma_*^2$ and $\operatorname{Var}(S_i)=\mathcal A\sigma_*^2+\omega_0=V$,
\[
r=\operatorname{sign}(D^2\hat w)\operatorname{corr}(S_i,b^*_i)=\frac{\mathcal A^{1/2}\sigma_*^2}{V^{1/2}\sigma_*}=\Big(\frac{\mathrm{SNR}}V\Big)^{1/2},\qquad u=\frac{|D|\hat wV^{1/2}}{|D|\sigma_*}=\frac{\hat wV^{1/2}}{\sigma_*}.
\]
\end{itemize}

\emph{Case $V>1$.} By (d1), $1+\mathcal A\hat v\to V$ almost surely. Hence $\hat w\to w^\dagger:=\mathcal A^{1/2}(V-1)/(\mathcal AV)=(V-1)/(\mathcal A^{1/2}V)>0$, and $\hat v>0$ for all large $n$. Then
\[
u\to\frac{V-1}{\mathcal A^{1/2}\sigma_*V^{1/2}}=\frac{V-1}{(\mathrm{SNR}\,V)^{1/2}},\qquad\frac ur\to\frac{V-1}{\mathrm{SNR}}=1+\frac{\omega_0-1}{\mathrm{SNR}}.
\]
\begin{itemize}
\item \emph{Shrinkage weight.} The best linear predictor of $b_i^*$ from $S_i$ has weight
\[
w^{\rm BL}:=\operatorname{Cov}(b^*_i,S_i)/\operatorname{Var}(S_i)=\mathcal A^{1/2}\sigma_*^2/V=\mathrm{SNR}/(\mathcal A^{1/2}V),
\]
so $w^\dagger/w^{\rm BL}=(V-1)/\mathrm{SNR}$, the limit of $u/r$.
\item \emph{Calibration.} Since $\mathrm{SNR}>0$, the limit has $u=r$ if and only if $\omega_0=1$.
\item \emph{Beating zero.} The limiting values have $u>0$ and $r>0$, so by (a2) the limiting predictor beats zero if and only if $u/r<2$, that is $\omega_0-1<\mathrm{SNR}$. $r$ is constant once $\hat v>0$ and $u$ converges almost surely. So if the limit of $u$ is below $2r$, then $0<u<2r$ for all large $n$; if it is above, then $u>2r$ for all large $n$.
\item \emph{cPEHE.} By (a1), the limiting $\mathrm{cPEHE}^2/\operatorname{Var}(\tau)$ is
\[
1-r^2+r^2(u/r-1)^2=\frac{\omega_0}V+\frac{\mathrm{SNR}\,(\omega_0-1)^2}{V\,\mathrm{SNR}^2}=\frac{\omega_0\mathrm{SNR}+(\omega_0-1)^2}{\mathrm{SNR}\,V}.
\]
\item \emph{Mean error.} $\mathbb E\hat\tau_i=D\hat\mu$ (because $\mathbb ES_i-\mathcal A^{1/2}\hat\mu$ is multiplied by $D\hat w$ and $\hat\mu\to\mathbb ES_i/\mathcal A^{1/2}$), so it tends to $D\,\mathbb ES_i/\mathcal A^{1/2}=D\,\mathbb Eb^*_i+D\,\mathbb E\varepsilon_i/\mathcal A^{1/2}$, while $\mathbb E\tau_i=D\,\mathbb Eb^*_i$. The mean error enters neither $r$ nor $u$, which are centered quantities.
\end{itemize}

\emph{Case $V<1$.} $\hat V_n\to V<1$ almost surely, so almost surely $\hat V_n<1$ for all large $n$. Then $\hat v=0$ and the prediction is the constant $D\hat\mu$, which does not beat zero.

\emph{Case $V=1$.} $1+\mathcal A\hat v=\max(\hat V_n,1)\to1$ almost surely, so $\hat w\to0$ and $u=\hat w/\sigma_*\to0$ almost surely.
\begin{itemize}
\item On $\{\hat v=0\}$ the prediction is constant and does not beat zero.
\item On $\{\hat v>0\}$, $r=\mathrm{SNR}^{1/2}>0$ is fixed. Almost surely $u<2r$ for all large $n$, and then (a2) says the predictor beats zero. By (a4) the gain is at most $u\sigma_\tau\to0$.
\item Write $\hat V_n=n^{-1}\sum_i(S_i-\mathbb ES_i)^2-(\bar S-\mathbb ES_i)^2$. If $\mathbb ES_i^4<\infty$ and $|S_i-\mathbb ES_i|$ is not almost surely constant, then $(S_i-\mathbb ES_i)^2$ has mean $1$ and positive finite variance. By the central limit theorem, $n^{1/2}\{n^{-1}\sum_i(S_i-\mathbb ES_i)^2-1\}$ converges to a centered normal law, while $n^{1/2}(\bar S-\mathbb ES_i)^2\to0$ in probability. Hence $P(\hat v>0)=P(\hat V_n>1)\to1/2$.
\end{itemize}

(d3) Now $\tau_i^c=D_\tau b_i^{*c}+\nu_i^c$ and $\operatorname{Var}(\tau_i)=D_\tau^2\sigma_*^2/\rho_{\rm lin}^2$. Because $\nu_i$ is uncorrelated with $b^*_i$ and $\varepsilon_i$, $\operatorname{Cov}(S_i,\nu_i)=0$ and $\operatorname{Cov}(\hat\tau_i,\tau_i)=D\hat wD_\tau\mathcal A^{1/2}\sigma_*^2$. Hence, for a fitted $\hat v>0$ (so $\hat w>0$) and $DD_\tau>0$,
\[
r=\frac{D\hat wD_\tau\mathcal A^{1/2}\sigma_*^2}{|D|\hat wV^{1/2}\,|D_\tau|\sigma_*/\rho_{\rm lin}}=\rho_{\rm lin}\Big(\frac{\mathrm{SNR}}V\Big)^{1/2},\qquad u=\frac{|D|\hat wV^{1/2}\rho_{\rm lin}}{|D_\tau|\sigma_*}.
\]
If $V>1$, then $\hat w\to w^\dagger$ almost surely as in (d2). So $u\to\rho_{\rm lin}(D/D_\tau)(V-1)/(\mathrm{SNR}\,V)^{1/2}$ and $u/r\to(D/D_\tau)(V-1)/\mathrm{SNR}$. Both limits are positive, and by (a2) the limiting predictor beats zero if and only if this ratio is below $2$. \hfill$\square$

\paragraph{Numerical checks.} The identities of (a) were checked on all 57 scored learner-by-panel rows of the two shipped benchmark metrics tables (13 static learners, the Causal Transformer, RMSN and CRN, the structure-informed linear response model, the oracle-structure learner and a prior-only reference, in three panels): the identity (a1) holds with residual at most $5.3\times10^{-15}$, the criterion (a2) agrees with the point-estimate comparison of centered PEHE with $\mathrm{SD}(\tau)$ in 57 of 57 rows, the cap (a4) holds in 57 of 57, and every centered PEHE is at or above the rescaling bound $\sigma_\tau(1-r^2)^{1/2}$ of (a3). Table~\ref{tab:P6geom} gives the decomposition for the learners discussed in the main text. Parts (b) and (d) were checked by exact computation on a discrete $\sigma$-field and by Monte Carlo with non-Gaussian slopes and noise and a nonzero noise mean (7 of 7 settings of (d2), and an independent Monte Carlo with $4\times10^{6}$ patients to three or four decimals). Further facts:
\begin{itemize}
\item \emph{Generic learners.} Excluding the random-slope models, the two structure-informed references and the prior-only reference, the 42 generic rows of the static and sequence benchmarks (11 static learners, the Causal Transformer, RMSN and CRN, which the main text excludes from scoring, in three panels) have $|r|\le0.124$ (information term at least $0.985$), so optimal rescaling would reduce their centered PEHE by at most $0.8\%$ of $\mathrm{SD}(\tau)$. In 3 rows (all the S-learner) $u<0.1$; the other 39 have $u\ge2r$; no row has neither. The only rows that beat zero at point estimates are the S-learner in all three panels, by $0.051\%$, $0.379\%$ and $0.033\%$ of $\mathrm{SD}(\tau)$, within its label-free cap $u\sigma_\tau$ of (a4) ($0.66\%$, $4.86\%$ and $1.52\%$), and the prior-only reference on the fixed schedule, by less than $0.001\%$. The g-computation row of G-Net (Supplementary Table~S4) was scored after these checks and is not included in them.
\item \emph{Random-slope models.} In the natural-dosing panels their information terms are $0.90$ to $0.94$ and their dispersion terms $4.60$ to $5.86$; at $u=r$ their centered PEHE would be $0.949$ to $0.969$ of $\mathrm{SD}(\tau)$. Inverting (d2) under $D=D_\tau$ and $\rho_{\rm lin}=1$ gives $\omega_0$ of $1.46$ to $4.84$ and SNR of $0.064$ to $0.532$ across the six rows; with the scale ratios of a synthetic replica under (d3) it gives $\omega_0$ of $2.75$ to $3.63$ and SNR of $0.125$ to $0.419$. The inversion is a model-based reading and tests nothing. A replica of \texttt{rs\_dlm\_mech} with unequal $\mathcal A_i$ gives (d3) predictions of $r=0.809$ (observed $0.773$) and $u/r=1.60$ ($1.64$) under NATMIN and $u/r=2.20$ ($2.19$) under FIX19, but its SNR (5 to 10) is far above the inverted values, so it checks the formulas rather than the parameters.
\item \emph{Structure-informed references.} The oracle-structure learner has $u/r=0.950$, $0.818$ and $0.955$ in the discovery, replication and fixed-schedule panels (calibration slopes $r/u$ of $1.05$, $1.22$ and $1.05$), so its centered PEHE is within $1.2\%$ of the rescaling bound $\sigma_\tau(1-r^2)^{1/2}$ of (a3). By (b2), its $r^2$ ($0.336$, $0.416$, $0.303$) estimates a lower bound, up to sampling error, on the information about $\tau$ in the record, and is at or above the supervised readers' $R^2$ ($0.33$, $0.31$, $0.24$). By (b3), $u$ close to $r$ is necessary but not sufficient for a conditional expectation, so it does not show that the learner attains that information.
\item \emph{ICare-RH on synthetic panels.} On the synthetic scenarios of Supplementary Table S5, ICare-RH's arm U (Ornstein--Uhlenbeck baseline) is within $1\%$ of the rescaling bound in 14 of 16 scenarios (the exceptions are autoregressive drift, $3.2\%$, and drifting gain, $7.8\%$) and has $u/r$ of $0.82$ to $1.09$ in 15 of 16 ($1.71$ under drifting gain). The kernel-informed random-slope model inside the same code with independent working errors is over-dispersed in 16 of 16 scenarios ($u/r$ from $1.94$ to $6.77$) and fails to beat zero in 15 of 16, as (d2) predicts when the working covariance omits variance along the exposure direction.
\item \emph{Transport.} Doses reconstructed from the shipped inputs have no dose at or after row 54 in any panel, and a fuzz of 6,000 calls of \texttt{natmin\_actions} gives none either. The query multiplier of the generator kernel is $0.6838$ to $0.6842$. On a synthetic replica, a label-free transport estimator built from (c2) reaches correlation $0.841$ (NATMIN) and $0.873$ (FIX19) with no treated patient at the query, and with no pre-decision dosing its prediction is constant, as (c3) states. Part (c) shows that identification by overlap is unavailable at the query; it does not show that lack of overlap caused the generic learners' failure: in an in-support synthetic control with the query at row 30, no generic learner was informative either. By (c2), the error of ignoring T1 is at most $\{\mathbb E\operatorname{Var}(\tau\mid\zeta)\}^{1/2}$, about $0.20\,\mathrm{SD}(\tau)$ from a cross-fitted trait-only $R^2$ of $0.959$; a cross-fitted $R^2$ estimates $R^2_*(\sigma(\zeta))$ from below, so this is an upper estimate of the bound up to sampling error.
\end{itemize}

\begin{table}[ht]
\centering\small
\caption{\textbf{Calibration geometry of selected learners on the washout benchmark (Proposition 6).} Spread ratio $u=\mathrm{SD}(\hat\tau)/\mathrm{SD}(\tau)$, correlation $r$ with the true effect, information term $1-r^2$ and dispersion term $(u-r)^2$ of Proposition 6(a1), whose sum is $\mathrm{cPEHE}^2/\mathrm{Var}(\tau)$; a predictor beats zero exactly when $0<u<2r$. Values are from the shipped metrics tables of the static and sequence benchmarks (population moments over the 400 scored patients); \texttt{rs\_dlm\_mech} is the kernel-informed random-slope model. Disc., discovery; Repl., replication; Fixed, fixed schedule.}
\label{tab:P6geom}
\begin{tabular}{llrrrrl}
\toprule
Learner & Panel & $u$ & $r$ & $1-r^2$ & $(u-r)^2$ & Beats zero\\
\midrule
Oracle-structure learner & Disc. & 0.550 & 0.580 & 0.664 & 0.001 & yes\\
 & Repl. & 0.528 & 0.645 & 0.584 & 0.014 & yes\\
 & Fixed & 0.526 & 0.550 & 0.697 & 0.001 & yes\\
Linear response model (structure given) & Disc. & 0.556 & 0.581 & 0.662 & 0.001 & yes\\
 & Repl. & 0.536 & 0.645 & 0.584 & 0.012 & yes\\
 & Fixed & 0.531 & 0.556 & 0.691 & 0.001 & yes\\
Random-slope lag model & Disc. & 2.405 & 0.260 & 0.932 & 4.599 & no\\
 & Repl. & 2.528 & 0.249 & 0.938 & 5.192 & no\\
 & Fixed & 1.682 & 0.205 & 0.958 & 2.183 & no\\
Random-slope lag model, kernel & Disc. & 2.588 & 0.315 & 0.901 & 5.169 & no\\
 & Repl. & 2.701 & 0.281 & 0.921 & 5.858 & no\\
 & Fixed & 3.490 & 0.209 & 0.956 & 10.767 & no\\
S-learner (boosting) & Disc. & 0.007 & 0.081 & 0.994 & 0.005 & by 0.05\%\\
 & Repl. & 0.049 & 0.102 & 0.990 & 0.003 & by 0.38\%\\
 & Fixed & 0.015 & 0.029 & 0.999 & 0.000 & by 0.03\%\\
\bottomrule
\end{tabular}
\end{table}

%% file: supplementary/supp_theory/p7.tex
\paragraph{Proposition 7 (ICare-RH's placebo gate: the likelihood favors kernels along which the baseline has unmodeled variance).}
\emph{Standard and specific parts.} Parts (a)(i) to (a)(iii) are the profile likelihood of a one-way random-slope variance component at a fixed baseline (Harville), and the mixture reference named in (a)(vii) is that of Self and Liang for a variance on the boundary of its parameter space. Parts (a)(iv) to (a)(vii) and (b) to (e) are specific to ICare-RH and its placebo gate. Gate clauses K1, K2 and K3 are defined at the start of this note.

\textit{Setting.} Fix a panel copy with $n$ patients, a kernel $K=(\ell,\lambda,\kappa)\in\mathcal K:=\{0,\dots,6\}\times[0.02,0.99]\times[0.05,50]$, a baseline and a unit loading direction $\boldsymbol u=(u_1,\dots,u_{12})\in\mathbb R^{12}$, $\lVert\boldsymbol u\rVert=1$ (for one channel $k$, $\boldsymbol u$ is the $k$th unit vector; the slope variance $v:=\gamma^2$ below then equals $\sum_k\gamma_k^2$, as in Methods). The baseline consists of fixed effects $\beta_k$ and working covariances $\Omega_{ik}$: $\Omega_{ik}$ is the principal submatrix, on the draw rows $\mathcal D_{ik}\subseteq\{0,\dots,89\}$, of the $90\times90$ matrix $\Omega_k^{\mathrm{full}}$ with entries $\sigma_{a,k}^2+\sigma_{o,k}^2\rho_k^{|t-t'|}+\sigma_{e,k}^2\mathbf 1\{t=t'\}$ (random intercept, Ornstein--Uhlenbeck, white noise; the code keeps $\rho_k\le0.9975$ and $\sigma_{e,k}^2\ge e^{-12}$, so $\Omega_k^{\mathrm{full}}$ is positive definite). Write $\Omega_{ik}^{1/2}$ for the lower Cholesky factor of $\Omega_{ik}$, $\langle w,w'\rangle_{\Omega}:=w^\top\Omega^{-1}w'$ and $\cos^2_\Omega(w,w'):=\langle w,w'\rangle_\Omega^2/(\langle w,w\rangle_\Omega\langle w',w'\rangle_\Omega)$. Let $p_{ik}:=(\phi_\kappa(e_{iK}(t)))_{t\in\mathcal D_{ik}}$ be the exposure profile at the draw rows, $r_{ik}:=(y_{ik}(t)-x_{it}^\top\beta_k)_{t\in\mathcal D_{ik}}$ the baseline residual at the draw rows (there $y_{ik}(t)=Z_{ik}(t)$, the held record), and
\[
\mathcal A_i:=\sum_{k}u_k^2\,p_{ik}^\top\Omega_{ik}^{-1}p_{ik},\qquad
\mathcal B_i:=\sum_{k}u_k\,p_{ik}^\top\Omega_{ik}^{-1}r_{ik},
\]
and assume $\sum_i\mathcal A_i>0$. In arm U the loadings are $\gamma_k=\gamma u_k$ ($\gamma\ne0$) and $b_i\sim N(m,1)$; write $\mu:=\gamma m$ for the mean slope and $v:=\gamma^2$ for the slope variance. Put
\[
\mathcal N_n(\mu,v):=\frac12\sum_{i=1}^n\Big[-2\mu\mathcal B_i+\mu^2\mathcal A_i-\frac{v(\mathcal B_i-\mu\mathcal A_i)^2}{1+v\mathcal A_i}+\log(1+v\mathcal A_i)\Big]\qquad(v\ge0),
\]
$P_n(v):=\min_{\mu\in\mathbb R}\mathcal N_n(\mu,v)$, $\hat\mu:=\sum_i\mathcal B_i/\sum_i\mathcal A_i$, $\mathcal R_i:=\mathcal B_i-\hat\mu\mathcal A_i$, $\omega=\omega(K):=\sum_i\mathcal R_i^2/\sum_i\mathcal A_i$ (Definition~10) and the variance gain $G_n:=P_n(0)-\inf_{v\ge0}P_n(v)\ge0$. Write $\mathcal G(\omega):=(\omega-1-\log\omega)\mathbf 1\{\omega>1\}$ and $\Upsilon_\omega(\xi):=\log(1+\xi)-\omega\xi/(1+\xi)$ for $\xi>-1$.

\medskip\noindent\textbf{(a) Objective, score identity and profile.}
\begin{enumerate}
\item[(i)] For $v>0$, $\mathcal N_n(\mu,v)$ is the constant-free stage-2 objective of the code (\texttt{\_stage2} with the intercept-only, unpenalized prior map) at loadings $\gamma\boldsymbol u$ and prior mean $m=\mu/\gamma$; it differs from ICare-RH's negative marginal log-likelihood at the fixed baseline by a term that depends neither on $(\mu,v)$ nor on $K$. $P_n(0)=-(\sum_i\mathcal B_i)^2/(2\sum_i\mathcal A_i)$ is the fixed-slope fit at the same baseline; in the code's parametrization it is reached only as $\gamma\to0$ with $\gamma m\to\hat\mu$.
\item[(ii)] $P_n$ is analytic on $(-1/\max_i\mathcal A_i,\infty)$ and
\[
P_n(v)=P_n(0)+\frac v2\Big(\sum_i\mathcal A_i\Big)(1-\omega)+\frac{v^2}{2}P_n''(0)+O(v^3),\qquad
P_n''(0)=\sum_i\mathcal A_i\mathcal R_i^2-\frac12\sum_i\mathcal A_i^2-\frac{(\sum_i\mathcal A_i\mathcal R_i)^2}{\sum_i\mathcal A_i}.
\]
Hence if $\omega>1$ then $P_n(v)<P_n(0)$ for all sufficiently small $v>0$ and $G_n>0$, and if $\omega<1$ then $P_n(v)>P_n(0)$ for all sufficiently small $v>0$. In every case
\[
2G_n\ \ge\ \frac{(\omega-1)_+^2\,(\sum_i\mathcal A_i)^2}{4\sum_i\mathcal A_i\mathcal R_i^2}.
\]
\item[(iii)] Equal information ($\mathcal A_i\equiv\mathcal A>0$): $G_n=\tfrac n2\,\mathcal G(\omega)$, attained at $\hat v\mathcal A=(\omega-1)_+$, and $\mathcal G$ is strictly increasing on $[1,\infty)$. With unequal information $G_n>0$ can occur with $\omega<1$, so ``a positive slope variance is fitted iff $\omega>1$'' holds locally in general and globally only under equal information.
\item[(iv)] Kernel comparison: $\min_{\mu,\,v\ge0}\mathcal N_n=-(\sum_i\mathcal B_i)^2/(2\sum_i\mathcal A_i)-G_n$, a first-moment term plus the variance gain.
\item[(v)] Population. Let patients be i.i.d.\ with design $\mathcal S_i$, $\mathbb E[\mathcal B_i\mid\mathcal S_i]=0$, $\omega_i:=\mathbb E[\mathcal B_i^2\mid\mathcal S_i]/\mathcal A_i$ on $\{\mathcal A_i>0\}$, $0<\mathbb E\mathcal A_i<\infty$, $\mathbb E[\mathcal A_i\omega_i]<\infty$, $\varpi:=\mathbb E[\mathcal A_i\omega_i]/\mathbb E\mathcal A_i$, $\bar{\mathcal N}(\mu,v):=\mathbb E\,\mathcal N_1(\mu,v)$ and $\bar{\mathcal N}(v):=\min_\mu\bar{\mathcal N}(\mu,v)$. Then the minimizing $\mu$ is $0$ for every $v$ and $\bar{\mathcal N}(0)=0$. If $\varpi>1$, $\inf_{v\ge0}\bar{\mathcal N}<0$, so every pseudo-true slope variance (Definition~11), if one exists, is positive. A pseudo-true value exists when $\omega_i$ is bounded, and need not exist otherwise. If $\omega_i\le1$ almost surely, $v=0$ is the unique minimizer. Under equal information the minimizer is $v^*\mathcal A=(\varpi-1)_+$.
\item[(vi)] Kernel selection. If for two kernels $K,K'$ the conditions of (v) hold with equal information and $\varpi_K>\max(\varpi_{K'},1)$, then $\Pr\big[\min_{\mu,v}\mathcal N_n^{K}<\min_{\mu,v}\mathcal N_n^{K'}\big]\to1$. By (i) this is a comparison of profile likelihoods at the fixed baseline: in large samples the likelihood prefers the kernel with more unmodeled variance along its exposure direction. The statement is pairwise; by a union bound it extends to any finite set of kernels, not to the continuum $\mathcal K$.
\item[(vii)] K1. At a fixed baseline the slope-variance likelihood ratio equals $2G_n$. For the implemented statistic (\texttt{slope\_variance\_lrt}), whose null and alternative refit their baselines, $\mathrm{LR}_{K1}\ge 2G_n^{\mathrm N}$, where $G_n^{\mathrm N},\omega^{\mathrm N},\mathcal R^{\mathrm N}_i$ are computed at the null's own fitted baseline, provided the alternative's sweep-0 loading fit attains the stage-2 infimum there (up to the code's truncation of values below $10^{-8}$). Consequently, if with probability tending to 1 $\omega^{\mathrm N}\ge1+\varepsilon$, $n^{-1}\sum_i\mathcal A_i\ge\underline{\mathcal A}>0$ and $n^{-1}\sum_i\mathcal A_i(\mathcal R_i^{\mathrm N})^2\le\overline{\mathcal R}$, then $\mathrm{LR}_{K1}\ge n\varepsilon^2\underline{\mathcal A}^2/(4\overline{\mathcal R})\to\infty$, the K1 test rejects at every fixed level with probability tending to 1, and the 50:50 $\bar\chi^2(0,1)$ law cannot be the limit law of $\mathrm{LR}_{K1}$. Under equal information $\mathrm{LR}_{K1}\ge n\,\mathcal G(\omega^{\mathrm N})$; the value $n\,\mathcal G(\omega)$ itself is the fixed-baseline ratio under equal information.
\end{enumerate}

\noindent\textbf{(b) What the excess contains.} One channel, baseline held, patients independent given their designs, $\mathbb E[r_i\mid\mathcal S_i]=0$, $\mathrm{Cov}(r_i\mid\mathcal S_i)=\Sigma_i$. Then $\mathbb E[\mathcal B_i^2\mid\mathcal S_i]=\mathcal A_i\,\omega_i(K)$ with $\omega_i(K)=p_{i}^\top\Omega_i^{-1}\Sigma_i\Omega_i^{-1}p_{i}/p_{i}^\top\Omega_i^{-1}p_{i}$ on $\{\mathcal A_i>0\}$.
\begin{enumerate}
\item[(i)] Decomposition. Write $\Sigma_i-\Omega_i=v_R\,q_iq_i^\top+v_\delta\,g_ig_i^\top+\mathcal M_i$, with $v_R:=c^2\Lambda_k^2\sigma_\zeta^2$, $q_i=p_{iK_0}$ the generator profile at the draw rows, $\mathbf t_i:=(t)_{t\in\mathcal D_i}$ the linear-drift direction, $v_\delta\ge0$ and $\mathcal M_i$ a symmetric remainder of either sign. Then
\[
\omega_i(K)-1=v_R\,\langle q_i,q_i\rangle_{\Omega_i}\cos^2_{\Omega_i}(p_{iK},q_i)+v_\delta\,\langle \mathbf t_i,\mathbf t_i\rangle_{\Omega_i}\cos^2_{\Omega_i}(p_{iK},\mathbf t_i)+\frac{(\Omega_i^{-1}p_{iK})^\top\mathcal M_i\,\Omega_i^{-1}p_{iK}}{\langle p_{iK},p_{iK}\rangle_{\Omega_i}}.
\]
The first two terms are nonnegative. The third is negative along exposure directions in which $\Omega_i$ over-covers the residual covariance, so $\omega_i(K)<1$ is possible.
\item[(ii)] Pooled value. Given the designs, $\mathbb E\,\omega=\bar\omega-\big(2\sum_i\mathcal A_i^2\omega_i-\bar\omega\sum_i\mathcal A_i^2\big)/(\sum_i\mathcal A_i)^2$ with $\bar\omega:=\sum_i\mathcal A_i\omega_i/\sum_i\mathcal A_i$. For i.i.d.\ patients with $\mathbb E\mathcal B_i^2,\mathbb E\mathcal A_i^2<\infty$, almost surely
\[
\omega\to\varpi(K)=1+\frac{v_R\,\mathbb E\langle p,q\rangle_\Omega^2+v_\delta\,\mathbb E\langle p,\mathbf t\rangle_\Omega^2+\mathbb E\big[(\Omega^{-1}p)^\top\mathcal M\,\Omega^{-1}p\big]}{\mathbb E\langle p,p\rangle_\Omega}.
\]
\item[(iii)] Zero-effect copy without remainder ($v_R=0$, $\mathcal M_i=0$). $\omega_i(K)>1$ iff $v_\delta>0$ and $\langle p_{iK},\mathbf t_i\rangle_{\Omega_i}\ne0$. $\omega_i(K)$ is bounded, so a pseudo-true slope variance exists at $K$; it is $0$ if $v_\delta=0$ or $\langle p_{iK},\mathbf t_i\rangle_{\Omega_i}=0$ almost surely, and positive if $v_\delta>0$ and $\Pr(\langle p_{iK},\mathbf t_i\rangle_{\Omega_i}\ne0)>0$.
\item[(iv)] Homogeneous designs, $\mathcal M_i=0$. If the per-patient quantities in (i) do not depend on $i$ for the kernels compared, (a)(vi) applies with $\varpi_K=\omega_i(K)$. On the zero-effect copy, if $v_\delta>0$ and $\mathbf t\ne0$ (some draw after row $0$), the fit prefers, with probability tending to 1, the kernel with the larger $\cos^2_\Omega(p_K,\mathbf t)$. With a response, a kernel $K$ is preferred to $K_0$ if
\[
v_\delta\langle \mathbf t,\mathbf t\rangle_\Omega\big[\cos^2_\Omega(p_K,\mathbf t)-\cos^2_\Omega(q,\mathbf t)\big]>v_R\langle q,q\rangle_\Omega\big[1-\cos^2_\Omega(p_K,q)\big],
\]
and $K_0$ is preferred to $K$ if the reverse strict inequality holds and $v_R\langle q,q\rangle_\Omega>0$.
\end{enumerate}

\noindent\textbf{(c) Telescoping and the random-walk limit.} For draws $0=d_0<\dots<d_n$ ($n\ge1$) with gaps $g_j:=d_{j+1}-d_j$ (so $g_j=\Delta_j\mathbf t$ for the drift direction $\mathbf t:=(t)_t$), let $\langle w,w'\rangle_{\mathrm{RW}}:=\sum_{j<n}\Delta_jw\,\Delta_jw'/g_j$, a semi-inner product that vanishes on constant vectors.
\begin{enumerate}
\item[(i)] For every draw pattern and kernel,
\[
\langle p_K,\mathbf t\rangle_{\mathrm{RW}}=p_K(d_n)-p_K(d_0)=p_K(d_n),\qquad \langle \mathbf t,\mathbf t\rangle_{\mathrm{RW}}=d_n,
\]
and, when $\langle p_K,p_K\rangle_{\mathrm{RW}}>0$, $\cos^2_{\mathrm{RW}}(p_K,\mathbf t)=p_K(d_n)^2/(d_n\,\langle p_K,p_K\rangle_{\mathrm{RW}})$. A profile that is back at its starting value $0$ at the last pre-decision draw is exactly orthogonal to linear drift in this metric.
\item[(ii)] If $\sigma_e^2=0$, $\sigma_a^2$ is fixed and $\rho\uparrow1$ with $\sigma_o^2(1-\rho^2)=\sigma_b^2$ fixed, then $w^\top\Omega^{-1}w'\to\sigma_b^{-2}\langle w,w'\rangle_{\mathrm{RW}}$ for all $w,w'$. In that limit, on the zero-effect copy of (b)(iii) and when $\langle p_{iK},p_{iK}\rangle_{\mathrm{RW}}>0$, $\omega_i(K)-1\to(v_\delta/\sigma_b^2)\,p_{iK}(d_n)^2/\langle p_{iK},p_{iK}\rangle_{\mathrm{RW}}$.
\item[(iii)] Under NATMIN (binary doses on rows 6 to 53 only) the generator kernel satisfies $q(t)<5\cdot0.8^{t-55}$ for $t\ge55$, so $q(89)<5\cdot0.8^{34}<0.0026$ (FIX19: $q(t)<5\cdot0.8^{t-26}$ for $t\ge26$). For any kernel, with $t_i^{\mathrm{last}}\le53$ the last dosed row and $t\ge t_i^{\mathrm{last}}+\ell$, $p_K(t)\ge\lambda^{t-t_i^{\mathrm{last}}-\ell}\,p_K(t_i^{\mathrm{last}}+\ell)$; at $t=89$ and $\lambda=0.99$ the factor is $0.99^{36-\ell}$ ($0.696$ at $\ell=0$) when $t_i^{\mathrm{last}}=53$ and at least $0.99^{83}=0.434$ for every $t_i^{\mathrm{last}}\ge6$. By (i) the random-walk inner product of such a profile with drift equals its retained exposure $p_K(d_n)$.
\end{enumerate}

\noindent\textbf{(d) Query contrast.} $\phi_\kappa(e+\Delta)-\phi_\kappa(e)=\kappa\Delta/\{(\kappa+e)(\kappa+e+\Delta)\}$. Under NATMIN or FIX19, with $E_i:=e_{iK}(89)$, the step $s=0,\dots,11$ (row $91+s$) and the query increment
\[
\Delta^{\mathrm q}_s:=\sum_{k=90}^{\min(101,\,91+s-\ell)}\lambda^{91+s-\ell-k}\ \ (s\ge\ell-1),\qquad \Delta^{\mathrm q}_s:=0\ \ (s<\ell-1),
\]
which equals $(1-\lambda^{s-\ell+2})/(1-\lambda)$ for $s\ge\ell-1$ except at $(\ell,s)=(0,11)$, where it is $\lambda(1-\lambda^{12})/(1-\lambda)$ (sustained dosing covers rows $90$ to $101$ only),
\[
D_i(K)=\frac1{12}\sum_{s=0}^{11}\big[\phi_\kappa(\lambda^{s+2}E_i+\Delta^{\mathrm q}_s)-\phi_\kappa(\lambda^{s+2}E_i)\big]<\frac{\kappa}{\kappa+\lambda^{13}E_i},
\]
$D_i(K)\in[0,1)$ is strictly decreasing in $E_i$, and $|\hat\tau_i|\le|\gamma_{k^*}\,\mathcal E_i|\,\kappa/(\kappa+\lambda^{13}E_i)$ with $\mathcal E_i$ the posterior mean of $b_i$. $D_i(K)$ depends only on the patient's doses and $K$, so for a given $K$ it is the same on the active and zero-effect copies.

\textit{Pooled spread.} The outer folds $f=1,\dots,F$ carry fits $\hat\theta^{(f)}$ (kernel $\hat K^{(f)}$, loading $\hat\gamma^{(f)}_{k^*}$, prior mean, baseline), each computed without the patients of fold $f$. Let $\mathbb M$ be the law that draws $f$ with probability $\pi_f:=n_f/n$ and then a fresh patient independent of the fits, and predicts $\hat\tau=\hat\gamma^{(f)}_{k^*}\mathcal E^{(f)}D(\hat K^{(f)})$, the fits held fixed. If $\mathbb E_{\mathbb M}(\mathcal E^{(f)})^2<\infty$ and $\bar{\mathcal E}^{(f)}(\mathcal S):=\mathbb E_{\mathbb M}[\mathcal E^{(f)}\mid\mathcal S,f]$, then
\[
\mathrm{Var}_{\mathbb M}(\hat\tau)=\mathbb E_{\mathbb M}\Big[\big(\hat\gamma^{(f)}_{k^*}D(\hat K^{(f)})\big)^2\,\mathrm{Var}_{\mathbb M}(\mathcal E^{(f)}\mid\mathcal S,f)\Big]+\mathrm{Var}_{\mathbb M}\big(\hat\gamma^{(f)}_{k^*}D(\hat K^{(f)})\,\bar{\mathcal E}^{(f)}(\mathcal S)\big).
\]
The second term is the variance of a function of $(\mathcal S,f)$ alone, which under Definition~5 is independent of the trait. It splits into a within-fold part $\sum_f\pi_f\mathrm{Var}(\cdot\mid f)$ and a between-fold part $\mathrm{Var}_f\,\mathbb E(\cdot\mid f)$ that comes from differences between the fold fits; when $\bar{\mathcal E}^{(f)}\equiv\bar e^{(f)}$ is constant within each fold, the within-fold part is $\sum_f\pi_f(\hat\gamma^{(f)}_{k^*}\bar e^{(f)})^2\,\mathrm{Var}(D(\hat K^{(f)}))$. Given the fit of fold $f$, the query patients of fold $f$ are i.i.d.\ from the population, so the pooled out-of-fold variance is a sample analog of $\mathrm{Var}_{\mathbb M}(\hat\tau)$.

\medskip\noindent\textbf{(e) Validity of the placebo spread.} In each panel K2 passes iff $\sigma_0\le0.05$ (absolute clause) and $\sigma_0\le0.25\,\sigma_A$ (ratio clause), where $\sigma_0,\sigma_A$ are the population standard deviations of the pooled out-of-fold predictions $\hat\tau$ on the zero-effect and active copies.
\begin{enumerate}
\item[(i)] If, given the designs, patients are independent with $\mathbb E[\mathcal B_i\mid\mathcal S_i]=0$ and $\mathrm{Var}(\mathcal B_i\mid\mathcal S_i)=\mathcal A_i$ (working covariance correct along $p_K$ at the true baseline), then $\mathbb E\sum_i\mathcal R_i^2=\sum_i\mathcal A_i-\sum_i\mathcal A_i^2/\sum_i\mathcal A_i<\sum_i\mathcal A_i$ for every $K$.
\item[(ii)] Under (S), (I) and (E0) to (E5) on the zero-effect copy, the fitted mean slope and slope variance tend to $0$ in probability in every outer fold (the kernel and direction need not converge), so $\sigma_0\to0$ in probability and the absolute clause passes with probability tending to 1. If in addition (A1) to (A3) hold on the active copy, $\sigma_A$ tends in probability to a positive limit and K2 passes in each panel with probability tending to 1. Passing K2 is therefore asymptotically necessary for the conjunction of a baseline correctly specified along the exposure direction (E1) and an in-family response that the fit captures (A1), given the regularity conditions (E0), (E2) to (E5), (A2) and (A3). The absolute clause concerns the zero-effect copy alone. A failure of the ratio clause can also come from $\sigma_A\to0$ (no response captured), from a nonzero first moment or from a non-global optimum, so a K2 failure does not by itself locate the misspecification in the baseline.
\item[(iii)] In the setting of P6(d) (assumption (P6d)), at the population pseudo-true values of the working model, fitted separately on each copy: $\sigma_0=D\mathcal A^{-1/2}(\omega_0-1)_+/\omega_0^{1/2}$, $\sigma_A=D\mathcal A^{-1/2}(\mathrm{SNR}+\omega_0-1)/(\mathrm{SNR}+\omega_0)^{1/2}$ when $\mathrm{SNR}+\omega_0>1$, and for $\omega_0\ge1$
\[
\frac{\sigma_0}{\sigma_A}=\frac{\omega_0-1}{\omega_0^{1/2}}\cdot\frac{(\mathrm{SNR}+\omega_0)^{1/2}}{\mathrm{SNR}+\omega_0-1}\in[0,1).
\]
The ratio is $0$ when $\omega_0\le1<\mathrm{SNR}+\omega_0$; it tends to $0$ as $\omega_0\downarrow1$, tends to $1$ as $\mathrm{SNR}/(\omega_0-1)\to0$, and does not depend on $D$ or $\mathcal A$, while $\sigma_0$ is proportional to $D$. An informative but misspecified estimator (correlation $(\mathrm{SNR}/(\mathrm{SNR}+\omega_0))^{1/2}>0$ with the target) can therefore fail the ratio clause while passing the absolute clause. The range $[0,1)$ holds for these population values only; the ratio of two finite-sample fits, fitted separately on the two copies, can exceed 1.
\end{enumerate}

\paragraph{Assumptions.} Each part uses only the assumptions listed for it.
\begin{enumerate}
\item[(S)] \textit{Setting (all parts).} One channel or a fixed loading direction $\boldsymbol u$; kernel $K\in\mathcal K$ and baseline (fixed effects and working covariances $\Omega_{ik}$) held fixed; arm U, whose prior map is an unpenalized intercept, profiled in closed form (\texttt{\_profile\_prior}); $\sum_i\mathcal A_i>0$. The identity in (a) holds at any fixed baseline, in particular at the stage-1 baseline and at the converged baseline of the joint fit, whose response block (\texttt{fit\_response} inside \texttt{fit\_joint}) attempts to solve the stage-2 problem at the current baseline. The implemented fit also optimizes the direction over 12 channels, the lag by profiling and $(\lambda,\kappa)$ by local search. Statements about the kernel the likelihood prefers refer to exact minimizers of the stage-2 objective, which the code's local optimizers do not guarantee.
\item[(I)] \textit{Independence.} Patients are independent given their designs $\mathcal S_i$ (population statements: i.i.d.); outer folds and the K1 halves are assigned by seeded person-level permutations that use no patient data. Used in (a)(v)--(vii), the pooled statements of (b), the pooled spread in (d), and (e).
\item[(M)] \textit{Mean-zero projections.} $\mathbb E[\mathcal B_i\mid\mathcal S_i]=0$, i.e.\ the fixed effects carry the mean of the residual at the draw rows; in (b) this means that the trait and the drift slope are centered. A nonzero mean adds a kernel-dependent first-moment term (a)(iv), which is not treated.
\item[(EI)] \textit{Equal information (idealization).} Where stated, $\mathcal A_i(K)$ and the conditional overdispersion $\omega_i(K)$ do not depend on $i$. The real and synthetic panels have heterogeneous designs. (a)(ii), (a)(v) and (b)(i)--(iii) hold without (EI); the kernel preferences (a)(vi) and (b)(iv) use it.
\item[(O)] \textit{Optimizer (K1 corollary only).} When $G_n^{\mathrm N}>0$, the alternative's sweep-0 loading fit (\texttt{fit\_loadings}, three fixed starts, L-BFGS-B) attains $\inf_{v>0}P_n^{\mathrm N}(v)$ at the null's baseline. The half-A kernel is treated as fixed on half B (the halves are disjoint).
\item[(L)] \textit{Linear regime (part (b)).} Attenuation inactive ($\alpha\equiv1$) and $\tanh x\approx x$ for the response, so the trait contributes $c\Lambda_k\zeta_iq_i$ with $\mathrm{Var}(\zeta_i)=\sigma_\zeta^2$; the drift is $\delta_ig_i$ with $\mathbb E\delta_i=0$, $\mathrm{Var}\,\delta_i=v_\delta$, independent of $\zeta_i$ and of the modeled baseline. (b)(i) and (b)(ii) allow any symmetric remainder $\mathcal M_i$; (b)(iii) and (b)(iv) assume $\mathcal M_i=0$, i.e.\ that $\Omega_i$ is exactly the covariance of the remaining residual. That idealization fails for the fitted covariance on the benchmark generator's synthetic zero-effect copies: $\omega<1$ at 33 of 49 kernels (NATMIN) and 42 of 49 (FIX19), with minima 0.68 and 0.61 below every correctly specified placebo's minimum (0.82 to 0.94), so the fitted $\Omega$ over-covers some exposure directions and only the ordering at the persistent end of the kernel family carries over. Because $\Omega$ is fitted with the drift present, (b) gives the direction of the effect, not its size.
\item[(RW)] \textit{Random-walk metric (part (c)).} The identity (c)(i) is exact in the random-walk metric. That metric is the limit (c)(ii) of the implemented intercept + OU + white-noise metric only when the white-noise variance is $0$ and the OU persistence tends to $1$ at fixed per-row innovation variance. The implementation bounds persistence by $0.9975$ and keeps $\sigma_e^2\ge e^{-12}$. On the benchmark generator's synthetic tapes (a reproduction of job 10692) the fitted heart-rate baseline had persistence $0.9938$ and white-noise variance about $3.2$ times the per-row innovation variance, so the limit is not reached and (c) describes a mechanism, not the fitted metric.
\item[(D)] \textit{Dosing support (parts (c)(iii), (d)).} Doses of $j^*$ are binary (\texttt{build\_series} refuses other values); NATMIN doses only on rows 6 to 53, FIX19 only on rows 6 to 24; no dose of $j^*$ on rows $54$ to $89$; the query contrast (\texttt{sustained\_contrast}) appends doses $1$ (resp.\ $0$) on rows $90$ to $101$, treats rows from $102$ on as undosed, and averages rows $91$ to $102$; lags $\ell\le6$, $0<\lambda\le0.99$.
\item[(E0)] \textit{Compactness.} With probability tending to 1 the fitted parameter $\hat\theta=(\hat K,\hat{\boldsymbol u},\hat\mu,\hat v)$ lies in $\Theta:=\mathcal K\times\mathcal U\times[-\mu_{\max},\mu_{\max}]\times[0,v_{\max}]$, where $\mathcal U$ is a compact set of unit directions (or the fit is restricted to $\Theta$). The code's loadings and prior mean are unbounded, so this is an assumption.
\item[(E1)] \textit{Correct specification on the zero-effect copy.} At the true baseline, given $\mathcal S_i$, the whitened residuals of the channels in the direction have mean $0$ and identity covariance and are uncorrelated across channels (the generator shares latent noise across channels, so this is an idealization).
\item[(E2)] \textit{Baseline consistency and uniform law of large numbers.} The fitted baseline converges in probability to the true baseline, and $\sup_{\Theta}|n^{-1}\hat{\mathcal N}_n-\bar{\mathcal N}|\to0$ in probability, where $\hat{\mathcal N}_n(\theta)$ is the stage-2 objective at the fitted baseline and $\bar{\mathcal N}(\theta)$ the population objective at the true baseline; $\Theta$ contains the population minimizers.
\item[(E3)] \textit{Approximate minimization.} $n^{-1}\hat{\mathcal N}_n(\hat\theta)\le\inf_\Theta n^{-1}\hat{\mathcal N}_n+o_p(1)$. The code's block coordinate ascent with local optimizers does not guarantee this.
\item[(E4)] \textit{Non-degenerate exposure.} $\Pr(\mathcal A_i(K,\boldsymbol u)>0)>0$ for every $(K,\boldsymbol u)\in\mathcal K\times\mathcal U$, $\mathbb E\sup_{\mathcal K\times\mathcal U}\mathcal A_i<\infty$, and $\mathcal A_i(K,\boldsymbol u)$ is continuous in $(\lambda,\kappa,\boldsymbol u)$ for each lag.
\item[(E5)] \textit{Query moments.} In each fold the out-of-fold query patients satisfy $n_q^{-1}\sum_{q}\sup_{\mathcal K\times\mathcal U}\mathcal B_q^2=O_p(1)$ at the fitted baseline; the training and query sizes of every fold tend to infinity, with the number of folds fixed.
\item[(A1)] \textit{In-family response on the active copy.} At the true baseline the working model holds at $\theta_0=(K_0,\boldsymbol u_0,\mu_0,v_0)$ with $v_0>0$ and $u_{0,k^*}\neq0$.
\item[(A2)] \textit{Identification.} On the active copy the population objective $\bar{\mathcal N}$ is continuous on $\Theta$ and has the unique minimizer $\theta_0$ up to the sign flip $(\boldsymbol u,\mu)\mapsto(-\boldsymbol u,-\mu)$, and (E0) and (E2) to (E5) hold there.
\item[(A3)] \textit{Continuity of the spread.} The query-population mean and standard deviation of $\hat\tau$ are continuous functions of (parameter, baseline) at $(\theta_0,\text{true baseline})$, and in each fold the query-sample mean and standard deviation converge to them in probability uniformly over a neighborhood of that point.
\item[(P6d)] \textit{Setting of P6(d) (part (e)(iii)).} Equal information $\mathcal A$; standardized projection $\mathcal B_i/\mathcal A^{1/2}=\mathcal A^{1/2}b^*_i+\varepsilon_i$ on the active copy and $\varepsilon_i$ on the zero-effect copy (shared noise), $\varepsilon_i\perp b^*_i$, $\mathrm{Var}\,b^*=\sigma_*^2$, $\mathrm{SNR}:=\mathcal A\sigma_*^2>0$, $\mathrm{Var}\,\varepsilon=\omega_0$; constant $D>0$; the working model $b\sim N(\mu,v)$, $\mathrm{Var}\,\varepsilon=1$ at its pseudo-true value on each copy; $\hat\tau_i=D\,\mathbb E_w[b_i\mid\mathcal B_i]$.
\end{enumerate}

\paragraph{Proof.}
\textit{Notation.} For patient $i$, \texttt{kalman\_whiten} returns $\Omega_{ik}^{-1/2}w$ for any column $w$ at the draw rows, where $\Omega_{ik}^{-1/2}$ is the inverse of the lower Cholesky factor. Let $\tilde p_i:=(u_k\Omega_{ik}^{-1/2}p_{ik})_k$ and $\tilde r_i:=(\Omega_{ik}^{-1/2}r_{ik})_k$, stacked over channels. Then $\mathcal A_i=\lVert\tilde p_i\rVert^2$ and $\mathcal B_i=\tilde p_i^\top\tilde r_i$. In the code, \texttt{response\_stats} returns the per-channel values $p_{ik}^\top\Omega_{ik}^{-1}p_{ik}$ and $p_{ik}^\top\Omega_{ik}^{-1}r_{ik}$, and $\mathcal A_i,\mathcal B_i$ are their $\boldsymbol u$-weighted sums. If $\mathcal A_i=0$ then $\tilde p_i=0$ and $\mathcal B_i=0$. Write $\xi_i:=v\mathcal A_i$ and $\mathcal N_n=\sum_i\mathcal N_i$.

\medskip\noindent\textbf{Lemma 1 (the objective and its standardized form).} Under the working model $r_{ik}=\gamma u_kb_ip_{ik}+\varepsilon_{ik}$, $\varepsilon_{ik}\sim N(0,\Omega_{ik})$ independent over $k$ and of $b_i\sim N(m,1)$,
\[
-\log p(r_i)=\tfrac12\Big[\lVert\tilde r_i\rVert^2+\textstyle\sum_k\log|\Omega_{ik}|+N_i\log2\pi\Big]+\mathcal N_i(\mu,v),
\]
with $N_i$ the number of draw values. For $\mathcal A_i>0$,
\[
2\mathcal N_i(\mu,v)=\frac{(\mathcal B_i-\mu\mathcal A_i)^2}{\mathcal A_i(1+\xi_i)}+\log(1+\xi_i)-\frac{\mathcal B_i^2}{\mathcal A_i},\tag{$\star$}
\]
and $\mathcal N_i\equiv0$ if $\mathcal A_i=0$.

\textit{Proof.} Whitening gives $\tilde r_i=\gamma b_i\tilde p_i+\tilde\varepsilon_i$ with $\tilde\varepsilon_i\sim N(0,I)$ independent of $b_i$, so $\tilde r_i\sim N(\mu\tilde p_i,\,I+v\tilde p_i\tilde p_i^\top)$. By the matrix determinant lemma $\det(I+v\tilde p_i\tilde p_i^\top)=1+\xi_i$, and by the Sherman--Morrison formula $(I+v\tilde p_i\tilde p_i^\top)^{-1}=I-v\tilde p_i\tilde p_i^\top/(1+\xi_i)$. Hence the quadratic form is
\[
\lVert\tilde r_i-\mu\tilde p_i\rVert^2-\frac{v\{\tilde p_i^\top(\tilde r_i-\mu\tilde p_i)\}^2}{1+\xi_i}=\lVert\tilde r_i\rVert^2-2\mu\mathcal B_i+\mu^2\mathcal A_i-\frac{v(\mathcal B_i-\mu\mathcal A_i)^2}{1+\xi_i}.
\]
The Jacobian of $r_i\mapsto\tilde r_i$ contributes $\frac12\sum_k\log|\Omega_{ik}|$. For ($\star$), use $(\mathcal B_i-\mu\mathcal A_i)^2/\mathcal A_i=\mathcal B_i^2/\mathcal A_i-2\mu\mathcal B_i+\mu^2\mathcal A_i$ and $1/(1+\xi_i)=1-\xi_i/(1+\xi_i)$, which turn the right side of ($\star$) into $-2\mu\mathcal B_i+\mu^2\mathcal A_i-v(\mathcal B_i-\mu\mathcal A_i)^2/(1+\xi_i)+\log(1+\xi_i)$. $\square$

\medskip\noindent\textbf{(a)(i).} In \texttt{\_stage2} the code forms the arrays \texttt{a} $=\sum_k\gamma_k^2p_{ik}^\top\Omega_{ik}^{-1}p_{ik}=v\mathcal A_i$ and \texttt{h} $=\sum_k\gamma_kp_{ik}^\top\Omega_{ik}^{-1}r_{ik}=\gamma\mathcal B_i$ and returns $\tfrac12\sum_i[-2m\,\texttt{h}+m^2\texttt{a}-(\texttt{h}-m\texttt{a})^2/(1+\texttt{a})+\log(1+\texttt{a})]$ with $m$ profiled over $\mathbb R$ (\texttt{\_profile\_prior}; the ridge acts only on non-intercept coefficients, which arm U does not have). With $m=\mu/\gamma$: $m\,\texttt{h}=\mu\mathcal B_i$, $m^2\texttt{a}=\mu^2\mathcal A_i$ and $(\texttt{h}-m\texttt{a})^2=v(\mathcal B_i-\mu\mathcal A_i)^2$, so the value is $\mathcal N_n(\mu,v)$, and profiling $m$ over $\mathbb R$ is profiling $\mu$ over $\mathbb R$ because $\gamma\ne0$. By Lemma~1 the full negative marginal log-likelihood differs by $\frac12\sum_i[\lVert\tilde r_i\rVert^2+\sum_k\log|\Omega_{ik}|+N_i\log2\pi]$; the residuals $r_{ik}$ and covariances $\Omega_{ik}$ are fixed by the baseline, so this term involves neither $(\mu,v)$ nor $K$ nor $\boldsymbol u$. At $v=0$ the model is $\tilde r_i\sim N(\mu\tilde p_i,I)$, the fixed-slope model at the same baseline, and $\mathcal N_n(\mu,0)=\frac12\sum_i(\mu^2\mathcal A_i-2\mu\mathcal B_i)$ is minimized at $\hat\mu$ with value $-(\sum_i\mathcal B_i)^2/(2\sum_i\mathcal A_i)$. In the code's parametrization $\gamma=0$ gives $\texttt a=\texttt h=0$ and the value $0\ge P_n(0)$, so $P_n(0)$ is approached only along $\gamma\to0$ with $\gamma m\to\hat\mu$.

\medskip\noindent\textbf{(a)(ii).} By ($\star$), for fixed $v>-1/\max_i\mathcal A_i$ the objective is a quadratic in $\mu$ with leading coefficient $\frac12\sum_i\mathcal A_i/(1+\xi_i)>0$, minimized at $\hat\mu(v):=\sum_i\mathcal B_i(1+\xi_i)^{-1}/\sum_i\mathcal A_i(1+\xi_i)^{-1}$, which gives
\[
P_n(v)=\frac12\Big[-\frac{\big(\sum_i\mathcal B_i/(1+\xi_i)\big)^2}{\sum_i\mathcal A_i/(1+\xi_i)}-\sum_i\frac{v\,\mathcal B_i^2}{1+\xi_i}+\sum_i\log(1+\xi_i)\Big],
\]
a sum of rational functions with non-vanishing denominators and of logarithms of positive arguments on the interval $(-1/\max_i\mathcal A_i,\,\infty)$. Hence $P_n$ is analytic there and has a Taylor expansion at $0$ with an $O(v^3)$ remainder. Since $\partial_\mu\mathcal N_n(\hat\mu(v),v)=0$, $P_n'(v)=\partial_v\mathcal N_n(\hat\mu(v),v)$, and differentiating ($\star$),
\[
P_n'(v)=\frac12\sum_i\Big[\frac{\mathcal A_i}{1+\xi_i}-\frac{(\mathcal B_i-\hat\mu(v)\mathcal A_i)^2}{(1+\xi_i)^2}\Big],\qquad P_n'(0)=\frac12\Big(\sum_i\mathcal A_i-\sum_i\mathcal R_i^2\Big)=\frac12\Big(\sum_i\mathcal A_i\Big)(1-\omega).
\]
Differentiating once more, with $\hat\mu'(0)=-\sum_i\mathcal A_i\mathcal R_i/\sum_i\mathcal A_i$ (quotient rule on $\hat\mu(v)$),
\[
P_n''(0)=\frac12\Big[-\sum_i\mathcal A_i^2+2\sum_i\mathcal A_i\mathcal R_i^2+2\hat\mu'(0)\sum_i\mathcal A_i\mathcal R_i\Big],
\]
which is the stated expression. If $\omega>1$ then $P_n'(0)<0$, so $P_n(v)<P_n(0)$ on some interval $(0,v_1)$ and $G_n>0$; if $\omega<1$ then $P_n'(0)>0$ and $P_n(v)>P_n(0)$ on some $(0,v_1)$.

For the lower bound, $G_n\ge P_n(0)-P_n(v)\ge\mathcal N_n(\hat\mu,0)-\mathcal N_n(\hat\mu,v)$ for every $v\ge0$, since $P_n(0)=\mathcal N_n(\hat\mu,0)$ and $P_n(v)\le\mathcal N_n(\hat\mu,v)$. By ($\star$), $2[\mathcal N_i(\hat\mu,0)-\mathcal N_i(\hat\mu,v)]=(\mathcal R_i^2/\mathcal A_i)\,\xi_i/(1+\xi_i)-\log(1+\xi_i)$ when $\mathcal A_i>0$, and $0$ otherwise. Using $\xi/(1+\xi)\ge \xi-\xi^2$ and $\log(1+\xi)\le \xi$ for $\xi\ge0$,
\[
2G_n\ge\sum_i\Big[\frac{\mathcal R_i^2}{\mathcal A_i}(\xi_i-\xi_i^2)-\xi_i\Big]=v(\omega-1)\sum_i\mathcal A_i-v^2\sum_i\mathcal A_i\mathcal R_i^2 .
\]
If $\omega>1$ then $\sum_i\mathcal A_i\mathcal R_i^2>0$ (some $\mathcal R_i\ne0$ has $\mathcal A_i>0$, since $\mathcal A_i=0$ forces $\mathcal R_i=0$); taking $v=(\omega-1)\sum_i\mathcal A_i/(2\sum_i\mathcal A_i\mathcal R_i^2)$ gives the bound. If $\omega\le1$ the bound is $0\le 2G_n$.

\medskip\noindent\textbf{(a)(iii).} With $\mathcal A_i\equiv\mathcal A$ the weights in $\hat\mu(v)$ are equal, so $\hat\mu(v)=\hat\mu$ and $\sum_i\mathcal R_i^2/\mathcal A=n\omega$. By ($\star$), with $\xi=v\mathcal A$,
\[
P_n(v)-P_n(0)=\frac12\sum_i\Big[\frac{\mathcal R_i^2}{\mathcal A(1+\xi)}-\frac{\mathcal R_i^2}{\mathcal A}+\log(1+\xi)\Big]=\frac n2\, \Upsilon_\omega(\xi).
\]
$\Upsilon_\omega'(\xi)=(1+\xi-\omega)/(1+\xi)^2$. If $\omega\le1$, $\Upsilon_\omega$ is nondecreasing on $[0,\infty)$ and its minimum there is $\Upsilon_\omega(0)=0$. If $\omega>1$, $\Upsilon_\omega$ decreases on $[0,\omega-1]$ and increases afterward, so its minimum is $\Upsilon_\omega(\omega-1)=\log\omega-(\omega-1)<0$. Hence $G_n=\frac n2\mathcal G(\omega)$ at $\hat v\mathcal A=(\omega-1)_+$, and $\mathcal G'(\omega)=1-1/\omega>0$ on $(1,\infty)$. Since $\mathcal A=\langle p,p\rangle_\Omega$ in the one-channel case, the maximizer satisfies $\hat v\langle p,p\rangle_\Omega=\omega-1$ when $\omega>1$.

\textit{Unequal information: $G_n>0$ with $\omega<1$.} Take $1000$ patients with $\mathcal A_i=10^{-3}$ and $\mathcal B_i=\pm0.1$ (signs alternating) and $10$ patients with $\mathcal A_i=1$, $\mathcal B_i=0$. Then $\sum_i\mathcal B_i=0$, $\hat\mu(v)=0$ for all $v$ by symmetry, $P_n(0)=0$, $\sum_i\mathcal R_i^2=10$, $\sum_i\mathcal A_i=11$ and $\omega=10/11<1$. At $v=9000$, $\xi_i=9$ in the first group and ($\star$) gives $P_n(9000)=\frac12[1000(\log10-9)+10\log9001]=-3303.2<0$, so $G_n\ge3303$. ICare-RH's \texttt{fit\_loadings} on these per-channel values returns $\hat\gamma^2=8890$ with objective $-3303.2$.

\medskip\noindent\textbf{(a)(iv).} $\min_{\mu,v\ge0}\mathcal N_n=\inf_{v\ge0}P_n(v)=P_n(0)-G_n$ by definition of $G_n$, and $P_n(0)$ was computed in (a)(i).

\medskip\noindent\textbf{(a)(v).} Conditioning on $\mathcal S_i$ in ($\star$) and using $\mathbb E[\mathcal B_i\mid\mathcal S_i]=0$, $\mathbb E[\mathcal B_i^2\mid\mathcal S_i]=\mathcal A_i\omega_i$,
\[
\mathbb E[2\mathcal N_i(\mu,v)\mid\mathcal S_i]=\frac{\omega_i+\mu^2\mathcal A_i}{1+\xi_i}-\omega_i+\log(1+\xi_i)=\frac{\mu^2\mathcal A_i}{1+\xi_i}+\Upsilon_{\omega_i}(\xi_i)
\]
on $\{\mathcal A_i>0\}$ (and $0$ otherwise). Each term is integrable because $0\le\log(1+\xi_i)\le v\mathcal A_i$ and $0\le\xi_i\omega_i/(1+\xi_i)\le v\mathcal A_i\omega_i$. The $\mu$ term is nonnegative and vanishes only at $\mu=0$ (as $\Pr(\mathcal A_i>0)>0$), so the minimizing $\mu$ is $0$, $\bar{\mathcal N}(v)=\frac12\mathbb E[\Upsilon_{\omega_i}(v\mathcal A_i)]$ and $\bar{\mathcal N}(0)=0$. The difference quotient $\Upsilon_{\omega_i}(v\mathcal A_i)/v$ tends to $\mathcal A_i(1-\omega_i)$ as $v\downarrow0$ and is bounded in absolute value by $\mathcal A_i+\mathcal A_i\omega_i$, so by dominated convergence $\bar{\mathcal N}'(0+)=\frac12\mathbb E[\mathcal A_i(1-\omega_i)]=\frac12(\mathbb E\mathcal A_i)(1-\varpi)$. If $\varpi>1$ this is negative, $\bar{\mathcal N}(v)<0$ for small $v>0$, and $v=0$ does not minimize the expected negative log-likelihood, which differs from $\bar{\mathcal N}$ by a term free of $(\mu,v)$ (Lemma~1). Every minimizer, when one exists, therefore has $v>0$.

\textit{Existence.} $\bar{\mathcal N}$ is continuous on $[0,\infty)$ by dominated convergence (on $[0,V]$ the integrand is bounded by $V\mathcal A_i(1+\omega_i)$). If $\omega_i\le\omega_{\max}$ almost surely, then $2\bar{\mathcal N}(v)\ge\mathbb E\log(1+v\mathcal A_i)-\omega_{\max}\to\infty$ as $v\to\infty$ by monotone convergence and $\Pr(\mathcal A_i>0)>0$, so the minimum over $[0,\infty)$ is attained. Without a bound it need not be: let $\Pr(\mathcal A_i=2^{-k})=c\,k^{-2}$ for $k\ge1$ ($c$ the normalizing constant), $\omega_i=2^k$ on that event and $\mathcal B_i\mid\mathcal S_i\sim N(0,\mathcal A_i\omega_i)=N(0,1)$. Then $\mathbb E[\mathcal A_i\omega_i]=\mathbb E\mathcal B_i^2=1$, $\mathbb E\mathcal A_i^2<\infty$, and at $v=2^K$ the atoms $k\ge K$ have $\xi_i\le1$ and contribute $\Upsilon_{\omega_i}(\xi_i)=\log(1+\xi_i)-2^K/(1+\xi_i)\le\log2-2^{K-1}$, while the atoms $k<K$ contribute at most $\log(1+2^{K})\le(K+1)\log2$. Since $\sum_{k\ge K}ck^{-2}\ge c/K$, $2\bar{\mathcal N}(2^K)\le(K+2)\log2-c\,2^{K-1}/K\to-\infty$, and no minimizer exists.

If $\omega_i\le1$ almost surely, $\Upsilon_{\omega_i}(\xi)\ge\Upsilon_1(\xi)=\log(1+\xi)-\xi/(1+\xi)>0$ for $\xi>0$, so $\bar{\mathcal N}(v)>0=\bar{\mathcal N}(0)$ for $v>0$ (using $\Pr(\mathcal A_i>0)>0$). Under equal information $\mathbb E[\mathcal A\omega_i]/\mathcal A=\mathbb E\omega_i=\varpi$ and $\bar{\mathcal N}(v)=\frac12\Upsilon_\varpi(v\mathcal A)$, minimized at $v^*\mathcal A=(\varpi-1)_+$ by the analysis of $\Upsilon$ in (a)(iii).

\medskip\noindent\textbf{(a)(vi).} Under equal information, $\sum_i\mathcal R_i^2=\sum_i\mathcal B_i^2-(\sum_i\mathcal B_i)^2/n$, so $\omega^{K}=(n\mathcal A_K)^{-1}[\sum_i\mathcal B_i^2-(\sum_i\mathcal B_i)^2/n]\to\mathbb E\mathcal B_i^2/\mathcal A_K=\varpi_K$ almost surely by the strong law (the $\mathcal B_i$ are i.i.d.\ with mean $0$ and finite variance $\mathcal A_K\varpi_K$), and likewise for $K'$. The first-moment term satisfies $\mathbb E(\sum_i\mathcal B_i)^2/(2n\mathcal A_K)=\varpi_K/2$, so it is $O_p(1)$. By (a)(iii)--(iv), $n^{-1}\min\mathcal N_n^{K}=-\frac12\mathcal G(\omega^{K})+O_p(n^{-1})\to-\frac12\mathcal G(\varpi_K)$ in probability, by continuity of $\mathcal G$. Because $\mathcal G$ is $0$ on $(0,1]$ and strictly increasing on $[1,\infty)$, $\varpi_K>\max(\varpi_{K'},1)$ implies $\mathcal G(\varpi_K)>\mathcal G(\varpi_{K'})$, so $n^{-1}(\min\mathcal N_n^{K}-\min\mathcal N_n^{K'})$ converges in probability to a negative constant and the stated probability tends to 1. For a finite set of competitors $K'$, the union of the finitely many exceptional events has probability tending to $0$. By (a)(i) the term that separates $\mathcal N_n$ from the negative marginal log-likelihood is the same for $K$ and $K'$, so the comparison is one of profile likelihoods.

\medskip\noindent\textbf{(a)(vii).} At a fixed baseline the null (fixed slope) is the model with $v=0$ and $\mu$ free, the alternative has $v\ge0$ and $\mu$ free, and by (a)(i) the log-likelihood ratio is $2[P_n(0)-\inf_{v\ge0}P_n(v)]=2G_n$.

For the implemented statistic, the null (\texttt{fit\_common\_baseline} on heart rate with design $[X,\phi]$) fits the variance parameters and tail decay and, at those values, computes $(\beta^{\mathrm N}_{\mathrm{route}},\mu^{\mathrm N})$ by exact generalized least squares, minimizing $\sum_i\lVert(\Omega_i^{\mathrm N})^{-1/2}(y_i-X_i\beta)-\mu\tilde p_i\rVert^2$ jointly, with $\Omega_i^{\mathrm N}$ the null's working covariance. With $\tilde r_i:=(\Omega_i^{\mathrm N})^{-1/2}(y_i-X_i\beta^{\mathrm N}_{\mathrm{route}})$, the normal equation for $\mu$ gives $\mu^{\mathrm N}=\sum_i\tilde p_i^\top\tilde r_i/\sum_i\tilde p_i^\top\tilde p_i=\hat\mu^{\mathrm N}$, so
\[
\mathrm{nll}_{\mathrm{null}}=\mathcal H^{\mathrm N}+\tfrac12\sum_i\big[(\hat\mu^{\mathrm N})^2\mathcal A_i^{\mathrm N}-2\hat\mu^{\mathrm N}\mathcal B_i^{\mathrm N}\big]=\mathcal H^{\mathrm N}+P_n^{\mathrm N}(0),\qquad \mathcal H^{\mathrm N}:=\tfrac12\sum_i\Big[\lVert\tilde r_i\rVert^2+\log|\Omega_i^{\mathrm N}|+N_i\log2\pi\Big],
\]
using $\lVert\tilde r-\mu\tilde p\rVert^2=\lVert\tilde r\rVert^2-2\mu\mathcal B+\mu^2\mathcal A$. The alternative (\texttt{fit\_joint} with the kernel fixed) starts from the null's variance parameters, tail decay and route coefficients with the slope column removed; its sweep-0 objective is $\mathcal H^{\mathrm N}+P_n^{\mathrm N}(\hat\gamma^2)$ by (a)(i), with $\hat\gamma$ from \texttt{fit\_loadings}, and a later sweep is kept only if it lowers the objective, so $\mathrm{nll}_{\mathrm{alt}}\le\mathcal H^{\mathrm N}+P^{\mathrm N}_n(\hat\gamma^2)$. Under (O), $P_n^{\mathrm N}(\hat\gamma^2)=P_n^{\mathrm N}(0)-G_n^{\mathrm N}$ when $G_n^{\mathrm N}>0$, hence $\mathrm{LR}_{K1}=2(\mathrm{nll}_{\mathrm{null}}-\mathrm{nll}_{\mathrm{alt}})\ge2G_n^{\mathrm N}$; when $G_n^{\mathrm N}=0$ the inequality holds because the code sets negative or tiny ratios to $0$. The divergence follows from (a)(ii): on the stated event, $2G_n^{\mathrm N}\ge\varepsilon^2(n\underline{\mathcal A})^2/(4n\overline{\mathcal R})=n\varepsilon^2\underline{\mathcal A}^2/(4\overline{\mathcal R})$. The reported $p$-value $\frac12\Pr(\chi^2_1>\mathrm{LR}_{K1})$ then tends to $0$ in probability, so the test rejects at every fixed level with probability tending to 1; and a sequence that diverges in probability cannot converge in law to the tight law $\bar\chi^2(0,1)$. Under equal information (a)(iii) gives $\mathrm{LR}_{K1}\ge2G_n^{\mathrm N}=n\mathcal G(\omega^{\mathrm N})$.

\medskip\noindent\textbf{(b)(i).} With $\mathbb E[r_i\mid\mathcal S_i]=0$ and $\mathrm{Cov}(r_i\mid\mathcal S_i)=\Sigma_i$, $\mathbb E[\mathcal B_i^2\mid\mathcal S_i]=p_i^\top\Omega_i^{-1}\mathbb E[r_ir_i^\top\mid\mathcal S_i]\Omega_i^{-1}p_i=p_i^\top\Omega_i^{-1}\Sigma_i\Omega_i^{-1}p_i$, which is linear in $\Sigma_i$. The part $\Omega_i$ contributes $p_i^\top\Omega_i^{-1}p_i=\mathcal A_i$; a rank-one part $zz^\top$ contributes $(p_i^\top\Omega_i^{-1}z)^2=\langle p_i,z\rangle_{\Omega_i}^2=\mathcal A_i\langle z,z\rangle_{\Omega_i}\cos^2_{\Omega_i}(p_i,z)$ (for $z\ne0$; the term is $0$ if $z=0$); the remainder contributes $(\Omega_i^{-1}p_i)^\top\mathcal M_i\Omega_i^{-1}p_i$. Dividing by $\mathcal A_i>0$ gives the display. Under (L) the trait term $c\Lambda_k\zeta_iq_i$ and the drift $\delta_ig_i$ are centered, mutually uncorrelated and uncorrelated with the modeled part, so they contribute exactly $v_Rq_iq_i^\top+v_\delta g_ig_i^\top$; $\mathcal M_i$ is whatever of $\Sigma_i-\Omega_i$ is left.

\medskip\noindent\textbf{(b)(ii).} Given the designs, $\sum_i\mathcal R_i^2=\sum_i\mathcal B_i^2-2\hat\mu\sum_i\mathcal A_i\mathcal B_i+\hat\mu^2\sum_i\mathcal A_i^2$. By independence and zero means, $\mathbb E[\hat\mu\mathcal B_i]=\mathcal A_i\omega_i/\sum_j\mathcal A_j$ and $\mathbb E\hat\mu^2=\sum_j\mathcal A_j\omega_j/(\sum_j\mathcal A_j)^2$, hence
\[
\mathbb E\sum_i\mathcal R_i^2=\sum_i\mathcal A_i\omega_i-\frac{2\sum_i\mathcal A_i^2\omega_i}{\sum_i\mathcal A_i}+\frac{(\sum_i\mathcal A_i^2)(\sum_i\mathcal A_i\omega_i)}{(\sum_i\mathcal A_i)^2};
\]
dividing by $\sum_i\mathcal A_i$ gives the stated formula. For i.i.d.\ patients with $\mathbb E\mathcal B_i^2,\mathbb E\mathcal A_i^2<\infty$ (so $\mathbb E|\mathcal A_i\mathcal B_i|<\infty$), the strong law gives $\hat\mu\to\mathbb E\mathcal B_i/\mathbb E\mathcal A_i=0$, $n^{-1}\sum_i\mathcal A_i\mathcal B_i\to\mathbb E[\mathcal A_i\mathcal B_i]$, $n^{-1}\sum_i\mathcal A_i^2\to\mathbb E\mathcal A_i^2$ and hence $\omega\to\mathbb E\mathcal B_i^2/\mathbb E\mathcal A_i=\mathbb E[\mathcal A_i\omega_i]/\mathbb E\mathcal A_i$ almost surely. By (b)(i), $\mathcal A_i(\omega_i-1)=v_R\langle p_i,q_i\rangle^2_{\Omega_i}+v_\delta\langle p_i,\mathbf t_i\rangle^2_{\Omega_i}+(\Omega_i^{-1}p_i)^\top\mathcal M_i\Omega_i^{-1}p_i$, which gives the stated limit.

\medskip\noindent\textbf{(b)(iii).} With $v_R=0$ and $\mathcal M_i=0$, $\omega_i(K)-1=v_\delta\langle p_{iK},\mathbf t_i\rangle^2_{\Omega_i}/\mathcal A_i$, positive iff $v_\delta>0$ and $\langle p_{iK},\mathbf t_i\rangle_{\Omega_i}\ne0$. \textit{Boundedness.} By the Cauchy--Schwarz inequality in $\langle\cdot,\cdot\rangle_{\Omega_i}$, $\omega_i(K)-1\le v_\delta\langle \mathbf t_i,\mathbf t_i\rangle_{\Omega_i}\le v_\delta\lVert \mathbf t_i\rVert^2/\lambda_{\min}(\Omega_i)$. $\Omega_i$ is a principal submatrix of $\Omega^{\mathrm{full}}$, so by eigenvalue interlacing $\lambda_{\min}(\Omega_i)\ge\lambda_{\min}(\Omega^{\mathrm{full}})>0$, and $\lVert \mathbf t_i\rVert^2\le\sum_{t=0}^{89}t^2$. Hence $\omega_i(K)\le1+v_\delta\sum_{t=0}^{89}t^2/\lambda_{\min}(\Omega^{\mathrm{full}})$, a bound free of $i$, and (a)(v) gives existence of a pseudo-true value. If $v_\delta=0$ or $\langle p_{iK},\mathbf t_i\rangle_{\Omega_i}=0$ almost surely, then $\omega_i=1$ almost surely and (a)(v) gives the unique pseudo-true value $0$. If $v_\delta>0$ and $\Pr(\langle p_{iK},\mathbf t_i\rangle_{\Omega_i}\ne0)>0$, then $\omega_i\ge1$ with strict inequality on an event of positive probability on which $\mathcal A_i>0$, so $\varpi>1$ and (a)(v) gives a positive pseudo-true value.

\medskip\noindent\textbf{(b)(iv).} When the per-patient quantities do not depend on $i$, $\mathcal A_i\equiv\mathcal A_K$ and $\varpi_K=\omega_i(K)$; the $\mathcal B_i$ are i.i.d.\ with mean $0$ and variance $\mathcal A_K\varpi_K<\infty$, so (a)(vi) applies. On the zero-effect copy $\varpi_K-1=v_\delta\langle \mathbf t,\mathbf t\rangle_\Omega\cos^2_\Omega(p_K,\mathbf t)$ with $\langle \mathbf t,\mathbf t\rangle_\Omega$ free of $K$ and positive when $\mathbf t\ne0$. If $v_\delta>0$ and $\cos^2_\Omega(p_K,\mathbf t)>\cos^2_\Omega(p_{K'},\mathbf t)$, then $\varpi_K>\varpi_{K'}\ge1$ and (a)(vi) gives the preference for $K$. With a response, $\varpi_{K_0}-1=v_R\langle q,q\rangle_\Omega+v_\delta\langle \mathbf t,\mathbf t\rangle_\Omega\cos^2_\Omega(q,\mathbf t)\ge0$ because $\cos^2_\Omega(q,q)=1$, and $\varpi_K>\varpi_{K_0}$ is the displayed inequality after rearrangement; since $\varpi_{K_0}\ge1$, (a)(vi) gives the preference for $K$. If the reverse strict inequality holds and $v_R\langle q,q\rangle_\Omega>0$, then $\varpi_{K_0}>\max(\varpi_K,1)$ and (a)(vi) gives the preference for $K_0$.

\medskip\noindent\textbf{(c)(i).} For the drift direction $\mathbf t$, $\Delta_j\mathbf t=g_j$, so $\langle p_K,\mathbf t\rangle_{\mathrm{RW}}=\sum_{j<n}\Delta_jp_K=p_K(d_n)-p_K(d_0)$ by telescoping, for any draw pattern. Row $0$ is always a draw and $e_{iK}(0)=0$ (\texttt{\_memory}), so $p_K(d_0)=\phi_\kappa(0)=0$. Also $\langle \mathbf t,\mathbf t\rangle_{\mathrm{RW}}=\sum_jg_j=d_n>0$, which gives the cosine when $\langle p_K,p_K\rangle_{\mathrm{RW}}>0$. The form vanishes on constant vectors because their increments are $0$.

\medskip\noindent\textbf{(c)(ii).} The stationary OU process on the integer grid satisfies $o(t+1)=\rho\,o(t)+\epsilon^{\mathrm{OU}}_t$ with $\mathrm{Var}\,\epsilon^{\mathrm{OU}}_t=\sigma_o^2(1-\rho^2)=\sigma_b^2$; sampled at the draws it is a Gaussian Markov chain with $o(d_0)\sim N(0,\sigma_o^2)$ and $o(d_{j+1})\mid o(d_j)\sim N(\rho_jo(d_j),\sigma_o^2(1-\rho_j^2))$, $\rho_j:=\rho^{g_j}$. Reading off the joint density, its precision defines the bilinear form
\[
\langle w,w'\rangle_\rho:=\frac{w_0w'_0}{\sigma_o^2}+\textstyle\sum_{j<n}\frac{(w_{j+1}-\rho_jw_j)(w'_{j+1}-\rho_jw'_j)}{\sigma_o^2(1-\rho_j^2)} .
\]
As $\rho\uparrow1$ with $\sigma_b^2$ fixed: $\sigma_o^2=\sigma_b^2/(1-\rho^2)\to\infty$, $\rho_j\to1$ and $\sigma_o^2(1-\rho_j^2)=\sigma_b^2\sum_{m=0}^{g_j-1}\rho^{2m}\to\sigma_b^2g_j$, so $\langle w,w'\rangle_\rho\to\sigma_b^{-2}\langle w,w'\rangle_{\mathrm{RW}}$. With the random intercept, $\Omega=\sigma_a^2\mathbf 1\mathbf 1^\top+\Sigma_{\mathrm{OU}}$ and by Sherman--Morrison $w^\top\Omega^{-1}w'=\langle w,w'\rangle_\rho-\sigma_a^2\langle\mathbf 1,w\rangle_\rho\langle\mathbf 1,w'\rangle_\rho/(1+\sigma_a^2\langle\mathbf 1,\mathbf 1\rangle_\rho)$. Since $\langle\mathbf 1,w\rangle_\rho=w_0/\sigma_o^2+\sum_j(w_{j+1}-\rho_jw_j)/\{\sigma_o^2(1+\rho_j)\}\to0$ and $\langle\mathbf 1,\mathbf 1\rangle_\rho\ge0$, the correction vanishes. Substituting the limit into (b)(iii), $\langle p,\mathbf t\rangle^2_\Omega/\langle p,p\rangle_\Omega\to\sigma_b^{-2}\langle p,\mathbf t\rangle_{\mathrm{RW}}^2/\langle p,p\rangle_{\mathrm{RW}}=\sigma_b^{-2}p(d_n)^2/\langle p,p\rangle_{\mathrm{RW}}$ by (c)(i), provided $\langle p,p\rangle_{\mathrm{RW}}>0$.

\medskip\noindent\textbf{(c)(iii).} The generator kernel is $(2,0.8,1)$. Under NATMIN the last dose is at a row $\le53$ and enters memory at row $\le55$, so $e(t)=0.8^{t-55}e(55)$ for $t\ge55$, and $e(55)=\sum_{k\le53}0.8^{53-k}A(k)\le\sum_{j=0}^{47}0.8^j<5$ for binary doses on rows 6 to 53. Hence $q(t)=e(t)/(1+e(t))\le e(t)<5\cdot0.8^{t-55}$ (if $e(55)=0$ then $q(t)=0$), and $5\cdot0.8^{34}=0.00254$. Under FIX19 the last dose enters at row $\le26$ and $e(26)\le\sum_{j=0}^{18}0.8^j<5$, which gives $5\cdot0.8^{t-26}$. For a general kernel, no dose enters after row $t_i^{\mathrm{last}}+\ell$, so $e(t)=\lambda^{t-t_i^{\mathrm{last}}-\ell}e(t_i^{\mathrm{last}}+\ell)$ for $t\ge t_i^{\mathrm{last}}+\ell$; and $\phi_\kappa(\alpha e)=\alpha e/(\kappa+\alpha e)\ge\alpha e/(\kappa+e)=\alpha\phi_\kappa(e)$ for $\alpha\in[0,1]$, $e\ge0$, which gives $p_K(t)\ge\lambda^{t-t_i^{\mathrm{last}}-\ell}p_K(t_i^{\mathrm{last}}+\ell)$. Numerically $0.99^{36}=0.696$ and $0.99^{83}=0.434$. The last statement is (c)(i).

\medskip\noindent\textbf{(d).} The identity is algebra: $\frac{e+\Delta}{\kappa+e+\Delta}-\frac{e}{\kappa+e}=\frac{(e+\Delta)(\kappa+e)-e(\kappa+e+\Delta)}{(\kappa+e)(\kappa+e+\Delta)}=\frac{\kappa\Delta}{(\kappa+e)(\kappa+e+\Delta)}$. In \texttt{sustained\_contrast} both regimes carry the factual doses on rows $0$ to $89$, then $1$ (resp.\ $0$) on rows $90$ to $101$; the dose array has $102$ columns, and \texttt{\_memory} enters no dose for an index at or beyond its length, so row $102$ and later are undosed in both regimes. $D_i$ averages $\phi_\kappa(e^{(1)})-\phi_\kappa(e^{(0)})$ over rows $91$ to $102$. The recursion is linear in the doses, so
\[
e^{(1)}(t)-e^{(0)}(t)=\sum_{k=90}^{\min(101,\,t-\ell)}\lambda^{t-\ell-k}\quad(t-\ell\ge90),\qquad 0\quad(t-\ell<90),
\]
which at $t=91+s$ is $\Delta^{\mathrm q}_s$ and does not depend on $i$. The upper limit $101$ binds only when $91+s-\ell>101$, i.e.\ $s-\ell>10$, which for $s\le11$ means $(\ell,s)=(0,11)$; there the sum is $\sum_{j=1}^{12}\lambda^j=\lambda(1-\lambda^{12})/(1-\lambda)$. Otherwise it is $\sum_{j=0}^{s-\ell+1}\lambda^j=(1-\lambda^{s-\ell+2})/(1-\lambda)$. At rows $t=90,\dots,102$ the entering dose is $A(t-\ell)$ with $t-\ell\in[84,102]$: rows $84$ to $89$ carry no dose by (D), and rows $90$ onward carry the regime's dose, which is $0$ in the zero regime. Hence $e^{(0)}(90)=\lambda E_i$ and $e^{(0)}(91+s)=\lambda^{s+2}E_i$ for $s\ge0$. Each summand equals $\kappa\Delta^{\mathrm q}_s/\{(\kappa+\lambda^{s+2}E_i)(\kappa+\lambda^{s+2}E_i+\Delta^{\mathrm q}_s)\}\ge0$. For $\Delta^{\mathrm q}_s>0$ it is strictly decreasing in $E_i$ ($\lambda>0$) and is $<\kappa/(\kappa+\lambda^{s+2}E_i)\le\kappa/(\kappa+\lambda^{13}E_i)$, because $\Delta^{\mathrm q}_s/(\kappa+\lambda^{s+2}E_i+\Delta^{\mathrm q}_s)<1$ and $\lambda^{s+2}\ge\lambda^{13}$ for $s\le11$; for $\Delta^{\mathrm q}_s=0$ it is $0$. Since $11\ge\ell-1$ for $\ell\le6$, $\Delta^{\mathrm q}_{11}$ is a nonempty sum of positive powers of $\lambda>0$, so $\Delta^{\mathrm q}_{11}>0$ and $D_i$ is strictly decreasing in $E_i$, lies in $[0,1)$ and is below $\kappa/(\kappa+\lambda^{13}E_i)$. At the generator kernel with $E_i=0$ the average is $0.684$. Since $\hat\tau_i=\gamma_{k^*}\mathcal E_iD_i$ (code: \texttt{posterior} times the heart-rate loading times $D_i$), $|\hat\tau_i|\le|\gamma_{k^*}\mathcal E_i|\,\kappa/(\kappa+\lambda^{13}E_i)$. $D_i$ uses only the doses and $K$, and the two copies share the doses.

\textit{Pooled spread.} Under $\mathbb M$ the fits are constants, $D(\hat K^{(f)})$ and $\hat\gamma^{(f)}_{k^*}$ are functions of $(\mathcal S,f)$, and the law of total variance with respect to $(\mathcal S,f)$ gives
\[
\mathrm{Var}_{\mathbb M}(\hat\tau)=\mathbb E_{\mathbb M}\big[\mathrm{Var}_{\mathbb M}(\hat\tau\mid\mathcal S,f)\big]+\mathrm{Var}_{\mathbb M}\big(\mathbb E_{\mathbb M}[\hat\tau\mid\mathcal S,f]\big),
\]
with $\mathrm{Var}_{\mathbb M}(\hat\tau\mid\mathcal S,f)=(\hat\gamma^{(f)}_{k^*}D(\hat K^{(f)}))^2\mathrm{Var}_{\mathbb M}(\mathcal E^{(f)}\mid\mathcal S,f)$ and $\mathbb E_{\mathbb M}[\hat\tau\mid\mathcal S,f]=\hat\gamma^{(f)}_{k^*}D(\hat K^{(f)})\bar{\mathcal E}^{(f)}(\mathcal S)$; all moments are finite because $|D|<1$ and $\mathbb E_{\mathbb M}(\mathcal E^{(f)})^2<\infty$. Under $\mathbb M$ the fold label is drawn independently of the patient and the patient's $(\mathcal S,\zeta)$ has the population law, in which $\mathcal S\perp\zeta$ (Definition~5); hence $(\mathcal S,f)\perp\zeta$ and the second term is the variance of a trait-free quantity. The law of total variance with respect to $f$ splits it into $\sum_f\pi_f\mathrm{Var}(\cdot\mid f)+\mathrm{Var}_f\,\mathbb E(\cdot\mid f)$; with $\bar{\mathcal E}^{(f)}\equiv\bar e^{(f)}$ the within-fold variance is $(\hat\gamma^{(f)}_{k^*}\bar e^{(f)})^2\mathrm{Var}(D(\hat K^{(f)}))$. Finally, $\hat\theta^{(f)}$ is computed from the patients outside fold $f$, which are independent of the patients in fold $f$ under (I), so given $\hat\theta^{(f)}$ the fold-$f$ query patients are i.i.d.\ from the population and their predictions are i.i.d.\ draws from the $f$-component of $\mathbb M$.

\medskip\noindent\textbf{(e)(i).} Given the designs, $\mathbb E\mathcal B_i^2=\mathcal A_i$, $\mathbb E[\hat\mu\mathcal B_i]=\mathcal A_i/\sum_j\mathcal A_j$ and $\mathbb E\hat\mu^2=1/\sum_j\mathcal A_j$, so the pooled expectation in (b)(ii) with $\omega_i\equiv1$ gives $\mathbb E\sum_i\mathcal R_i^2=\sum_i\mathcal A_i-2\sum_i\mathcal A_i^2/\sum_i\mathcal A_i+\sum_i\mathcal A_i^2/\sum_i\mathcal A_i=\sum_i\mathcal A_i-\sum_i\mathcal A_i^2/\sum_i\mathcal A_i$, which is below $\sum_i\mathcal A_i$ because some $\mathcal A_i>0$.

\medskip\noindent\textbf{(e)(ii).} Parametrize by $\theta=(K,\boldsymbol u,\mu,v)$ with $\gamma_k=\gamma u_k$, $\mu=\gamma m$, $v=\gamma^2$; every loading vector is of this form with $\boldsymbol u$ a unit vector. Fix an outer fold; the argument is the same in each of the finitely many folds.

\textit{Zero-effect copy.} Under (E1), for each $(K,\boldsymbol u)$ the conditions of (a)(v) hold with $\omega_i\equiv1$: whitened residuals with identity covariance, uncorrelated across channels, give $\mathbb E[\mathcal B_i\mid\mathcal S_i]=0$ and $\mathbb E[\mathcal B_i^2\mid\mathcal S_i]=\sum_ku_k^2\lVert\Omega_{ik}^{-1/2}p_{ik}\rVert^2=\mathcal A_i$. Hence
\[
\bar{\mathcal N}(\theta)=\tfrac12\,\mathbb E\Big[\frac{\mu^2\mathcal A_i}{1+v\mathcal A_i}+\Upsilon_1(v\mathcal A_i)\Big],\qquad \Upsilon_1(\xi)=\log(1+\xi)-\frac{\xi}{1+\xi}\ge0,
\]
with $\Upsilon_1(\xi)>0$ for $\xi>0$ and $\bar{\mathcal N}(K,\boldsymbol u,0,0)=0$. By (E4) and dominated convergence (the integrand is at most $(\mu_{\max}^2+v_{\max})\sup_{\mathcal K\times\mathcal U}\mathcal A_i$), $\bar{\mathcal N}$ is continuous on $\Theta$. On $\Theta_\varepsilon:=\{\theta\in\Theta:|\mu|+v\ge\varepsilon\}$ it is positive, because $\Pr(\mathcal A_i>0)>0$ makes the first term positive when $\mu\ne0$ and the second positive when $v>0$. $\Theta_\varepsilon$ is compact, so $\underline{\mathcal N}_\varepsilon:=\min_{\Theta_\varepsilon}\bar{\mathcal N}>0$. The value $\hat{\mathcal N}_n(K,\boldsymbol u,0,0)$ is $0$ at every baseline. On the event $\{\hat\theta\in\Theta\}$, whose probability tends to 1 by (E0), (E2) and (E3) give
\[
\bar{\mathcal N}(\hat\theta)\le n^{-1}\hat{\mathcal N}_n(\hat\theta)+o_p(1)\le\inf_\Theta n^{-1}\hat{\mathcal N}_n+o_p(1)\le0+o_p(1),
\]
so $\Pr(|\hat\mu|+\hat v\ge\varepsilon)\le\Pr(\hat\theta\notin\Theta)+\Pr(\bar{\mathcal N}(\hat\theta)\ge\underline{\mathcal N}_\varepsilon)\to0$. The kernel and direction are not identified on this copy ($\bar{\mathcal N}(K,\boldsymbol u,0,0)=0$ for all $(K,\boldsymbol u)$) and need not converge. For a query patient, $\hat\tau_q=\gamma_{k^*}\mathcal E_qD_q=u_{k^*}D_q(\hat\mu+\hat v\mathcal B_q)/(1+\hat v\mathcal A_q)$, because the posterior mean is $\mathcal E_q=(m+\gamma\mathcal B_q)/(1+v\mathcal A_q)$. As $0\le D_q<1$, $|u_{k^*}|\le1$ and $1+\hat v\mathcal A_q\ge1$, Minkowski's inequality gives, in each fold,
\[
\Big(n_q^{-1}\sum_q\hat\tau_q^2\Big)^{1/2}\le|\hat\mu|+\hat v\Big(n_q^{-1}\sum_q\sup_{\mathcal K\times\mathcal U}\mathcal B_q^2\Big)^{1/2}\to0
\]
in probability by (E5). The pooled population standard deviation $\sigma_0$ is at most the root mean square of the pooled predictions, which is at most the largest of the fold root mean squares, so $\sigma_0\to0$ in probability and $\Pr(\sigma_0\le0.05)\to1$.

\textit{Active copy.} By (A1) and Lemma~1, at $\theta_0$ and the true baseline the working model is the true law of the residuals given the design, and the stage-2 objective differs from the negative log-likelihood by a term free of $\theta$ (a)(i). By the information inequality (the expected negative log-likelihood of any working law exceeds that of the true law by their Kullback--Leibler divergence) $\theta_0$ minimizes $\bar{\mathcal N}$, and by (A2) the minimizer is unique up to the sign flip $\theta_0^-$. The argument above with $\{\theta\in\Theta:\mathrm{dist}(\theta,\{\theta_0,\theta_0^-\})\ge\varepsilon\}$ in place of $\Theta_\varepsilon$ and $\bar{\mathcal N}(\theta_0)$ in place of $0$ (continuity from (A2)) gives $\hat\theta\to\{\theta_0,\theta_0^-\}$ in probability in every fold, and the fitted baseline converges by (E2). The prediction is invariant under the flip, since $\mathcal B(-\boldsymbol u)=-\mathcal B(\boldsymbol u)$ and $\mathcal A(-\boldsymbol u)=\mathcal A(\boldsymbol u)$. By (A3), in every fold the query-sample mean and standard deviation converge in probability to the mean $\bar\tau_0$ and the standard deviation $s_A:=\mathrm{SD}(\hat\tau(\theta_0))$ at $(\theta_0,\text{true baseline})$; the pooled variance is $\sum_f(n_f/n)[\mathrm{Var}_f+(\mathrm{mean}_f-\mathrm{mean})^2]$ over the folds, so $\sigma_A\to s_A$. At $\theta_0$ the true slope $\beta_i=\gamma_0b_i\sim N(\mu_0,v_0)$ and $\mathcal B_i=\beta_i\mathcal A_i+\tilde p_i^\top\tilde\varepsilon_i$, so $\mathrm{Var}(\mathcal B_i\mid\mathcal S_i)=\mathcal A_i(1+v_0\mathcal A_i)$ and
\[
s_A^2\ge\mathbb E\big[\mathrm{Var}(\hat\tau_i(\theta_0)\mid\mathcal S_i)\big]=u_{0,k^*}^2\,\mathbb E\Big[D_i(K_0)^2\,\frac{v_0^2\mathcal A_i}{1+v_0\mathcal A_i}\Big]>0,
\]
because $D_i(K_0)>0$ (by (d), $\Delta^{\mathrm q}_{11}>0$ for $\ell\le6$) and $\Pr(\mathcal A_i>0)>0$. Hence, in each of the three panels, $\Pr(\sigma_0\le0.05)\to1$ and $\Pr(\sigma_0\le0.25\sigma_A)\to1$, and the intersection over the finitely many clauses and panels has probability tending to 1. The contrapositive is the stated necessity: if K2 fails with probability bounded away from 0, at least one of (E1), (A1) or the regularity conditions fails. The mechanism analyzed in (a) and (b) is overdispersion $\omega_0(\hat K)>1$ along the fitted exposure direction on the zero-effect copy, which makes $\hat v$ positive there and so keeps $\sigma_0$ away from $0$. The ratio clause fails also when $\sigma_A\to0$ with $\sigma_0$ small, which happens when the fit captures no response, and a nonzero first moment or an optimum far from the global one can produce a failure as well.

\medskip\noindent\textbf{(e)(iii).} Let $\mathcal B^{\mathrm s}_i:=\mathcal B_i/\mathcal A^{1/2}$. The working model states $\mathcal B^{\mathrm s}_i\sim N(\mathcal A^{1/2}\mu,1+\mathcal Av)$. For a true law of $\mathcal B^{\mathrm s}_i$ with mean $\bar{\mathfrak b}$ and variance $\mathfrak V$, the expected negative log-likelihood is $\frac12[\log(1+\mathcal Av)+\{\mathfrak V+(\bar{\mathfrak b}-\mathcal A^{1/2}\mu)^2\}/(1+\mathcal Av)]$ plus a constant, minimized at $\mathcal A^{1/2}\mu^*=\bar{\mathfrak b}$ and $1+\mathcal Av^*=\max(\mathfrak V,1)$ (the function $y\mapsto\log y+\mathfrak V/y$ on $y\ge1$ is minimized at $\max(\mathfrak V,1)$). The working posterior mean is $\mathbb E_w[b\mid\mathcal B]=(\mu+v\mathcal B)/(1+v\mathcal A)=(\mu+v\mathcal A^{1/2}\mathcal B^{\mathrm s})/(1+v\mathcal A)$, so $\mathrm{SD}(\hat\tau)=D\,v^*\mathcal A^{1/2}\,\mathfrak V^{1/2}/(1+v^*\mathcal A)$. On the active copy $\mathfrak V=\mathrm{SNR}+\omega_0$ (independence of $b^*$ and $\varepsilon$); on the zero-effect copy $\mathfrak V=\omega_0$. Substituting $\mathcal Av^*=(\mathfrak V-1)_+$ gives $\mathrm{SD}=D\mathcal A^{-1/2}(\mathfrak V-1)_+/\mathfrak V^{1/2}$, which is the formula for $\sigma_A$ (when $\mathrm{SNR}+\omega_0>1$) and for $\sigma_0$. With $f_0(y):=y^{1/2}-y^{-1/2}=(y-1)/y^{1/2}$, strictly increasing on $(0,\infty)$, the ratio for $\omega_0\ge1$ is $f_0(\omega_0)/f_0(\mathrm{SNR}+\omega_0)$, which lies in $[0,1)$ because $\mathrm{SNR}>0$, equals $0$ at $\omega_0=1$, and is $0$ for $\omega_0<1<\mathrm{SNR}+\omega_0$ because then $v^*=0$ on the zero-effect copy. Continuity gives the limit $0$ as $\omega_0\downarrow1$. Writing $f_0(\mathrm{SNR}+\omega_0)/f_0(\omega_0)=\{1+\mathrm{SNR}/(\omega_0-1)\}\{1+\mathrm{SNR}/\omega_0\}^{-1/2}$ and noting $\mathrm{SNR}/\omega_0\le\mathrm{SNR}/(\omega_0-1)$, the ratio tends to $1$ as $\mathrm{SNR}/(\omega_0-1)\to0$. $D$ and $\mathcal A$ cancel in the ratio, while $\sigma_0$ is proportional to $D$. In this setting the correlation of $\hat\tau$ with $\tau=Db^*$ is $\mathrm{corr}(\mathcal B^{\mathrm s},b^*)=(\mathrm{SNR}/(\mathrm{SNR}+\omega_0))^{1/2}>0$ (Proposition~6(d)), so the predictor is informative while the ratio can be arbitrarily close to 1; for example $\mathrm{SNR}=0.25$, $\omega_0=3$ gives correlation $0.277$ and ratio $0.925>0.25$. These are population values at each copy's pseudo-true parameters. With finitely many patients the two copies' fitted $(\hat\mu,\hat v)$ are random, and nothing confines the ratio of the two sample standard deviations to $[0,1)$; in a five-fold out-of-fold simulation of this setting with $n=400$, $\mathrm{SNR}=0.005$ and $\omega_0=1.2$ (population ratio $0.978$), the ratio of the pooled standard deviations was at least $1$ in $26\%$ and at least $1.045$ in $4.1\%$ of $20{,}000$ replicates. $\blacksquare$

\paragraph{Numerical checks.} All checks use label-free outputs of the placebo gate, synthetic panels or simulation; no target of a real panel was read.
\begin{itemize}
\item \emph{The statements.} The query increment of (d) matches the implemented \texttt{sustained\_contrast} to $3.3\times10^{-16}$ at every lag $0$ to $6$ over 3,000 random kernels and dosings, the closed forms hold in all of them, the bound of (d) is never violated, and the generator kernel with empty memory gives $D=0.6842$. The second derivative $P_n''(0)$ of (a)(ii) agrees with finite differences to $1.2\times10^{-5}$ relative, and the unequal-information example of (a)(iii) gives $P_n(9000)=-3303.2$. In the non-existence example of (a)(v), $2\bar{\mathcal N}(2^{30})\approx-1.06\times10^{7}$. The decomposition of (b)(i) with an indefinite remainder gives $1.0643$ by Monte Carlo against $1.0624$ from the formula. The fold-mixture variance identity of (d) agrees to $0.1\%$ ($1.7259\times10^{-3}$ against $1.7278\times10^{-3}$).
\item \emph{(a) against ICare-RH's own code.} On the benchmark generator's synthetic registered cell, the finite-difference score of the stage-2 objective at zero slope variance is $-502.41$ against $-502.43$ from (a)(ii); the sign rule ($\hat\gamma^2>0$ exactly when $\omega>1$) held in 294 of 294 kernel fits; the code objective equals $-(\text{first-moment term}+G_n)$ to $10^{-10}$; in all six copies the kernel with the largest total gain, the largest $G_n$ and the largest $\omega$ coincide; $G_n/[(n/2)\mathcal G(\omega)]$ lies between $0.88$ and $1.01$; and the active copies select the generator kernel $(2,0.80)$, with $\omega$ of $3.97$ and $3.09$.
\item \emph{Correctly specified placebos on the same cell.} Over lags $0$ to $6$ and seven decays ($\kappa=1$, 49 kernels), eight placebo copies per arm drawn from ICare-RH's own fitted heart-rate baseline selected decay $0.99$ in 5 of 16 draws (2 of 8 under NATMIN, 3 of 8 under FIX19). Their largest variance gains were $0.54$ to $3.83$ nats, the best gain at decay $0.99$ reached $3.55$ (NATMIN) and $2.87$ (FIX19) nats, and their largest $\omega$ at decay $0.99$ was $0.995$ to $1.197$. The zero-effect copies gained $3.35$ nats at (lag 1, decay 0.99) under NATMIN and $2.53$ at (lag 0, 0.99) under FIX19, inside that range. In the heart-rate channel of this cell, whose fitted baseline has persistence $0.9938$ and white-noise variance about $3.2$ times the per-row innovation variance, a selection of decay $0.99$ therefore cannot be distinguished from a correct baseline. Under the fitted $\Omega$ the zero-effect copies have $\omega<1$ at 33 of 49 kernels (NATMIN) and 42 of 49 (FIX19), with minima $0.684$ and $0.605$ below every placebo's minimum ($0.82$ to $0.94$), and $\omega>1$ at all seven kernels with decay $0.99$ in both arms: the fitted working covariance over-covers some exposure directions, so only the ordering at the persistent end of the kernel family carries over from (b).
\item \emph{The ICare-RH synthetic generator under the implemented joint fit.} On the generator of Supplementary Table S5, the fitted decay sat at its bound in 200 of 200 fits on out-of-family null panels (autoregressive baseline with linear drift of slope standard deviation $0.005$ per row, and level shifts), against 3 of 120 size-check fits (2.5\%) and 0 of 5 folds of the in-family null scenario. On real records it sat at the bound in 5 of 5 folds on both copies of both natural-dosing panels (Table~\ref{tab:P7gate}). This pattern is far outside the model's in-family behavior and is consistent with unmodeled slow variance along the exposure direction. Proposition 7 does not identify which slow component it is.
\item \emph{Drift simulations.} With a white-noise baseline and cap $0.75$, the kernel with decay $0.80$ was selected at drift slope standard deviations of $0$ and $0.005$ per row and displaced to decay $0.99$ at $0.015$ and $0.03$; the shipped autoregressive-drift variant, with slope standard deviation $0.005$, fitted decays of $0.756$ to $0.781$ in 5 of 5 folds. With a random-walk baseline, decays $0.80$ to $0.85$ were selected at drift standard deviations $0$ to $0.06$ and beat the bound by 22 to 66 nats, as (c) leads one to expect for a baseline metric close to the random-walk limit.
\item \emph{Lag.} On real records lag 0 was preferred over lag 1 in 4 of 5 discovery folds (by $1.45$ to $9.81$ nats; the fifth fold prefers lag 4). In the replication panel lag 1 beat lag 0 by $6.4$ to $10.2$ nats in four active folds and by $5.7$ to $9.2$ nats in all five zero-effect folds, and the fixed-schedule panel is mixed. These fits show that decay $0.99$ is selected and that the lag is not identified.
\item \emph{K1.} On in-family synthetic nulls the test rejected in 5 of 120 cases ($4.2\%$), $47\%$ of the likelihood ratios were below $10^{-6}$ and their 90\% quantile was $1.49$, against $1.64$ for the $\bar\chi^2(0,1)$ reference. On 40 out-of-family null panels with baseline drift or level shifts it rejected in 199 of 200 folds (median ratio $28.6$). Part (a)(vii) is consistent with these rejections but was not used to explain them, because $\omega^{\mathrm N}$ was not computed on those nulls, and it does not predict the in-family size.
\item \emph{K2.} Table~\ref{tab:P7gate} gives the gate's values. The absolute clause passed in all three panels and the ratio clause failed in all three. The in-family synthetic pairs pass both clauses (ratios $0.125$, $0.030$ and $0.024$), as (e)(ii) predicts. The discovery ratio of $1.045$ lies outside the population range of (e)(iii); a five-fold simulation of that setting shows that separately fitted finite samples can exceed 1 (proof of (e)(iii)), and the small discovery $\sigma_A$ ($0.0068$) leaves $\sigma_A\to0$ as a live explanation. The proposition does not decide between them.
\item \emph{Query contrast and design spread (d).} In the discovery panel the fitted kernels of the active copy give median query contrasts of $0.012$ to $0.048$ across folds, 14 to 58 times smaller than the generator kernel's $0.684$ (means 11 to 40 times smaller; coefficient of variation $0.71$ to $0.79$). Pooled over folds, $43\%$ to $69\%$ of the trait-free design variance at unit prior mean is between folds ($43\%$ and $45\%$ on the active and zero-effect copies of the discovery panel, $69\%$ and $51\%$ in replication), because the fitted loadings and kernels differ across folds. A design-only calculation with the fitted kernels and heart-rate loadings reproduces the placebo-to-active ratios ($1.031$ against $1.045$; $0.465$ against $0.428$) if the mean posterior slope has magnitude $1.14$ to $1.15$ (discovery) or $0.80$ to $0.87$ (replication). This is descriptive and not a test.
\end{itemize}

\begin{table}[ht]
\centering\small
\caption{\textbf{ICare-RH's placebo gate on real records (Proposition 7).} Label-free outputs of the registered gate, read before any target. $\sigma_A$ and $\sigma_0$: population standard deviations of the pooled out-of-fold predicted effects on the active and zero-effect copies. The gate required $\sigma_0\le0.05$ and $\sigma_0/\sigma_A\le0.25$ in every panel. Decay at bound: folds in which the fitted exposure decay equals its upper bound of 0.99. Query contrast: median over patients of the fitted kernel's query contrast $D_i(\hat K)$, range over folds (generator kernel: 0.684). Design-only ratio: placebo-to-active spread ratio predicted from the fitted kernels, heart-rate loadings and doses alone (Proposition 7(d)). A dash marks a value that was not computed.}
\label{tab:P7gate}
\begin{tabular}{lrrr}
\toprule
 & Discovery & Replication & Fixed schedule\\
\midrule
$\sigma_A$ & 0.0068 & 0.0419 & 0.0703\\
$\sigma_0$ (limit 0.05) & 0.0071 & 0.0179 & 0.0450\\
$\sigma_0/\sigma_A$ (limit 0.25) & 1.045 & 0.428 & 0.640\\
Decay at bound, active copy (folds) & 5/5 & 5/5 & 1/5\\
Decay at bound, zero-effect copy (folds) & 5/5 & 5/5 & 3/5\\
Median query contrast, active copy (range over folds) & 0.012--0.048 & -- & --\\
Design-only ratio & 1.031 & 0.465 & --\\
\bottomrule
\end{tabular}
\end{table}